\documentclass{article}

\usepackage{arxiv}

\usepackage[utf8]{inputenc} \usepackage[T1]{fontenc}    \usepackage[colorlinks]{hyperref}       \usepackage{url}            \usepackage{booktabs}       \usepackage{amsmath}
\usepackage{amsfonts}       \usepackage{nicefrac}       \usepackage{microtype}      \usepackage{xspace}
\usepackage{graphicx}
\usepackage{subcaption}
\usepackage{multirow}
\usepackage{wrapfig}
\usepackage[font=small,labelfont=bf,tableposition=top]{caption}
\DeclareCaptionLabelFormat{andtable}{\tablename~\thetable}

\newcommand{\bc}{\mathbf{c}}\newcommand{\bC}{\mathbf{C}}

\newcommand{\bF}{\mathbf{F}} 

\newcommand{\bI}{\mathbf{I}}

\newcommand{\bq}{\mathbf{q}}
\newcommand{\br}{\mathbf{r}}
\newcommand{\bs}{\mathbf{s}}
\newcommand{\bt}{\mathbf{t}}

\newcommand{\cO}{\mathcal{O}}

\newcommand{\cX}{\mathcal{X}}

\usepackage{amsfonts}

\makeatletter
\DeclareRobustCommand\onedot{\futurelet\@let@token\@onedot}
\def\@onedot{\ifx\@let@token.\else.\null\fi\xspace}
\def\eg{e.g\onedot} 
\def\ie{i.e\onedot} 
 
\def\etc{etc\onedot}

\def\wrt{wrt\onedot}

\def\etal{et~al\onedot}

\makeatother

\newcommand{\boldparagraph}[1]{\vspace{0.2cm}\noindent{\bf #1:} }

\newcommand{\figref}[1]{Fig.~\ref{#1}}
\newcommand{\tabref}[1]{Tab.~\ref{#1}}
\newcommand{\secref}[1]{Sec.~\ref{#1}}

\usepackage{xcolor}
\definecolor{darkred}{rgb}{0.7,0.2,0.1}
\definecolor{darkgreen}{rgb}{0,0.7,0}
\definecolor{orange}{RGB}{255,127,0}
\definecolor{ourpurple}{RGB}{127,127,204}
\definecolor{palgreen}{RGB}{51,179,179}
\definecolor{magenta}{RGB}{199,21,133}

\title{ATISS: Autoregressive Transformers for Indoor Scene Synthesis}

\author{%
  Despoina Paschalidou$^{1,3,4}$ \quad Amlan Kar$^{4,5,6}$ \quad Maria Shugrina$^{4}$ \quad Karsten Kreis$^{4}$ \\[-1em]
  \And Andreas Geiger$^{1,2,3}$ \quad Sanja Fidler$^{4,5,6}$\\[1em]
  $^1$Max Planck Institute for Intelligent Systems T{\"u}bingen\quad
  $^2$University of T{\"u}bingen\\
  $^3$Max Planck ETH Center for Learning Systems\\
  $^4$NVIDIA\quad
  $^5$University of Toronto\quad
  $^6$Vector Institute \\
  {\tt\small \{firstname.lastname\}@tue.mpg.de\quad \{amlank, mshugrina, kkreis, sfidler\}@nvidia.com}
                                            }

\begin{document}

\maketitle

\begin{abstract}
The ability to synthesize realistic and diverse indoor furniture layouts
automatically or based on partial input, unlocks many applications, from better
interactive 3D tools to data synthesis for training and simulation. In this
paper, we present ATISS, a novel autoregressive transformer architecture for
creating diverse and plausible synthetic indoor environments, given only the
room type and its floor plan. In contrast to prior work, which poses scene
synthesis as sequence generation, our model generates rooms as unordered sets
of objects. We argue that this formulation is more natural, as it makes ATISS
generally useful beyond fully automatic room layout synthesis. For example, the
same trained model can be used in interactive applications for general scene
completion, partial room re-arrangement with any objects specified by the user,
as well as object suggestions for any partial room. To enable this, our model
leverages the permutation equivariance of the transformer when conditioning on
the partial scene, and is trained to be permutation-invariant across object
orderings. Our model is trained end-to-end as an autoregressive generative
model using only labeled $3$D bounding boxes as supervision. Evaluations on
four room types in the 3D-FRONT dataset demonstrate that our model consistently
generates plausible room layouts that are more realistic than existing methods.
In addition, it has fewer parameters, is simpler to implement and train and
runs up to $8$x faster than existing methods.

\end{abstract}

\vspace{-4mm}
\section{Introduction}
\label{sec:intro}
\vspace{-1mm}

Generating synthetic 3D content that is both realistic and diverse is a
long-standing problem in computer vision and graphics.  In the last decade,
there has been increased demand for tools that automate the creation of 3D
artificial environments for applications like video games and AR/VR, as well as
general 3D content creation \cite{Yu2011SIGGRAPH, Fisher2011SIGGRAPH,
Merrell2011SIGGRAPH, Chaudhuri2013UIST, Yu20216VisComputGraph}. These tools can
also synthesize data to train computer vision models, avoiding expensive and
laborious annotations. Generative models \cite{Kingma2014ICLR,
Goodfellow2014NIPS, Dinh2017ICLR, Kingma2018NIPS, Vaswani2017NIPS} have
demonstrated impressive results on synthesizing photorealistic images
\cite{Choi2018CVPR, Brock2019ICLR, Karras2019CVPR, Choi2020CVPR,
Karras2020CVPR} and intelligible text \cite{Radford2019ARXIV,
Brown2020NeurIPS}, and are beginning to be adopted for the generation of 3D
environments.

Recent works proposed to solve the scene synthesis task by incorporating
procedural modeling techniques \cite{Qi2018CVPRa,
Prakash2019ICRA, Kar2019ICCV, Devaranjan2020ECCV} or by generating scene graphs with generative models
\cite{Li2019SIGGRAPH, Wang2019SIGGRAPH, Zhou2019ICCV, Luo2020CVPR,
Purkait2020ECCV, Zhang2020SIGGRAPH, Zhang2020ARXIVa, Keshavarzi2020ARXIV,
Di2020ARXIV}. Procedural modeling requires specifying a set of rules for
the scene formation process, but acquiring these rules is a time-consuming
task, requiring skills of experienced artists. Similarly, graph-based approaches require scene graph annotations, which may be laborious to obtain.

\begin{figure*}
\vspace{-2mm}
\centering
\includegraphics[width=1.0\textwidth]{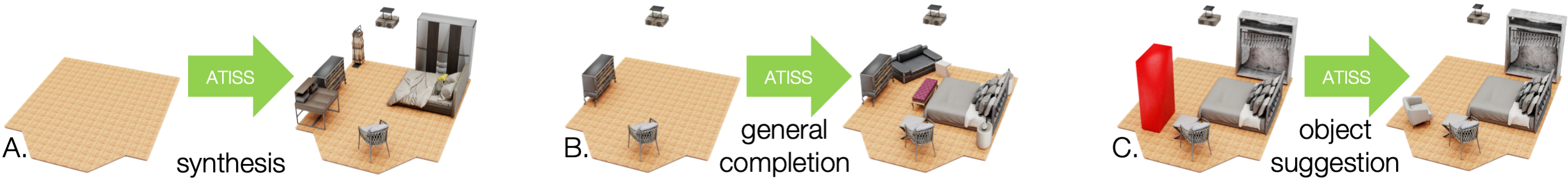}
\vspace{-3.6mm}
\caption{{\bf{Motivation}} In addition to fully automatic layout synthesis (A),
our formulation in terms of unordered sets of objects allows our model to be
used for novel interactive applications with versatile user control: scene
completion given any number of existing furniture pieces of any class pinned to
a specific location by the user (B), and object suggestions with user-provided
constraints (object centroid constraint shown in red) (C).}  
\label{fig:teaser}
\vspace{-1.2em}
\end{figure*}

Another line of research utilizes CNN-based~\cite{Wang2018SIGGRAPH, Ritchie2019CVPR} and transformer-based~\cite{Wang2020ARXIV} architectures to generate rooms by autoregressively
selecting and placing objects in a scene, \ie one after
the other.
These approaches represent scenes as ordered
sequences of objects. Typically, the ordering is defined using the spatial
arrangement of objects in a room (\eg left-to-right)
\cite{Jyothi2019ICCV} or the object class frequency (\eg most to least
probable) \cite{Ritchie2019CVPR, Wang2020ARXIV}. Such orderings impose unnatural constraints on the scene generation process, inhibiting practical applications. For example, in~\cite{Ritchie2019CVPR, Wang2020ARXIV}, which order objects by class frequency, the
probability of a bed (more common) appearing after an ottoman (less common) in the training set is zero. As a result, these
methods cannot generate more common objects after less common
objects, which makes them impractical for
interactive tasks like general room completion and partial room re-arrangement, where input is unconstrained (\eg Fig.\ref{fig:teaser}B).

To address these limitations, we pose scene synthesis as an unordered set
generation problem and introduce ATISS, a novel autoregressive transformer architecture to
model this process. Given a room type (\eg bedroom, living room) and its shape, our
model generates meaningful furniture arrangements by sequentially
placing objects in a permutation-invariant fashion. We train ATISS
to maximize the log-likelihood of all possible permutations of object
arrangements in a collection of training scenes, labeled only with object
classes and $3D$ bounding boxes, which are easier to obtain, than costly
support relationship \cite{Wang2019SIGGRAPH} or scene graph annotations \cite{Li2019SIGGRAPH}. Unlike
existing works~\cite{Wang2018SIGGRAPH, Ritchie2019CVPR, Wang2020ARXIV},
we propose the first model to perform scene synthesis as an autoregressive \emph{set generation} task.
ATISS is significantly simpler to implement and train, requires
fewer parameters and is up to $8\times$ faster at run-time than the fastest available baseline
\cite{Wang2020ARXIV}. Furthermore, we
demonstrate that our model generates more plausible object arrangments
without any post-processing on the
predicted layout. Our  formulation allows applying a single trained model to
automatic layout synthesis and to a number of interactive scenarios with
versatile user input (Fig.\ref{fig:teaser}), such as automatic placement of
user-provided objects, object suggestion with user-provided constraints, and room completion.
Code and data are publicaly available at \url{https://nv-tlabs.github.io/ATISS}.

\vspace{-3mm}
\section{Related Work}
\label{sec:related}
\vspace{-1mm}

In this section, we discuss the
most relevant literature on interior scene synthesis, as well as transformer
architectures \cite{Vaswani2017NIPS} in the context of generative modeling.

\vspace{-2mm}
\boldparagraph{Procedural Modeling with Grammars}%
Procedural modeling describes methods that recursively apply a set of functions for content synthesis. Grammars are a formal instantiation of this idea and have been used for modeling 3D structures such as plants \cite{Talton2011SIGGRAPH}, buildings and cities \cite{Muller2006SIGGRAPH, Parish2001SIGGRAPH}, indoor~\cite{Qi2018CVPRa} and outdoor~\cite{Prakash2019ICRA} scenes. \cite{Talton2011SIGGRAPH} employed reversible-jump MCMC to control the output of stochastic context-free grammars. Meta-Sim~\cite{Kar2019ICCV} learned a model that modifies attributes of scene graphs sampled from a known probabilistic context-free grammar to match visual statistics between generated and real data. \cite{Devaranjan2020ECCV} extended this model to also learn to sample from the grammar, allowing context dependent relationships to be learnt. Concurrently, \cite{Purkait2020ECCV} employed Grammar-VAE~\cite{kusner2017grammar} to generate scenes using a scene grammar generated from annotated data.
In contrast, our model implicitly encapsulates inter-object relationships, without having to impose hand-crafted constraints. 
\vspace{-2mm}
\boldparagraph{Graph-based Scene Synthesis}%
Representing scenes as graphs has been extensively studied in literature
\cite{Li2019SIGGRAPH, Wang2019SIGGRAPH, Zhou2019ICCV, Luo2020CVPR,
Purkait2020ECCV, Zhang2020SIGGRAPH, Zhang2020ARXIVa, Keshavarzi2020ARXIV,
Di2020ARXIV}.
Zhou \etal
\cite{Zhou2019ICCV} introduced a neural message passing algorithm for scene
graphs that predicts the category of the next object to be placed at a
specific location. Similarly, \cite{Li2019SIGGRAPH, Zhang2020SIGGRAPH, Purkait2020ECCV,
Luo2020CVPR} utilized a VAE~\cite{Kingma2014ICLR} to synthesize 3D scenes as
parse trees~\cite{Purkait2020ECCV}, adjacency matrices~\cite{Zhang2020SIGGRAPH}, 
scene graphs~\cite{Luo2020CVPR} and scene hierarchies~\cite{Li2019SIGGRAPH}.
Concurrently, \cite{Wang2019SIGGRAPH, Zhang2020ARXIVa} adopted a
two-stage generation process that disentangles planning the
scene layout from instantiating the scene based on this plan. Note
that graph-based models require supervision either in the form of relation
graphs \cite{Wang2019SIGGRAPH, Zhang2020ARXIVa, Luo2020CVPR} or scene
hierarchies \cite{Li2019SIGGRAPH}. In contrast, ATISS infers functional and
spatial relations between objects directly from data labeled only with object
classes and 3D bounding boxes.

\vspace{-2mm}
\boldparagraph{Autoregressive Scene Synthesis}%
Closely related to our work are autoregressive indoor scene generation
models~\cite{Wang2018SIGGRAPH, Ritchie2019CVPR, Wang2020ARXIV}. Ritchie
\etal~\cite{Ritchie2019CVPR} introduced a CNN-based architecture that operates
on a top-down image-based representation of a scene and inserts objects in it
sequentially by predicting their category, location, orientation and size with
separate network modules. \cite{Ritchie2019CVPR} requires supervision in the
form of 2D bounding boxes as well as auxiliary supervision such as depth maps
and object segmentation masks.  In concurrent work, Wang
\etal~\cite{Wang2020ARXIV} introduced SceneFormer, a series of transformers
that autoregressively add objects in a scene similar to
\cite{Ritchie2019CVPR}. Both \cite{Ritchie2019CVPR, Wang2020ARXIV}
use separate models to generate object attributes (\eg category, location) that are trained independently and represent
scenes as ordered sequences of objects, ordered by the category frequency.  In
contrast, we propose a simpler architecture that consists of a single model
trained end-to-end to predict all attributes. We provide experimental
evidence that our model generates more realistic object arrangements while being significantly faster. While~\cite{Ritchie2019CVPR, Wang2020ARXIV} assume a fixed ordering
of the objects in each scene, our model does not impose any
constraint on the ordering of objects. Instead, during training,
we enforce that our model generates objects with all orderings, in a
permutation invariant fashion.  This allows us to represent scenes as unordered
sets of objects and perform various interactive tasks such as rearranging any
object in a room or suggesting new objects given any room.

\vspace{-2mm}
\boldparagraph{Transformers for Set Generation}%
Transformer models \cite{Vaswani2017NIPS} demonstrated impressive results on
various tasks such as machine translation \cite{Shaw2018NAACL, Ott2018WMT},
language-modeling \cite{Brown2020NeurIPS, Devlin2019NAACL}, object detection
\cite{Li2020IROS, Carion2020ECCV, Zhu2021ICLR}, image recognition
\cite{Dosovitskiy2021ICLR, Touvron2020ARIV}, semantic segmentation
\cite{Ye2019CVPR} as well as on image \cite{Parmar2018ICML,
Katharopoulos2020ICML, Chen2020ICML, Esser2021CVPR, Tulsiani2021ICML} and
music \cite{Dhariwal2020ARXIV} generation tasks. While there are works
\cite{Lee2019ICML, Kosiorek2020ARXIV} that utilize the permutation equivariance property of
transformers for unordered set processing and prediction, existing generative
models with transformers assume ordered sequences \cite{Brown2020NeurIPS,
Chen2020ICML, Child2019ARXIV} even when there exists no natural order such as
for pointclouds \cite{Nash2020ICML} and objects in a scene
\cite{Wang2020ARXIV}. Instead, we introduce an autoregressive transformer for
unordered set generation that enforces that the probability of adding a new element
in the set is invariant to the order of the elements already in the set. We
show that for the scene synthesis task, our model outperforms
transformers that consider ordered sets of elements in
every metric.

\section{Method}
\label{sec:method}

\begin{figure*}
\vspace{-2mm}
    \centering
    \includegraphics[width=0.95\textwidth]{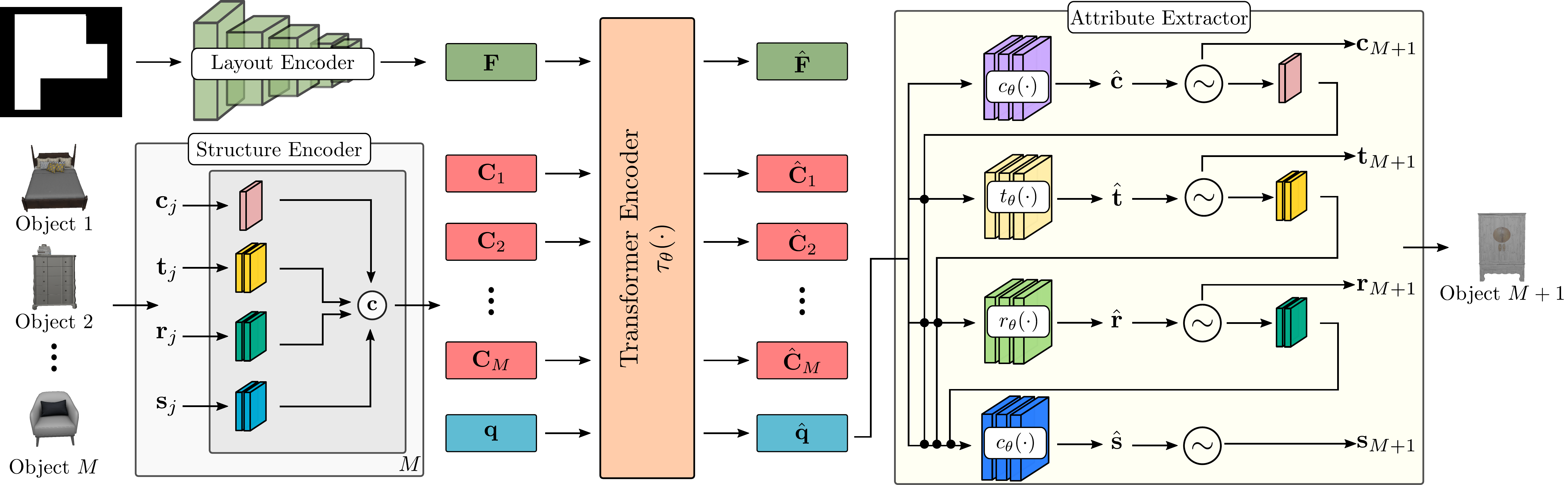}
    \caption{\small 
        {\bf{Method Overview.}} Starting from a scene with
        $M$ objects and a floor layout, the
        \textbf{layout encoder} maps the floor into a feature representation $\bF$ and
        the \textbf{structure encoder} maps the objects into a
        context embedding $\bC=\{\bC_j\}_{j=1}^M$. The floor
        layout feature $\bF$, the context embedding $\bC$ and a learnable
        query vector $\bq$ are then passed to the \textbf{transformer
        encoder} that predicts $\hat{\bq}$. Using $\hat{\bq}$ the \textbf{attribute
        extractor} autoregressively predicts the attribute distributions that are used
        to sample the attributes for the next object to be generated.
        }
    \label{fig:method_overview}
    \vspace{-1.2em}
\end{figure*}

Given an empty or a partially complete room of a specific type (\eg bedroom)
together with its shape, as a top-down orthographic projection of its
floor, we want to learn a generative model that populates the room with objects,
whose functional composition and spatial arrangement is plausible. To this end,
we propose an autoregressive model that represents scenes as \emph{unordered
sets of objects} (\secref{subsec:set_gen}) and describe our implementation using a
transformer network (\secref{subsec:network_architecture}). Finally, we analyse
the training and inference details of our method (\secref{subsec:training_inference}).

\vspace{-2mm}
\subsection{Autoregressive Set Generation}\label{subsec:set_gen}
Let $\cX = \{\cX_1, \dots, \cX_N\}$ denote a collection of scenes where each
$\cX_i=\left(\cO_i, \bF^i\right)$ comprises the unordered set of objects in the
scene $\cO_i=\{o_j^i\}_{j=1}^M$ and its floor layout $\bF^i$. To compute the
likelihood of generating $\cO_i$ we need to accumulate the likelihood of
generating $\{o_j^i\}_{j=1}^M$ autoregressively in \emph{any order}. This is
formally written as
\begin{equation}
    p_{\theta}(\cO_i | \bF^i) = 
        \sum_{\hat{\cO}\in \pi(\cO_i)}
        \prod_{j\in \hat{\cO}}
            p_{\theta}(o_j^i \mid o_{<j}^i, \bF^i),
    \label{eq:scene_probability}
\end{equation}
where $p_{\theta}(o_j^i \mid o_{<j}^i, \bF^i)$ is the probability of the $j$-th
object, conditioned on the previously generated objects and the floor layout,
and $\pi(\cdot)$ is a permutation function that computes the set of
permutations of all objects in the scene. As a result, the log-likelihood of
the whole collection $\cX$ is
\begin{equation}
    \log p_{\theta}(\cX) = 
    \sum_{i=1}^N\log\left(\sum_{\hat{\cO}\in \pi(\cO_i)}\prod_{j\in \hat{\cO}}
    p_{\theta}(o_j^i \mid o_{<j}^i, \bF^i)\right).
    \label{eq:data_likelihood}
\end{equation}
However, training our generative model to maximize the log-likelihood of
\eqref{eq:data_likelihood} poses two problems: (a) the summation over all
permutations is intractable and (b) \eqref{eq:data_likelihood} does not ensure
that all orderings will have high probability. The second problem is crucial
because we want our generative model to be able to complete \emph{any
partial set} in a plausible way, namely we want any generation order to have
high probability. To this end, instead of maximizing
\eqref{eq:data_likelihood}, we maximize the likelihood of generating a scene in
all possible orderings, $\hat{p}_\theta(\cdot)$, which is defined as
\begin{equation}
    \log \hat{p}_{\theta}(\cX)
    = \sum_{i=1}^N\log\left(\prod_{\hat{\cO}\in \pi(\cO_i)}
       \prod_{j\in \hat{\cO}}p_{\theta}(o_j^i \mid o_{<j}^i, \bF^i)\right)
    = \sum_{i=1}^N\sum_{\hat{\cO}\in \pi(\cO_i)}
       \sum_{j\in \hat{\cO}}\log p_{\theta}(o_j^i \mid o_{<j}^i, \bF^i).
    \label{eq:data_likelihood_ours}
\end{equation}
Note that training our generative model with \eqref{eq:data_likelihood_ours}
allows us to approximate the summation over all permutations using Monte Carlo
sampling thus solving both problems of \eqref{eq:data_likelihood}.

\boldparagraph{Modelling Object Attributes}%
We represent objects in a scene as labeled 3D bounding boxes and model them
with four random variables that describe their category, size, orientation and
location, $o_j = \{\bc_j, \bs_j, \bt_j, \br_j\}$. The category
$\bc_j$ is modeled using a categorical variable over the total number of
object categories $C$ in the dataset. For the size $\bs_j \in \mathbb{R}^3$, the
location $\bt_j \in \mathbb{R}^3$ and the orientation $\br_j \in \mathbb{R}^1$,
we follow \cite{Salimans2017ICLR, Oord2016SSW} and model them with mixture
of logistics distributions
\begin{equation}
    \bs_j \sim \sum_{k=1}^K \, \pi_k^s \text{logistic}(\mu_k^s, \sigma_k^s)\quad
    \bt_j \sim \sum_{k=1}^K \, \pi_k^t \text{logistic}(\mu_k^t, \sigma_k^t)\quad
    \br_j \sim \sum_{k=1}^K \, \pi_k^r \text{logistic}(\mu_k^r, \sigma_k^r)
\end{equation}
where $\pi_k^s$, $\mu_k^s$ and $\sigma_k^s$ denote the
weight, mean and variance of the $k$-th logistic distribution used for
modeling the size. Similarly, $\pi_k^t$, $\mu_k^t$ and $\sigma_k^t$ and
$\pi_k^r$, $\mu_k^r$ ans $\sigma_k^r$ refer to the weight, mean and variance of
the $k$-th logistic of the location and orientation, respectively. In our
setup, the orientation is the angle of rotation around the up vector and the
location is the 3D centroid of the bounding box.

Similar to prior work \cite{Ritchie2019CVPR, Wang2020ARXIV}, we predict the
object attributes in an autoregressive manner: object category first, followed
by position, orientation and size as follows:
\begin{equation}
    p_{\theta}(o_j \mid o_{<j}, \bF) = p_{\theta}(\bc_j | o_{<j}, \bF)p_{\theta}(\bt_j |\bc_j, o_{<j}, \bF)
               p_{\theta}(\br_j |\bc_j, \bt_j, o_{<j}, \bF)p_{\theta}(\bs_j |\bc_j, \bt_j, \br_j, o_{<j}, \bF).
    \label{eq:bbox_likelihood}
\end{equation}
This is a natural choice, since we want our model to consider the object class
before reasoning about the size and the position of an object. To avoid
notation clutter, we omit the scene index $i$ from \eqref{eq:bbox_likelihood}.

\vspace{-2mm}
\subsection{Network Architecture}\label{subsec:network_architecture}

The input to our model is a collection of scenes in the form of $3$D labeled
bounding boxes with their corresponding room shape. Our network
consists of four main components: (i) the \emph{layout encoder} that maps the
room shape to a global feature representation $\bF$, (ii) the
\emph{structure encoder} $h_{\theta}$ that maps the $M$ objects in the
scene into per-object context embeddings $\bC=\{\bC_j\}_{j=1}^M$, (iii) the
\emph{transformer encoder} $\tau_{\theta}$ that takes $\bF$, $\bC$ and a
query embedding $\bq$ and predicts the features $\hat{\bq}$ for the next object
to be generated and (iv) the \emph{attribute extractor} that predicts the attributes of
the next object. Our model is illustrated in \figref{fig:method_overview}. The
layout encoder is simply a ResNet-18~\cite{He2016CVPR} that extracts
a feature representation $\bF \in \mathbb{R}^{64}$ for the top-down orthographic projection of the floor.

\vspace{-2mm}
\boldparagraph{Structure Encoder}%
The structure encoder $h_{\theta}$ maps the attributes of the $j$-th object into a per-object
context embedding $\bC_j$ as follows:
\begin{equation}
    \begin{split}
    h_{\theta}: \mathbb{R}^C \times \mathbb{R}^3 \times \mathbb{R}^3 \times \mathbb{R}^1 &\rightarrow \mathbb{R}^{L_c} \times \mathbb{R}^{L_s} \times \mathbb{R}^{L_t} \times \mathbb{R}^{L_r} \\
    (\bc, \bs, \bt, \br) &\mapsto [\lambda(\bc); \gamma(\bs); \gamma(\bt); \gamma(\br)]
    \end{split}
    \label{eq:structure_encoder}
\end{equation}
where $L_c, L_s, L_t, L_r$ are the output dimensionalities of the
embeddings used to map the category, the size, the location and the
orientation into a higher dimensional space respectively and $[\cdot\,;\cdot]$ denotes concatenation.
For the object category $\bc_j$ we use a learnable embedding $\lambda(\cdot)$,
whereas for the size $\bs_j$, the position $\bt_j$ and the orientation
$\br_j$, we use the positional encoding of \cite{Vaswani2017NIPS} as follows
\begin{equation}
    \gamma(p) = (\sin(2^0\pi p), \cos(2^0\pi p), \dots, \sin(2^{L-1}\pi p), \cos(2^{L-1}\pi p))
\end{equation}
where $p$ can be any of the size, position or orientation attributes and
$\gamma(\cdot)$ is applied separately in each attribute's dimension.
The set of per-object context vectors synthesizes
the context embedding $\bC$ that encapsulates information for
the existing objects in the scene and is used to condition the
next object to be generated. Before passing the output of
\eqref{eq:structure_encoder} to the transformer encoder, we map each $\bC_j$ to
$64$ dimensions using a linear projection.

\vspace{-2mm}
\boldparagraph{Transformer Encoder}%
We follow \cite{Vaswani2017NIPS, Devlin2019NAACL} and implement our encoder
$\tau_{\theta}$ as a multi-head attention transformer without any positional
encoding. This allows us to learn a parametric function that computes features
that are invariant to the order of $\bC_j$ in $\bC$. We use these features to
predict the next object to be added in the scene, creating an
autoregressive model. The input set of the transformer is

$\bI=\{\bF\}\cup\{\bC_j\}_{j=1}^M \cup{\bq}$, with $M$ the number of
objects in the scene. $\bq \in \mathbb{R}^{64}$ is a learnable object
query vector that allows the transformer to predict output features
$\hat{\bq} \in \mathbb{R}^{64}$ used for generating the next
object to be added in the scene.  The use of a query token is akin to the use of a mask
embedding in Masked Language Modelling \cite{Devlin2019NAACL} or the class
embedding for the Vision Transformer \cite{Sharir2021ICLR}.

\vspace{-2mm}
\boldparagraph{Attribute Extractor}%
We autoregressively predict the attributes
of the next object to be added in the scene using one MLP for each attribute.
More formally, the attribute extractor is defined as follows:
\begin{align}
    c_{\theta}: \mathbb{R}^{64} &\rightarrow \mathbb{R}^C &\hat{\bq} \mapsto \hat{\bc} \label{eq:att_ectractor_1} \\
    t_{\theta}: \mathbb{R}^{64}\times \mathbb{R}^{L_c} &\rightarrow \mathbb{R}^{3\times 3\times K}
        & (\hat{\bq}, \lambda(\bc)) \mapsto \hat{\bt} \\
    r_{\theta}: \mathbb{R}^{64}\times \mathbb{R}^{L_c} \times\mathbb{R}^{L_t} &\rightarrow \mathbb{R}^{1\times 3\times K}
    & (\hat{\bq}, \lambda(\bc), \gamma(\bt)) \mapsto \hat{\br} \\
    s_{\theta}: \mathbb{R}^{64}\times \mathbb{R}^{L_c} \times\mathbb{R}^{L_t} \times \mathbb{R}^{L_r} &\rightarrow \mathbb{R}^{3\times 3\times K}
    & (\hat{\bq}, \lambda(\bc), \gamma(\bt), \gamma(\br)) \mapsto \hat{\bs} \label{eq:att_ectractor_2}
\end{align}
where $\hat{\bc}, \hat{\bs}, \hat{\bt}, \hat{\br}$ are the predicted
attribute distributions and $c_\theta$, $t_\theta$, $r_\theta$ and $s_\theta$ are mappings between the latent space and the low-dimensional space of attributes.
For the object category, $c_\theta$ predicts $C$ class probabilities, whereas,
$t_\theta$, $r_\theta$ and $s_\theta$ predict the mean, variance and
mixing coefficient for the $K$ logistic distributions for each attribute.
To predict the object properties in an autoregressive manner, we need to
condition the prediction of a property on the previously predicted properties.
Thus, instead of only passing $\hat{\bq}$ to each MLP, we concatenate it with
the previously predicted attributes, mapped in a
higher-dimensional space using the embeddings
$\lambda(\cdot)$ and $\gamma(\cdot)$ from \eqref{eq:structure_encoder}.

\begin{figure*}
    \centering
    \includegraphics[width=1\textwidth]{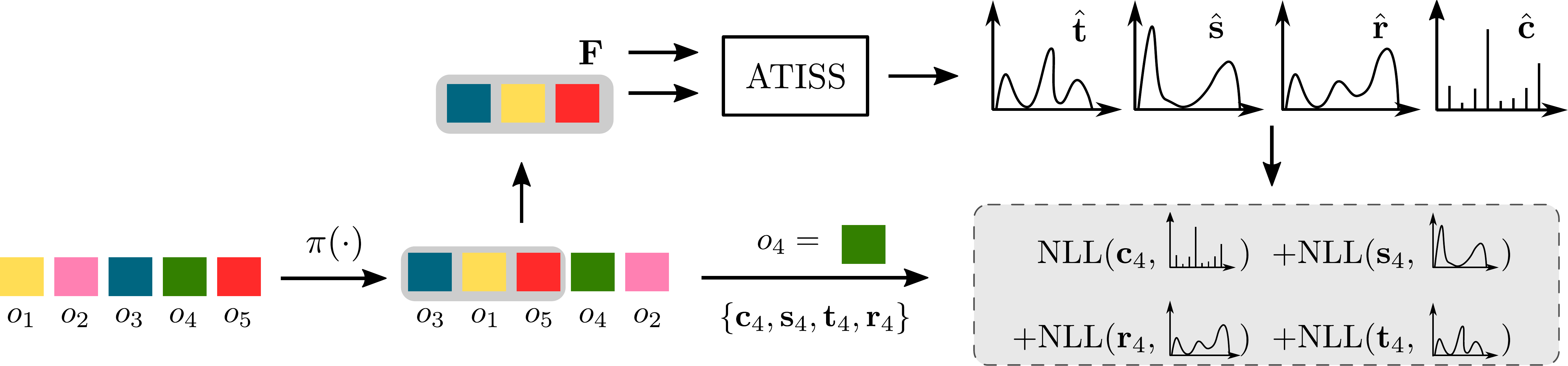}
    \caption{\small
        \textbf{Training Overview:} Given a scene with $M$ objects (coloured
        squares), we first randomly permute them and then keep the first $T$
        objects (here $T=3$). We task our network to predict
        the next object to be added in the scene given the subset of kept objects
        (highlighted with grey) and its floor layout feature
        $\bF$. Our loss function is the negative log-likelihood
        (NLL) of the next object in the permuted sequence (green square).
    }
    \label{fig:training_overview}
    \vspace{-1.2em}
\end{figure*}

\vspace{-2mm}
\subsection{Training and Inference}\label{subsec:training_inference}

During training, we choose a scene from the dataset and apply a random
permutation $\pi(\cdot)$ on its $M$ objects. Then, we randomly select the first
$T$ objects to compute the context embedding $\bC$. Conditioned on $\bC$ and
$\bF$, our network predicts the attribute distributions of the
next object to be added in the scene and is trained to maximize the
log-likelihood of the $T+1$ object from the permuted scene. A pictorial
representation of the training process is provided in
\figref{fig:training_overview}. To indicate the end of sequence, we
augment the $C$ object classes with an additional class, which we refer to as
\emph{end symbol}.

During inference, we start with an empty context embedding $\bC=\emptyset$ and
the floor representation $\bF$ of the room to be populated and autoregressively
sample attribute values from the predicted distributions of
\eqref{eq:att_ectractor_1}-\eqref{eq:att_ectractor_2} for the next object.
Once a new object is
generated, it is appended to the context $\bC$ to be used in the next
step of the generation process until the \emph{end symbol} is generated.  A
pictorial representation of the generation process can be found in
\figref{fig:method_overview}. In order to transform the predicted
labeled bounding boxes to 3D models we use object retrieval. In particular, we
retrieve the closest object from the dataset in terms of the euclidean distance
of the bounding box dimensions.

\vspace{-2mm}
\section{Experimental Evaluation}
\label{sec:results}

\begin{figure}
    \centering
    \vspace{-1.2em}
        \begin{subfigure}[b]{0.17\linewidth}
		\centering
        \small Scene Layout
    \end{subfigure}%
        \begin{subfigure}[b]{0.17\linewidth}
		\centering
        \small Training Sample
    \end{subfigure}%
        \begin{subfigure}[b]{0.17\linewidth}
		\centering
        \small FastSynth
    \end{subfigure}%
        \begin{subfigure}[b]{0.17\linewidth}
		\centering
        \small SceneFormer
    \end{subfigure}%
        \begin{subfigure}[b]{0.17\linewidth}
        \centering
        \small Ours
    \end{subfigure}
    \hfill%
    \vskip\baselineskip%
    \vspace{-0.75em}
    \hfill%
    \begin{subfigure}[b]{0.17\linewidth}
		\centering
		\includegraphics[width=0.8\linewidth]{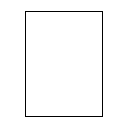}
    \end{subfigure}%
        \begin{subfigure}[b]{0.17\linewidth}
		\centering
		\includegraphics[width=\linewidth, trim=550 200 600 200, clip]{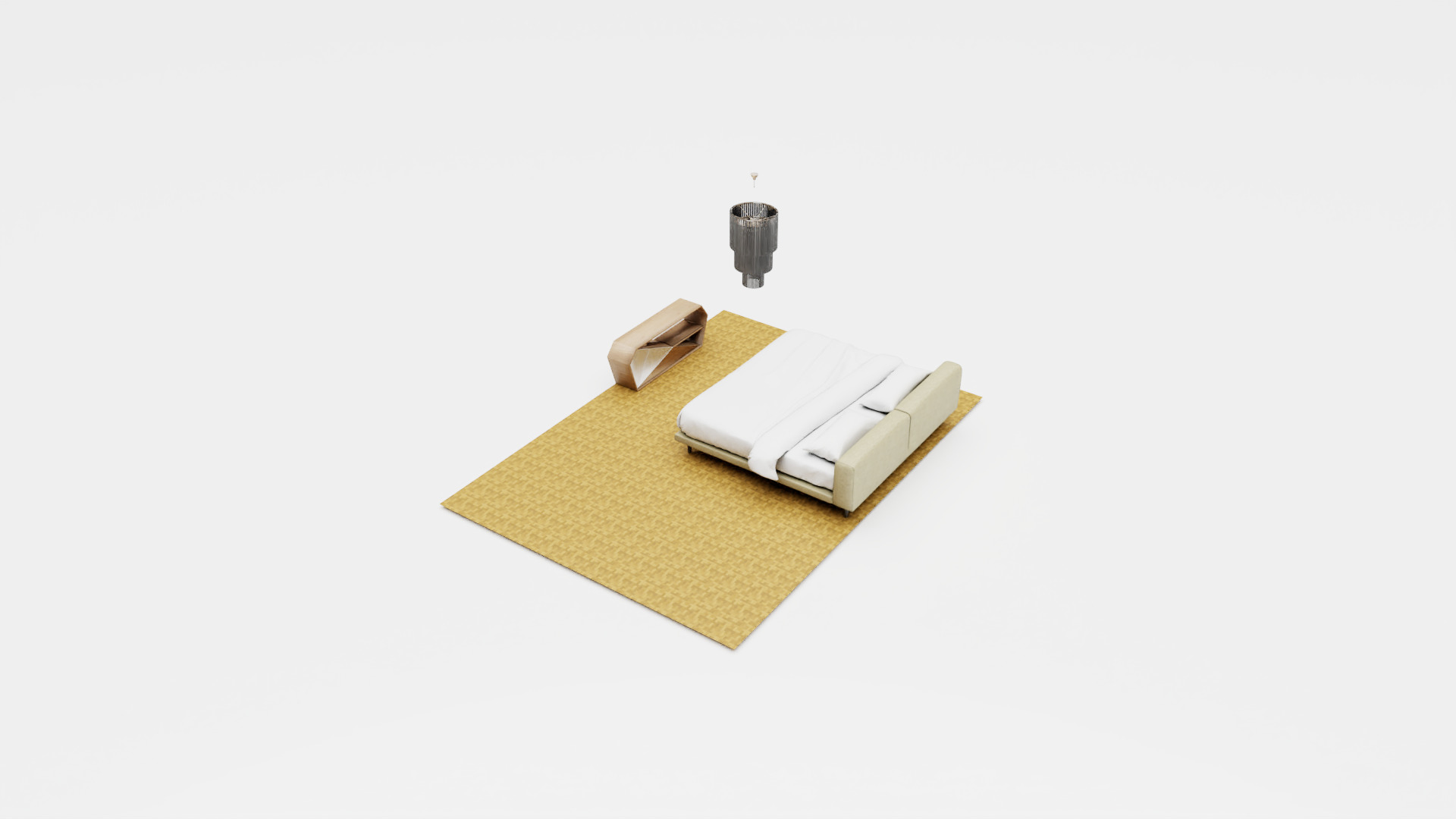}
    \end{subfigure}%
        \begin{subfigure}[b]{0.17\linewidth}
		\centering
		\includegraphics[width=\linewidth, trim=550 200 550 150, clip]{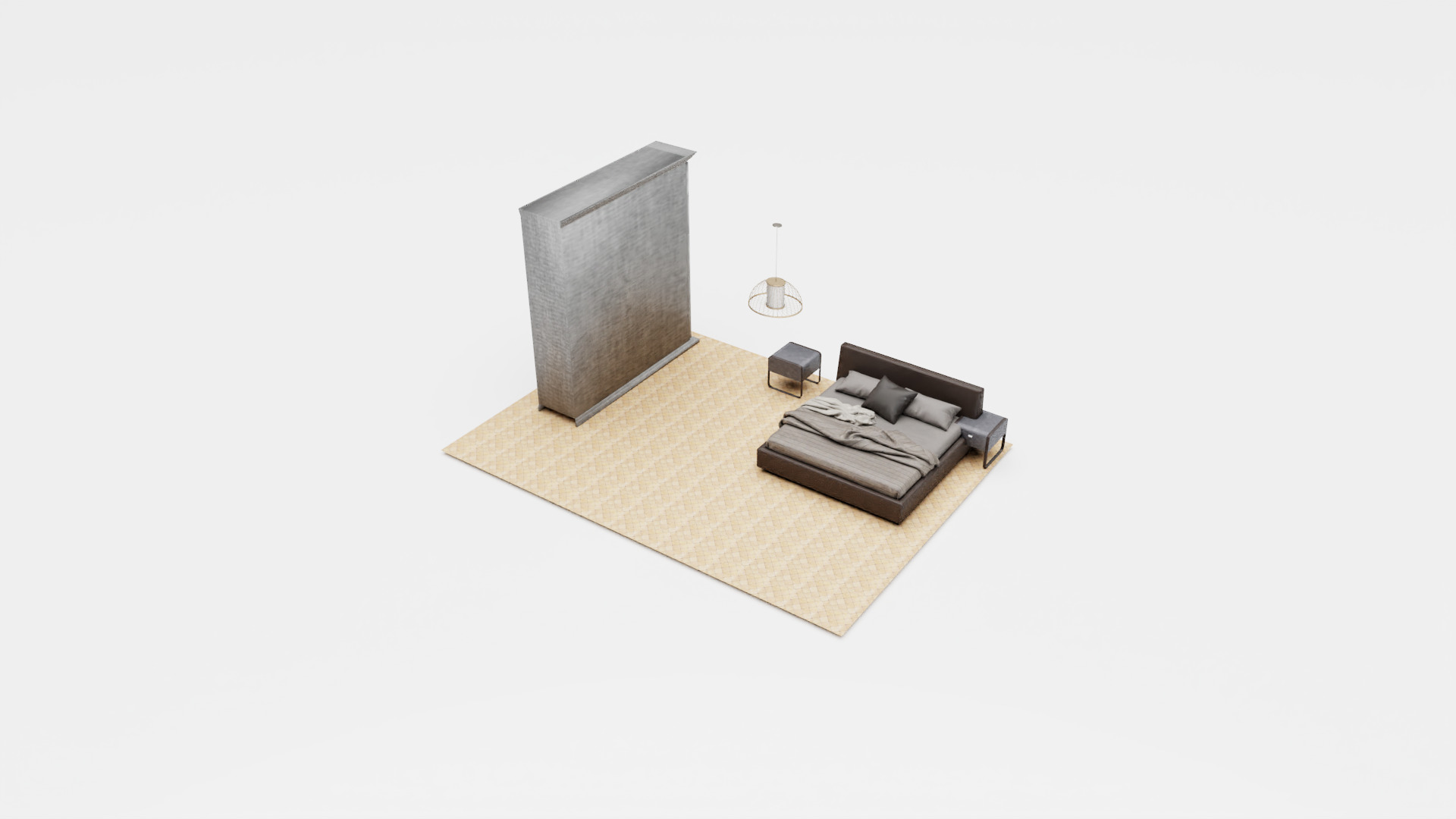}
    \end{subfigure}%
        \begin{subfigure}[b]{0.17\linewidth}
		\centering
		\includegraphics[width=\linewidth, trim=550 200 600 200, clip]{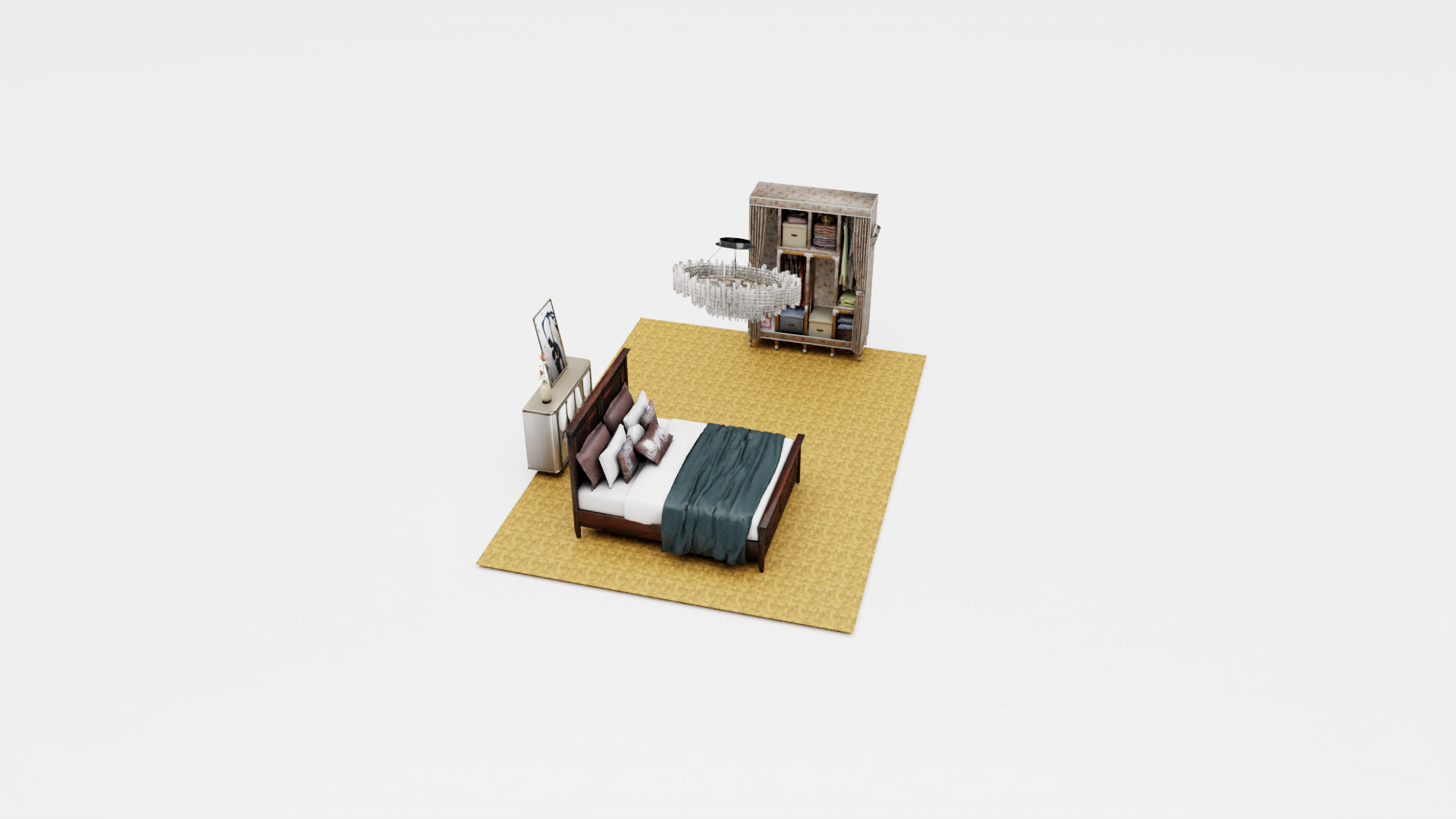}
    \end{subfigure}%
    \begin{subfigure}[b]{0.17\linewidth}
		\centering
		\includegraphics[width=\linewidth, trim=550 200 600 200, clip]{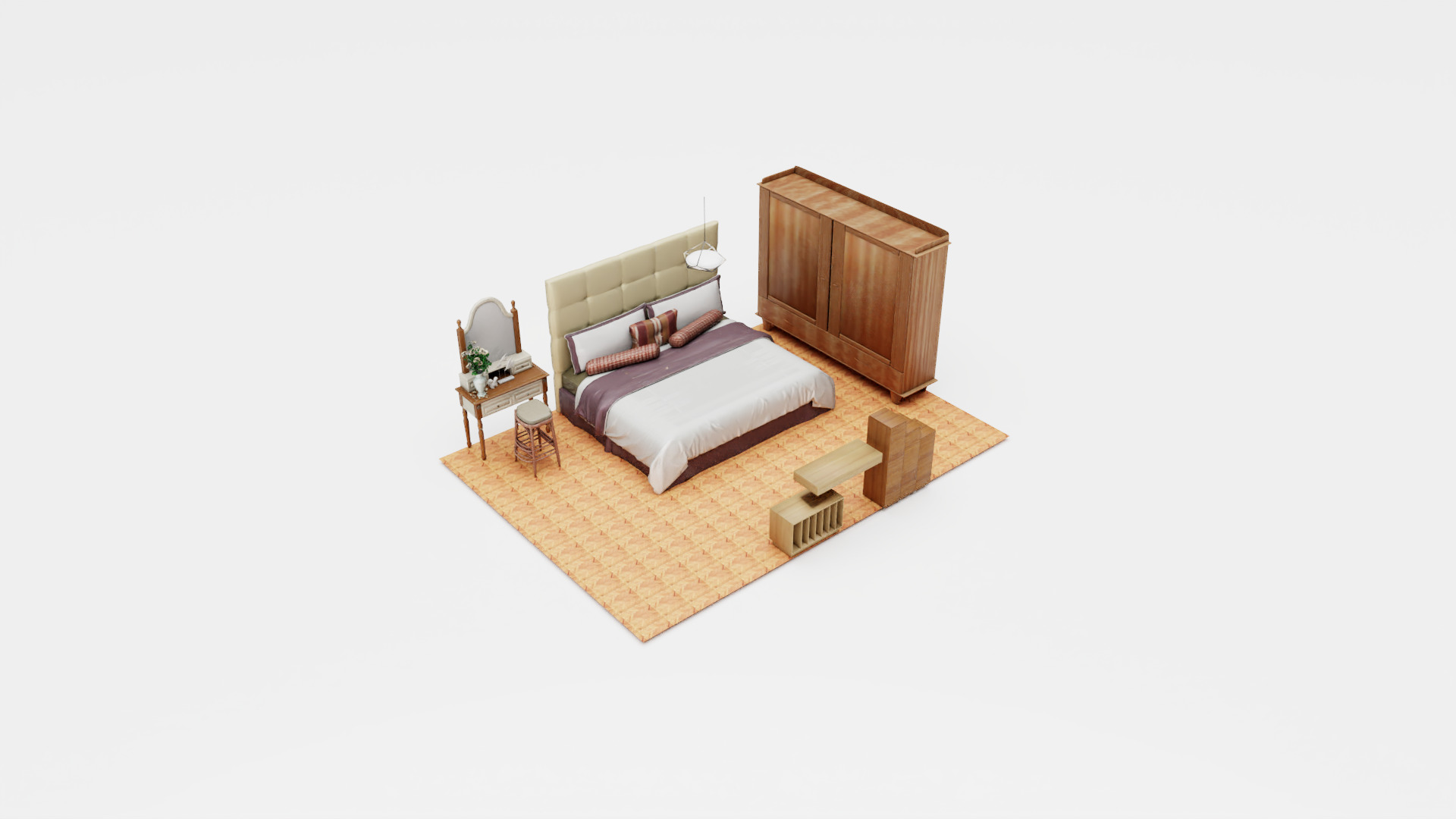}
    \end{subfigure}%
    \hfill%
    \vspace{-1.2em}
    \vskip\baselineskip%
    \hfill%
    \begin{subfigure}[b]{0.17\linewidth}
		\centering
		\includegraphics[width=0.8\linewidth]{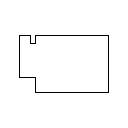}
    \end{subfigure}%
        \begin{subfigure}[b]{0.17\linewidth}
		\centering
		\includegraphics[width=\linewidth, trim=600 250 600 100, clip]{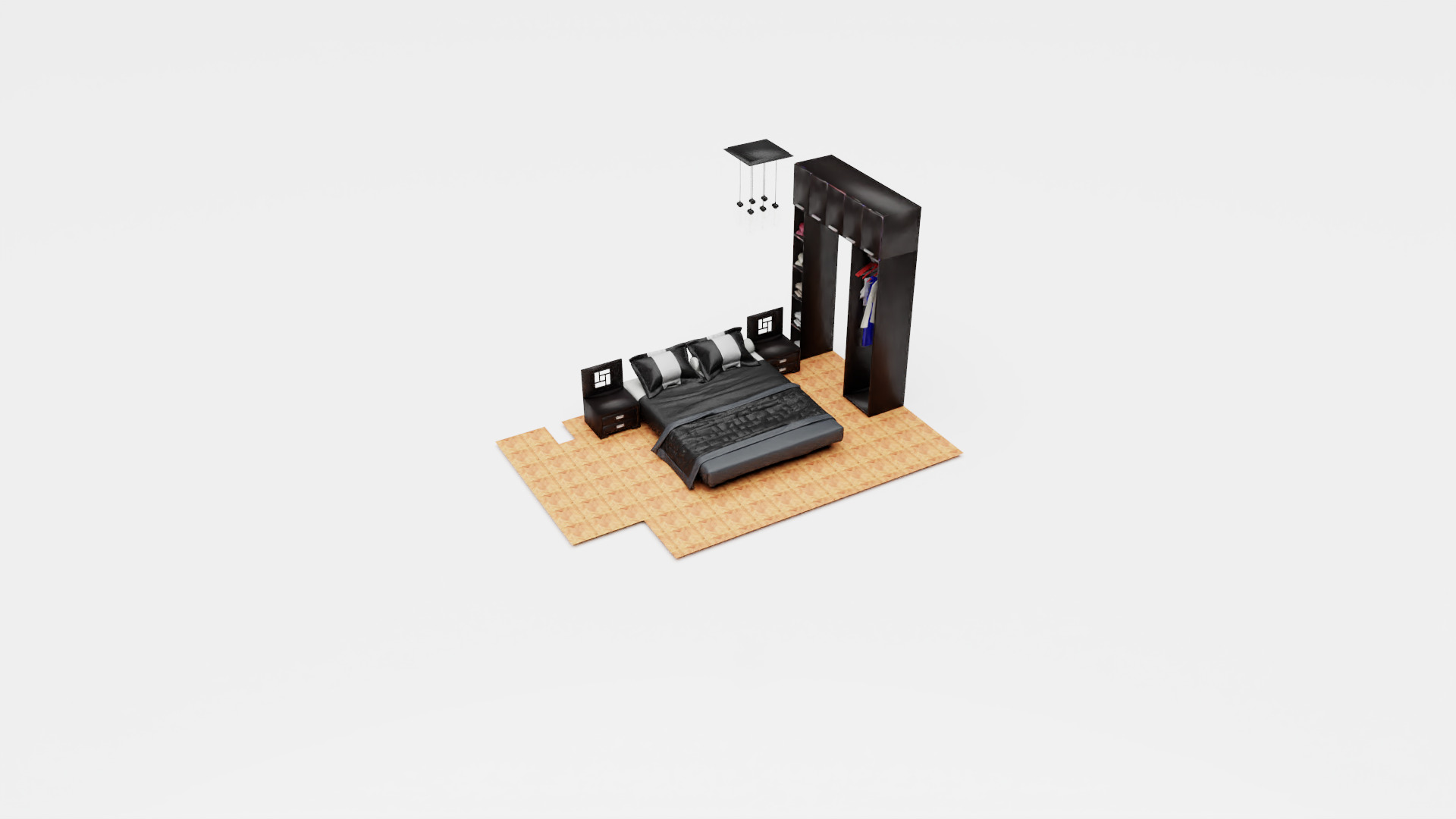}
    \end{subfigure}%
        \begin{subfigure}[b]{0.17\linewidth}
		\centering
		\includegraphics[width=\linewidth, trim=600 250 600 100, clip]{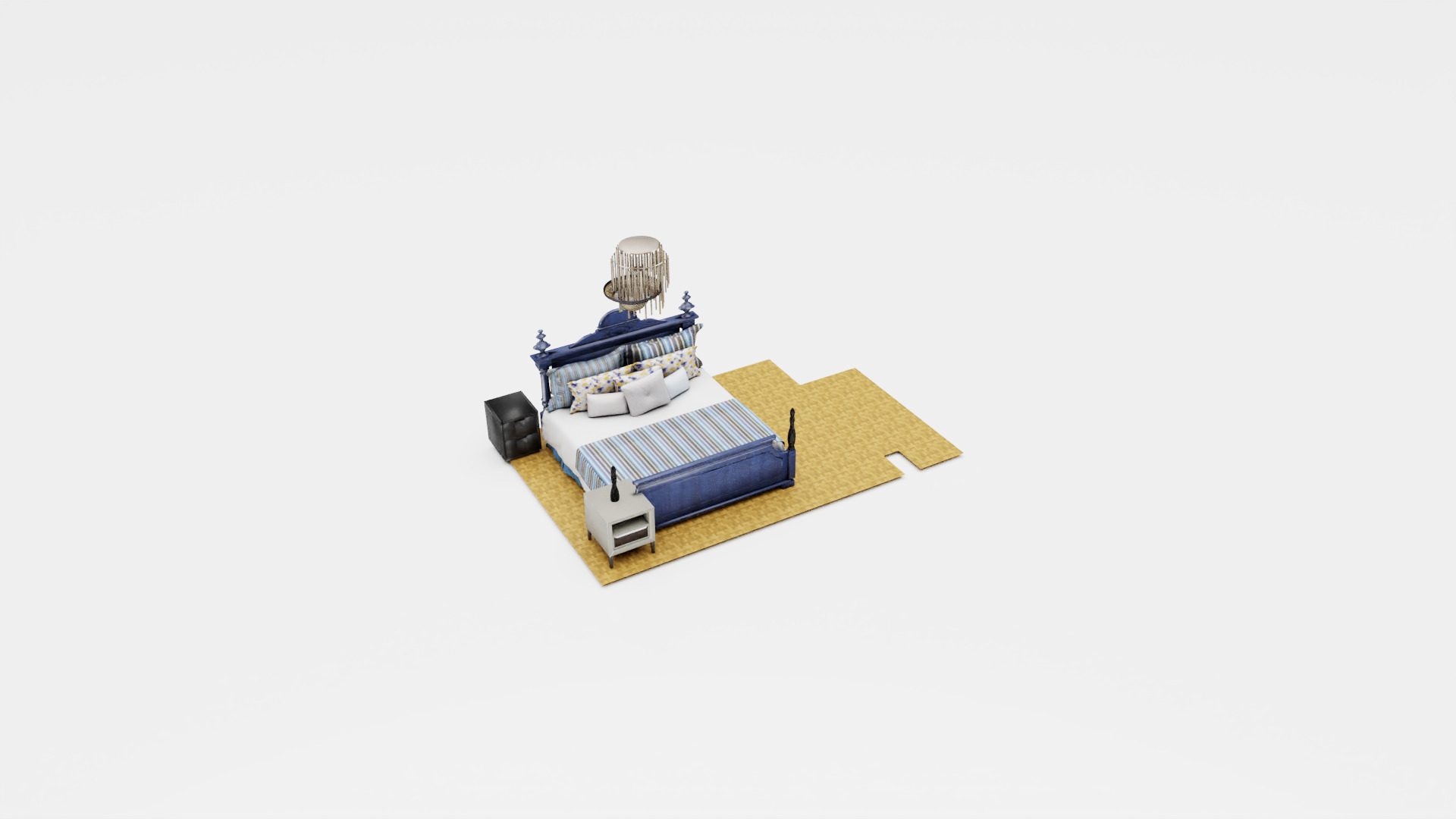}
    \end{subfigure}%
        \begin{subfigure}[b]{0.17\linewidth}
		\centering
		\includegraphics[width=\linewidth, trim=600 250 600 100, clip]{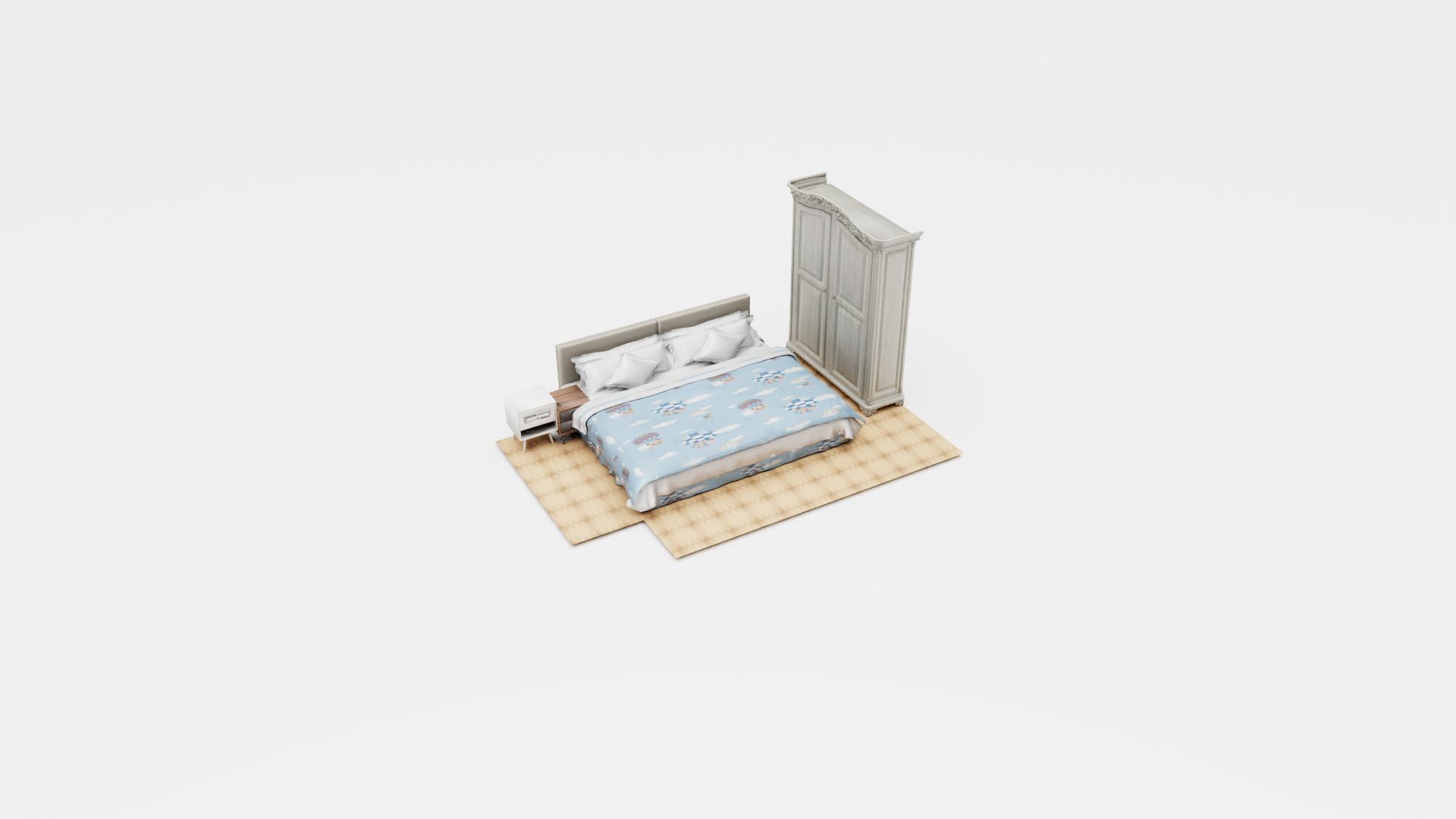}
    \end{subfigure}%
    \begin{subfigure}[b]{0.17\linewidth}
		\centering
		\includegraphics[width=\linewidth, trim=600 250 600 100, clip]{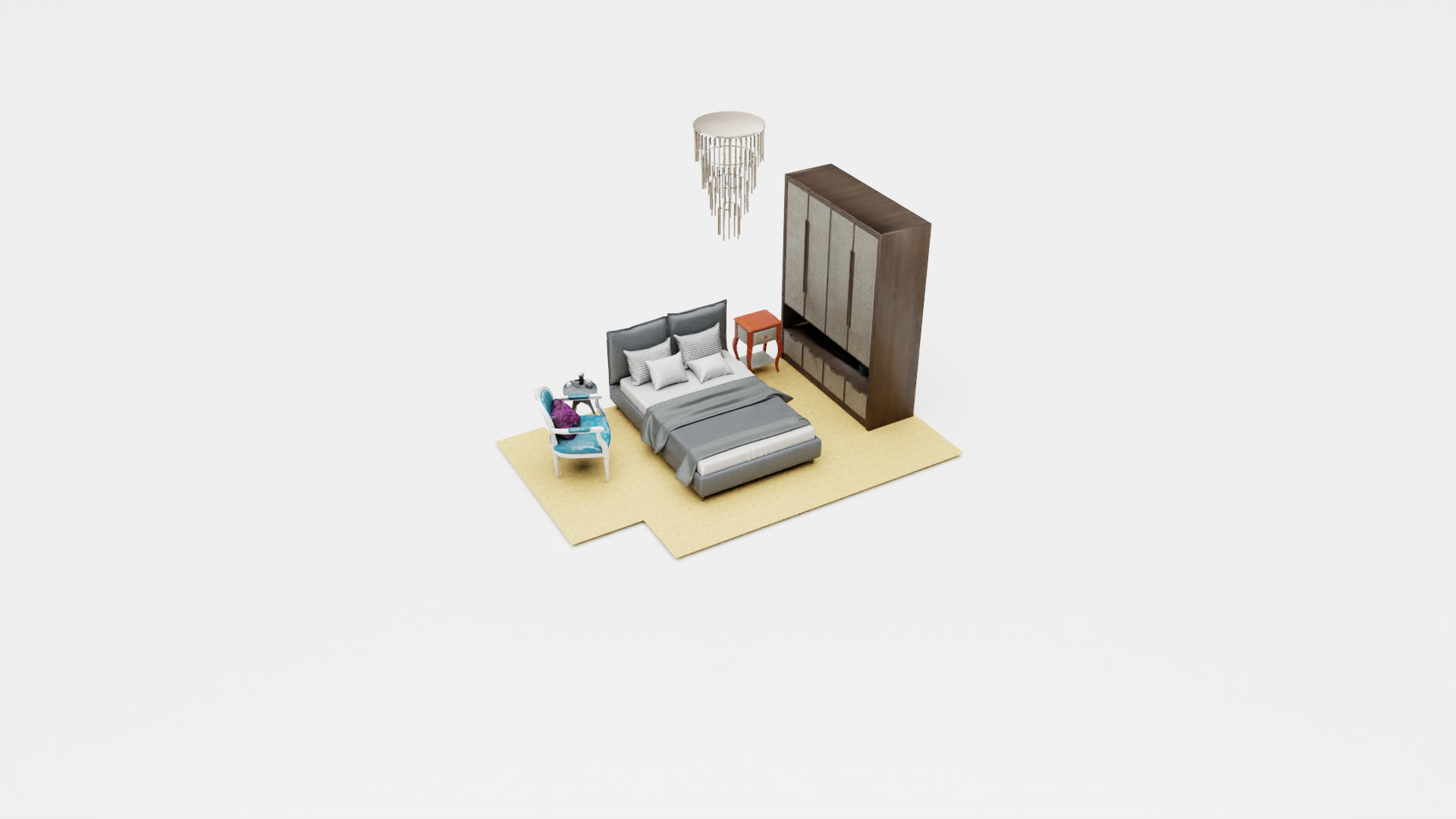}
    \end{subfigure}%
    \hfill%
    \vspace{-1.5em}
    \vskip\baselineskip%
    \hfill%
    \begin{subfigure}[b]{0.17\linewidth}
		\centering
		\includegraphics[width=\linewidth]{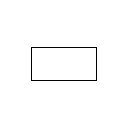}
    \end{subfigure}%
        \begin{subfigure}[b]{0.17\linewidth}
		\centering
		\includegraphics[width=\linewidth, trim=600 100 500 100, clip]{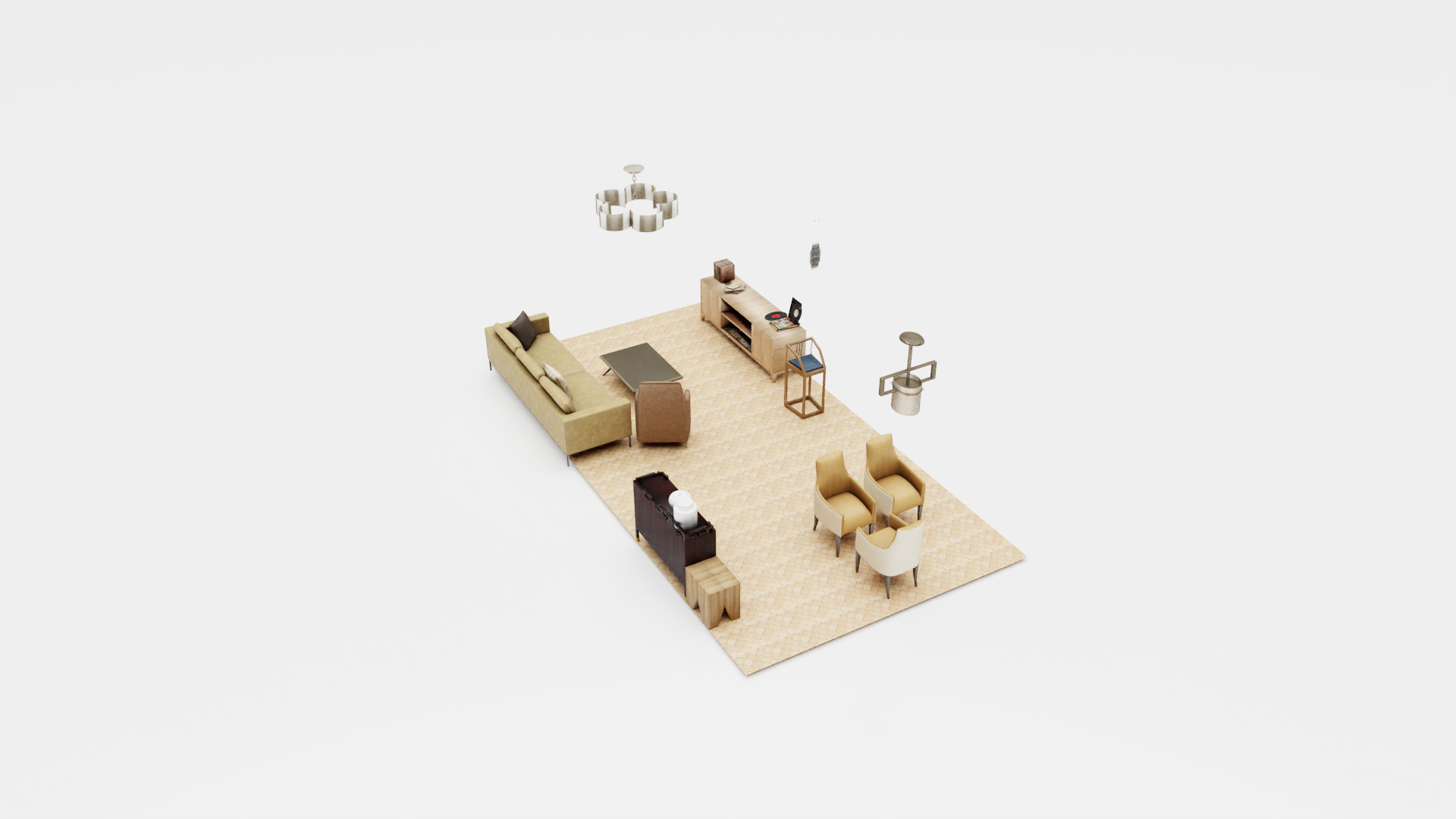}
    \end{subfigure}%
        \begin{subfigure}[b]{0.17\linewidth}
		\centering
		\includegraphics[width=\linewidth, trim=600 100 500 100, clip]{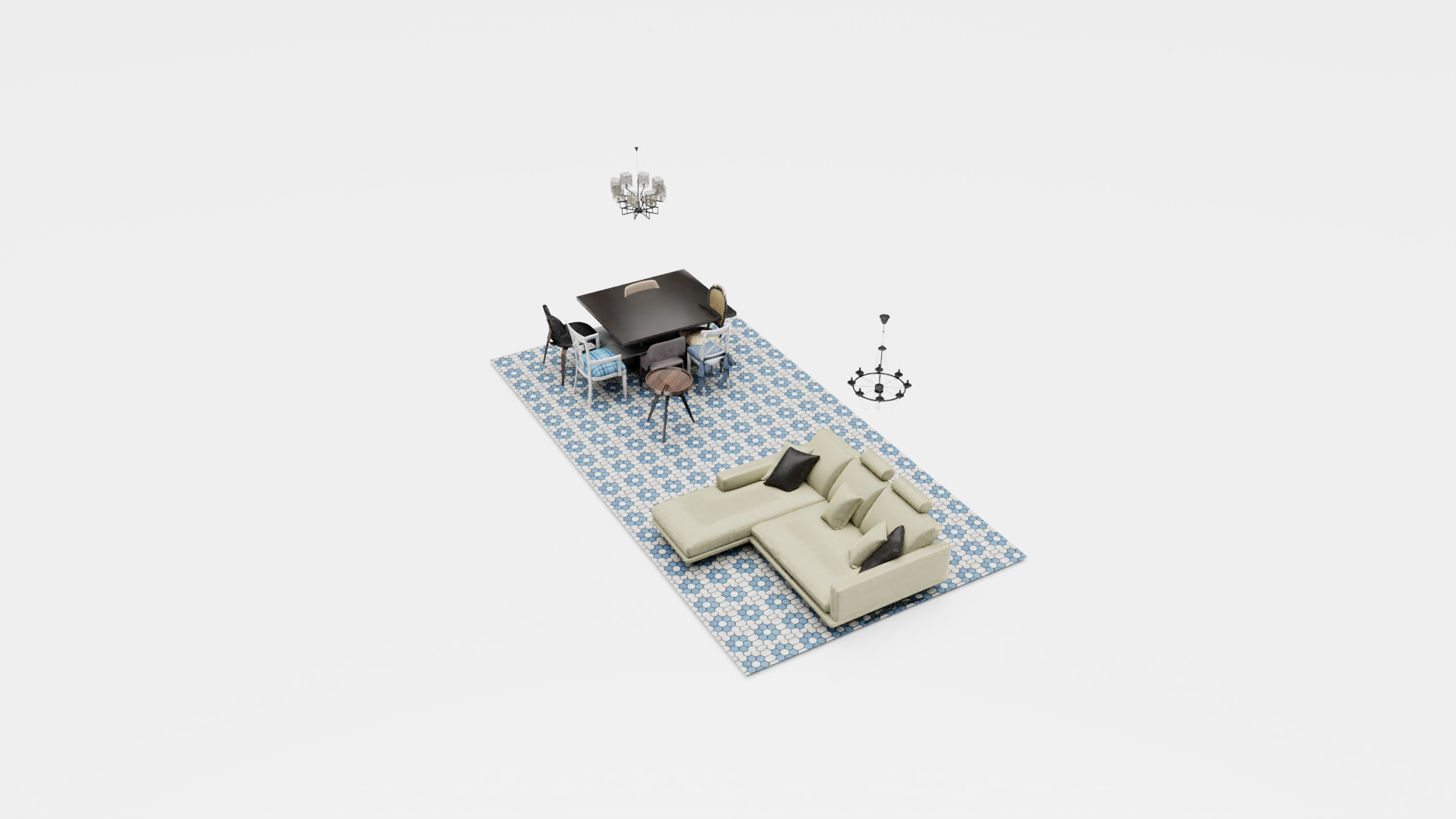}
    \end{subfigure}%
        \begin{subfigure}[b]{0.17\linewidth}
		\centering
		\includegraphics[width=\linewidth, trim=600 100 500 100, clip]{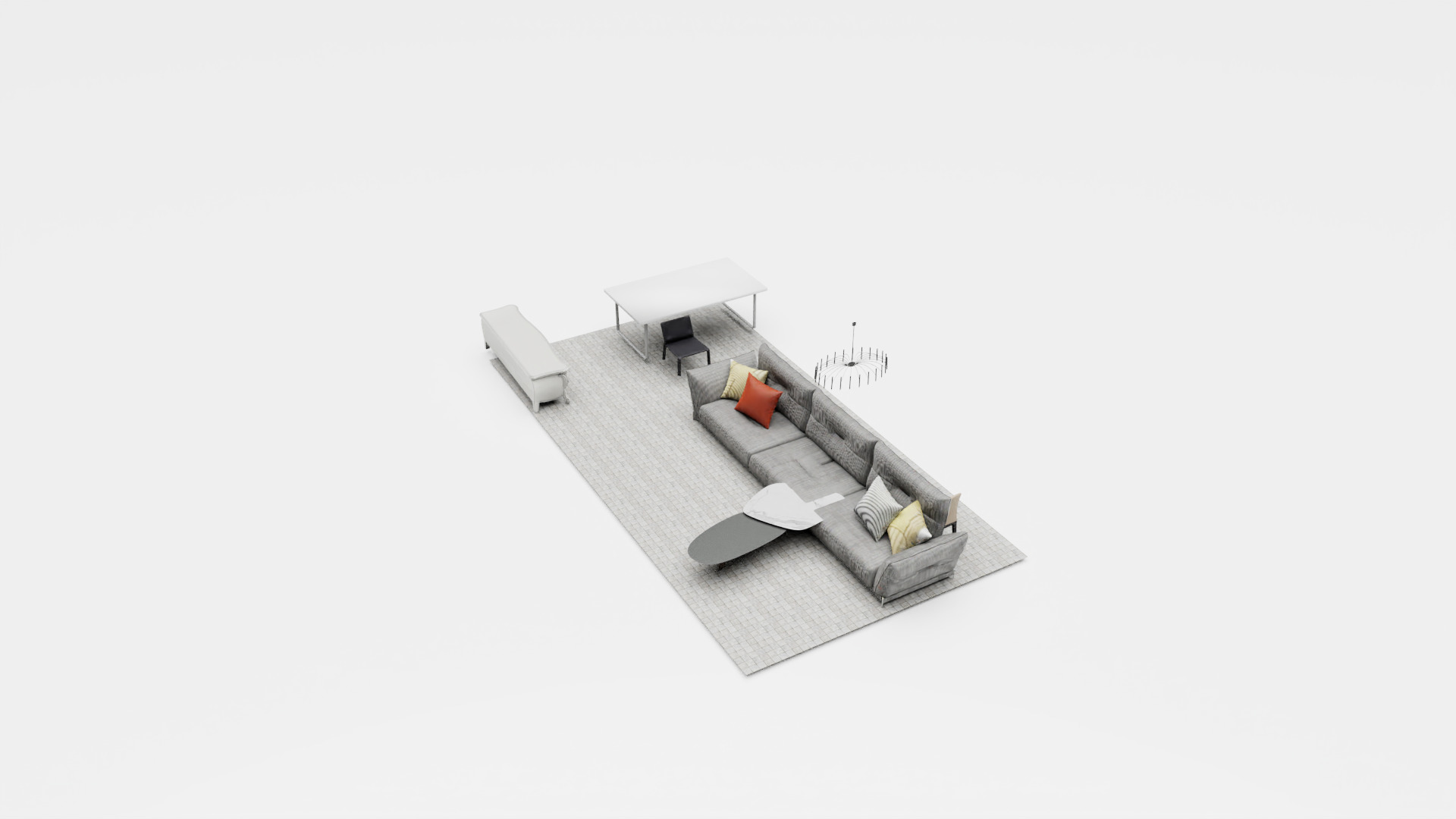}
    \end{subfigure}%
    \begin{subfigure}[b]{0.17\linewidth}
		\centering
		\includegraphics[width=\linewidth, trim=600 100 500 100, clip]{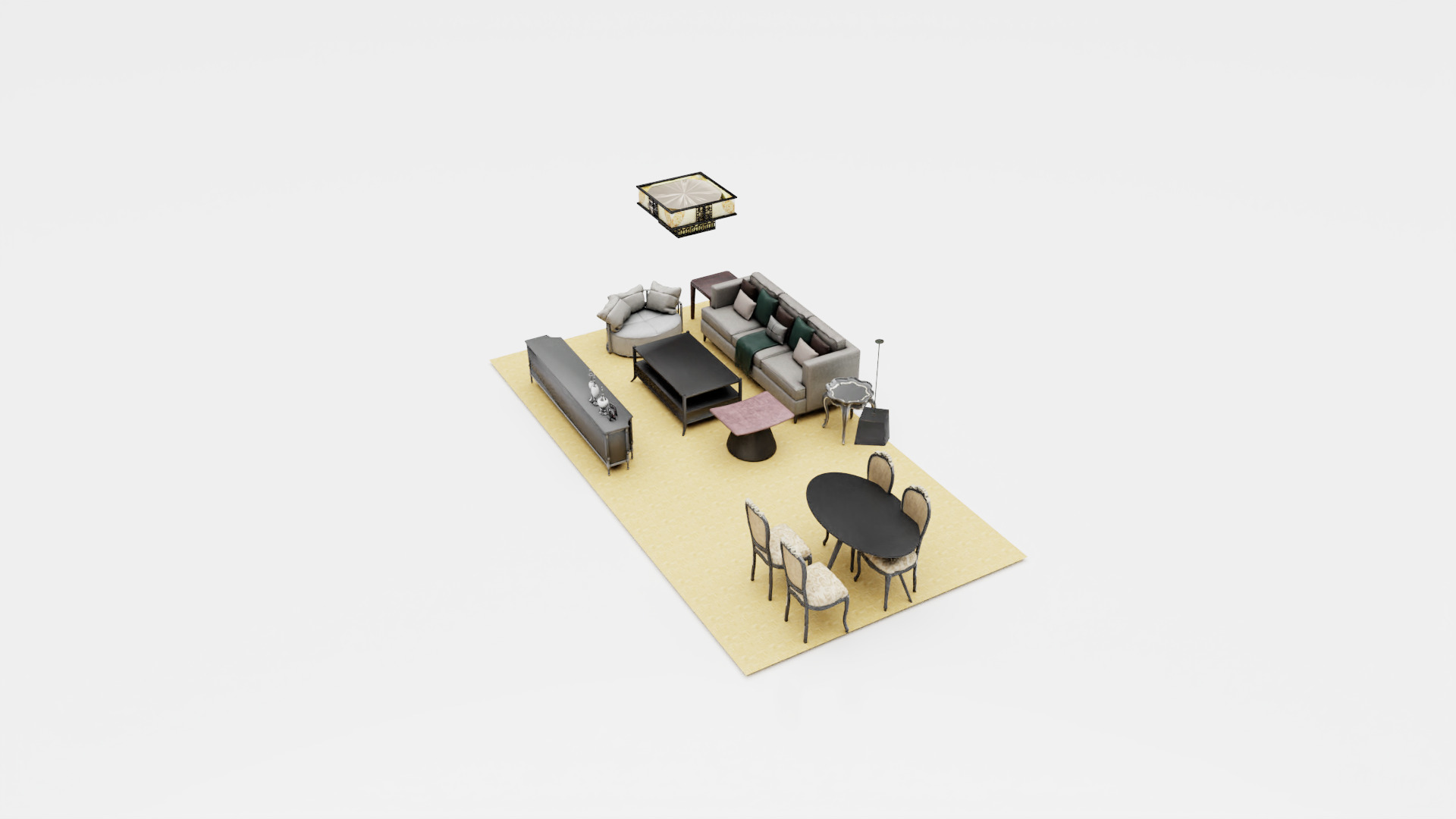}
    \end{subfigure}%
    \hfill%
    \vspace{-1.5em}
        \hfill%
    \begin{subfigure}[b]{0.17\linewidth}
		\centering
		\includegraphics[width=\linewidth]{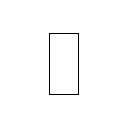}
    \end{subfigure}%
        \begin{subfigure}[b]{0.17\linewidth}
		\centering
		\includegraphics[width=\linewidth, trim=600 150 500 250, clip]{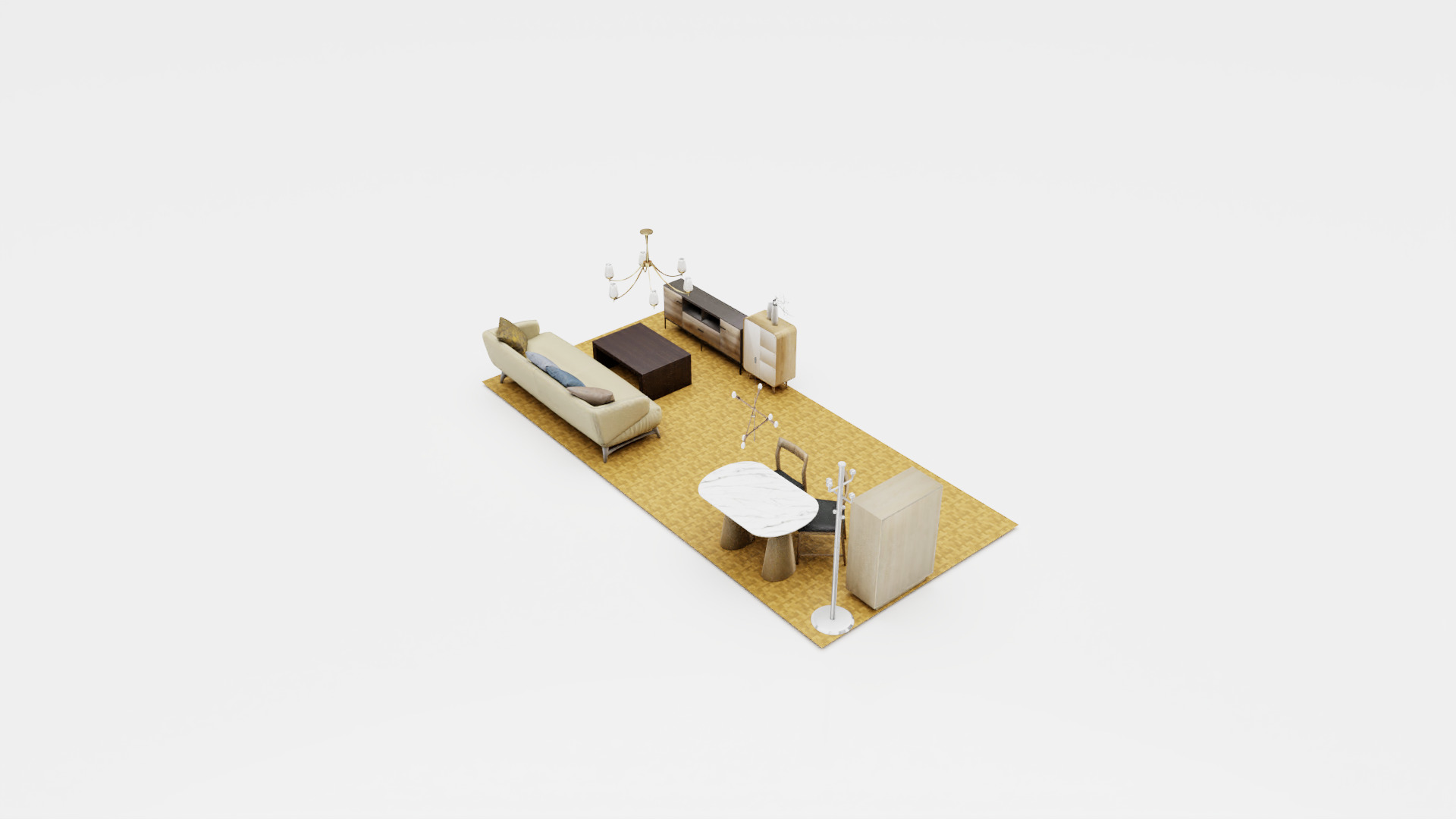}
    \end{subfigure}%
        \begin{subfigure}[b]{0.17\linewidth}
		\centering
		\includegraphics[width=\linewidth, trim=600 200 500 200, clip]{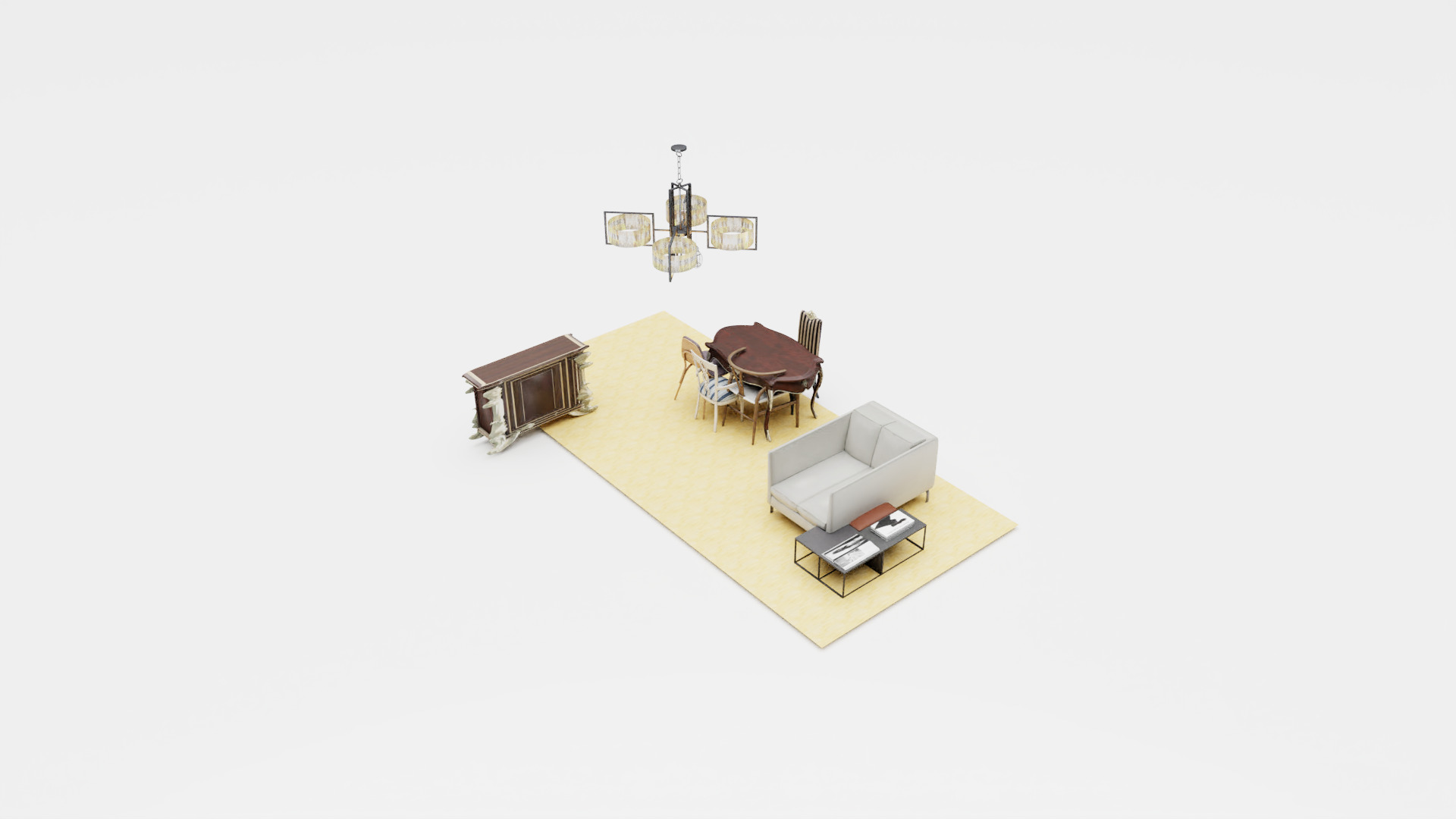}
    \end{subfigure}%
        \begin{subfigure}[b]{0.17\linewidth}
		\centering
		\includegraphics[width=\linewidth, trim=600 150 500 250, clip]{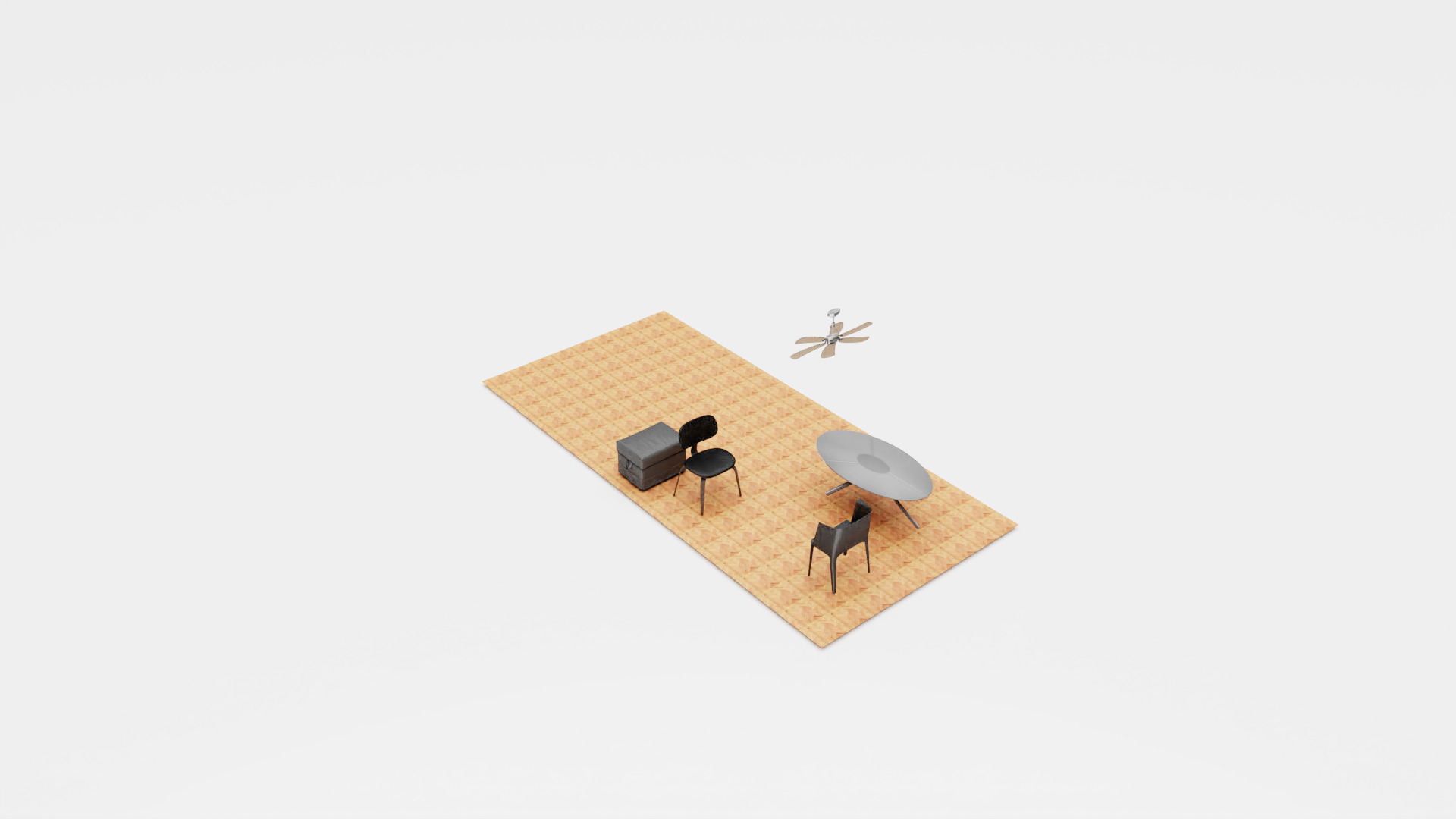}
    \end{subfigure}%
        \begin{subfigure}[b]{0.17\linewidth}
		\centering
		\includegraphics[width=\linewidth, trim=600 150 500 250, clip]{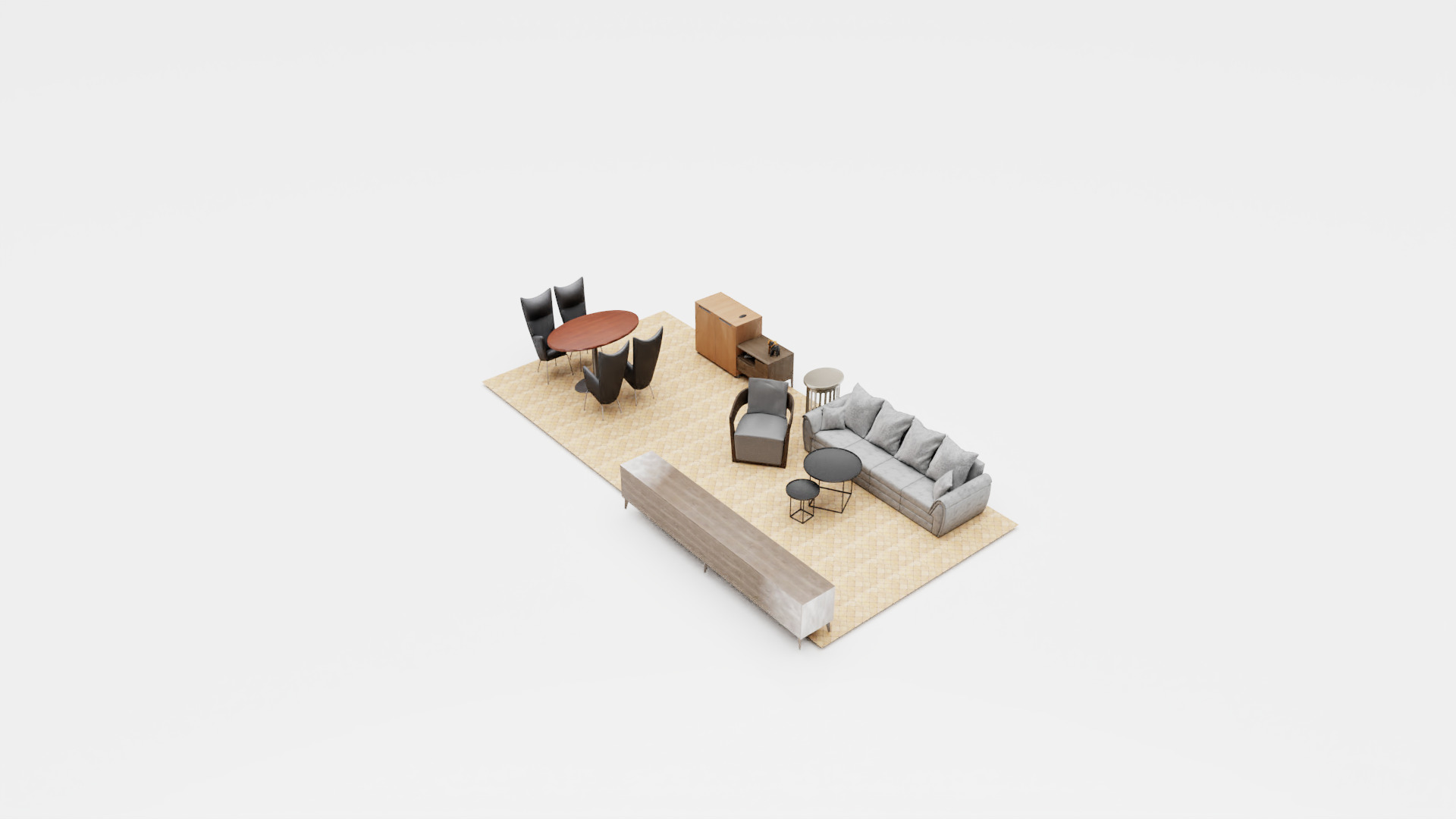}
    \end{subfigure}%
    \hfill%
    \caption{\small {\bf Qualitative Scene Synthesis Results}. Synthesized scenes for three room types: bedrooms (1st$+$2nd row), living room (3rd row), dining room (4th row) using FastSynth, SceneFormer and our method. To showcase the generalization abilities of our model we also show the closest scene from the training set (2nd column).}
    \label{fig:scene_synthesis_qualitative}
    \vspace{-1.2em}
\end{figure}

In this section, we provide an extensive evaluation of our method, comparing it to existing baselines. We further showcase several interactive use cases enabled by our method, not previously possible. Additional results as well as implementation details are provided in the supplementary.

\vspace{-2mm}
\boldparagraph{Datasets}%
We train our model on the 3D-FRONT dataset \cite{Fu2020ARXIVa} which contains a collection
of $6,813$ houses with roughly $14,629$ rooms,
populated with 3D furniture objects from the 3D-FUTURE dataset
\cite{Fu2020ARXIVb}. In our evaluation, we focus on four room types: (i) bedrooms,
(ii) living rooms, (iii) dining rooms and (iv) libraries. After
pre-processing to filter out uncommon object arrangements and rooms with
unnatural sizes, we obtained $5996$ bedrooms, $2962$ living rooms,
$2625$ dining rooms and $622$ libraries. We use $21$ object categories for the
bedrooms, $24$ for the living and dining rooms and $25$ for the
libraries. The preprocessing steps are discussed in the
supplementary. 

\vspace{-2mm}
\boldparagraph{Baselines}%
We compare our approach to
FastSynth~\cite{Ritchie2019CVPR}
and 
SceneFormer~\cite{Wang2020ARXIV}
using the authors' implementations. Note that both approaches were
originally evaluated on the SUNCG dataset \cite{Song2017CVPR},
which is now unavailable. Thus, we retrained both on 3D-FRONT.
We also compare with a variant of our model that generates scenes
as ordered sequences of objects (Ours+Order). To incorporate the order
information to the input, we utilize a positional embedding
\cite{Vaswani2017NIPS} and a fixed ordering based on the object frequency as
described in \cite{Wang2020ARXIV}.

\vspace{-2mm}
\boldparagraph{Evaluation Metrics}%
To measure the realism of the generated scenes, we
follow prior work \cite{Ritchie2019CVPR} and report
the KL divergence between the object category distributions of synthesized
and real scenes from the test set and the classification accuracy of a
classifier trained to discriminate real from synthetic scenes. We also report the FID~\cite{Heusel2017NIPS} between
top-down orthographic projections of synthesized and real scenes from the test
set, which we compute using \cite{Parmar2021ARXIV} on $256^2$ images.
We repeat the metric computation for FID and classification accuracy $10$ times and report the average.   

\begin{figure}[t]
    \centering
    \hfill%
    \begin{minipage}[b]{0.03\linewidth}
    \rotatebox{90}{~ \small{FastSynth}}
    \end{minipage}
    \begin{minipage}[b]{0.1\linewidth}
		\centering
		\includegraphics[width=\linewidth, trim=600 0 500 0, clip]{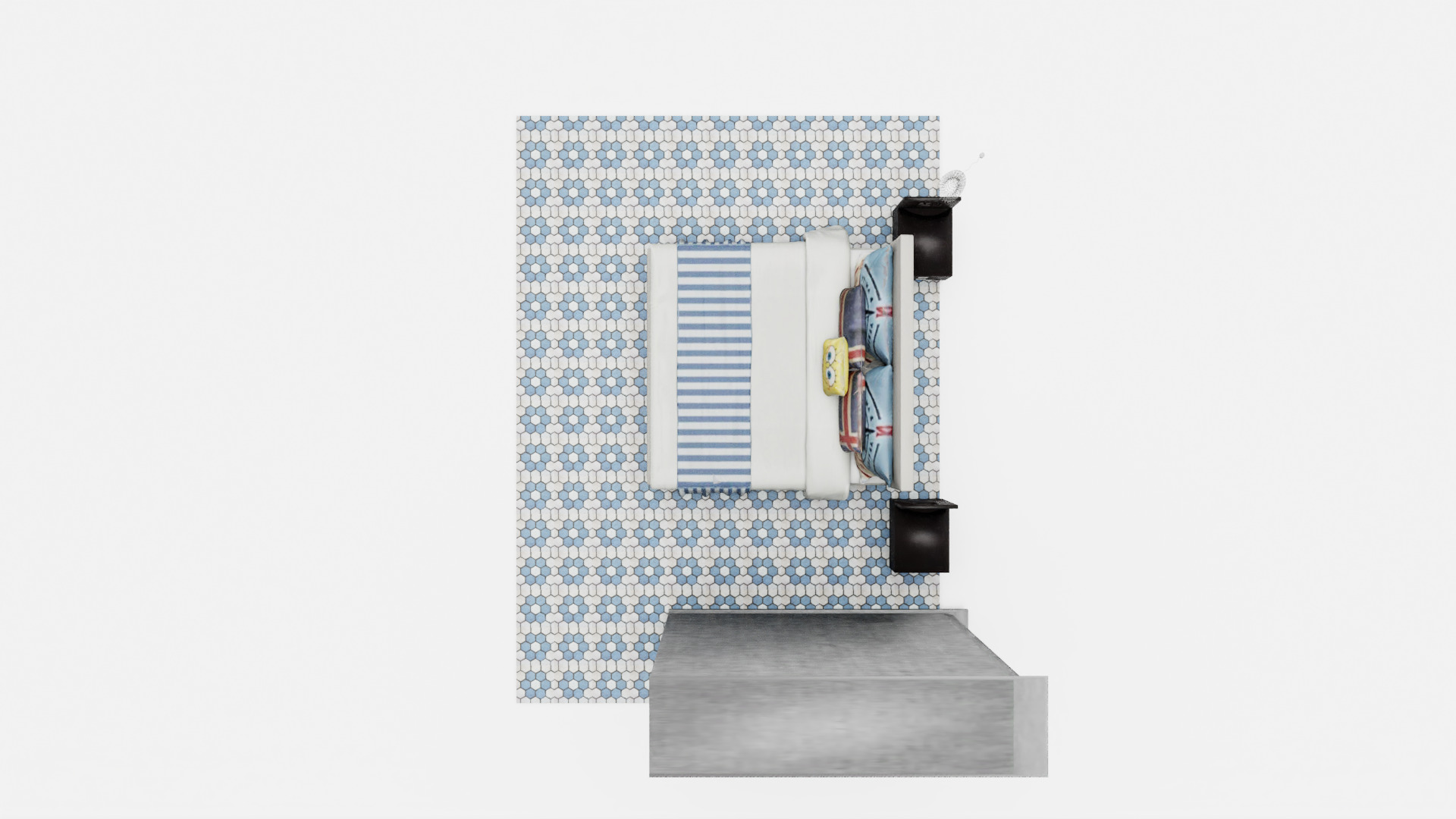}
    \end{minipage}%
        \begin{minipage}[b]{0.1\linewidth}
		\centering
		\includegraphics[width=\linewidth, trim=600 0 500 0, clip]{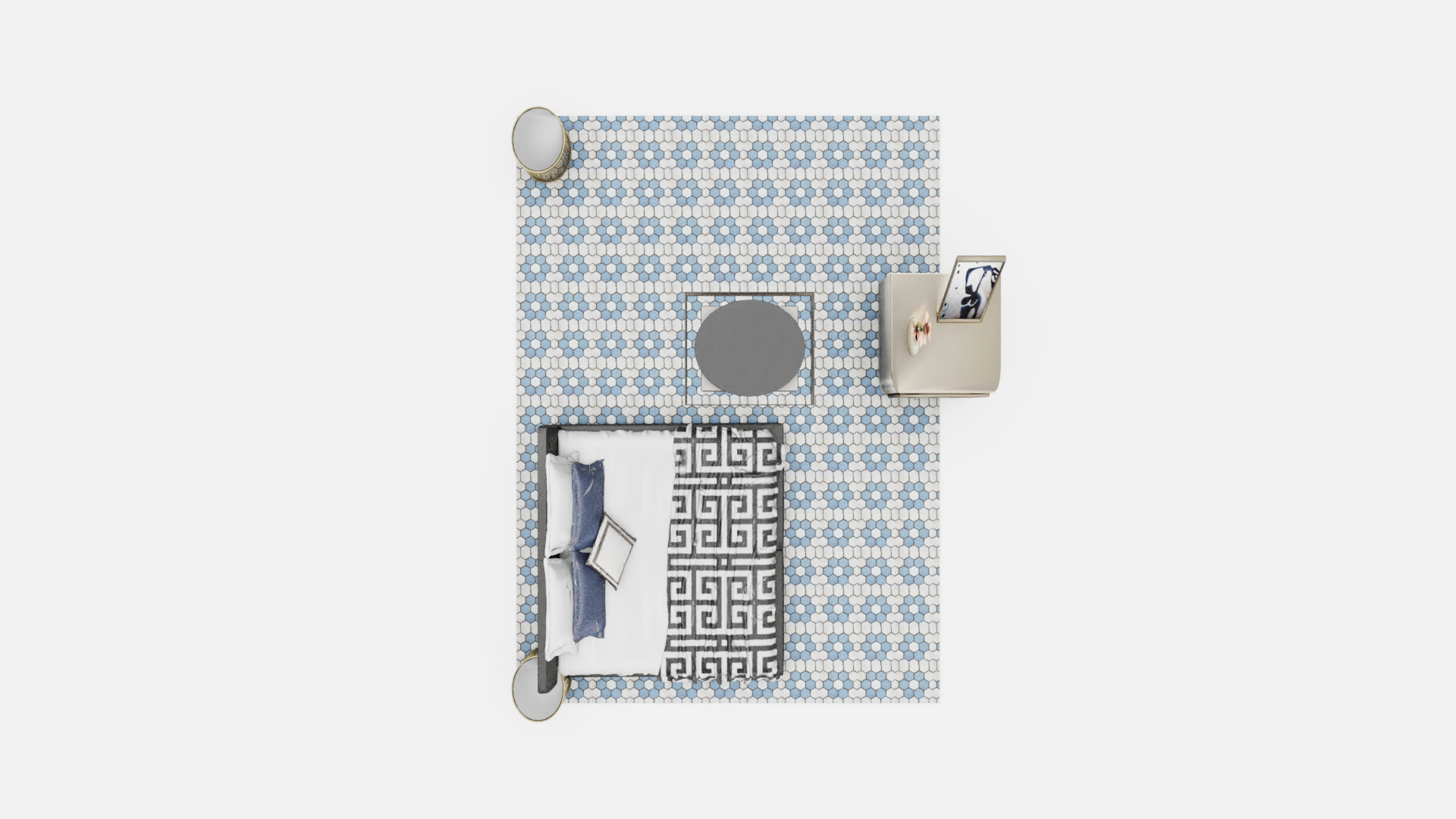}
    \end{minipage}%
        \begin{minipage}[b]{0.1\linewidth}
		\centering
		\includegraphics[width=\linewidth, trim=600 0 500 0, clip]{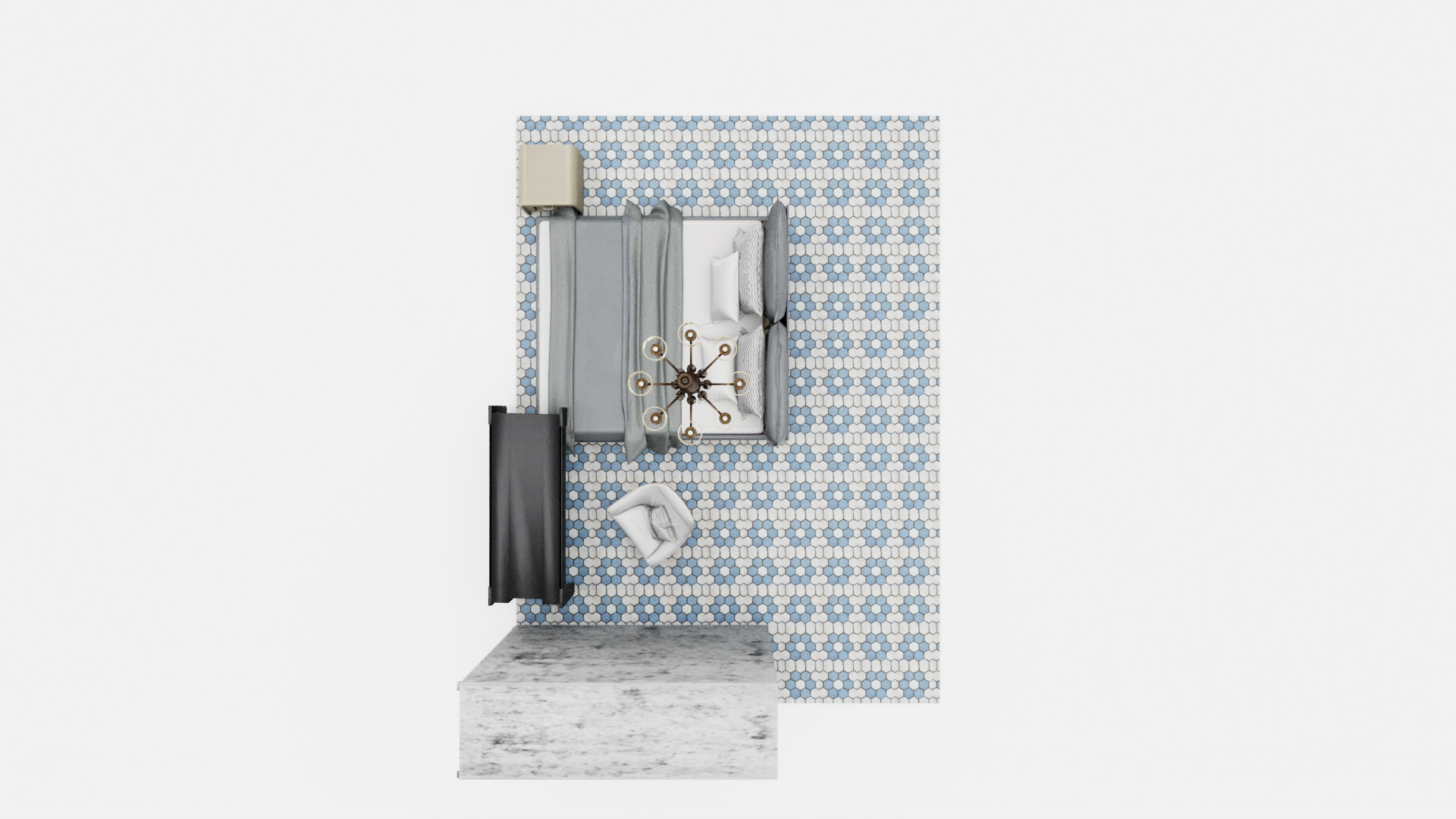}
    \end{minipage}%
    				        \begin{minipage}[b]{0.1\linewidth}
		\centering
		\includegraphics[width=\linewidth, trim=600 0 500 0, clip]{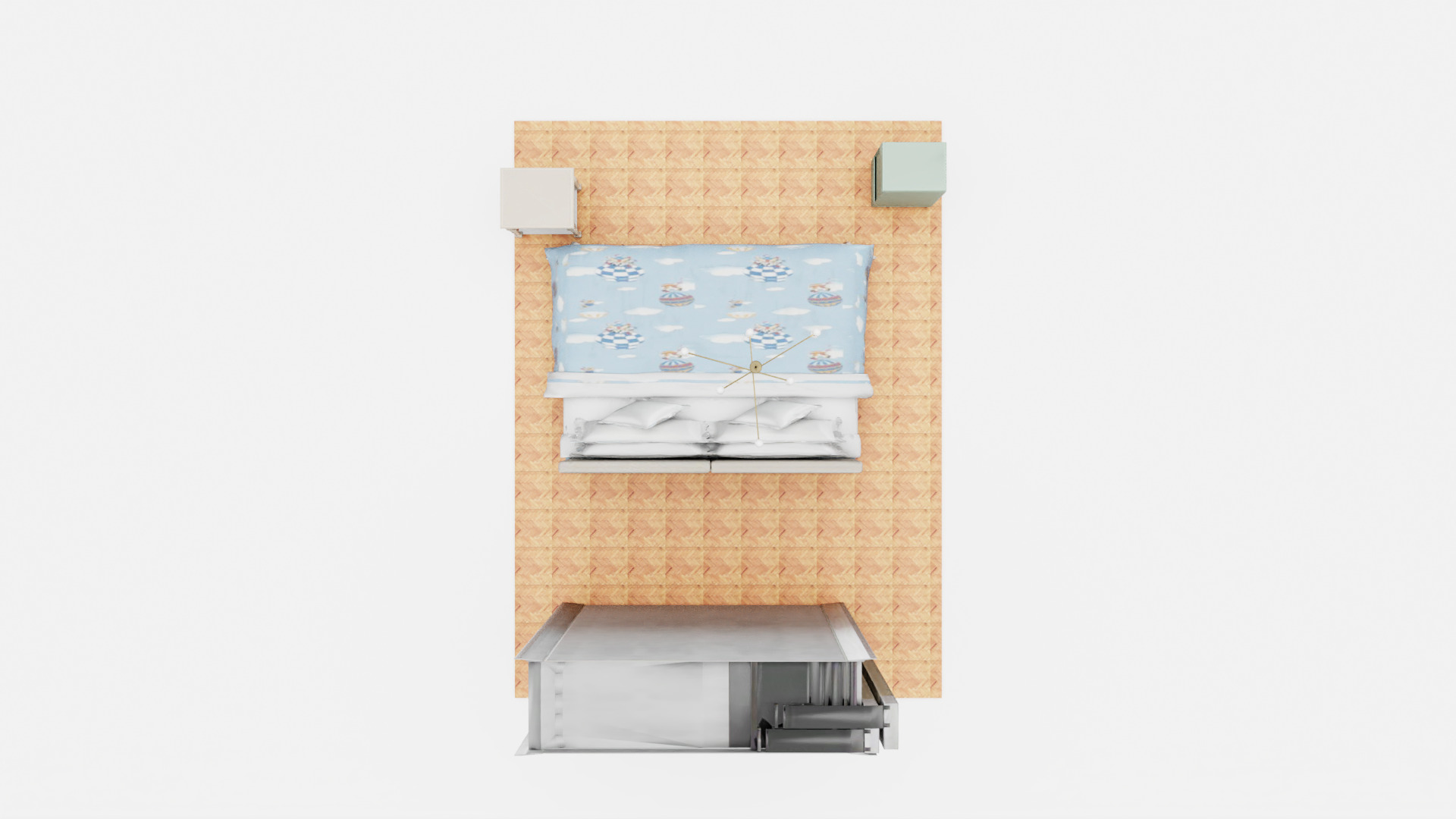}
    \end{minipage}%
        \begin{minipage}[b]{0.1\linewidth}
		\centering
		\includegraphics[width=\linewidth, trim=600 0 500 0, clip]{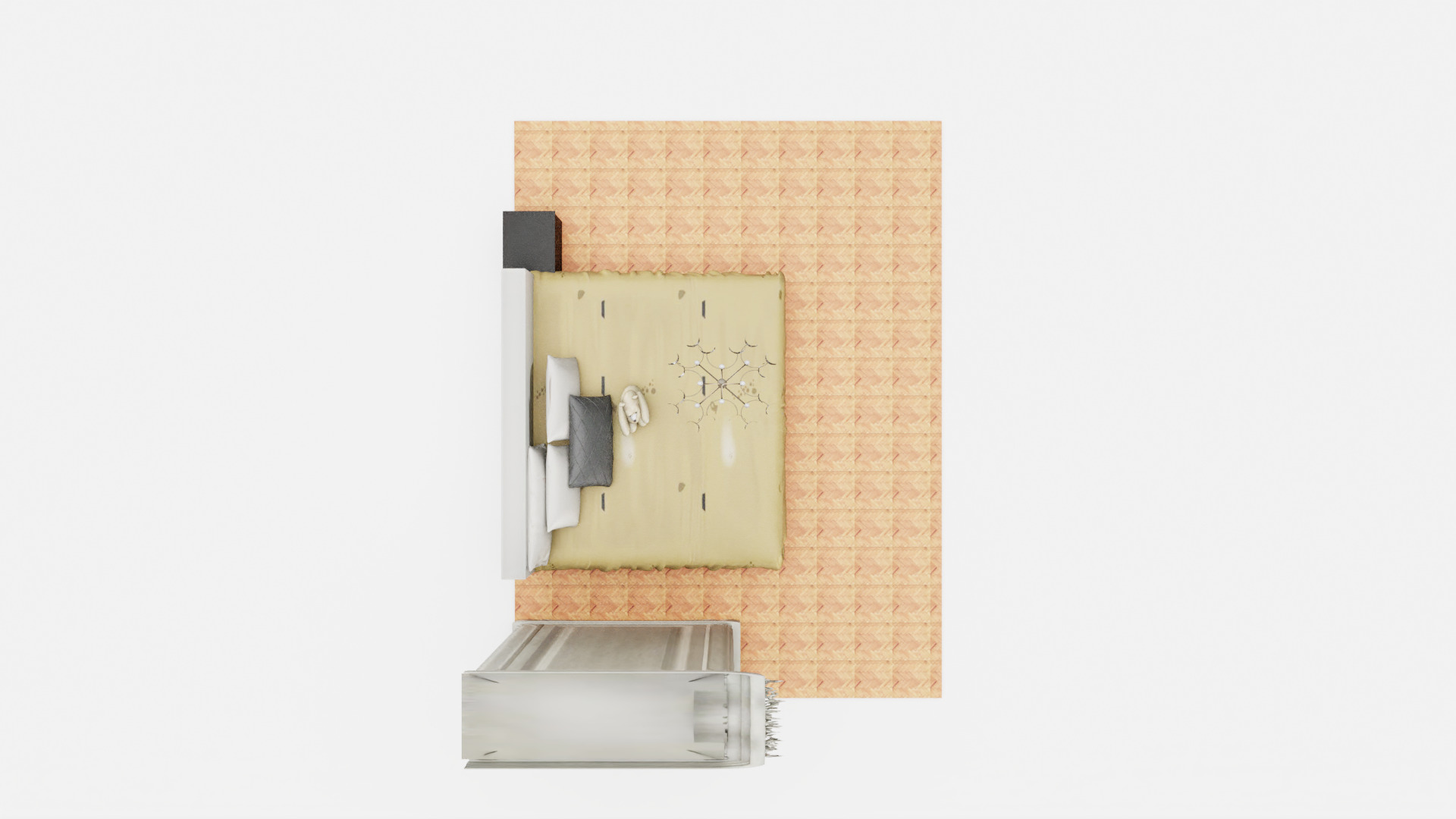}
    \end{minipage}%
        \begin{minipage}[b]{0.1\linewidth}
		\centering
		\includegraphics[width=\linewidth, trim=600 0 500 0, clip]{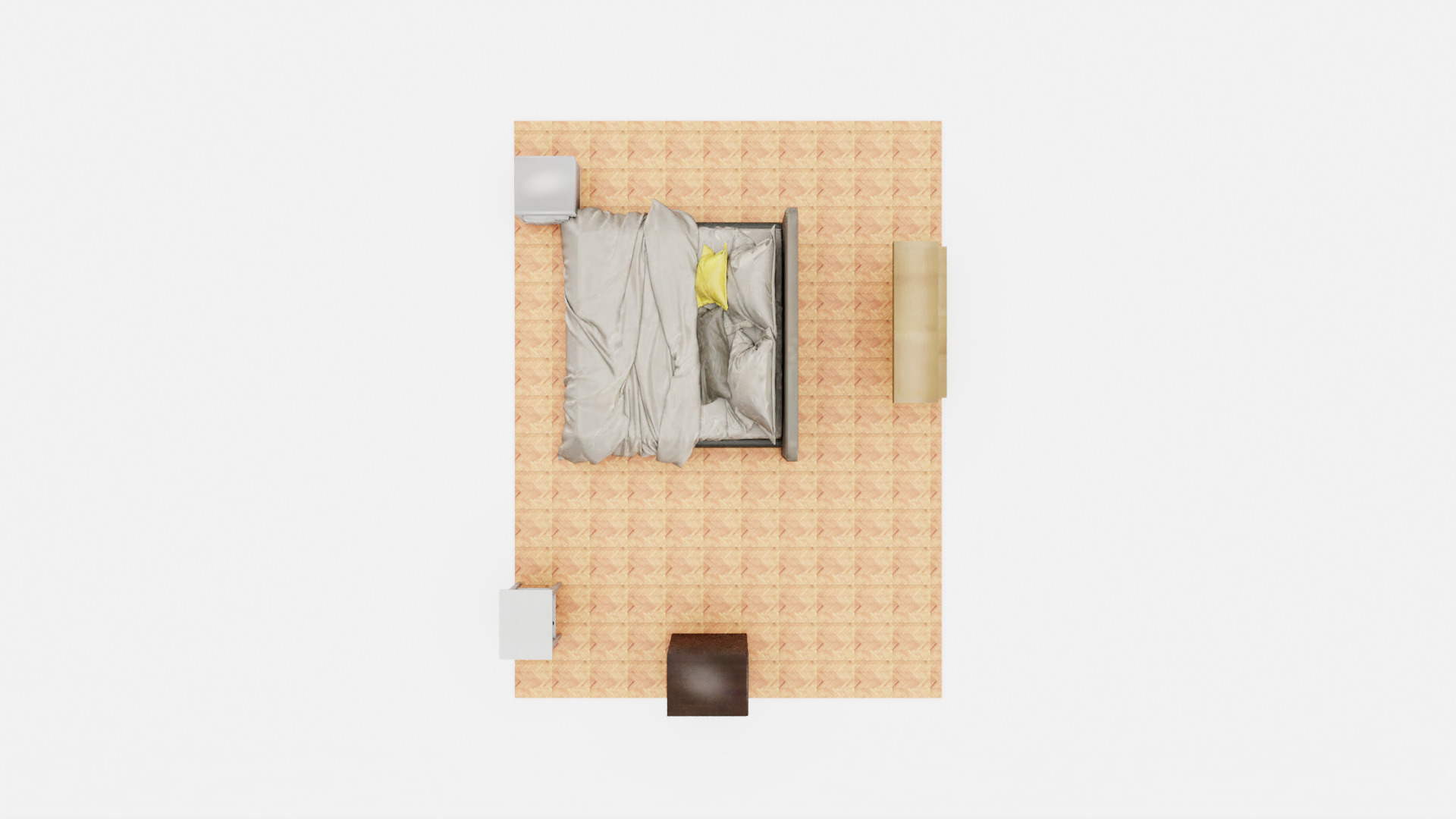}
    \end{minipage}%
    				        \begin{minipage}[b]{0.1\linewidth}
		\centering
		\includegraphics[width=\linewidth, trim=600 0 500 0, clip]{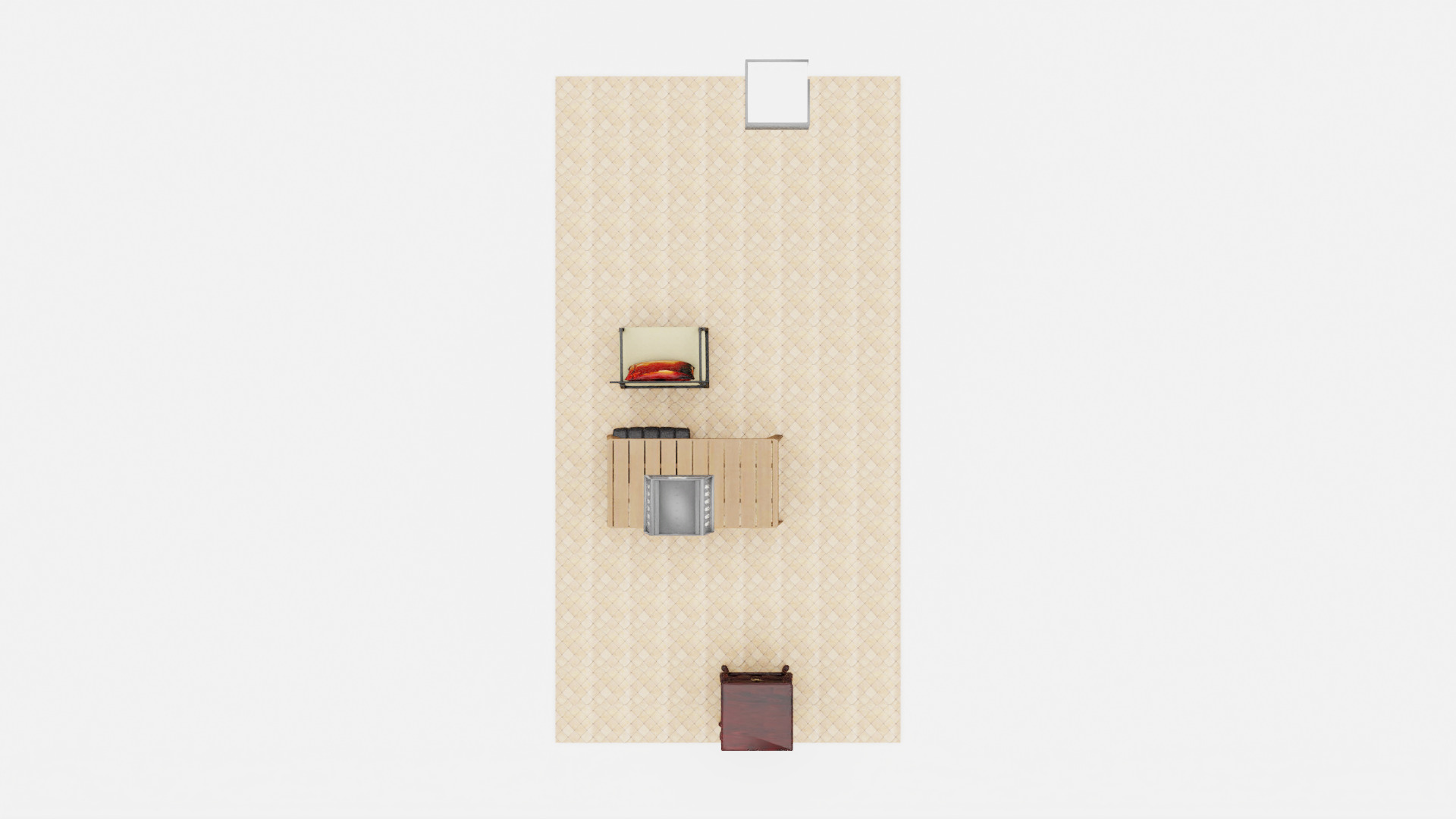}
    \end{minipage}%
        \begin{minipage}[b]{0.1\linewidth}
		\centering
		\includegraphics[width=\linewidth, trim=600 0 500 0, clip]{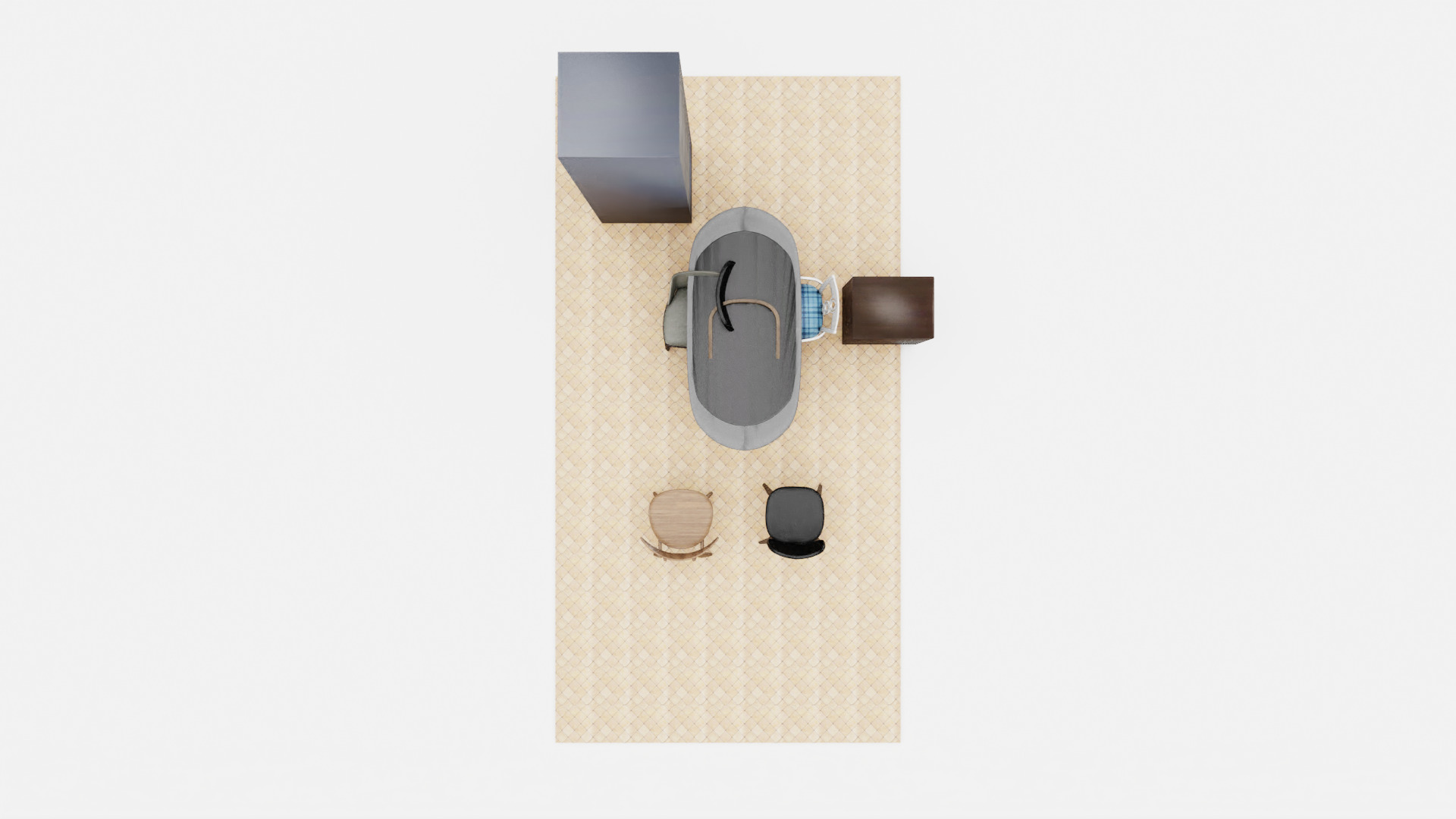}
    \end{minipage}%
        \begin{minipage}[b]{0.1\linewidth}
		\centering
		\includegraphics[width=\linewidth, trim=600 0 500 0, clip]{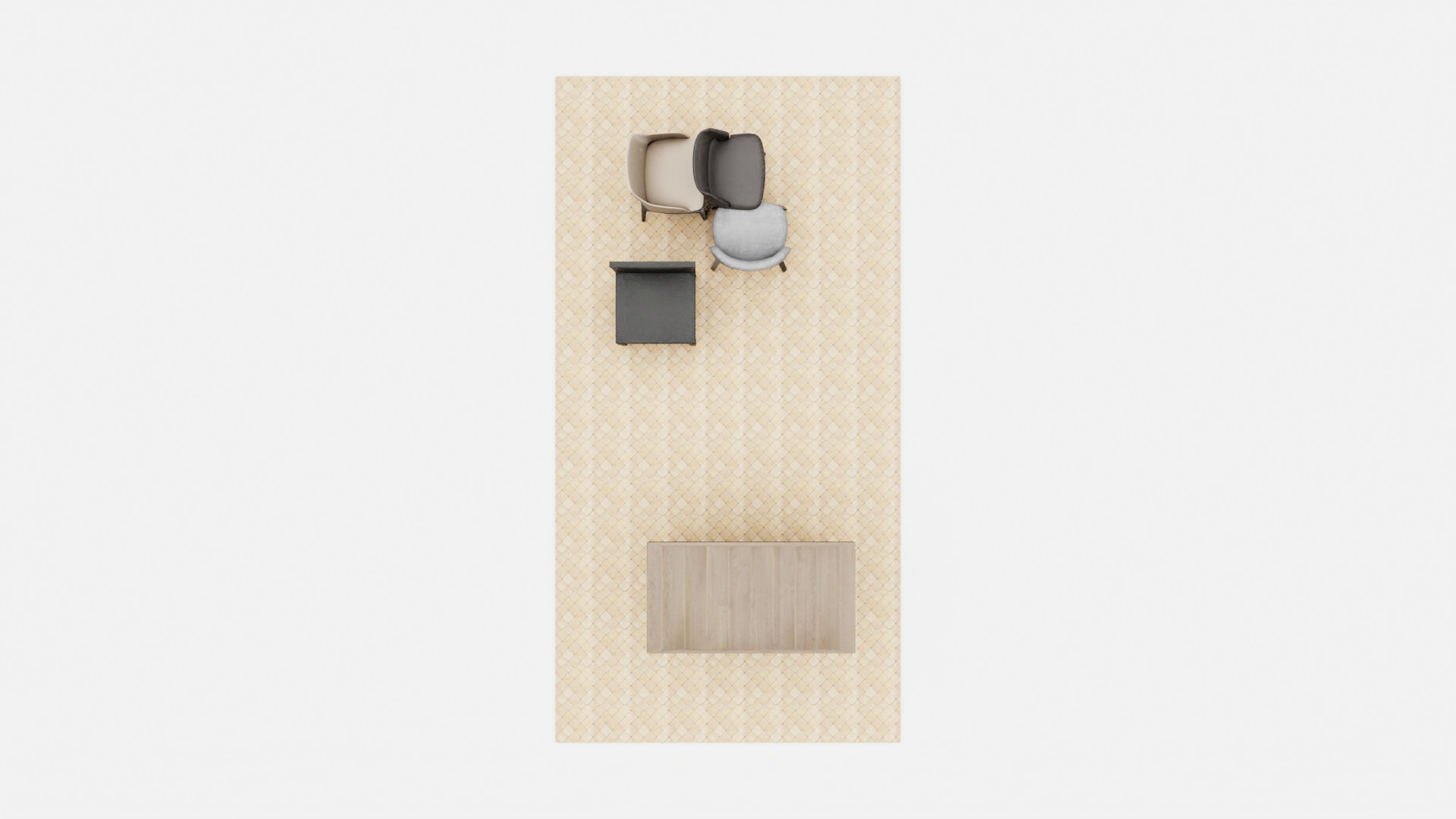}
    \end{minipage}%
    				        				            				            				        				        \hfill%
    \vspace{-1.2em}
    \vskip\baselineskip%
    		                \hfill%
    \begin{minipage}[b]{0.03\linewidth}
    \rotatebox{90}{\small{SceneFormer}}
    \end{minipage}
    \begin{minipage}[b]{0.1\linewidth}
		\centering
		\includegraphics[width=\linewidth, trim=600 0 500 0, clip]{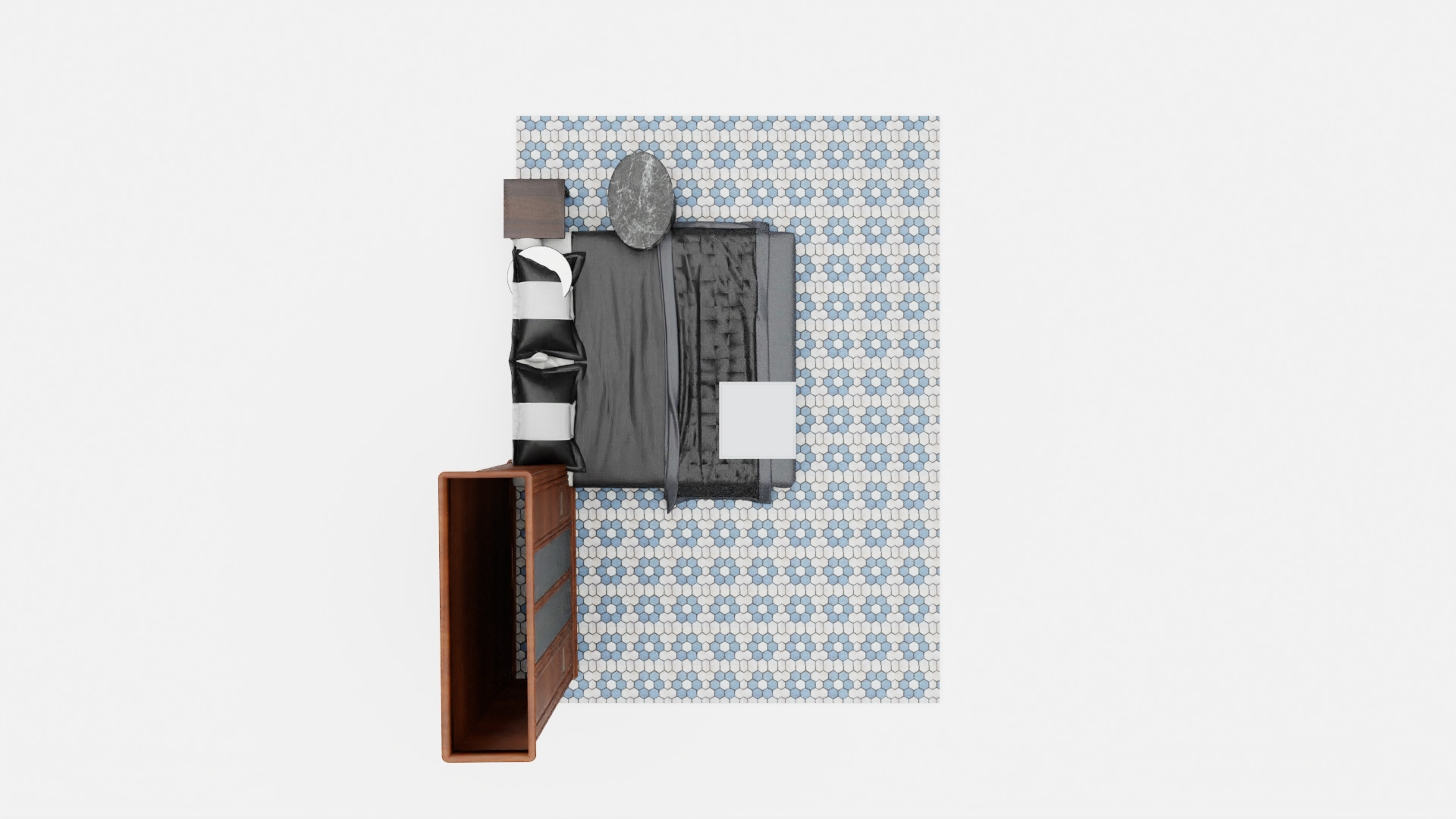}
    \end{minipage}%
        \begin{minipage}[b]{0.1\linewidth}
		\centering
		\includegraphics[width=\linewidth, trim=600 0 500 0, clip]{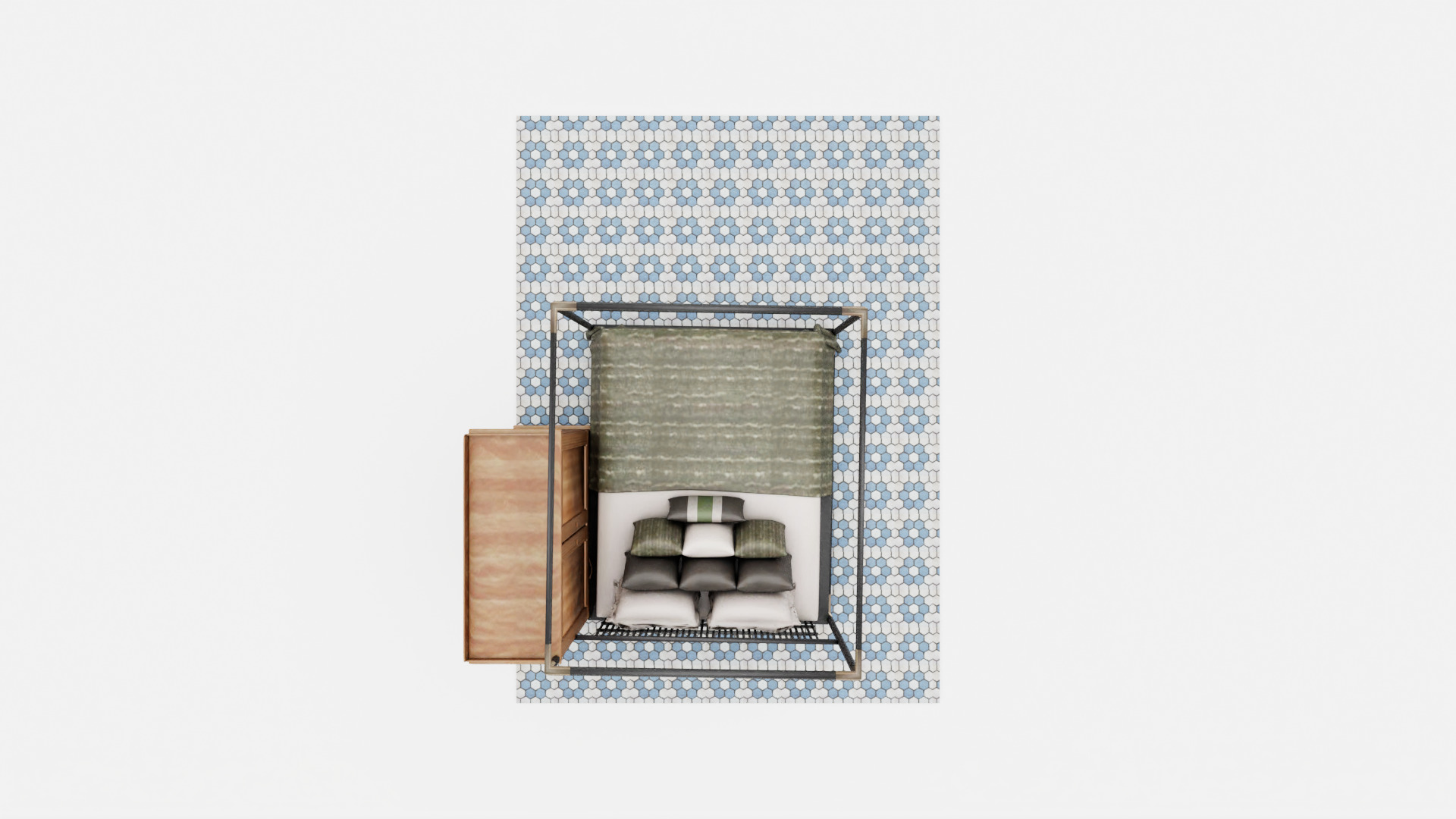}
    \end{minipage}%
        \begin{minipage}[b]{0.1\linewidth}
		\centering
		\includegraphics[width=\linewidth, trim=600 0 500 0, clip]{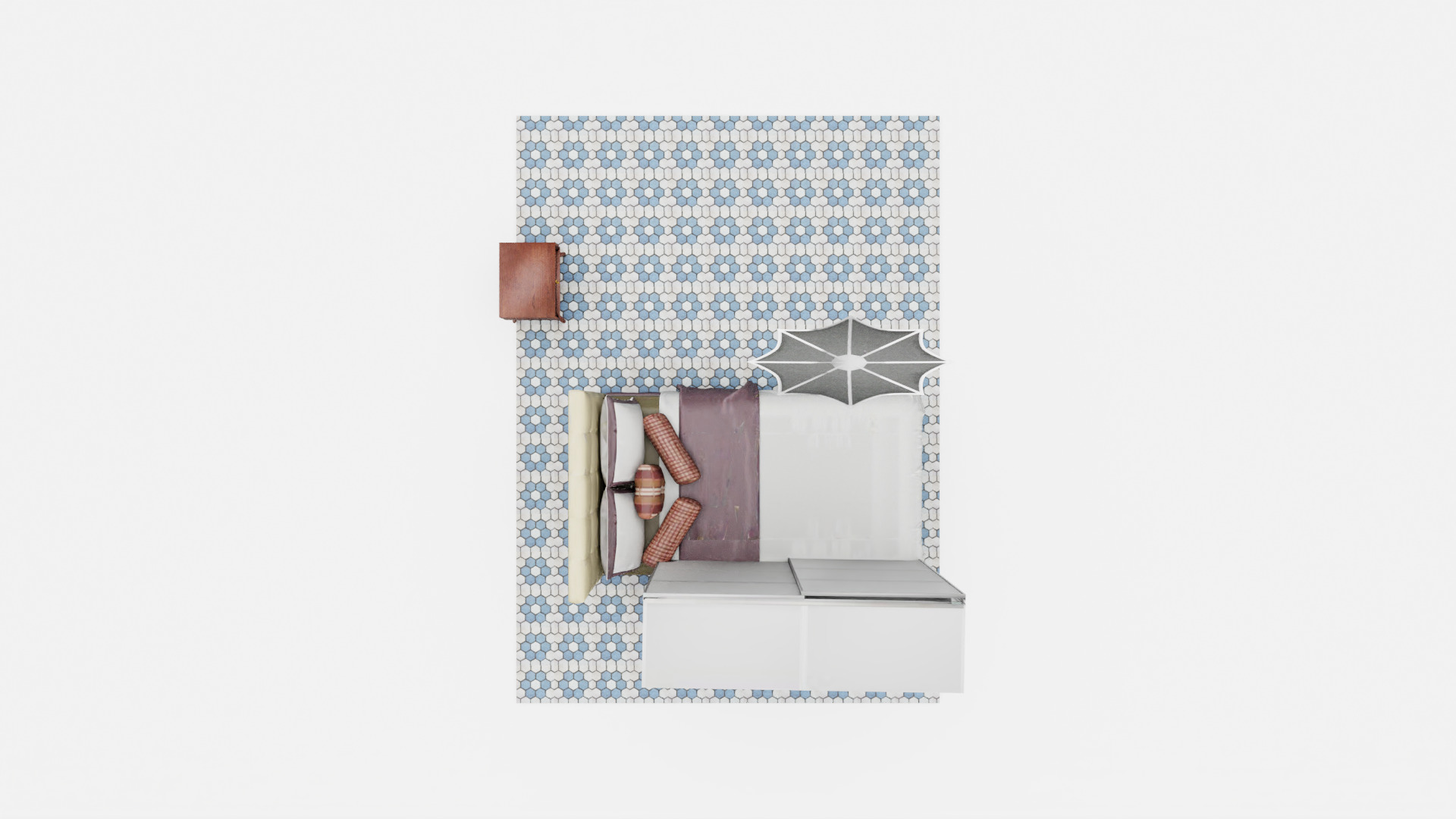}
    \end{minipage}%
    				        \begin{minipage}[b]{0.1\linewidth}
		\centering
		\includegraphics[width=\linewidth, trim=600 0 500 0, clip]{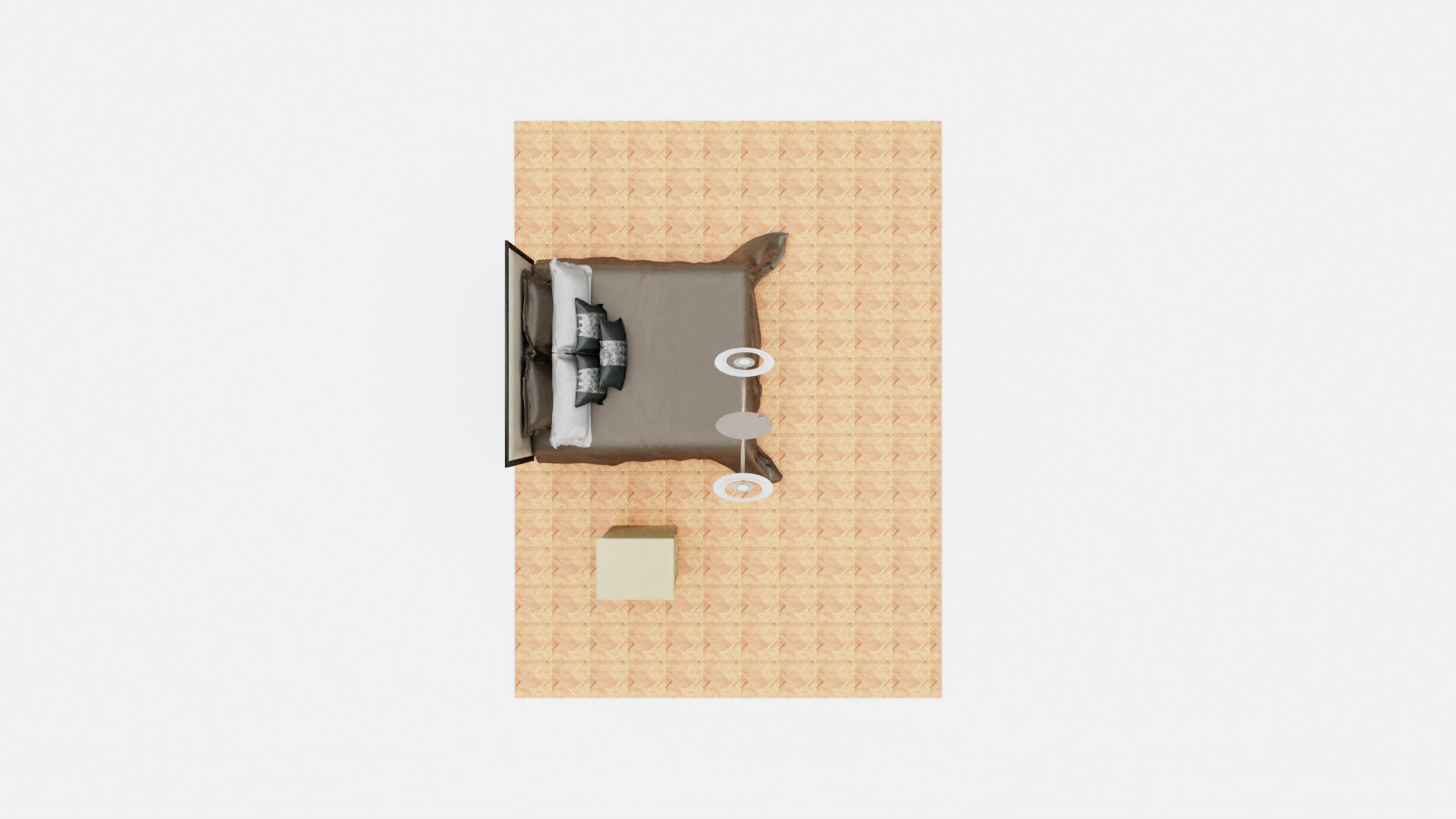}
    \end{minipage}%
        \begin{minipage}[b]{0.1\linewidth}
		\centering
		\includegraphics[width=\linewidth, trim=600 0 500 0, clip]{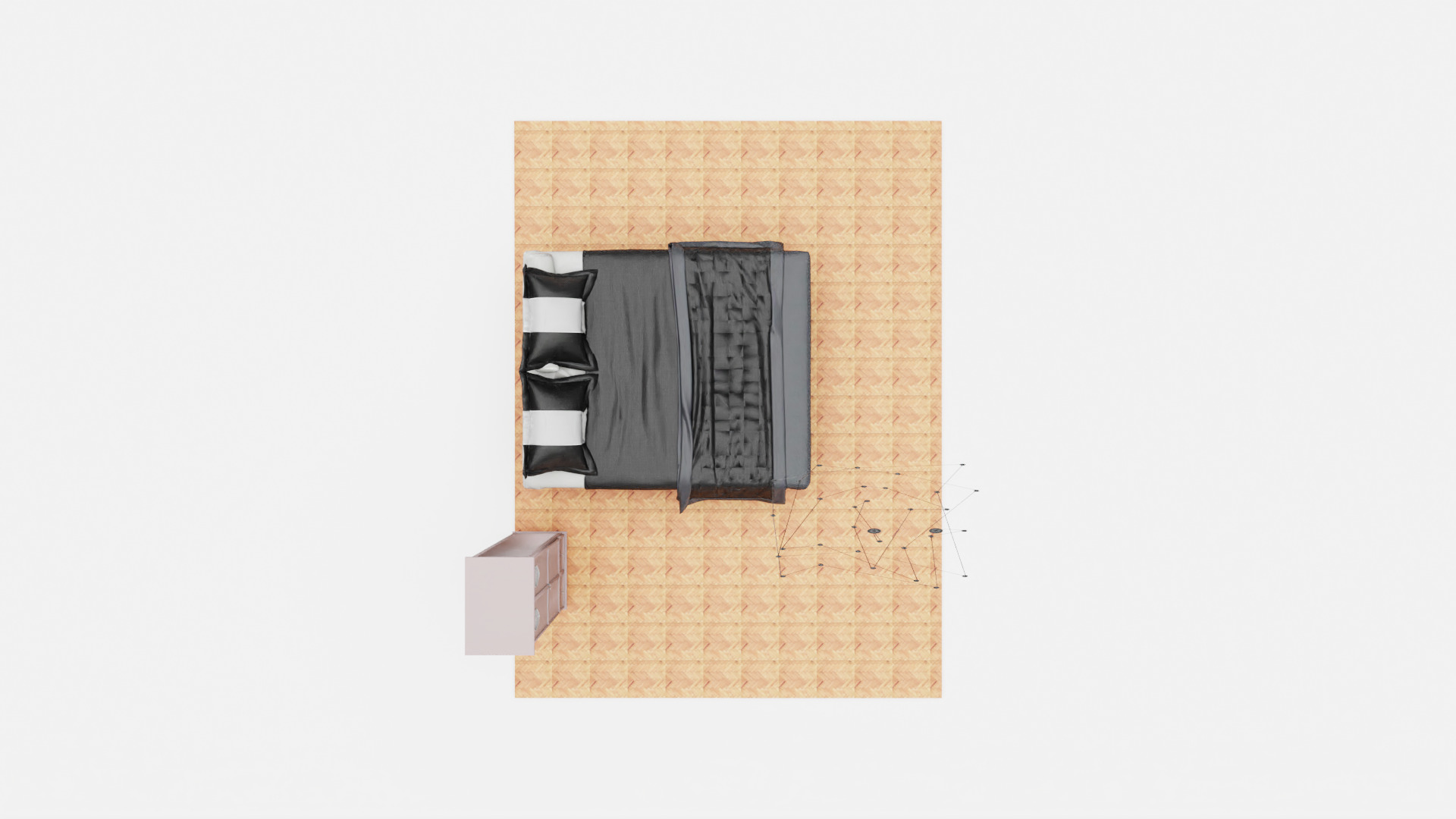}
    \end{minipage}%
        \begin{minipage}[b]{0.1\linewidth}
		\centering
		\includegraphics[width=\linewidth, trim=600 0 500 0, clip]{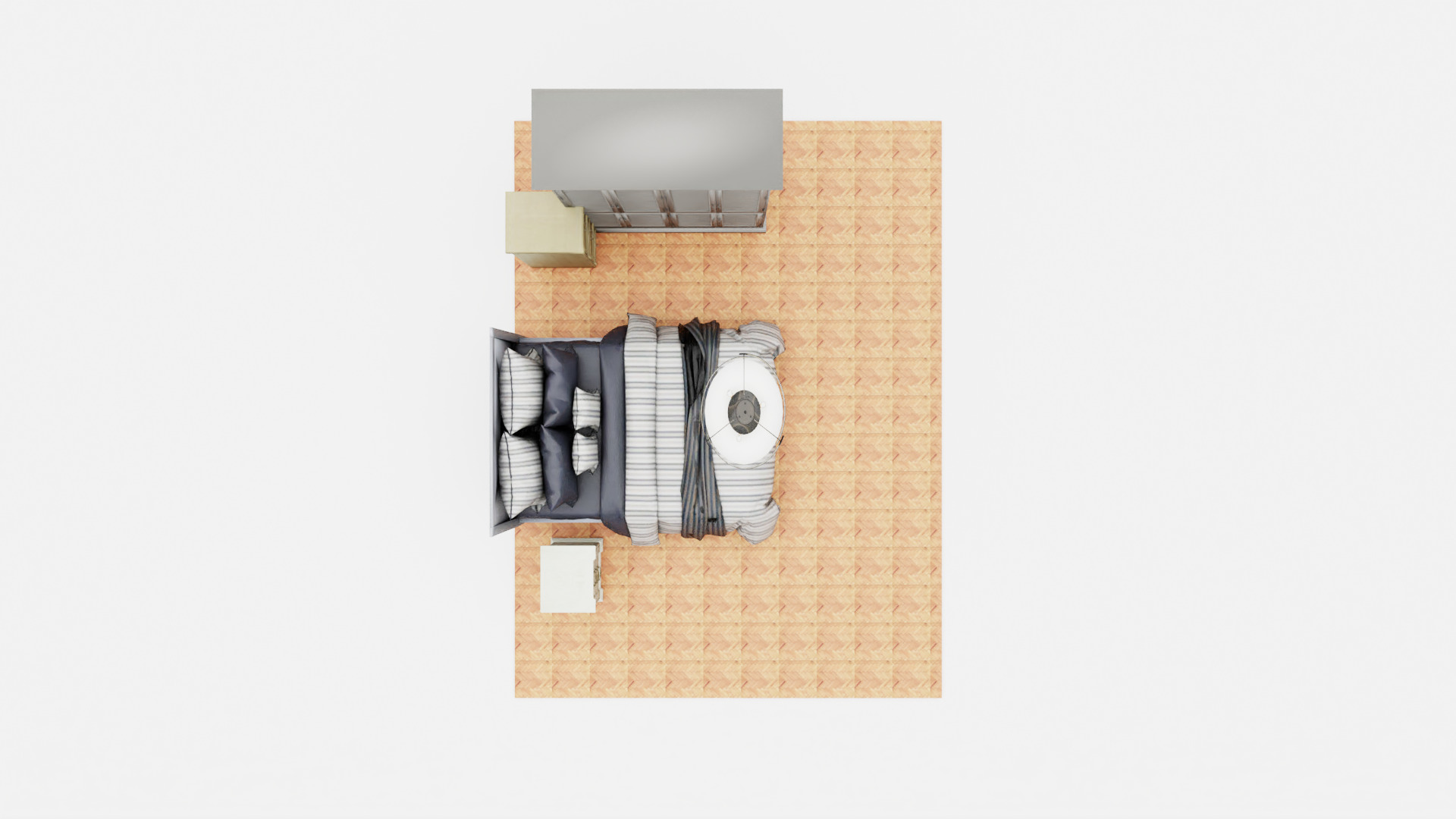}
    \end{minipage}%
    				                \begin{minipage}[b]{0.1\linewidth}
		\centering
		\includegraphics[width=\linewidth, trim=600 0 500 0, clip]{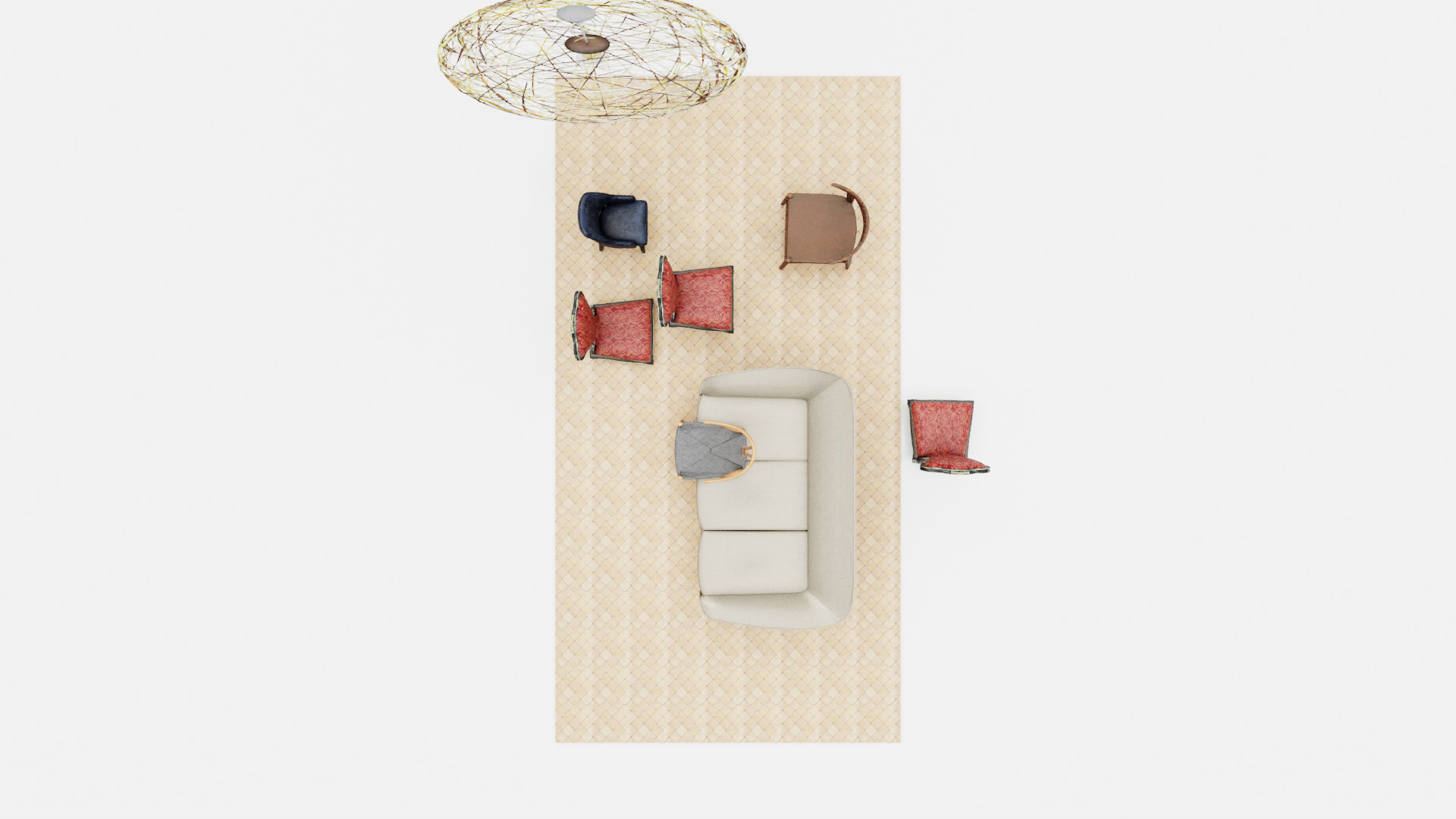}
    \end{minipage}%
        \begin{minipage}[b]{0.1\linewidth}
		\centering
		\includegraphics[width=\linewidth, trim=600 0 500 0, clip]{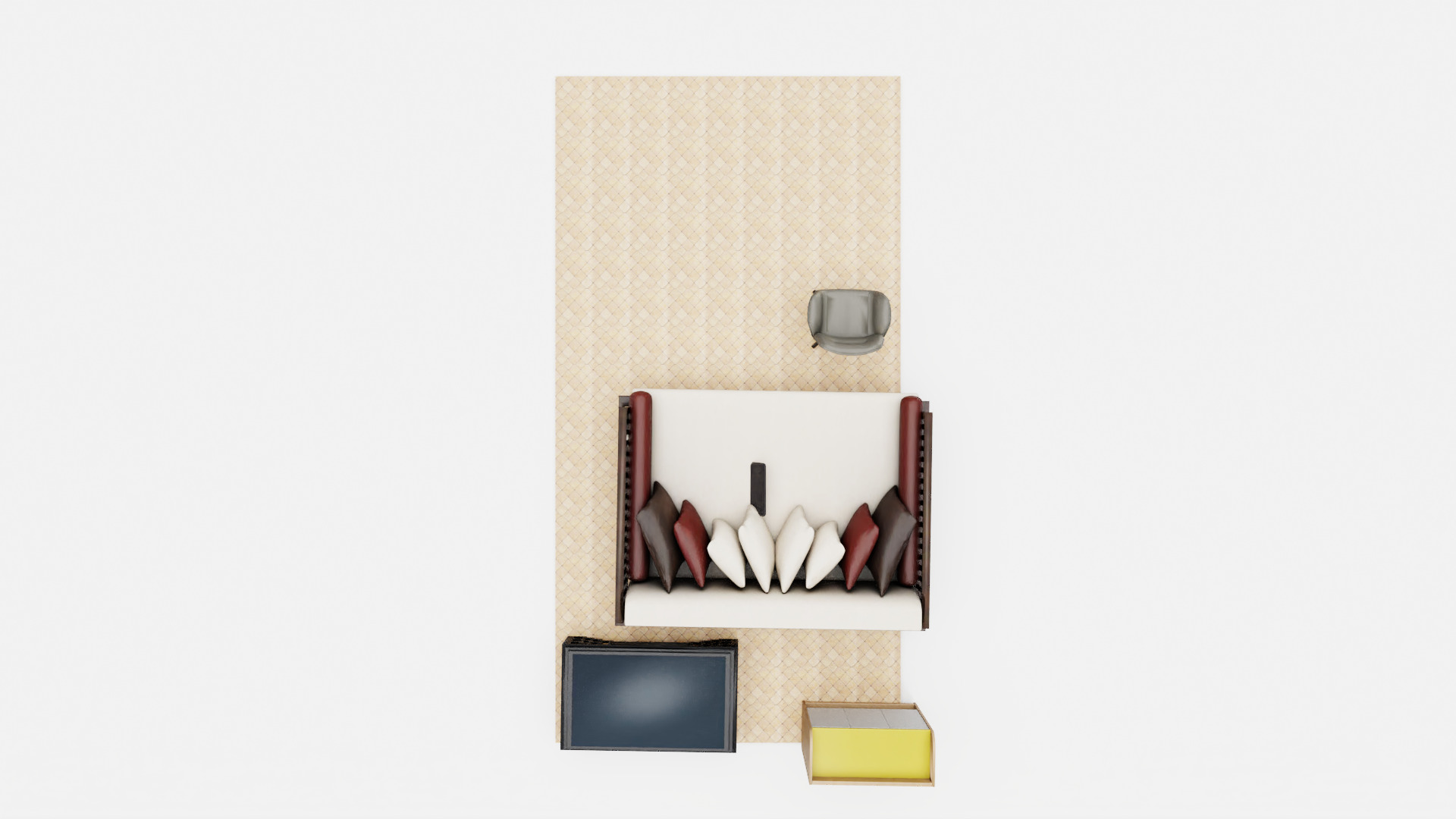}
    \end{minipage}%
        \begin{minipage}[b]{0.1\linewidth}
		\centering
		\includegraphics[width=\linewidth, trim=600 0 500 0, clip]{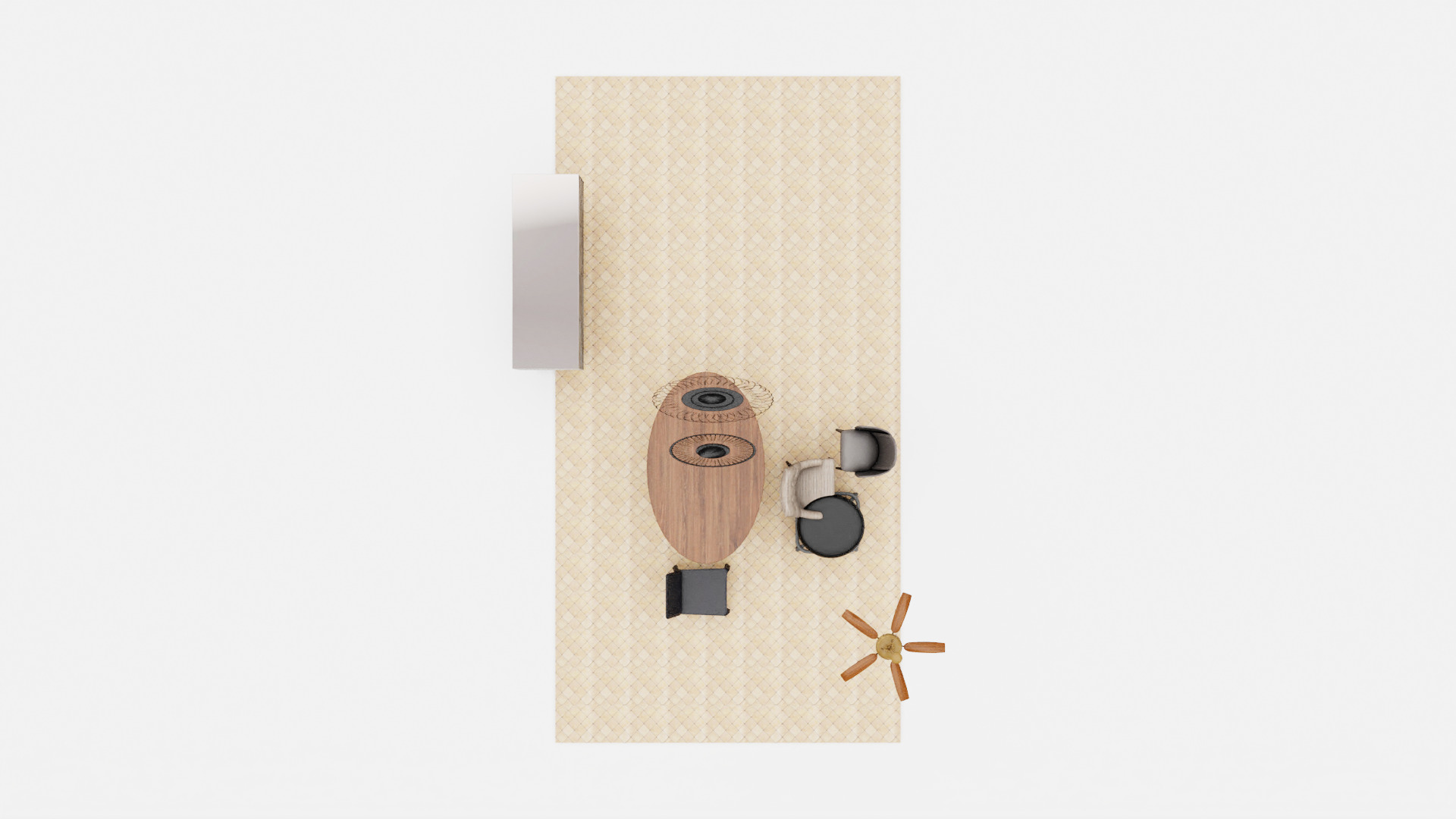}
    \end{minipage}%
    				        				            				            				        				        \hfill%
    \vspace{-1.2em}
    \vskip\baselineskip%
    		                \hfill%
    \begin{minipage}[b]{0.03\linewidth}
    \rotatebox[origin=lt]{90}{~~ \small{Ours}}
    \end{minipage}
    \begin{minipage}[b]{0.1\linewidth}
		\centering
		\includegraphics[width=\linewidth, trim=600 0 500 0, clip]{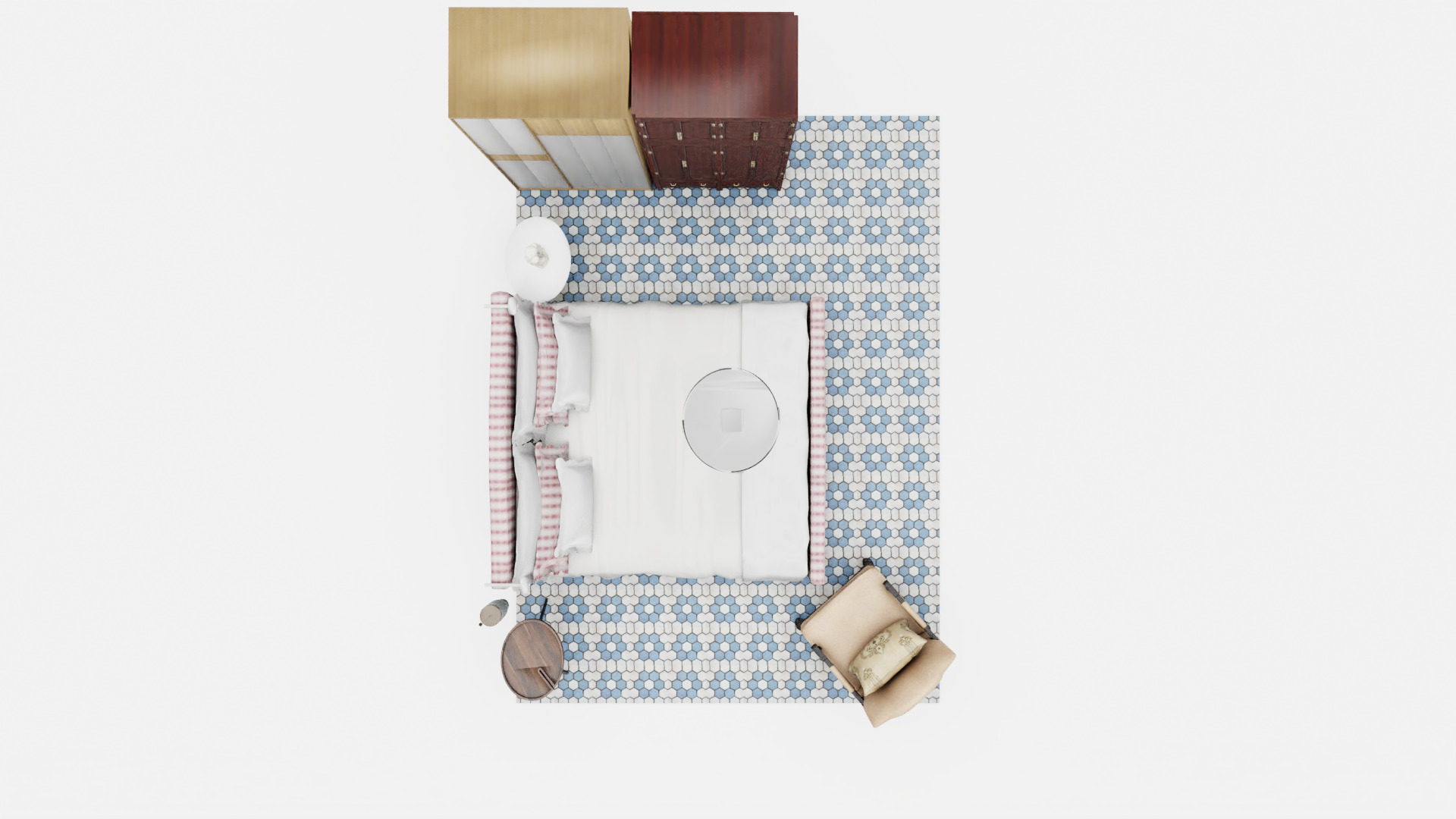}
    \end{minipage}%
        \begin{minipage}[b]{0.1\linewidth}
		\centering
		\includegraphics[width=\linewidth, trim=600 0 500 0, clip]{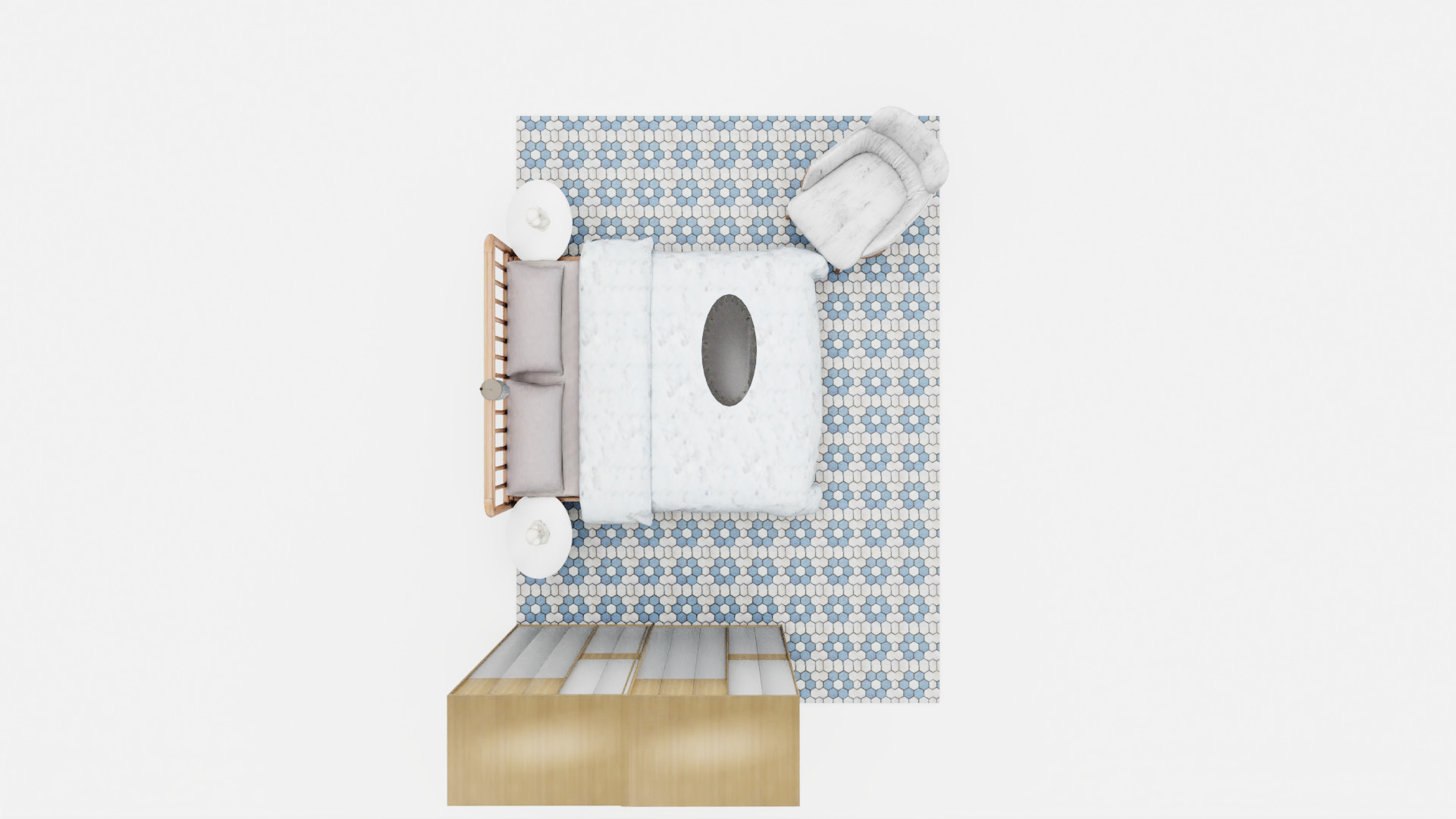}
    \end{minipage}%
        \begin{minipage}[b]{0.1\linewidth}
		\centering
		\includegraphics[width=\linewidth, trim=600 0 500 0, clip]{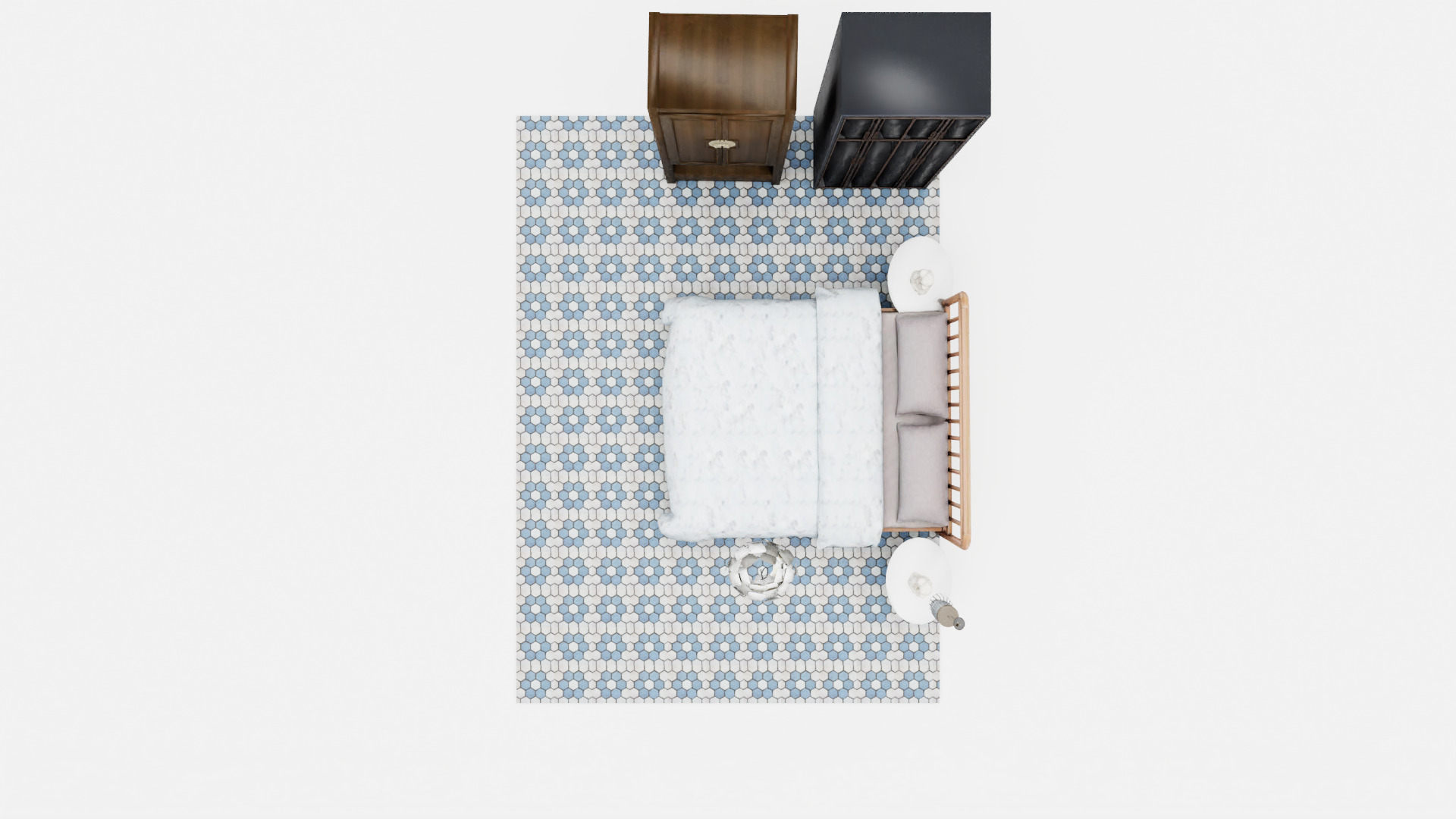}
    \end{minipage}%
    				        \begin{minipage}[b]{0.1\linewidth}
		\centering
		\includegraphics[width=\linewidth, trim=600 0 500 0, clip]{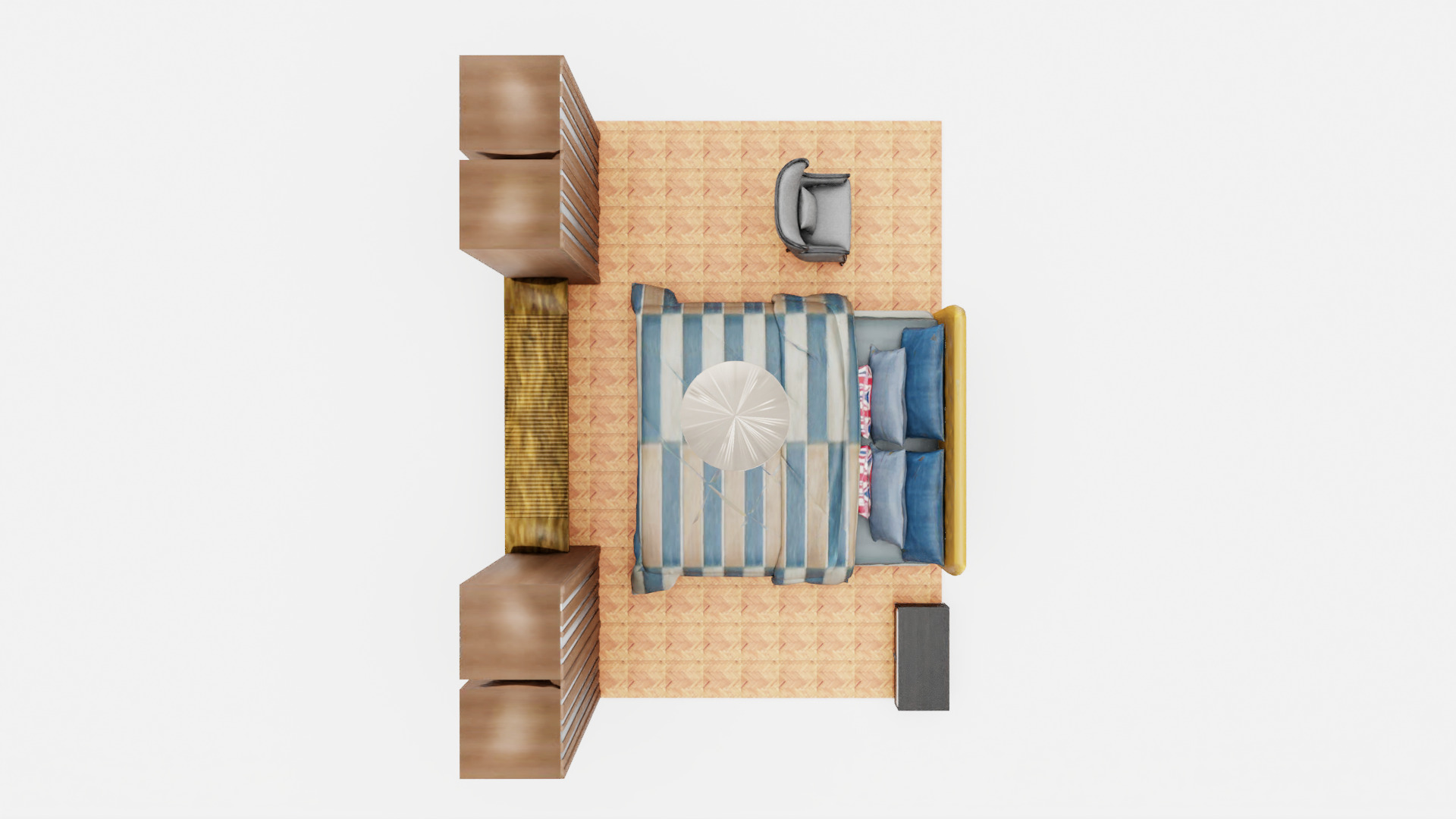}
    \end{minipage}%
        \begin{minipage}[b]{0.1\linewidth}
		\centering
		\includegraphics[width=\linewidth, trim=600 0 500 0, clip]{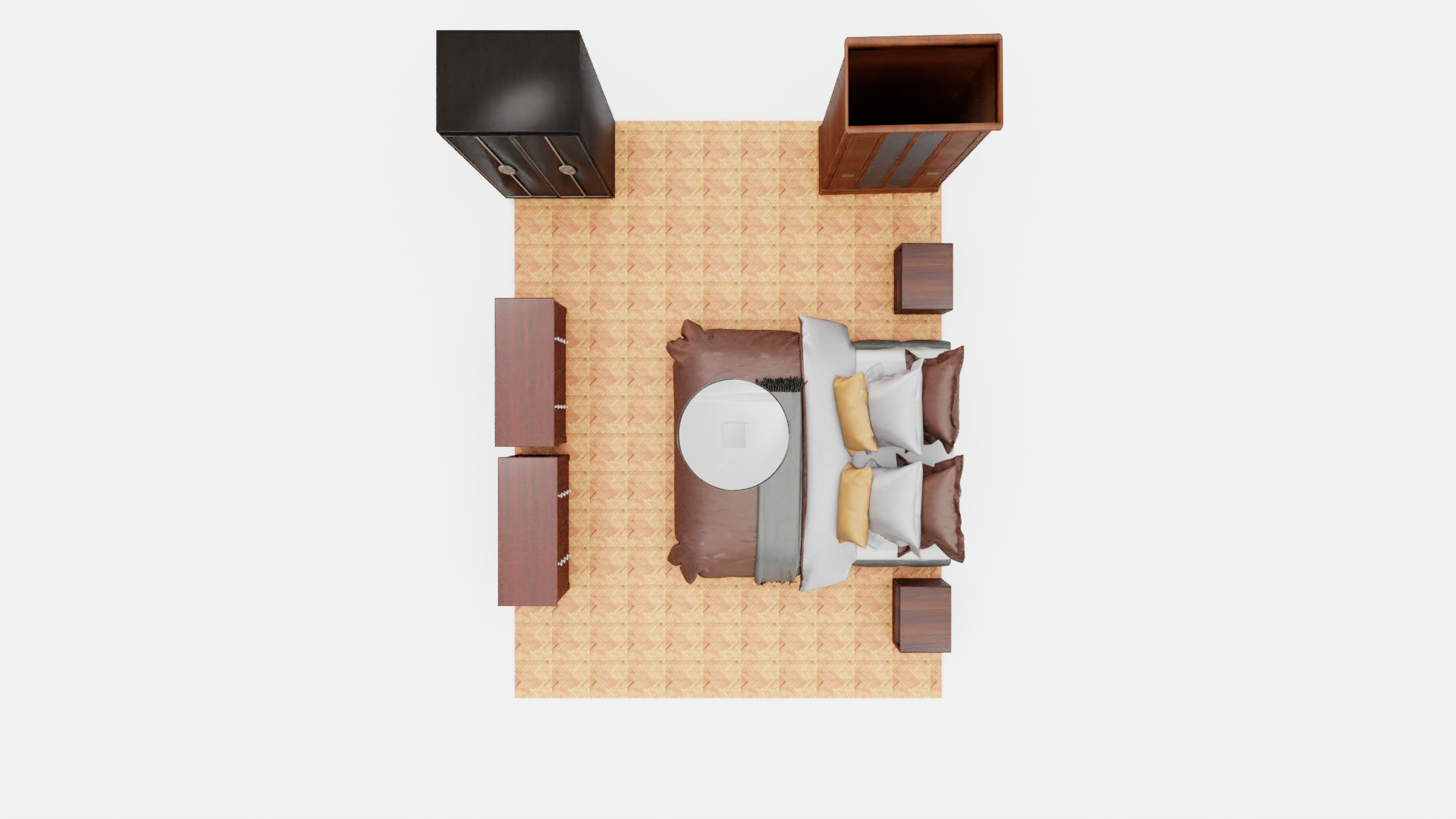}
    \end{minipage}%
        \begin{minipage}[b]{0.1\linewidth}
		\centering
		\includegraphics[width=\linewidth, trim=600 0 500 0, clip]{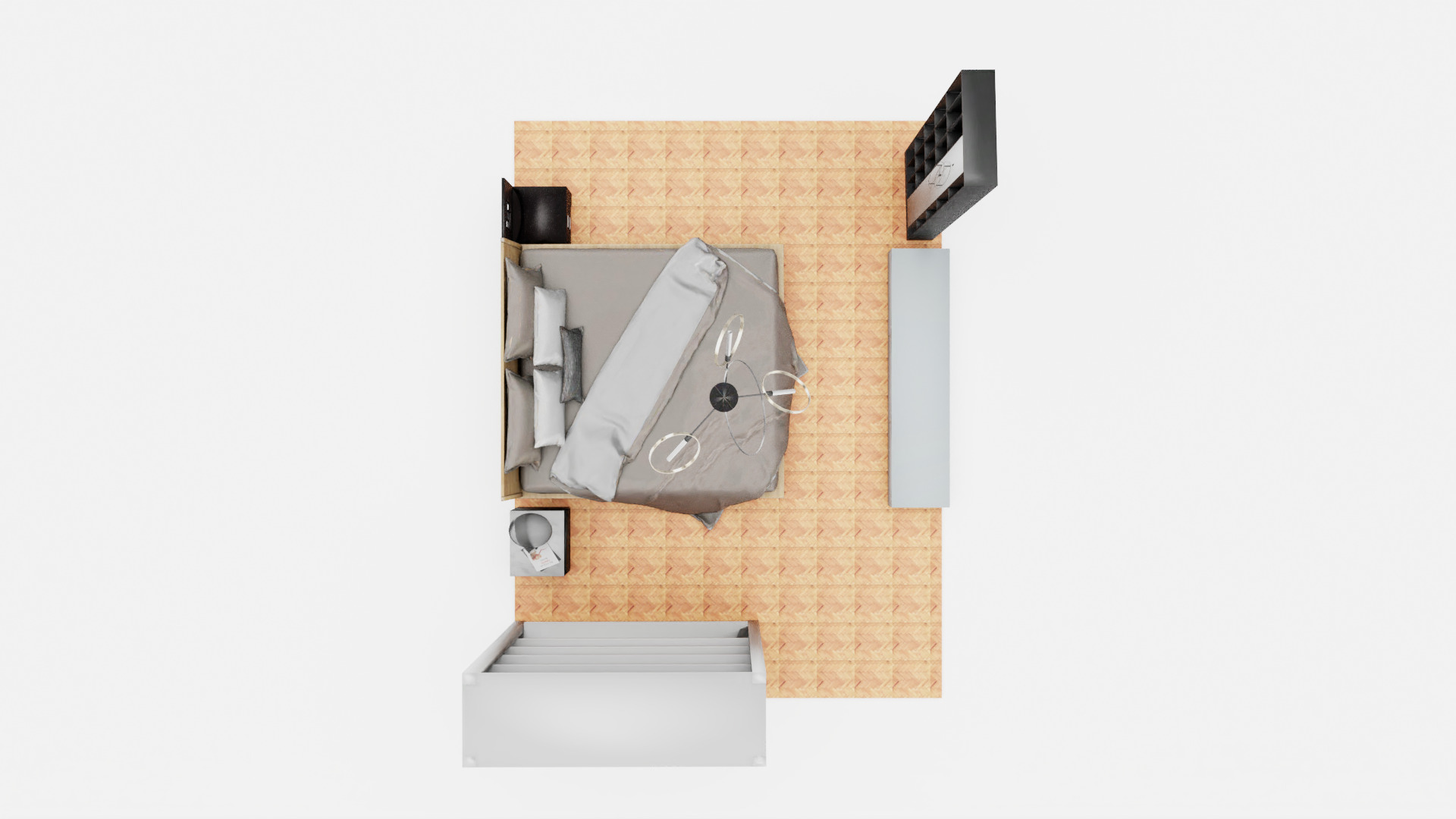}
    \end{minipage}%
    				                \begin{minipage}[b]{0.1\linewidth}
		\centering
		\includegraphics[width=\linewidth, trim=600 0 500 0, clip]{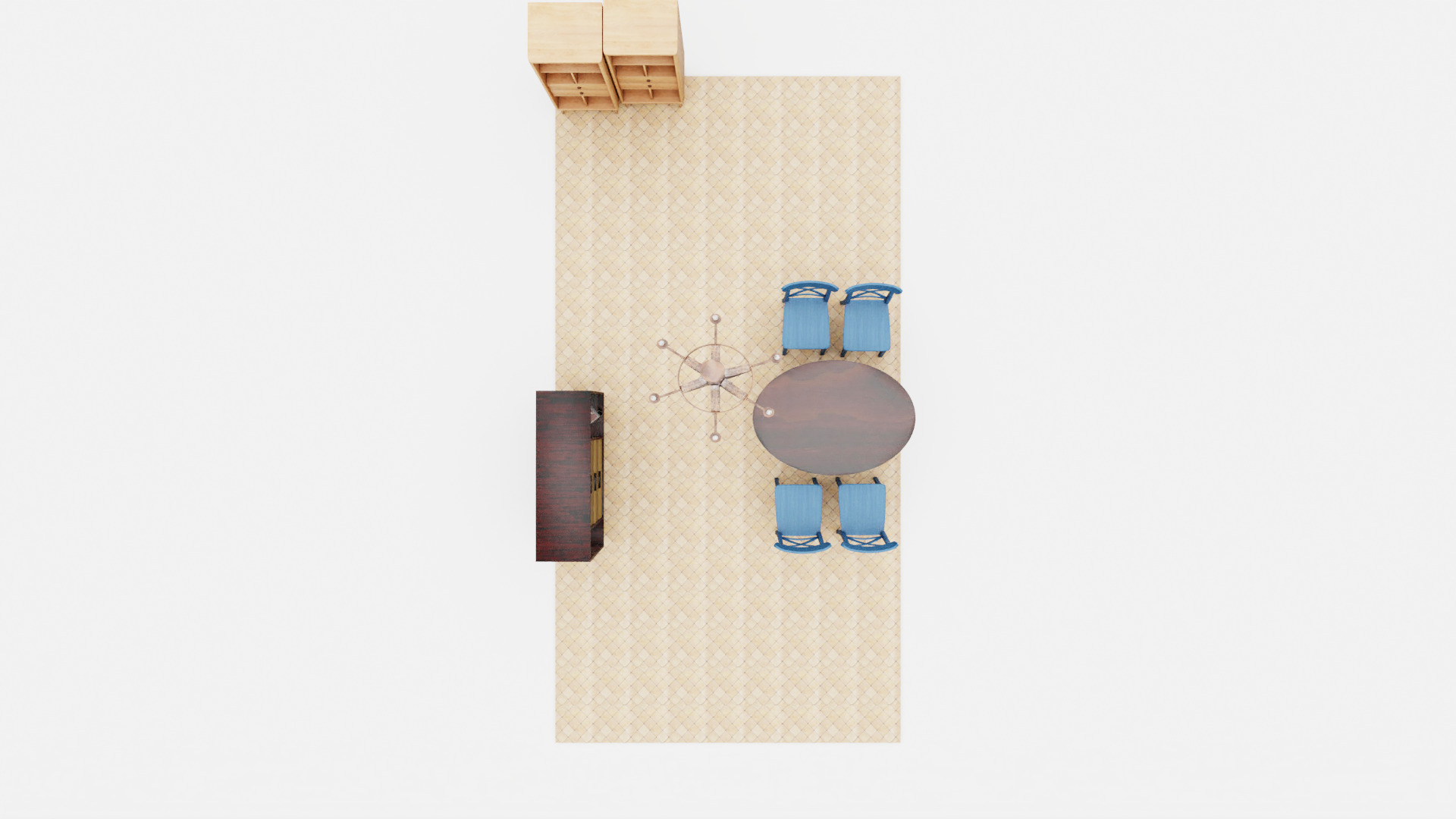}
    \end{minipage}%
        \begin{minipage}[b]{0.1\linewidth}
		\centering
		\includegraphics[width=\linewidth, trim=600 0 500 0, clip]{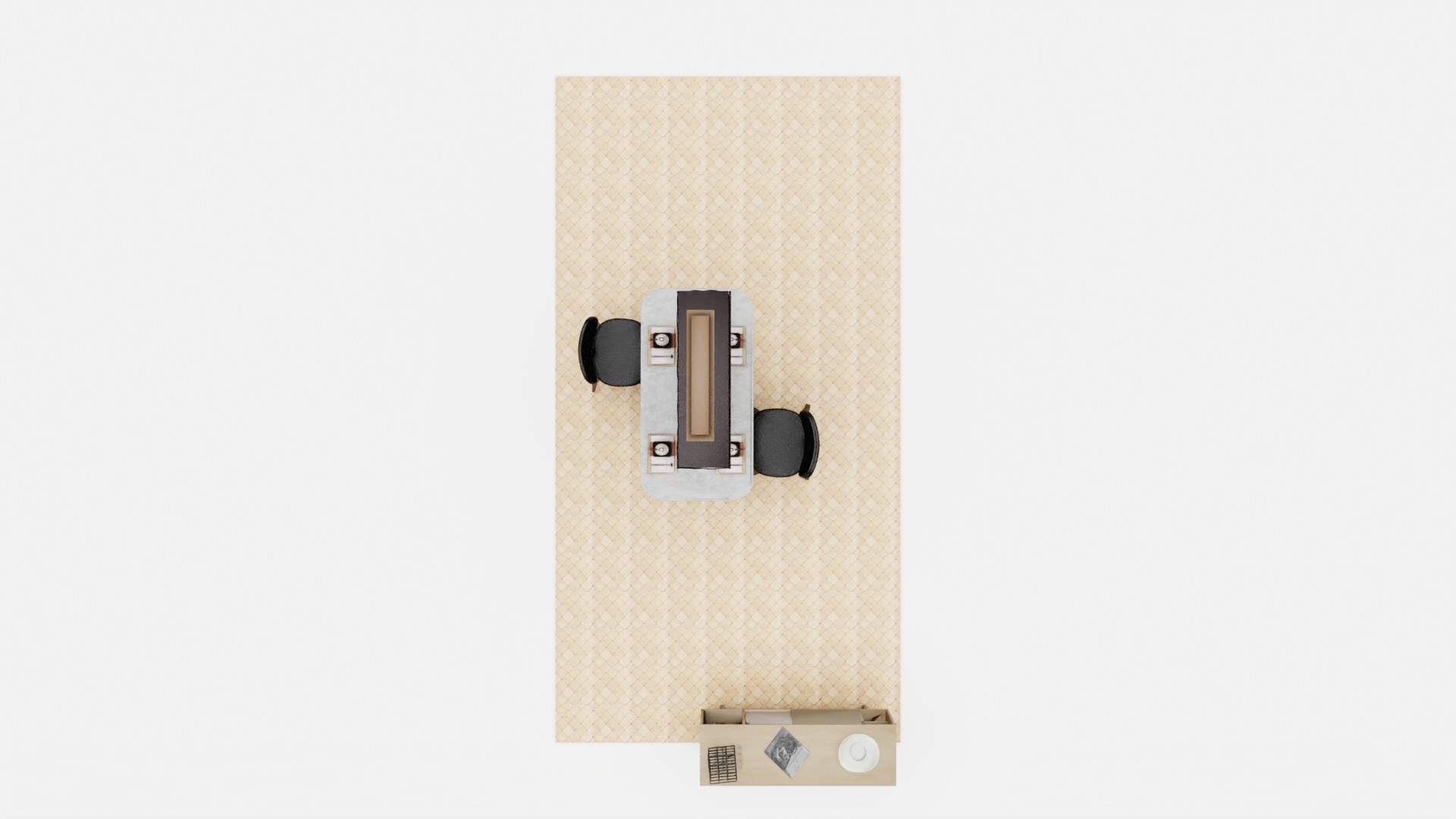}
    \end{minipage}%
        				        \begin{minipage}[b]{0.1\linewidth}
		\centering
		\includegraphics[width=\linewidth, trim=600 0 500 0, clip]{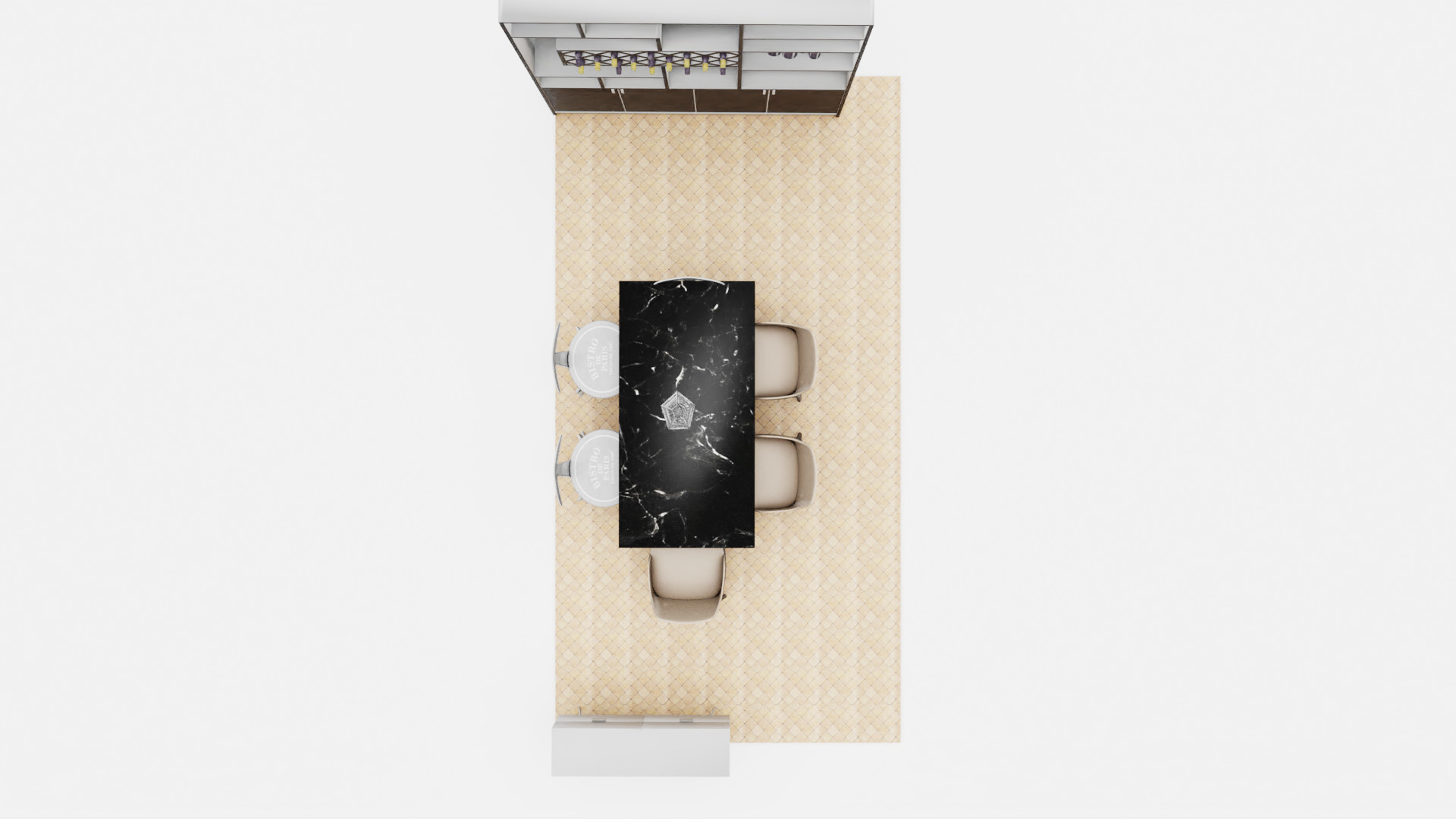}
    \end{minipage}%
    \hfill%
    				            				            				        				        \vspace{-1.4em}
    \vskip\baselineskip%
    		                    \caption{\small{\bf Scene Diversity}. We show three generated scenes conditioned on
    three different floor plans for bedrooms and dining rooms. Every triplet of
    columns corresponds to a different floor plan.}
    \label{fig:scene_synthesis_diversity}
    \vspace{-0.8em}
\end{figure}

\begin{table}
    \centering
    \resizebox{\columnwidth}{!}{
    \begin{tabular}{l|cccc|cccc|cccc}
        \toprule
        \multicolumn{1}{c}{\,} & \multicolumn{4}{c}{FID Score ($\downarrow$)}& \multicolumn{4}{c}{Scene Classification Accuracy}& \multicolumn{4}{c}{Category KL Divergence ($\downarrow$)}\\
        \toprule
        \multicolumn{1}{c}{\,} & FastSynth & SceneFormer & Ours+Order & Ours & FastSynth & SceneFormer & Ours+Order & Ours & FastSynth& SceneFormer & Ours+Order & Ours \\
        \midrule
        Bedrooms & 40.89 & 43.17 & 38.67 & \bf{38.39} & 0.883 & 0.945 & 0.760 & \bf{0.562} & 0.0064 & \bf{0.0052} & 0.0533 & 0.0085\\
        Living   & 61.67 & 69.54 & 35.37 & \bf{33.14} & 0.945 & 0.972 & 0.694 & \bf{0.516} & 0.0176 & 0.0313 & 0.0372 & \bf{0.0034}\\
        Dining   & 55.83 & 67.04 & 35.79 & \bf{29.23} & 0.935 & 0.941 & 0.623 & \bf{0.477} & 0.0518 & 0.0368 & 0.0278 & \bf{0.0061}\\
        Library  & 37.72 & 55.34 & 35.60 & \bf{35.24} & 0.815 & 0.880 & 0.572 & \bf{0.521} & 0.0431 & 0.0232 & 0.0183 & \bf{0.0098}\\
        \bottomrule
    \end{tabular}}
    \vspace{0.2em}
    \caption{\small {\bf Quantitative Comparison.} We report the FID score
    ($\downarrow$) at $256^2$ pixels, the KL divergence ($\downarrow$) between the
    distribution of object categories of synthesized and real scenes and the real vs. synthetic classification
    accuracy for all methods. Classification accuracy closer to $0.5$ 
    is better.}
    \vspace{-1.2em}
    \label{tab:scene_synthesis_quantitative}
\end{table}

\vspace{-2mm}
\subsection{Scene Synthesis}
We start by evaluating the performance of our model on generating
plausible object configurations for various room types, conditioned on
different floor plans. \figref{fig:scene_synthesis_qualitative} provides a
qualitative comparison of four scenes generated with our model and baselines.
In some cases, both \cite{Ritchie2019CVPR, Wang2020ARXIV} generate invalid room
layouts with objects positioned outside room boundaries or overlapping.
Instead, our model consistently synthesizes realistic object
arrangements. We validate this quantitatively in
\tabref{tab:scene_synthesis_quantitative}, where we compare the generated
scenes \wrt their similarity to the original data from 3D-FRONT. Synthesized scenes
sampled from our model are almost indistinguishable from scenes from the test
set, as indicated by the classification accuracy in
\tabref{tab:scene_synthesis_quantitative}, which is consistently around $50\%$.
Our model also achieves lower FID scores for all room types and
generates category distributions that are more faithful to the category
distributions of the test set, expressed as lower KL divergence.

\begin{figure}
    \vspace{-0.8em}
    \centering
    \begin{minipage}[b]{0.124\linewidth}
		\centering
        \small Scene Layout
    \end{minipage}%
    \hfill%
    \begin{minipage}[b]{0.124\linewidth}
		\centering
        \small FastSynth
    \end{minipage}%
    \hfill%
    \begin{minipage}[b]{0.124\linewidth}
		\centering
        \small SceneFormer
    \end{minipage}%
    \hfill%
    \begin{minipage}[b]{0.124\linewidth}
        \centering
        \small Ours
    \end{minipage}
    \hfill%
    \begin{minipage}[b]{0.124\linewidth}
		\centering
        \small Scene Layout
    \end{minipage}%
    \hfill%
    \begin{minipage}[b]{0.124\linewidth}
		\centering
        \small FastSynth
    \end{minipage}%
    \hfill%
    \begin{minipage}[b]{0.124\linewidth}
		\centering
        \small SceneFormer
    \end{minipage}%
    \hfill%
    \begin{minipage}[b]{0.124\linewidth}
        \centering
        \small Ours
    \end{minipage}
    \vskip\baselineskip%
    \vspace{-1.2em}
    \begin{minipage}[b]{0.125\linewidth}
		\centering
		\includegraphics[width=\linewidth]{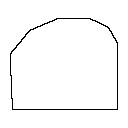}
    \end{minipage}%
    \hfill%
    \begin{minipage}[b]{0.125\linewidth}
		\centering
		\includegraphics[width=\linewidth, trim=600 250 600 200, clip]{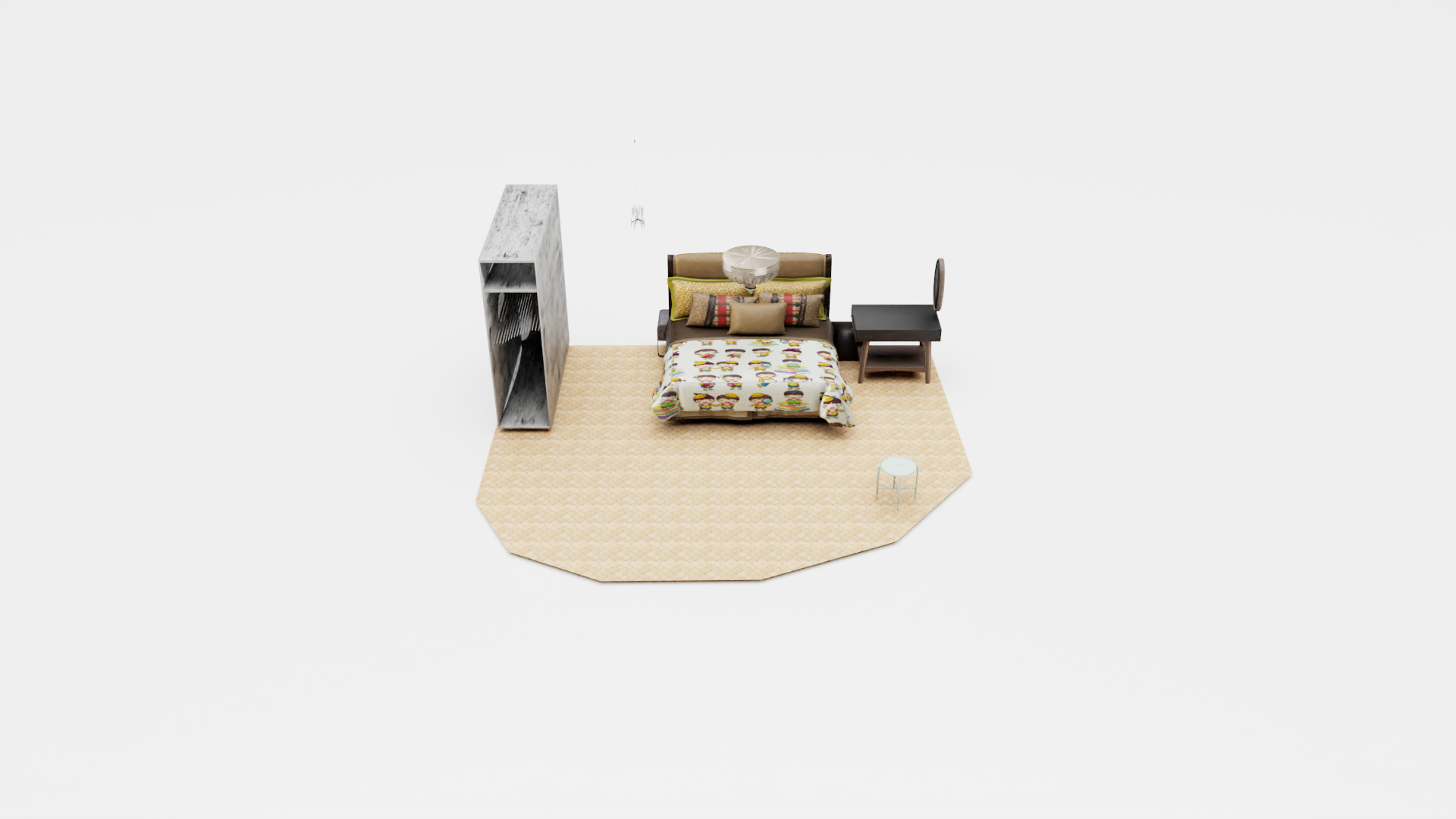}
    \end{minipage}%
    \hfill%
    \begin{minipage}[b]{0.125\linewidth}
		\centering
		\includegraphics[width=\linewidth, trim=600 250 600 200, clip]{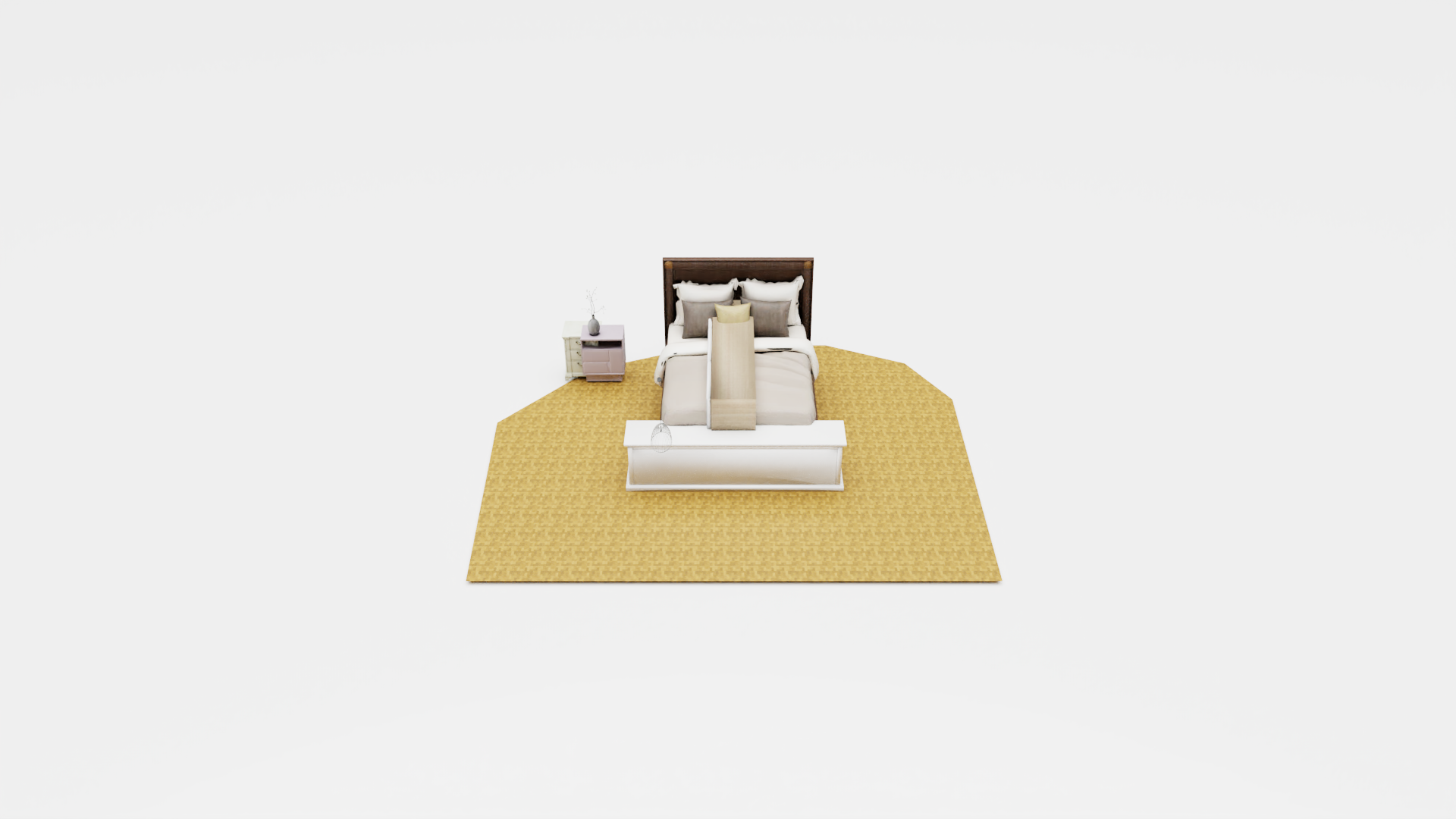}
    \end{minipage}%
    \begin{minipage}[b]{0.125\linewidth}
		\centering
		\includegraphics[width=\linewidth, trim=600 250 600 200, clip]{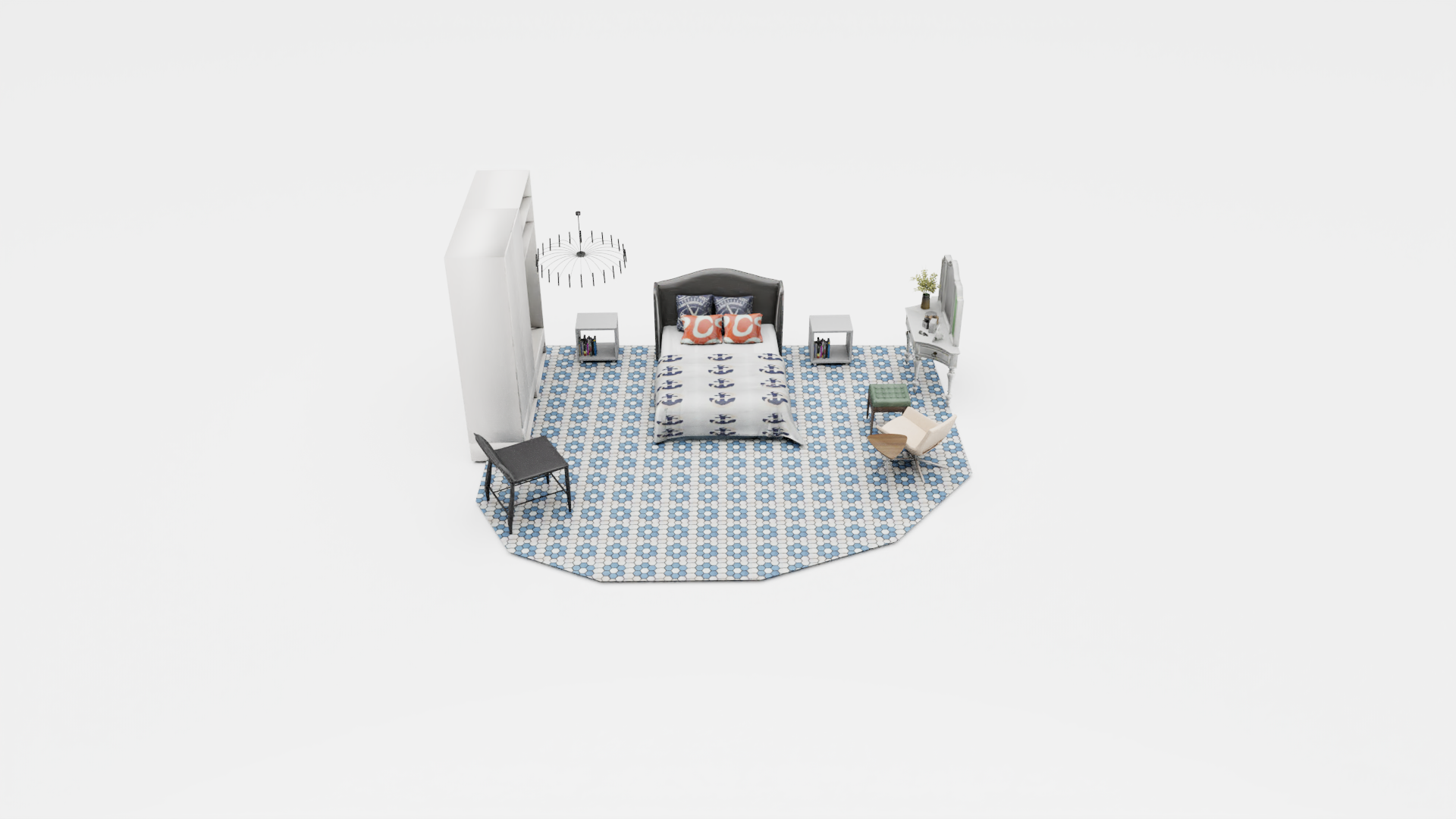}
    \end{minipage}%
    \begin{minipage}[b]{0.125\linewidth}
		\centering
		\includegraphics[width=\linewidth]{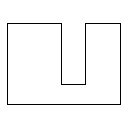}
    \end{minipage}%
    \hfill%
    \begin{minipage}[b]{0.125\linewidth}
		\centering
		\includegraphics[width=\linewidth, trim=550 200 550 150, clip]{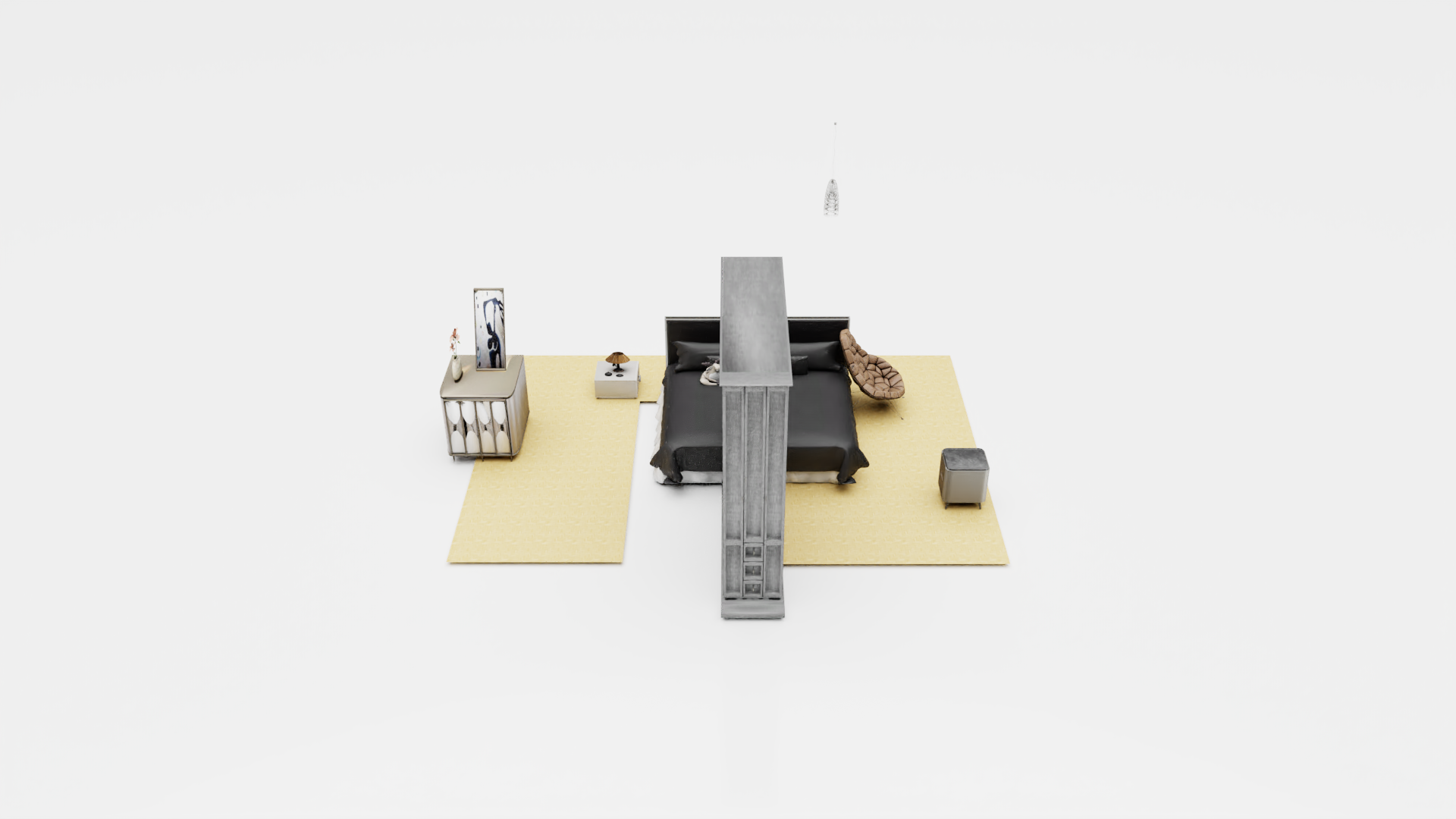}
    \end{minipage}%
    \hfill%
    \begin{minipage}[b]{0.125\linewidth}
		\centering
		\includegraphics[width=\linewidth, trim=550 200 550 150, clip]{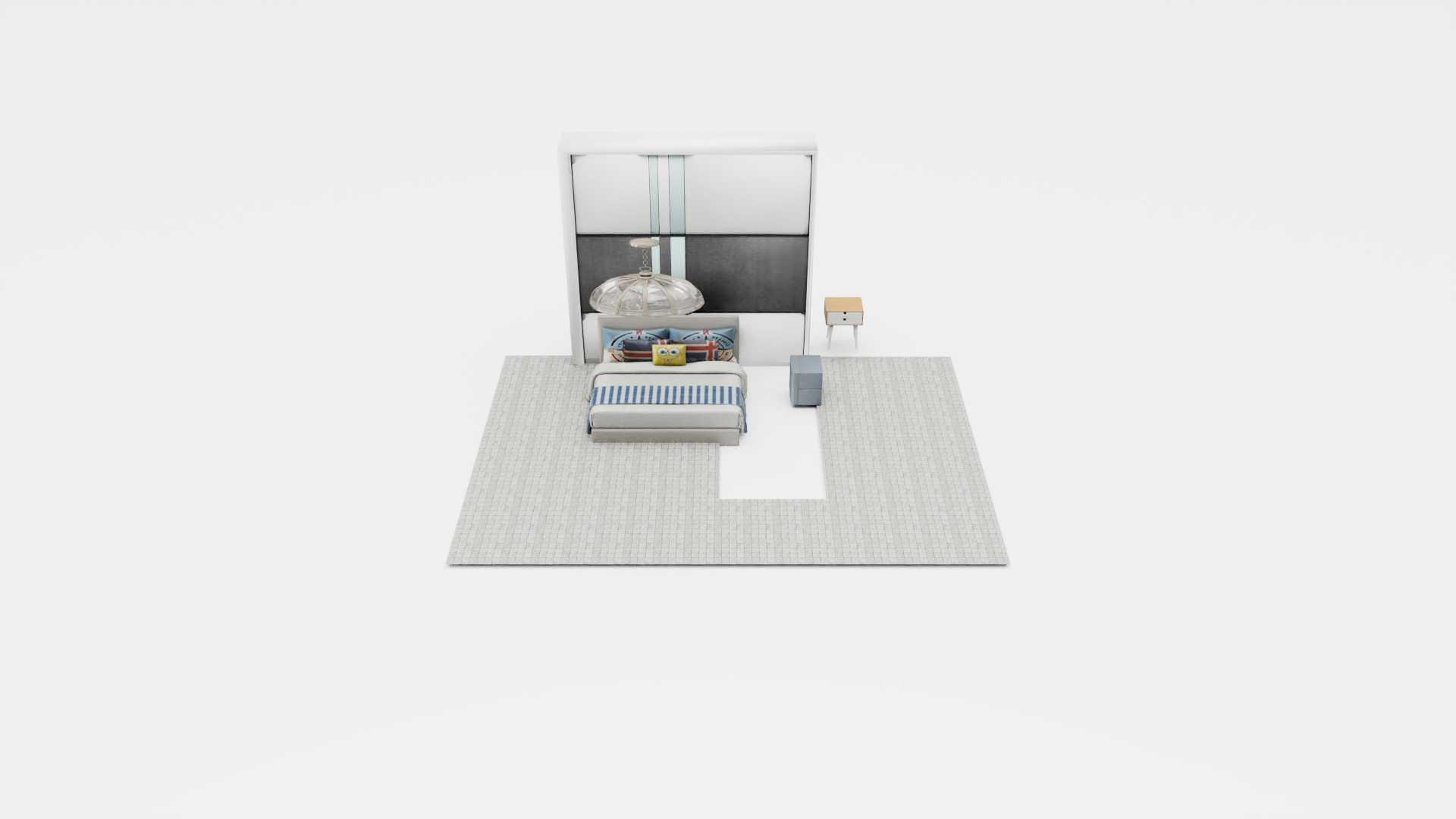}
    \end{minipage}%
    \begin{minipage}[b]{0.125\linewidth}
		\centering
		\includegraphics[width=\linewidth, trim=550 200 550 150, clip]{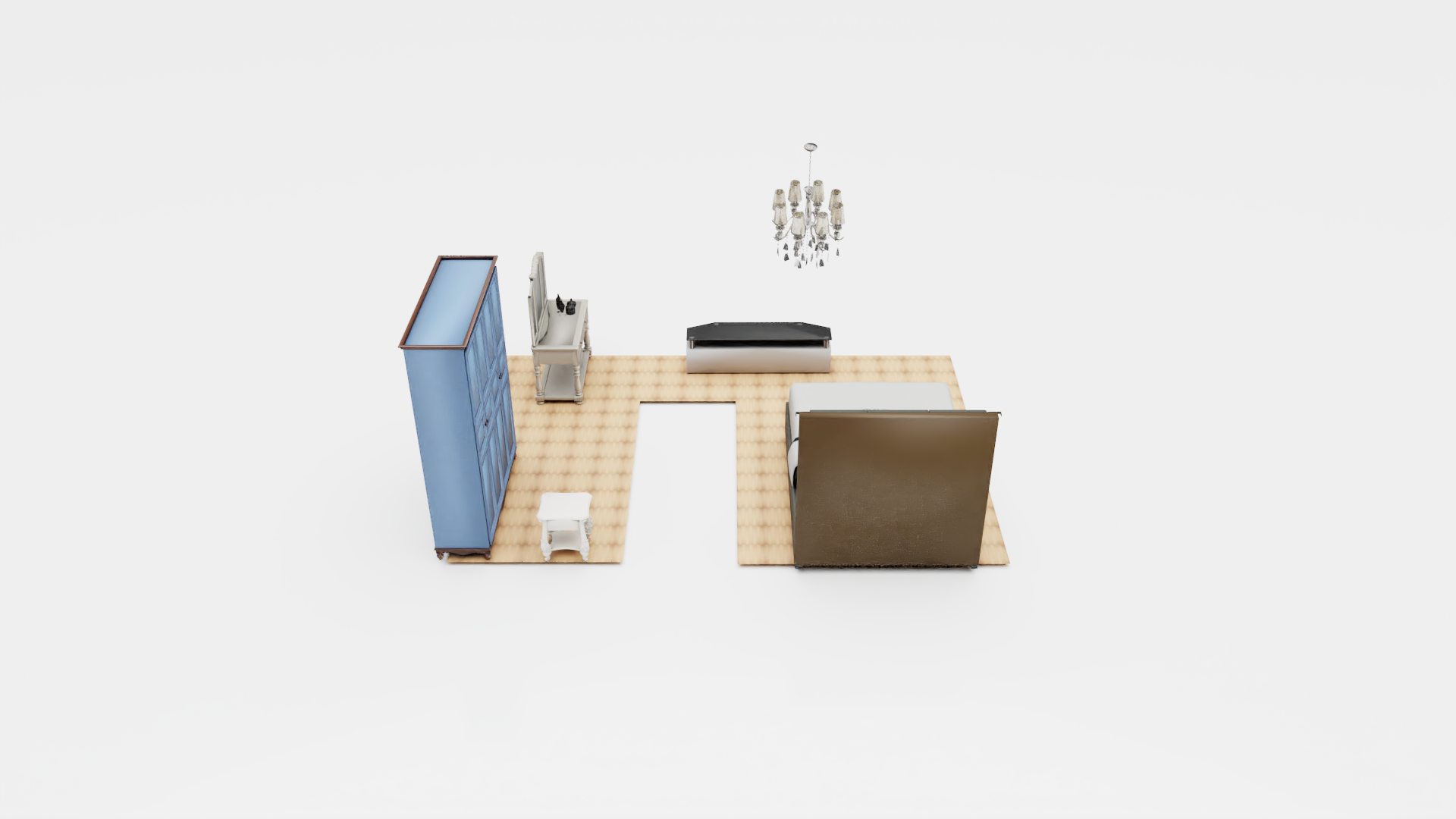}
    \end{minipage}%
    \vspace{-1.2em}
    \vskip\baselineskip%
    \begin{minipage}[b]{0.125\linewidth}
		\centering
		\includegraphics[width=\linewidth]{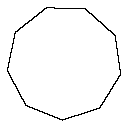}
    \end{minipage}%
    \hfill%
    \begin{minipage}[b]{0.125\linewidth}
		\centering
		\includegraphics[width=\linewidth, trim=625 150 575 80, clip]{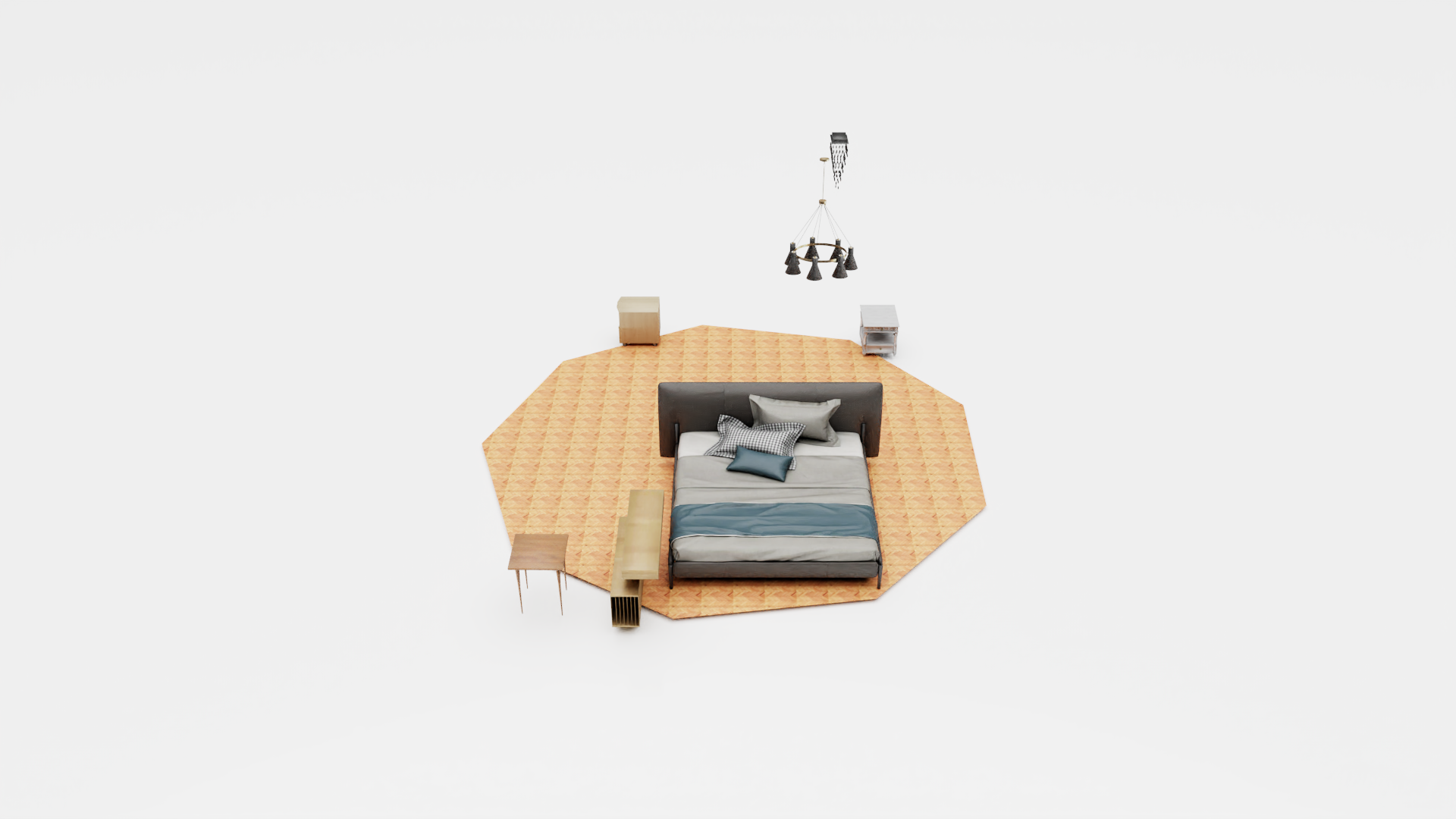}
    \end{minipage}%
    \hfill%
    \begin{minipage}[b]{0.125\linewidth}
		\centering
		\includegraphics[width=\linewidth, trim=625 150 575 80, clip]{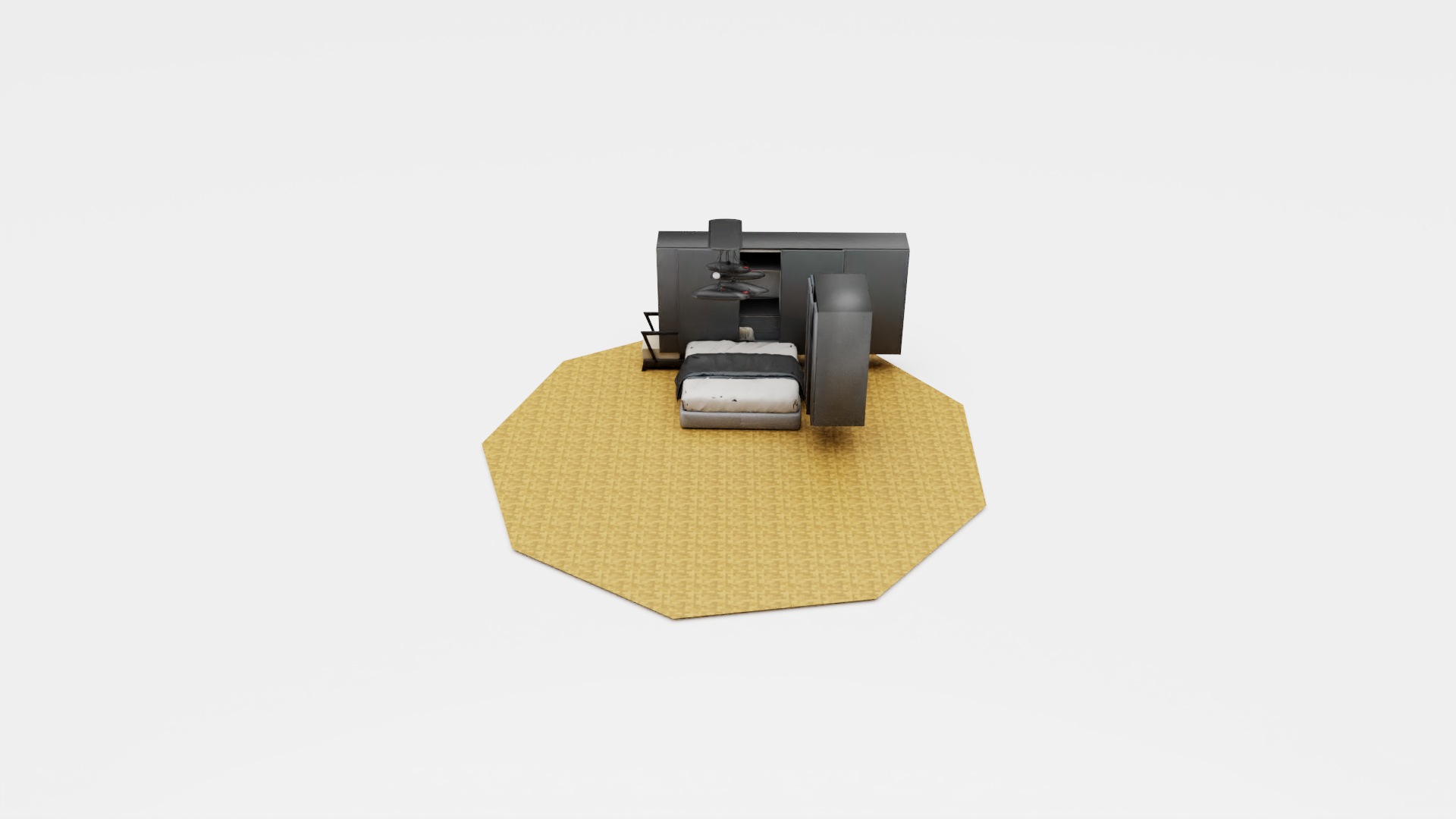}
    \end{minipage}%
    \begin{minipage}[b]{0.125\linewidth}
		\centering
		\includegraphics[width=\linewidth, trim=625 150 575 80, clip]{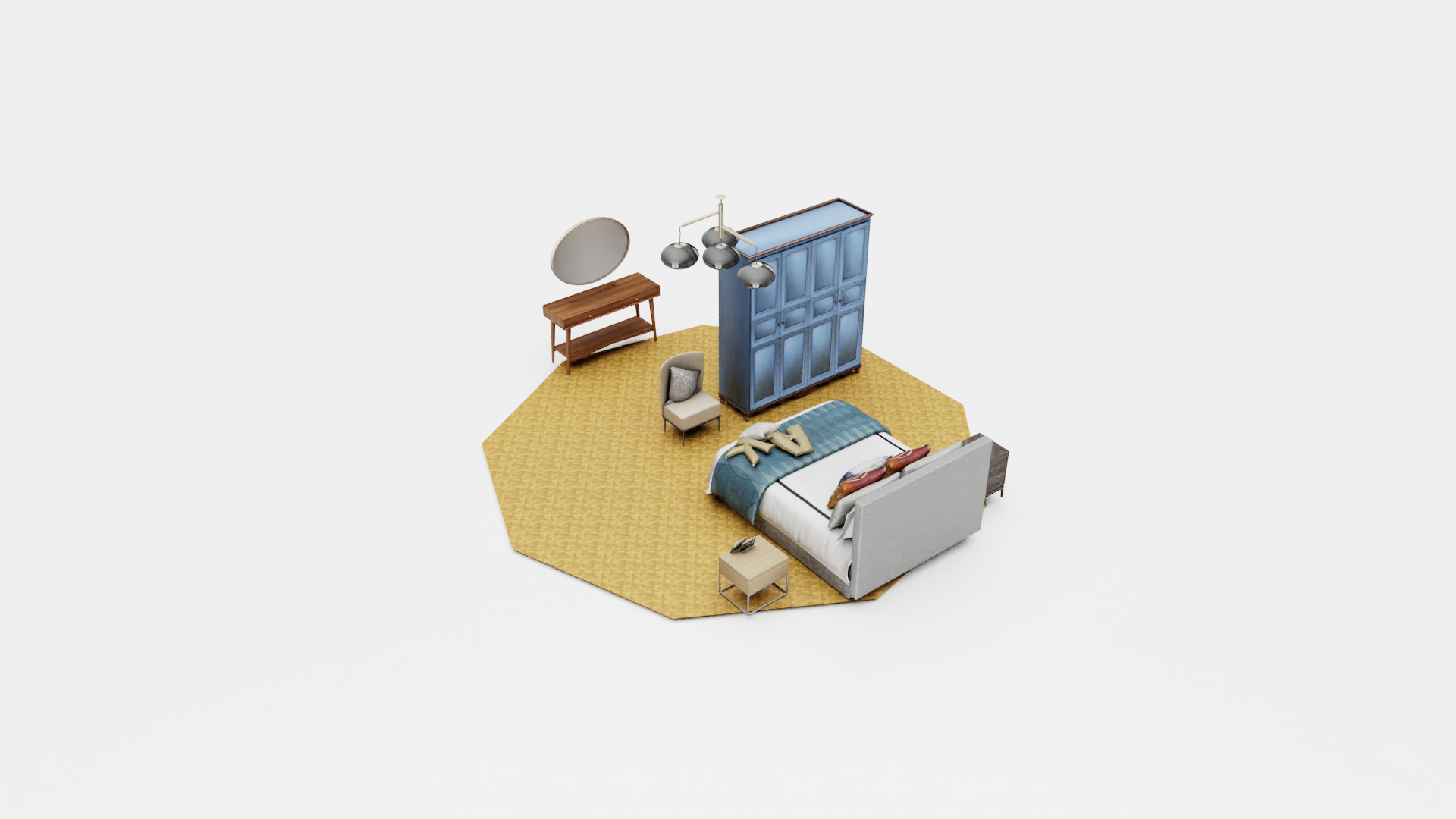}
    \end{minipage}%
    \begin{minipage}[b]{0.125\linewidth}
		\centering
		\includegraphics[width=\linewidth]{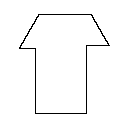}
    \end{minipage}%
    \hfill%
    \begin{minipage}[b]{0.125\linewidth}
		\centering
		\includegraphics[width=\linewidth, trim=650 250 650 100, clip]{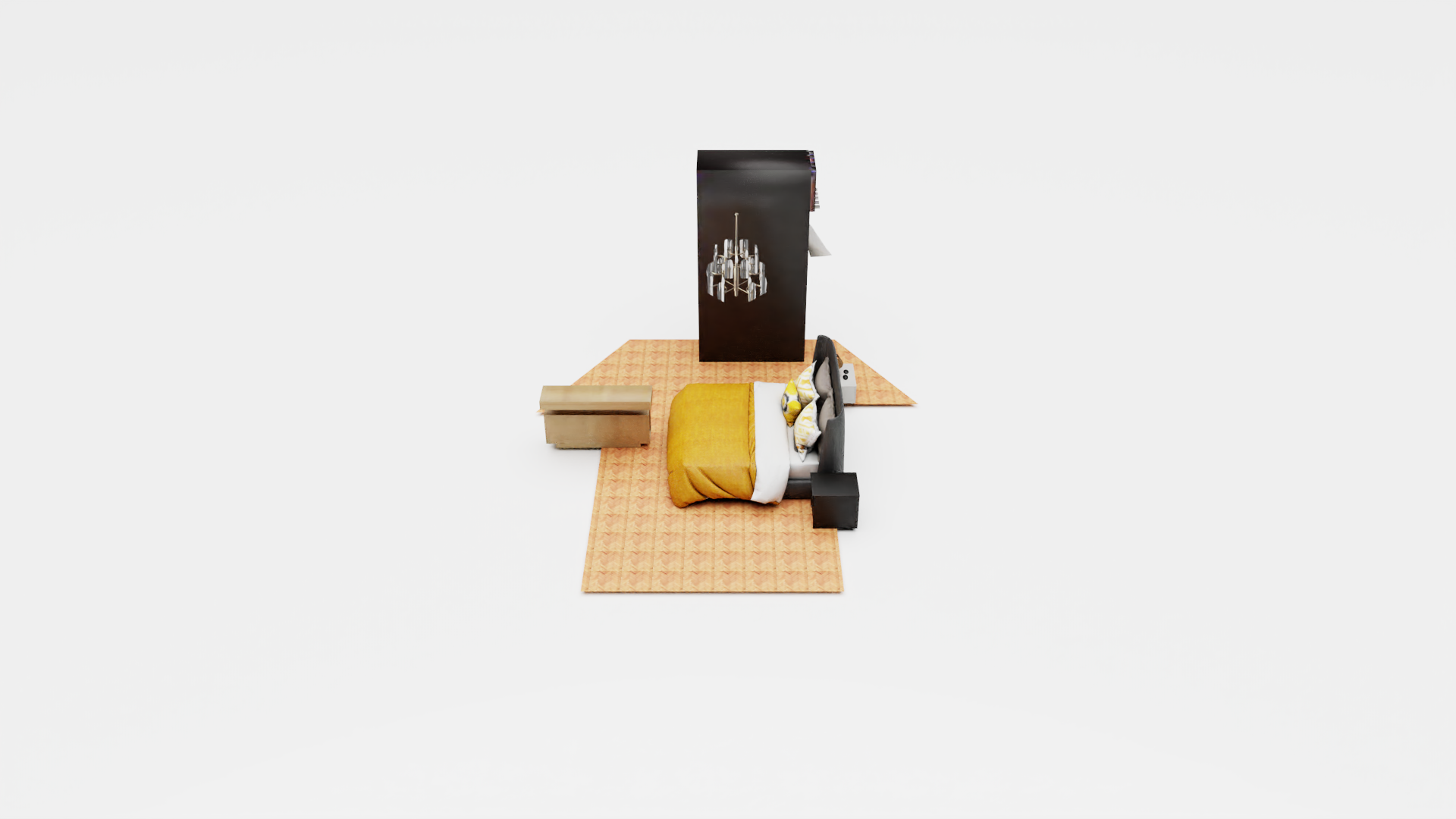}
    \end{minipage}%
    \hfill%
    \begin{minipage}[b]{0.125\linewidth}
		\centering
		\includegraphics[width=\linewidth, trim=700 250 600 100, clip]{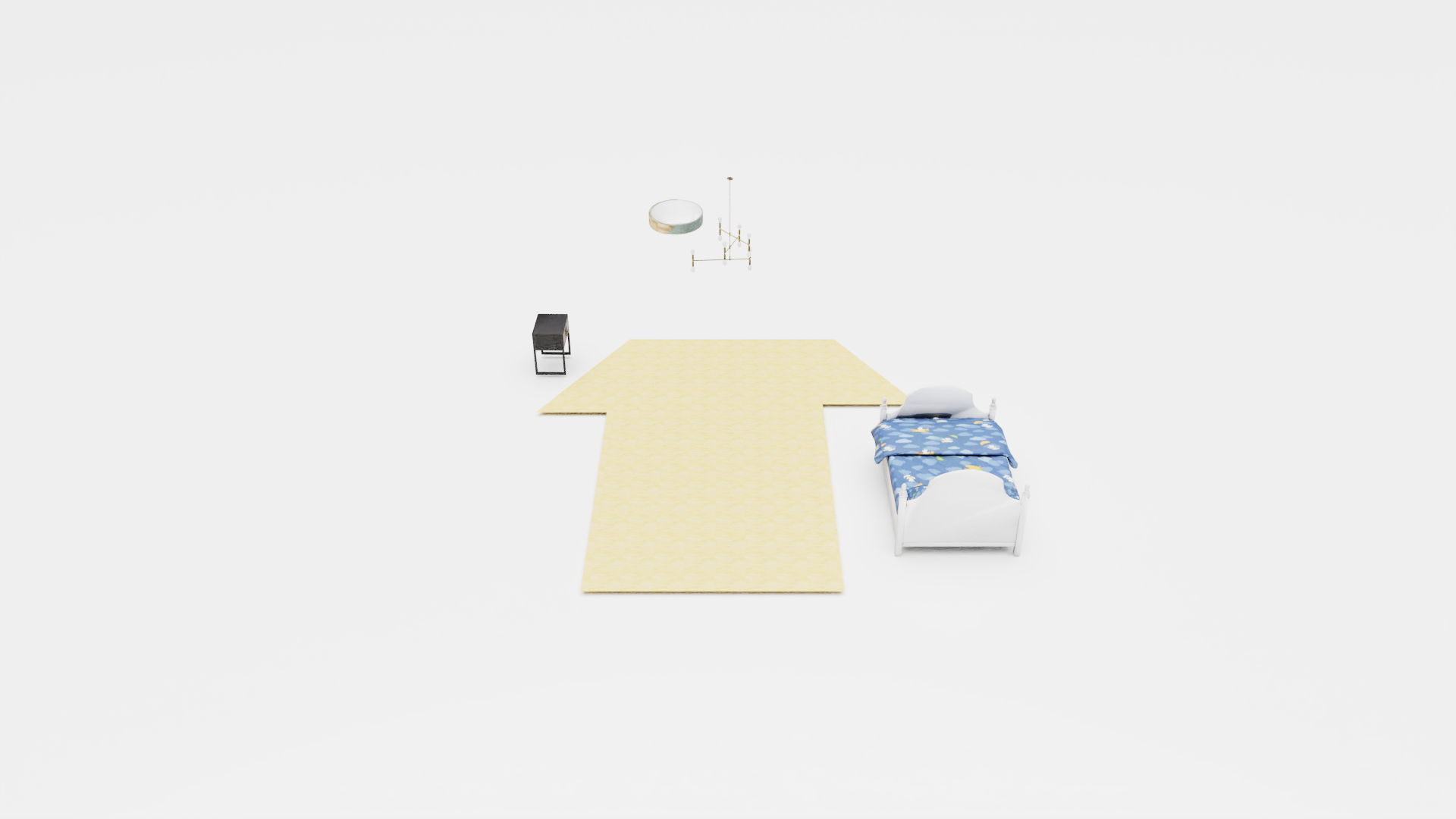}
    \end{minipage}%
    \begin{minipage}[b]{0.125\linewidth}
		\centering
		\includegraphics[width=\linewidth, trim=650 250 650 100, clip]{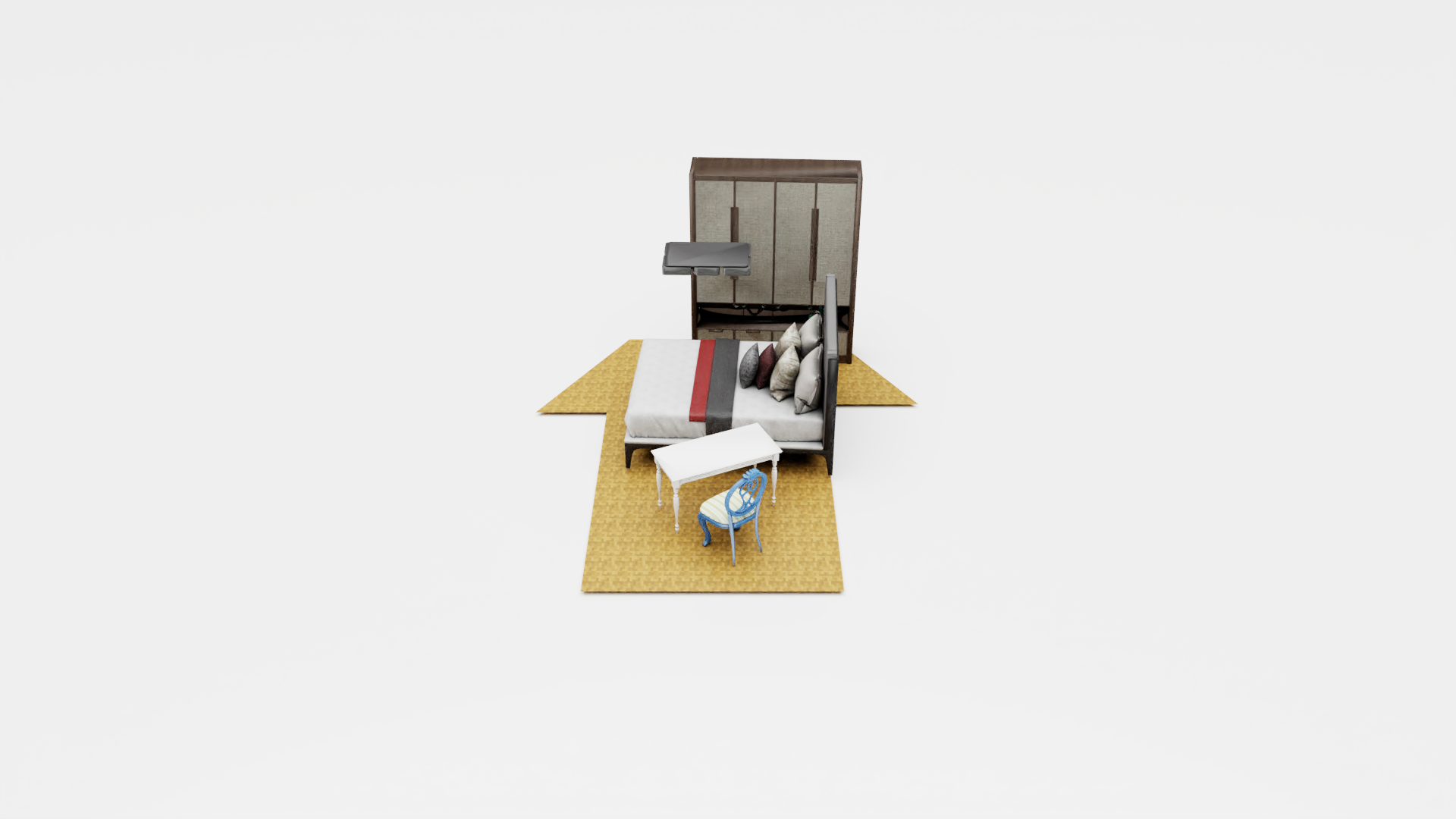}
    \end{minipage}%
    \vspace{-1.2em}
    \vskip\baselineskip%
    \caption{\small{\bf Generalization Beyond Training Data}. We show four synthesized bedrooms
    conditioned on four room layouts that we manually designed.}
    \label{fig:scene_synthesis_generalization}
    \vspace{-1.2em}
\end{figure}

To showcase that our model generates diverse object arrangements we
visualize $3$ generated scenes conditioned on the same floor plan for all
methods (\figref{fig:scene_synthesis_diversity}). We observe that our generated
scenes are consistently valid and contain diverse object arrangements. In comparison
\cite{Ritchie2019CVPR, Wang2020ARXIV} struggle to generate plausible layouts
particularly for the case of living rooms and dining rooms. We hypothesize that these rooms are more challenging than bedrooms, for the baselines, due to their significantly smaller volume of training data,
and the large number of constituent objects per scene ($20$ on average, as opposed to $8$).
To investigate whether our model 
also generates plausible layouts conditioned on floor plans with
uncommon shapes that are not in the training set, we manually design unconventional floor plans (\figref{fig:scene_synthesis_generalization}) and generate bedroom layouts.
While both \cite{Ritchie2019CVPR, Wang2020ARXIV} fail to generate valid scenes,
our model synthesizes diverse object layouts that are consistent with the floor plan.
Finally, we compare the computational requirements of our architecture to
\cite{Ritchie2019CVPR, Wang2020ARXIV}. Our model is significantly faster
(\tabref{tab:generation_time}), while having fewer parameters
(\tabref{tab:network_parameters}) than both \cite{Ritchie2019CVPR, Wang2020ARXIV}. Note
that \cite{Ritchie2019CVPR} is orders of magnitude slower because it requires
rendering every individual object added in the scene.
\begin{figure}
    \centering
    \vspace{-1.2em}
    \begin{minipage}[b]{0.19\linewidth}
		\centering
        \small Partial Scene
    \end{minipage}%
    \hfill%
    \begin{minipage}[b]{0.19\linewidth}
		\centering
        \small FastSynth
    \end{minipage}%
    \hfill%
    \begin{minipage}[b]{0.19\linewidth}
		\centering
        \small SceneFormer
    \end{minipage}%
    \hfill%
    \begin{minipage}[b]{0.19\linewidth}
        \centering
        \small Ours+Order
    \end{minipage}
    \hfill%
    \begin{minipage}[b]{0.19\linewidth}
        \centering
        \small Ours
    \end{minipage}
    \vskip\baselineskip%
    \vspace{-1em}
    \begin{minipage}[b]{0.2\linewidth}
		\centering
		\includegraphics[width=\linewidth, trim=500 300 500 200, clip]{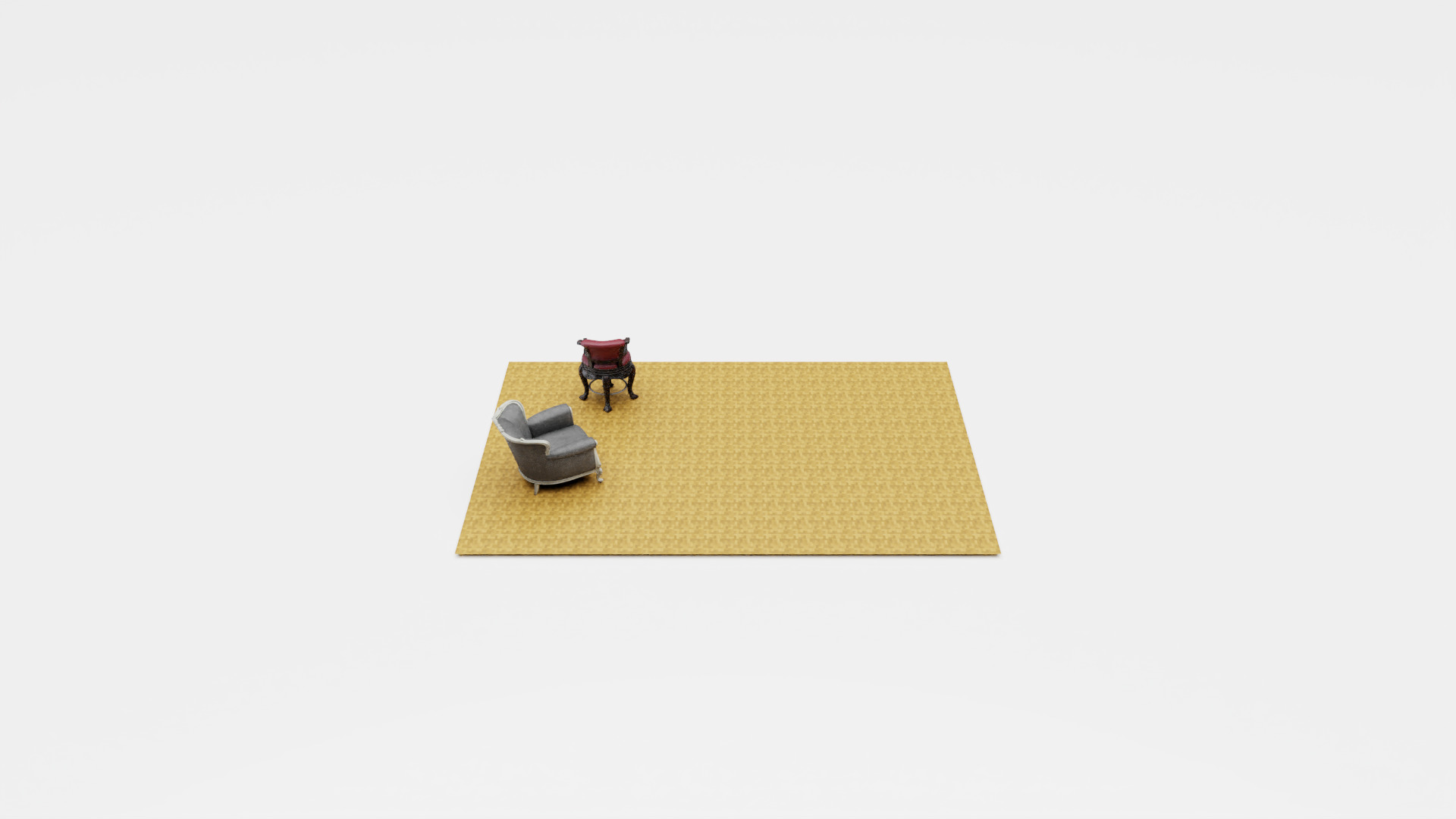}
    \end{minipage}%
    \hfill%
    \begin{minipage}[b]{0.2\linewidth}
		\centering
		\includegraphics[width=\linewidth, trim=500 300 500 200, clip]{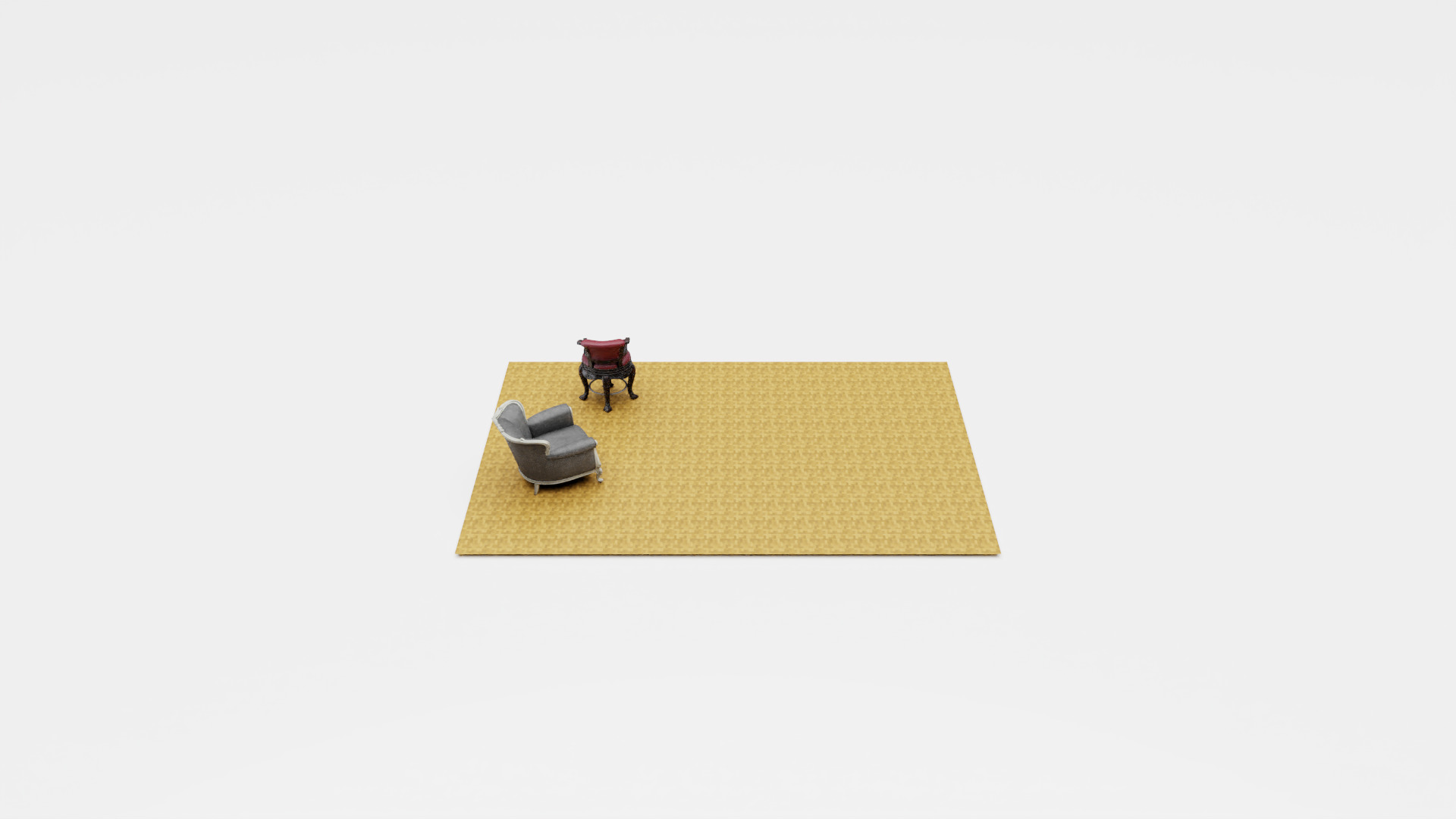}
    \end{minipage}%
    \hfill%
    \begin{minipage}[b]{0.2\linewidth}
		\centering
		\includegraphics[width=\linewidth, trim=500 300 500 200, clip]{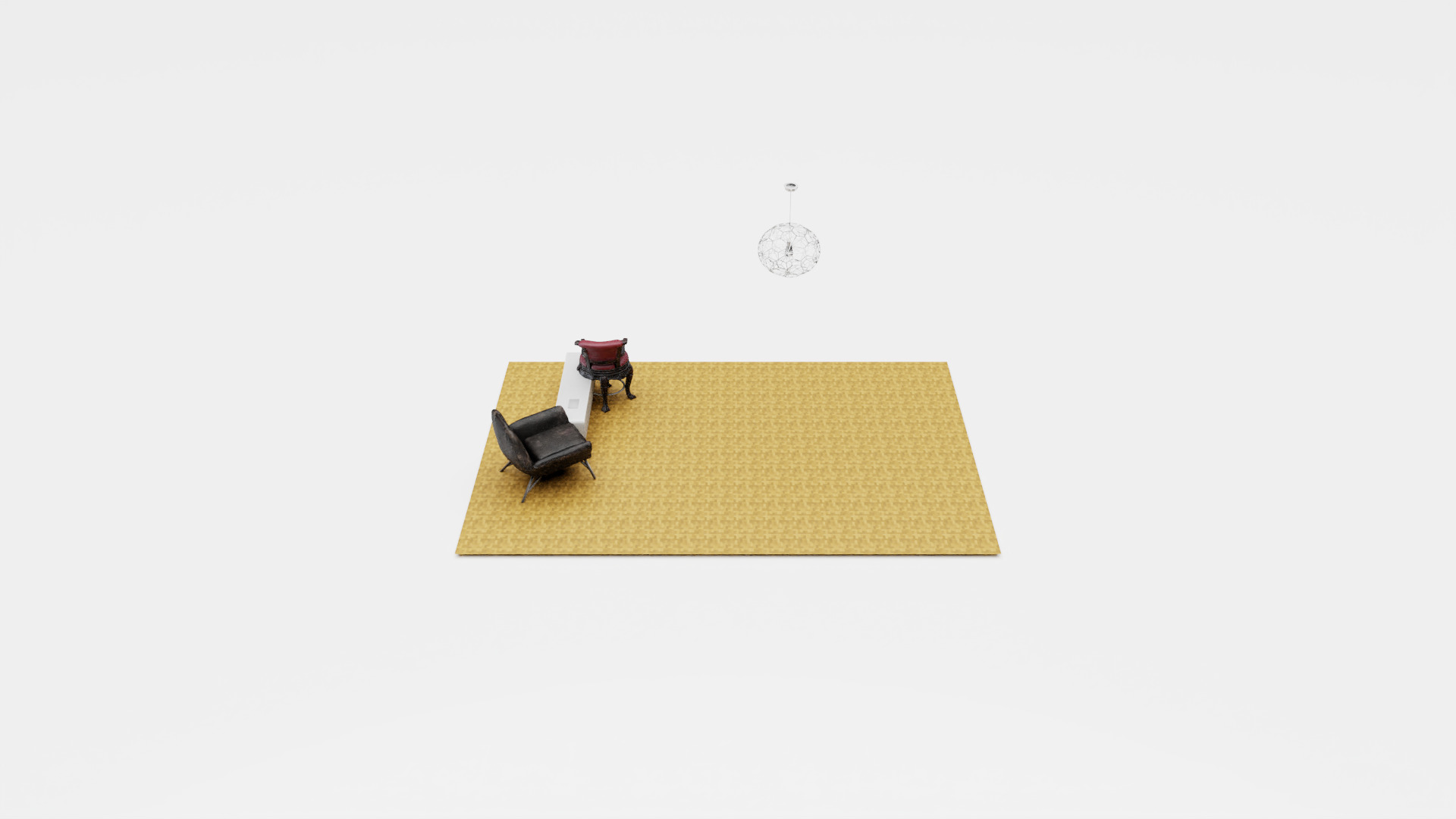}
    \end{minipage}%
    \begin{minipage}[b]{0.2\linewidth}
		\centering
		\includegraphics[width=\linewidth, trim=500 300 500 200, clip]{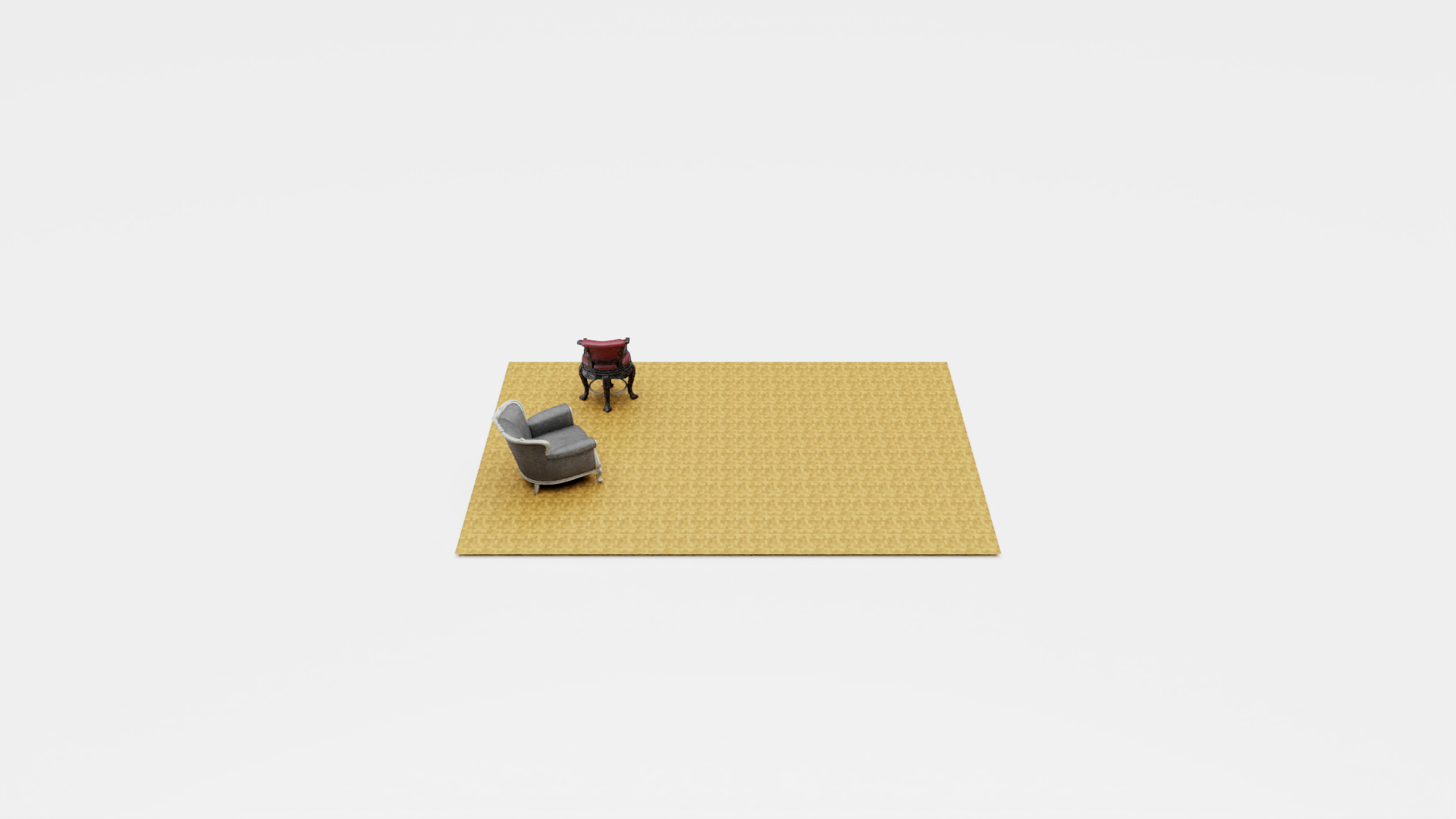}
    \end{minipage}%
    \begin{minipage}[b]{0.2\linewidth}
		\centering
		\includegraphics[width=\linewidth, trim=500 300 500 200, clip]{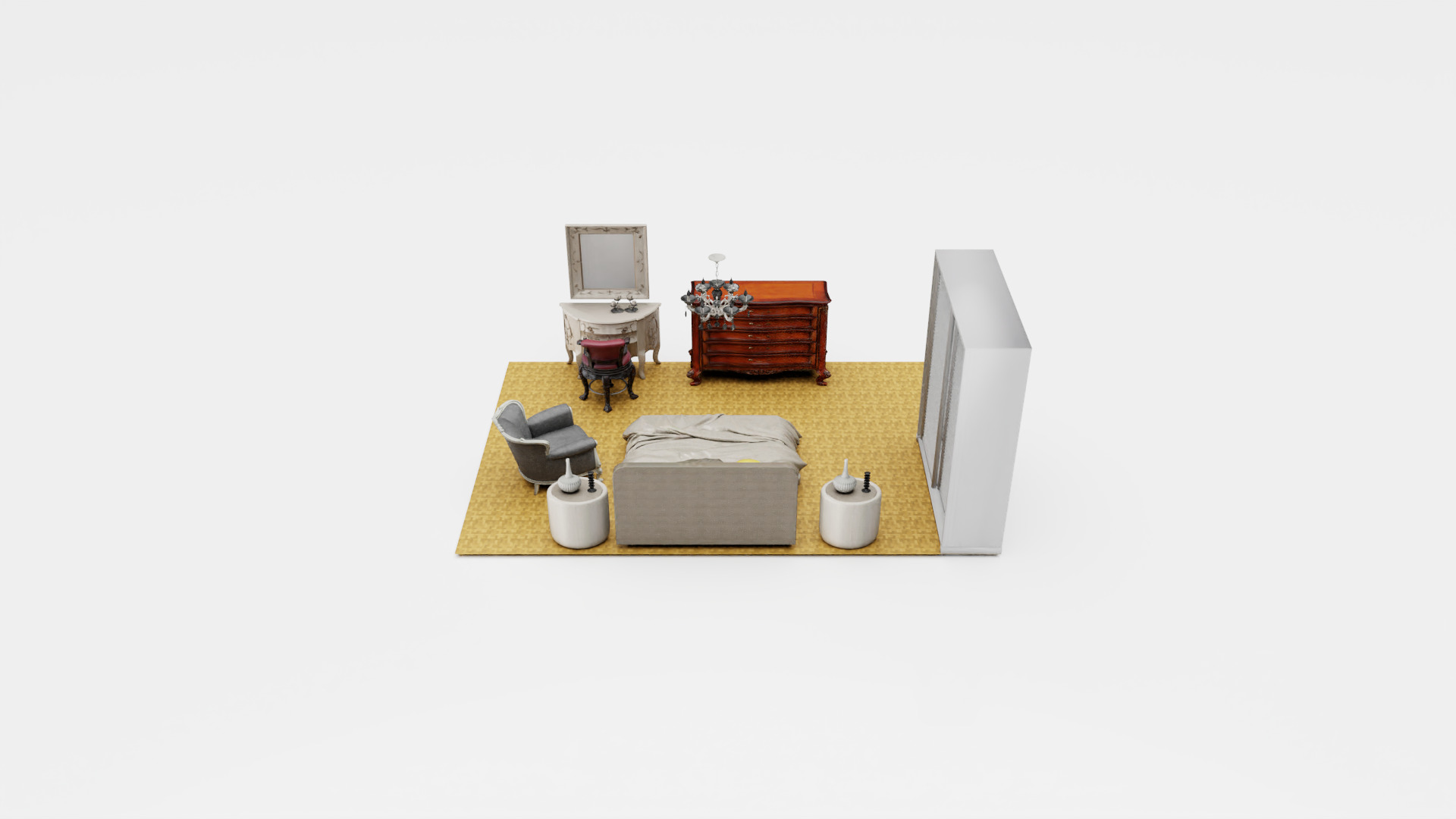}
    \end{minipage}%
    \vskip\baselineskip%
    \vspace{-1.5em}
    \begin{minipage}[b]{0.2\linewidth}
		\centering
		\includegraphics[width=\linewidth, trim=500 300 500 200, clip]{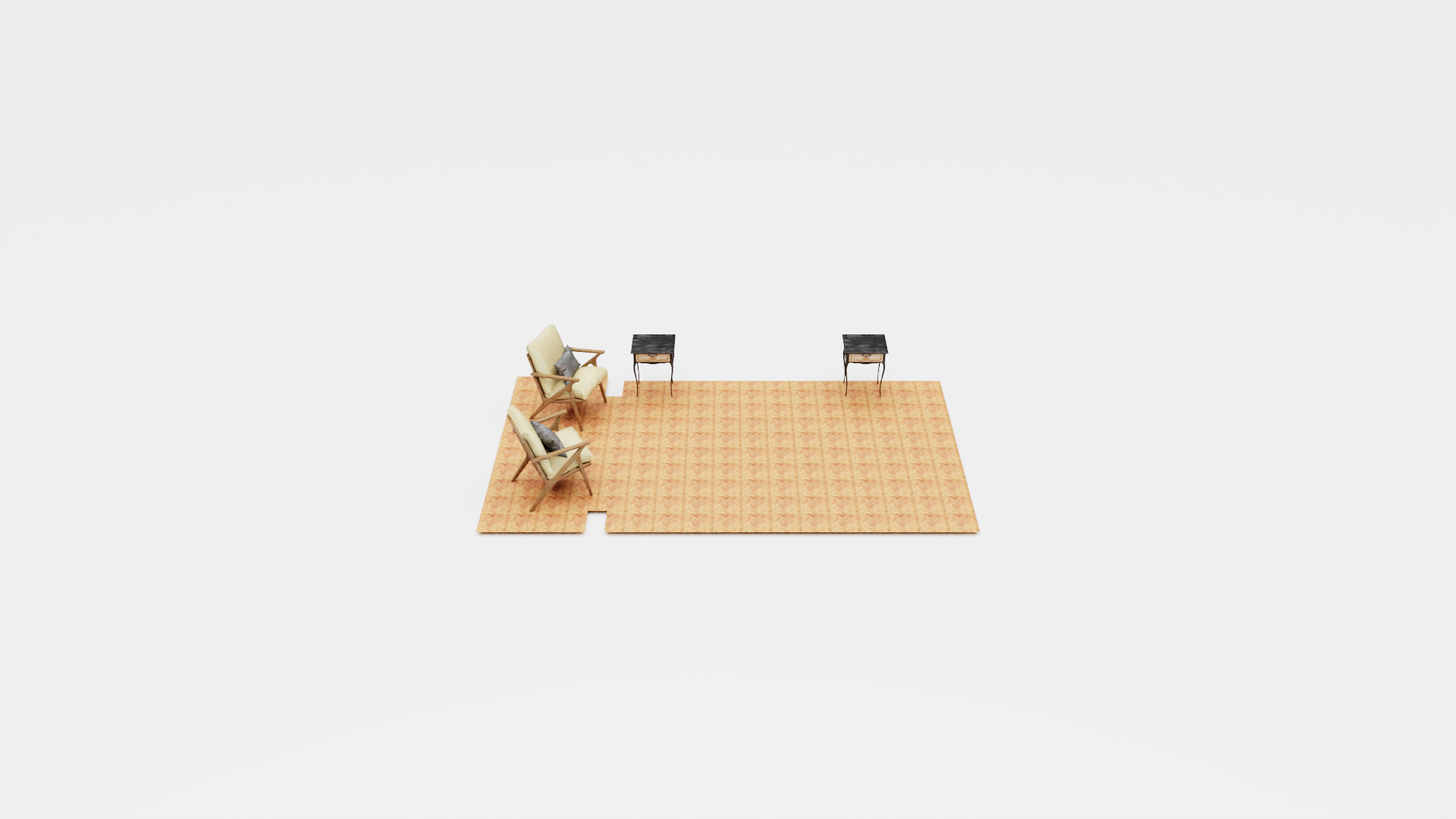}
    \end{minipage}%
    \hfill%
    \begin{minipage}[b]{0.2\linewidth}
		\centering
		\includegraphics[width=\linewidth, trim=500 300 500 200, clip]{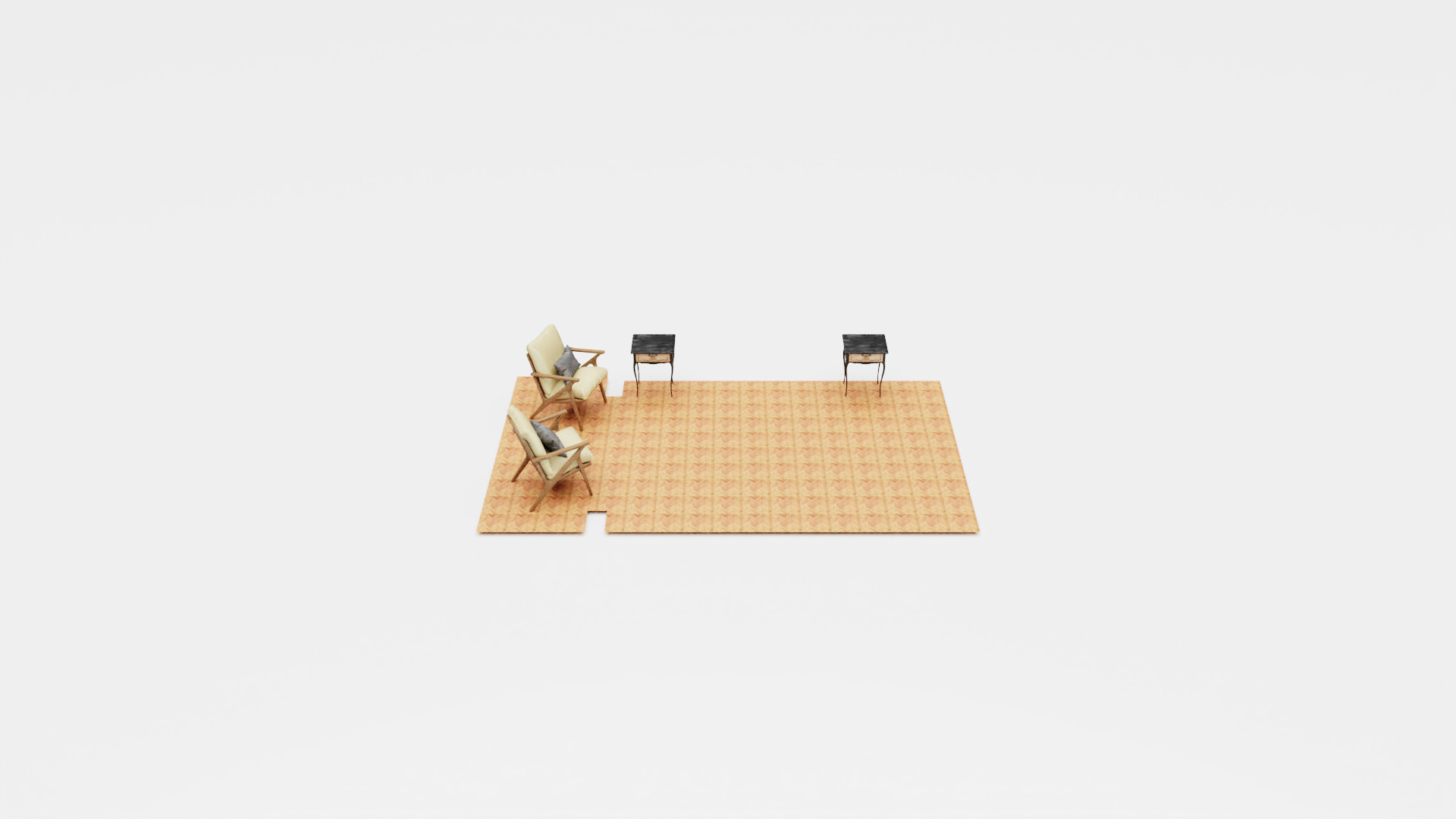}
    \end{minipage}%
    \hfill%
    \begin{minipage}[b]{0.2\linewidth}
		\centering
		\includegraphics[width=\linewidth, trim=500 300 500 200, clip]{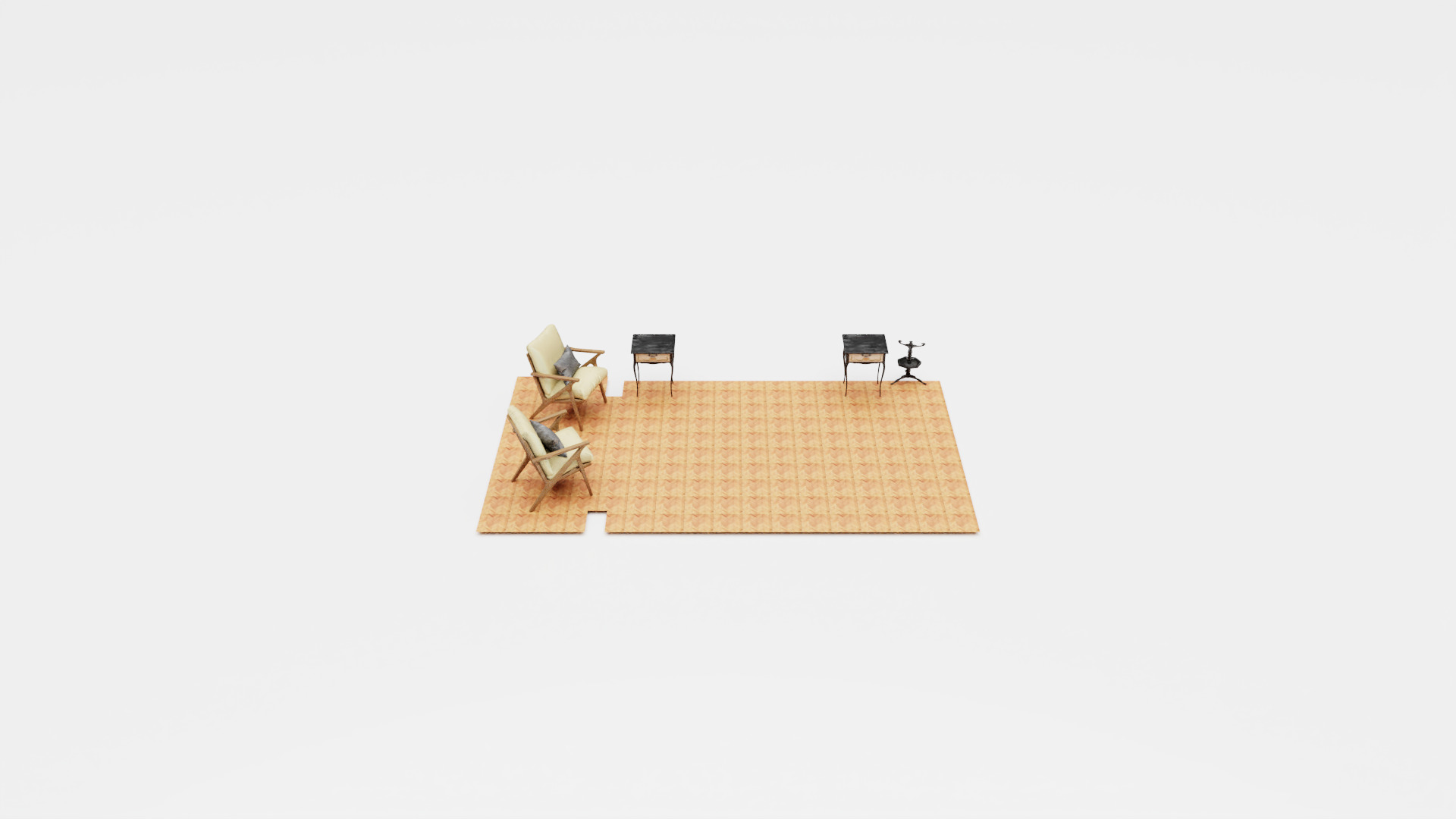}
    \end{minipage}%
    \begin{minipage}[b]{0.2\linewidth}
		\centering
		\includegraphics[width=\linewidth, trim=500 300 500 200, clip]{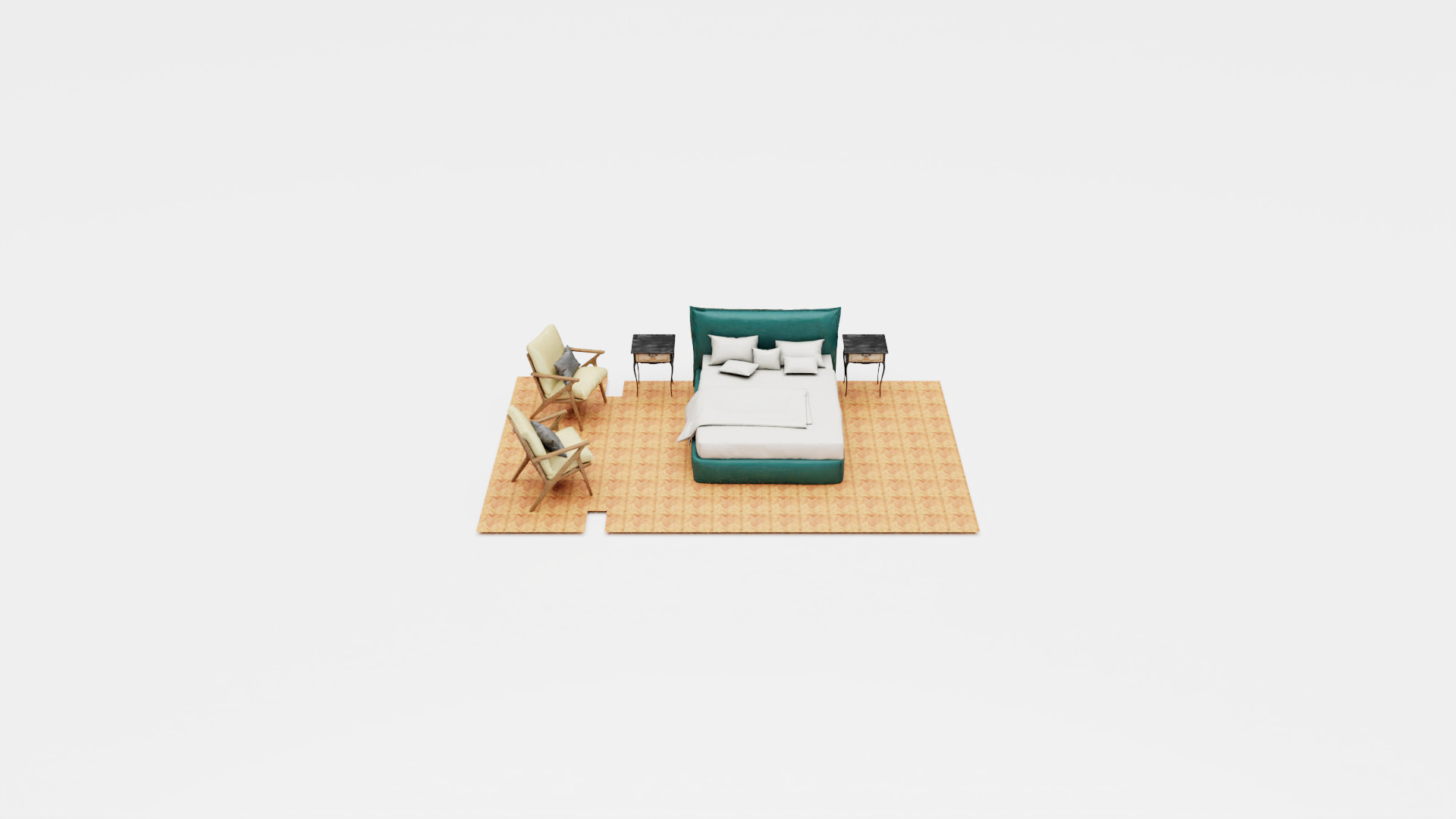}
    \end{minipage}%
    \begin{minipage}[b]{0.2\linewidth}
		\centering
		\includegraphics[width=\linewidth, trim=500 300 500 200, clip]{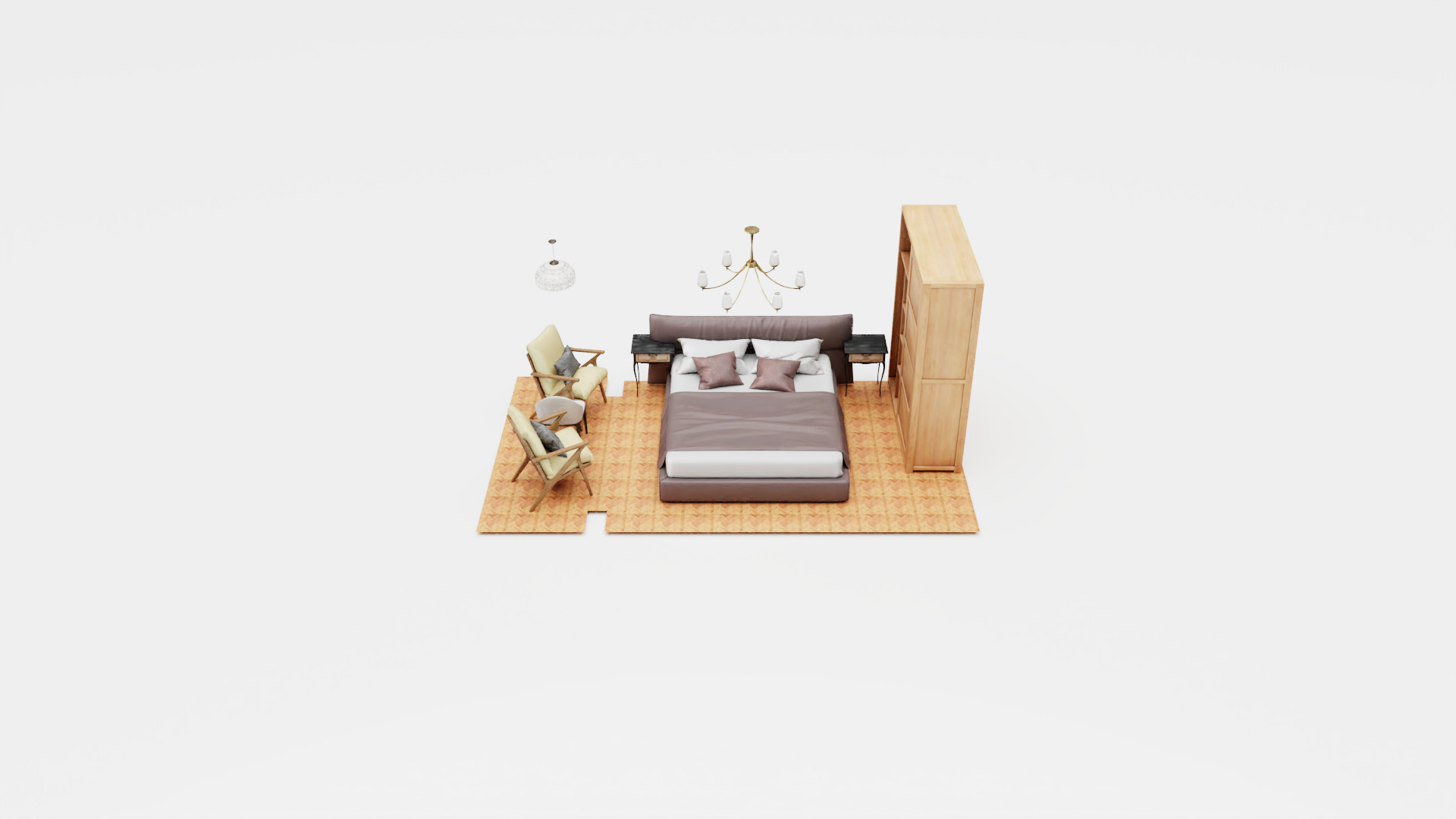}
    \end{minipage}%
    \vspace{-1.4em}
    \vskip\baselineskip%
    \caption{\small{\bf Scene Completion}.
             Given a partial scene (left column), we visualize scene
             completions using our model and our baselines. Our model
             consistently generates plausible layouts.}
    \label{fig:scene_completion}
    \vspace{-0.8em}
\end{figure}
\begin{table}
\begin{minipage}[b!]{0.51\linewidth}
    \centering
    \resizebox{\linewidth}{!}{
    \begin{tabular}{l|cccc}
    \toprule
    & Bedroom & Living & Dining & Library \\
    \midrule
    FastSynth \cite{Ritchie2019CVPR} & 13193.77    & 30578.54 & 26596.08 & 10813.87\\
    SceneFormer \cite{Wang2020ARXIV} & 849.37      & 731.84 & 901.17 & 369.74\\
    Ours \cite{Ritchie2019CVPR}      & \bf{102.38} & \bf{201.59} & \bf{201.84} & \bf{88.24}\\
    \bottomrule
    \end{tabular}}
    \vspace{0.1em}
    \caption{\small {\bf Generation Time Comparison.} We measure time (ms) to
    generate a scene, conditioned on a floor plan.}     \label{tab:generation_time}
    \vspace{-0.8em}
\end{minipage}
\hfill%
\begin{minipage}[b!]{0.465\linewidth}
    \centering
    \vspace{-1.2em}
    \resizebox{0.95\linewidth}{!}{
    \begin{tabular}{ccc}
    \toprule
    FastSynth \cite{Ritchie2019CVPR} & SceneFormer \cite{Wang2020ARXIV} & Ours\\
    \midrule
    38.180 & 129.298 & 36.053\\
    \bottomrule
    \end{tabular}}
    \vspace{0.1em}
    \caption{\small {\bf Network Parameters Comparison.} We report the number of network parameters in millions.}
    \label{tab:network_parameters}
    \vspace{-0.8em}
\end{minipage}

\end{table}

\subsection{Applications}
In this section, we present three applications that greatly benefit by our unordered set formulation and are crucial for creating an interactive scene synthesis tool.

\boldparagraph{Scene Completion}%
Starting from a partial scene, the task is to populate the empty space in a
meaningful way. Since both \cite{Ritchie2019CVPR, Wang2020ARXIV} are trained on
sorted sequences of objects, they first generate frequent objects (\eg beds,
wardrobes) followed by less common objects. As a result, incomplete scenes that
contain less common objects cannot be correctly populated.  This is illustrated
in \figref{fig:scene_completion}, where \cite{Ritchie2019CVPR, Wang2020ARXIV}
either fail to add any objects in the scene or place furnitures in unatural
positions, thus resulting in bedrooms without beds (see 1st row
\figref{fig:scene_completion}) and scenes with overlapping furniture (see 2nd
row \figref{fig:scene_completion}). In contrary, our model successfullly
generates plausible completions with multiple objects such as lamps, wardrobes
and dressing tables.

\begin{figure}
\vspace{-3mm}
    \centering
    \vskip\baselineskip%
    \vspace{-0.5em}
    \hfill%
    \begin{subfigure}[b]{0.16\linewidth}
    \centering
    \includegraphics[width=\linewidth, trim=500 270 500 125, clip]{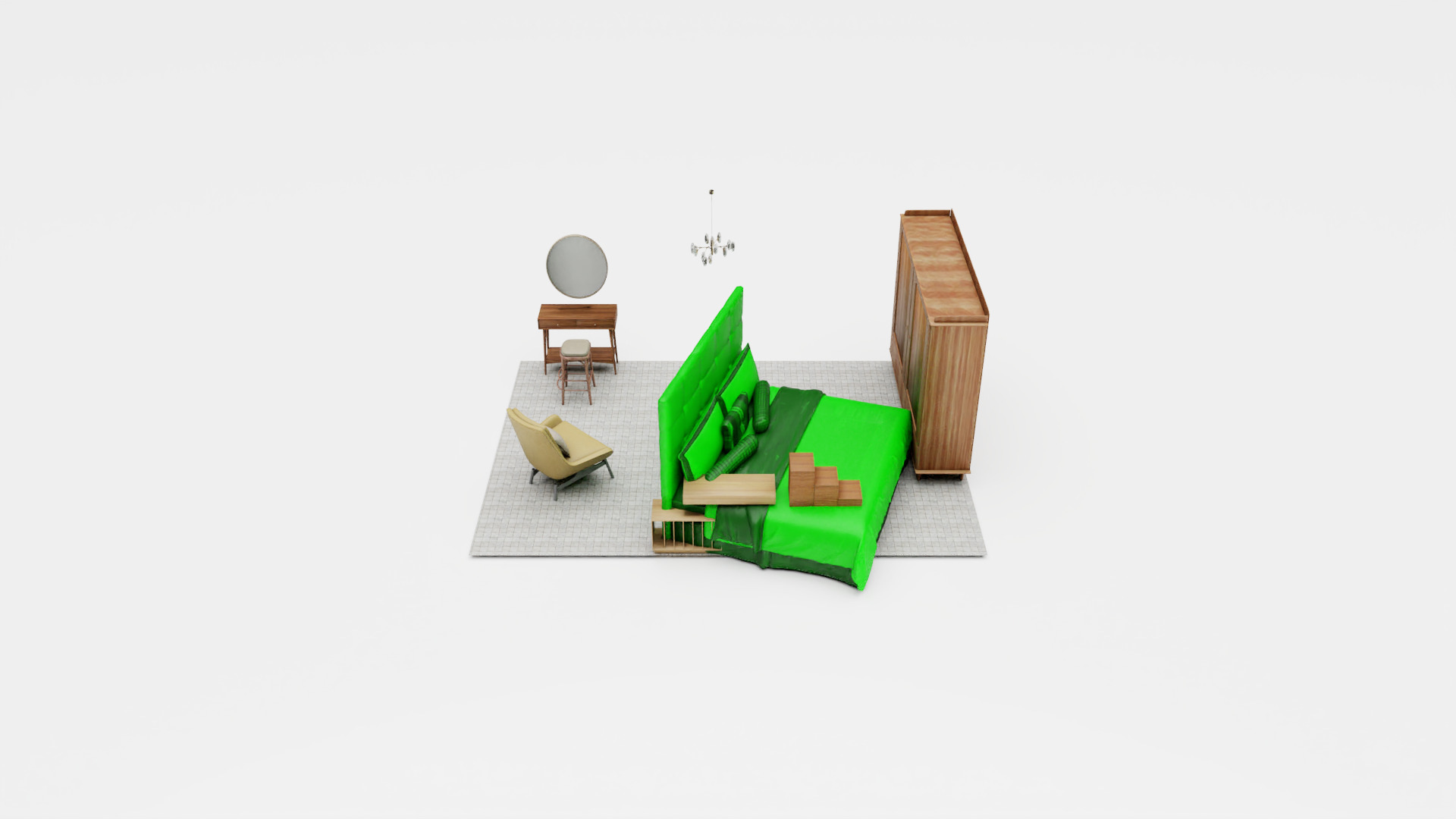}
    \end{subfigure}%
    \begin{subfigure}[b]{0.16\linewidth}
		\centering
		\includegraphics[width=\linewidth, trim=500 270 500 125, clip]{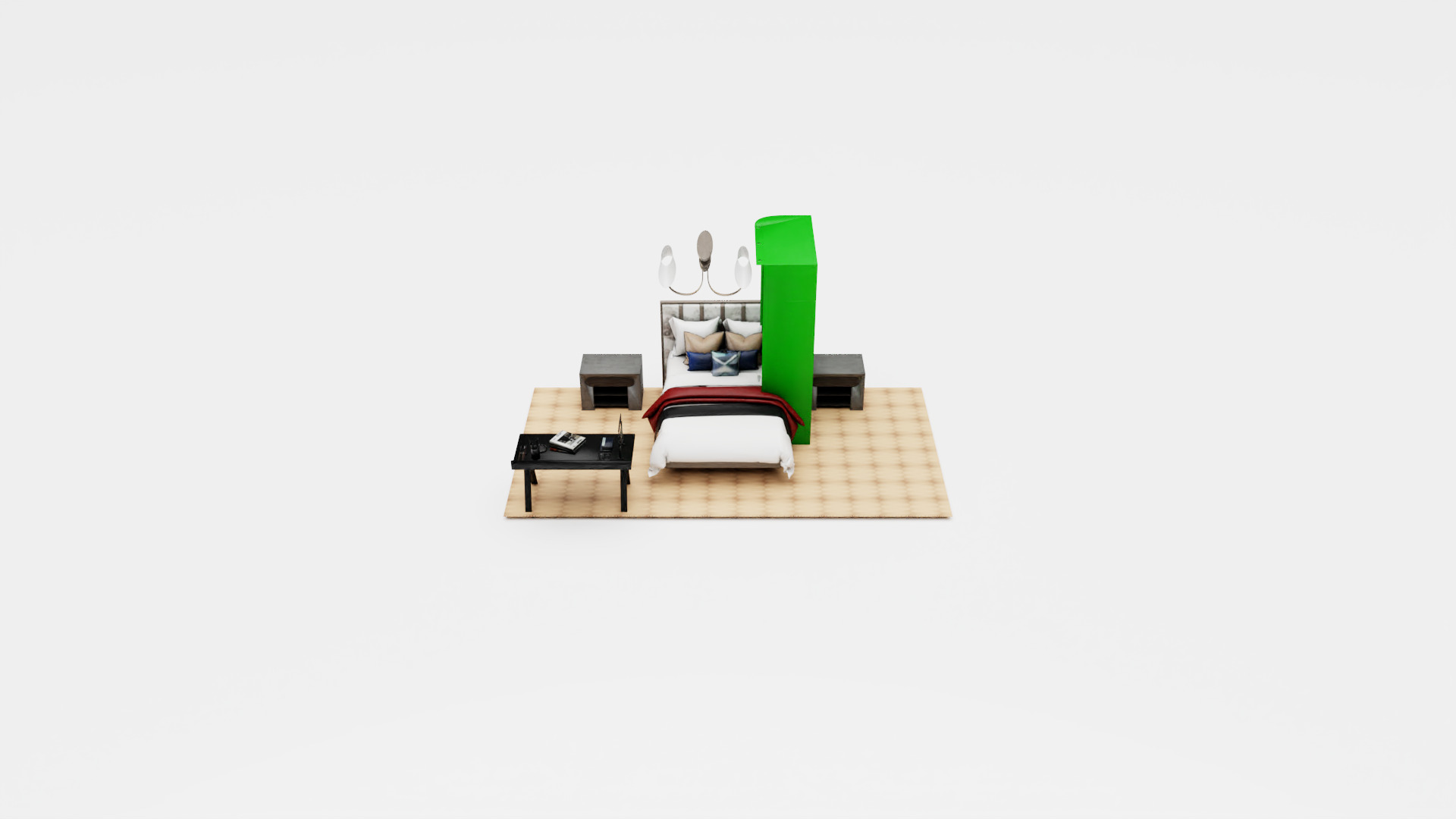}
    \end{subfigure}%
    \begin{subfigure}[b]{0.16\linewidth}
    \centering
    \includegraphics[width=\linewidth, trim=200 0 300 20, clip]{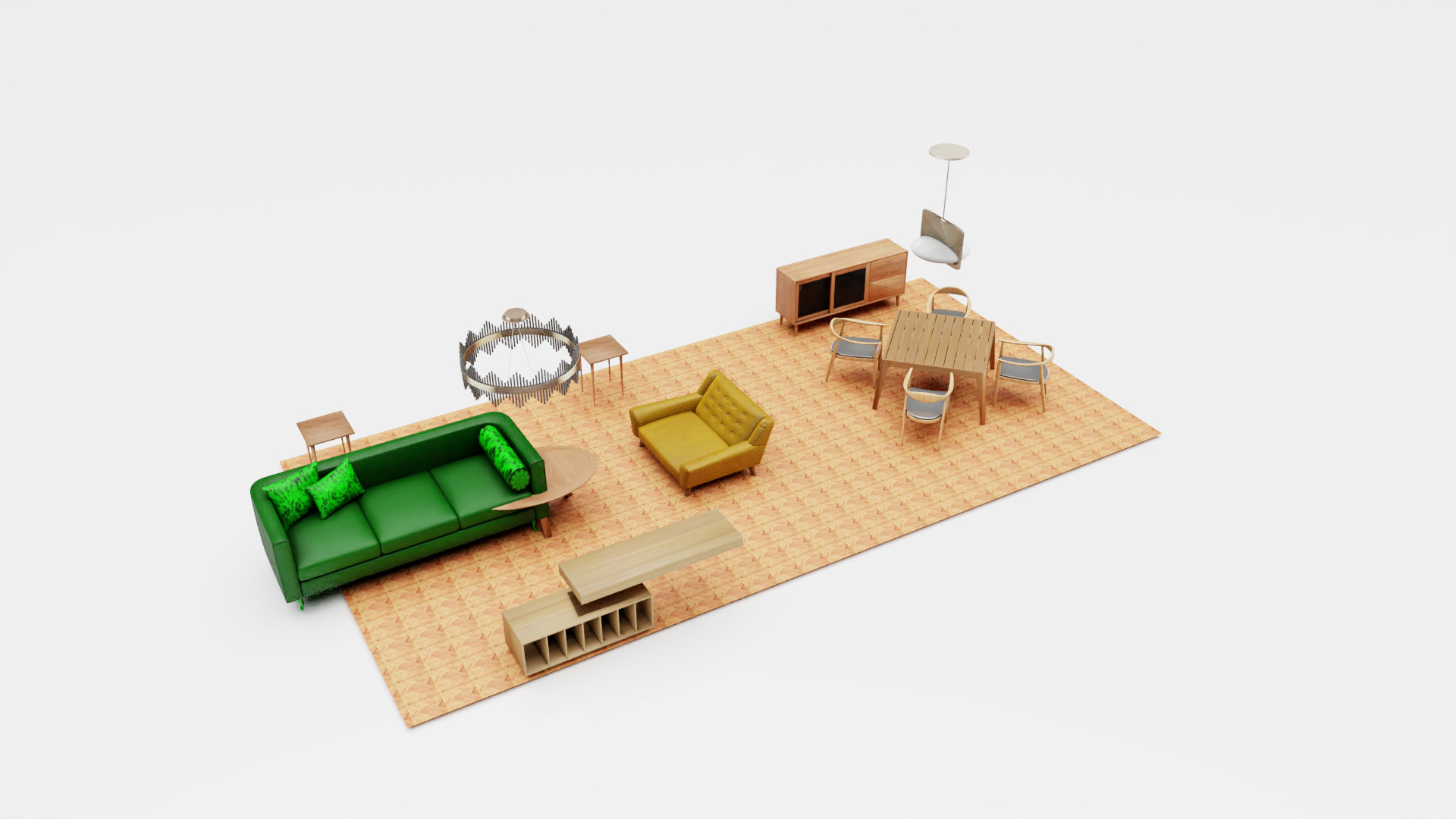}
    \end{subfigure}%
    \begin{subfigure}[b]{0.16\linewidth}
		\centering
		\includegraphics[width=\linewidth, trim=500 270 500 125, clip]{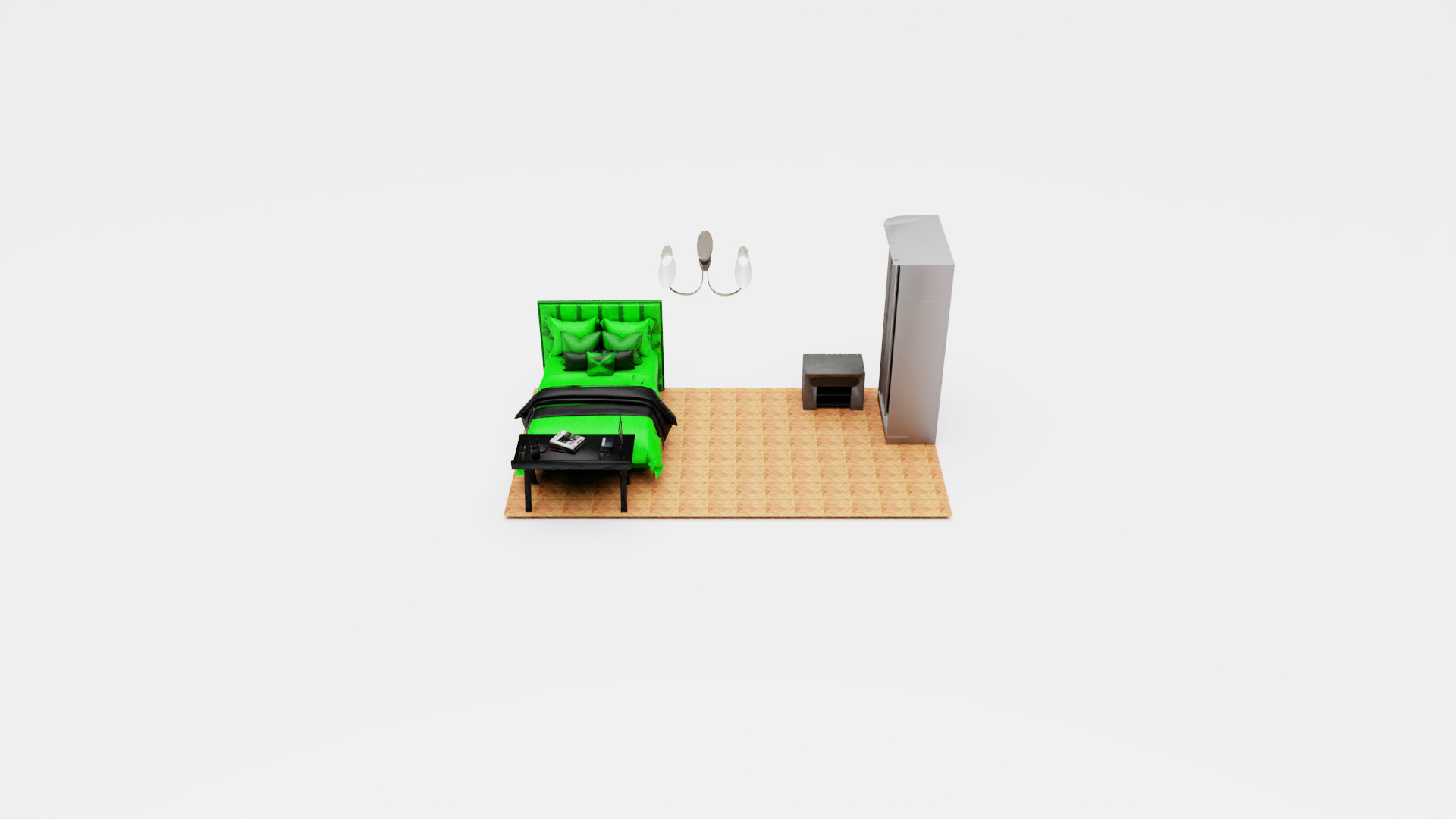}
    \end{subfigure}%
    \begin{subfigure}[b]{0.16\linewidth}
    \centering
    \includegraphics[width=\linewidth, trim=400 0 300 170, clip]{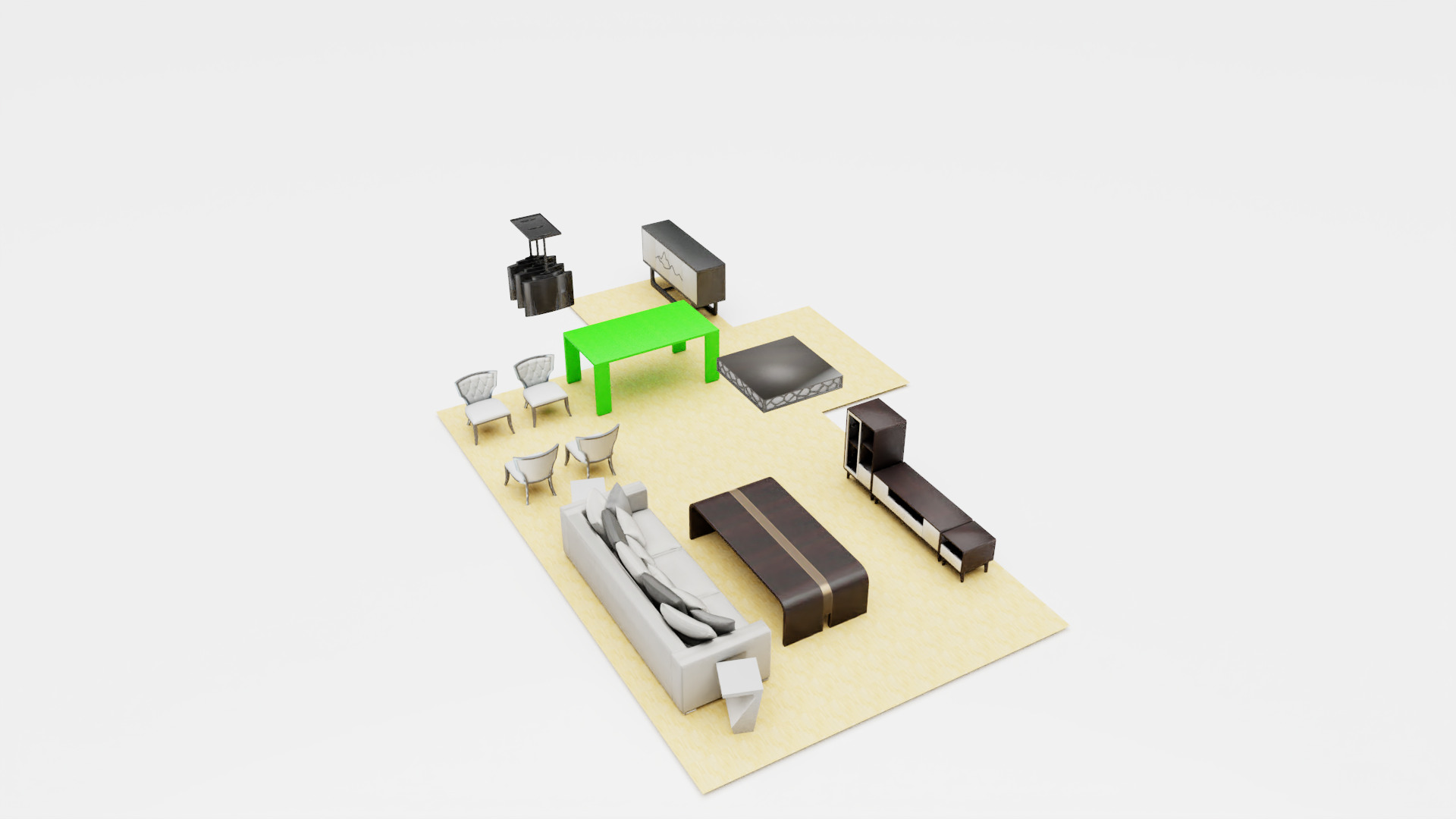}
    \end{subfigure}%
    \begin{subfigure}[b]{0.16\linewidth}
		\centering
		\includegraphics[width=\linewidth, trim=500 270 500 125, clip]{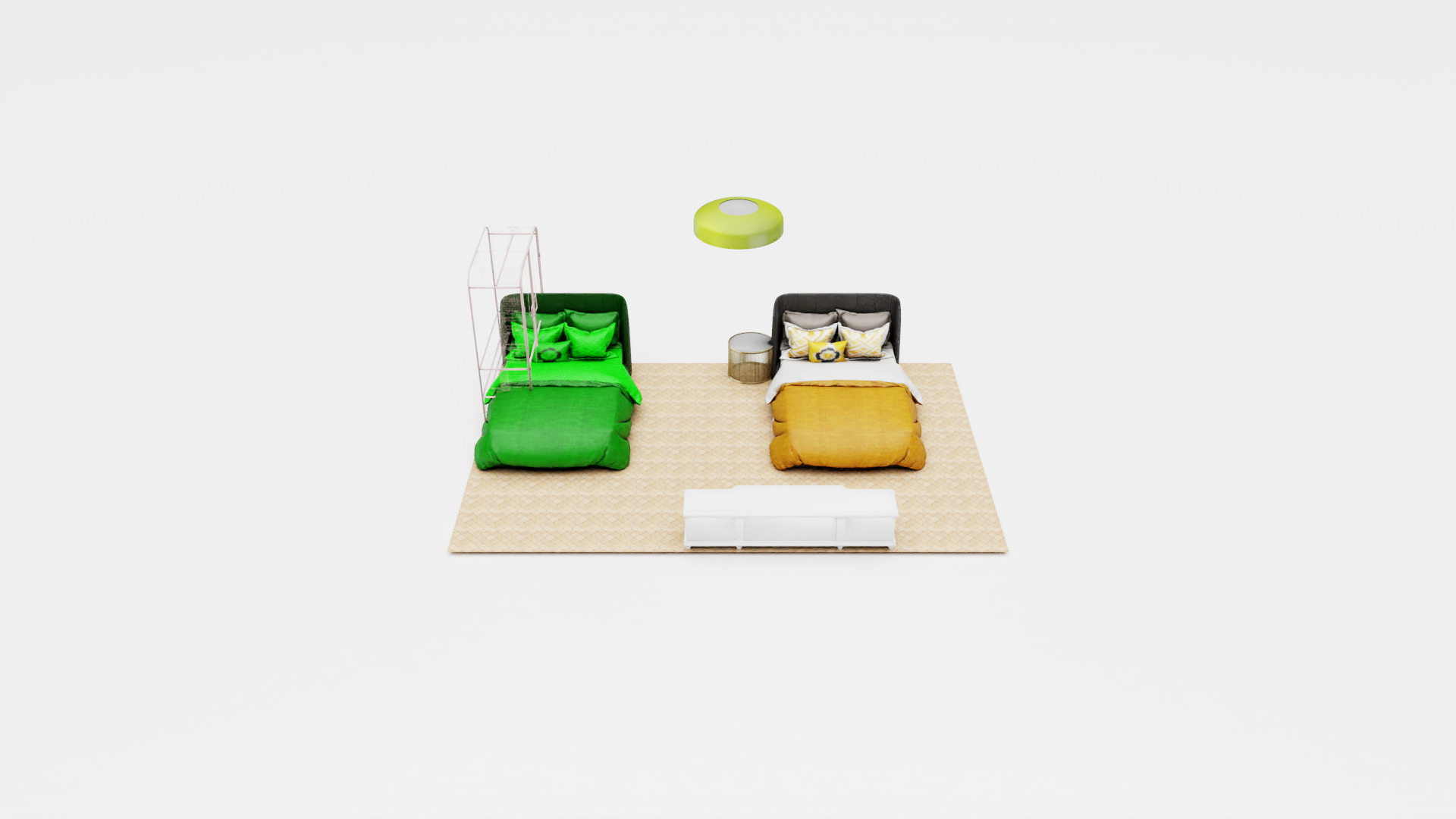}
    \end{subfigure}%
    \hfill%
    \vspace{-1.2em}
    \vskip\baselineskip%
    \hfill%
    \begin{subfigure}[b]{0.16\linewidth}
    \centering
    \includegraphics[width=\linewidth, trim=500 270 500 125, clip]{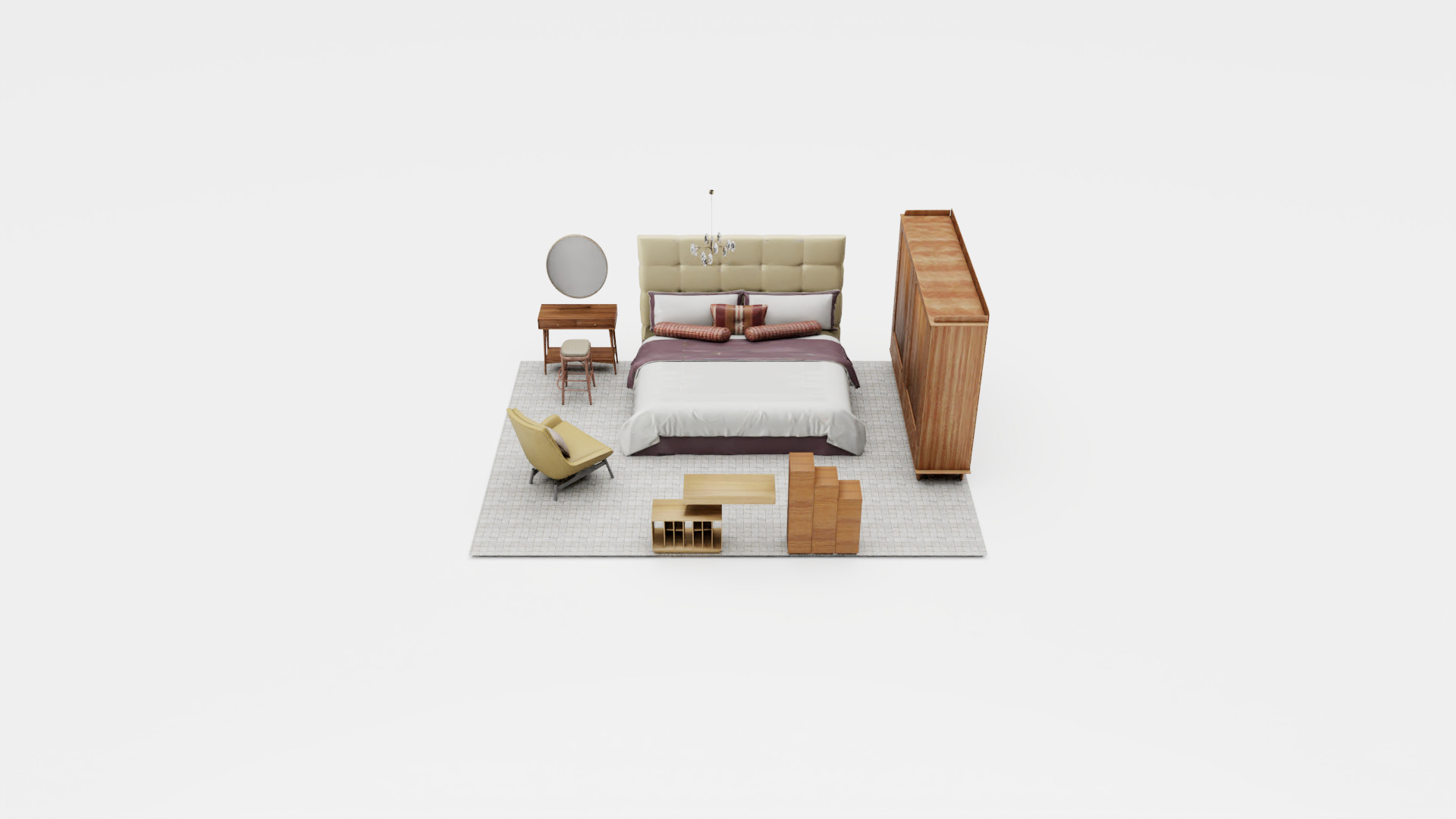}
    \end{subfigure}%
    \begin{subfigure}[b]{0.16\linewidth}
		\centering
		\includegraphics[width=\linewidth, trim=500 270 500 125, clip]{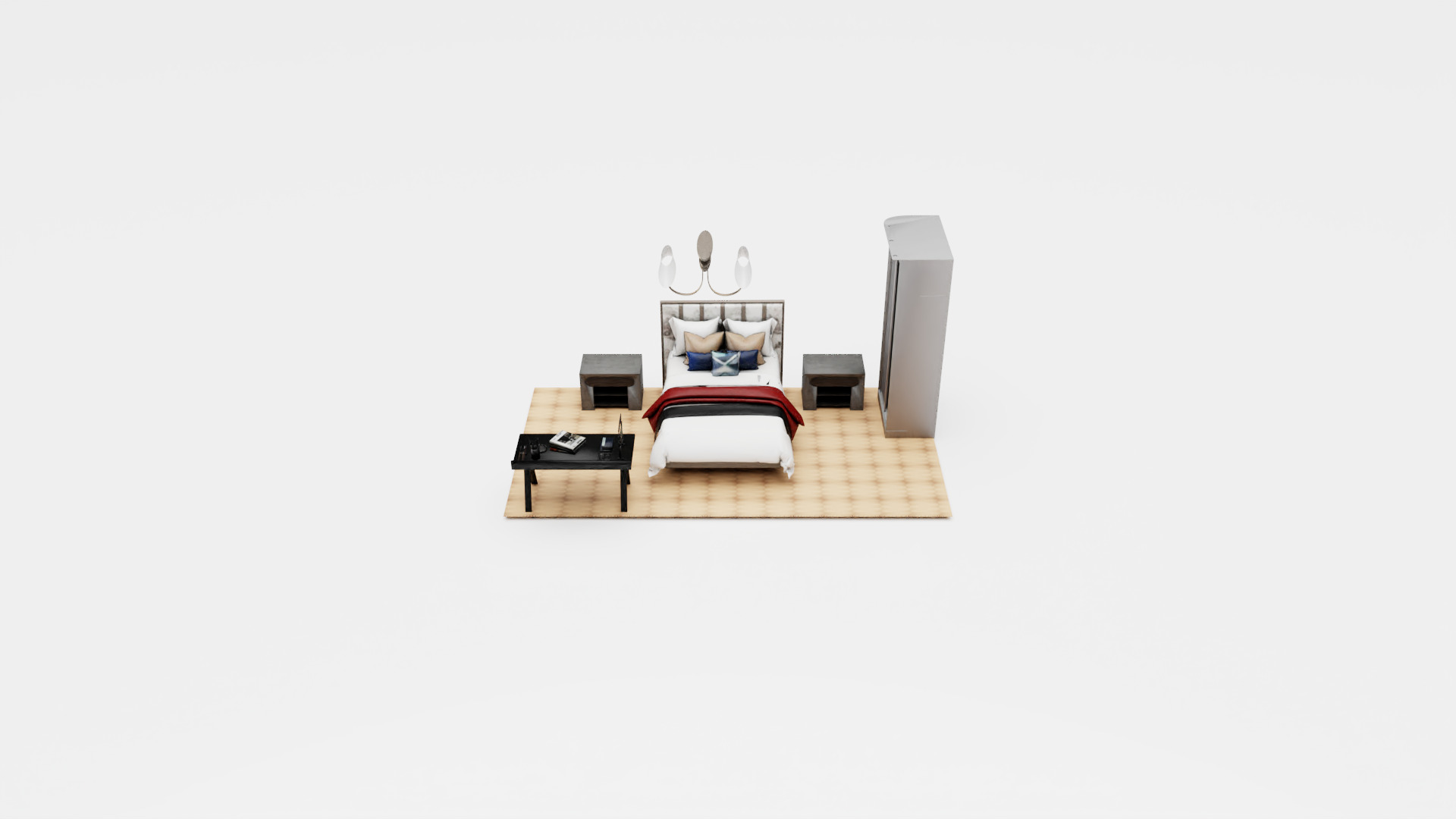}
    \end{subfigure}%
    \begin{subfigure}[b]{0.16\linewidth}
    \centering
    \includegraphics[width=\linewidth, trim=200 0 300 20, clip]{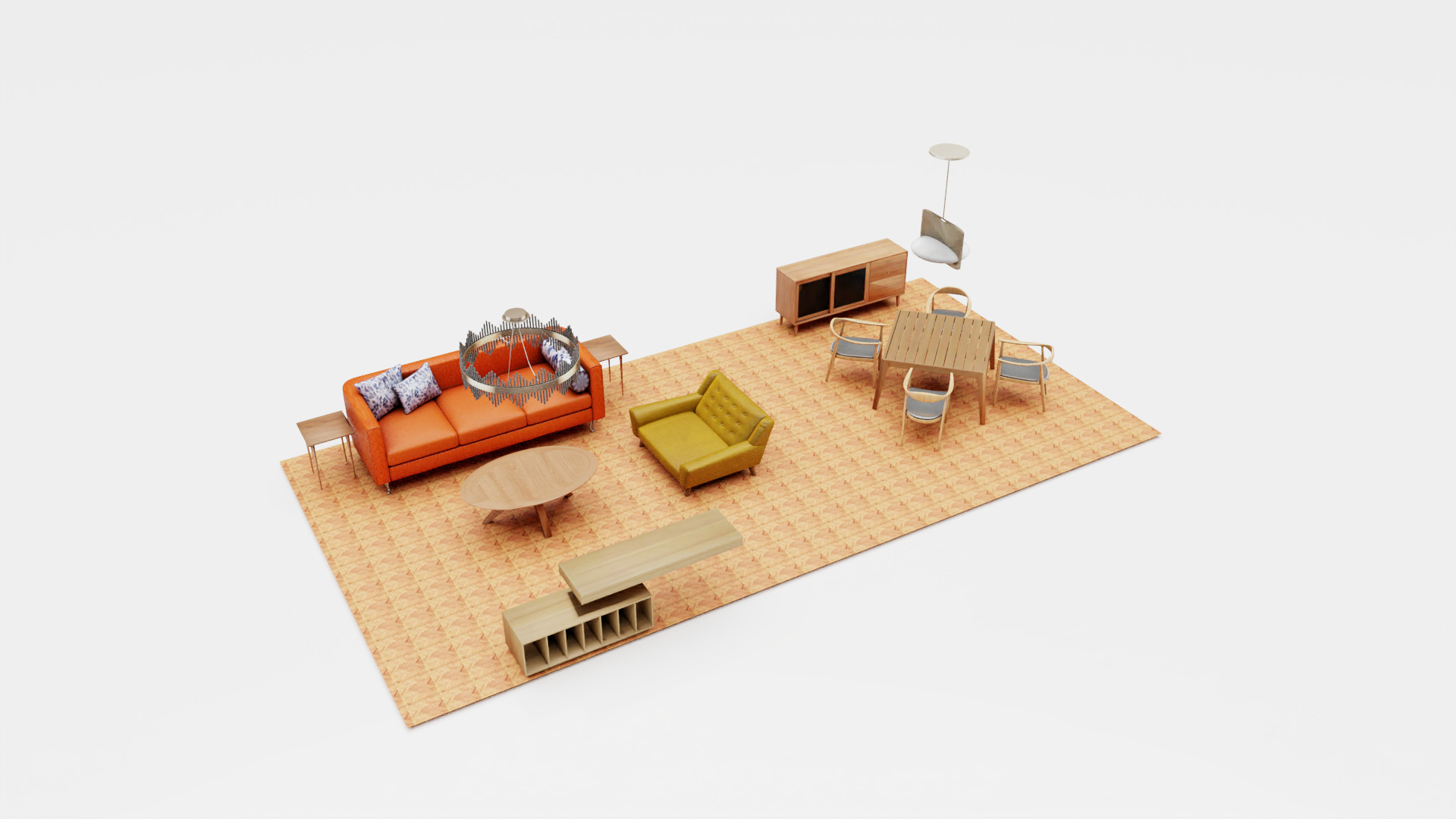}
    \end{subfigure}%
    \begin{subfigure}[b]{0.16\linewidth}
		\centering
		\includegraphics[width=\linewidth, trim=500 270 500 125, clip]{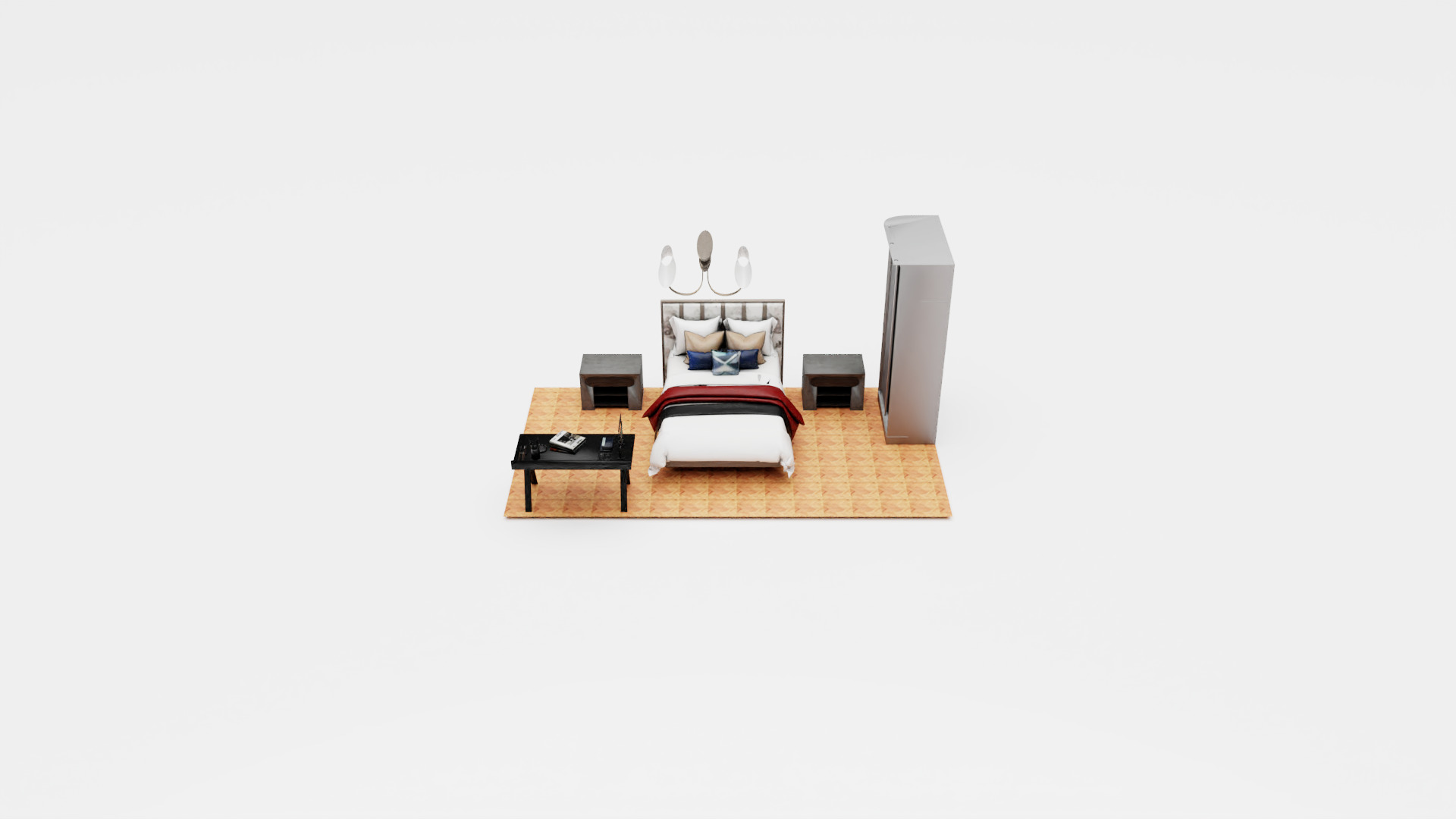}
    \end{subfigure}%
    \begin{subfigure}[b]{0.16\linewidth}
    \centering
    \includegraphics[width=\linewidth, trim=400 0 300 170, clip]{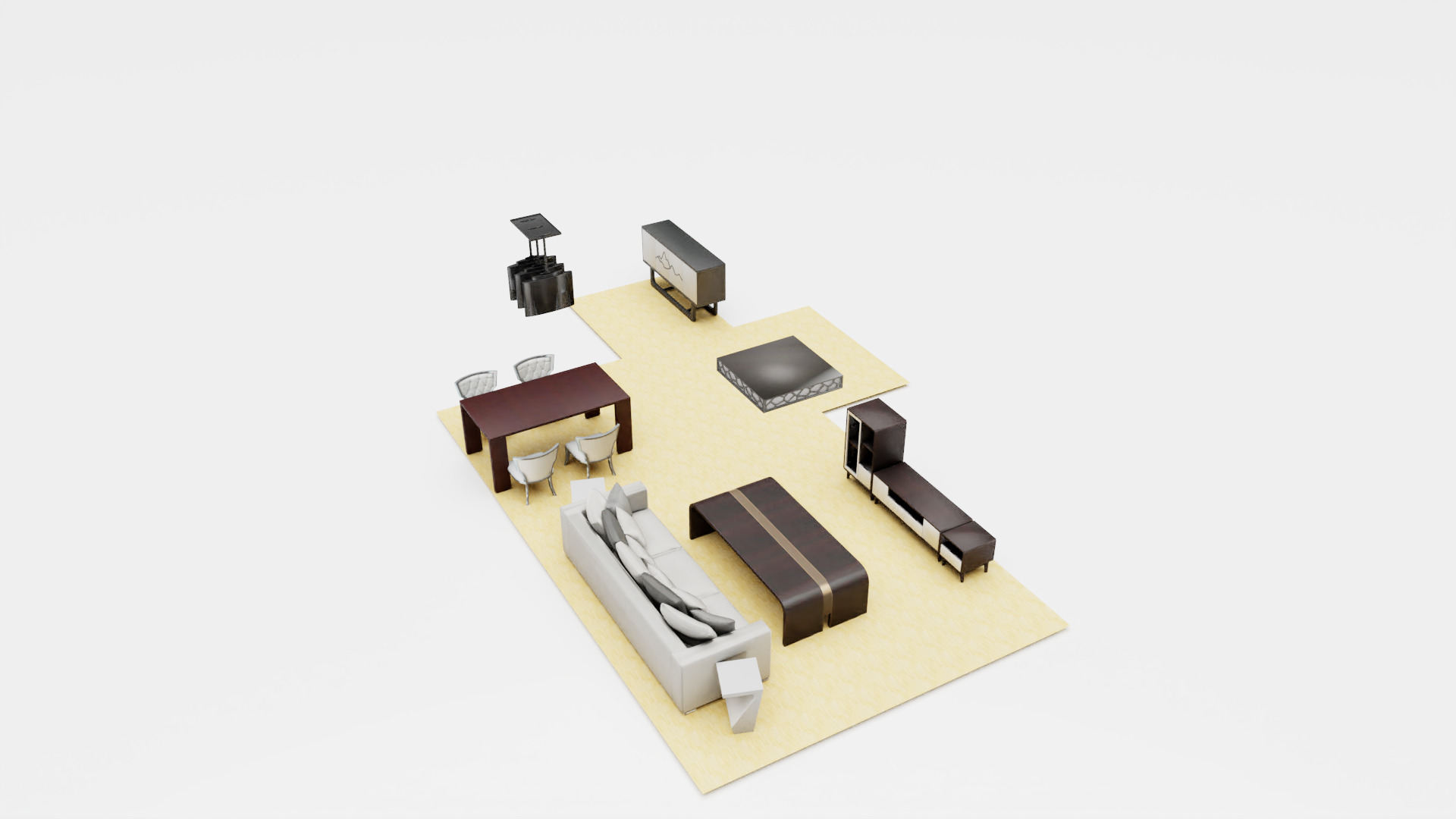}
    \end{subfigure}    \begin{subfigure}[b]{0.16\linewidth}
		\centering
		\includegraphics[width=\linewidth, trim=500 270 500 125, clip]{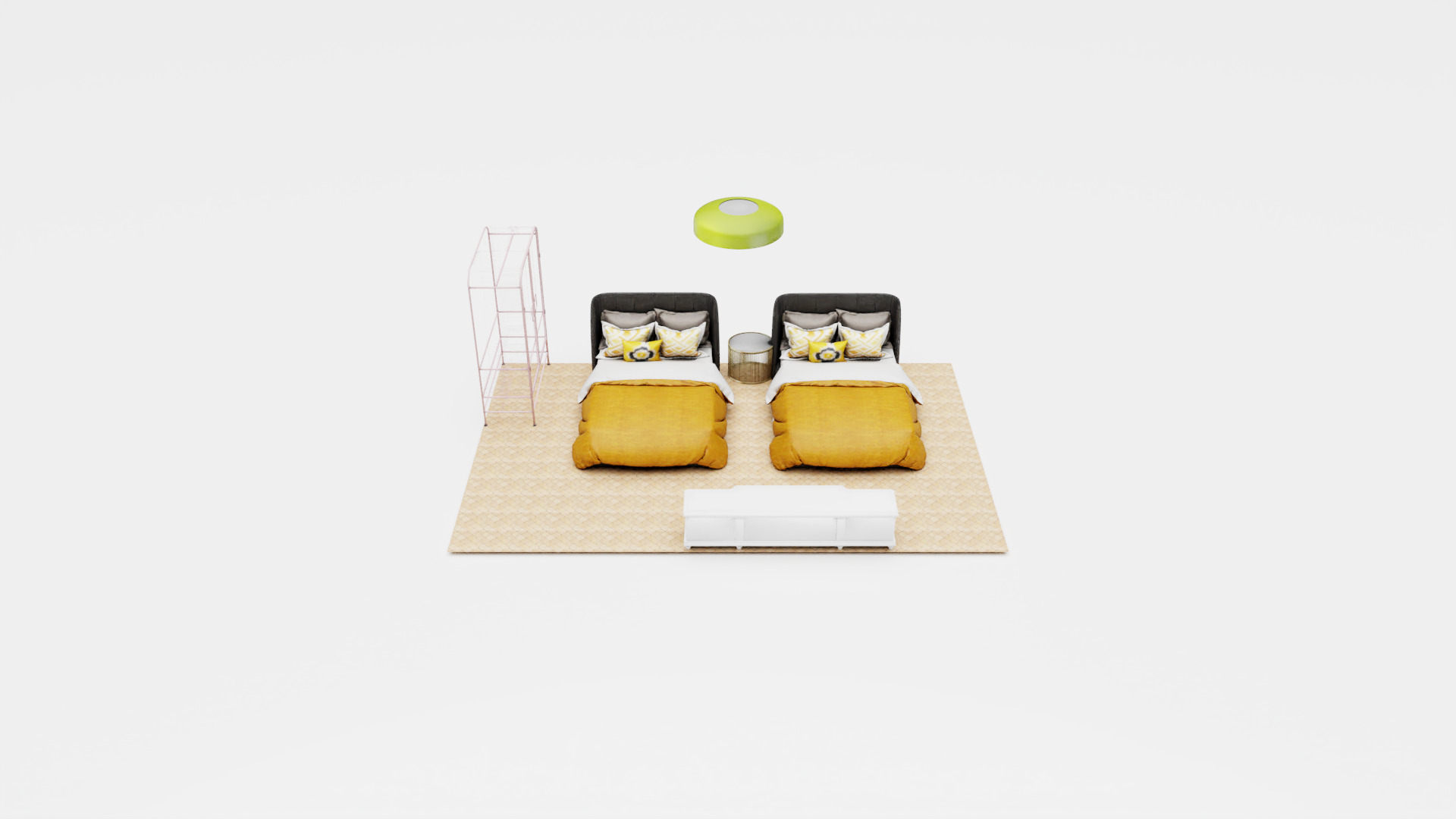}
    \end{subfigure}%
    \hfill%
    \vspace{-1.2em}
    \vskip\baselineskip%
    \caption{\small {\bf Failure Case Detection and Correction}. We use a partial room with unnatural object arrangements. Our model identifies the problematic objects (first row, in green) and relocates them into meaningful positions.}
    \label{fig:failure_cases_correction}
    \vspace{-1.5em}
\end{figure}
\boldparagraph{Failure Case Detection and Correction}%
We showcase that our model is able to identify and
correct unnatural object arrangements. Given a scene, we compute the
likelihood of each object, according to our model, conditioned on the other
objects in the scene. We identify problematic objects as those with low likelihood and
sample a new location from our generative model to rearrange it.
We test our model in various scenarios such as overlapping objects, objects
outside the room boundaries and objects in unnatural positions and show
that it successfully identifies problematic objects (highlighted in
green in \figref{fig:failure_cases_correction}) and rearranges them into a more
plausible position. Note that this task cannot be performed by methods that
consider ordering because they assign very low likelihood to common objects
appearing after rare objects \eg beds after cabinets.

\begin{figure}
    \centering
        \vskip\baselineskip%
    \vspace{-0.8em}
    \hfill%
    \begin{subfigure}[b]{0.16\linewidth}
		\centering
		\includegraphics[width=\linewidth, trim=500 200 500 100, clip]{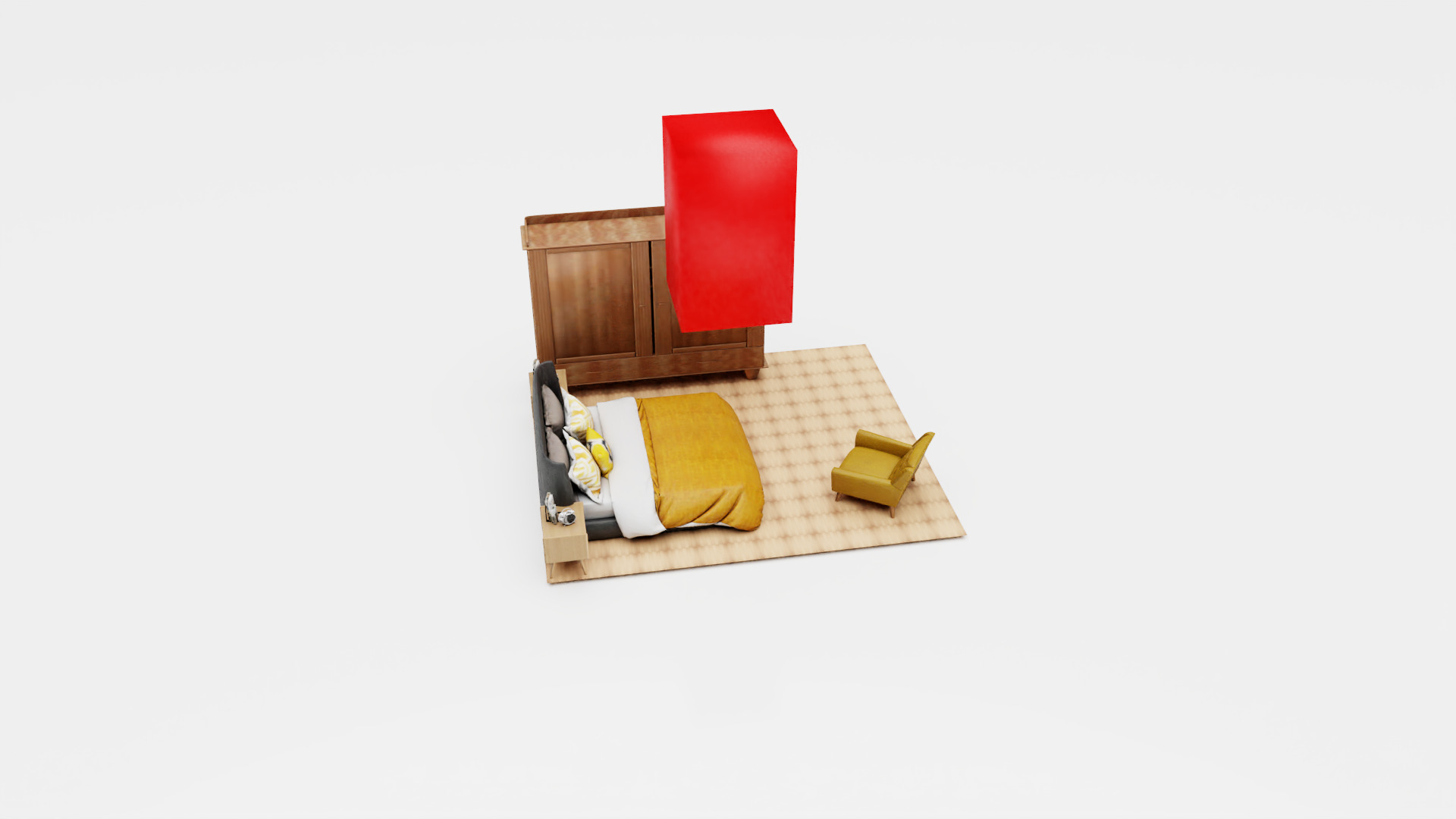}
    \end{subfigure}%
    \begin{subfigure}[b]{0.16\linewidth}
		\centering
		\includegraphics[width=\linewidth, trim=500 200 500 100, clip]{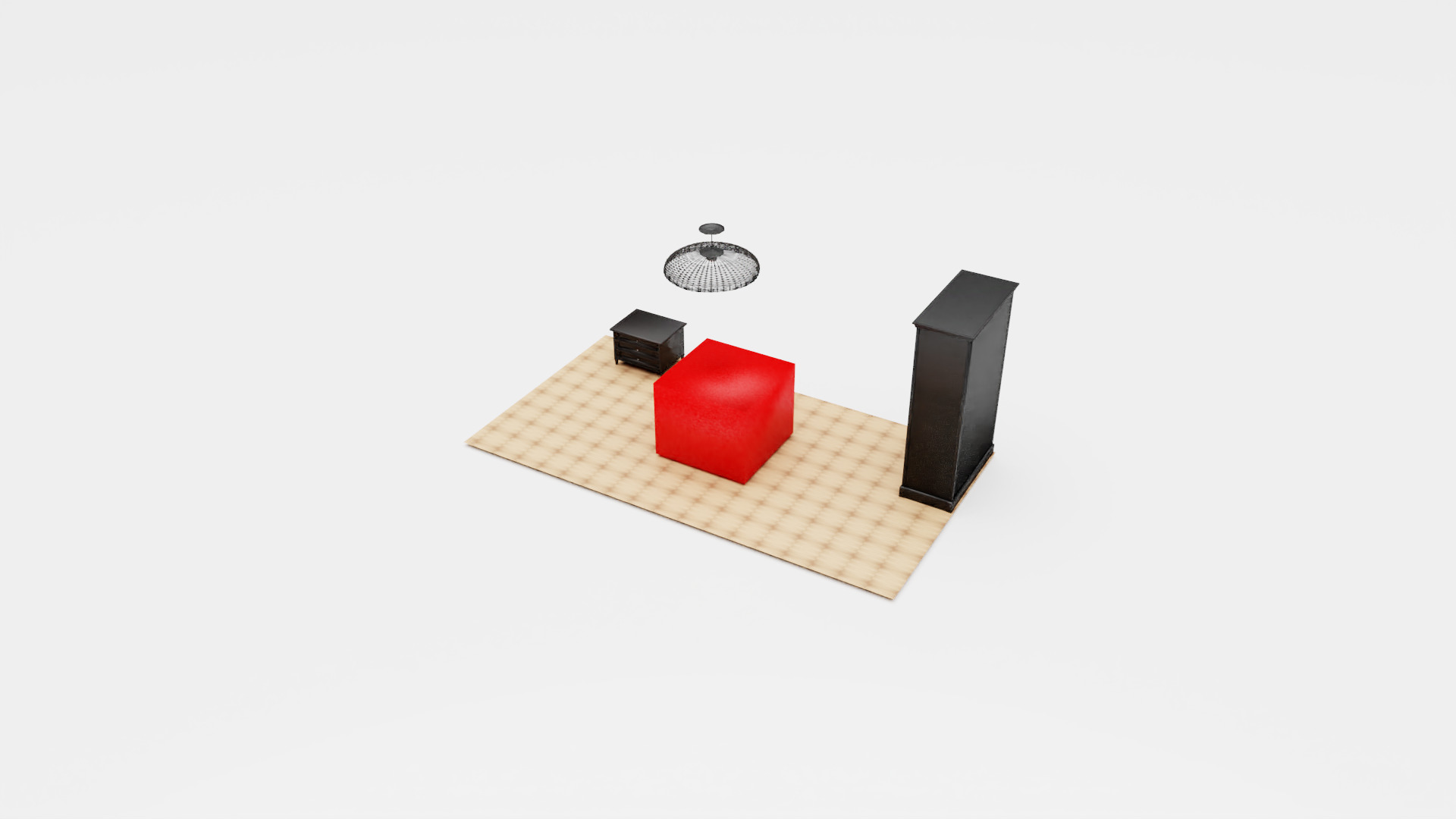}
    \end{subfigure}%
    \begin{subfigure}[b]{0.16\linewidth}
		\centering
		\includegraphics[width=\linewidth, trim=370 100 450 50, clip]{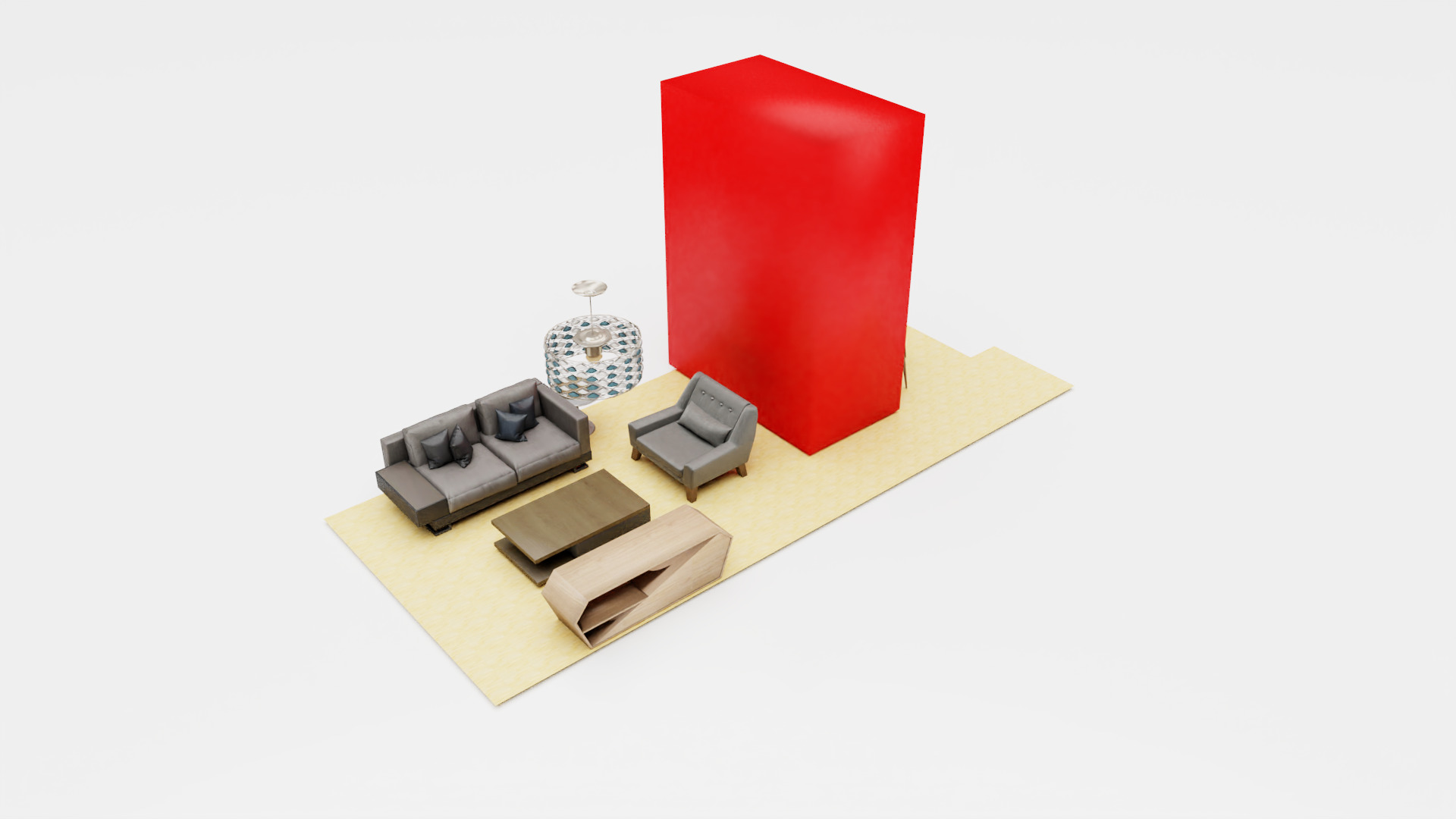}
    \end{subfigure}%
    \begin{subfigure}[b]{0.16\linewidth}
		\centering
		\includegraphics[width=\linewidth, trim=370 0 450 150, clip]{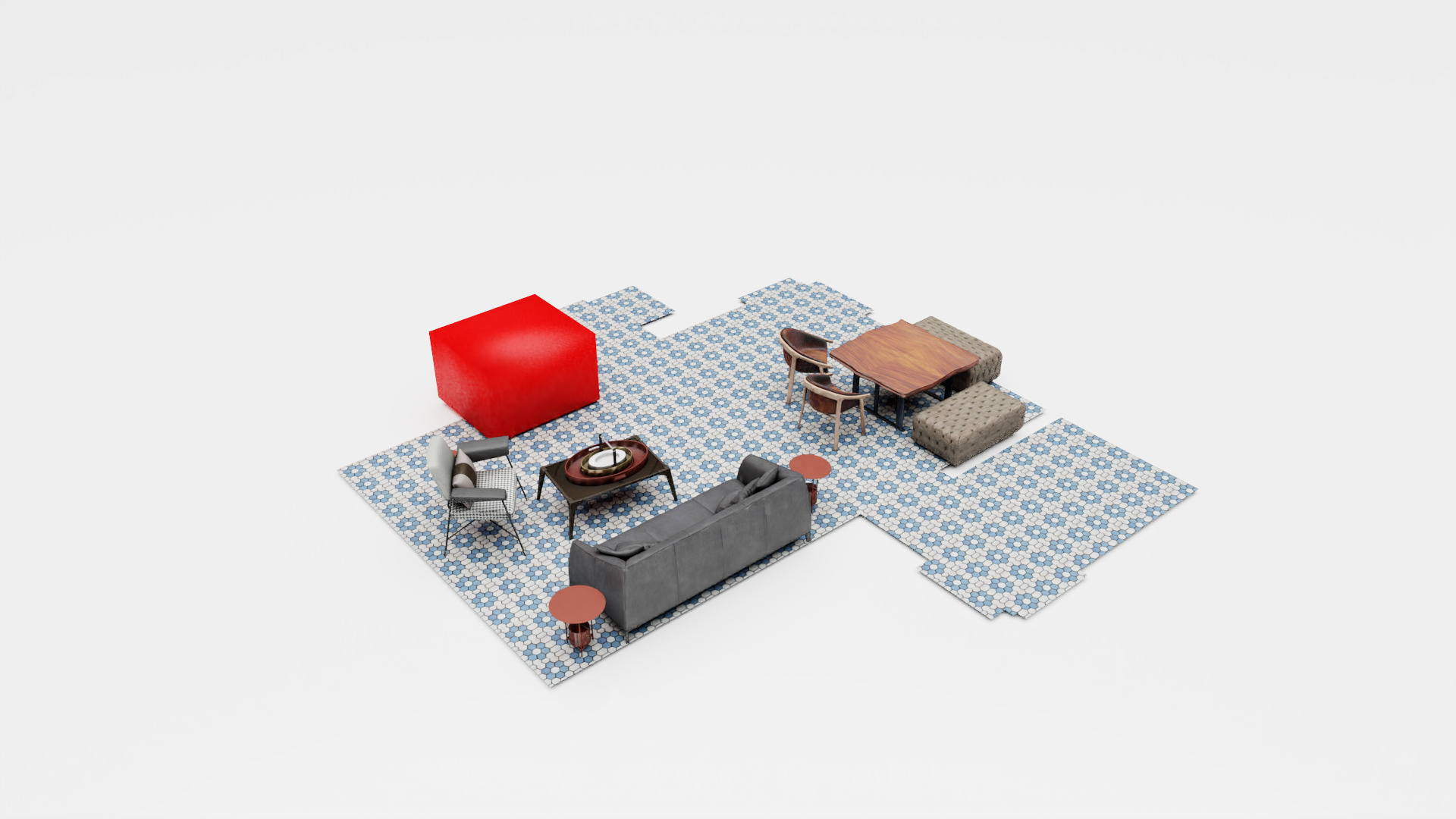}
    \end{subfigure}%
    \begin{subfigure}[b]{0.16\linewidth}
		\centering
		\includegraphics[width=\linewidth, trim=500 200 500 100, clip]{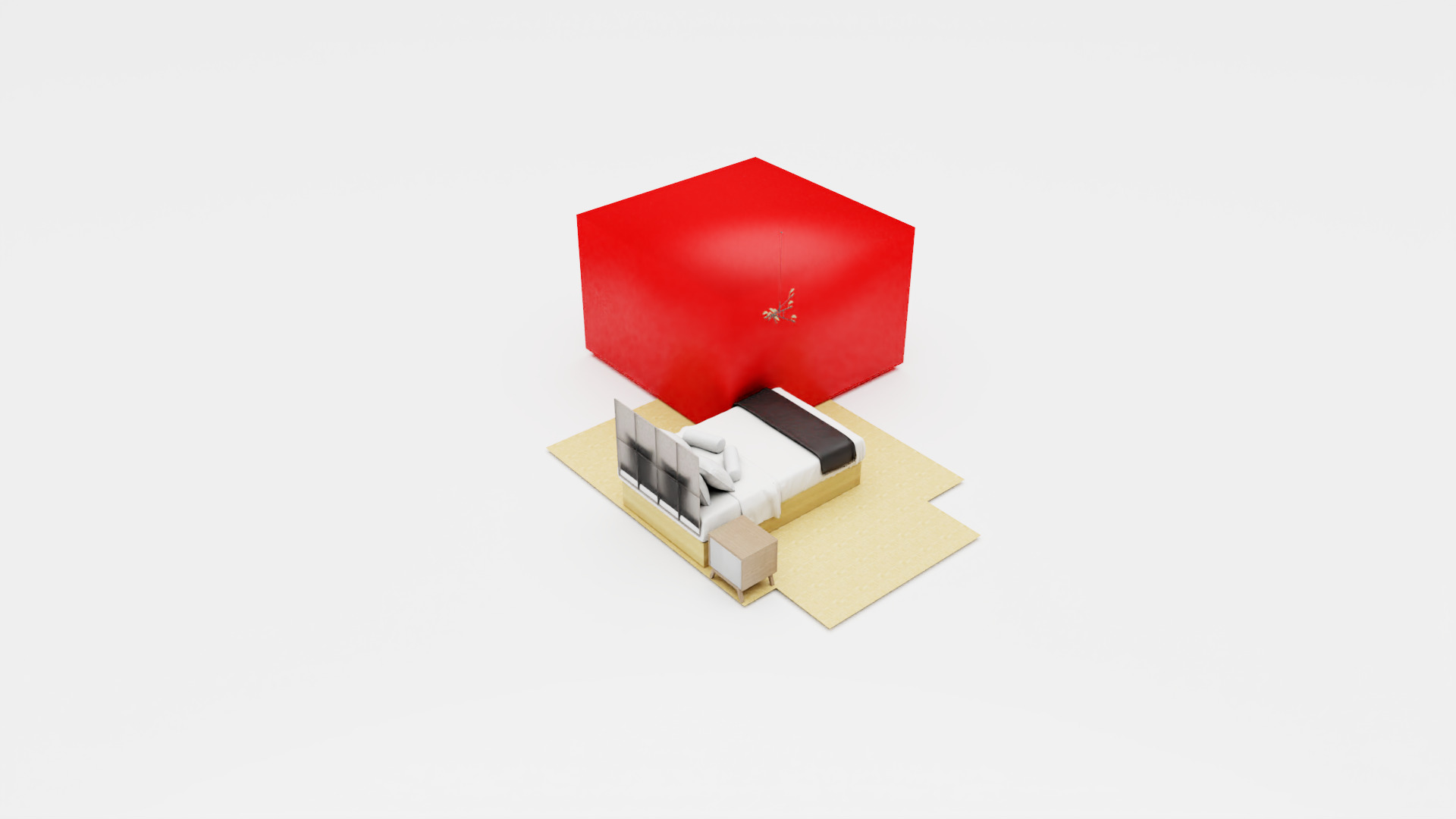}
    \end{subfigure}%
    \begin{subfigure}[b]{0.16\linewidth}
		\centering
		\includegraphics[width=\linewidth, trim=500 200 500 100, clip]{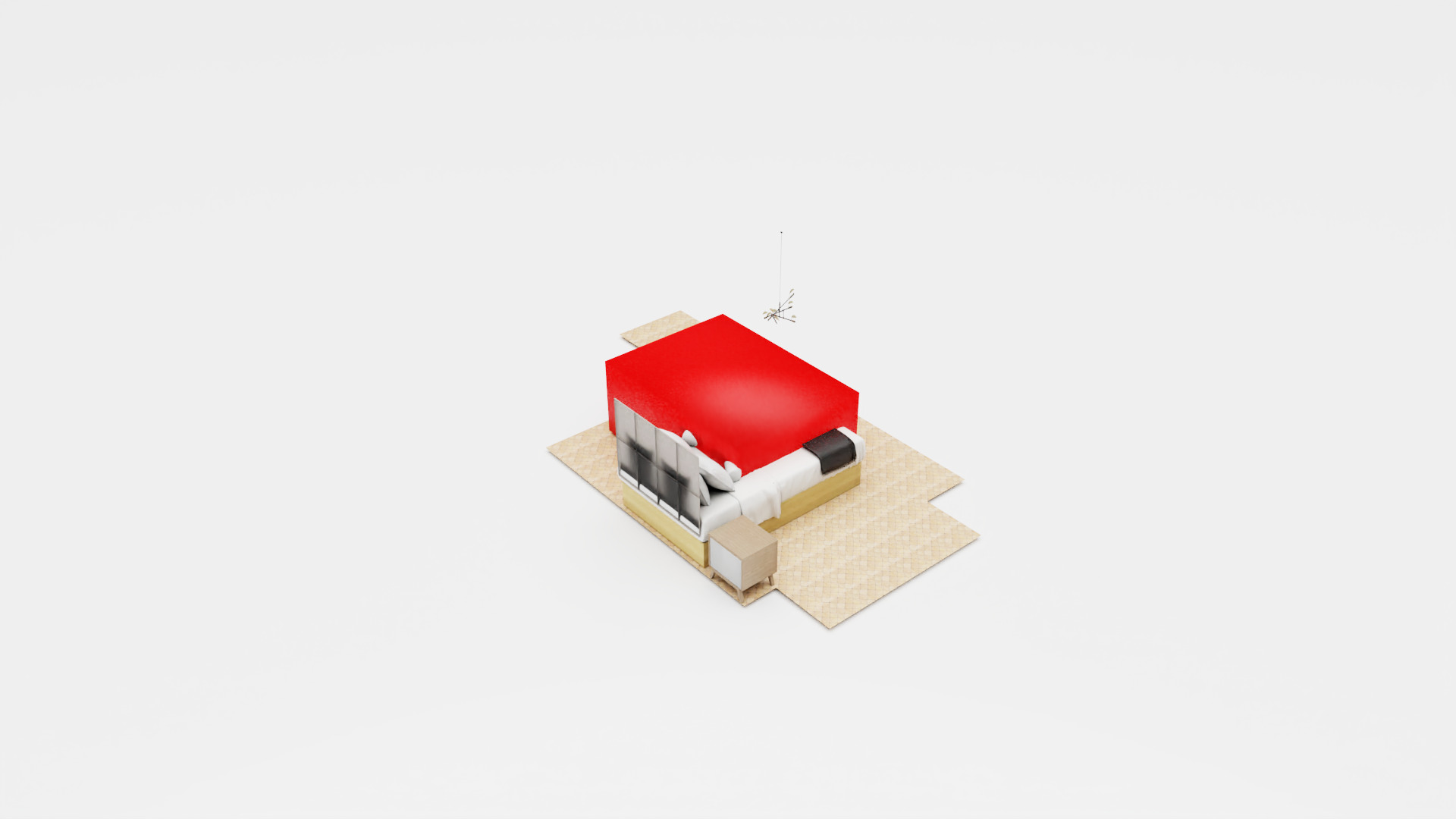}
    \end{subfigure}%
    \hfill%
    \vspace{-1.2em}
    \vskip\baselineskip%
    \begin{subfigure}[b]{0.16\linewidth}
		\centering
		\includegraphics[width=\linewidth, trim=500 200 500 100, clip]{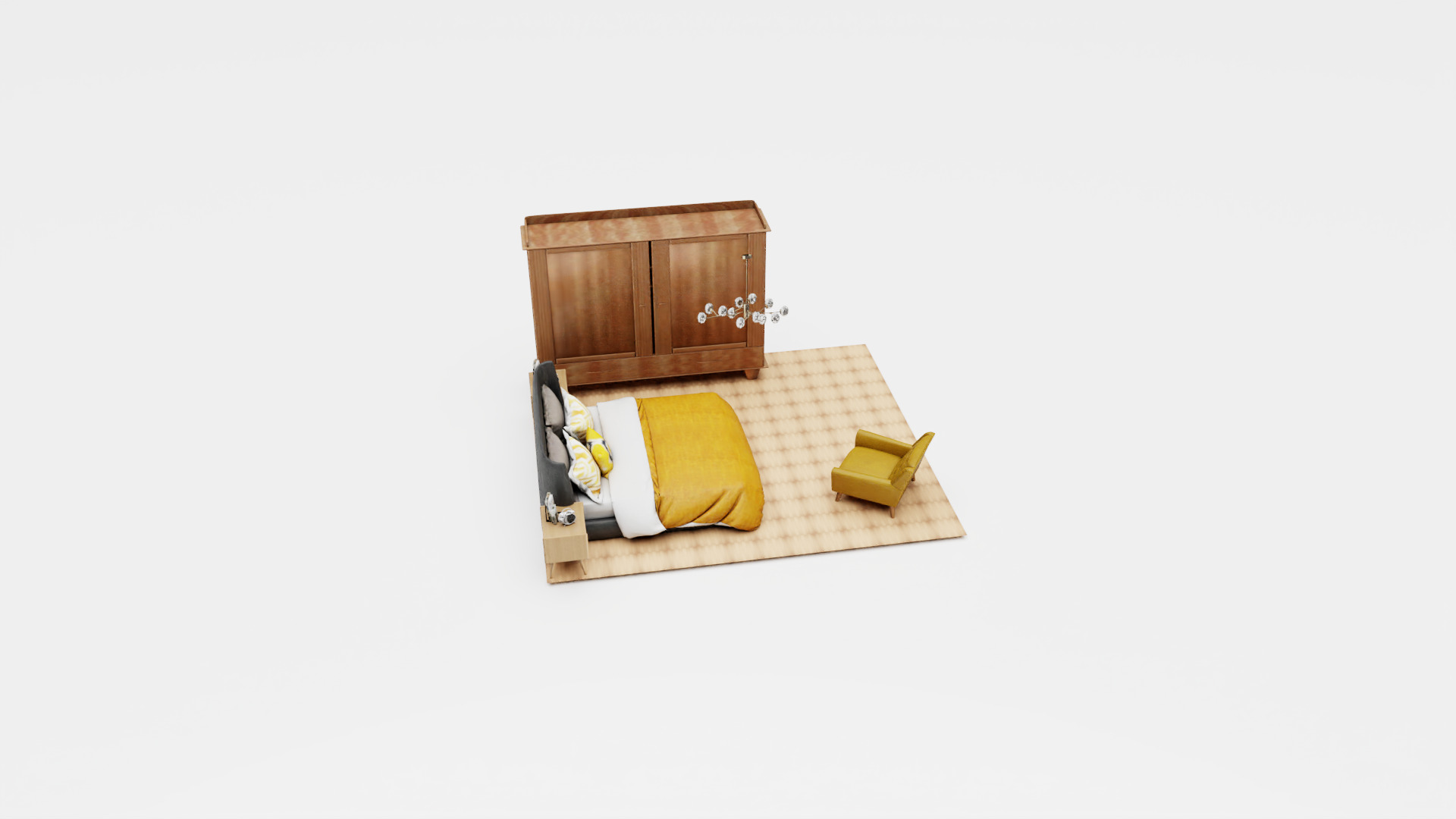}
    \end{subfigure}%
    \begin{subfigure}[b]{0.16\linewidth}
		\centering
		\includegraphics[width=\linewidth, trim=500 200 500 100, clip]{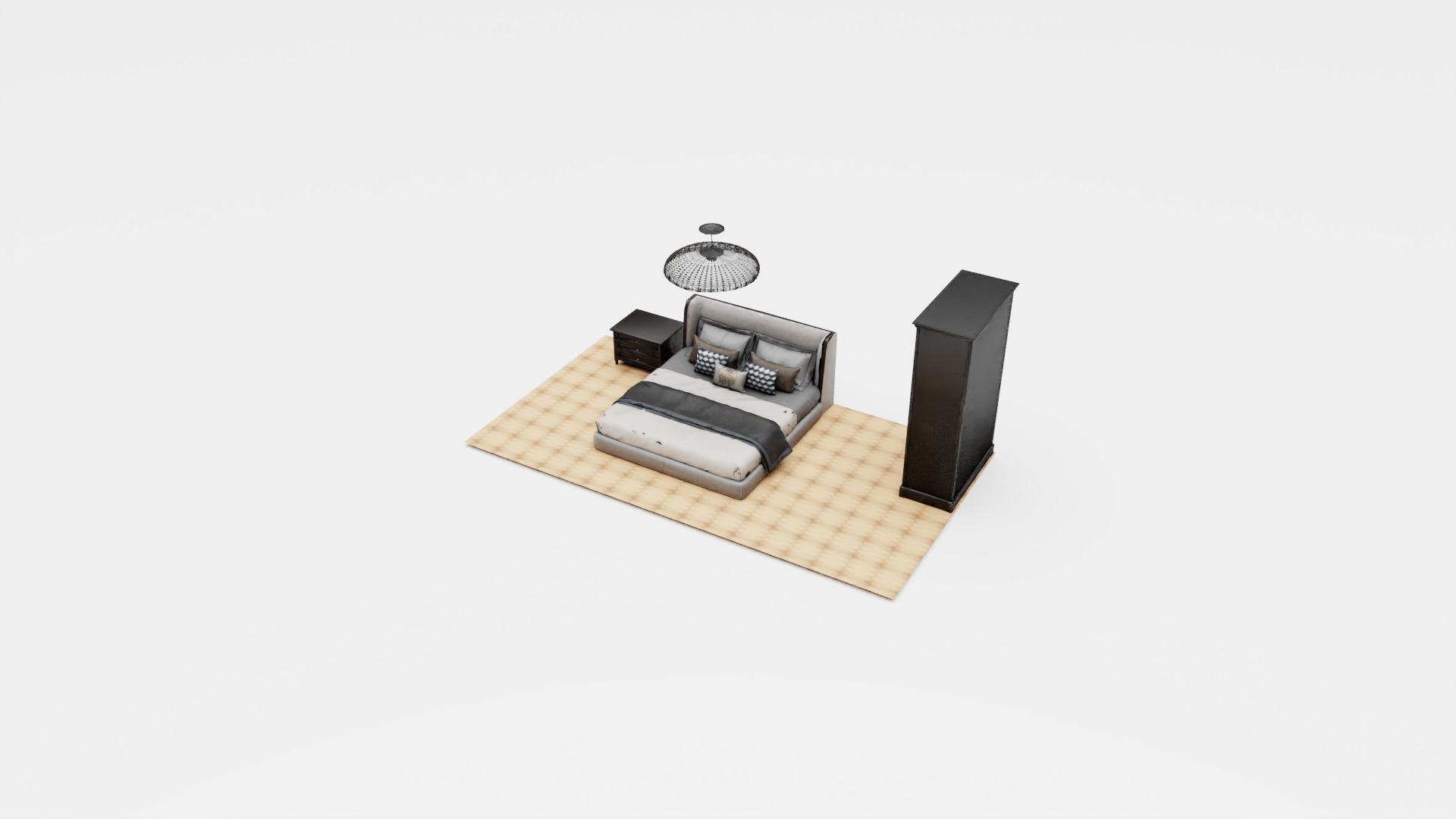}
    \end{subfigure}%
    \begin{subfigure}[b]{0.16\linewidth}
		\centering
		\includegraphics[width=\linewidth, trim=370 100 450 50, clip]{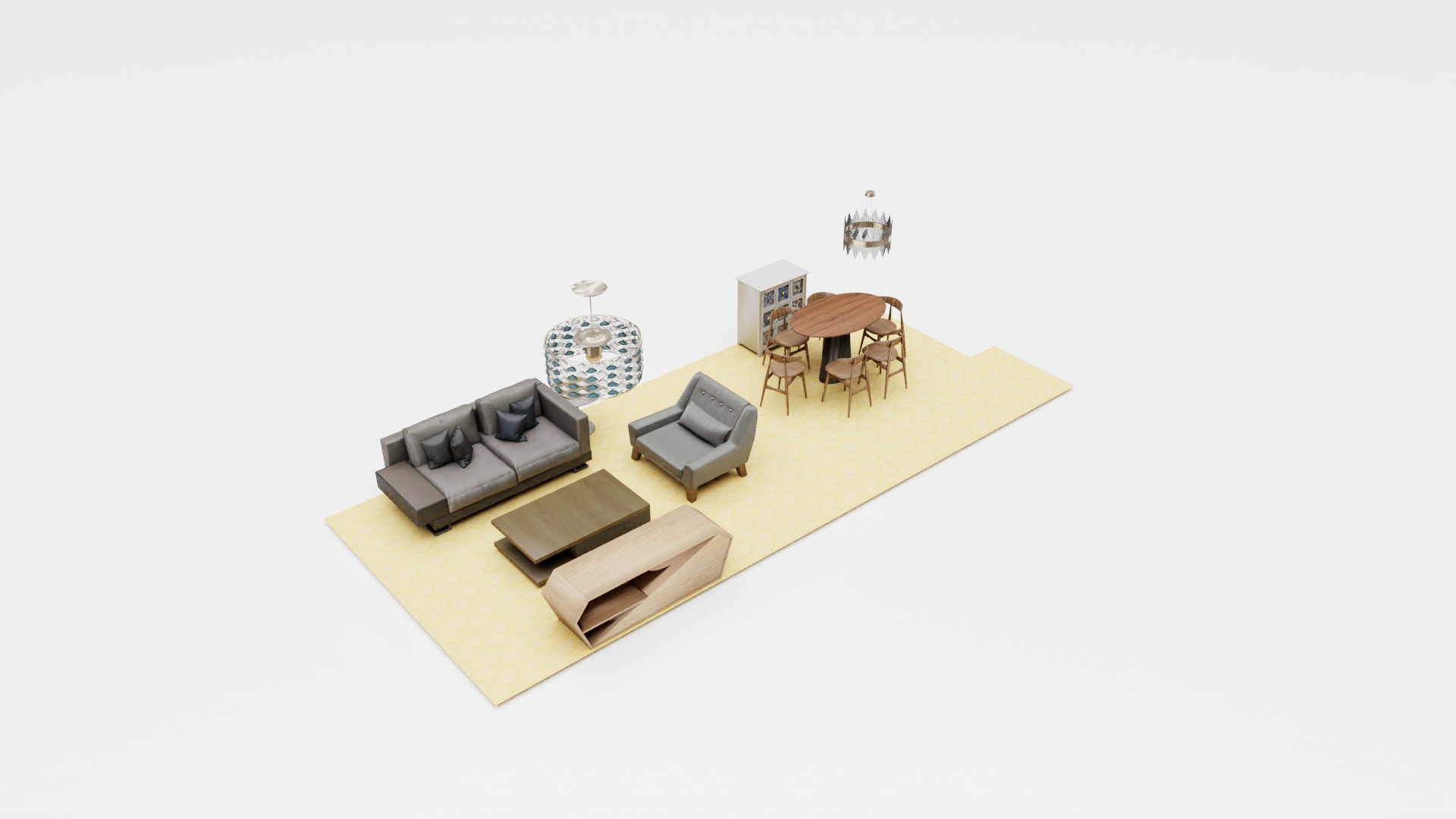}
    \end{subfigure}%
    \begin{subfigure}[b]{0.16\linewidth}
		\centering
		\includegraphics[width=\linewidth, trim=370 0 450 150, clip]{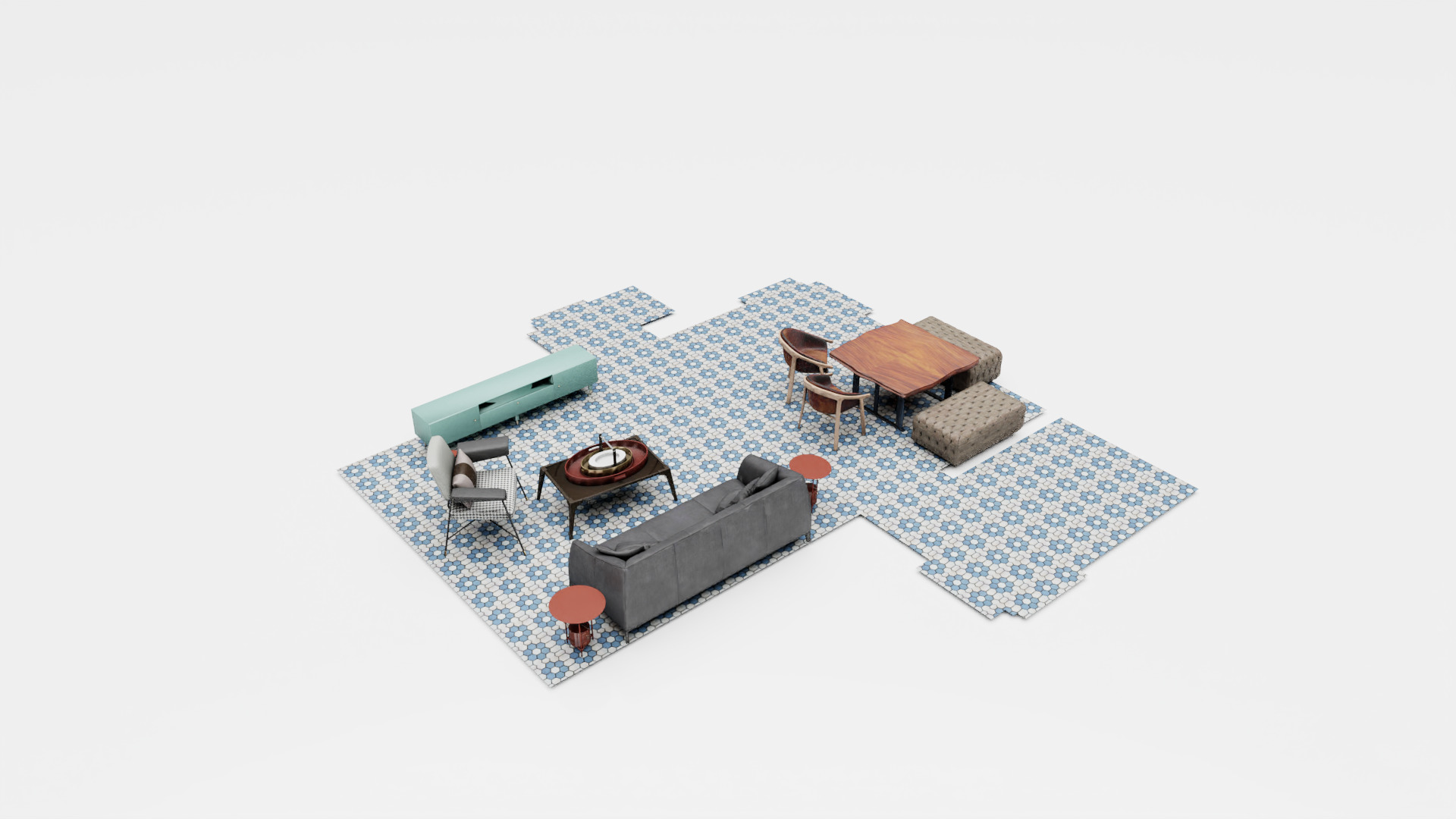}
    \end{subfigure}%
    \begin{subfigure}[b]{0.16\linewidth}
		\centering
		\includegraphics[width=\linewidth, trim=500 200 500 100, clip]{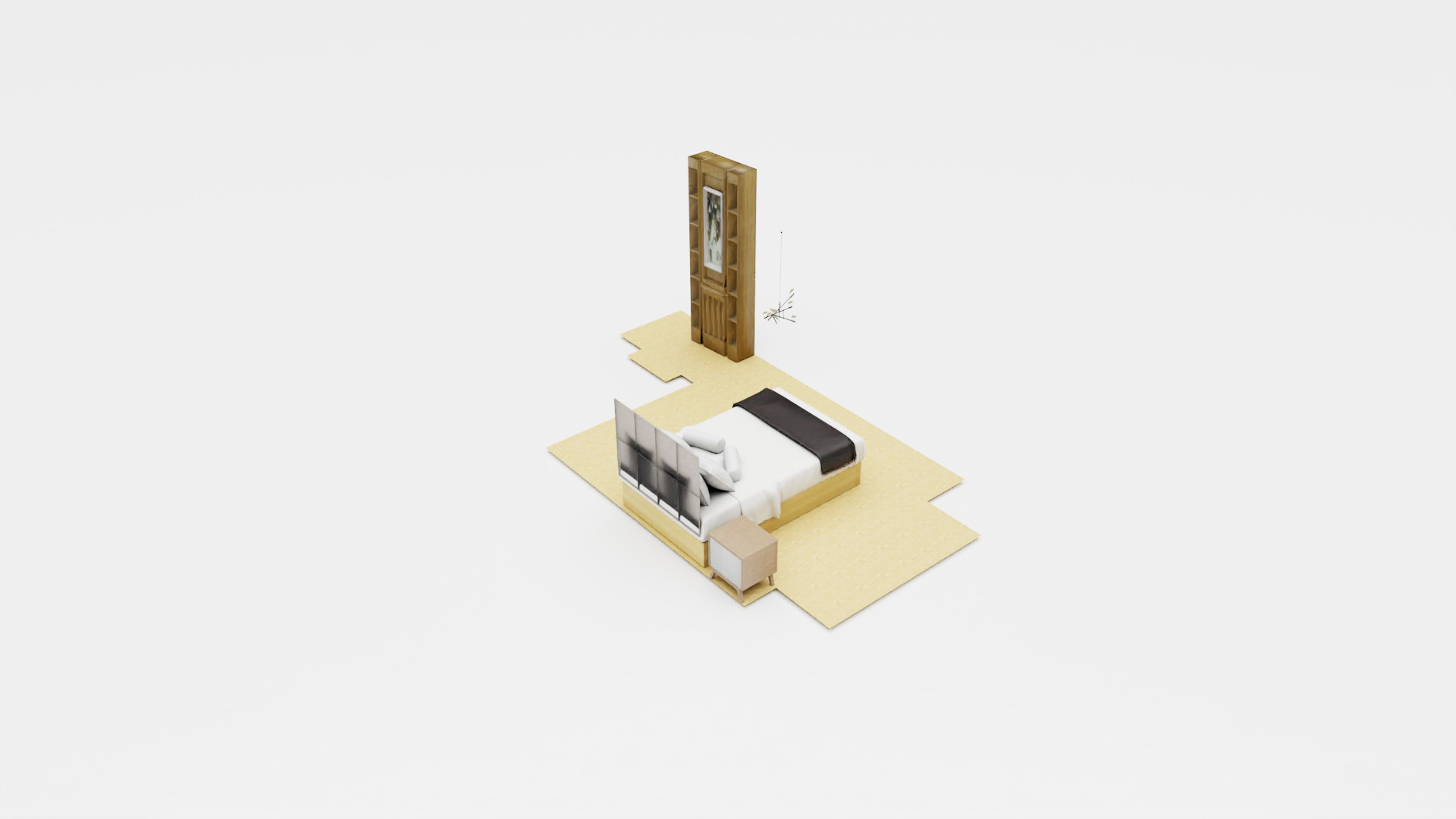}
    \end{subfigure}%
    \begin{subfigure}[b]{0.16\linewidth}
		\centering
		\includegraphics[width=\linewidth, trim=500 200 500 100, clip]{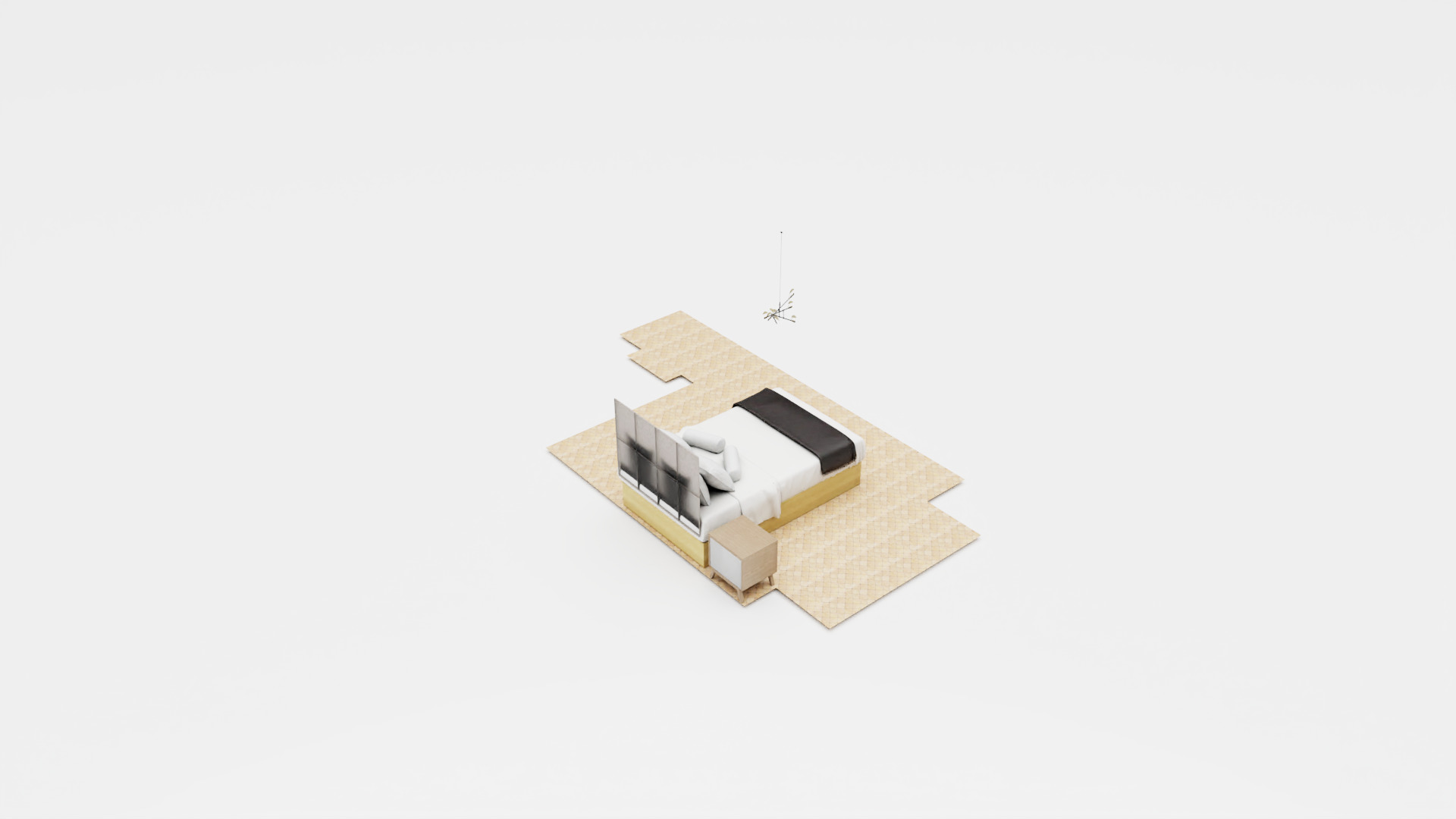}
    \end{subfigure}%
    \hfill%
    \vskip\baselineskip%
    \vspace{-1.5em}
    \hfill%
    \begin{subfigure}[b]{0.16\linewidth}
		\centering
        \small Lamp
    \end{subfigure}%
    \begin{subfigure}[b]{0.16\linewidth}
		\centering
        \small Double Bed
    \end{subfigure}%
    \begin{subfigure}[b]{0.16\linewidth}
		\centering
        \small Cabinet
    \end{subfigure}%
    \begin{subfigure}[b]{0.16\linewidth}
		\centering
        \small TV-stand
    \end{subfigure}%
    \begin{subfigure}[b]{0.16\linewidth}
		\centering
        \small Wardrobe
    \end{subfigure}%
    \begin{subfigure}[b]{0.16\linewidth}
		\centering
        \small Nothing
    \end{subfigure}%
    \hfill%
    \vspace{-1.2em}
    \vskip\baselineskip%
    \caption{\small{\bf Object Suggestion}. A user specifies a region of acceptable
    positions to place an object (marked as red boxes, 1st row) and our model suggests
    suitable objects (2nd row) to be placed in this location.}
    \label{fig:object_suggestions}
    \vspace{-1.2em}
\end{figure}
\boldparagraph{Object Suggestion}%
We now test the ability of our model to provide object suggestions given a scene and user specified location constraints.
To perform this task we sample objects from our generative model and accept the ones that fullfill the constraints provided by the user.
\figref{fig:object_suggestions} shows examples of location constraints (red box
in top row) and the corresponding objects suggested (bottom row). Note that
even when the user provided region is partially outside the room
boundaries (4th, 5th column), suggested objects always reside in the room.
Moreover, if the acceptable region overlaps with another object, our model
suggests adding nothing (6th column). This task requires computing the likelihood of an object conditioned on an
arbitrary scene, which \cite{Wang2020ARXIV,Ritchie2019CVPR}
cannot perform due to ordering.

\subsection{Perceptual Study}

We conducted two paired Amazon Mechanical Turk perceptual studies to evaluate
the quality of our generated layouts against \cite{Ritchie2019CVPR} and
\cite{Wang2020ARXIV}. We sample 6 bedroom layouts for each method from the
same 211 test set floor plans. Users saw 2 different rotating 3D scenes per
method randomly selected from 6 pre-rendered layouts. Random layouts for each
floor plan were assessed by 5 different workers to evaluate agreement and
diversity across samples for a total of 1055 question sets per paired study.
Generated scenes of \cite{Ritchie2019CVPR} were judged to contain errors like
interpenetrating furniture 41.4\% of the time, nearly twice as frequently as
our method, while \cite{Wang2020ARXIV} performs significantly worse
(\tabref{tab:ustudy_res}). Regarding realism, the scenes of
\cite{Ritchie2019CVPR} were more realistic than ours in only 26.9\% of the
cases. We conclude that our method outperforms the baselines in
the key metric, generation of realistic indoor scenes, by a large margin.
Additional details are provided in the supplementary.
\begin{table}[h!]
    \centering
    \resizebox{0.5\textwidth}{!}{
    \begin{tabular}{@{}l@{}l||c|c|c@{}}
    Method & Condition
    &%
    \begin{tabular}{@{}c@{}}
    Mean Error \\
    Frequency $\downarrow$ \\
    \end{tabular}
    &%
    \begin{tabular}{@{}c@{}}
    More $\uparrow$ \\
    Realistic
    \end{tabular}&
    \begin{tabular}{@{}c@{}}
    Realism \\
    CI $99\%$
    \end{tabular}\\
    \hline
    FastSynth \cite{Ritchie2019CVPR} & \;vs. Ours & 0.414 & 0.269 & $[0.235, 0.306]$  \\ 
    SceneFormer \cite{Wang2020ARXIV} & \;vs. Ours & 0.713 & 0.165 & $[0.138, 0.196]$  \\
    Ours & \;vs. Both & \textbf{0.232} & \textbf{0.783} & $[0.759, 0.805]$ \\
    \end{tabular}}
    \captionof{table}{\textbf{Perceptual Study Results}. Aggregated results
    for two A/B paired tests. Our method was judged more realistic
    with high confidence (binomial confidence interval with $\alpha=0.01$ reported) and contained fewer errors.}
    \label{tab:ustudy_res}
    \vspace{-0.8em}
\end{table}

\vspace{-2mm}
\section{Conclusion}\label{sec:conclusion}
\vspace{-2mm}

We introduced ATISS,
 a novel autoregressive transformer architecture for
synthesizing 3D rooms as unordered sets of objects. Our method generates realistic scenes that advance the state-of-the-art for scene synthesis. In addition, our novel formulation enables new interactive applications for semi-automated scene authoring, such as general scene completion, object suggestions, anomaly detection and more. We believe that our model is an important step not only toward automating the generation of 3D environments, with impact on simulation and virtual testing, but also toward a new generation of tools for user-driven content generation. By accepting a wide range of user inputs, our model mitigates societal risks of task automation, and promises to usher in tools that enhance the workflow of skilled laborers, rather than replacing them. In future work, we plan to extend order invariance to object attributes to further expand interactive possibilities of this model, and to incorporate style information. As any machine learning model, our model can introduce learned biases for indoor scenes, and we plan to investigate learning from less structured and more widely available data sources to make this model applicable to a wider range of cultures and environments.

{\small
	\bibliographystyle{plain}
	\bibliography{bibliography_long,bibliography,bibliography_custom}
}

\newpage
\clearpage
\appendix

\standalonetitle{Supplementary Material for\\ATISS: Autoregressive Transformers for Indoor\\ Scene Synthesis}

\begin{abstract}
In this \textbf{supplementary document}, we provide a detailed overview of our
network architecture and the training procedure. Subsequently, we describe
the preprocessing steps that we followed to filter out problematic rooms from
the 3D-FRONT dataset~\cite{Fu2020ARXIVa}. Next, we provide ablations on how
different components of our system impact the performance of our model on the
scene synthesis task and we compare ATISS with various transformer models that
consider ordering. Finally, we provide additional qualitative and quantitative
results as well as additional details for our perceptual study presented in Sec
$4.3$ in our main submission.
\end{abstract}

\section{Implementation Details}

In this section, we provide a detailed description of our network architecture.
We then describe our training protocol and provide
details on the metrics computation during training and testing.
Finally, we also provide additional details regarding our baselines.

\subsection{Network Architecture}
Here we describe the architecture of each individual component of our model
(from Fig. $2$ in the main submission). Our architecture comprises four components:
(i) the \emph{layout encoder} that maps the
room shape to a global feature representation $\bF$, (ii) the
\emph{structure encoder} that maps the $M$ objects in a
scene into per-object context embeddings $\bC=\{\bC_j\}_{j=1}^M$, (iii) the
\emph{transformer encoder} that takes $\bF$, $\bC$ and a
query embedding $\bq$ and predicts the features $\hat{\bq}$ for the next object
to be generated and (iv) the \emph{attribute extractor} that autoregressively
predicts the attributes of the next object. 

\boldparagraph{Layout Encoder}%
The first part of our architecture is the \emph{layout encoder} that is used to
map the room's floor into a global feature representation $\bF$.
We follow \cite{Wang2018SIGGRAPH} and we model the
floor plan with its top-down orthographic projection. This projection
maps the floor plan into an image, where pixel values of $1$
indicate regions inside the room and pixel values of $0$ otherwise.
The layout encoder is implemented with a ResNet-18
architecture~\cite{He2016CVPR} that is not pre-trained on
ImageNet~\cite{Deng2009CVPR}. We empirically observed that using a pre-trained
ResNet resulted in worse performance.  From the original architecture, we
remove the final fully connected layer and replace it with a linear projection to
$64$ dimensions, after average pooling.

\begin{figure}
    \centering
    \includegraphics[width=0.6\textwidth]{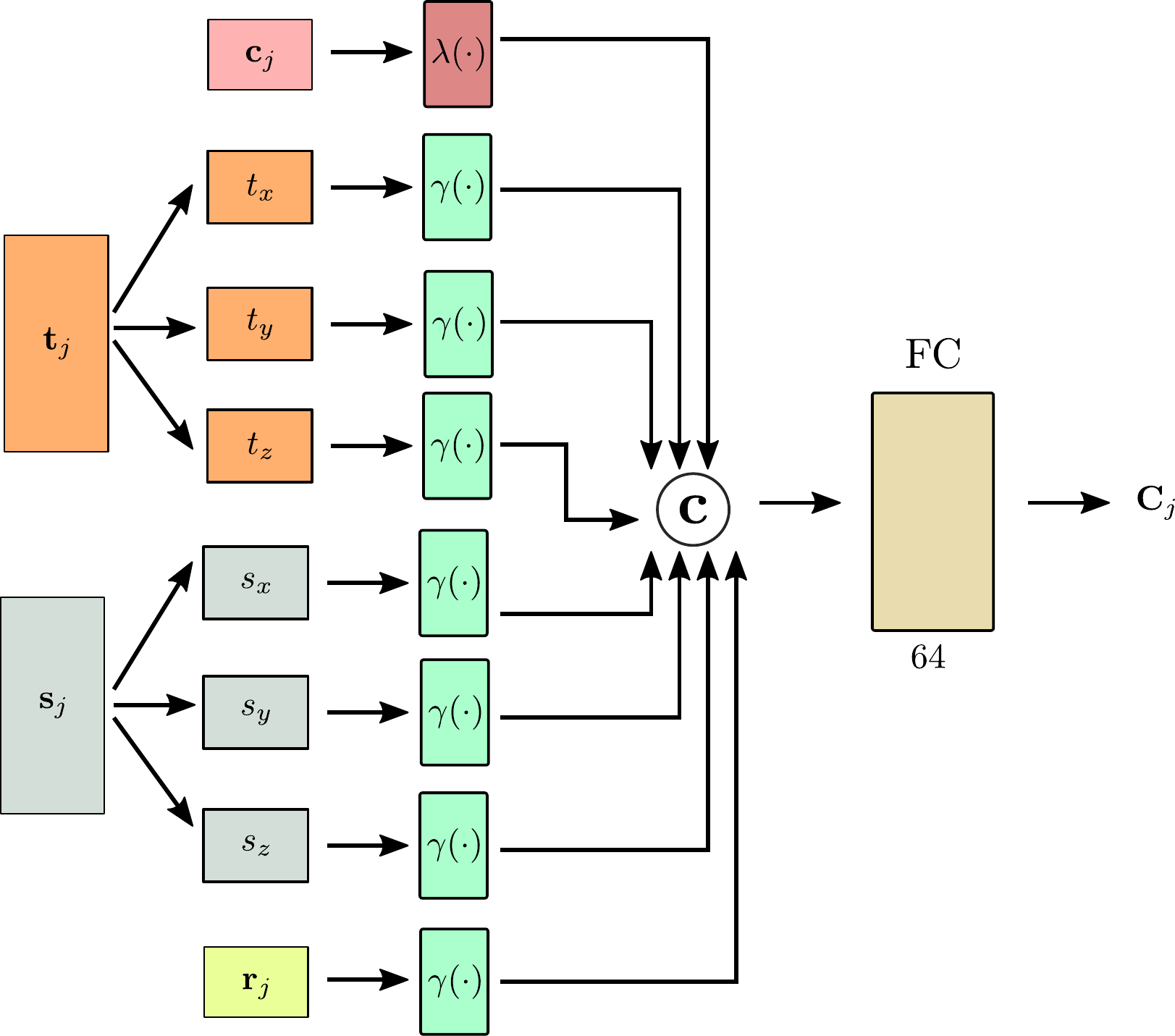}
    \caption{{\bf{Structure Encoder.}} The structure encoder predicts the
    per-object context embeddings $\bC_j$ conditioned on the object attributes.
    For the object category $\bc_j$, we use a learnable embedding
    $\lambda(\cdot)$, whereas for the location $\bt_j$, the size $\bs_j$ and
    orientation $\br_j$ we employ the positional encoding from
    \eqref{eq:positional_encoding}.
    Note that the positional encoding $\gamma(\cdot)$ is applied separately in
    each dimension of $\bt_j$ and $\bs_j$.}
    \label{fig:structure_encoder}
    \vspace{-0.8em}
\end{figure}

\boldparagraph{Structure Encoder}%
The structure encoder maps the attributes of each object into a per-object
context embedding $\bC_j$. For the object category $\bc_j$, we use a learnable
embedding, which is simply a matrix of size $C \times 64$, that stores a
per-object category vector, for all $C$ object categories in the dataset. 
For the size $\bs_j$, the position $\bt_j$ and the orientation
$\br_j$, we use the positional encoding of \cite{Vaswani2017NIPS} as follows
\begin{equation}
    \gamma(p) = (\sin(2^0\pi p), \cos(2^0\pi p), \dots, \sin(2^{L-1}\pi p), \cos(2^{L-1}\pi p))
    \label{eq:positional_encoding}
\end{equation}
where $p$ can be any of the size, position or orientation attributes and
$\gamma(\cdot)$ is applied separately in each attribute's dimension. In our
experiments, $L$ is set to $32$. The output of each embedding layer, used to map the
category, size, location and orientation in a higher dimensional space, are
concatenated into an $512$-dimensional feature vector, which is then mapped
to the per-object context embedding.
A pictorial representation
of the structure encoder is provided in \figref{fig:structure_encoder}.

\boldparagraph{Transformer Encoder}%
We follow \cite{Vaswani2017NIPS, Devlin2019NAACL} and implement our transformer 
encoder as a multi-head attention transformer without any positional
encoding. Our transformer consists of $4$ layers with $8$ attention heads. The
queries, keys and values have $64$ dimensions and the intermediate
representations for the MLPs have $1024$ dimensions. To implement the
transformer architecture we use the transformer library provided by Katharopoulos \etal
\cite{Katharopoulos2020ICML}\footnote{\href{https://github.com/idiap/fast-transformers}{https://github.com/idiap/fast-transformers}}.
The input set of the transformer is $\bI=\{\bF\}\cup\{\bC_j\}_{j=1}^M \cup{\bq}$, where $M$ denotes the number of
objects in the scene and $\bq \in \mathbb{R}^{64}$ is a learnable object
query vector that allows the transformer to predict output features
$\hat{\bq} \in \mathbb{R}^{64}$ used for generating the next object to be added
in the scene.

\begin{figure}[!h]
    \vspace{-0.8em}
    \centering
    \begin{subfigure}[t]{0.48\columnwidth}
        \centering
        \includegraphics[width=\textwidth]{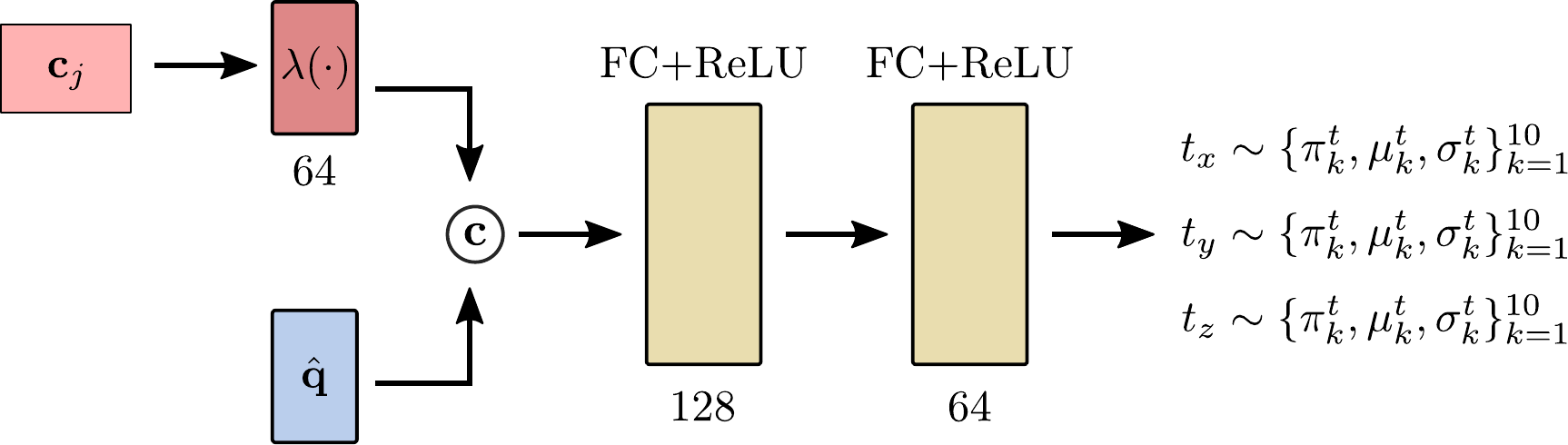}
        \vspace{-1.2em}
        \caption{$t_{\theta}(\cdot)$ predicts the parameters of the mixture of logistics distribution for the location $\bt$.}
        \label{fig:mlp_translation}
    \end{subfigure}%
    \hfill
    \begin{subfigure}[t]{0.48\columnwidth}
        \centering
        \includegraphics[width=\textwidth]{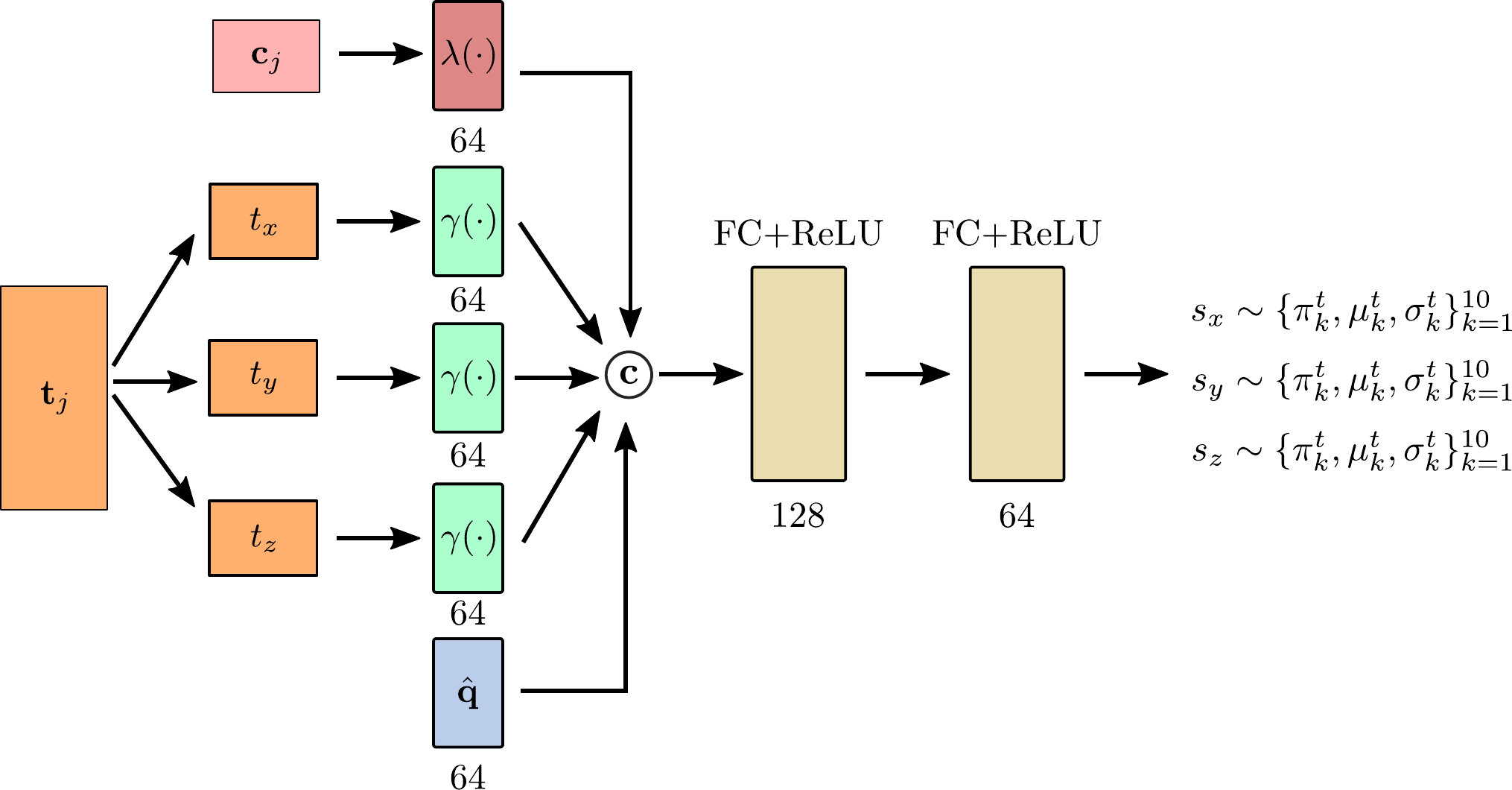}
        \vspace{-1.2em}
        \caption{$s_{\theta}(\cdot)$ predicts the parameters of the mixture of logistics distribution for the size $\bs$.}
        \label{fig:mlp_size}
    \end{subfigure}
    \caption{{\bf{Attribute Extractor.}} The attribute extractor consists of
    four MLPs that autoregressively predict the object attributes. Here we
    visualize the MLP $t_\theta(\cdot)$ for the location attribute (left side)
    and the MLP $s_\theta(\cdot)$ for the size attribute (right side).}
    \label{fig:attribute_extractor}
    \vspace{-0.8em}
\end{figure}

\boldparagraph{Attribute Extractor}%
The attribute extractor autoregressively predicts the attributes of the next
object to be added in the scene. The MLP for the object category is a linear
layer with $64$ hidden dimensions that predicts $C$ class probabilities per
object. The MLPs for the location, orientation and size predict the mean,
variance and mixing coefficient for the $K$ logistic distributions for each
attribute. In our experiments we set $K=10$.  The size, location and
orientation attributes are predicted using a 2-layer MLP with RELU
non-linearities with hidden size $128$ and output size $64$. A pictorial
representation for the MLPs $t_\theta(\cdot)$ and $\sigma_\theta(\cdot)$ used to predict the
parameters of the mixture of logistics distribution for the location and the
size is provided in \figref{fig:attribute_extractor}. Note that
$r_\theta$ is defined in a similar manner.

\subsection{Object Retrieval}
During inference, we select 3D models from the 3D-FUTURE
dataset~\cite{Fu2020ARXIVb} to be placed in the scene based on the predicted
category, location, orientation and size. In particular, we perform nearest
neighbor search through the 3D-FUTURE dataset\cite{Fu2020ARXIVb} to find the
closest model in terms of object dimensions. While prior work
\cite{Ritchie2019CVPR, Wang2020ARXIV} explored more complex object retrieval
schemes based on object dimensions and object cooccurrences (\ie favor 3D model
of objects that frequently co-occur in the dataset), we note that our simple
object retrieval strategy consistently resulted in visually
plausible rooms. We leave more advanced object retrieval schemes for future
research.

\subsection{Training Protocol}
In all our experiments, we use the Adam optimizer \cite{Kingma2015ICLR} with
learning rate $\eta=10^{-4}$ and no weight decay. For the other
hyperparameters of Adam we use the PyTorch defaults: $\beta_1=0.9$, $\beta_2 =
0.999$ and $\epsilon= 10^{-8}$. We train all models with a batch size of $128$
for $100$k iterations. During training, we perform rotation augmentation with
random rotations between $[0, 360]$ degrees.
To determine when to stop training, we follow common practice and evaluate the
validation metric every 1000 iterations and use the model that performed
best as our final model.

\subsection{Metrics Computation}\label{subsec:evaluation_metrics}
As mentioned in our main submission, we evaluate our model and our baselines using
the KL divergence between the object category distributions of synthesized
and real scenes and the classification accuracy of a
classifier trained to discriminate real from synthetic scenes as well as the
FID~\cite{Heusel2017NIPS} score between $256^2$
top-down orthographic projections of synthesized and real scenes using the code provided by Parmar \etal
\cite{Parmar2021ARXIV}\footnote{\href{https://github.com/GaParmar/clean-fid}{https://github.com/GaParmar/clean-fid}}.
For the metrics computation, we generate the same amount of
scenes as in the test set and we compute each metric using real scenes from the
test set. In particular, for the KL divergence,  we measure the frequency of object category
occurrences in the generated scenes and compare it with the frequency of
object occurrences in real scenes. Regarding the scene classification accuracy, we
train a classifier to distinguish real from generated scenes. Our classifier is
an Alexnet \cite{Krizhevsky2012NIPS} pre-trained on ImageNet, that takes as
input a $256^2$ top-down image-based representation of a room and predicts whether this
scene is real or synthetic. Both for the FID and the classification accuracy,
we repeat the metric computation 10 times and report the average.

\subsection{Baselines}

In this section, we provide additional details regarding our baselines. We
compare our model with FastSynth~\cite{Ritchie2019CVPR} and
SceneFormer~\cite{Wang2020ARXIV}. Both methods were originally
evaluated on the SUNCG dataset \cite{Song2017CVPR}, which is currently
unavailable, thus we retrained both on 3D-FRONT using the augmentation
techniques described in the original papers. To ensure fair comparison, we use
the same object retrieval for all methods and no rule-based post-processing on
the generated layouts.

\boldparagraph{FastSynth}%
In FastSynth~\cite{Ritchie2019CVPR}, the authors employ a series of image-based
CNNs to sequentially predict the attributes of the next object to be added in
the scene. In addition to $2$D labeled bounding boxes they have auxiliary
supervision in the form of object segmentation masks, depth maps, wall masks
\etc. For more details, we refer the reader to \cite{Wang2018SIGGRAPH}. During
training, they assume that there exists an ordering of objects in each scene,
based on the average size of each category multiplied by its frequency of
occurrences in the dataset. Each CNN module is trained separately and the object
properties are predicted in an autoregressive manner: object category first,
followed by location, orientation and size. We train
\cite{Ritchie2019CVPR}\footnote{\href{https://github.com/brownvc/fast-synth}{https://github.com/brownvc/fast-synth}}
using the provided PyTorch \cite{Paszke2016ARXIV} implementation with the
default parameters until convergence.

\boldparagraph{SceneFormer}%
In SceneFormer~\cite{Wang2020ARXIV}, the authors utilize a series of
transformers to autoregressively add objects in the scene, similar to
\cite{Ritchie2019CVPR}. In particular, they train a separate transformer for
each attribute and they predict the object properties in an autoregressive
manner: object category first, followed by orientation, location and size.
Similar to \cite{Ritchie2019CVPR}, they also treat scenes as ordered sequences of
objects ordered by the frequency of their categories.
We train \cite{Wang2020ARXIV}\footnote{\href{https://github.com/cy94/sceneformer}{https://github.com/cy94/sceneformer}} using the provided PyTorch \cite{Paszke2016ARXIV} implementation with the
default parameters until convergence.

\section{3D-FRONT Dataset Filtering}
\label{sec:3d_front_filtering}

We evaluate our model on the 3D-FRONT dataset \cite{Fu2020ARXIVa},
which is one of the few available datasets that contain indoor environments.
3D-FRONT contains a collection
of $6813$ houses with roughly $14629$ designed rooms,
populated with 3D furniture objects from the 3D-FUTURE dataset
\cite{Fu2020ARXIVb}. In our experiments, we focused on four room types: (i)
bedrooms, (ii) living rooms, (iii) dining rooms and (iv) libraries.
Unfortunately, 3D-FRONT contains multiple problematic rooms with unnatural
sizes, misclassified objects as well as objects in unnatural positions \eg
outside the room boundaries, lamps on the floor, overlapping objects \etc.
Therefore, in order to be able to use it, we had to perform thorough filtering
to remove problematic scenes. In this section, we present in detail the
pre-processing steps for each room type. We plan to release the names/ids of
the filtered rooms, when the paper is published.

The 3D-FRONT dataset provides scenes for the following room types: \emph{bedroom},
\emph{diningroom}, \emph{elderlyroom}, \emph{kidsroom}, \emph{library},
\emph{livingdiningroom}, \emph{livingroom}, \emph{masterbedroom},
\emph{nannyroom}, \emph{secondbedroom} that contain $2287$, $3233$, $233$,
$951$, $967$, $2672$, $1095$, $3313$, $16$ and $2534$ rooms respectively. Since
some room types have very few rooms we do not consider them in our evaluation.

\begin{figure}[h!]
    \begin{subfigure}[t]{0.5\textwidth}
        \centering
        \includegraphics[width=1.0\textwidth]{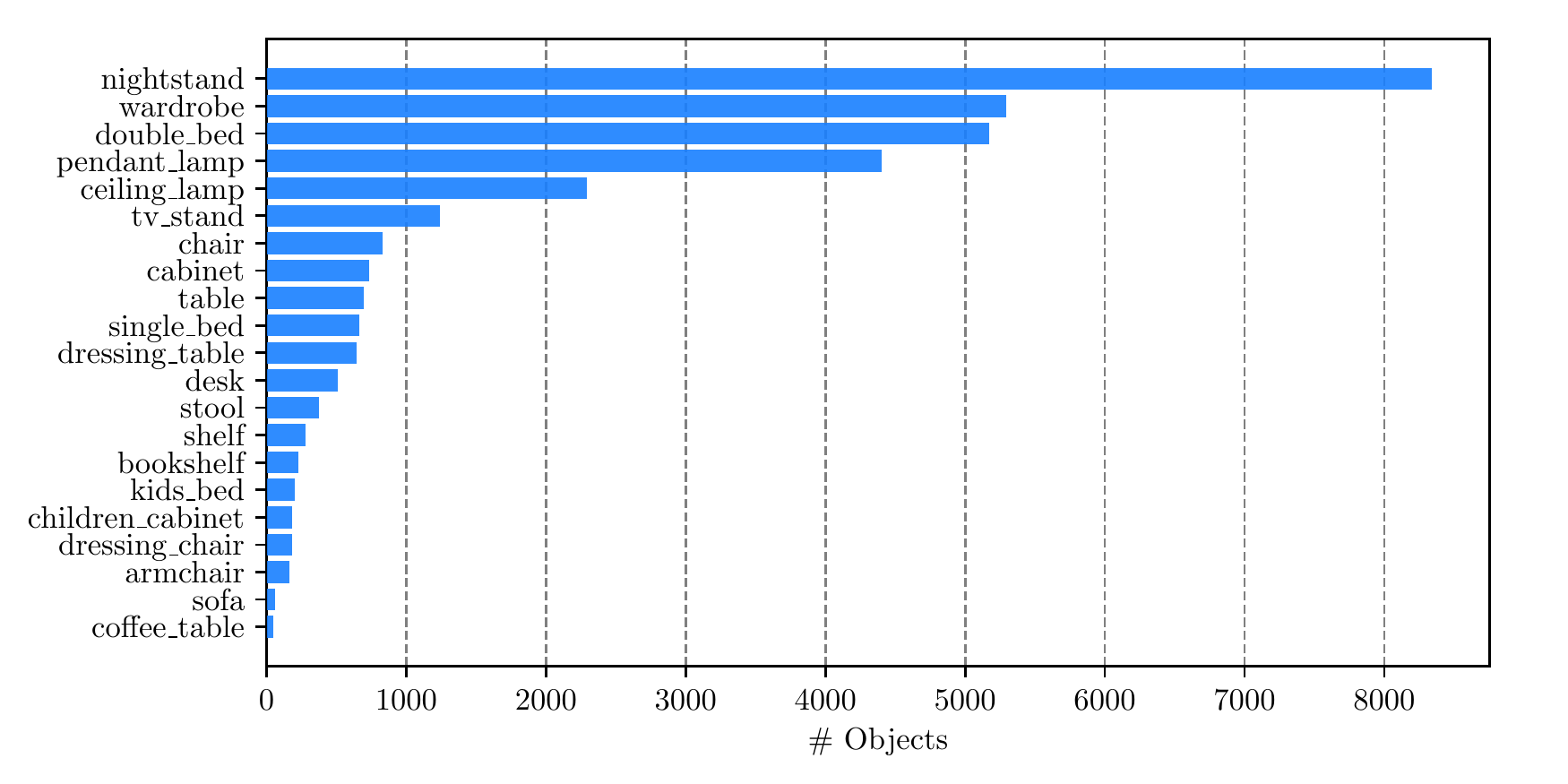}
        \vspace{-1.2em}
        \caption{{\bf{Bedrooms}}}
        \label{fig:bedroom_object_frequencies}
    \end{subfigure}%
    \begin{subfigure}[t]{0.5\textwidth}
        \centering
        \includegraphics[width=1.0\textwidth]{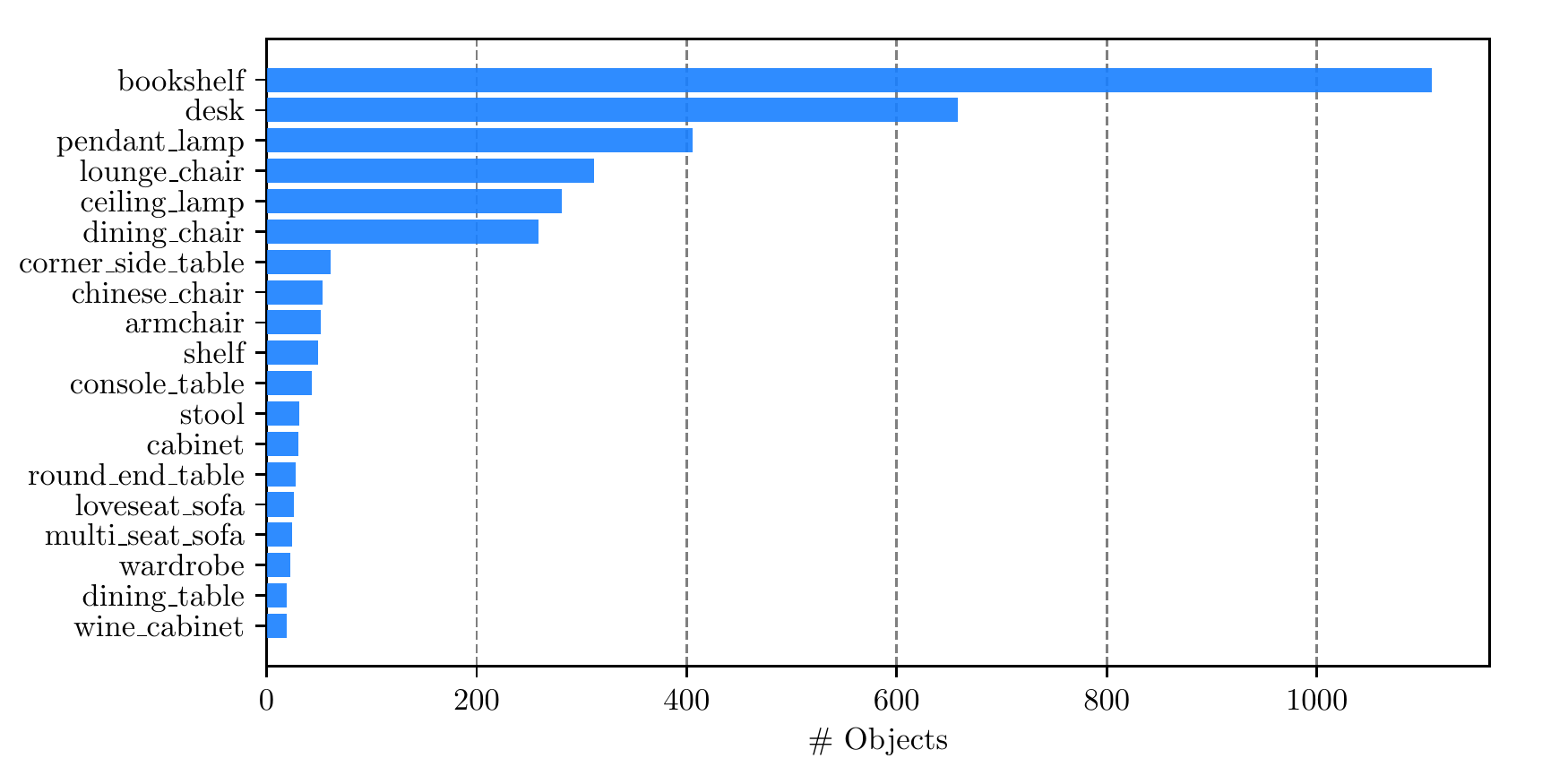}
        \vspace{-1.2em}
        \caption{{\bf{Libraries}}}
        \label{fig:library_object_frequencies}
    \end{subfigure}
    \caption{\bf{Number of object occurrences in Bedrooms and Libraries.}}
    \vspace{-1.2em}
\end{figure}

\boldparagraph{Bedroom}%
To create training and test data for bedroom scenes, we consider rooms of type
\emph{bedroom, secondbedroom} and \emph{masterbedroom}, which amounts to $8134$
rooms in total. We start by removing rooms of unnatural sizes, namely rooms
that are larger than $6$m $\times$ $6$m in floor size and taller than $4$m.
Next, we remove infrequent objects that appear in less than $15$ rooms, such as
chaise lounge sofa, l-shaped sofa, barstool, wine cabinet \etc.  Subsequently,
we filter out rooms that contain fewer than $3$ and more than $13$ objects,
since they amount to a small portion of the dataset.  Since the original
dataset contained various rooms with problematic object arrangements such
overlapping objects, we also remove rooms that have objects that are
overlapping as well as misclassified objects \eg beds being classified as
wardrobes. This results in $5996$ bedrooms with $21$ object categories in
total. \figref{fig:bedroom_object_frequencies} illustrates the
number of appearances of each object category in the $5996$ bedroom scenes and
we remark that the most common category is the nightstand with $8337$
occurrences and the least common is the coffee table with $45$.

\boldparagraph{Library}%
We consider rooms of type \emph{library} that amounts to $967$ scenes in total.
For the case of libraries, we start by filtering out rooms with unnatural sizes
that are larger than $6$m $\times$ $6$m in floor size and taller than $4$m.
Again we remove rooms that contain overlapping objects, objects positioned
outside the room boundaries as well as rooms with unnatural layouts \eg
single chair positioned in the center of the room. We also filter out rooms
that contain less than $3$ objects and more than $12$ objects since they appear
less frequently. Our pre-processing resulted in $622$ rooms with $19$ object categories 
in total. \figref{fig:library_object_frequencies} shows the number of
appearances of each object category in the $622$ libraries. The most common
category is the bookshelf with $1109$ occurrences and the least common is the
wine cabinet with $19$.

\begin{figure}[t!]
    \begin{subfigure}[t]{0.5\textwidth}
        \centering
        \includegraphics[width=1.0\textwidth]{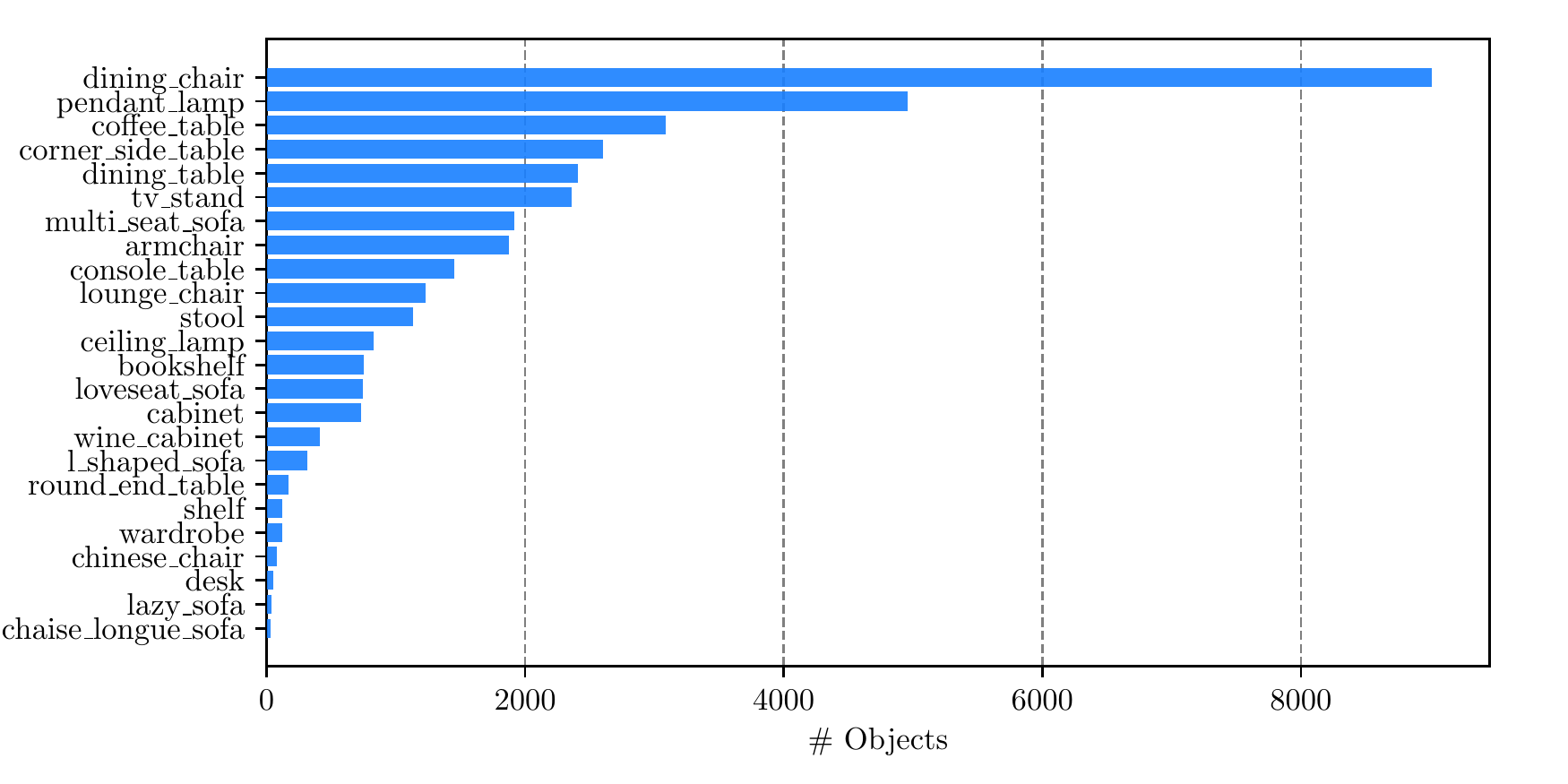}
        \vspace{-1.2em}
        \caption{{\bf{Living Rooms}}}
        \label{fig:livingroom_object_frequencies}
    \end{subfigure}%
    \begin{subfigure}[t]{0.5\textwidth}
        \centering
        \includegraphics[width=1.0\textwidth]{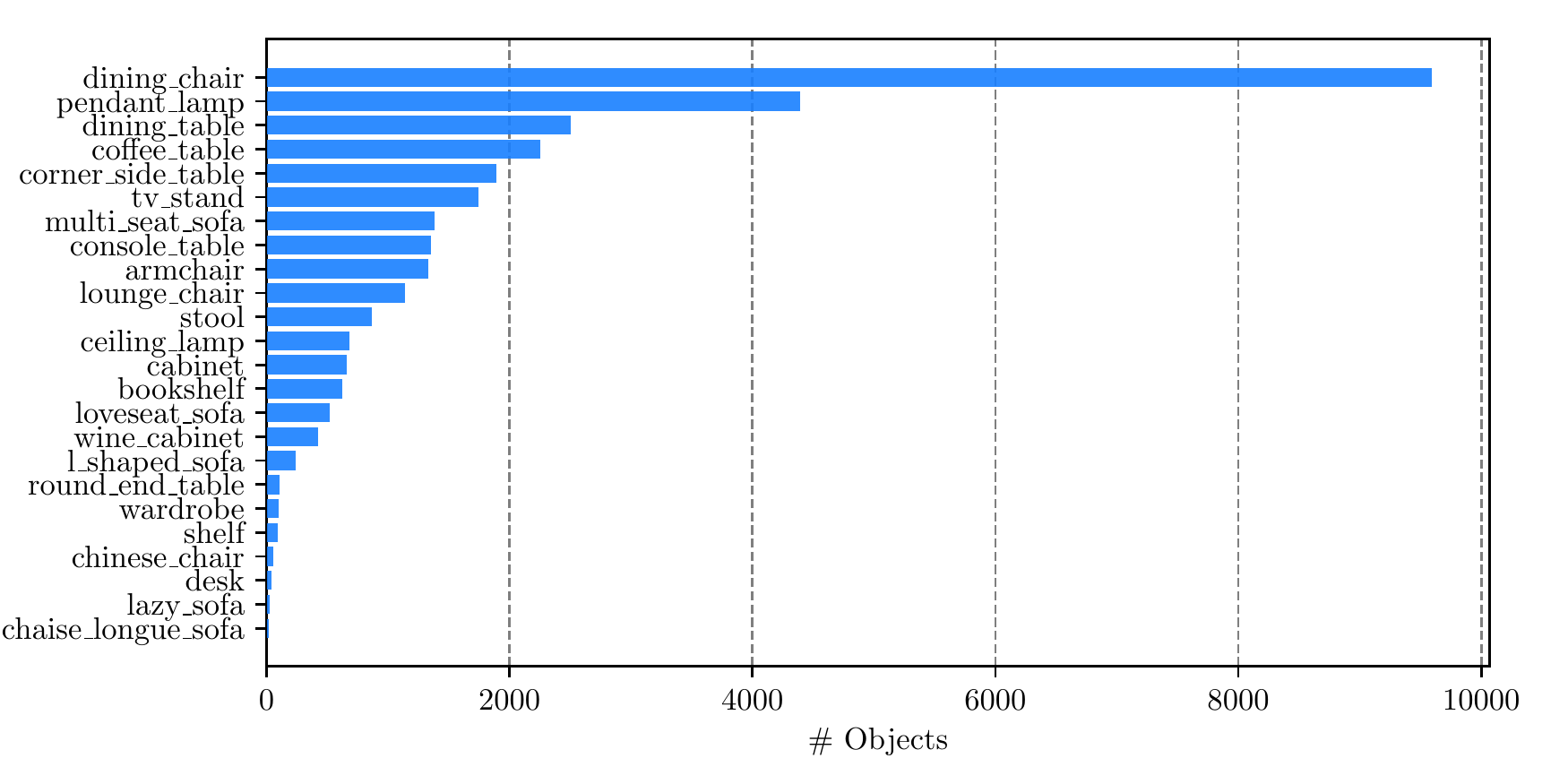}
        \vspace{-1.2em}
        \caption{{\bf{Dining Rooms}}}
        \label{fig:diningroom_object_frequencies}
    \end{subfigure}
    \caption{\bf{Number of object occurrences in Living Rooms and Dining Rooms.}}
    \vspace{-1.2em}
\end{figure}

\boldparagraph{Living Room}%
For the living rooms, we consider rooms of type \emph{livingroom} and
\emph{livingdiningroom}, which amounts to $3767$ rooms. We follow a similar
process as before and we start by filtering out rooms with unnatural sizes.  In
particular, we discard rooms that are larger than $12$m $\times$ $12$m in floor
size and taller than $4$m. We also remove uncommon objects that appear in less
than $15$ rooms such as bed and bed frame. Next, we filter out rooms that
contain less than $3$ objects and more than $13$ objects, since they are
significantly less frequent. For the case of living rooms, we observed that
some of the original scenes contained multiple lamps without having any other
furniture. Since this is unnatural, we also removed these scenes together with
some rooms that had either overlapping objects or objects positioned outside
the room boundaries. Finally, we also remove any scenes that contain any kind
of bed \eg double bed, single bed, kid bed \etc.  After our pre-processing, we
ended up with $2962$ living rooms with $24$ object categories in total.
\figref{fig:livingroom_object_frequencies} visualizes the number of occurrences
of each object category in the living rooms. We observe that the most common
category is the dining chair with $9009$ occurrences and the least common is
the chaise lounge sofa with $30$.

\boldparagraph{Dining Room}%
For the dining rooms, we consider rooms of type \emph{diningroom} and
\emph{livingdiningroom}, since the \emph{diningroom} scenes amount to only $233$ scenes.
This results in $3233$ rooms in total. For the dining rooms,
we follow the same filtering process as for the living rooms and we keep
$2625$ rooms with $24$ objects in total. \figref{fig:diningroom_object_frequencies}
shows the number of occurrences of each object category in the dining rooms.
The most common category is the dining chair with $9589$
occurrences and the least common is the chaise lounge sofa with $19$.

To generate the train, test and validation splits, we split the preprocessed
rooms such that $70\%$ is used for training, $20\%$ for testing and $10\%$ for
validation. Note that the 3D-FRONT dataset comprises multiple houses that may
contain the same room, e.g the exact same object arrangement might appear in
multiple houses. Thus splitting train and test scenes solely based on whether
they belong to different houses could result in the same room appearing both in
train and test scenes. Therefore, instea of randomly selecting rooms from
houses but we select from the set of rooms with distinct object arrangements.

\section{Ablation Study}
\label{sec:ablations}

In this section, we investigate how various components of our model affect its
performance on the scene synthesis task. In
\secref{subsec:mixture_of_logistic_distributions}, we investigate the impact of
the number of logistic distributions in the performance of our model. 
Next, in
\secref{subsec:layout_encoder}, we examine the impact of the architecture of
the layout encoder. In \secref{subsec:ordered_transformers}, we compare
ATISS with two variants of our model that consider ordered sets of objects.
Unless stated otherwise, all ablations are conducted on the bedroom scenes of
the 3D-FRONT~\cite{Fu2020ARXIVa} dataset.

\subsection{Mixture of Logistic distributions}
\label{subsec:mixture_of_logistic_distributions}

We represent objects in a scene as labeled 3D bounding boxes and model them
with four random variables that describe their category, size, orientation and
location, $o_j = \{\bc_j, \bs_j, \bt_j, \br_j\}$. The category
$\bc_j$ is modeled using a categorical variable over the total number of
object categories $C$ in the dataset. For the size $\bs_j \in \mathbb{R}^3$, the
location $\bt_j \in \mathbb{R}^3$ and the orientation $\br_j \in \mathbb{R}^1$,
we follow \cite{Salimans2017ICLR, Oord2016SSW} and model them with a mixture
of logistic distributions
\begin{equation}
    \bs_j \sim \sum_{k=1}^K \, \pi_k^s \text{logistic}(\mu_k^s, \sigma_k^s)\quad
    \bt_j \sim \sum_{k=1}^K \, \pi_k^t \text{logistic}(\mu_k^t, \sigma_k^t)\quad
    \br_j \sim \sum_{k=1}^K \, \pi_k^r \text{logistic}(\mu_k^r, \sigma_k^r)
\end{equation}
where $\pi_k^s$, $\mu_k^s$ and $\sigma_k^s$ denote the
weight, mean and variance of the $k$-th logistic distribution used for
modeling the size. Similarly, $\pi_k^t$, $\mu_k^t$ and $\sigma_k^t$ and
$\pi_k^r$, $\mu_k^r$ ans $\sigma_k^r$ refer to the weight, mean and variance of
the $k$-th logistic of the location and orientation, respectively.

In this experiment, we test our model with different numbers for logistic distributions for
modelling the object attributes. Results are summarized in
\tabref{tab:ablation_n_mixtures}.
\begin{table}[!h]
    \centering
    \resizebox{0.7\textwidth}{!}{
    \begin{tabular}{l|ccc}
        \toprule
        & FID ($\downarrow$) & Classification Accuracy ($\downarrow$) & Category Distribution ($\downarrow$)\\
        \midrule
        $K=1$  & 41.71 $\pm$ 0.4008 & 0.7826 $\pm$ 0.0080 & 0.0491 \\
        $K=5$  & 40.41 $\pm$ 0.2491 & 0.5667 $\pm$ 0.0405 & 0.0105 \\
        $K=10$ & {\bf{38.39}} $\pm$ 0.3392 & {\bf{0.5620}} $\pm$ 0.0228 & 0.0085\\
        $K=15$ & 40.41 $\pm$ 0.4504 & 0.5980 $\pm$ 0.0074 & 0.0095\\
        $K=20$ & 40.39 $\pm$ 0.3964 & 0.6680 $\pm$ 0.0035 & \bf{0.0076} \\
        \bottomrule
    \end{tabular}}
    \vspace{0.6em}
    \caption{{\bf Ablation Study on the Number of Logistic Distributions.} This
    table shows a quantitative comparison of our approach with different
    numbers of $K$ logistic distributions for modelling the size, the location
    and the orientation of each object.}
    \label{tab:ablation_n_mixtures}
    \vspace{-1.2em}
\end{table}

As it is expected, using a single logistic distribution (first row in
\tabref{tab:ablation_n_mixtures}) results in worse performance, since it
does not have enough representation capacity for modelling the object
attributes. We also note that increasing the number of logistic
distributions beyond $10$ hurts performance \wrt FID and
classification accuracy. We hypothesize that this is due to overfitting. In our
experiments we set $K=10$.

\subsection{Layout Encoder}
\label{subsec:layout_encoder}

We further examine the impact of the layout encoder on the performance of our
model. To this end, we replace the ResNet-18 architecture~\cite{He2016CVPR},
with an AlexNet~\cite{Krizhevsky2012NIPS}. From the original architecture, we
remove the final classifier layers and keep only the feature vector of length
9216 after max pooling. We project this feature vector to $64$ dimensions with
a linear projection layer. Similar to our vanilla model, we do not use an AlexNet pre-trained on
ImageNet because we empirically observed that it resulted in worse performance.
\begin{table}[!h]
    \centering
    \resizebox{0.7\textwidth}{!}{
    \begin{tabular}{l|ccc}
        \toprule
        & FID ($\downarrow$) & Classification Accuracy ($\downarrow$) & Category Distribution ($\downarrow$)\\
        \midrule
        AlexNet & 40.40 $\pm$ 0.2637 & 0.6083 $\pm$ 0.0034  & \bf{0.0064}\\
        ResNet-18 & {\bf{38.39}} $\pm$ 0.3392 & {\bf{0.5620}} $\pm$ 0.0228 & 0.0085\\
        \bottomrule
    \end{tabular}}
    \vspace{0.6em}
    \caption{{\bf Ablation Study on the Layout Encoder Architecture.} This table shows a quantitative 
    comparison of ATISS with two different layout encoders.}
    \label{tab:ablation_layout_encoder}
    \vspace{-1.2em}
\end{table}

\tabref{tab:ablation_layout_encoder} compares the two variants of our model
\wrt to the FID score, the classification accuracy and the KL-divergence.
We remark that our method is not particularly sensitive to the choice of the layout encoder.
However, using an AlexNet results in slightly worse performance, hence we
utilize a ResNet-18 in all our experiments.

\subsection{Transformers with Ordering}
\label{subsec:ordered_transformers}
\begin{table}
    \centering
    \resizebox{0.8\linewidth}{!}{
    \begin{tabular}{l|ccc}
        \toprule
        & FID ($\downarrow$) & Classification Accuracy ($\downarrow$) & Category Distribution ($\downarrow$)\\
        \midrule
        Ours+Perm+Order & 40.18 $\pm$ 0.2831 & 0.6019 $\pm$  0.0060      & 0.0089\\
        Ours+Order      & 38.67 $\pm$ 0.5552 & 0.7603 $\pm$ 0.0010 & 0.0533\\
        Ours            & {\bf{38.39}} $\pm$ 0.3392 & {\bf{0.5620}} $\pm$ 0.0228 & 0.0085\\
        \bottomrule
    \end{tabular}}
    \vspace{0.6em}
    \caption{{\bf Ablation Study on Ordering.} This table shows a quantitative
    comparison of our approach \wrt two variants of our model that represent rooms as
    ordered sequence of objects.}
    \label{tab:ablations}
\end{table}

In this section, we analyse the benefits of synthesizing rooms as unordered
sets of objects in contrast to ordered sequences. To this end, we train two
variants of our model that utilize a positional embedding
\cite{Vaswani2017NIPS} to incorporate order information to the input. The first
variant is trained with random permutations of the input (Ours+Perm+Order),
similar to our model,  whereas the second with a fixed ordering based on the
object frequency (Ours+Order) as described in \cite{Ritchie2019CVPR,
Wang2020ARXIV}. We compare these variants to our model on the scene synthesis
task and observe that the variant with the fixed ordering
(second row \tabref{tab:ablations}) performs significantly worse as the classifier can identify
synthesized scenes with $76\%$ accuracy. Moreover, we remark that besides enabling all
the applications presented in our main submission, training with random
permutations also improves the quality of the synthesized scenes (first row
\tabref{tab:ablations}). However, our model that is permutation
invariant, namely the prediction is the same regardless of the order of the
partial scene, performs even better (third row \tabref{tab:ablations}). We
conjecture that the invariance of our model will be more even more crucial for
training with either larger datasets or larger scenes \ie scenes with more
objects, because observing a single order allows reasoning about all
permutations of the partial scene.

\section{Applications}

In this section, we provide additional qualitative results for various
interactive applications that benefit greatly by our unordered set
formulation.

\subsection{Failure Case Detection And Correction}
\begin{figure}
    \centering
    \begin{subfigure}[b]{0.16\linewidth}
    \centering
    \includegraphics[width=\linewidth, trim=500 270 500 125, clip]{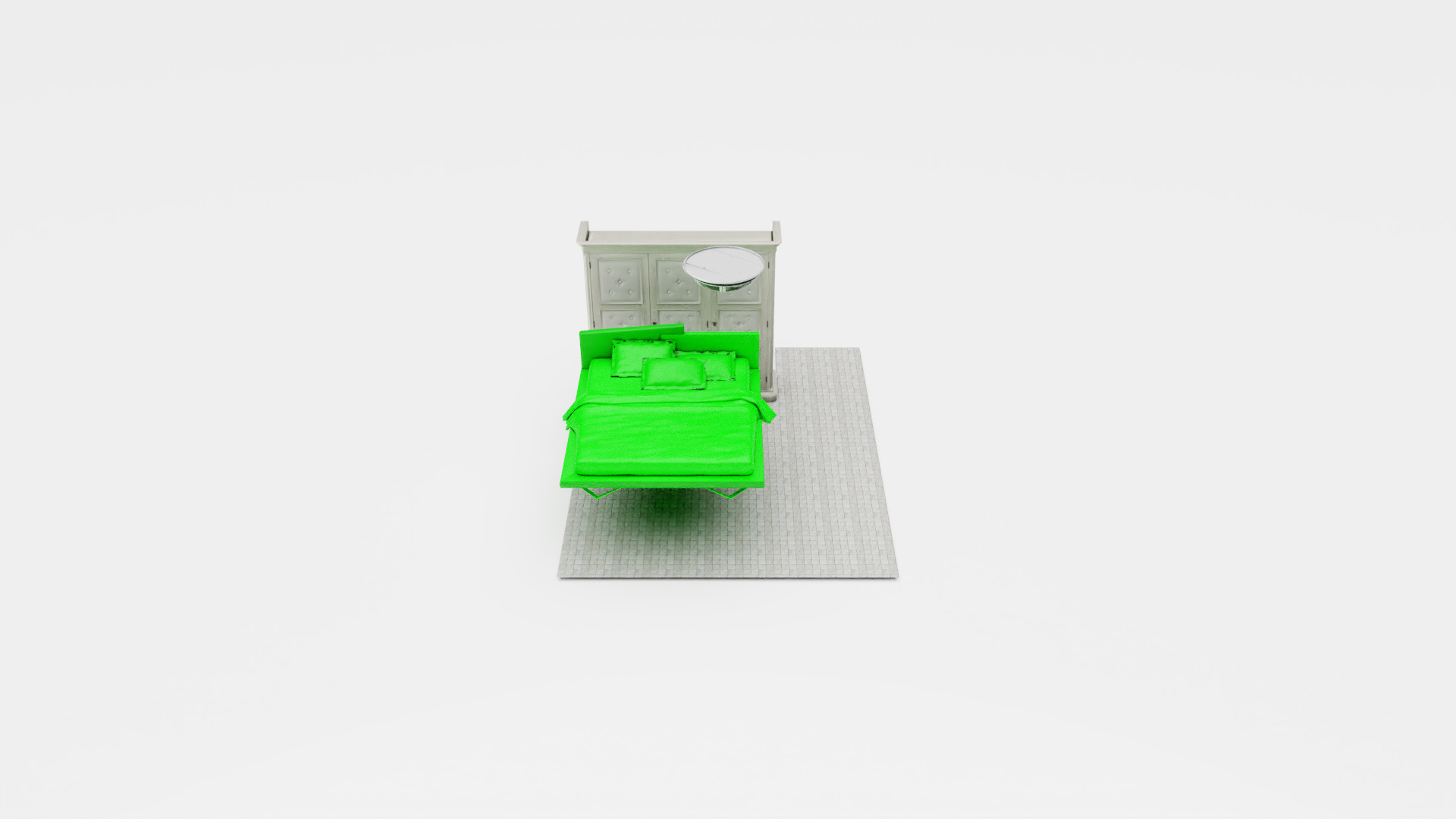}
    \end{subfigure}%
    \begin{subfigure}[b]{0.16\linewidth}
		\centering
        \includegraphics[width=\linewidth, trim=500 270 500 125, clip]{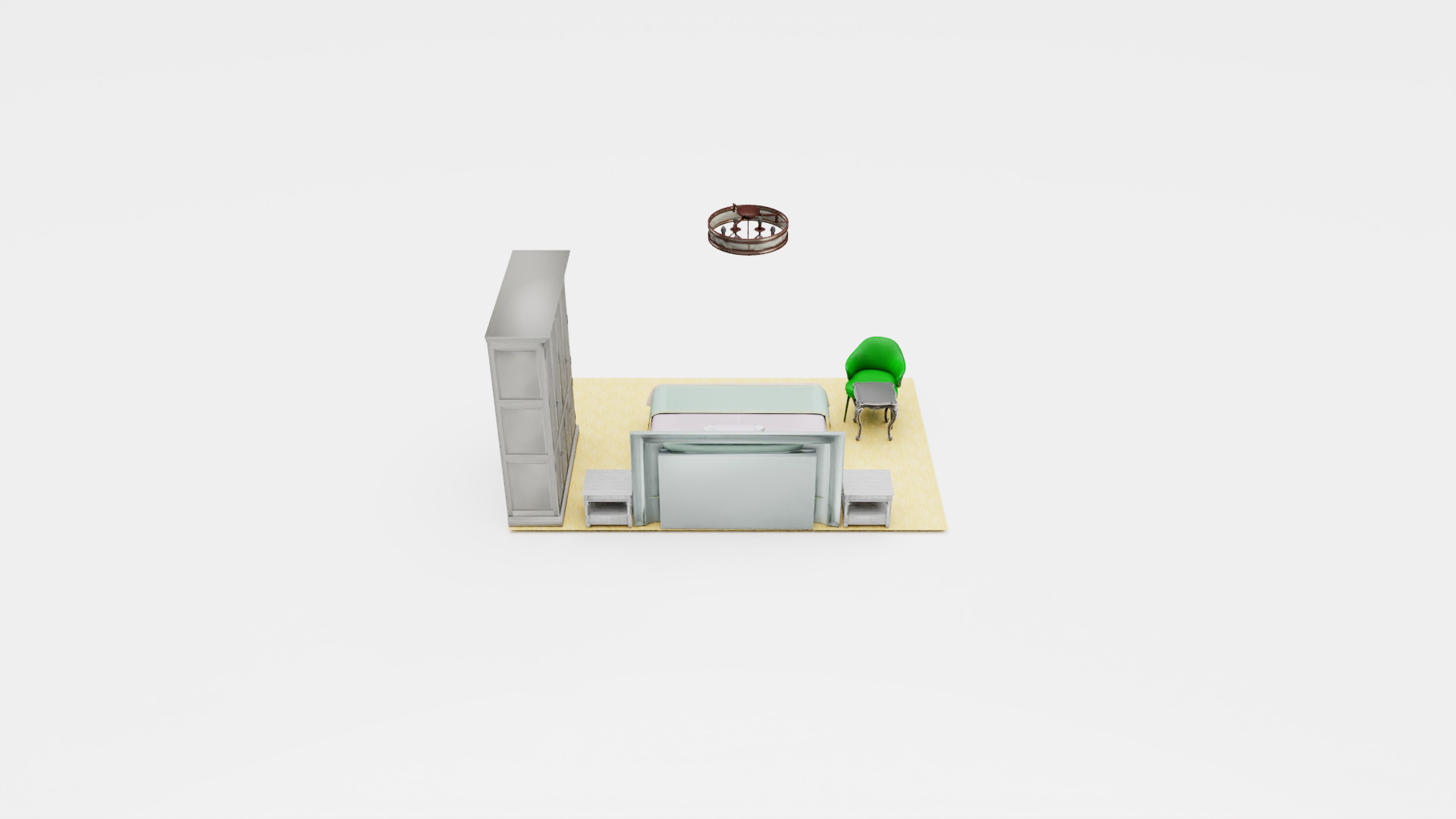}
    \end{subfigure}%
    \begin{subfigure}[b]{0.16\linewidth}
        \centering
        \includegraphics[width=\linewidth, trim=500 270 500 125, clip]{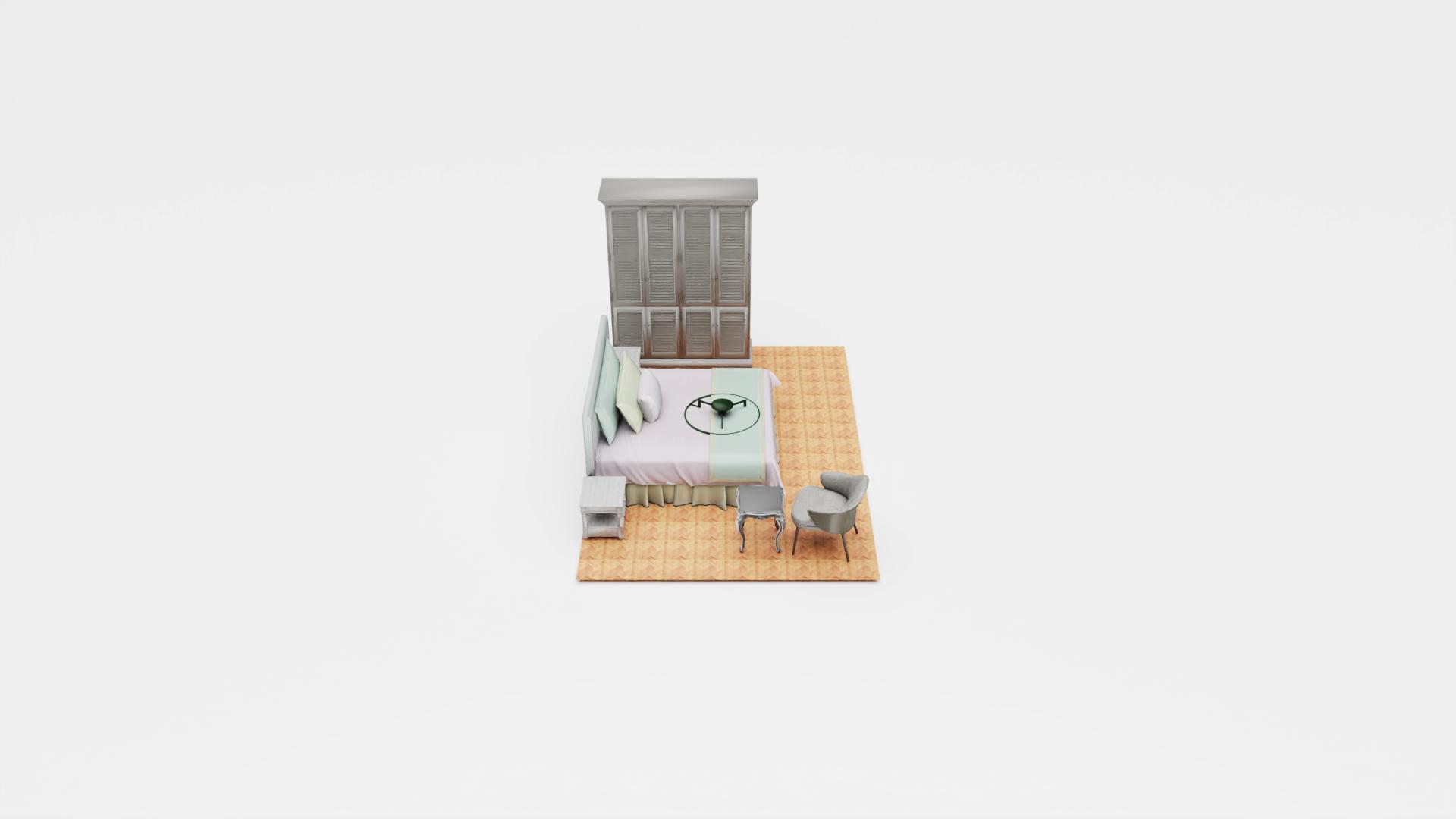}
    \end{subfigure}%
    \begin{subfigure}[b]{0.16\linewidth}
		\centering
        \includegraphics[width=\linewidth, trim=500 270 500 125, clip]{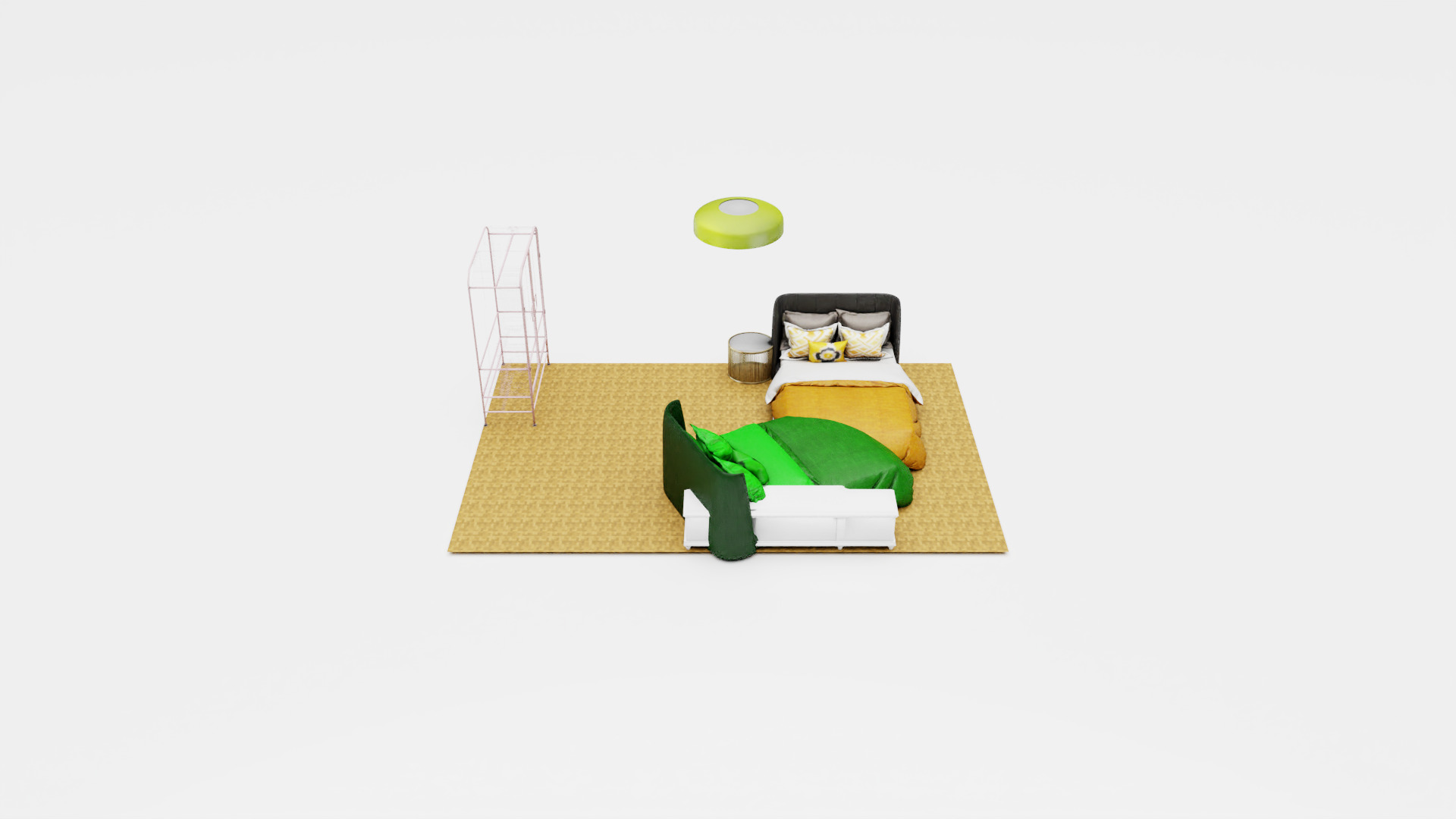}
    \end{subfigure}%
    \begin{subfigure}[b]{0.16\linewidth}
        \centering
        \includegraphics[width=\linewidth, trim=500 270 500 125, clip]{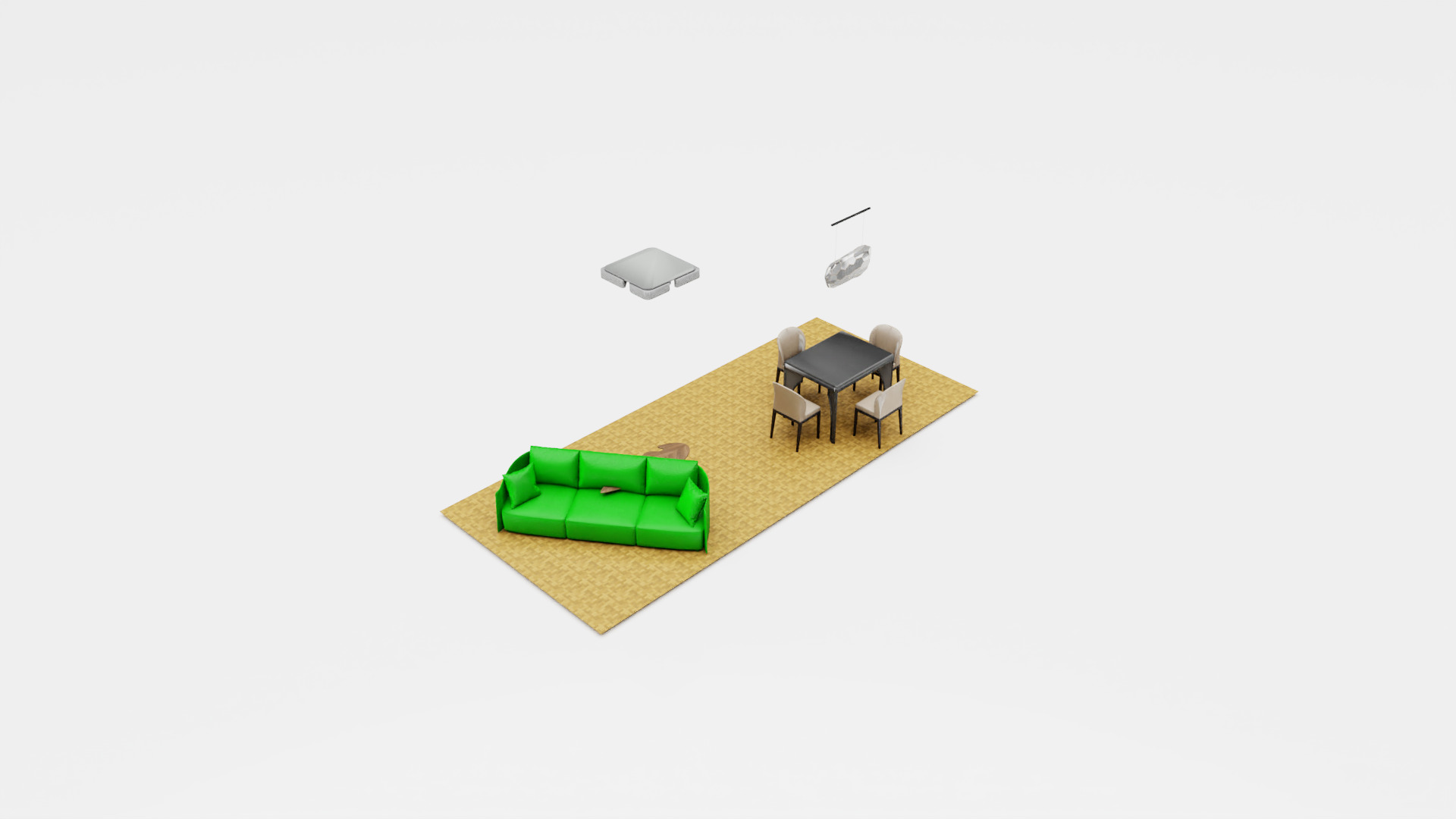}
    \end{subfigure}%
    \begin{subfigure}[b]{0.16\linewidth}
		\centering
		\includegraphics[width=\linewidth, trim=200 10 300 10, clip]{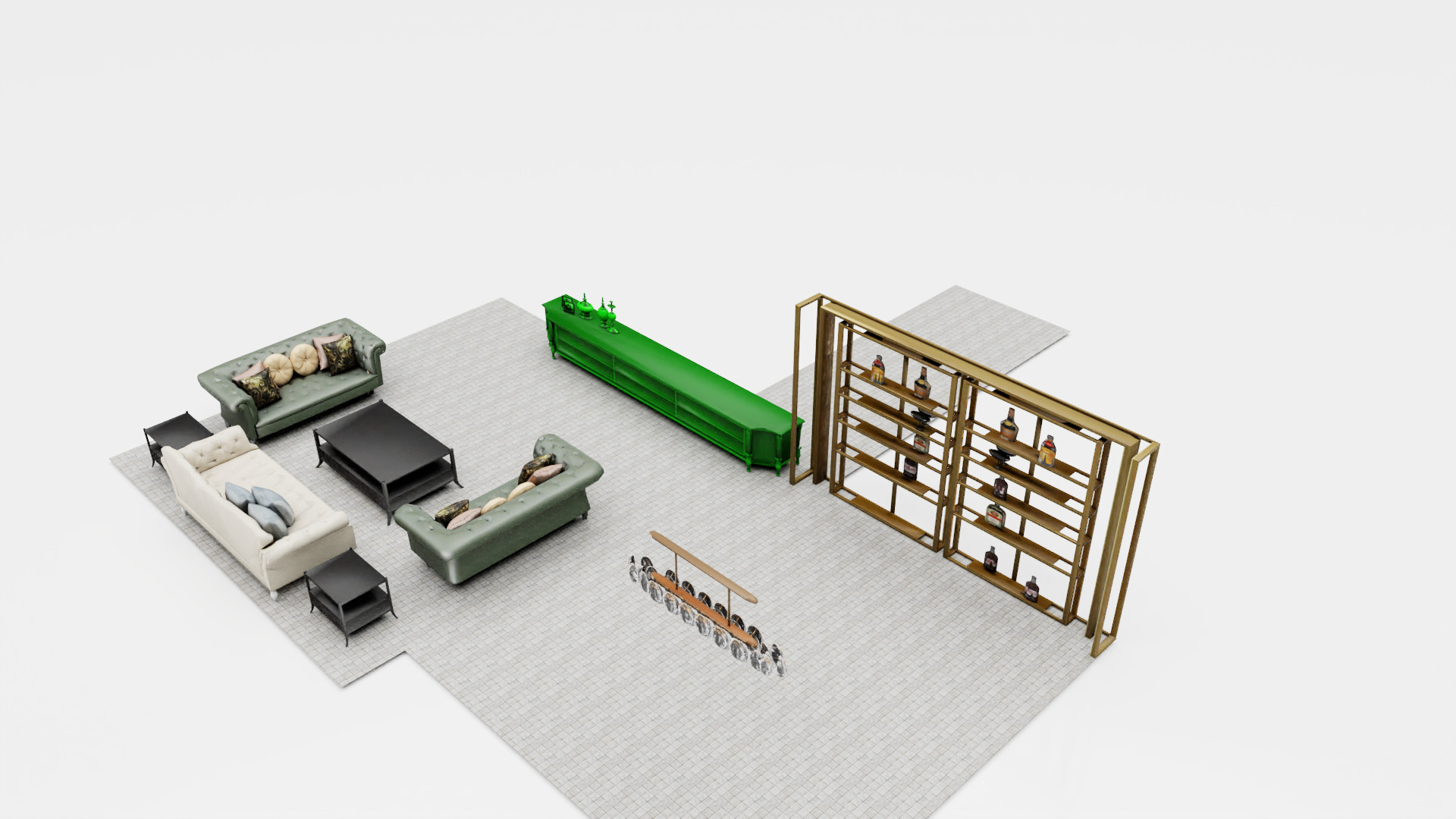}
    \end{subfigure}%
    \vspace{-1.2em}
    \vskip\baselineskip%
    \begin{subfigure}[b]{0.16\linewidth}
        \centering
        \includegraphics[width=\linewidth, trim=500 270 500 125, clip]{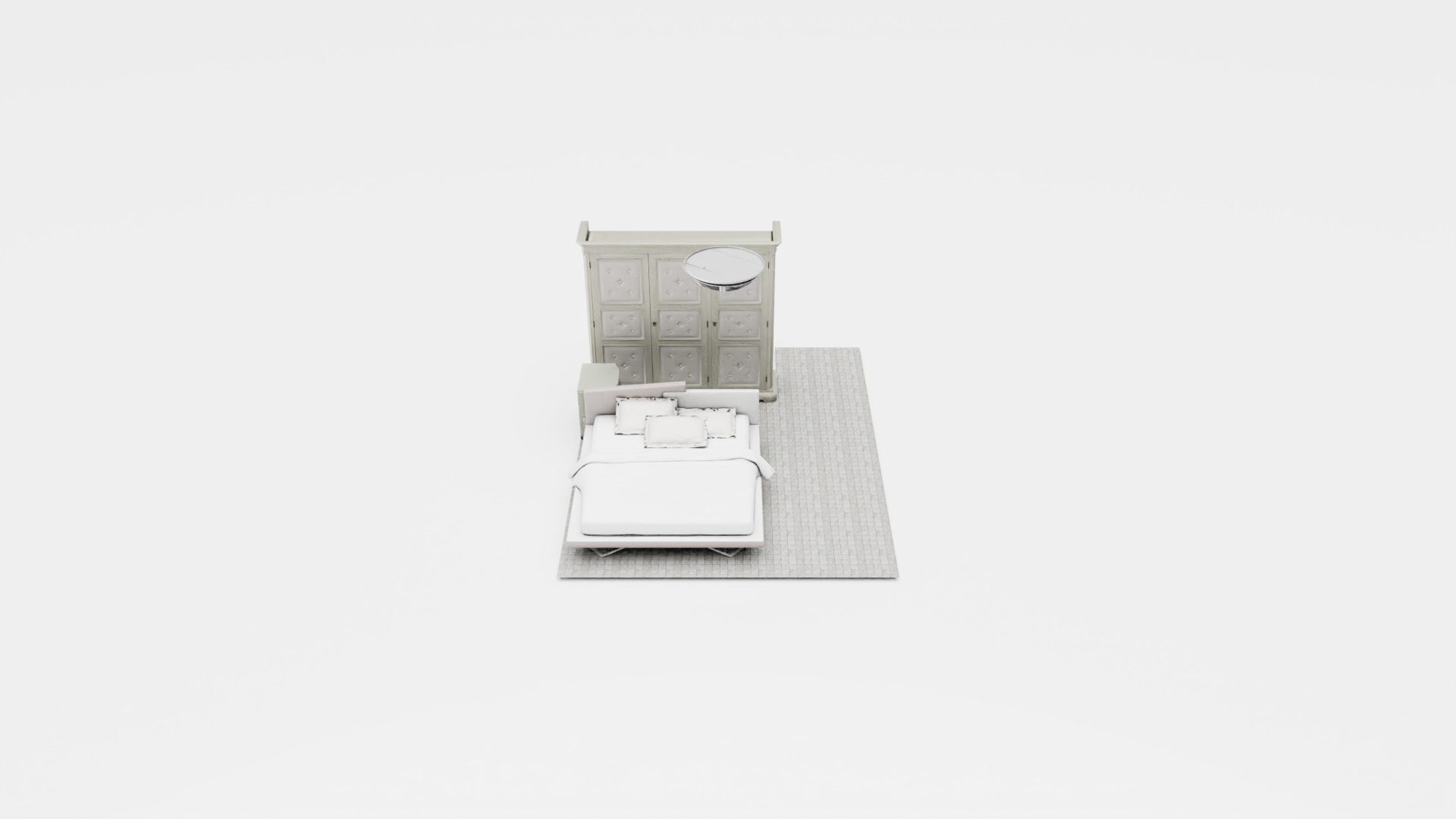}
    \end{subfigure}%
    \begin{subfigure}[b]{0.16\linewidth}
		\centering
        \includegraphics[width=\linewidth, trim=500 270 500 125, clip]{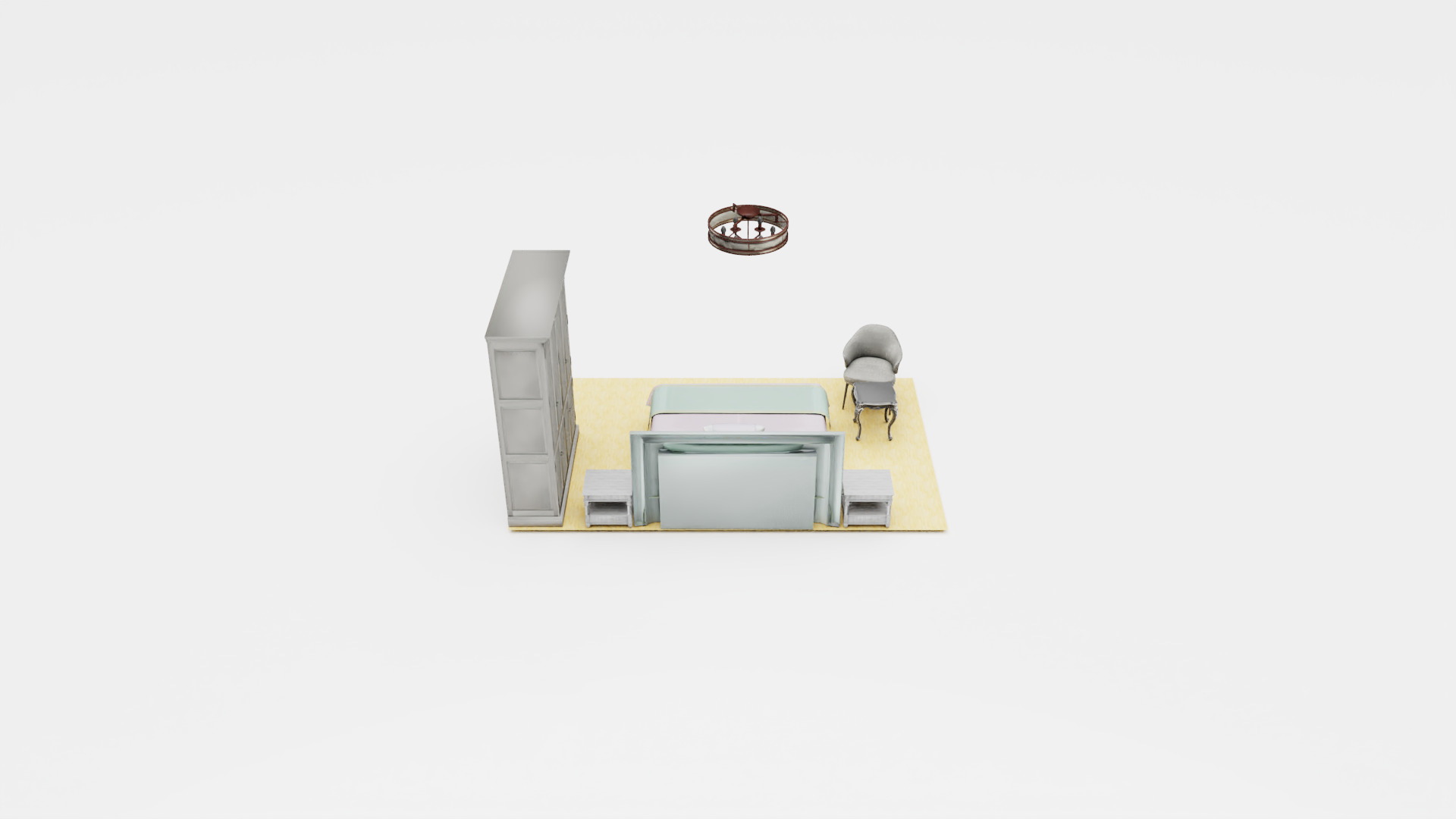}
    \end{subfigure}%
    \begin{subfigure}[b]{0.16\linewidth}
        \centering
        \includegraphics[width=\linewidth, trim=500 270 500 125, clip]{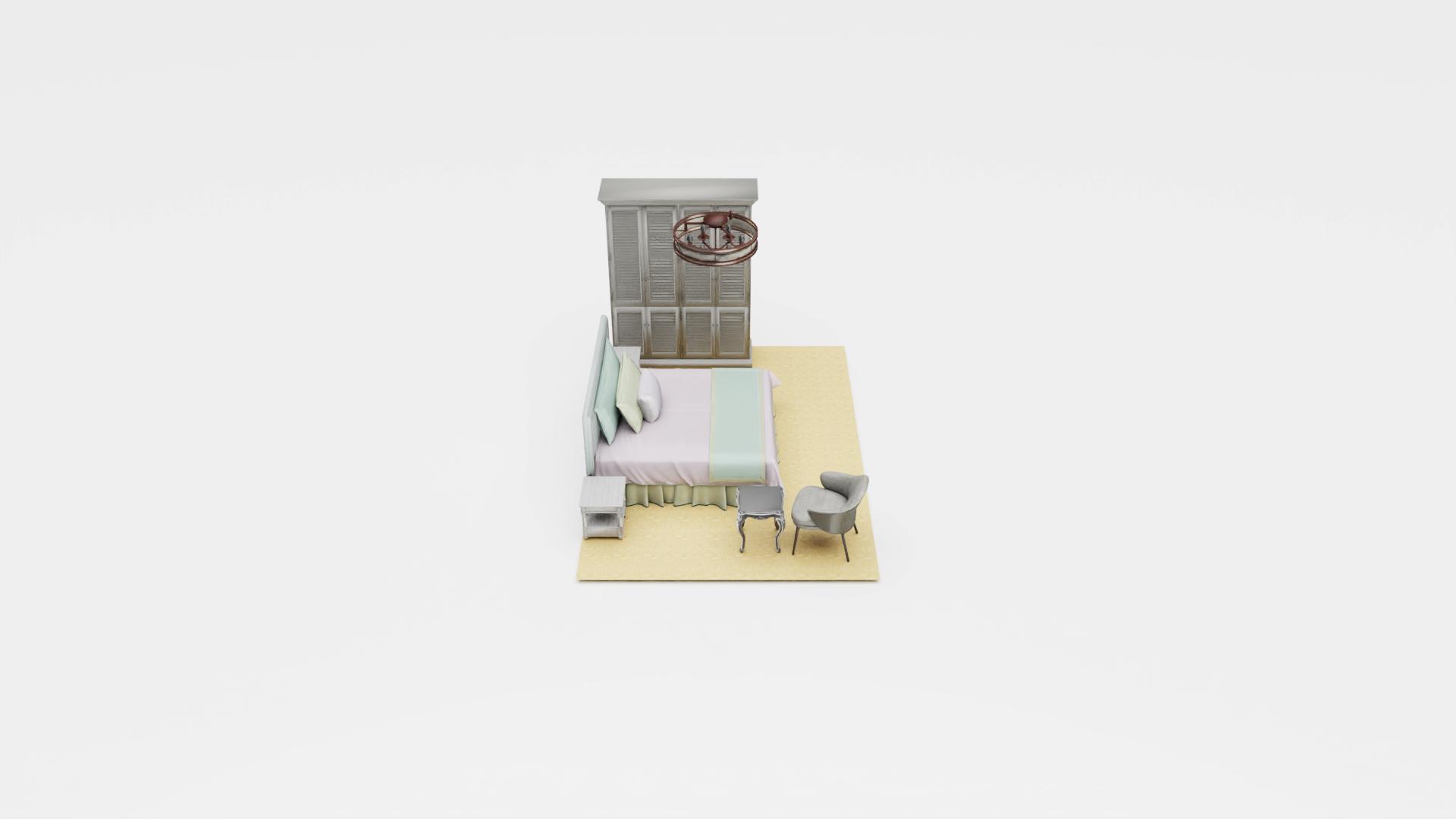}
    \end{subfigure}%
    \begin{subfigure}[b]{0.16\linewidth}
		\centering
        \includegraphics[width=\linewidth, trim=500 270 500 125, clip]{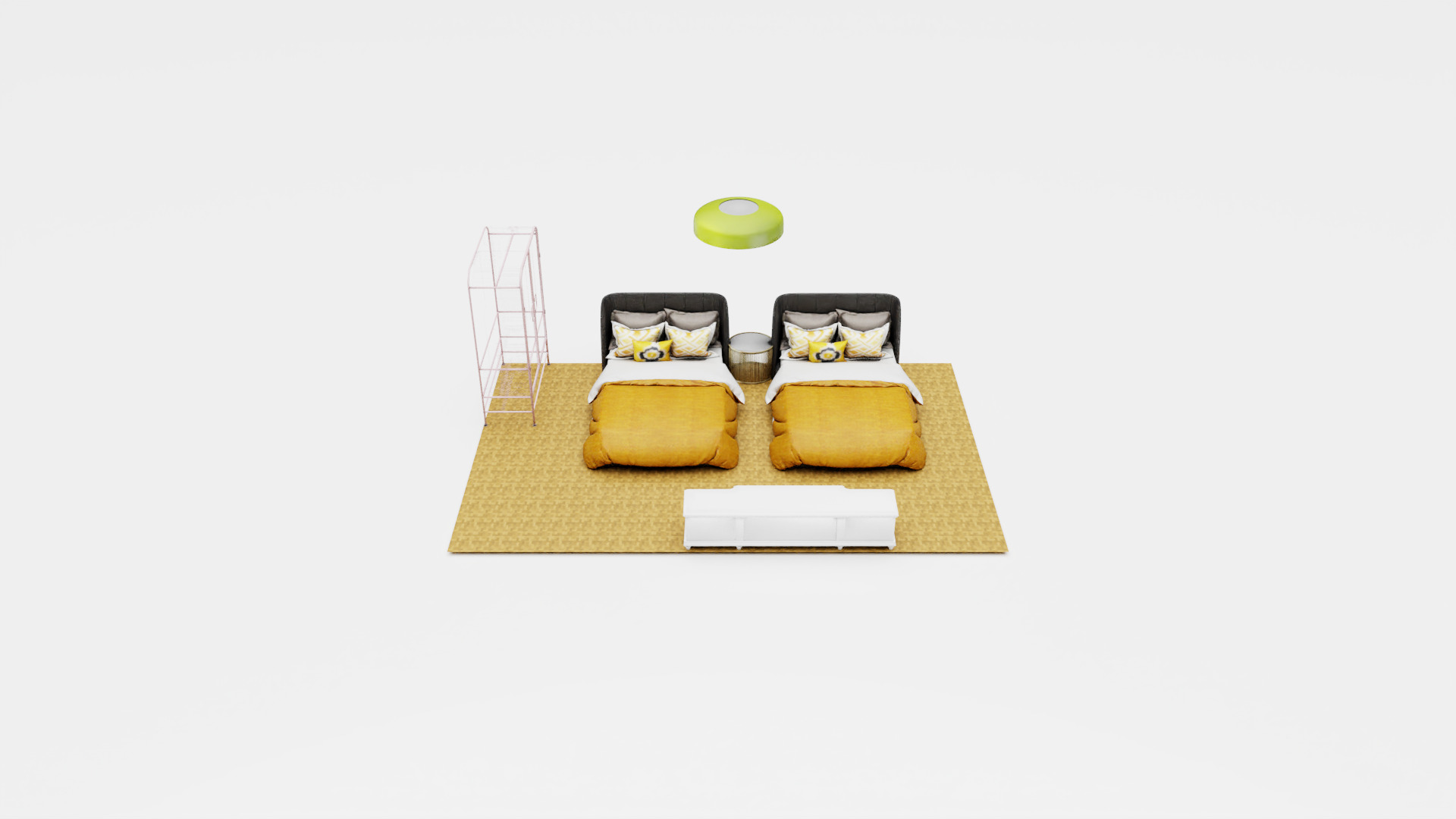}
    \end{subfigure}%
    \begin{subfigure}[b]{0.16\linewidth}
    \centering
        \includegraphics[width=\linewidth, trim=500 270 500 125, clip]{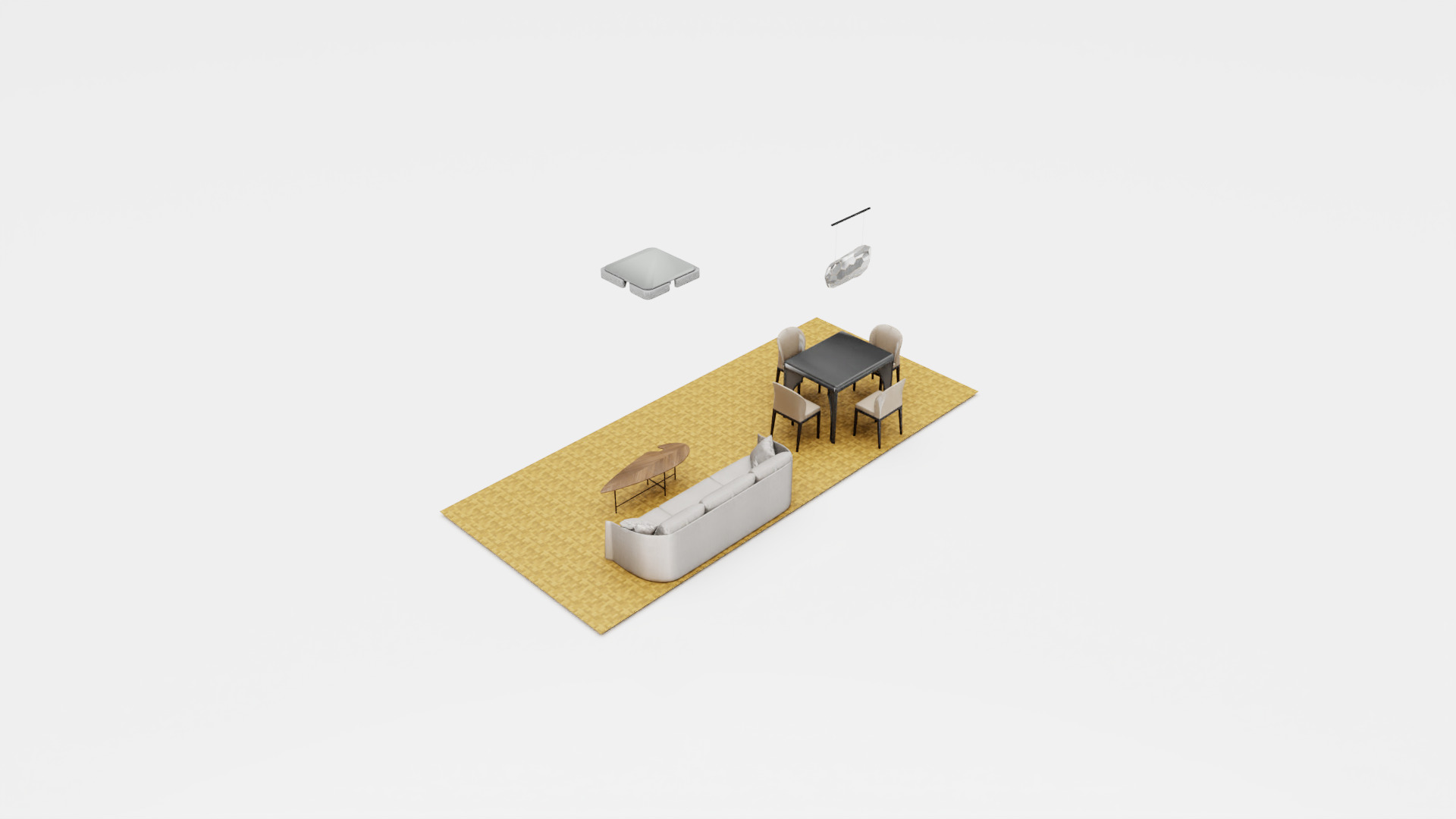}
    \end{subfigure}    \begin{subfigure}[b]{0.16\linewidth}
		\centering
		\includegraphics[width=\linewidth, trim=200 10 300 10, clip]{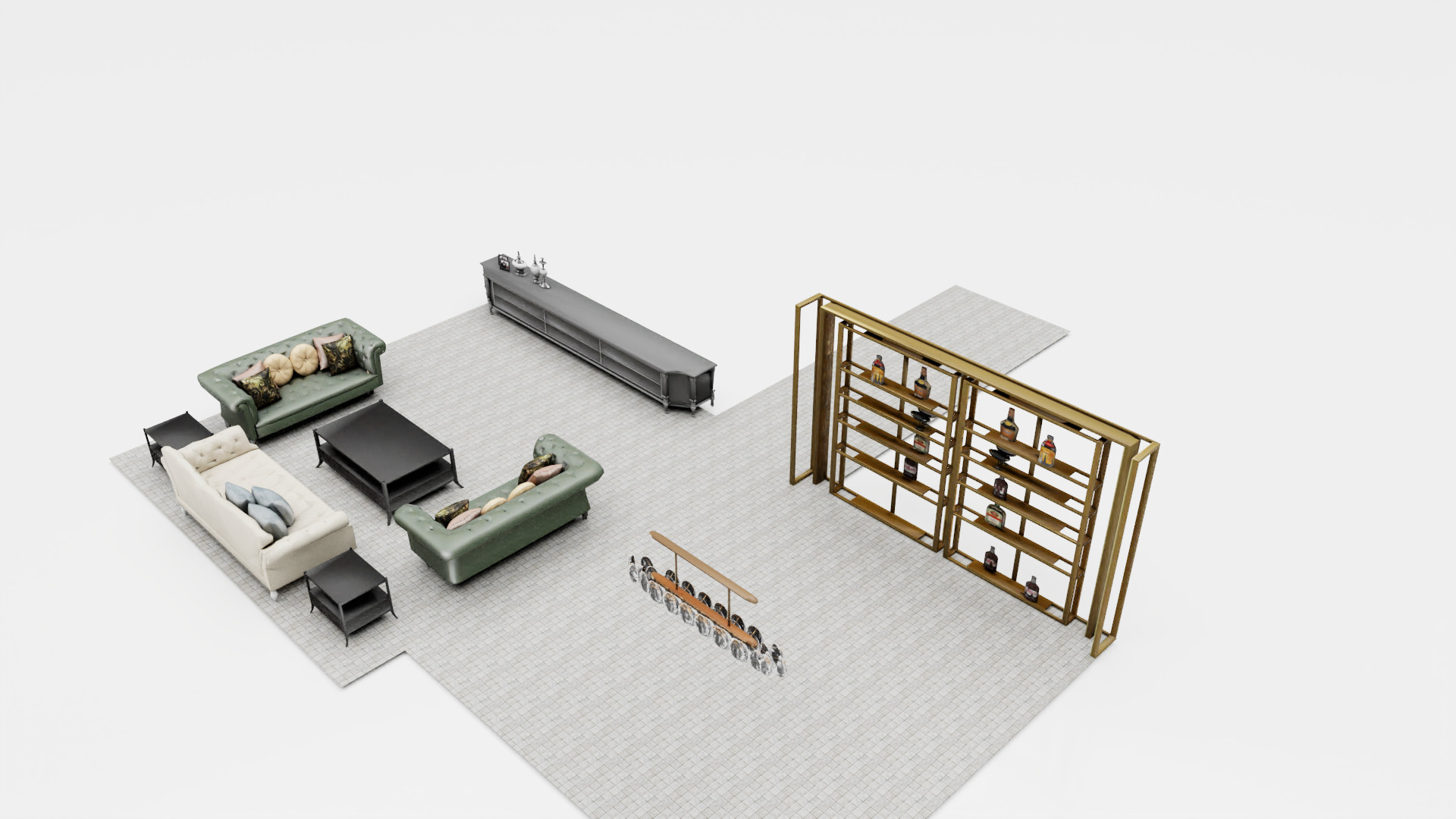}
    \end{subfigure}%
    \vskip\baselineskip%
    \vspace{-1.2em}
    \vskip\baselineskip%
    \begin{subfigure}[b]{0.16\linewidth}
    \centering
    \includegraphics[width=\linewidth, trim=500 190 400 125, clip]{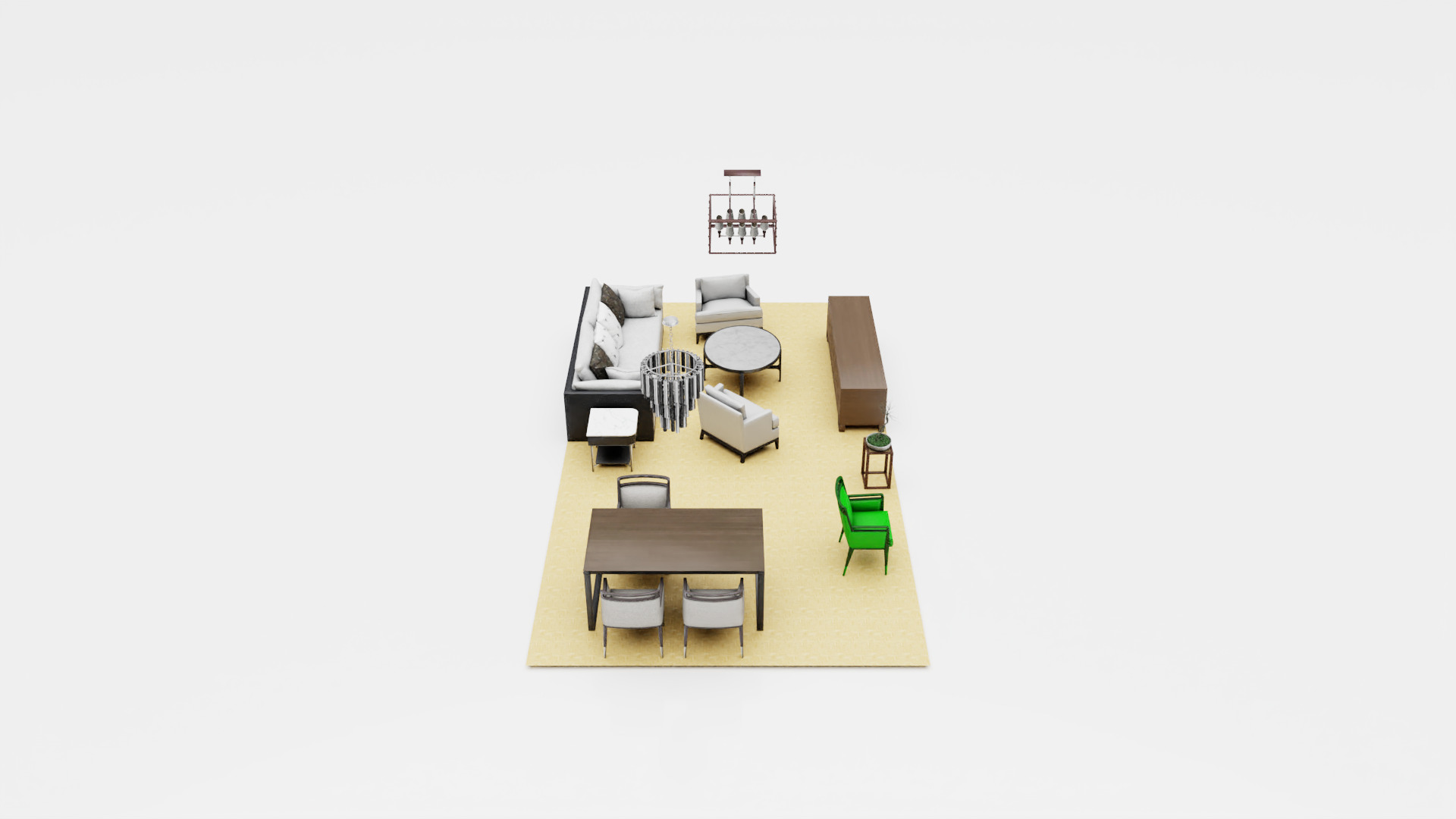}
    \end{subfigure}%
    \begin{subfigure}[b]{0.16\linewidth}
		\centering
        \includegraphics[width=\linewidth, trim=350 10 150 10, clip]{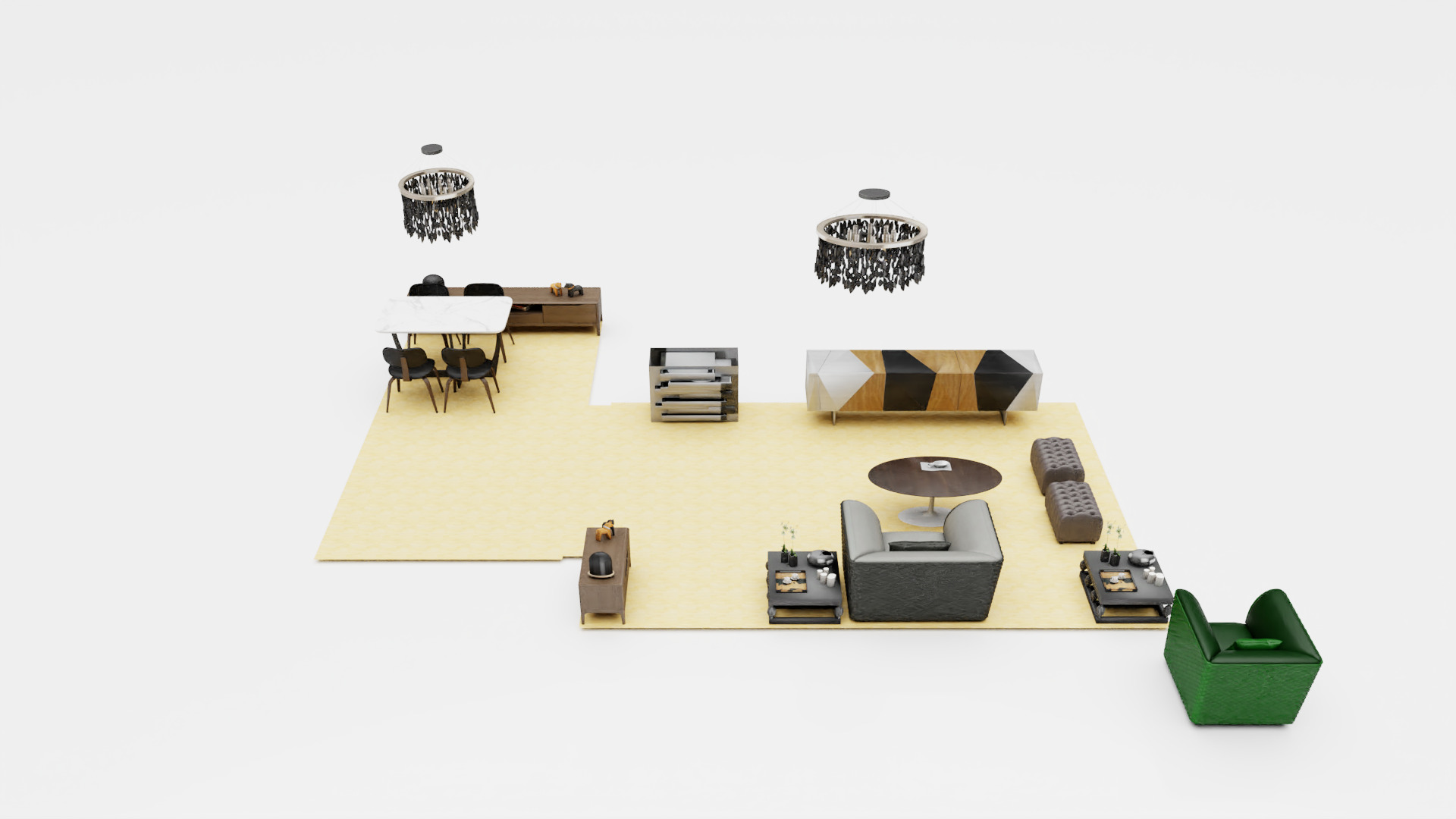}
    \end{subfigure}%
    \begin{subfigure}[b]{0.16\linewidth}
        \centering
        \includegraphics[width=\linewidth, trim=500 270 500 125, clip]{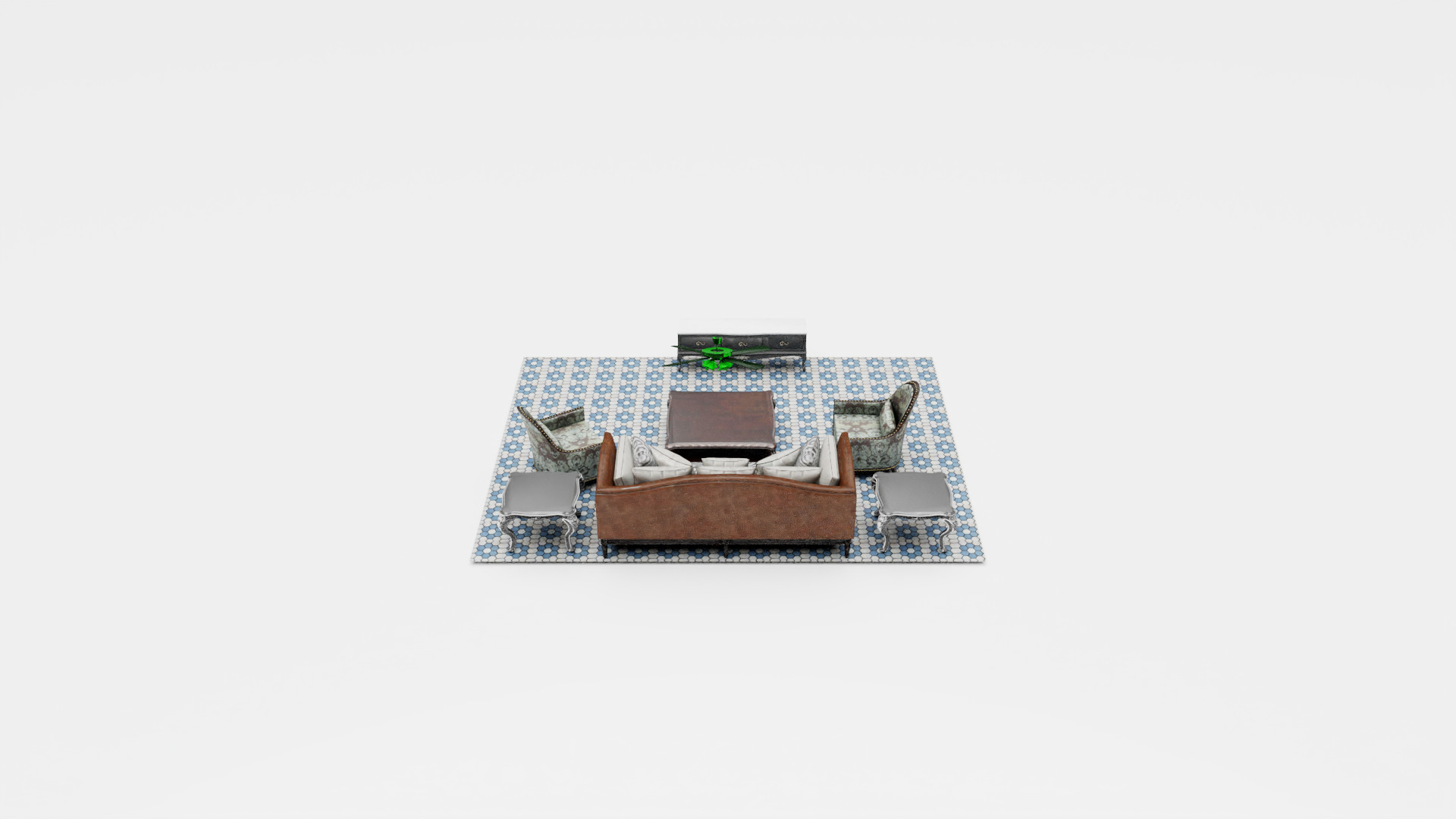}
    \end{subfigure}%
    \begin{subfigure}[b]{0.16\linewidth}
		\centering
        \includegraphics[width=\linewidth, trim=500 270 500 125, clip]{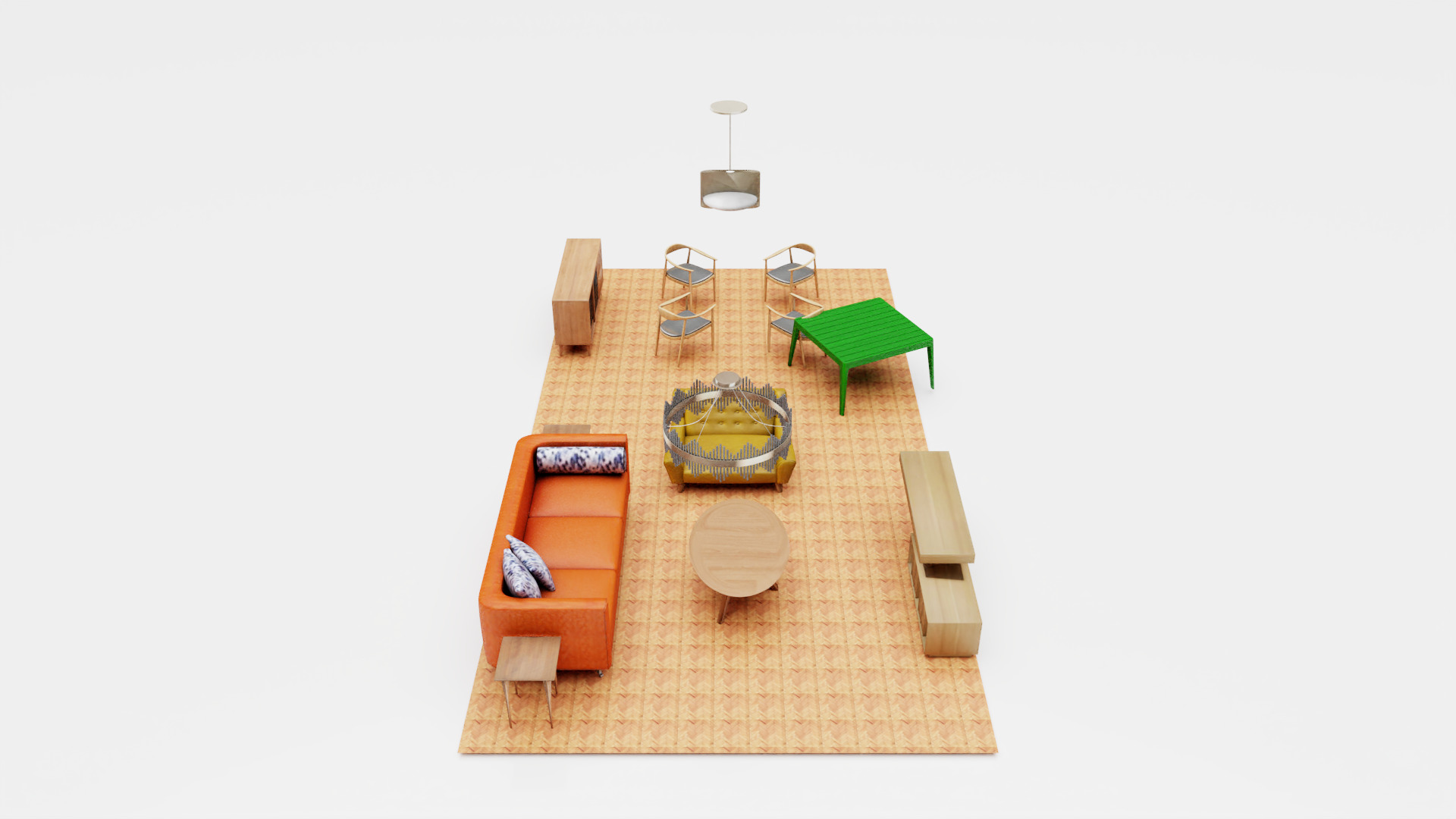}
    \end{subfigure}%
    \begin{subfigure}[b]{0.16\linewidth}
        \centering
        \includegraphics[width=\linewidth, trim=500 270 500 125, clip]{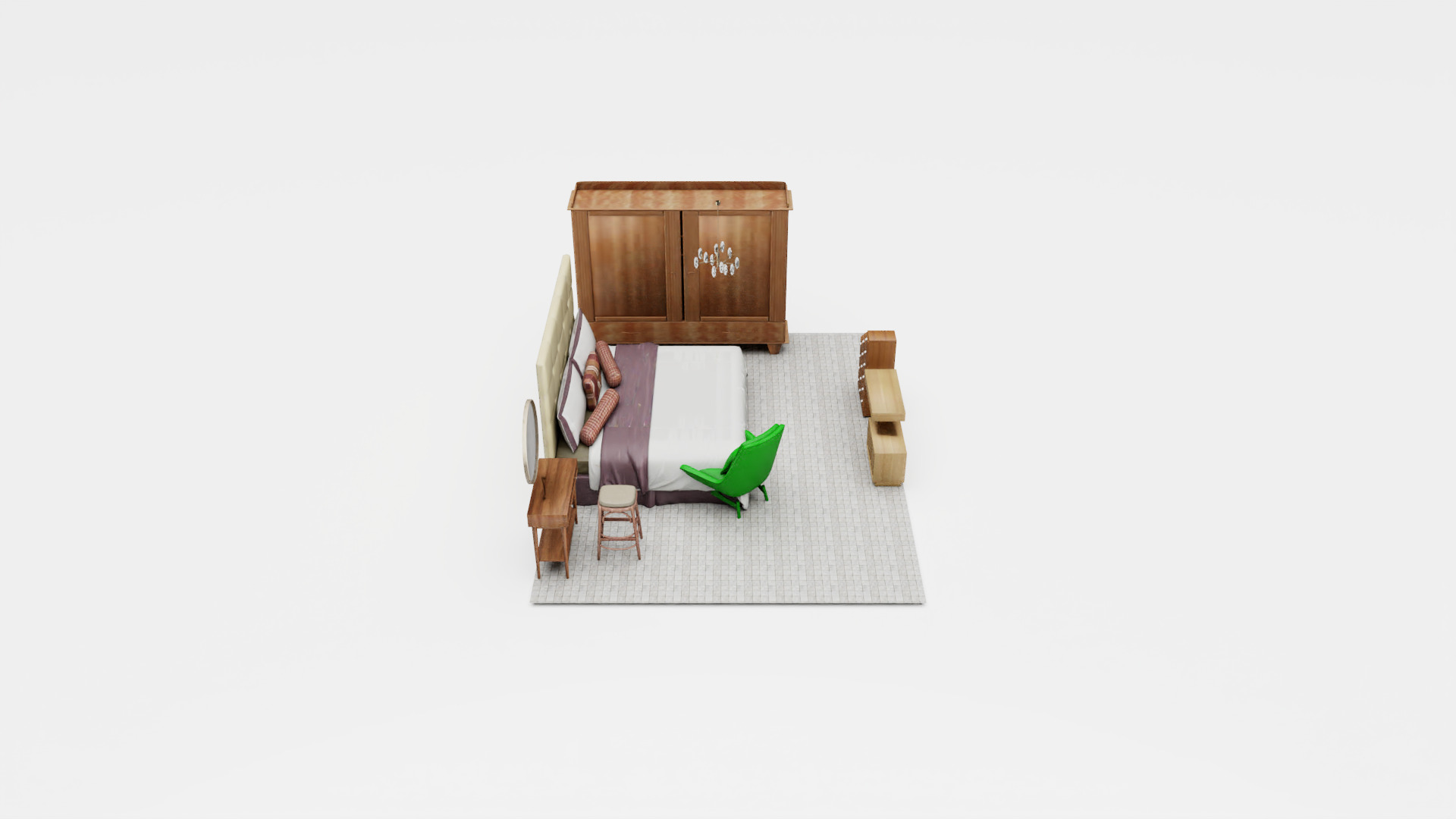}
    \end{subfigure}%
    \begin{subfigure}[b]{0.16\linewidth}
		\centering
		\includegraphics[width=\linewidth, trim=500 270 500 125, clip]{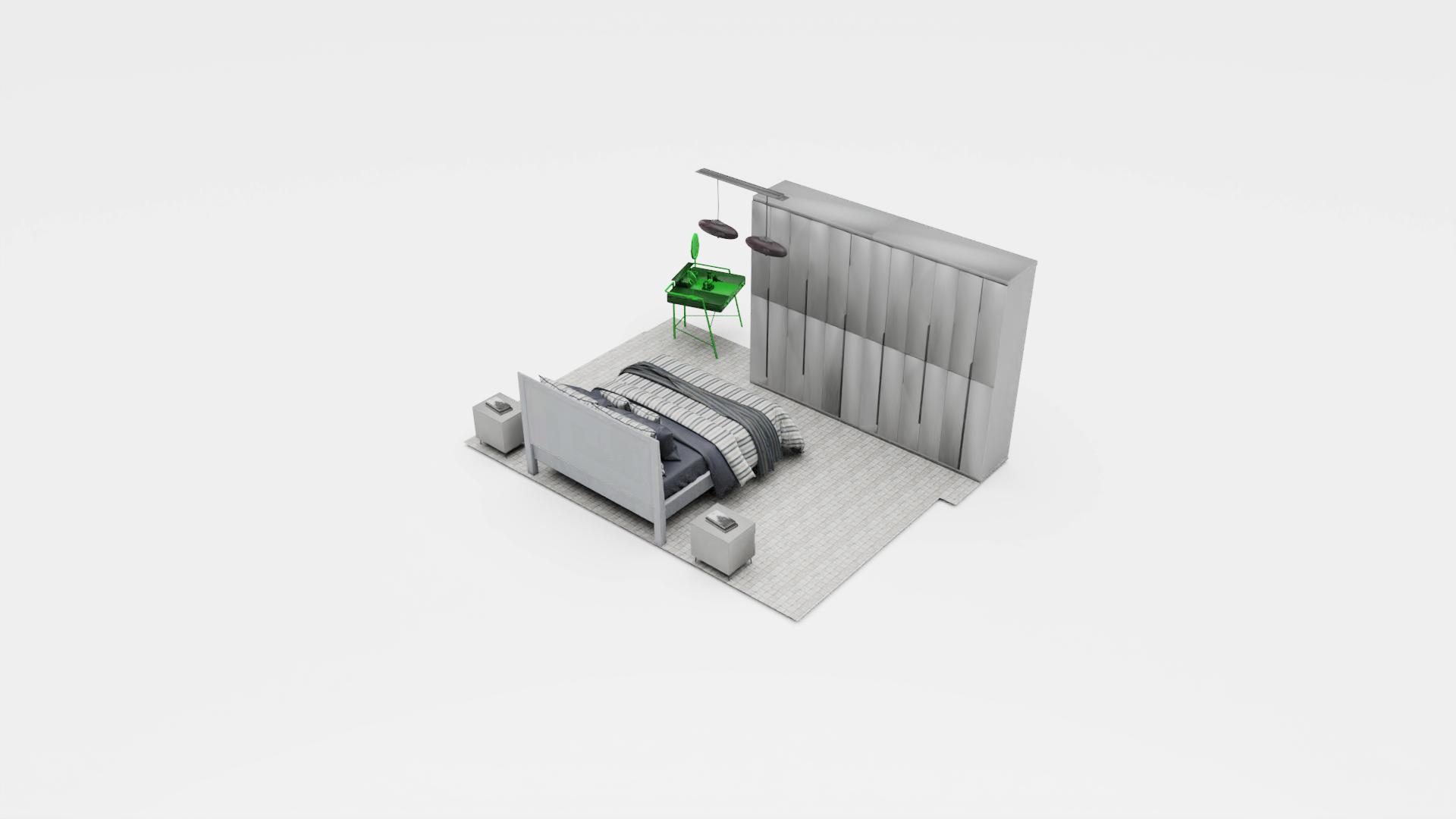}
    \end{subfigure}%
    \vspace{-1.2em}
    \vskip\baselineskip%
    \begin{subfigure}[b]{0.16\linewidth}
    \centering
    \includegraphics[width=\linewidth, trim=500 190 400 125, clip]{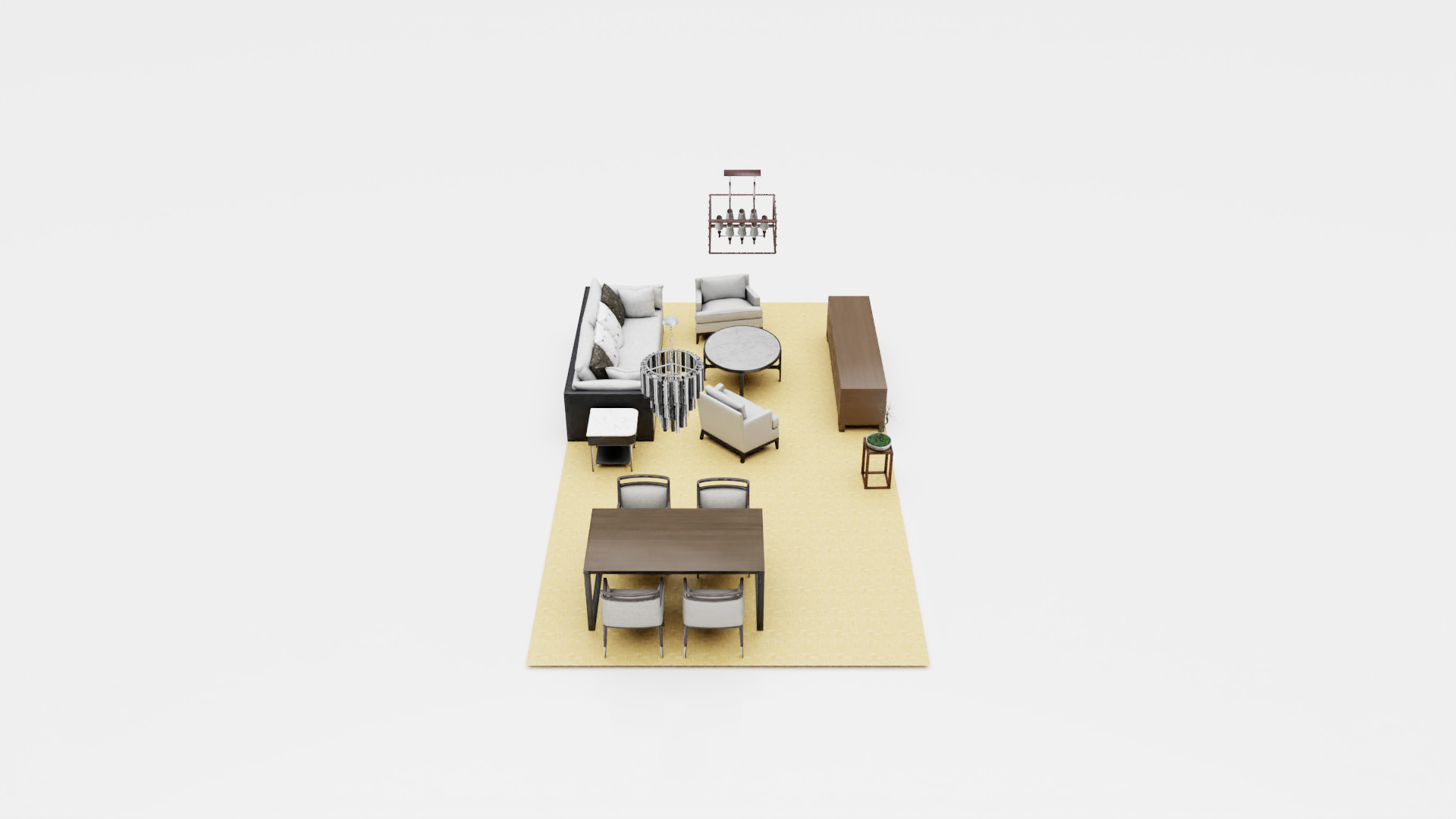}
    \end{subfigure}%
    \begin{subfigure}[b]{0.16\linewidth}
		\centering
        \includegraphics[width=\linewidth, trim=350 10 150 10, clip]{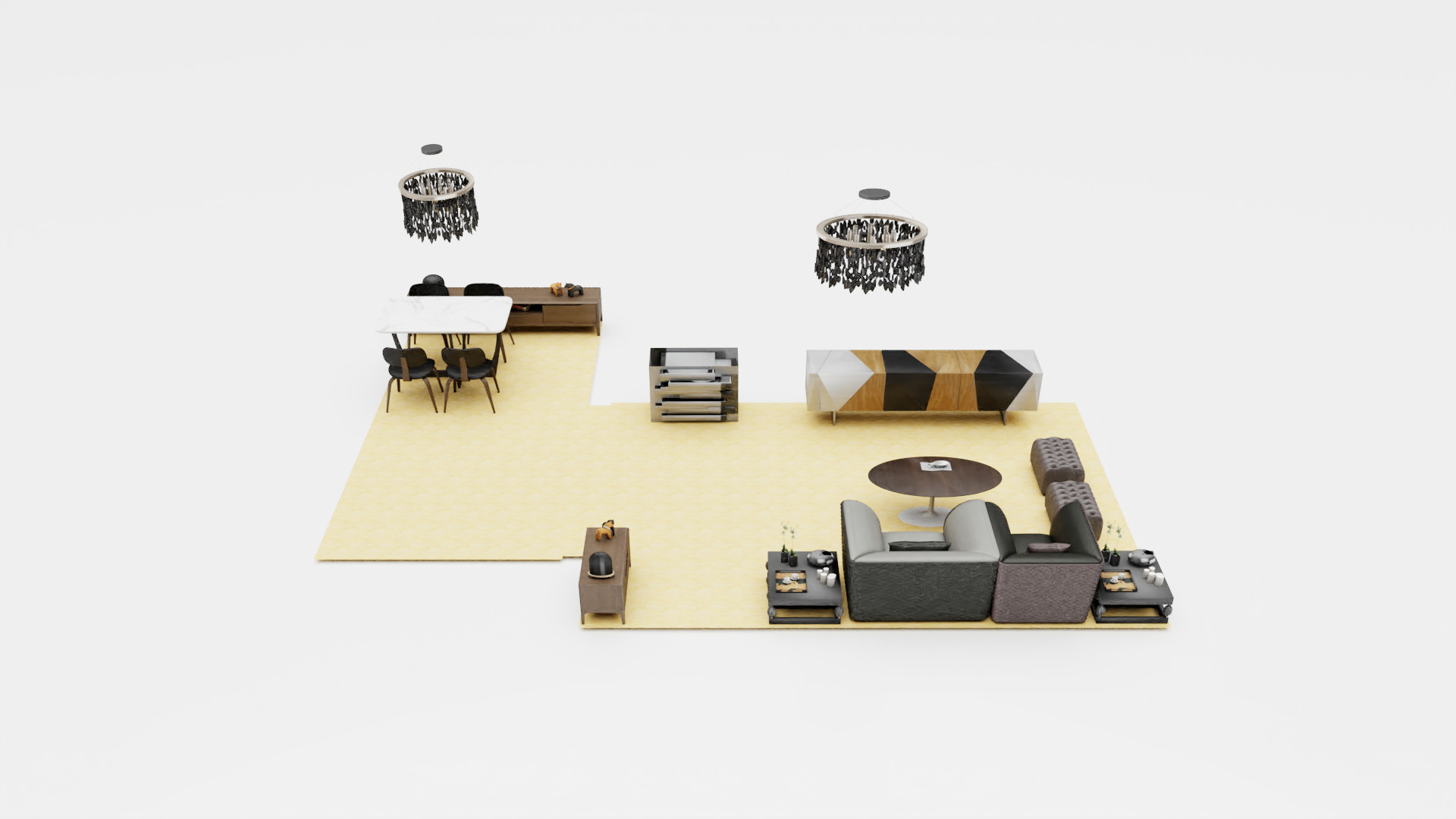}
    \end{subfigure}%
    \begin{subfigure}[b]{0.16\linewidth}
        \centering
        \includegraphics[width=\linewidth, trim=500 270 500 125, clip]{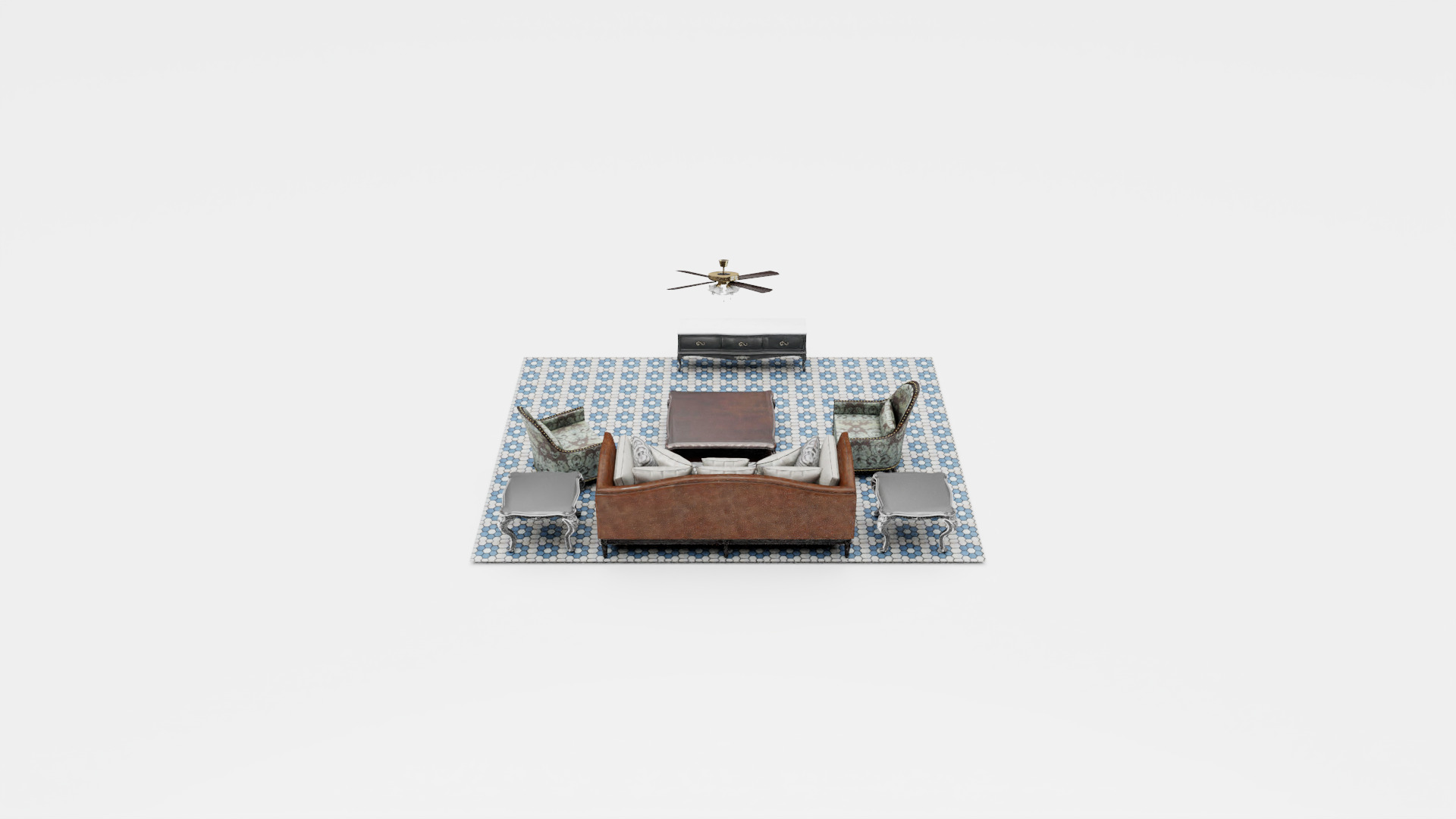}
    \end{subfigure}%
    \begin{subfigure}[b]{0.16\linewidth}
		\centering
        \includegraphics[width=\linewidth, trim=500 270 500 125, clip]{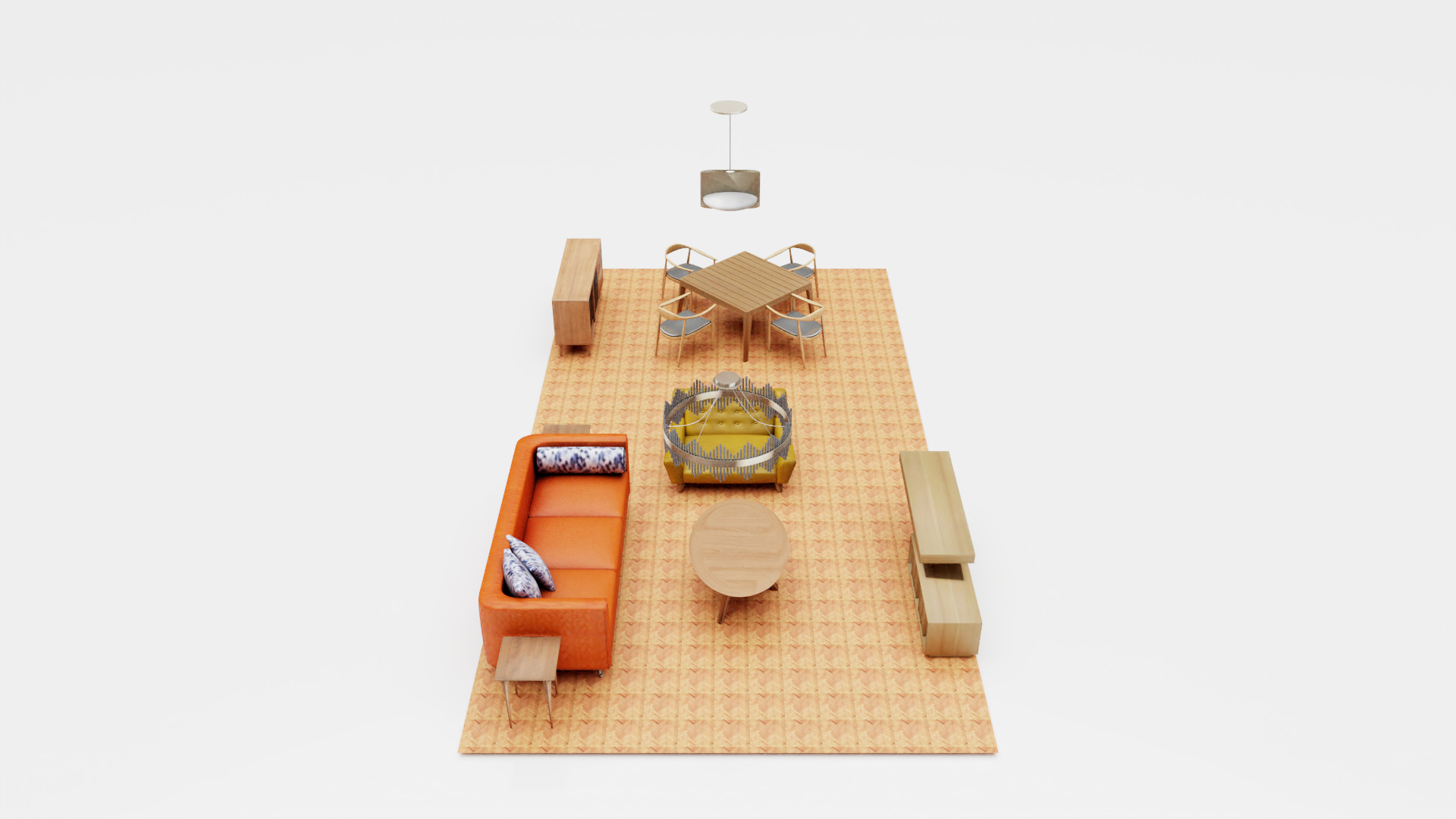}
    \end{subfigure}%
    \begin{subfigure}[b]{0.16\linewidth}
    \centering
        \includegraphics[width=\linewidth, trim=500 270 500 125, clip]{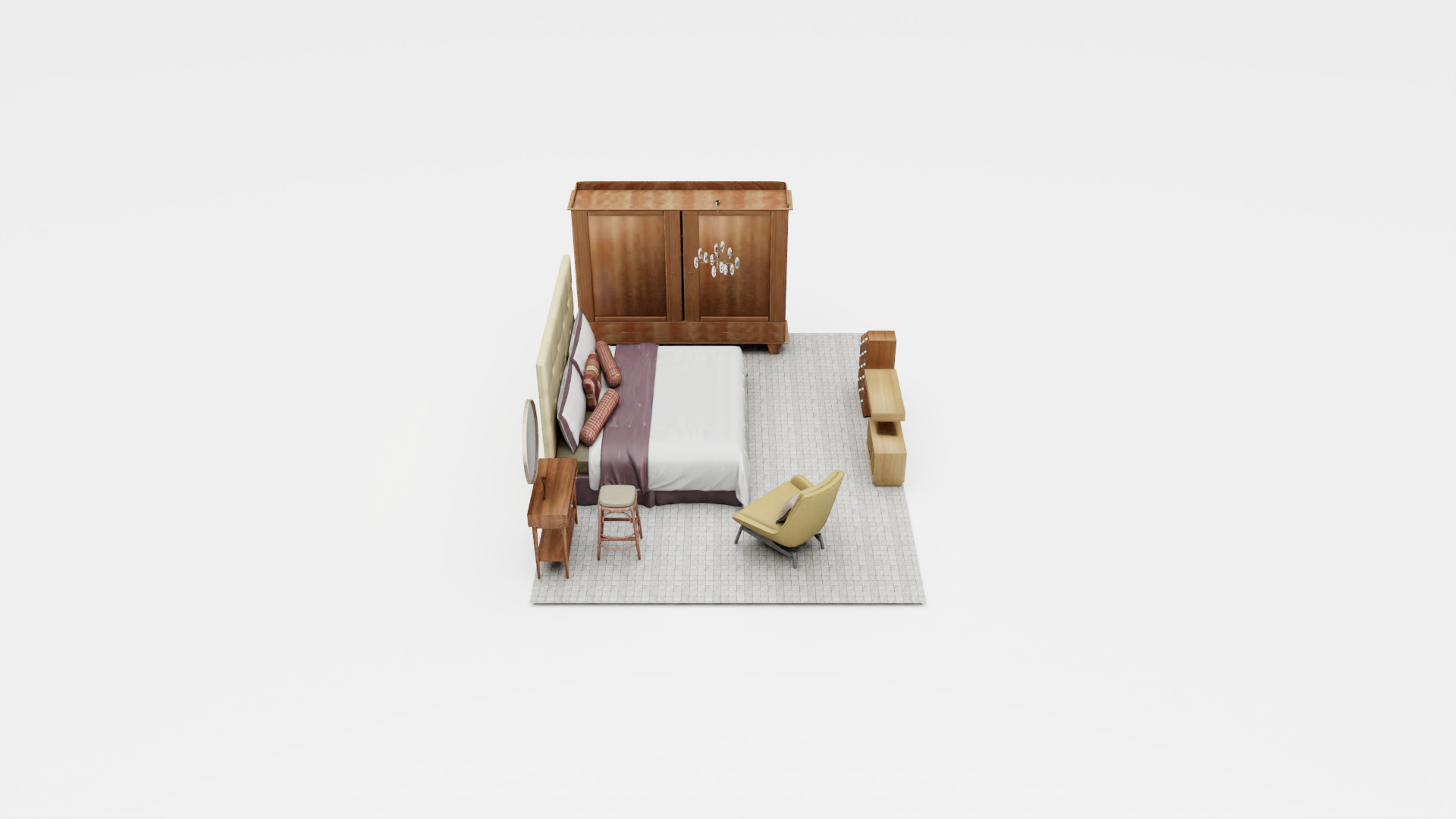}
    \end{subfigure}    \begin{subfigure}[b]{0.16\linewidth}
		\centering
		\includegraphics[width=\linewidth, trim=500 270 500 125, clip]{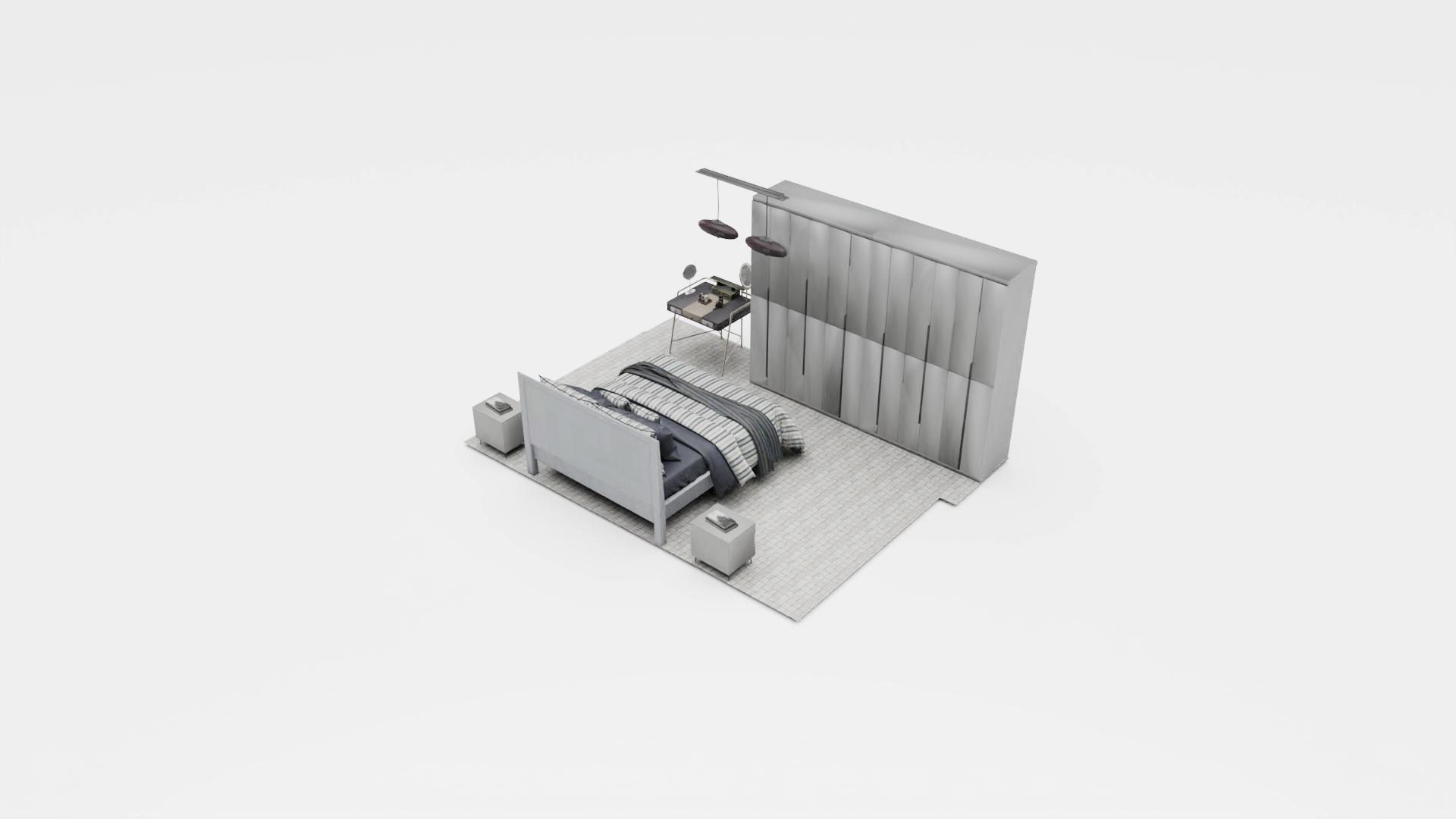}
    \end{subfigure}%
    \vspace{-1.2em}
    \vskip\baselineskip%
    \caption{{\bf Failure Case Detection and Correction}. Starting from
    a room with an unnatural object arrangement, our model identifies the
    problematic objects (first row and third row, in green) and relocates them into
    meaningful positions (second and fourth row).}
    \label{fig:failure_cases_correction_supp}
\end{figure}

In this experiment, we investigate whether our model is able to identify
unnatural furniture layouts and reposition the problematic objects such that
they preserve their functional properties. As we described in our main
submission, we identify problematic objects as those with low likelihood and as
soon as a problematic object is identified, we sample a new location from our
generative model to reposition it. \figref{fig:failure_cases_correction_supp}
shows additional qualitative results on this task. The first and third row show
examples of unnatural object arrangements, together with the problematic
object, highlighted in green, for each scenario. We note that our model
successfully identifies objects in unnatural positions \eg flying bed (first
row, first column \figref{fig:failure_cases_correction_supp}), light inside the
bed (first row, third column \figref{fig:failure_cases_correction_supp}) or table outside the room boundaries (third row,
fourth column \figref{fig:failure_cases_correction_supp} ) as well as problematic objects that do not necessarily look
unnatural, such as a cabinet blocking the corridor (first row, sixth column \figref{fig:failure_cases_correction_supp}), a
chair facing the wall (third row, first column \figref{fig:failure_cases_correction_supp}) or a lamp being too close to
the table (third row, third column \figref{fig:failure_cases_correction_supp}). After having identified the problematic
object, our model consistently repositions it at plausible position.

\subsection{Object Suggestion}
\begin{figure}
    \centering
    \begin{subfigure}[b]{0.16\linewidth}
        \centering
        \includegraphics[width=\linewidth, trim=500 150 500 150 , clip]{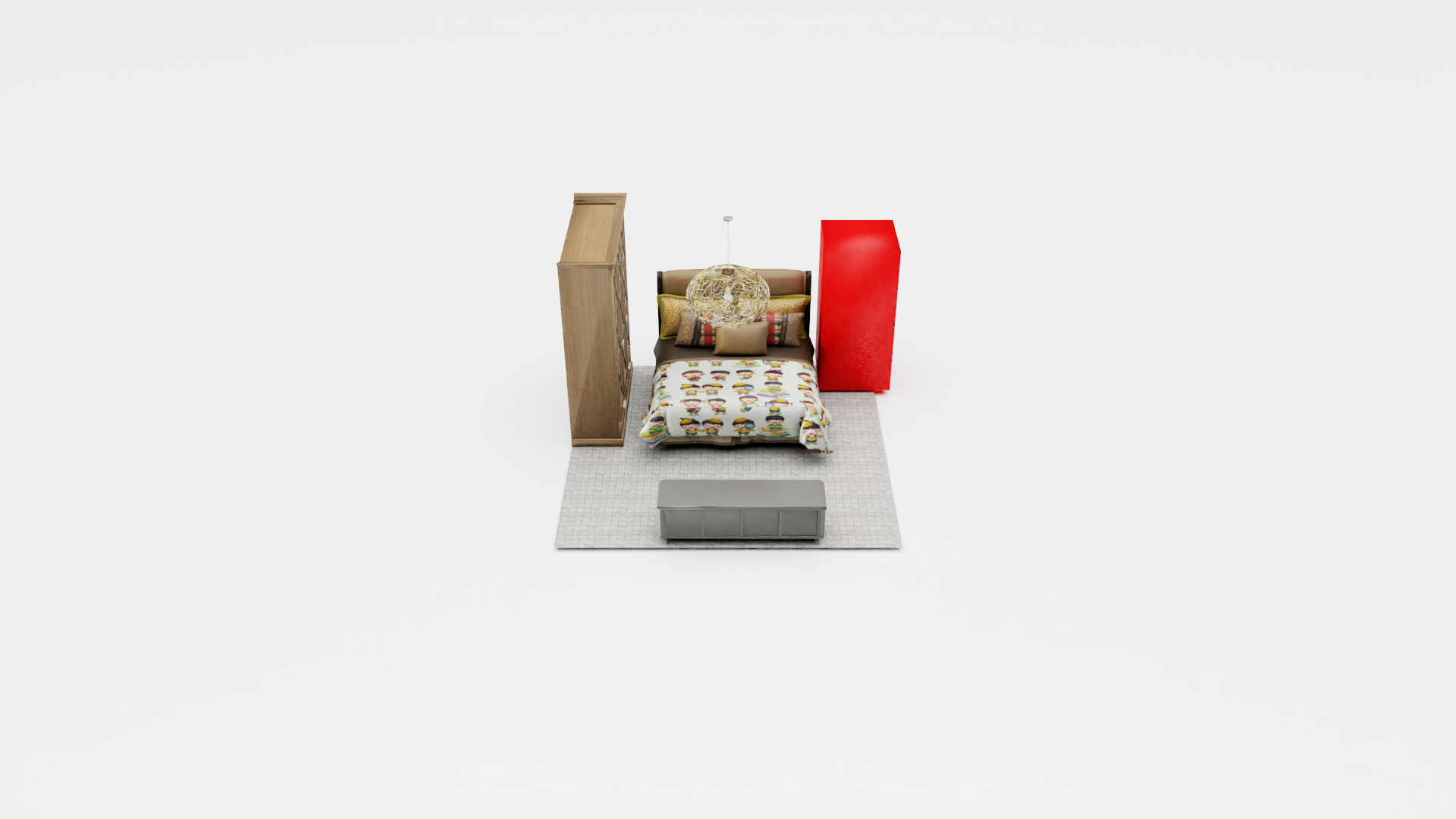}
    \end{subfigure}%
    \begin{subfigure}[b]{0.16\linewidth}
		\centering
        \includegraphics[width=\linewidth, trim=500 150 500 150 , clip]{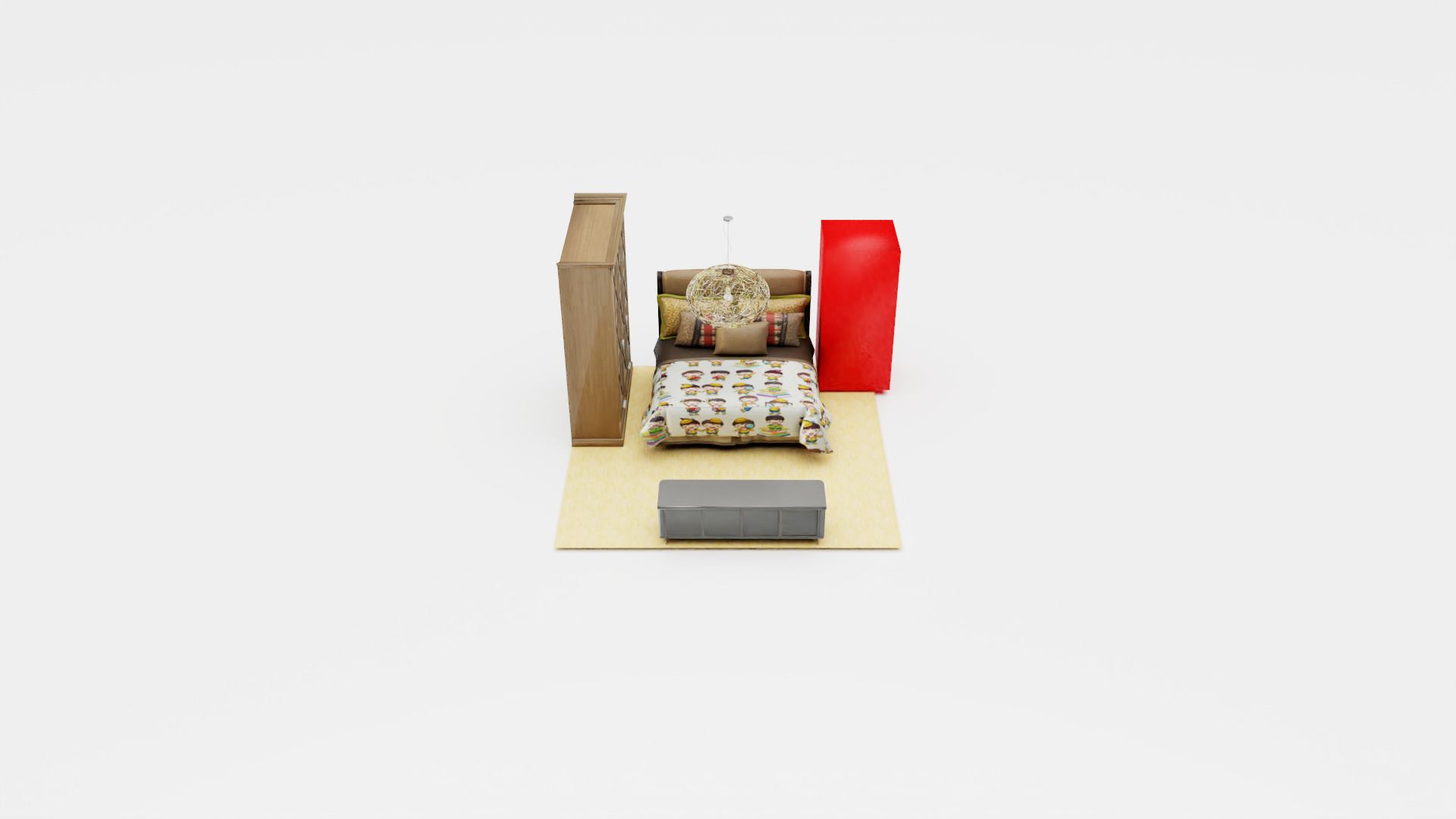}
    \end{subfigure}%
    \begin{subfigure}[b]{0.16\linewidth}
        \centering
        \includegraphics[width=\linewidth, trim=500 150 500 150 , clip]{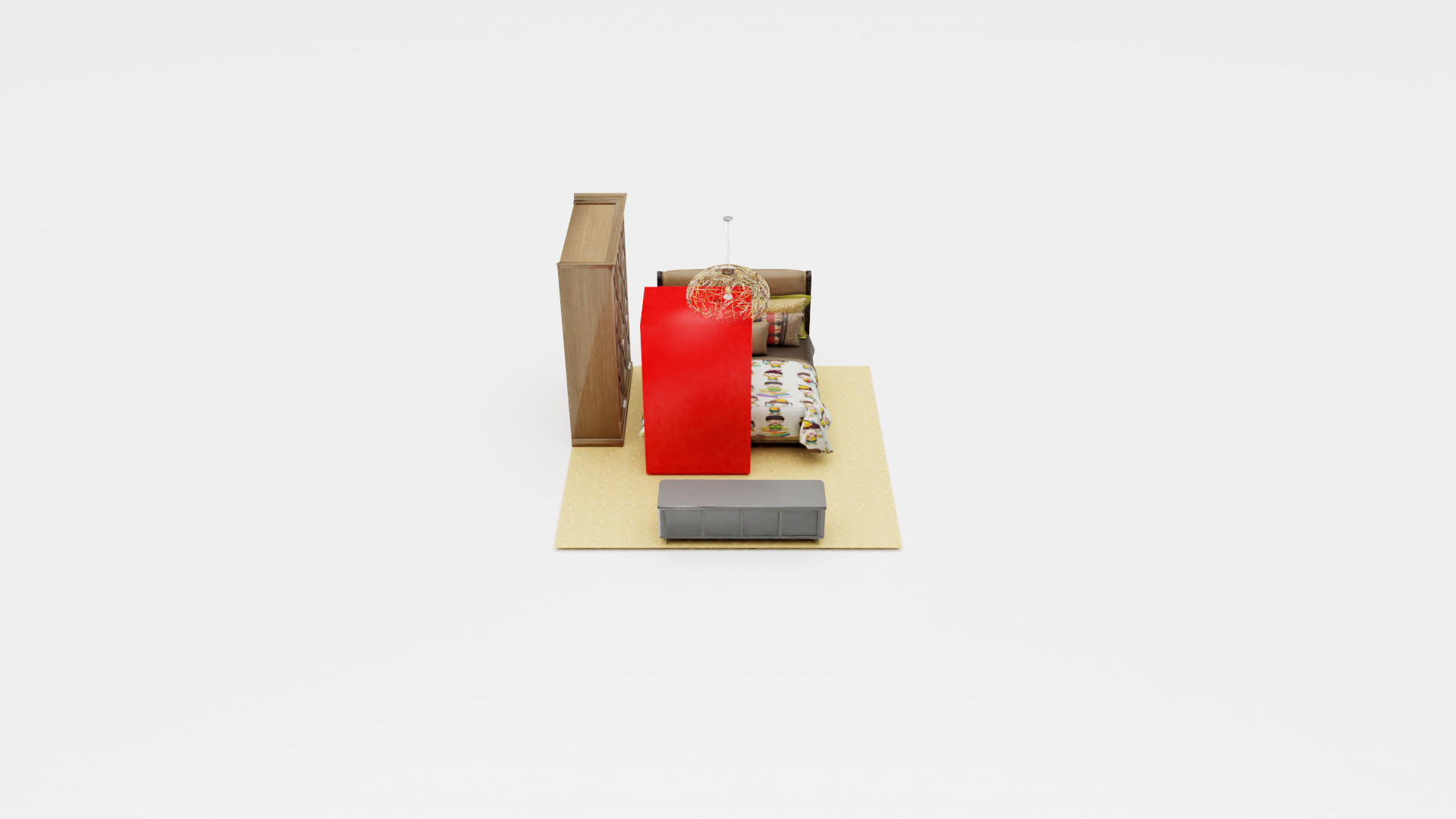}
    \end{subfigure}%
    \begin{subfigure}[b]{0.16\linewidth}
		\centering
        \includegraphics[width=\linewidth, trim=500 170 400 50 , clip]{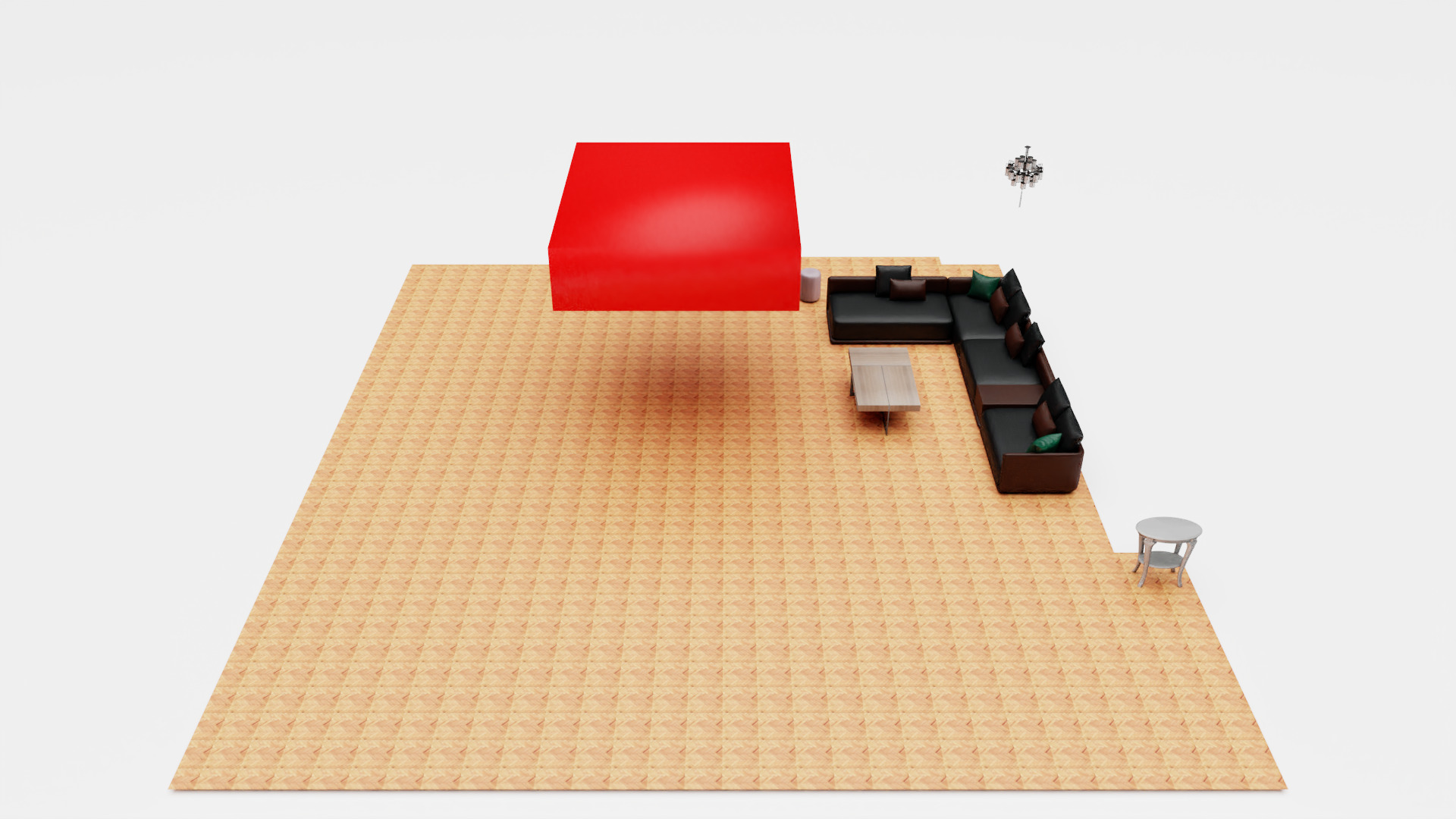}
    \end{subfigure}%
    \begin{subfigure}[b]{0.16\linewidth}
        \centering
        \includegraphics[width=\linewidth, trim=500 50 500 250 , clip]{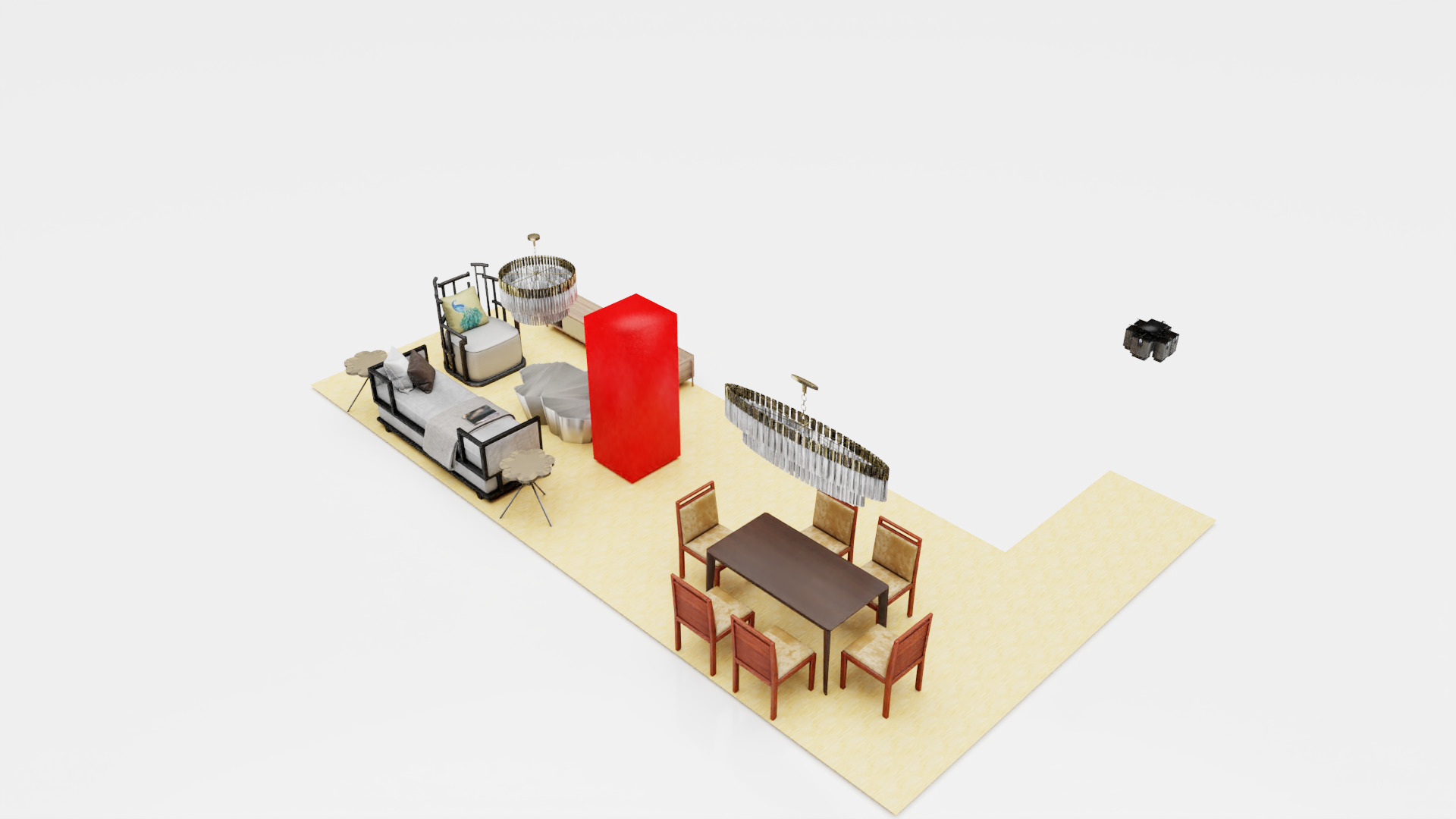}
    \end{subfigure}%
    \begin{subfigure}[b]{0.16\linewidth}
		\centering
        \includegraphics[width=\linewidth, trim=500 50 500 250 , clip]{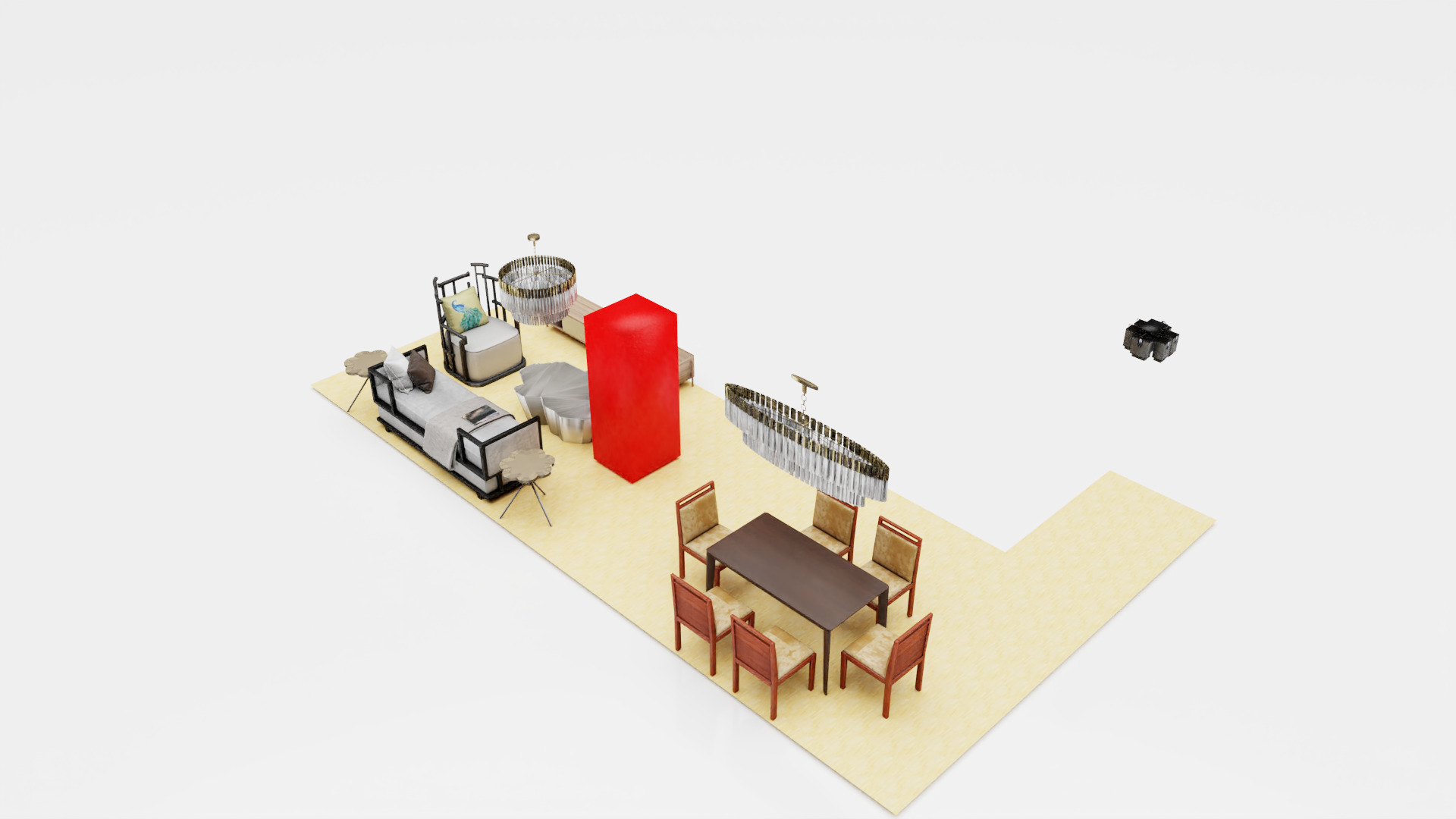}
    \end{subfigure}%
    \vspace{-1.2em}
    \vskip\baselineskip%
    \begin{subfigure}[b]{0.16\linewidth}
        \centering
        \includegraphics[width=\linewidth, trim=500 150 500 150 , clip]{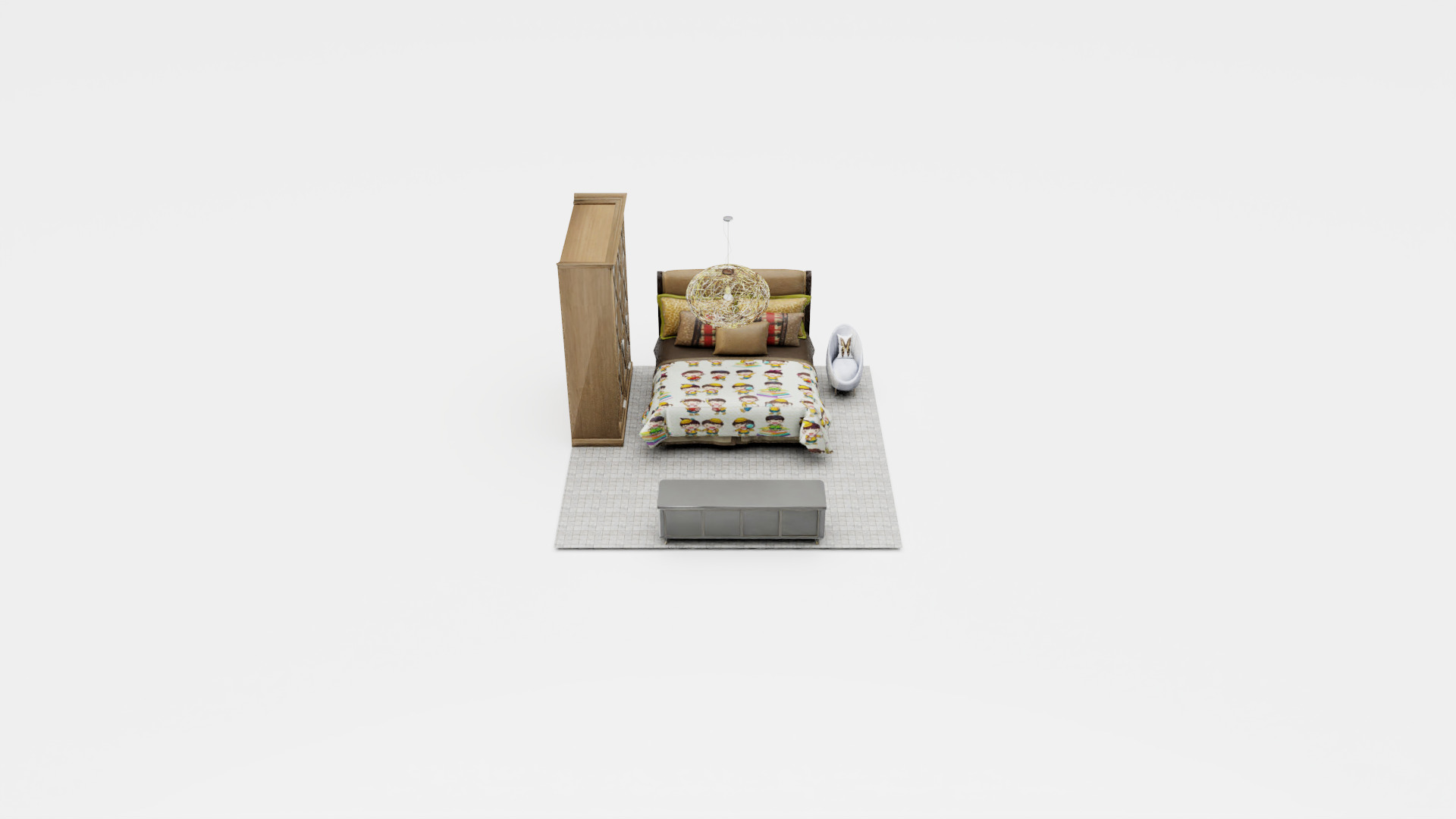}
    \end{subfigure}%
    \begin{subfigure}[b]{0.16\linewidth}
		\centering
        \includegraphics[width=\linewidth, trim=500 150 500 150 , clip]{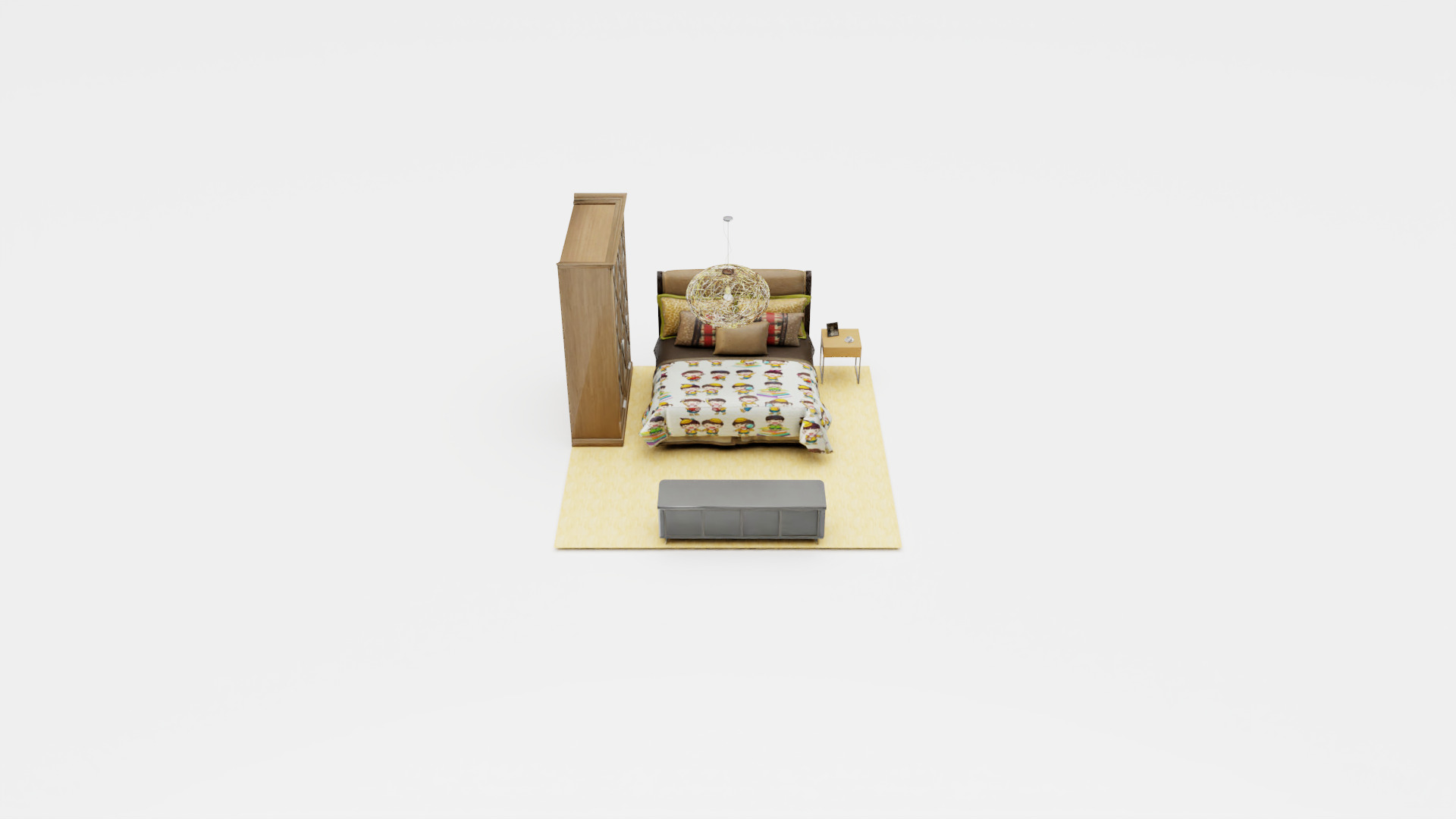}
    \end{subfigure}%
    \begin{subfigure}[b]{0.16\linewidth}
        \centering
        \includegraphics[width=\linewidth, trim=500 150 500 150 , clip]{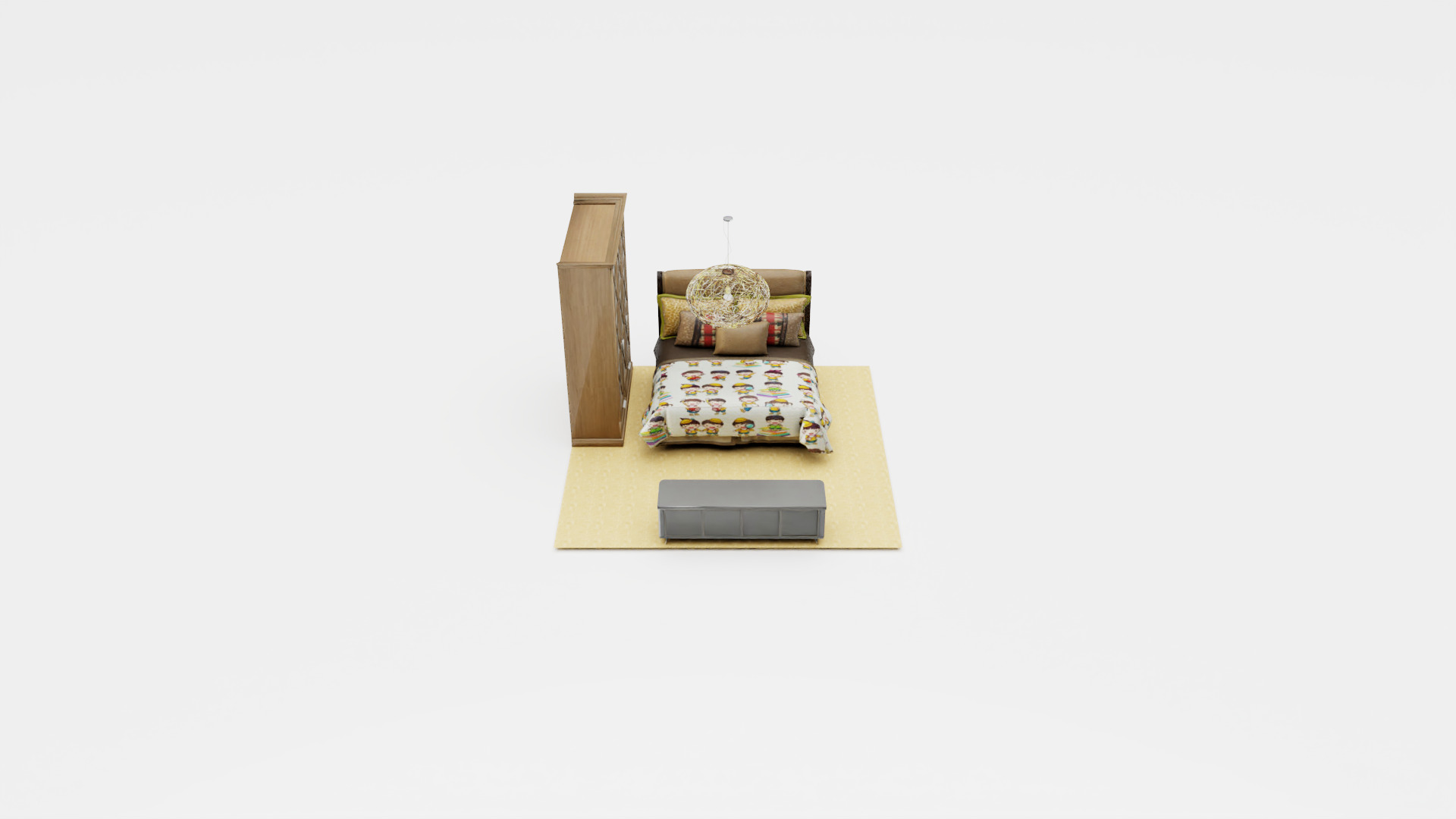}
    \end{subfigure}%
    \begin{subfigure}[b]{0.16\linewidth}
		\centering
        \includegraphics[width=\linewidth, trim=500 170 400 50 , clip]{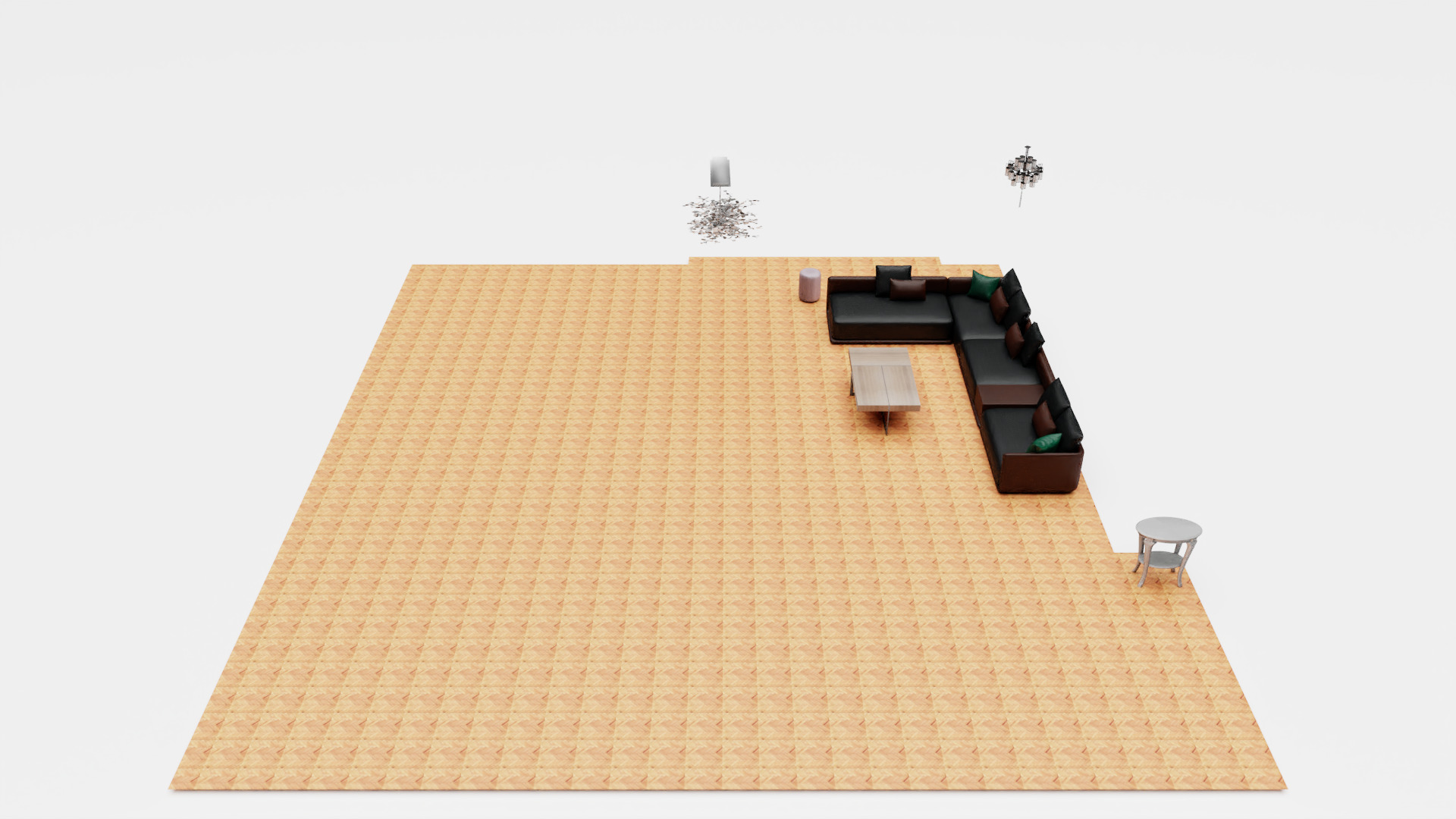}
    \end{subfigure}%
    \begin{subfigure}[b]{0.16\linewidth}
    \centering
        \includegraphics[width=\linewidth, trim=500 50 500 250 , clip]{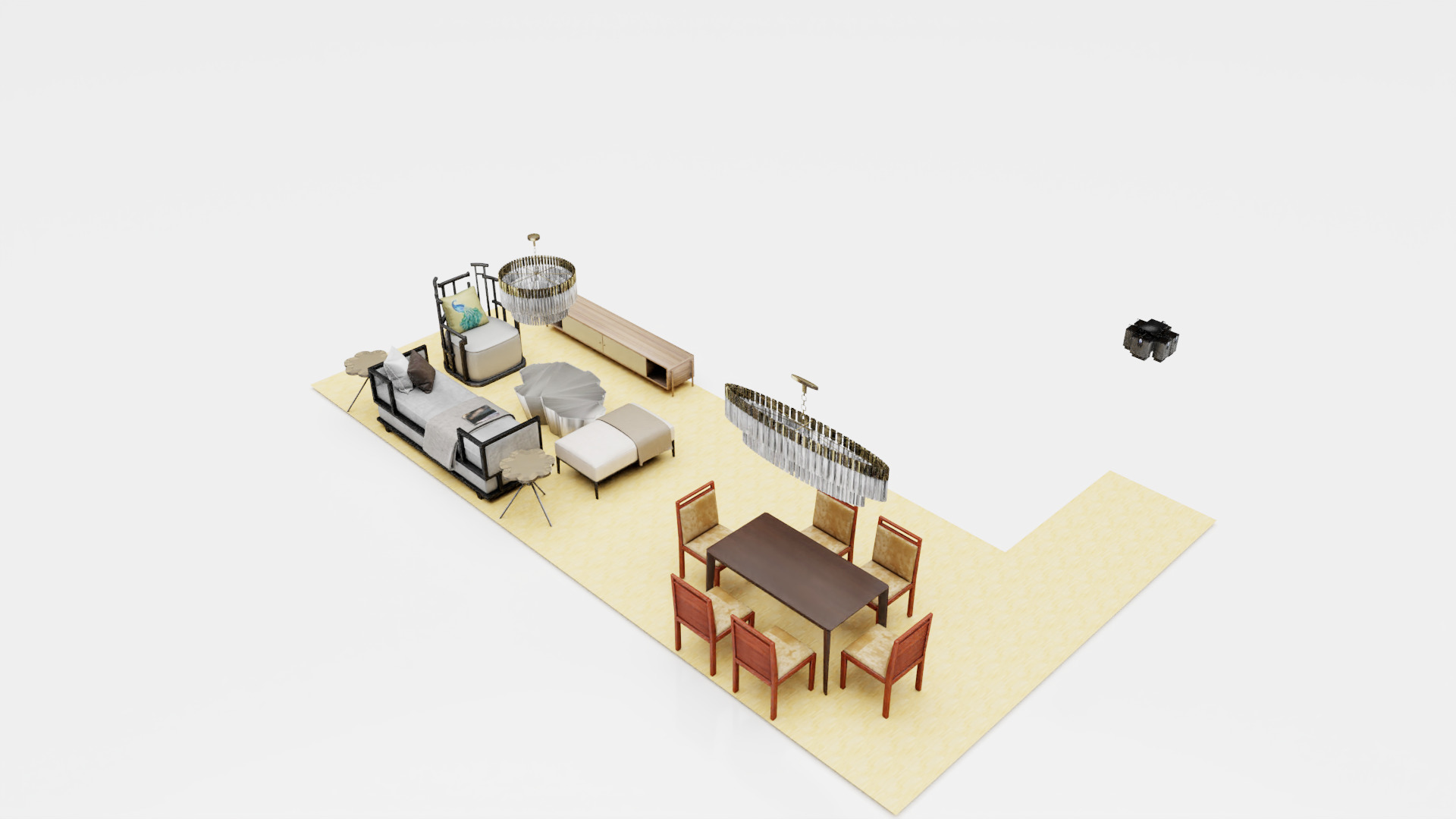}
    \end{subfigure}    \begin{subfigure}[b]{0.16\linewidth}
		\centering
        \includegraphics[width=\linewidth, trim=500 50 500 250 , clip]{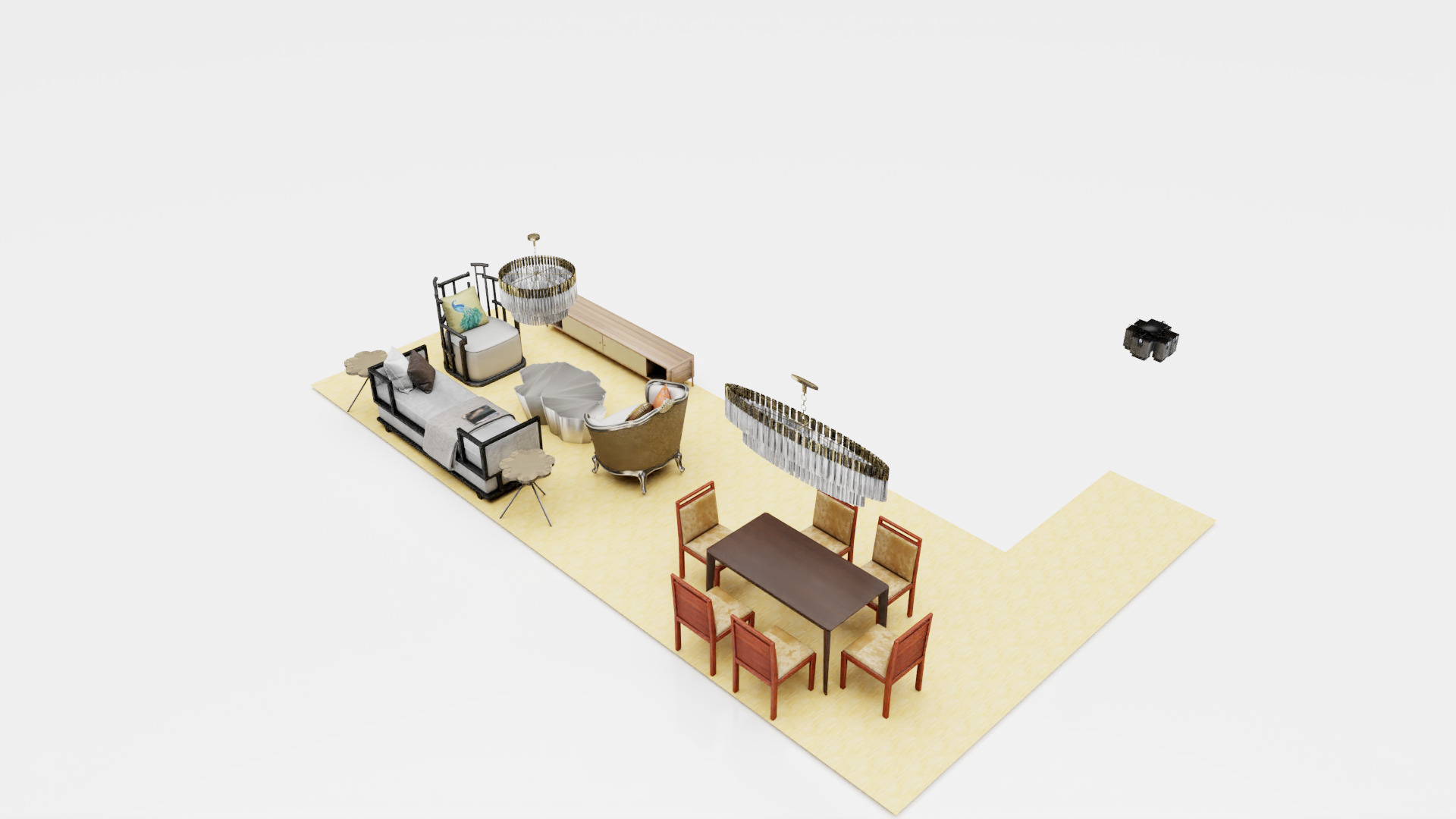}
    \end{subfigure}%
    \vskip\baselineskip%
    \vspace{-1.5em}
    \hfill%
    \begin{subfigure}[b]{0.16\linewidth}
		\centering
        \small Sofa
    \end{subfigure}%
    \begin{subfigure}[b]{0.16\linewidth}
		\centering
        \small Nightstand
    \end{subfigure}%
    \begin{subfigure}[b]{0.16\linewidth}
		\centering
        \small Nothing
    \end{subfigure}%
    \begin{subfigure}[b]{0.16\linewidth}
		\centering
        \small Lamp
    \end{subfigure}%
    \begin{subfigure}[b]{0.16\linewidth}
		\centering
        \small Stool
    \end{subfigure}%
    \begin{subfigure}[b]{0.16\linewidth}
		\centering
        \small Armchair
    \end{subfigure}%
    \hfill%
        \vskip\baselineskip%
    \begin{subfigure}[b]{0.16\linewidth}
    \centering
    \includegraphics[width=\linewidth, trim=500 150 500 150 , clip]{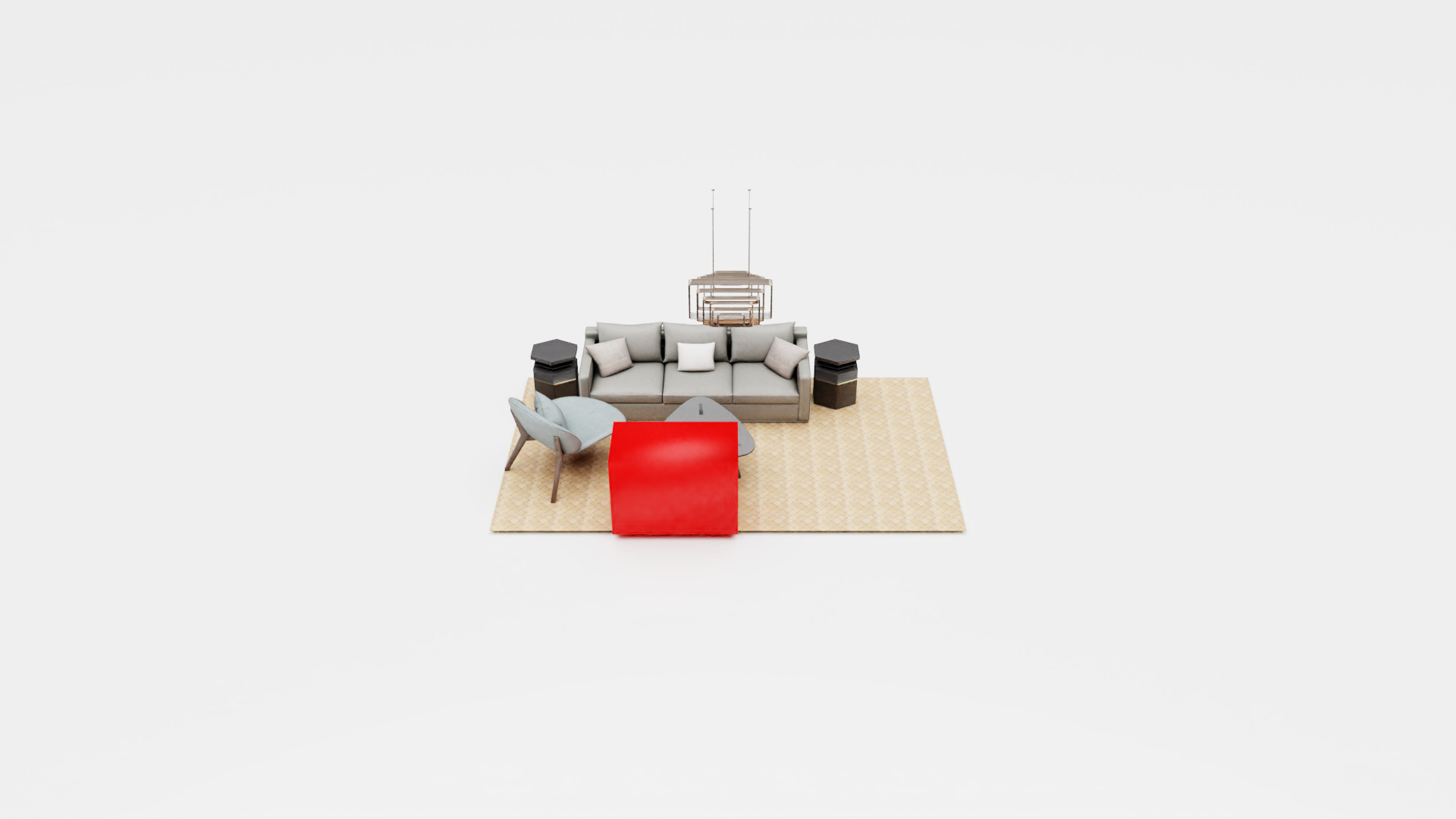}
    \end{subfigure}%
    \begin{subfigure}[b]{0.16\linewidth}
		\centering
        \includegraphics[width=\linewidth, trim=500 150 500 150 , clip]{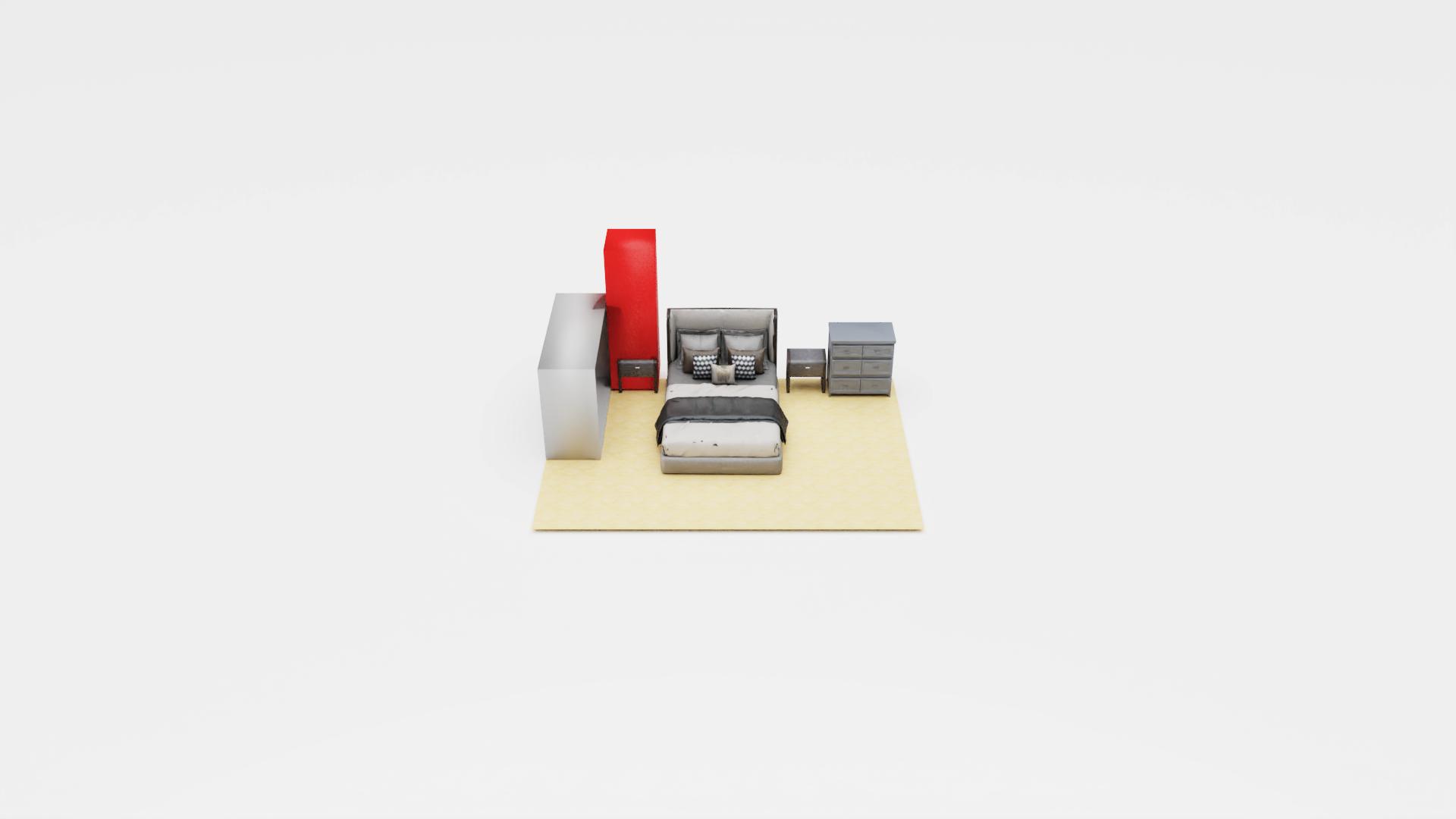}
    \end{subfigure}%
    \begin{subfigure}[b]{0.16\linewidth}
        \centering
        \includegraphics[width=\linewidth, trim=500 150 500 150 , clip]{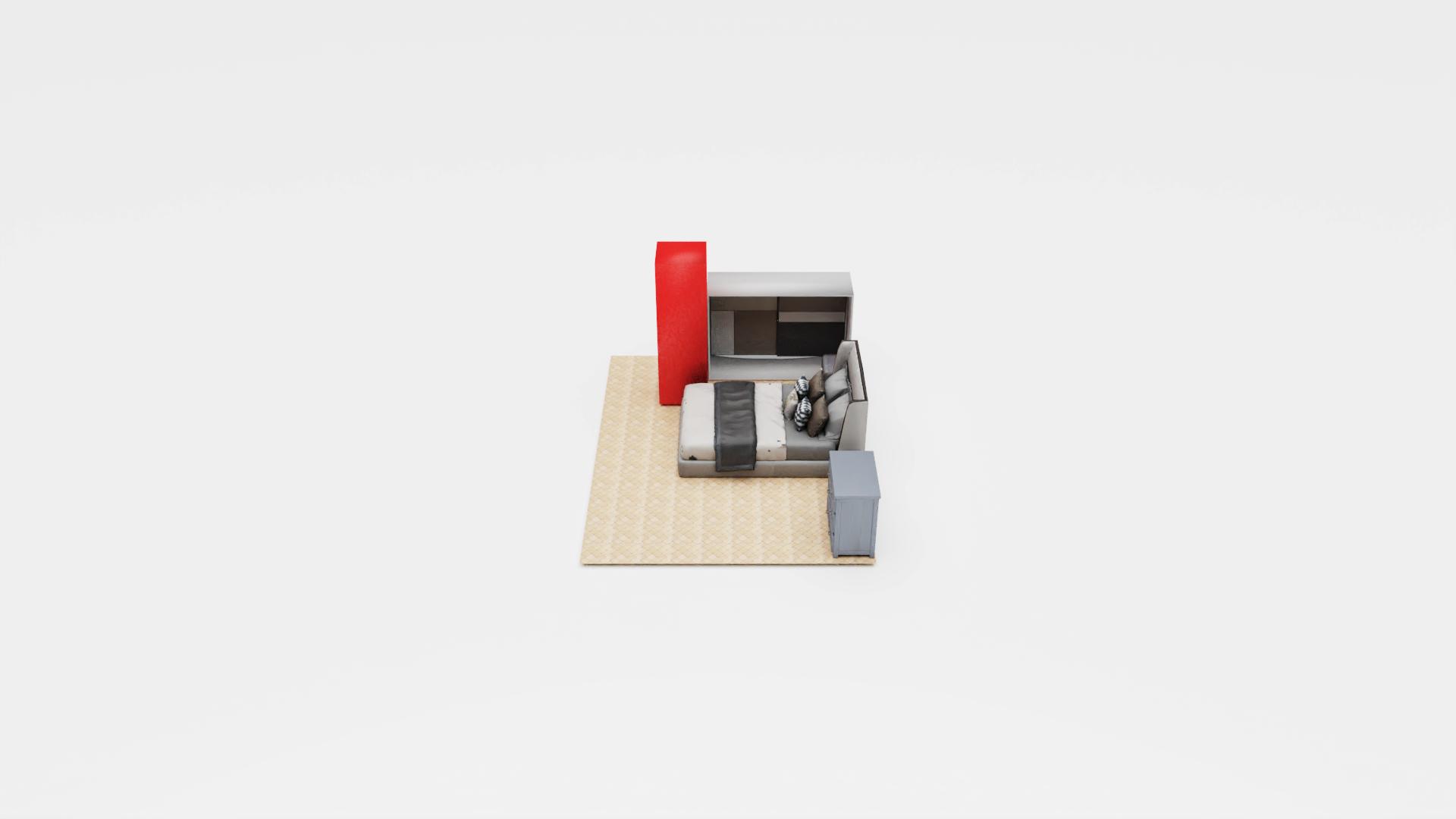}
    \end{subfigure}%
    \begin{subfigure}[b]{0.16\linewidth}
		\centering
        \includegraphics[width=\linewidth, trim=500 150 500 150 , clip]{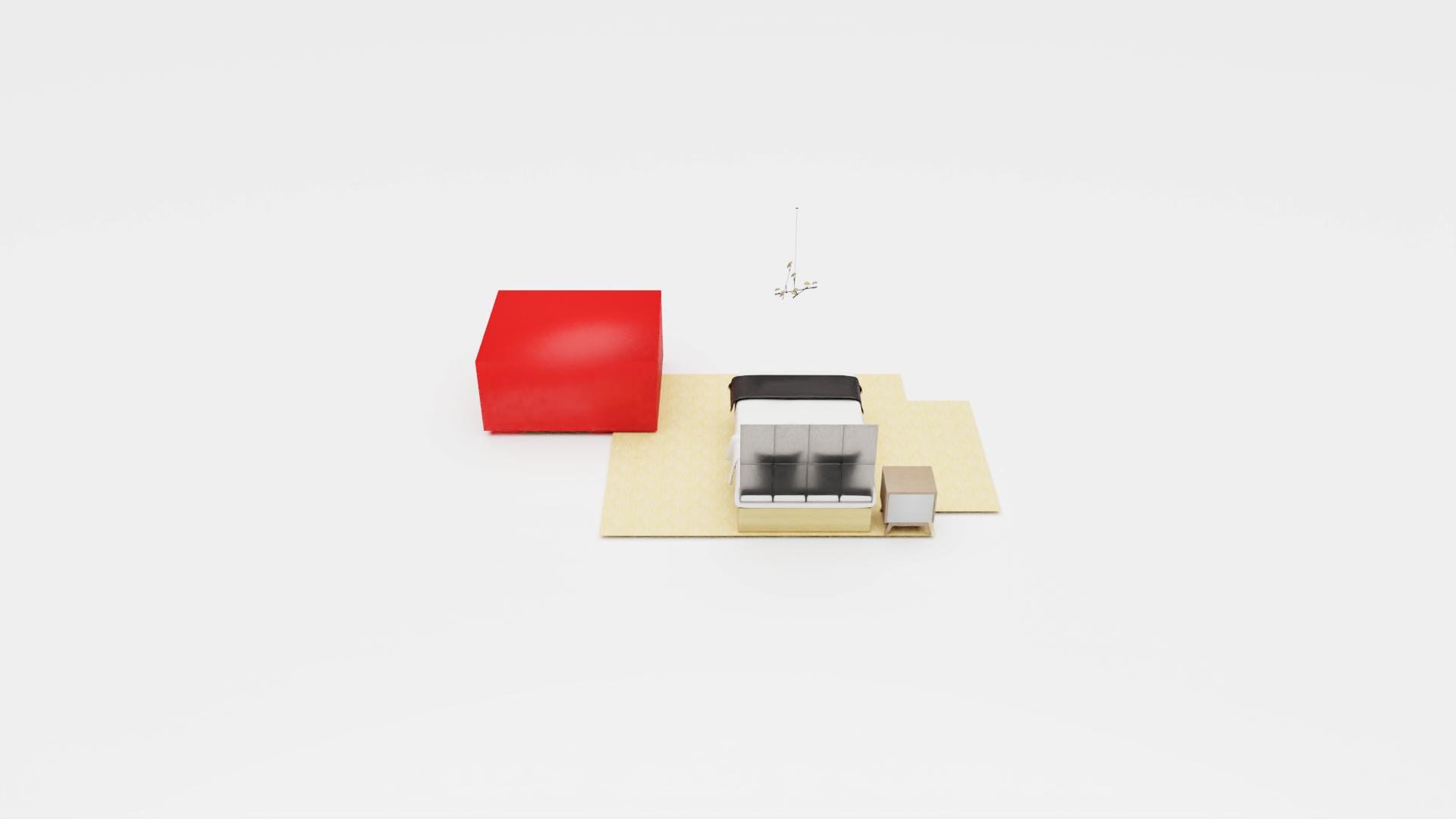}
    \end{subfigure}%
    \begin{subfigure}[b]{0.16\linewidth}
        \centering
        \includegraphics[width=\linewidth, trim=500 150 500 150 , clip]{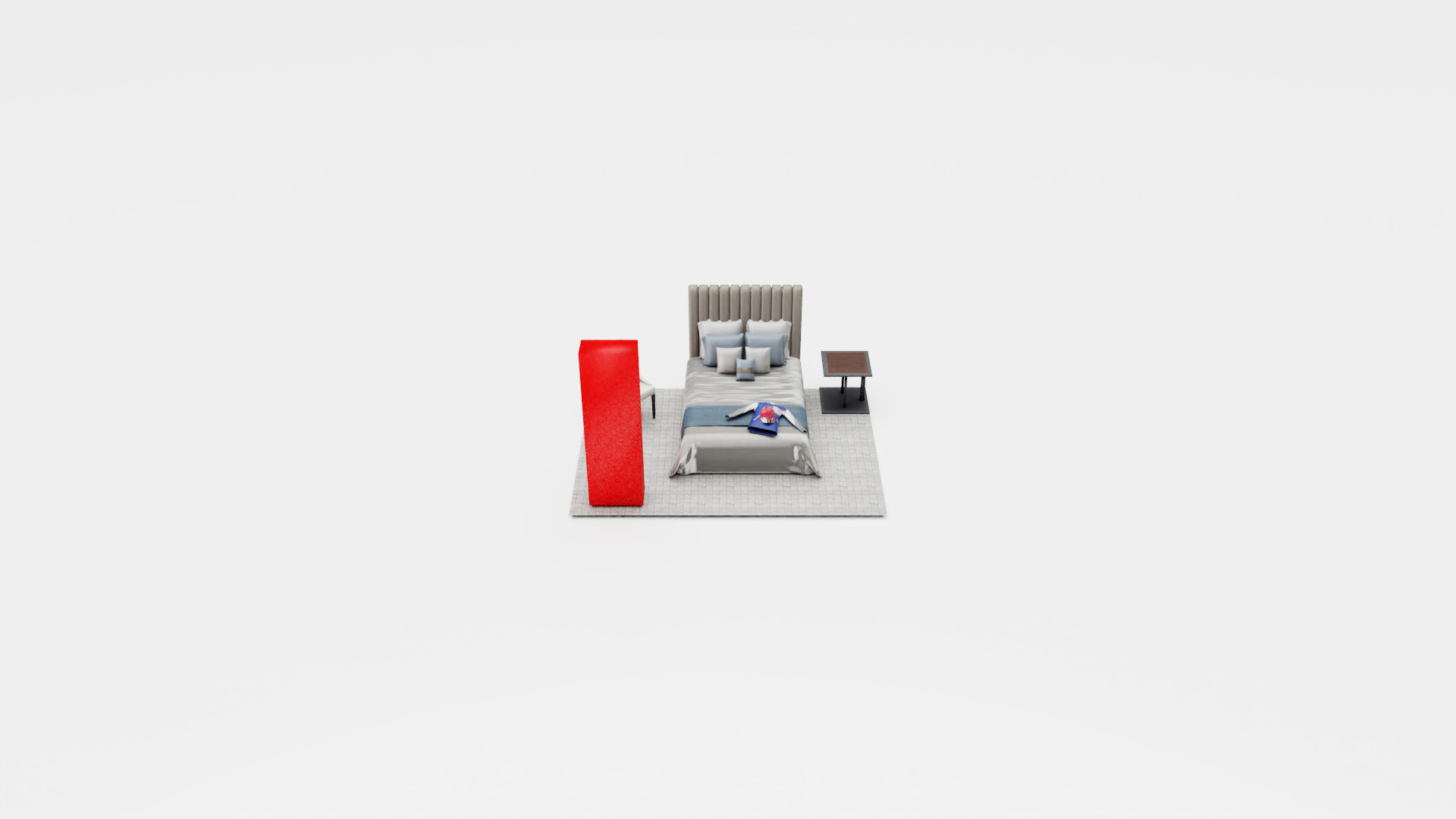}
    \end{subfigure}%
    \begin{subfigure}[b]{0.16\linewidth}
		\centering
        \includegraphics[width=\linewidth, trim=500 150 500 150 , clip]{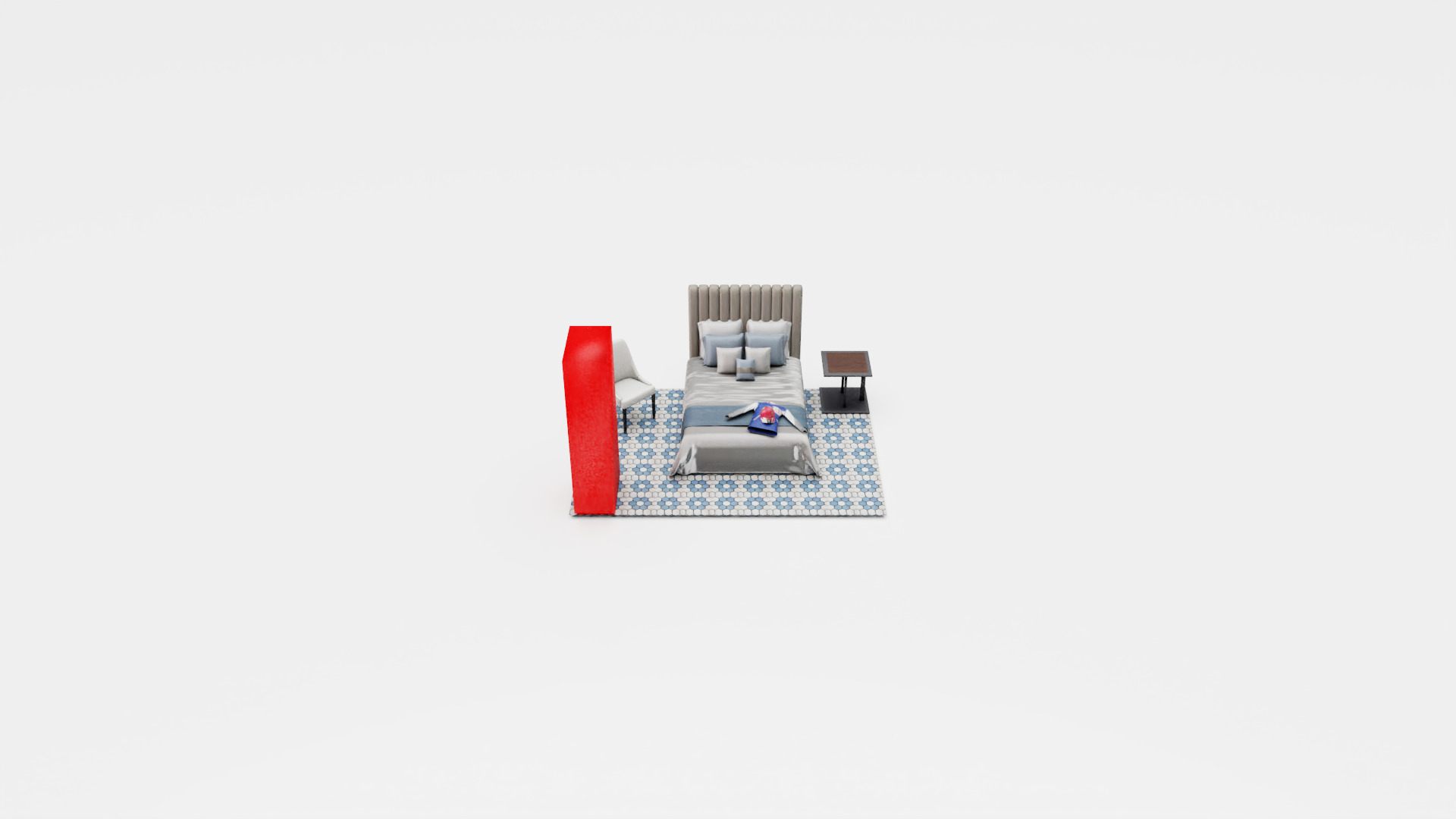}
    \end{subfigure}%
    \vspace{-1.2em}
    \vskip\baselineskip%
    \begin{subfigure}[b]{0.16\linewidth}
    \centering
    \includegraphics[width=\linewidth, trim=500 150 500 150 , clip]{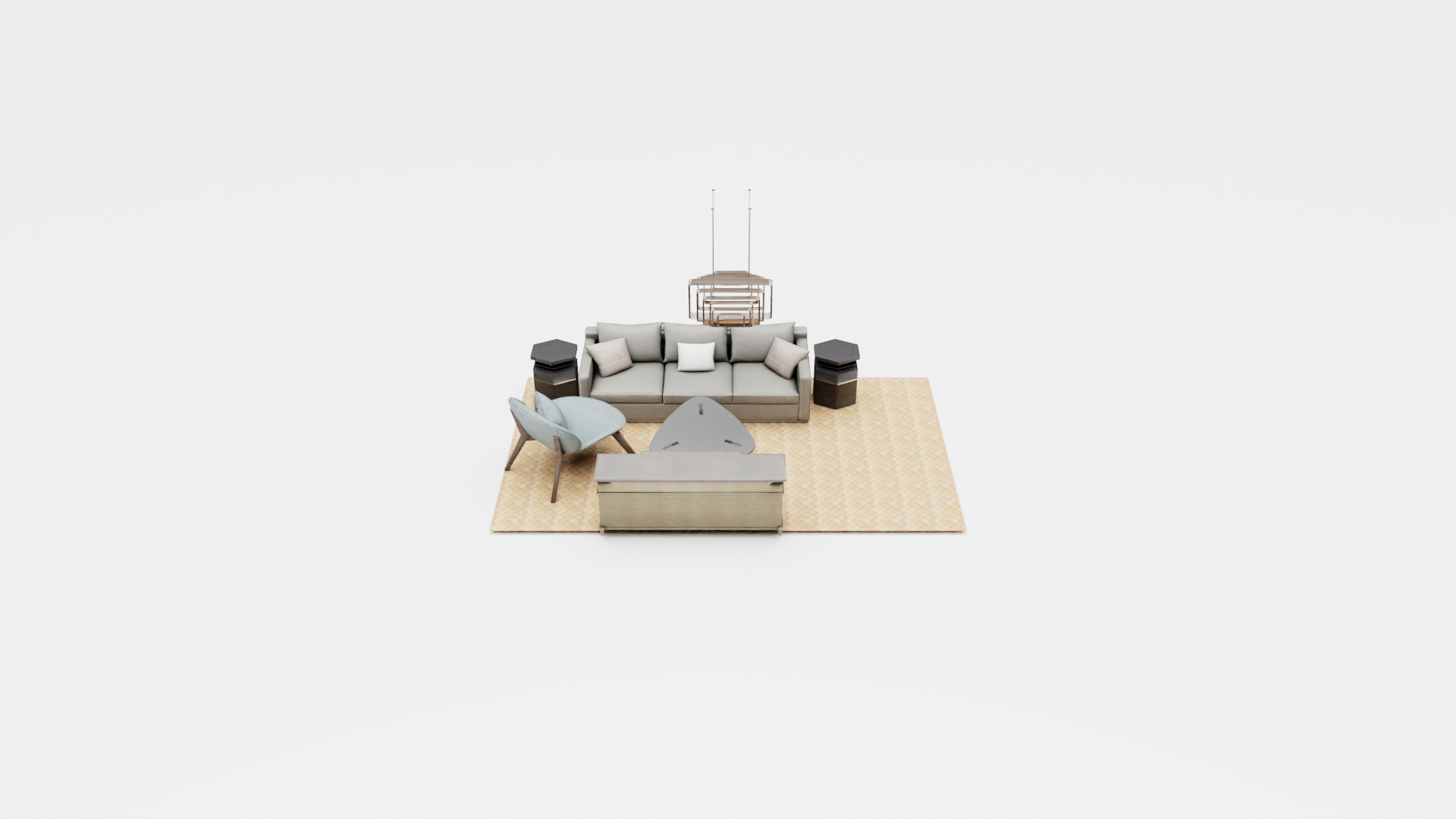}
    \end{subfigure}%
    \begin{subfigure}[b]{0.16\linewidth}
		\centering
        \includegraphics[width=\linewidth, trim=500 150 500 150 , clip]{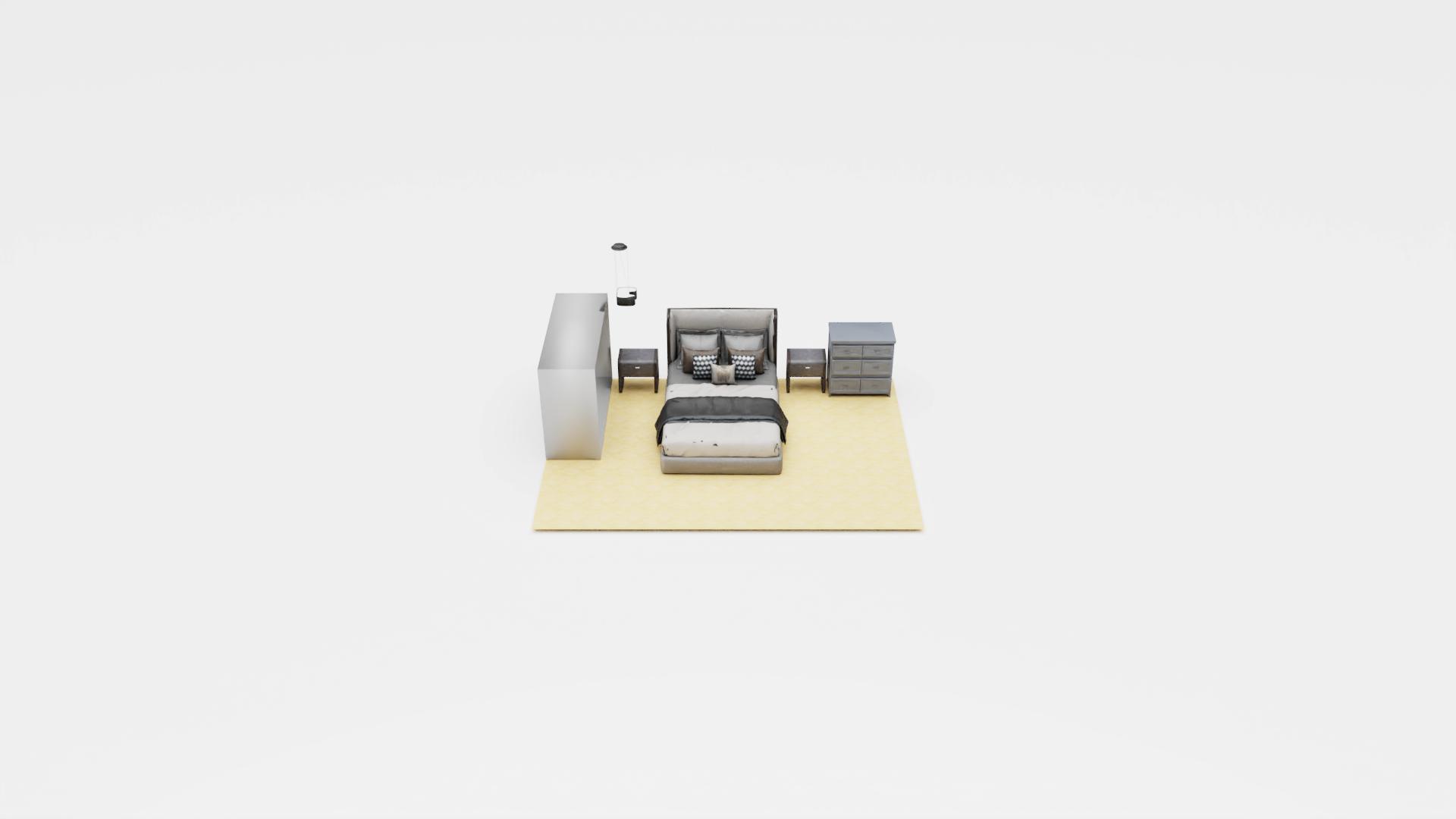}
    \end{subfigure}%
    \begin{subfigure}[b]{0.16\linewidth}
        \centering
        \includegraphics[width=\linewidth, trim=500 150 500 150 , clip]{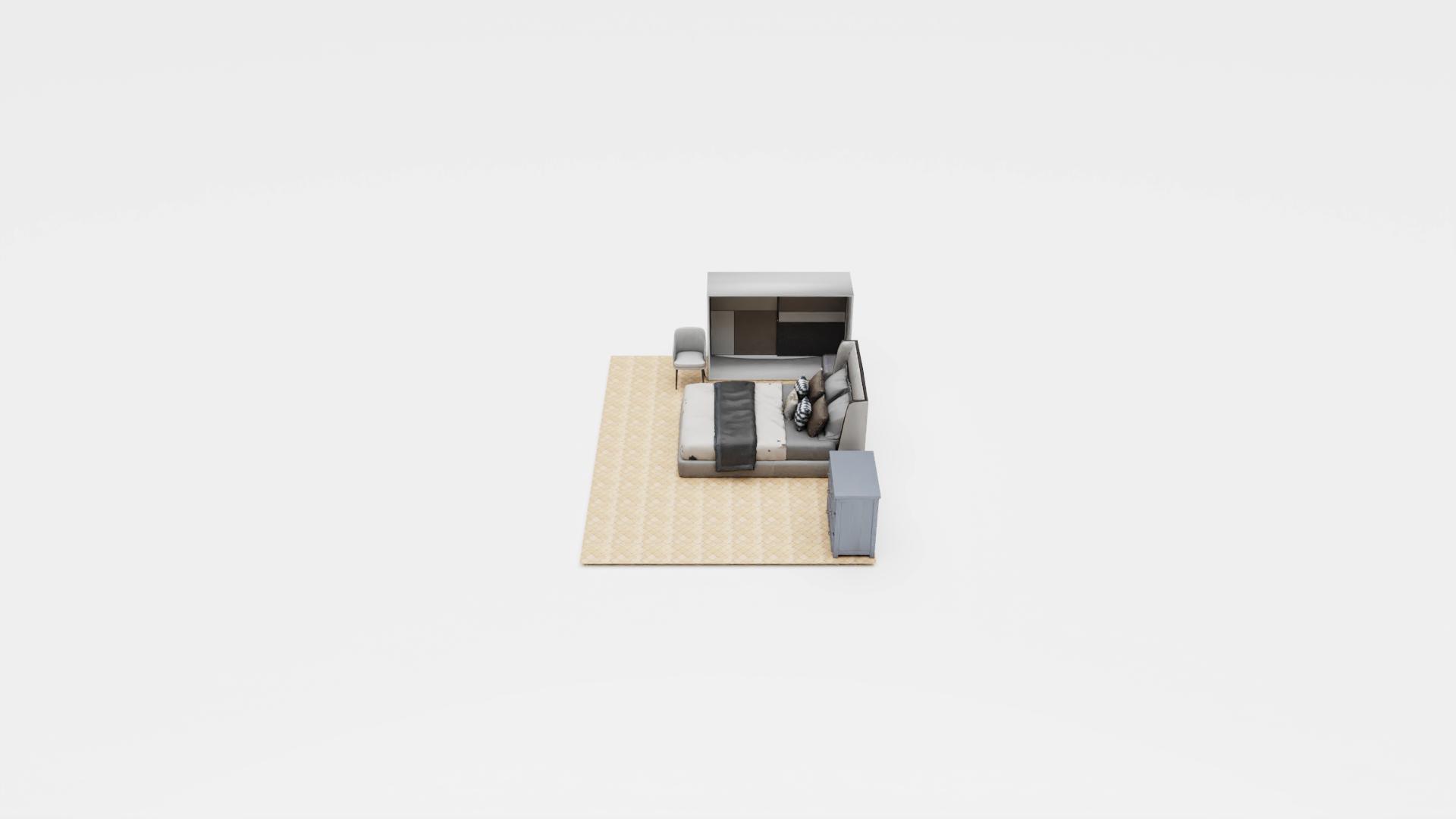}
    \end{subfigure}%
    \begin{subfigure}[b]{0.16\linewidth}
		\centering
        \includegraphics[width=\linewidth, trim=500 150 500 150 , clip]{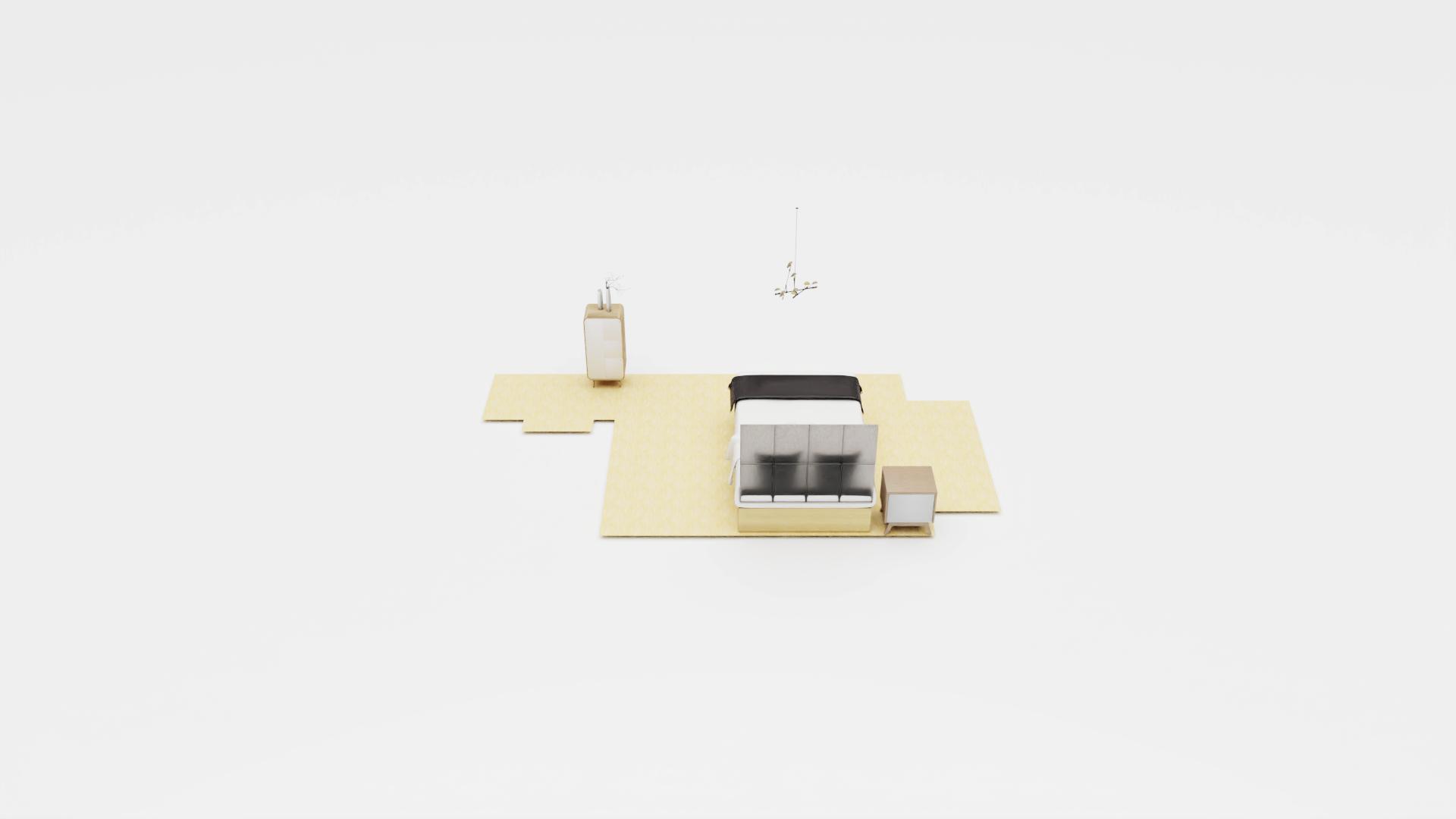}
    \end{subfigure}%
    \begin{subfigure}[b]{0.16\linewidth}
    \centering
        \includegraphics[width=\linewidth, trim=500 150 500 150 , clip]{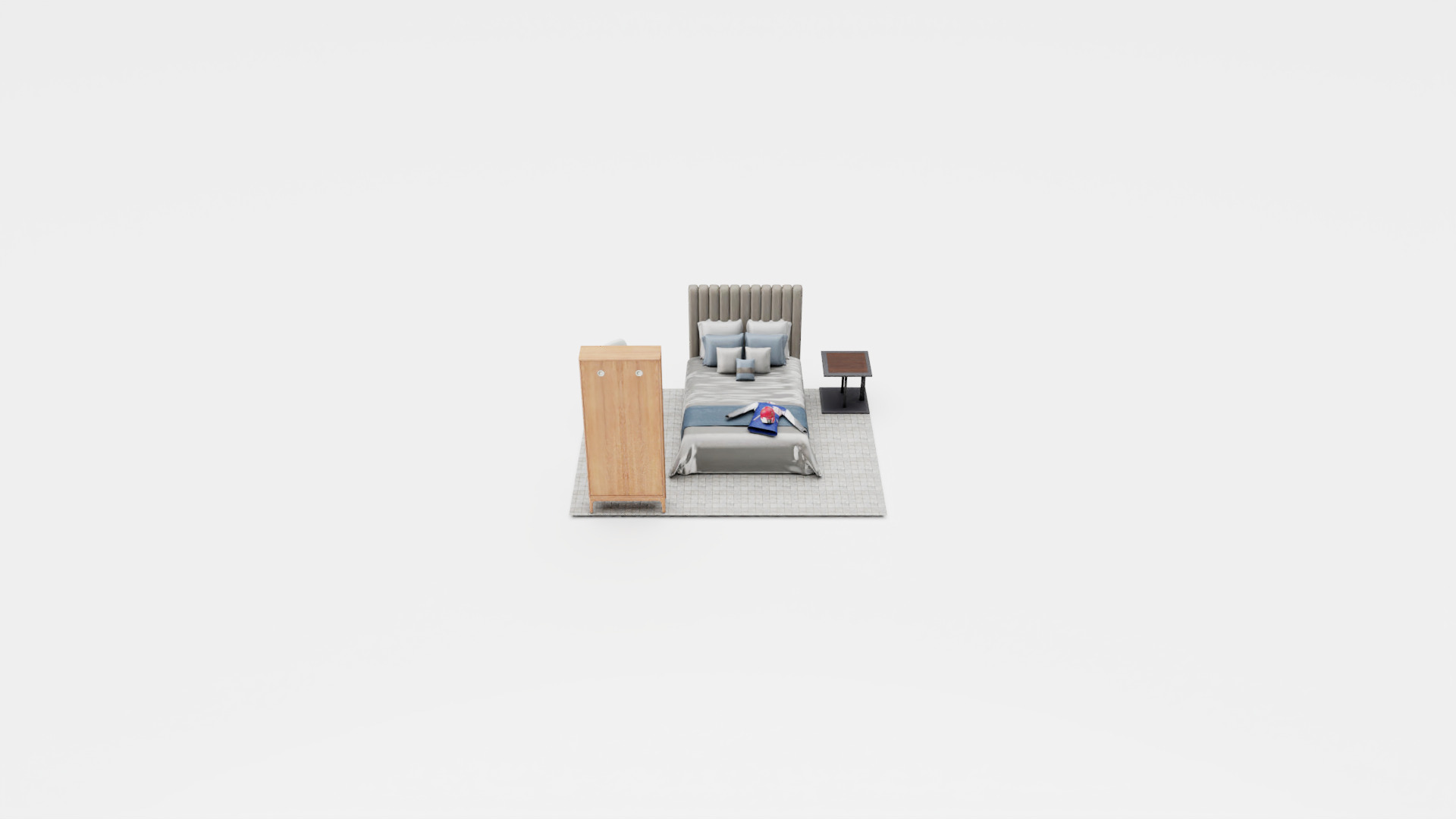}
    \end{subfigure}    \begin{subfigure}[b]{0.16\linewidth}
		\centering
        \includegraphics[width=\linewidth, trim=500 150 500 150 , clip]{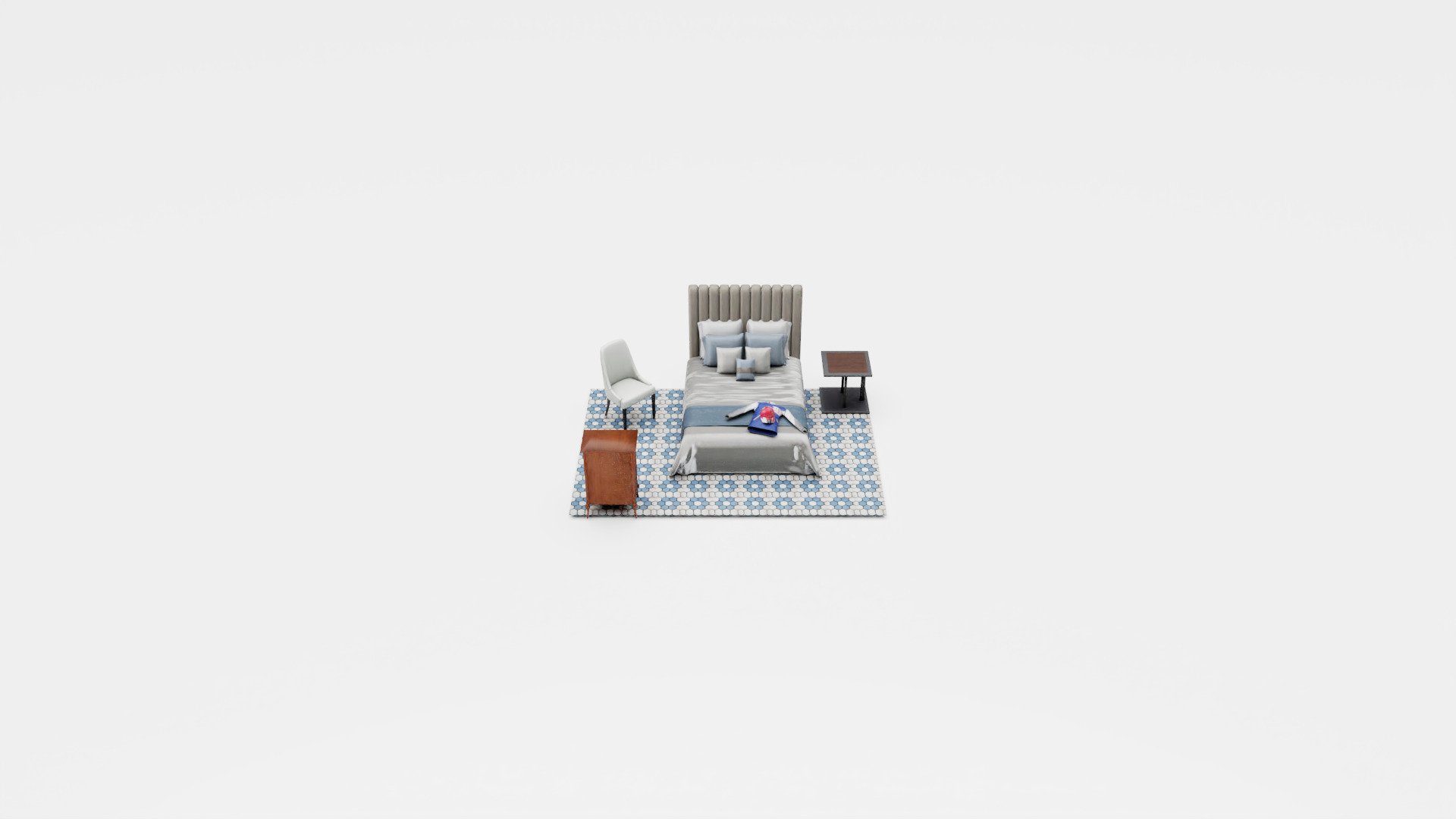}
    \end{subfigure}%
    \vskip\baselineskip%
    \vspace{-1.5em}
    \hfill%
    \begin{subfigure}[b]{0.16\linewidth}
		\centering
        \small TV-stand
    \end{subfigure}%
    \begin{subfigure}[b]{0.16\linewidth}
		\centering
        \small Lamp
    \end{subfigure}%
    \begin{subfigure}[b]{0.16\linewidth}
		\centering
        \small Sofa
    \end{subfigure}%
    \begin{subfigure}[b]{0.16\linewidth}
		\centering
        \small Cabinet
    \end{subfigure}%
    \begin{subfigure}[b]{0.16\linewidth}
		\centering
        \small Bookshelf
    \end{subfigure}%
    \begin{subfigure}[b]{0.16\linewidth}
		\centering
        \small Cabinet
    \end{subfigure}%
    \vspace{-1.2em}
    \vskip\baselineskip%
    \caption{{\bf Object Suggestion}. A user specifies a region of acceptable
    positions to place an object (marked as red boxes, first and third row) and our model suggests
    suitable objects (second and fourth row) to be placed in this location.}
    \label{fig:object_suggestions_supp}
\end{figure}

For this task, we examine the ability of our model to provide object
suggestions given a scene and user specified location constraints. For this
experiment, the user only provides location constraints, namely valid positions
for the centroid of the object to be generated.
\figref{fig:object_suggestions_supp} shows examples of the
location constraints, marked with red boxes, (first and third row) and the
corresponding objects suggested by our model (second and fourth row). We
observe that our model consistently makes plausible suggestions, and for the
cases that a user specifies a region that overlaps with other objects in the
scene, our model suggests adding nothing (first row, third column
\figref{fig:object_suggestions_supp}). In \figref{fig:object_suggestions_supp},
we also provide two examples, where our model makes different suggestions based
on the same location constraints, such as sofa and nightstand for the scenario
illustrated in the first and second column and stool and armchair for the
scenario illustrated in the fifth and sixth column in the first row. 

\subsection{Scene Completion}
\begin{figure}
    \centering
    \begin{subfigure}[b]{0.05\linewidth}
        \hfill
        \rotatebox{90}{\,\,\, \small Partial Scene}
        \,
    \end{subfigure}%
    \begin{subfigure}[b]{0.19\linewidth}
		\centering
		\includegraphics[width=\linewidth, trim=500 200 500 100, clip]{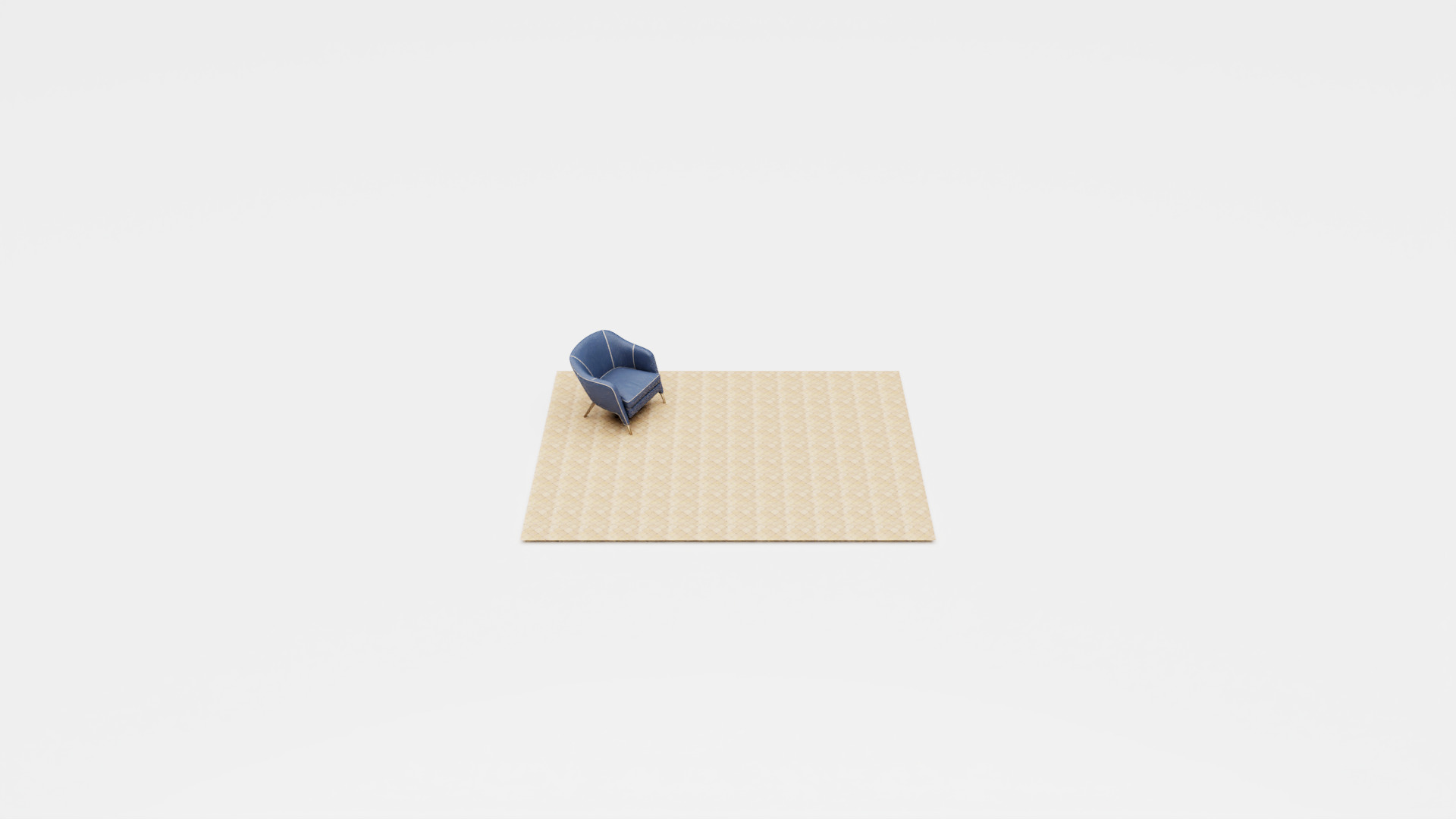}
    \end{subfigure}%
    \begin{subfigure}[b]{0.19\linewidth}
		\centering
		\includegraphics[width=\linewidth, trim=500 200 500 100, clip]{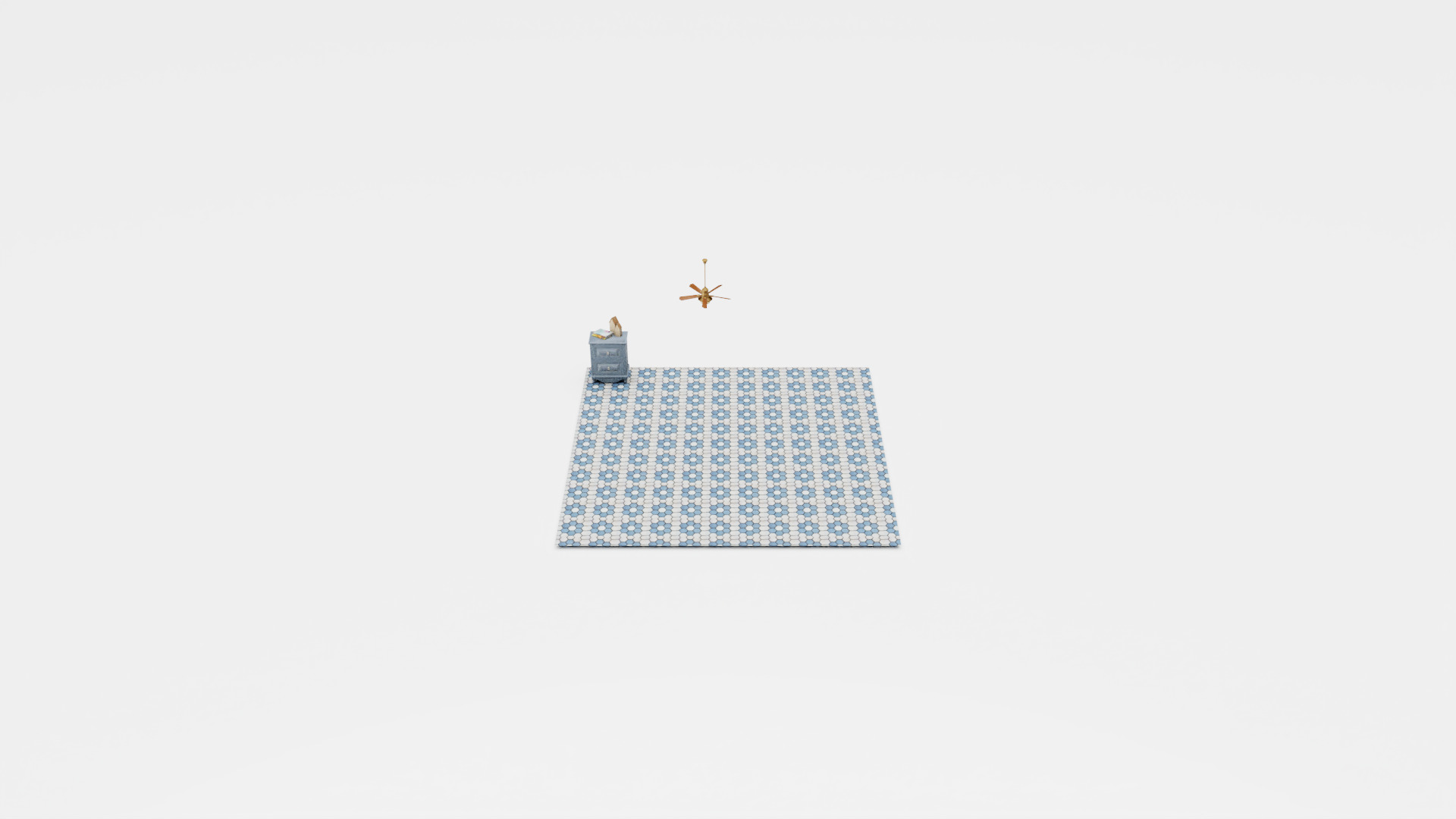}
    \end{subfigure}%
    \begin{subfigure}[b]{0.19\linewidth}
		\centering
		\includegraphics[width=\linewidth, trim=500 200 500 100, clip]{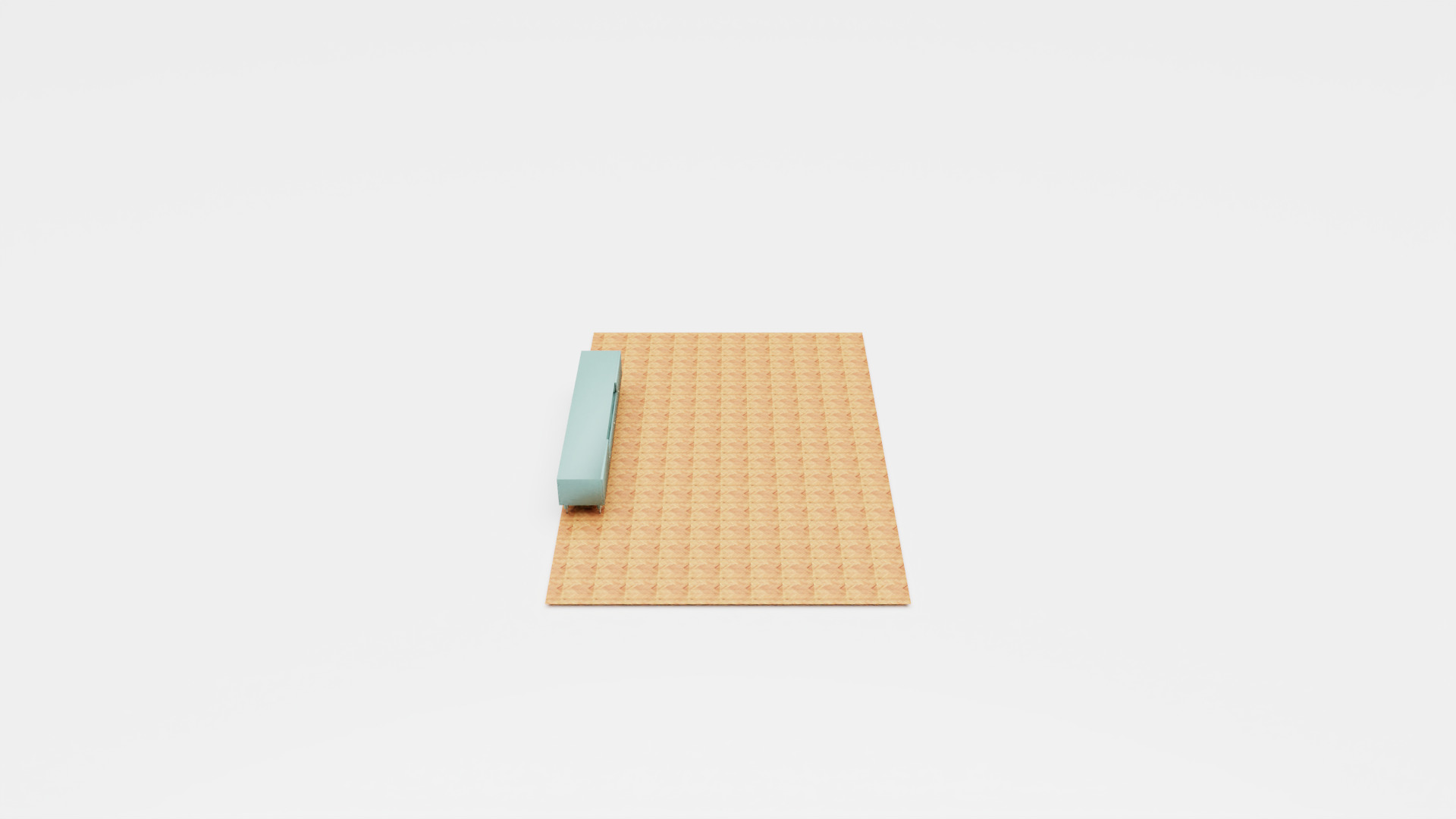}
    \end{subfigure}%
    \begin{subfigure}[b]{0.19\linewidth}
		\centering
		\includegraphics[width=\linewidth, trim=500 200 500 100, clip]{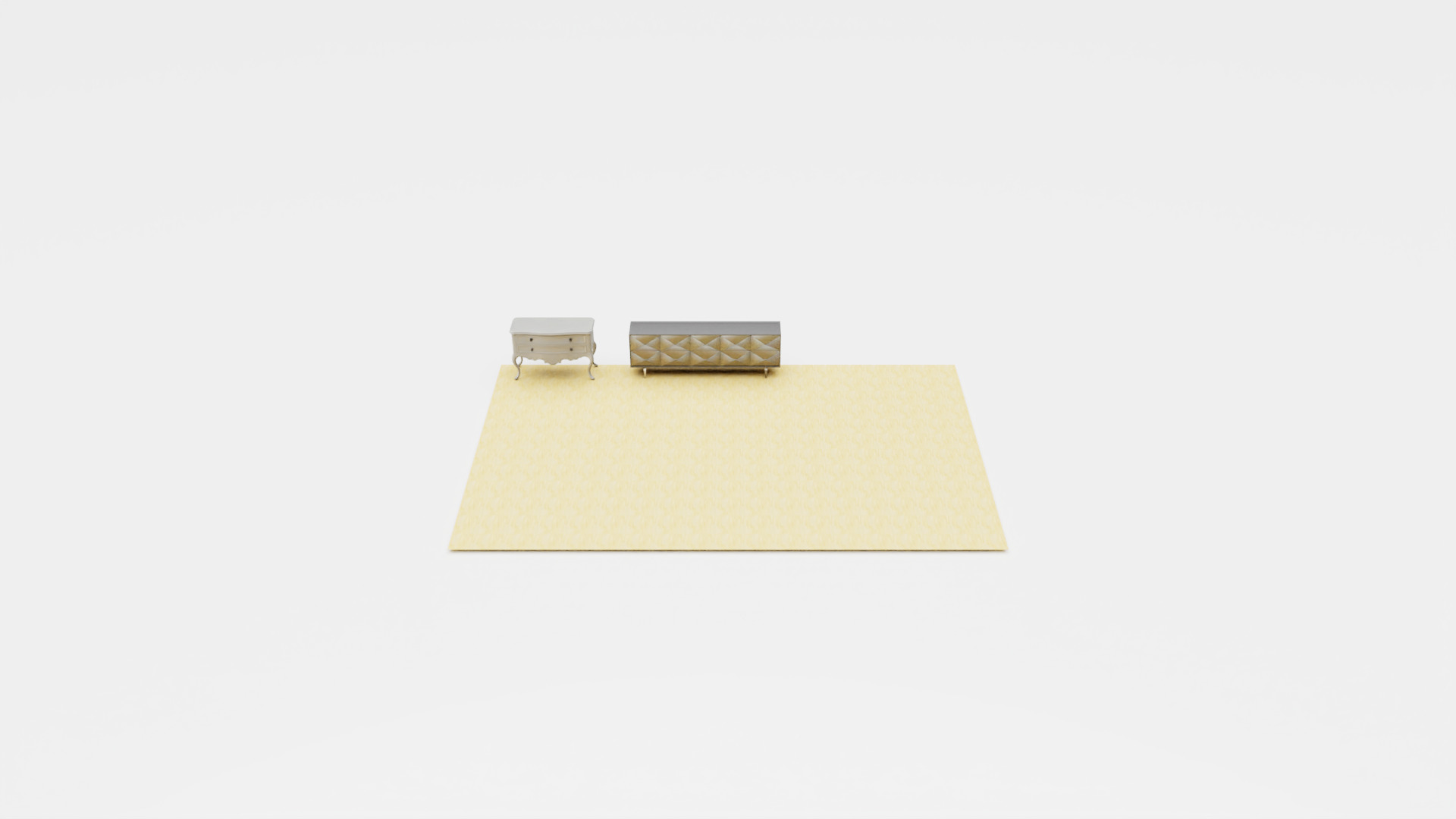}
    \end{subfigure}%
    \begin{subfigure}[b]{0.19\linewidth}
		\centering
		\includegraphics[width=\linewidth, trim=500 200 500 100, clip]{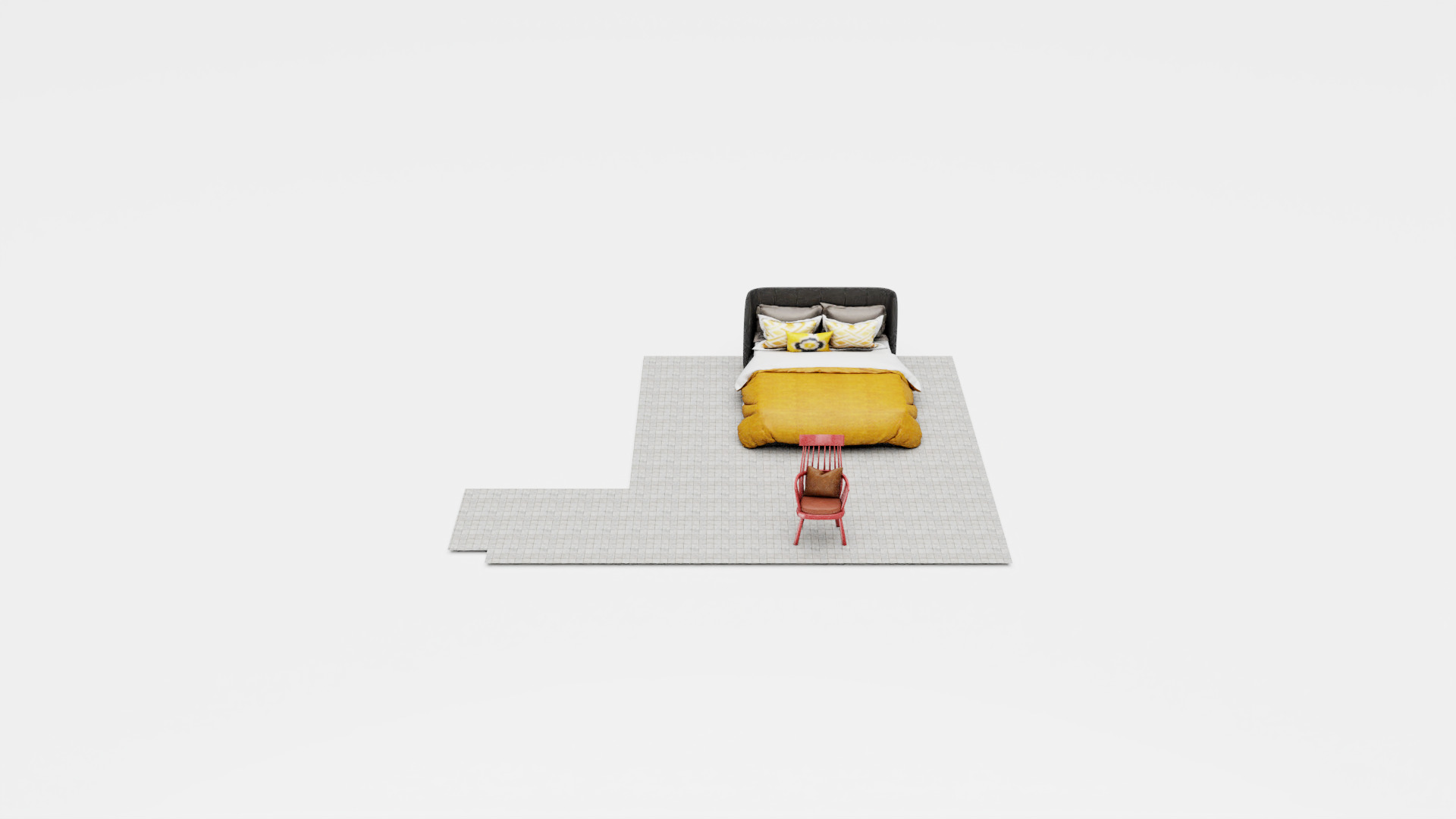}
    \end{subfigure}%
    \hfill%
    \vspace{-2.2em}
    \vskip\baselineskip%
    \begin{subfigure}[b]{0.05\linewidth}
        \hfill
        \rotatebox{90}{\,\,\, \small Completion 1}
        \,
    \end{subfigure}%
    \begin{subfigure}[b]{0.19\linewidth}
		\centering
		\includegraphics[width=\linewidth, trim=500 200 500 100, clip]{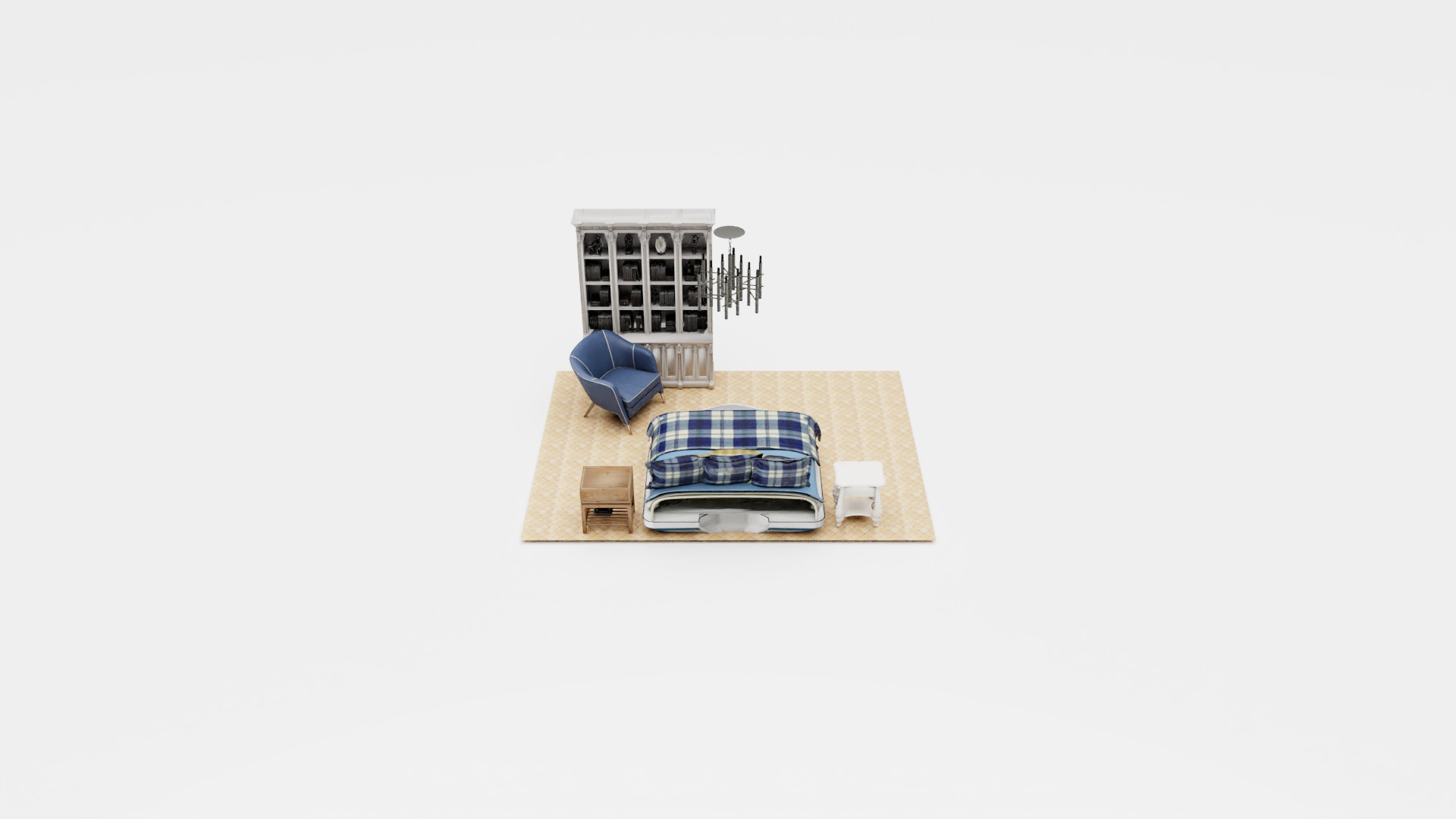}
    \end{subfigure}%
    \begin{subfigure}[b]{0.19\linewidth}
		\centering
		\includegraphics[width=\linewidth, trim=500 200 500 100, clip]{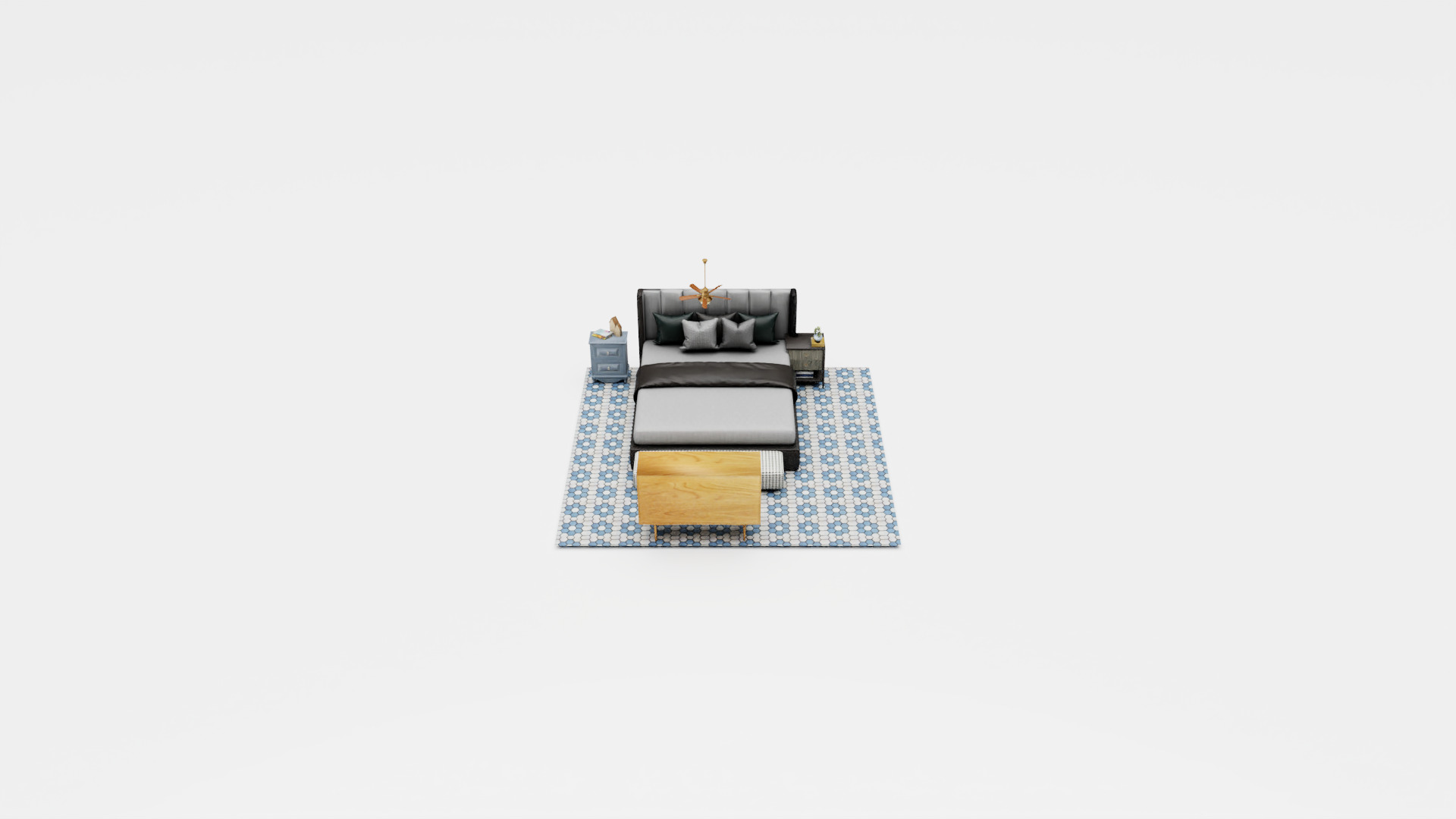}
    \end{subfigure}%
    \begin{subfigure}[b]{0.19\linewidth}
		\centering
		\includegraphics[width=\linewidth, trim=500 200 500 100, clip]{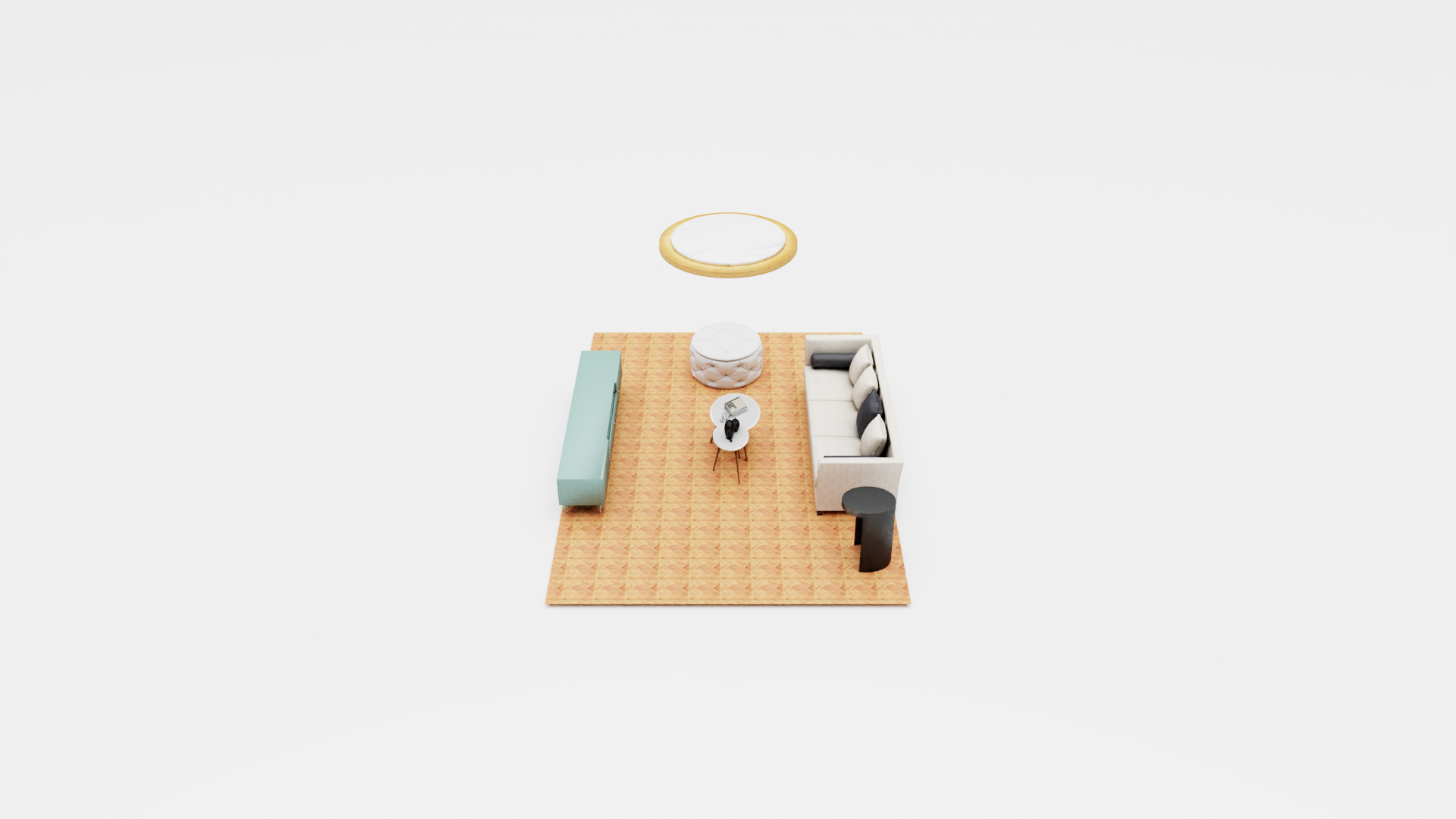}
    \end{subfigure}%
    \begin{subfigure}[b]{0.19\linewidth}
		\centering
		\includegraphics[width=\linewidth, trim=500 200 500 100, clip]{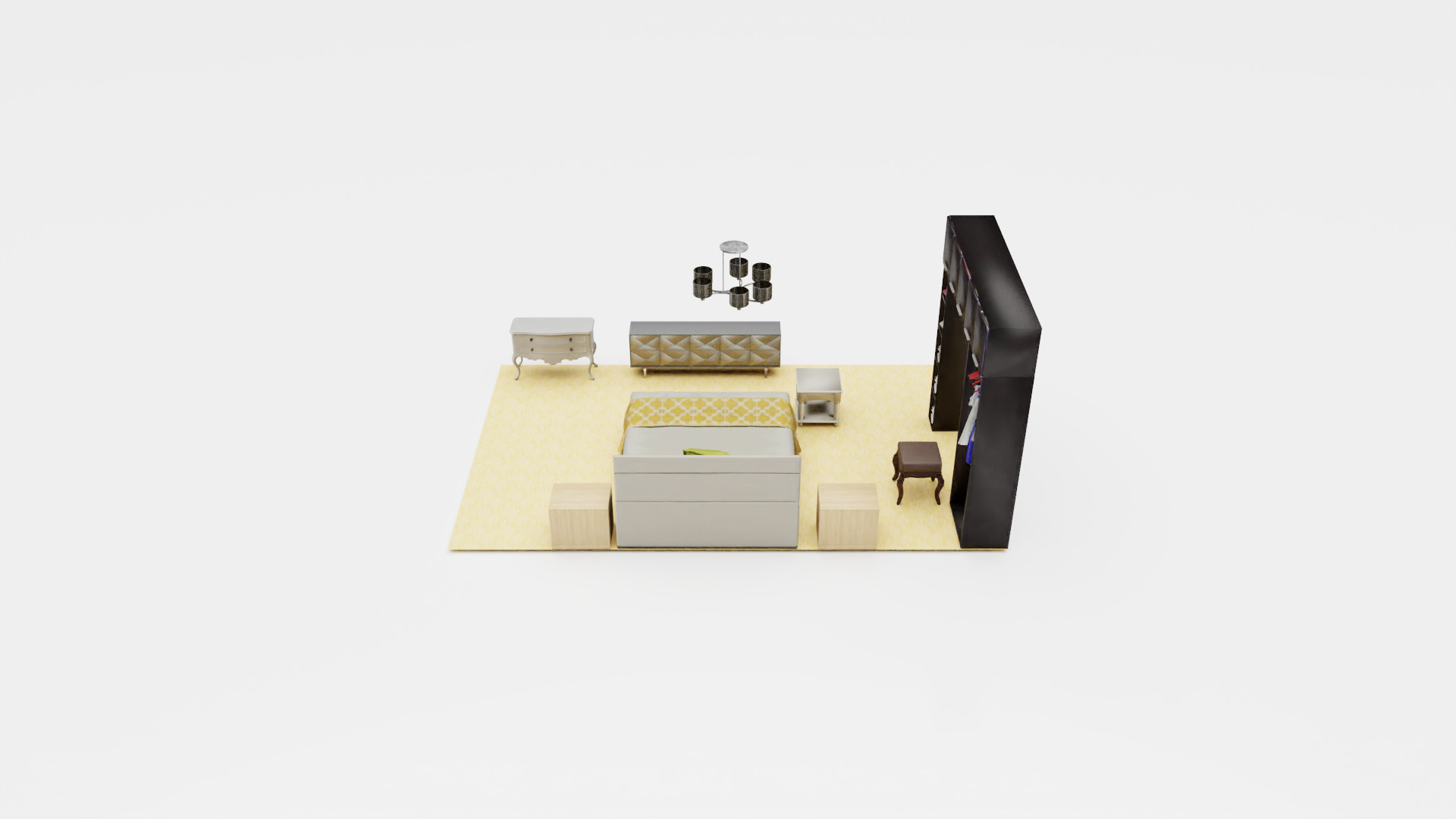}
    \end{subfigure}%
    \begin{subfigure}[b]{0.19\linewidth}
		\centering
		\includegraphics[width=\linewidth, trim=500 200 500 100, clip]{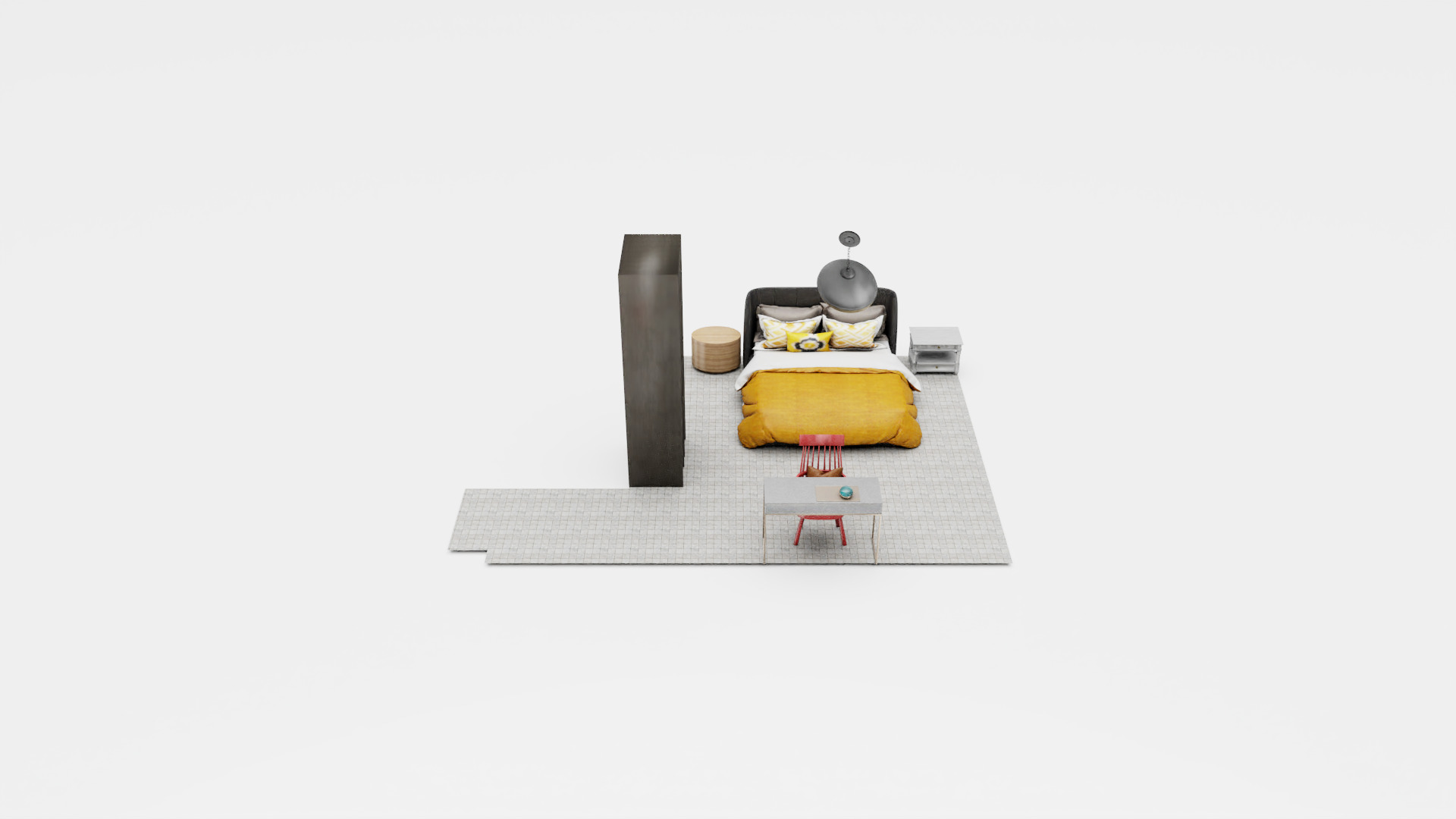}
    \end{subfigure}%
    \hfill%
    \vspace{-2.2em}
    \vskip\baselineskip%
    \begin{subfigure}[b]{0.05\linewidth}
        \hfill
        \rotatebox{90}{\,\,\, \small Completion 2}
        \,
    \end{subfigure}%
    \begin{subfigure}[b]{0.19\linewidth}
		\centering
		\includegraphics[width=\linewidth, trim=500 200 500 100, clip]{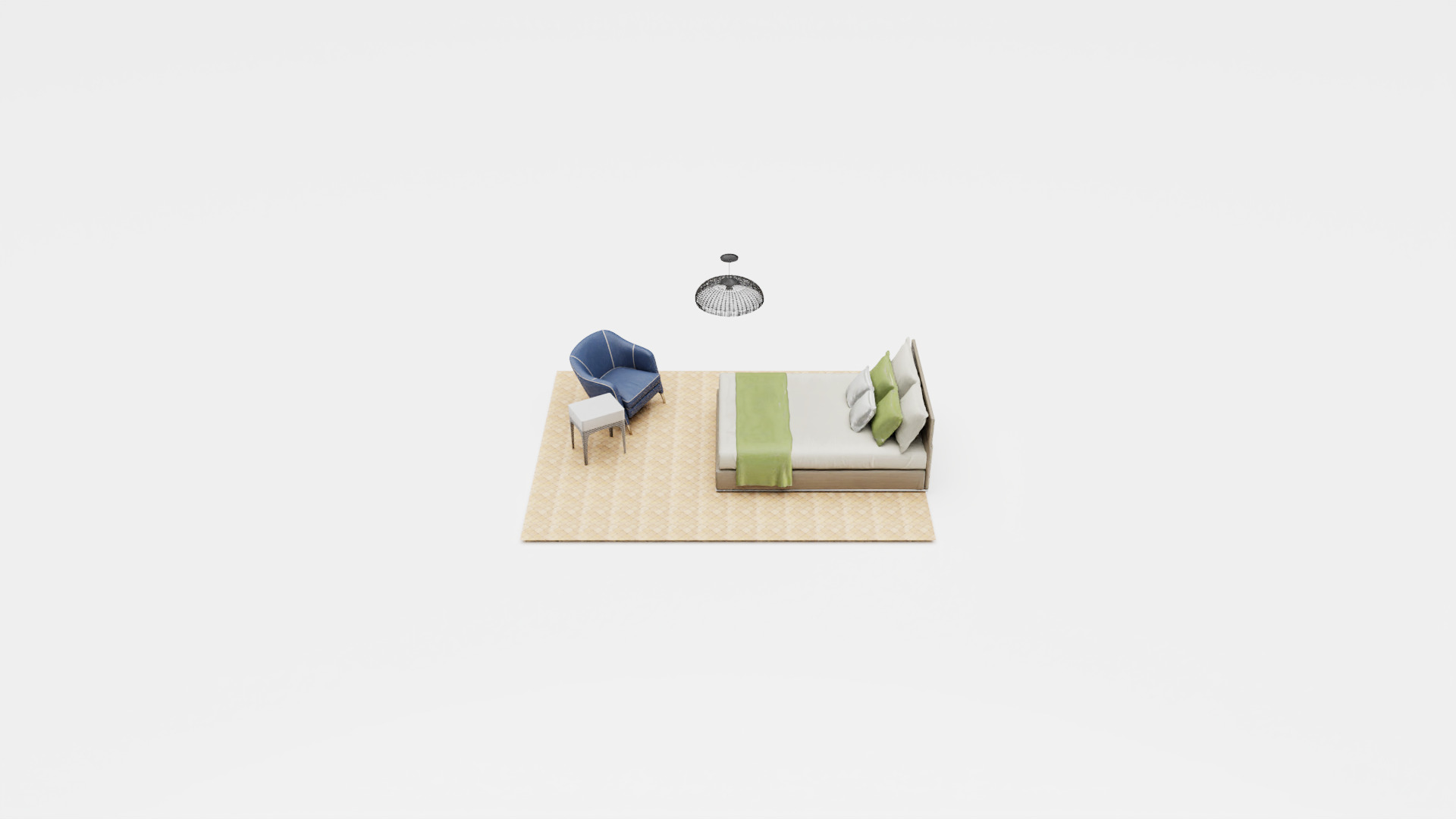}
    \end{subfigure}%
    \begin{subfigure}[b]{0.19\linewidth}
		\centering
		\includegraphics[width=\linewidth, trim=500 200 500 100, clip]{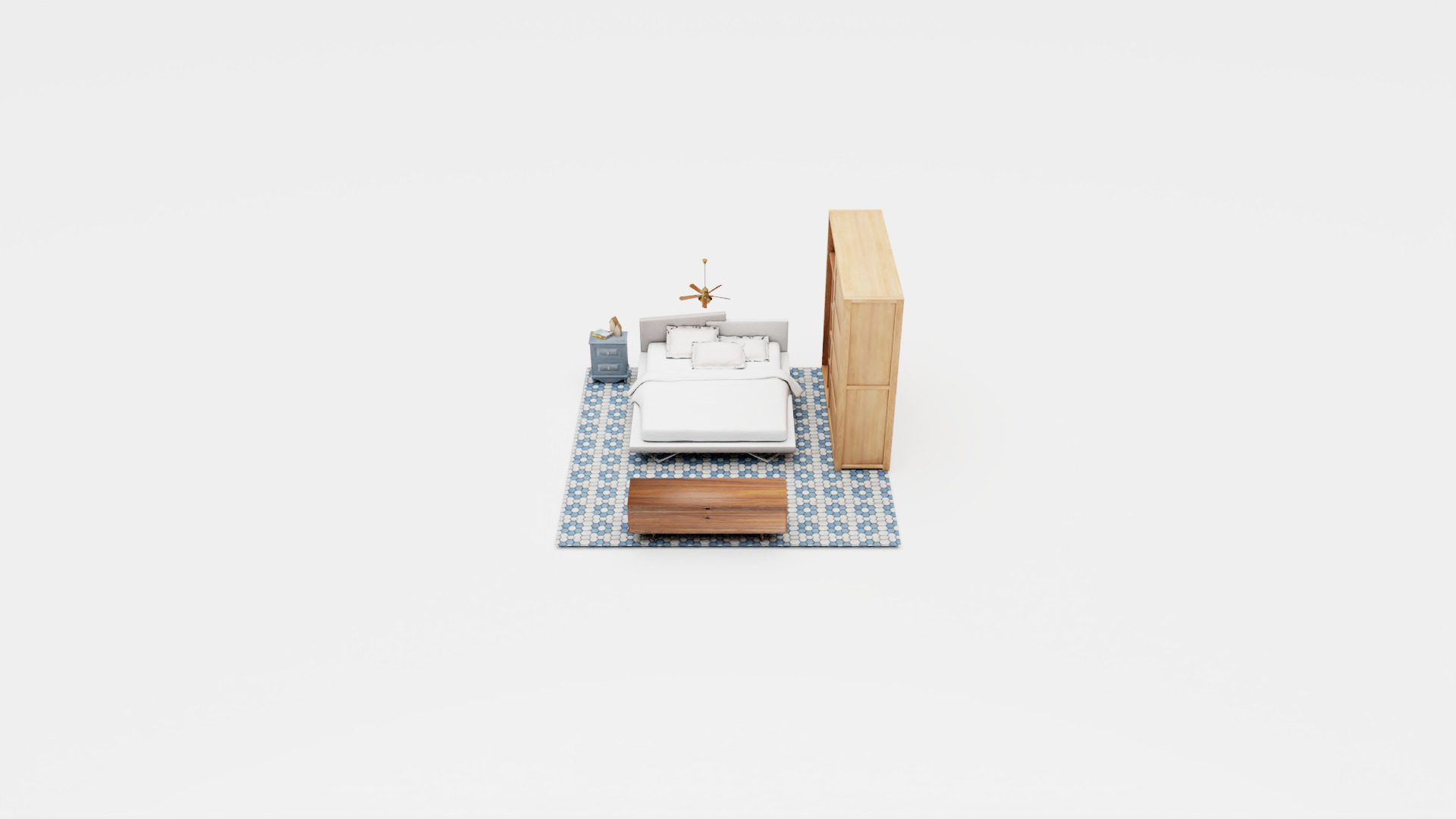}
    \end{subfigure}%
    \begin{subfigure}[b]{0.19\linewidth}
		\centering
		\includegraphics[width=\linewidth, trim=500 200 500 100, clip]{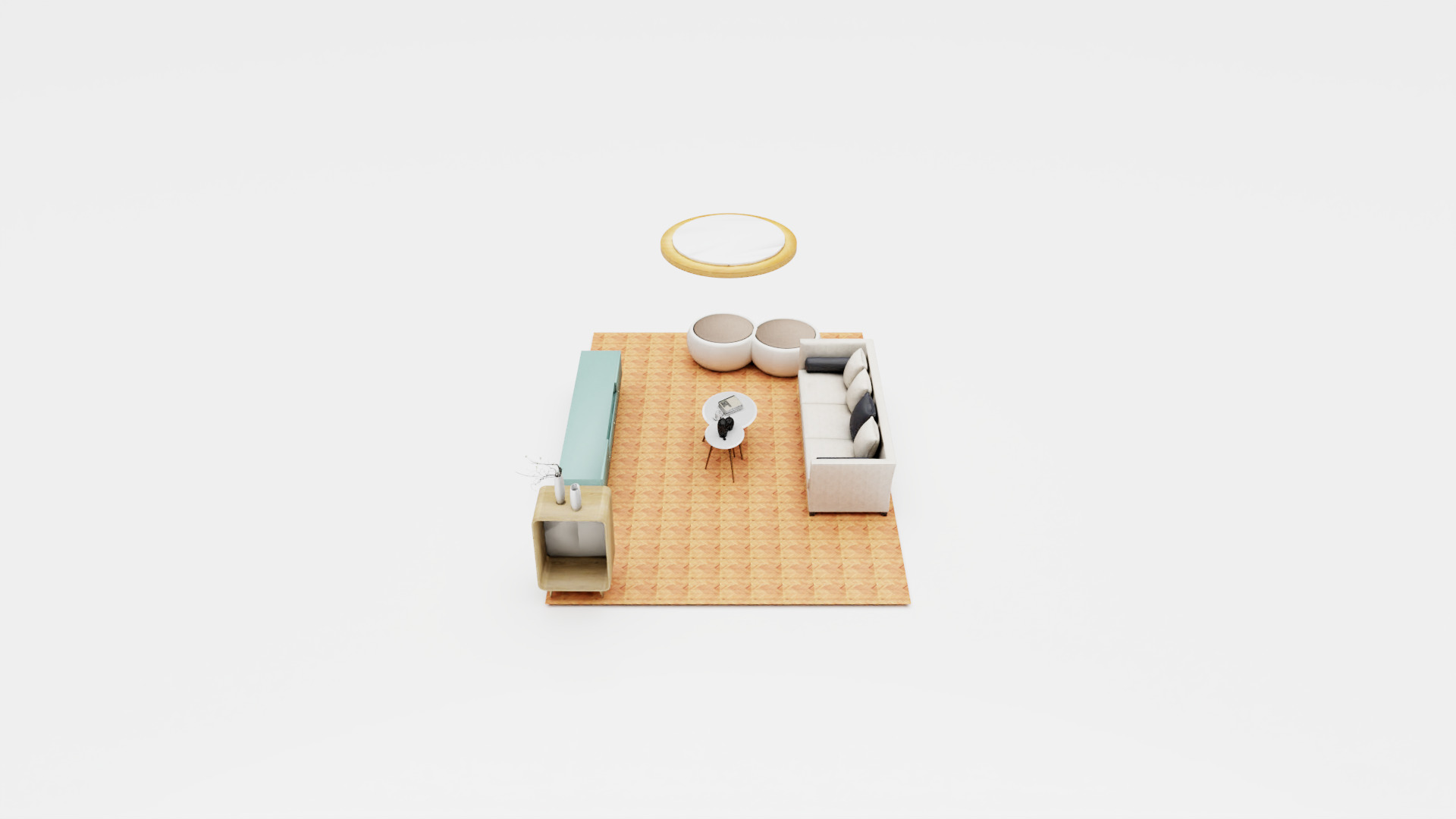}
    \end{subfigure}%
    \begin{subfigure}[b]{0.19\linewidth}
		\centering
		\includegraphics[width=\linewidth, trim=500 200 500 100, clip]{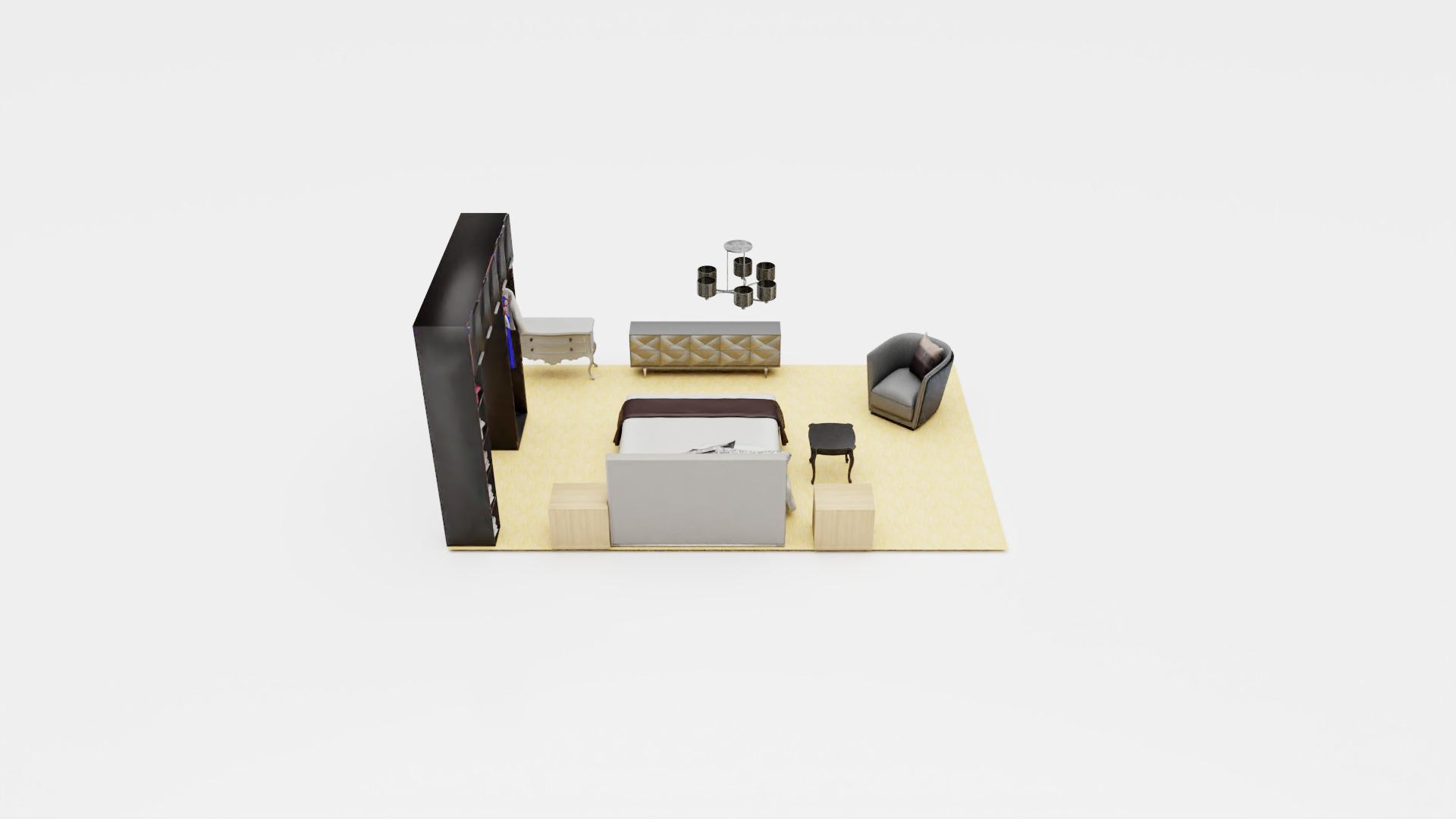}
    \end{subfigure}%
    \begin{subfigure}[b]{0.19\linewidth}
		\centering
		\includegraphics[width=\linewidth, trim=500 200 500 100, clip]{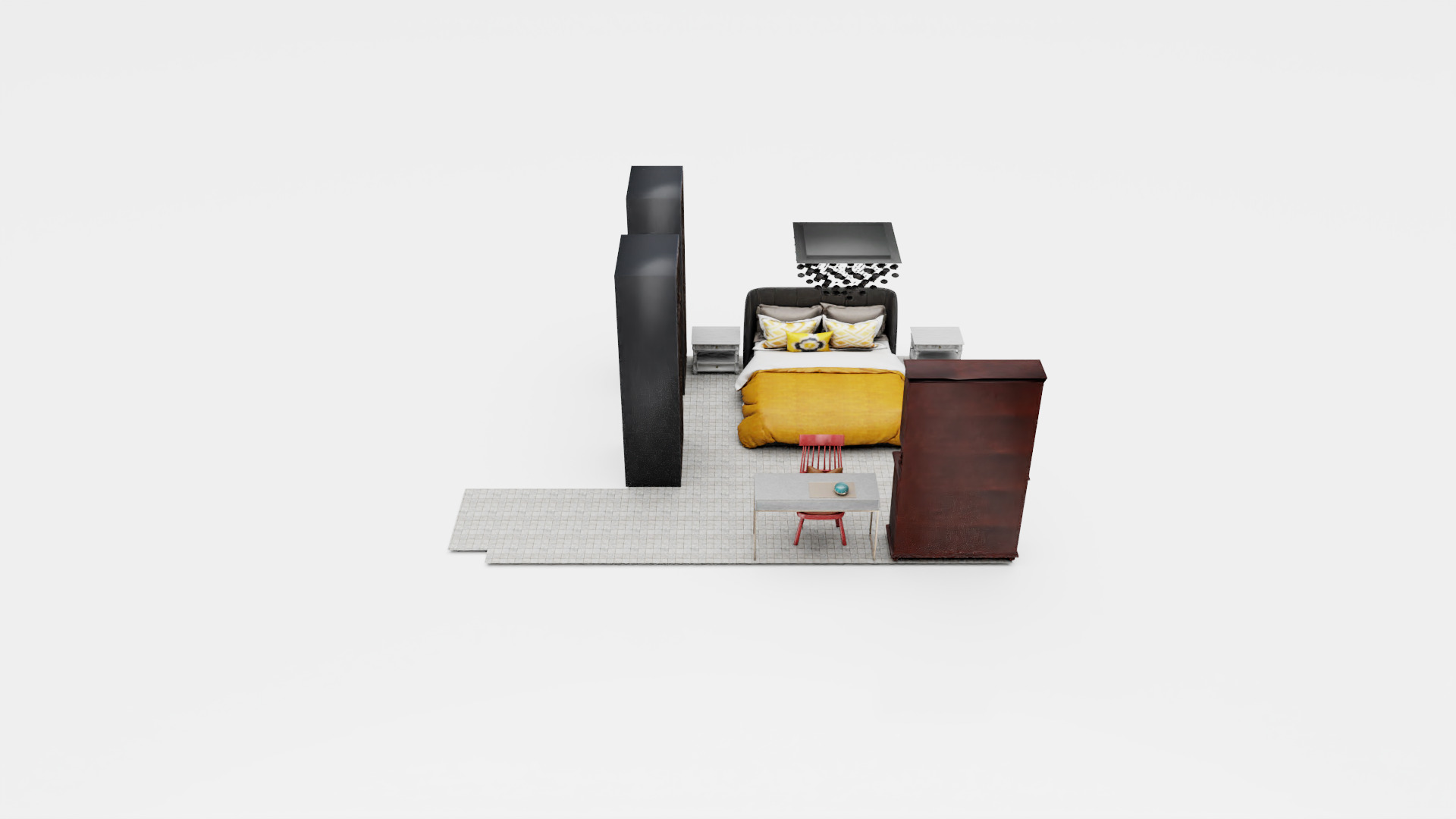}
    \end{subfigure}%
    \vspace{-1.2em}
    \vskip\baselineskip%
    \caption{{\bf Scene Completion}. Starting from a partially complete
    scene (first row), we visualize two examples of scene completions using our
    model (second and third row).}
    \label{fig:scene_completion_supp}
\end{figure}

Starting from a partial scene, we want to evaluate the ability of our model to
generate plausible object arrangements. To generate the partial scenes, we
randomly sample scenes from the test set and remove the majority of the objects
in them. \figref{fig:scene_completion_supp} shows examples for various partial
rooms (first row \figref{fig:scene_completion_supp}), as well as two
alternative scene completions using our model (second and third row
\figref{fig:scene_completion_supp}). We observe that our model generates
diverse arrangements of objects that are consistently meaningful. For example,
for the case where the partial scene consists of a chair and a bed (last column
\figref{fig:scene_completion_supp}), our model generates completions that have
nightstands surrounding the bed as well as a desk in front of the chair.

\begin{figure}
    \hfill%
    \centering
    \begin{subfigure}[b]{0.16\linewidth}
		\centering
		\includegraphics[width=\linewidth, trim=500 200 500 200, clip]{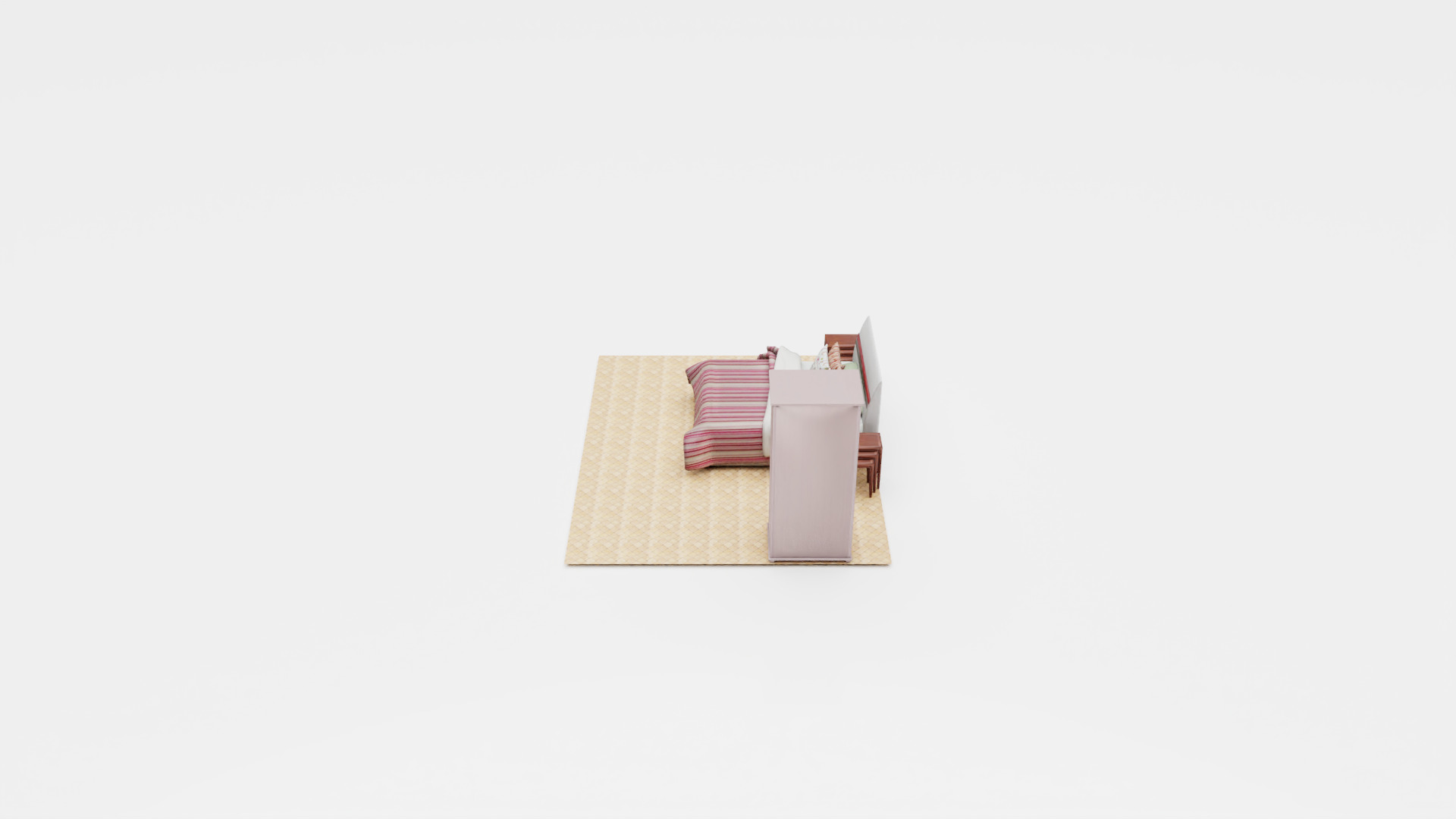}
    \end{subfigure}%
    \begin{subfigure}[b]{0.16\linewidth}
		\centering
		\includegraphics[width=\linewidth, trim=500 200 500 200, clip]{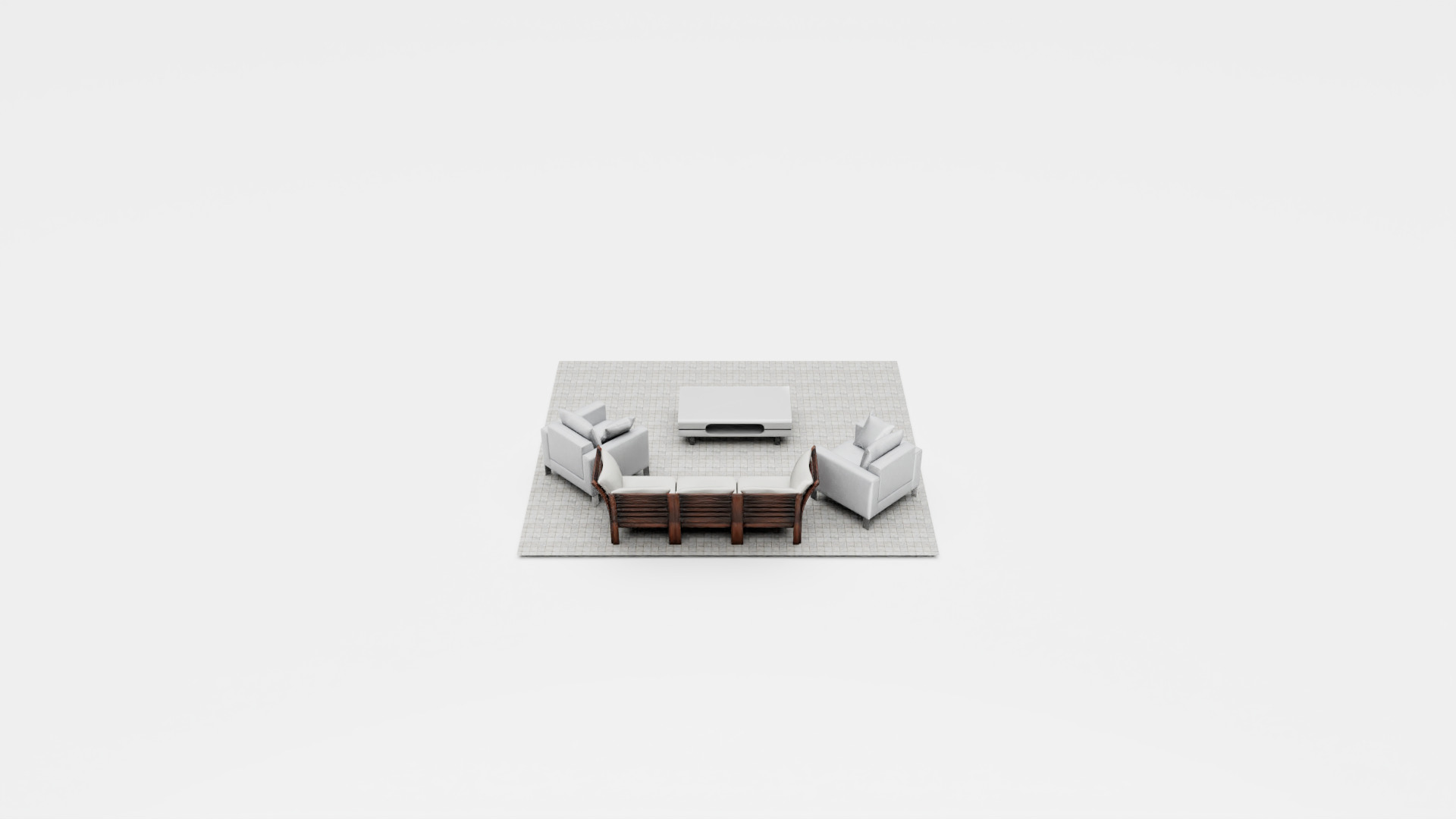}
    \end{subfigure}%
    \begin{subfigure}[b]{0.16\linewidth}
		\centering
		\includegraphics[width=\linewidth, trim=500 200 500 200, clip]{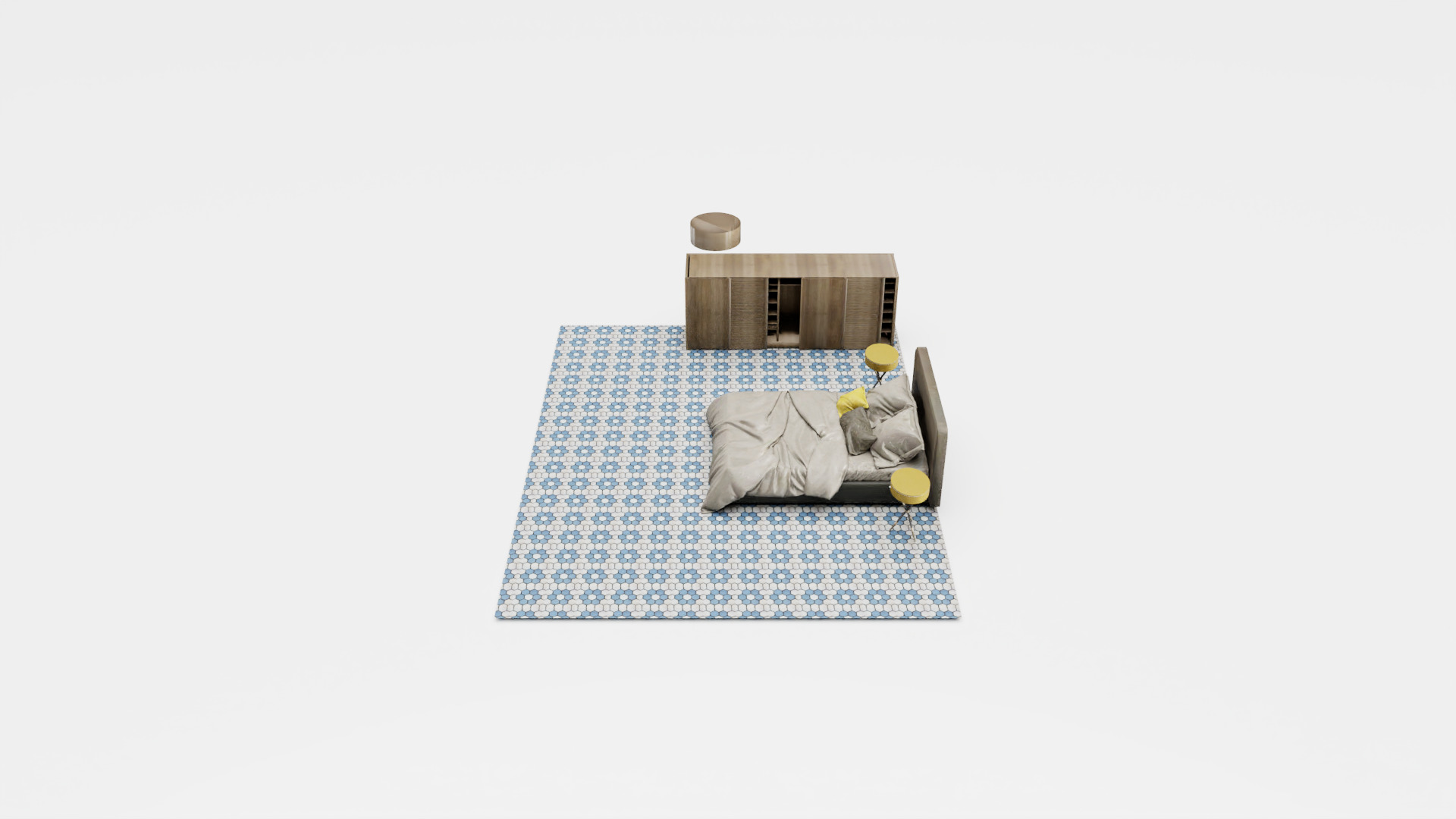}
    \end{subfigure}%
    \begin{subfigure}[b]{0.16\linewidth}
		\centering
		\includegraphics[width=\linewidth, trim=500 200 500 200, clip]{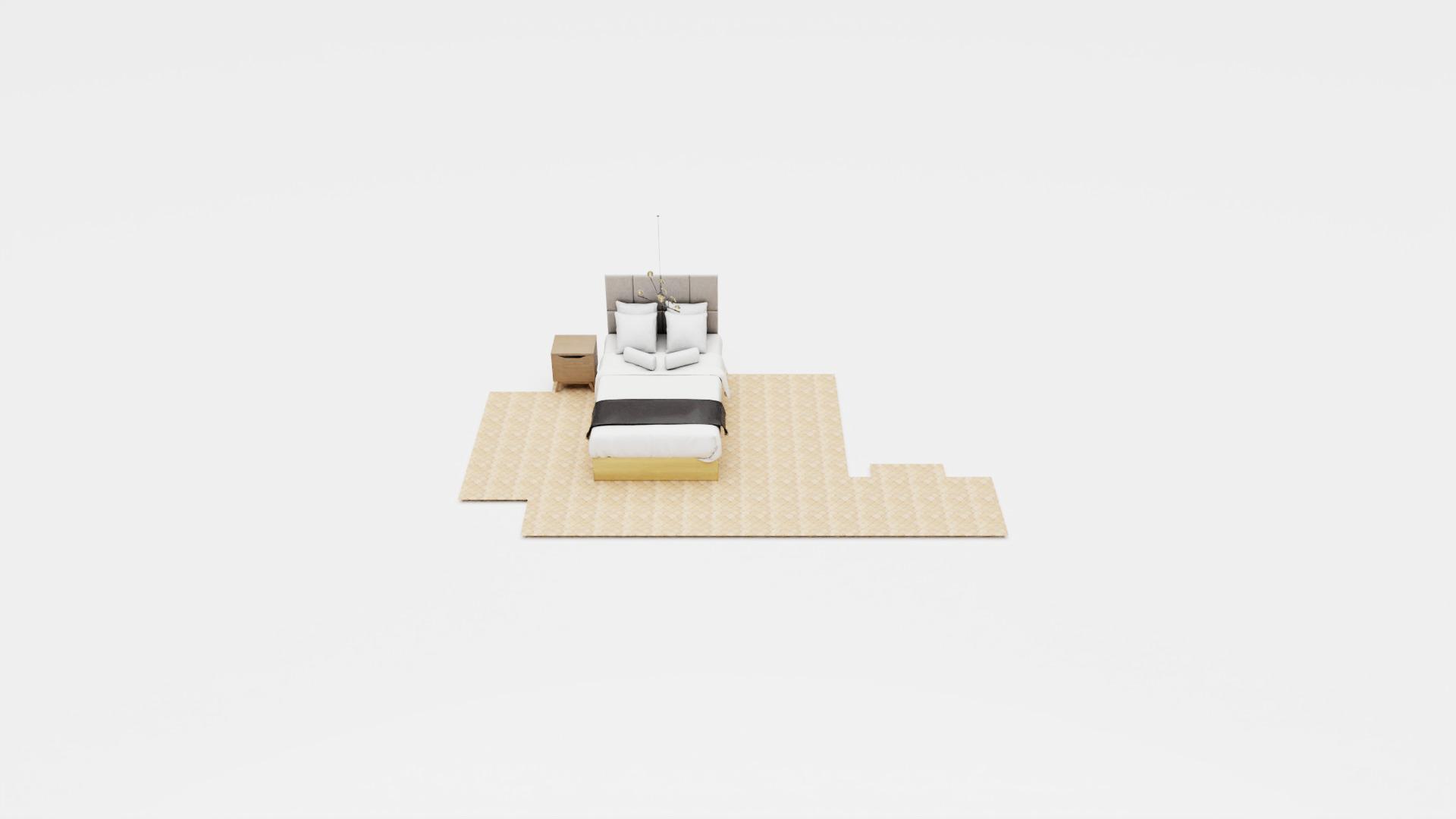}
    \end{subfigure}%
    \begin{subfigure}[b]{0.16\linewidth}
		\centering
		\includegraphics[width=\linewidth, trim=500 200 500 200, clip]{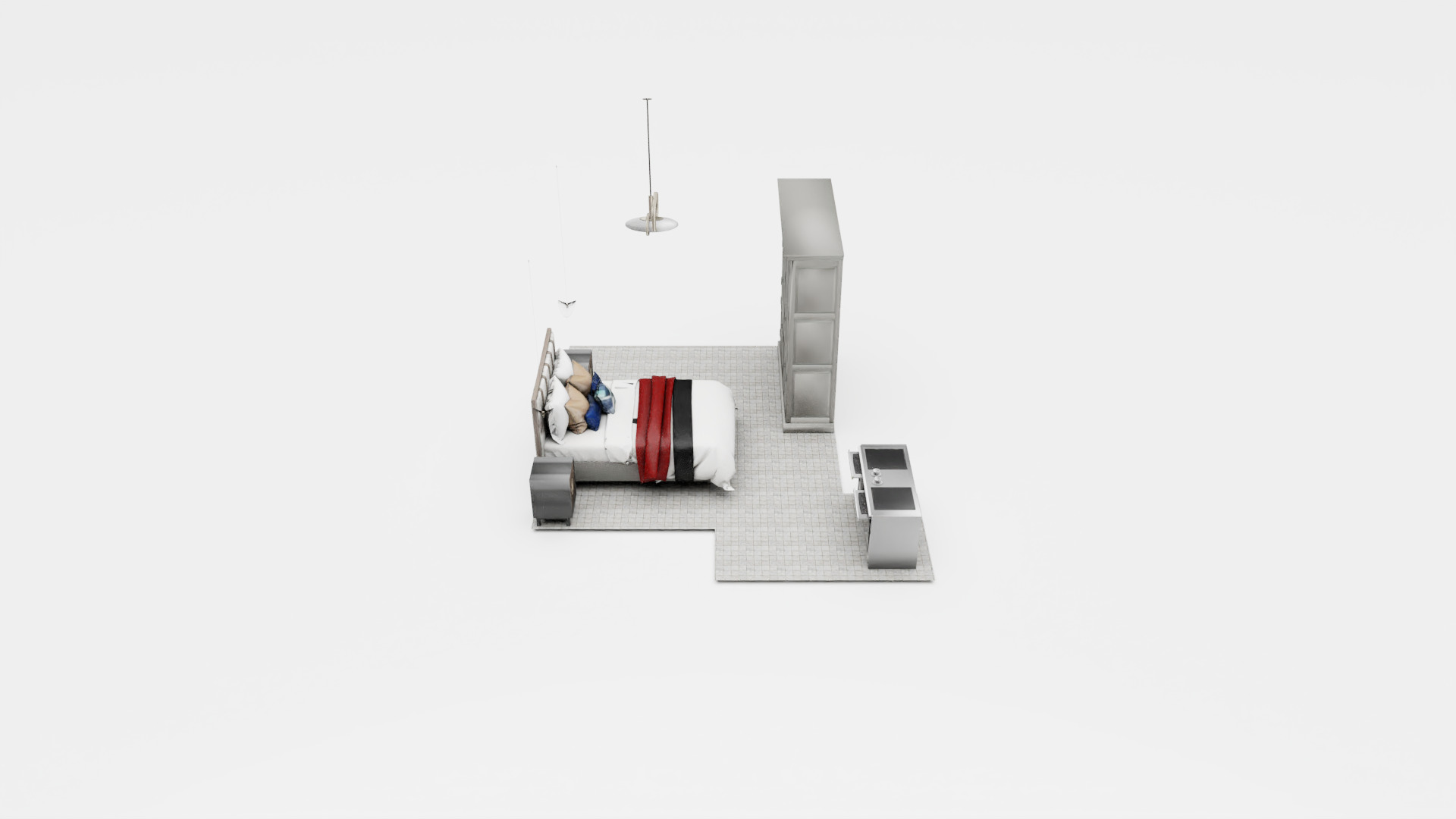}
    \end{subfigure}%
    \begin{subfigure}[b]{0.16\linewidth}
		\centering
		\includegraphics[width=\linewidth, trim=500 200 500 200, clip]{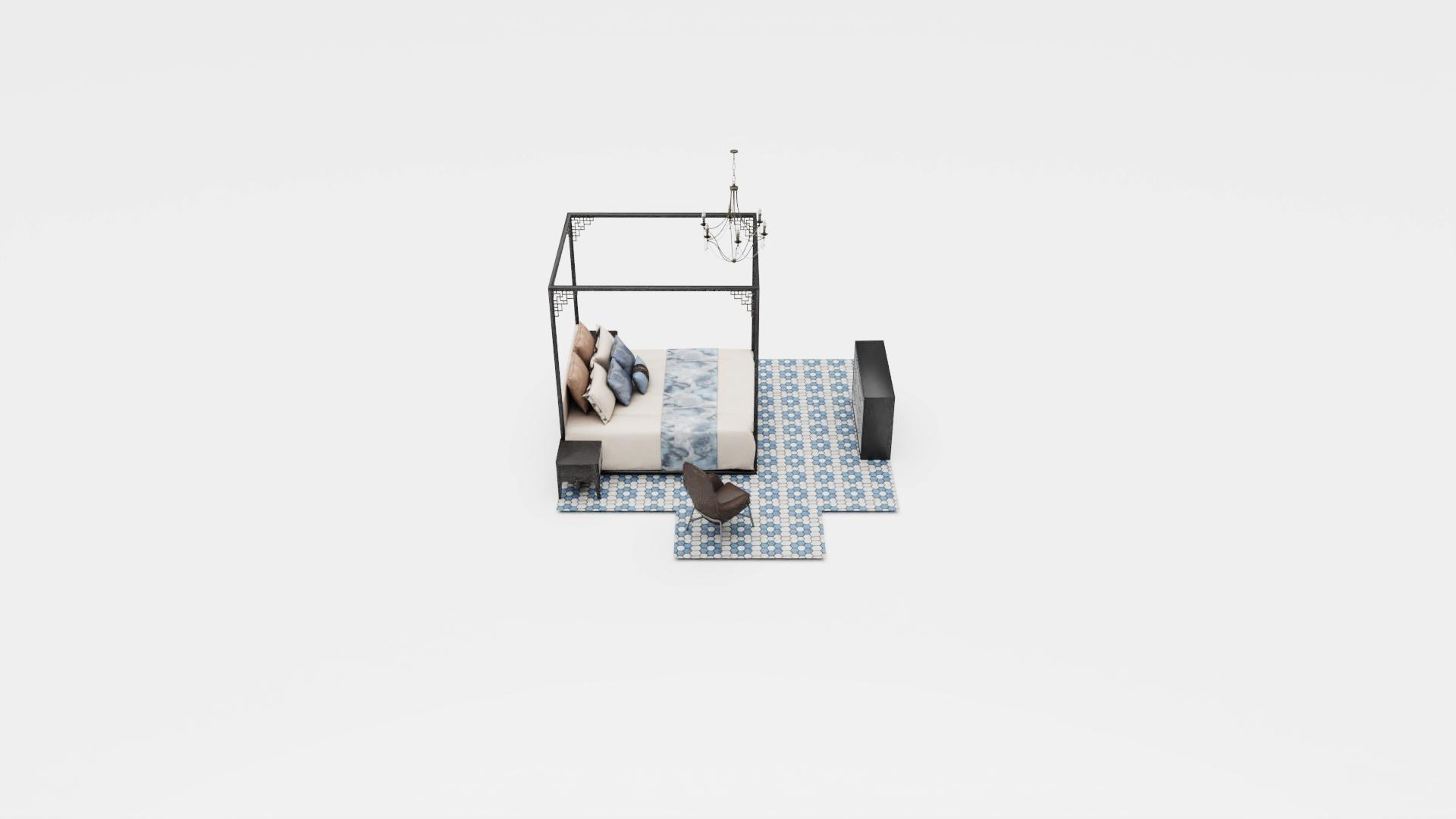}
    \end{subfigure}%
    \hfill%
    \vspace{-1.2em}
    \vskip\baselineskip%
    \begin{subfigure}[b]{0.16\linewidth}
		\centering
		\includegraphics[width=\linewidth, trim=500 200 500 200, clip]{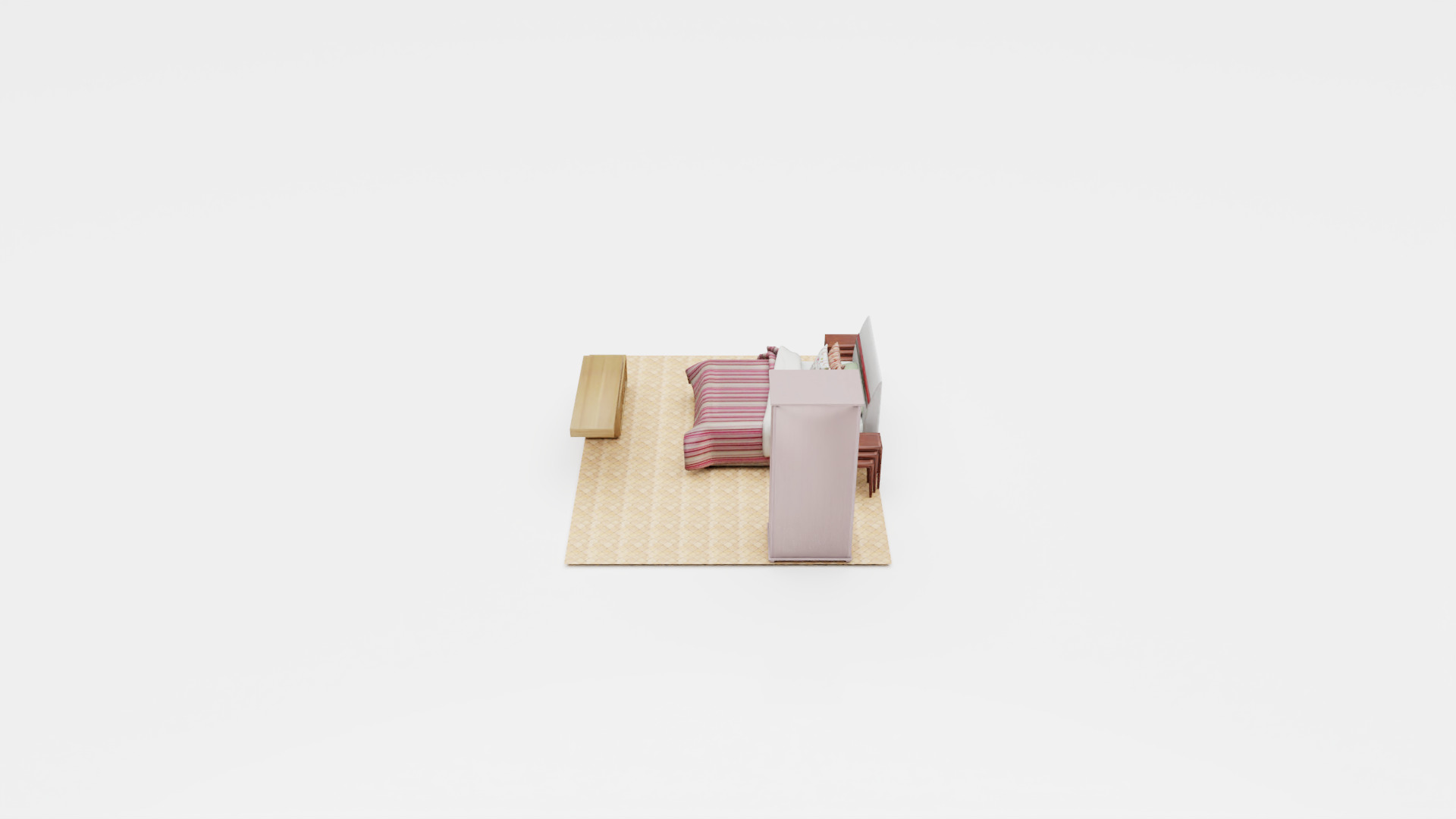}
    \end{subfigure}%
    \begin{subfigure}[b]{0.16\linewidth}
		\centering
		\includegraphics[width=\linewidth, trim=500 200 500 200, clip]{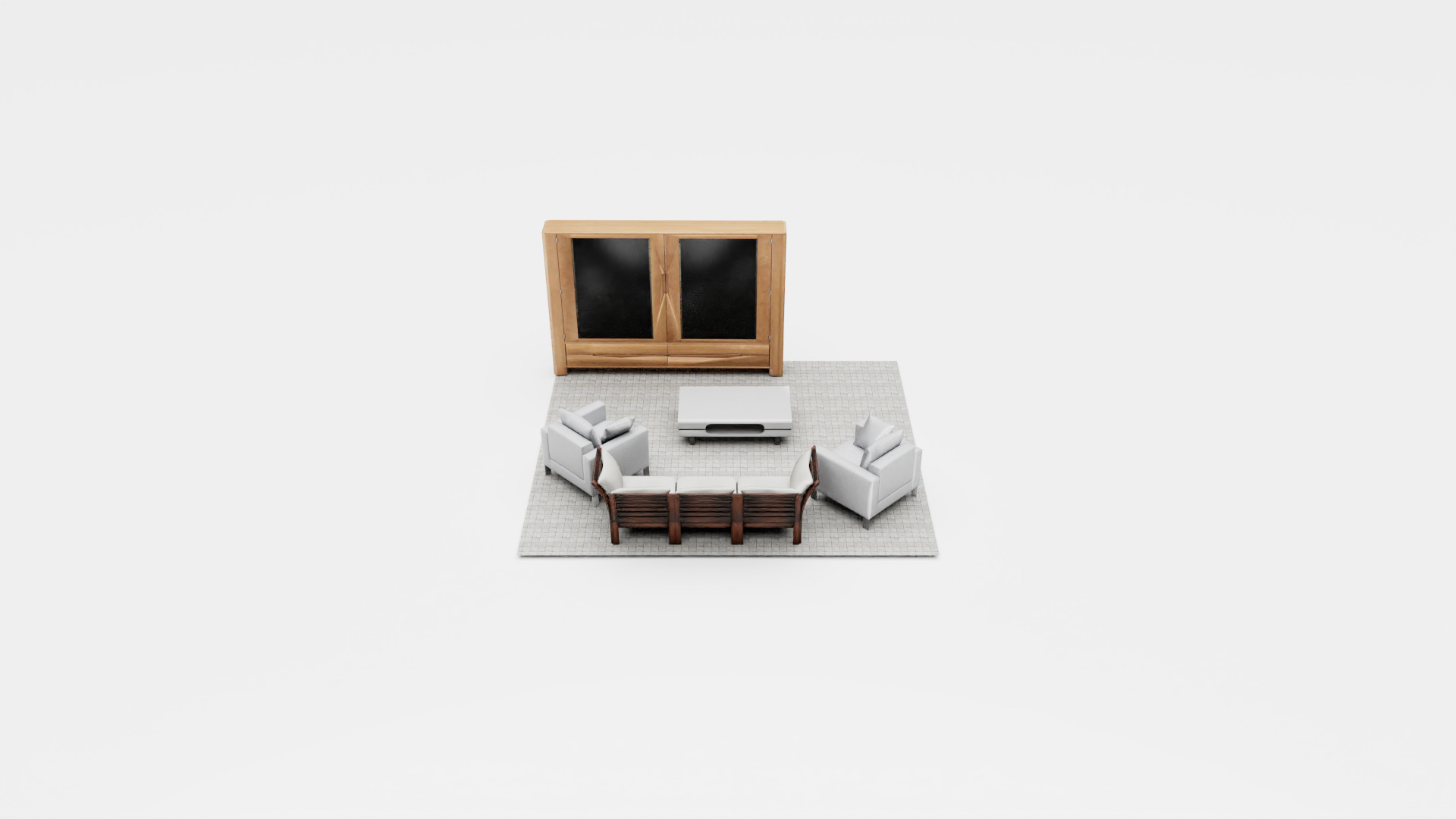}
    \end{subfigure}%
    \begin{subfigure}[b]{0.16\linewidth}
		\centering
		\includegraphics[width=\linewidth, trim=500 200 500 200, clip]{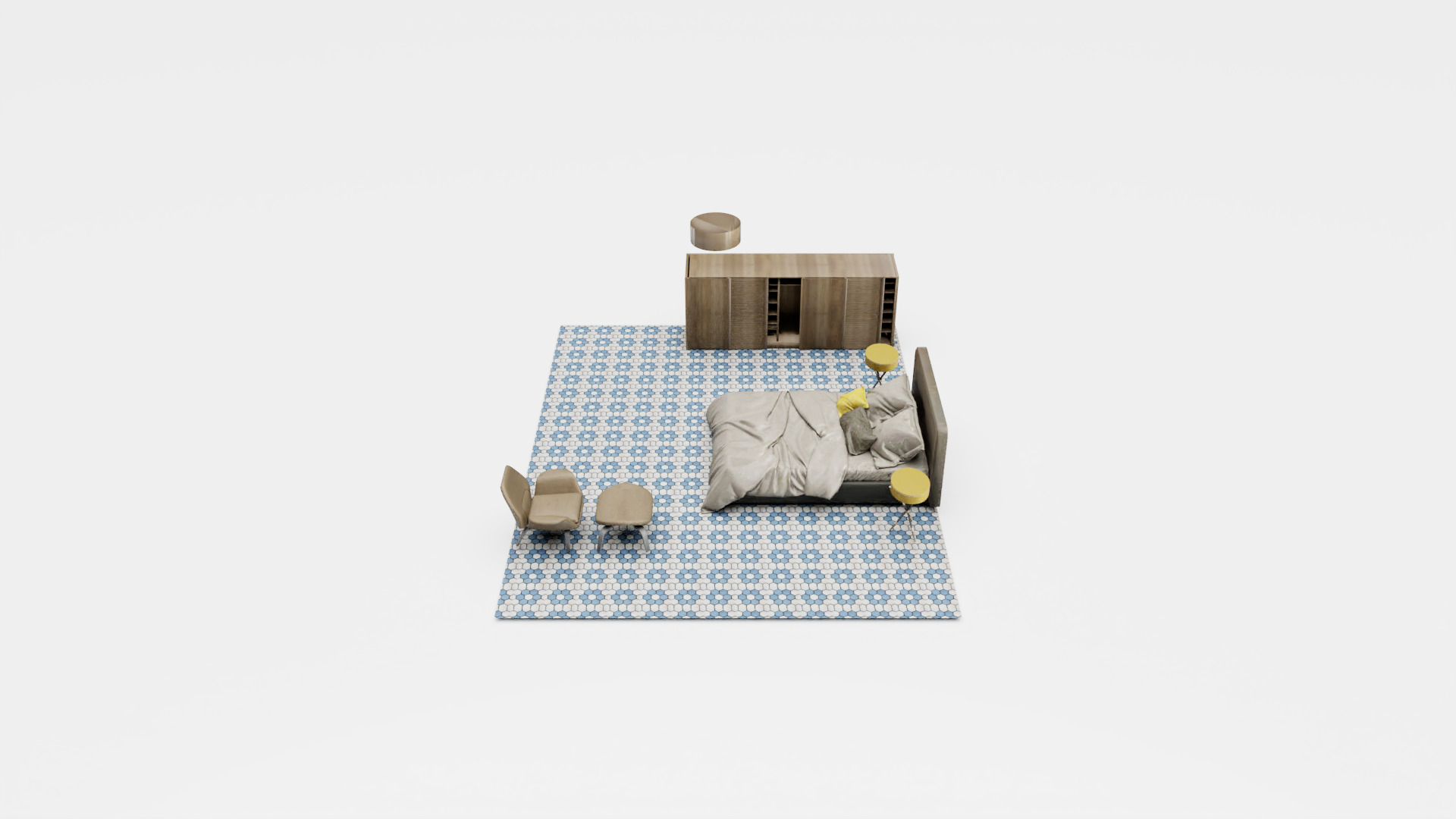}
    \end{subfigure}%
    \begin{subfigure}[b]{0.16\linewidth}
		\centering
		\includegraphics[width=\linewidth, trim=500 200 500 200, clip]{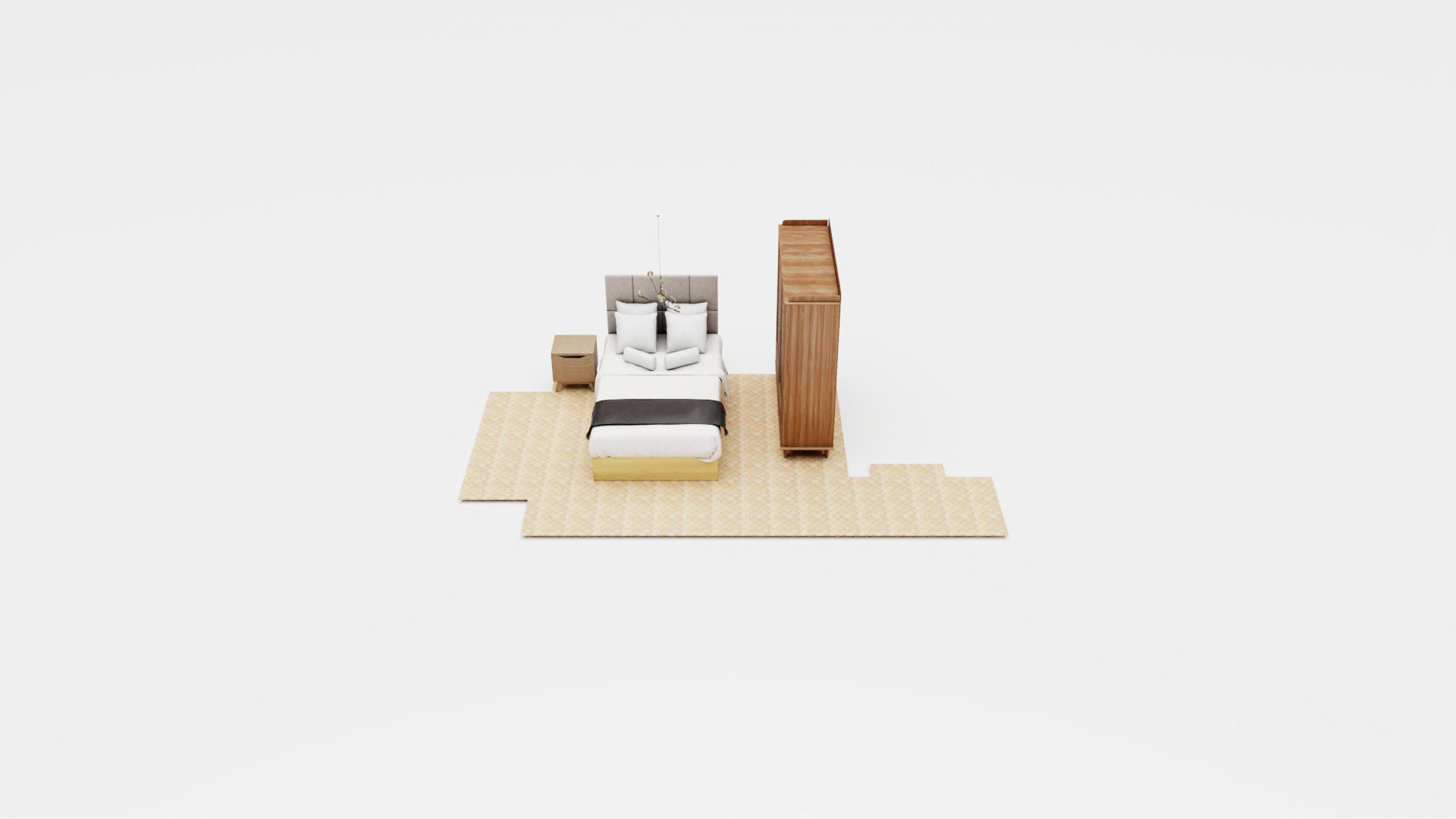}
    \end{subfigure}%
    \begin{subfigure}[b]{0.16\linewidth}
		\centering
		\includegraphics[width=\linewidth, trim=500 200 500 200, clip]{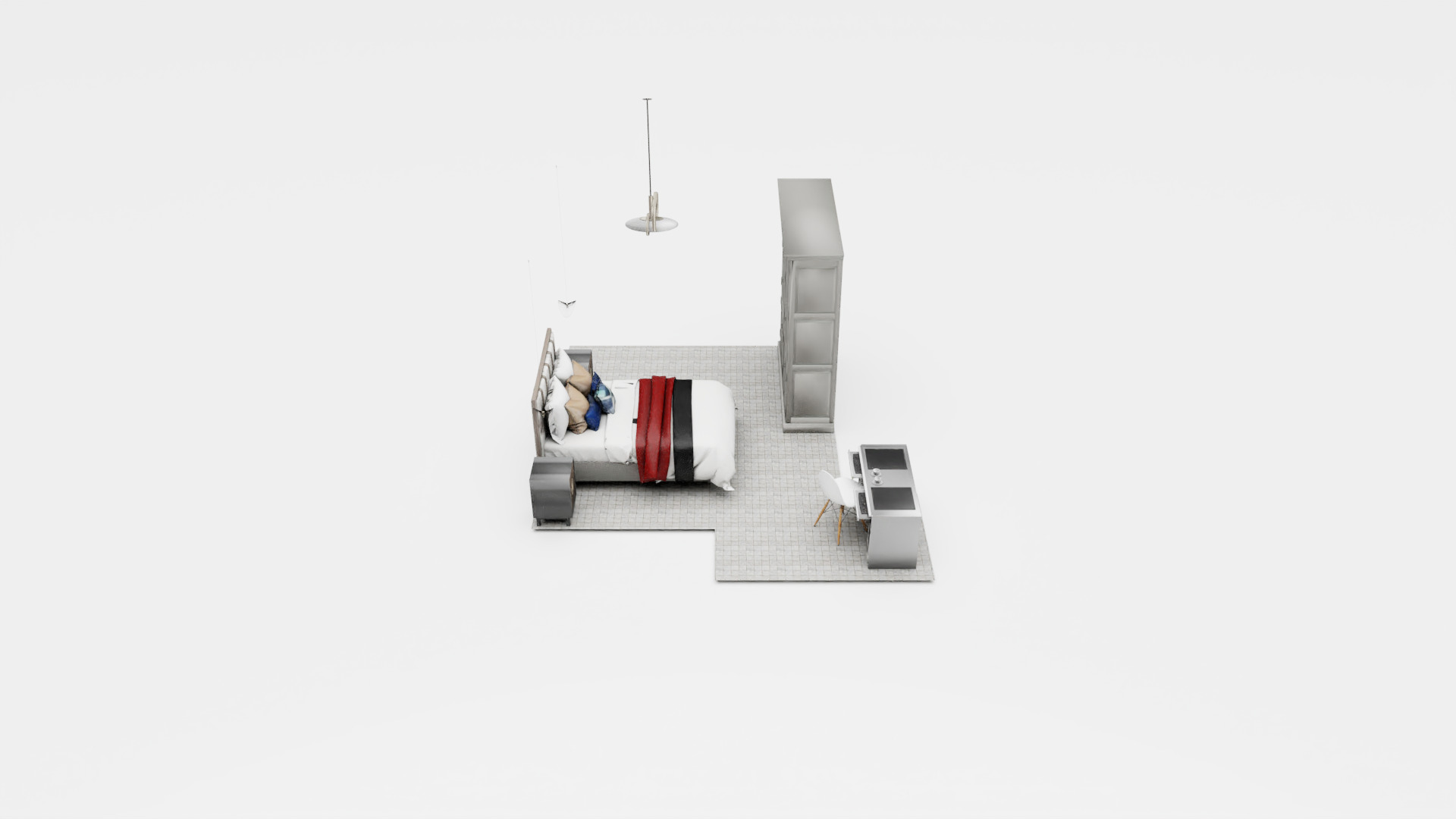}
    \end{subfigure}%
    \begin{subfigure}[b]{0.16\linewidth}
		\centering
		\includegraphics[width=\linewidth, trim=500 200 500 200, clip]{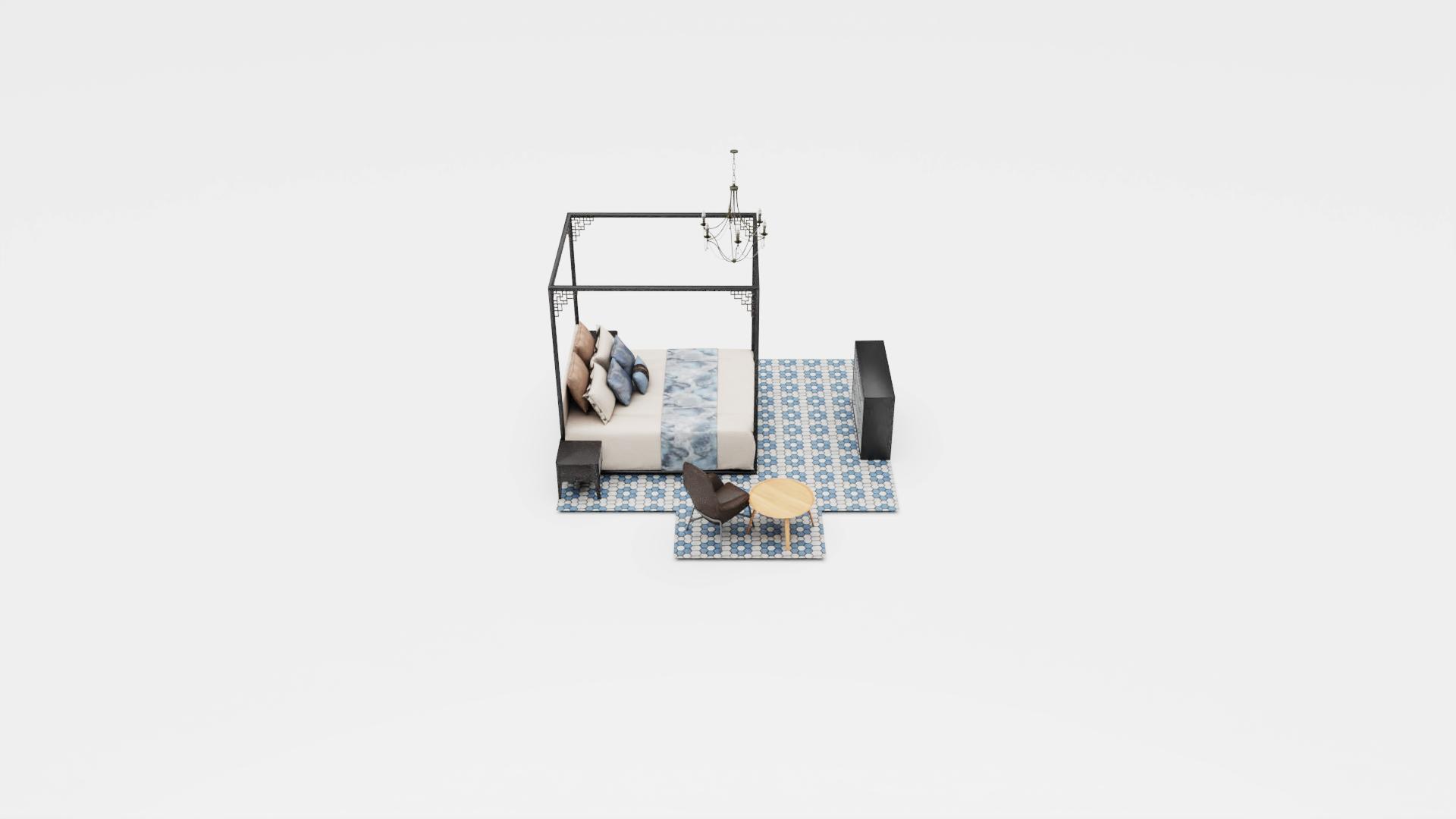}
    \end{subfigure}%
    \hfill%
    \vskip\baselineskip%
    \vspace{-1.5em}
    \hfill%
    \begin{subfigure}[b]{0.16\linewidth}
		\centering
        \small TV-stand
    \end{subfigure}%
    \begin{subfigure}[b]{0.16\linewidth}
		\centering
        \small Bookshelf
    \end{subfigure}%
    \begin{subfigure}[b]{0.16\linewidth}
		\centering
        \small Sofa
    \end{subfigure}%
    \begin{subfigure}[b]{0.16\linewidth}
		\centering
        \small Wardrobe
    \end{subfigure}%
    \begin{subfigure}[b]{0.16\linewidth}
		\centering
        \small Chair
    \end{subfigure}%
    \begin{subfigure}[b]{0.16\linewidth}
		\centering
        \small Coffee table
    \end{subfigure}%
    \vspace{-1.2em}
    \vskip\baselineskip%
    \caption{{\bf Object Placement}. Starting from a partially complete
    scene, the user specifies an object to be added in the scene and our model
    places it at a reasonable position. The first rows illustrates the
    starting scene and the second row the generated scened using the
    user specified object (third row).}
    \label{fig:object_placement}
\end{figure}

\subsection{Object Placement}
Finally, we showcase the ability of our model to add a specific object in a
scene on demand.  \figref{fig:object_placement} illustrates the original scene
(first row) and the complete scene (second row) using the user specified object
(third row). To perform this task, we condition on the given scene and instead
of sampling from the predicted object category distribution, we use the user
provided object category and sample the rest of the object attributes \ie
translation, size and orientation. Also in this task, we note that the
generated objects are realistic and match the room layout.

\section{Scene Synthesis}

In this section, we provide additional qualitative results for our scene
synthesis experiment on the four 3D-FRONT rooms. Moreover, since, we repeat the FID score and
classification accuracy computation $10$ times, in
\tabref{tab:scene_synthesis_quantitative_supp}, we also report the standard
deviation for completeness.
\begin{table}[!h]
    \centering
    \resizebox{\columnwidth}{!}{
    \begin{tabular}{l|ccc|ccc|ccc}
        \toprule
        \multicolumn{1}{c}{\,} & \multicolumn{3}{c}{FID Score ($\downarrow$)}& \multicolumn{3}{c}{Scene Classification Accuracy}& \multicolumn{3}{c}{Category KL Divergence ($\downarrow$)}\\
        \toprule
        \multicolumn{1}{c}{\,} & FastSynth & SceneFormer & Ours & FastSynth & SceneFormer & Ours & FastSynth& SceneFormer & Ours \\
        \midrule
        Bedrooms & 40.89 $\pm$ 0.5098 & 43.17 $\pm$ 0.6921 & {\bf{38.39}} $\pm$ 0.3392 &
        0.883 $\pm$ 0.0010 & 0.945 $\pm$ 0.0009 & {\bf{0.562}} $\pm$ 0.0228 & 0.0064 & \bf{0.0052} & 0.0085\\
        Living   & 61.67 $\pm$ 1.2136 & 69.54 $\pm$ 0.9542 & {\bf{33.14}} $\pm$ 0.4204 &
        0.945 $\pm$ 0.0010 & 0.972 $\pm$ 0.0010 & {\bf{0.516}} $\pm$ 0.0075 & 0.0176 & 0.0313 & \bf{0.0034}\\
        Dining   & 55.83 $\pm$ 1.0078 & 67.04 $\pm$ 1.3043 & {\bf{29.23}} $\pm$ 0.3533 &
        0.935 $\pm$ 0.0019 & 0.941 $\pm$ 0.0008 & {\bf{0.477}} $\pm$ 0.0027 & 0.0518 & 0.0368 & \bf{0.0061}\\
        Library  & 37.72 $\pm$ 0.4501 & 55.34 $\pm 0.1056 $ & {\bf{35.24}} $\pm$ 0.2683 &
        0.815 $\pm$ 0.0032 & 0.880 $\pm$ 0.0009 & {\bf{0.521}} $\pm$ 0.0048 & 0.0431 & 0.0232 & \bf{0.0098}\\
        \bottomrule
    \end{tabular}}
    \vspace{0.2em}
    \caption{\small {\bf Quantitative Comparison.} We report the FID score
    ($\downarrow$) at $256^2$ pixels, the KL divergence ($\downarrow$) between the
    distribution of object categories of synthesized and real scenes and the real vs. synthetic classification
    accuracy for all methods. Classification accuracy closer to $0.5$ 
    is better.}
    \vspace{-1.2em}
    \label{tab:scene_synthesis_quantitative_supp}
\end{table}

\begin{figure}[!h]
    \centering
    \begin{subfigure}[b]{0.20\linewidth}
		\centering
        \small Scene Layout
    \end{subfigure}%
    \begin{subfigure}[b]{0.20\linewidth}
		\centering
        \small Training Sample
    \end{subfigure}%
    \begin{subfigure}[b]{0.20\linewidth}
		\centering
        \small FastSynth
    \end{subfigure}%
    \begin{subfigure}[b]{0.20\linewidth}
		\centering
        \small SceneFormer
    \end{subfigure}%
    \begin{subfigure}[b]{0.20\linewidth}
        \centering
        \small Ours
    \end{subfigure}
    \hfill%
    \vskip\baselineskip%
    \vspace{-1.5em}
    \hfill%
    \begin{subfigure}[b]{0.20\linewidth}
		\centering
		\includegraphics[width=0.8\linewidth]{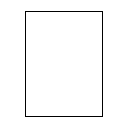}
    \end{subfigure}%
        \begin{subfigure}[b]{0.20\linewidth}
		\centering
		\includegraphics[width=\linewidth, trim=500 200 500 100, clip]{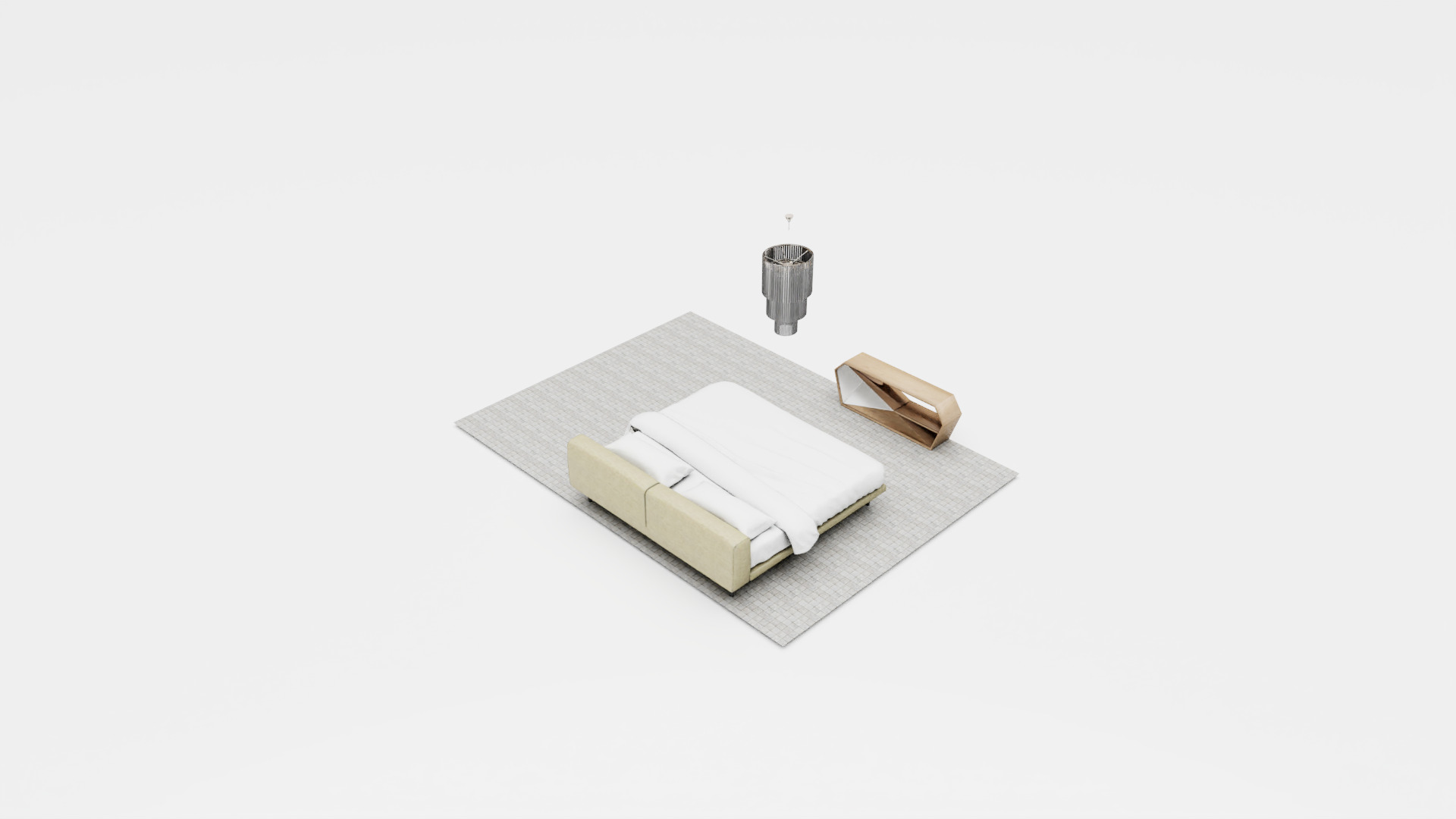}
    \end{subfigure}%
        \begin{subfigure}[b]{0.20\linewidth}
		\centering
		\includegraphics[width=\linewidth, trim=500 200 500 100, clip]{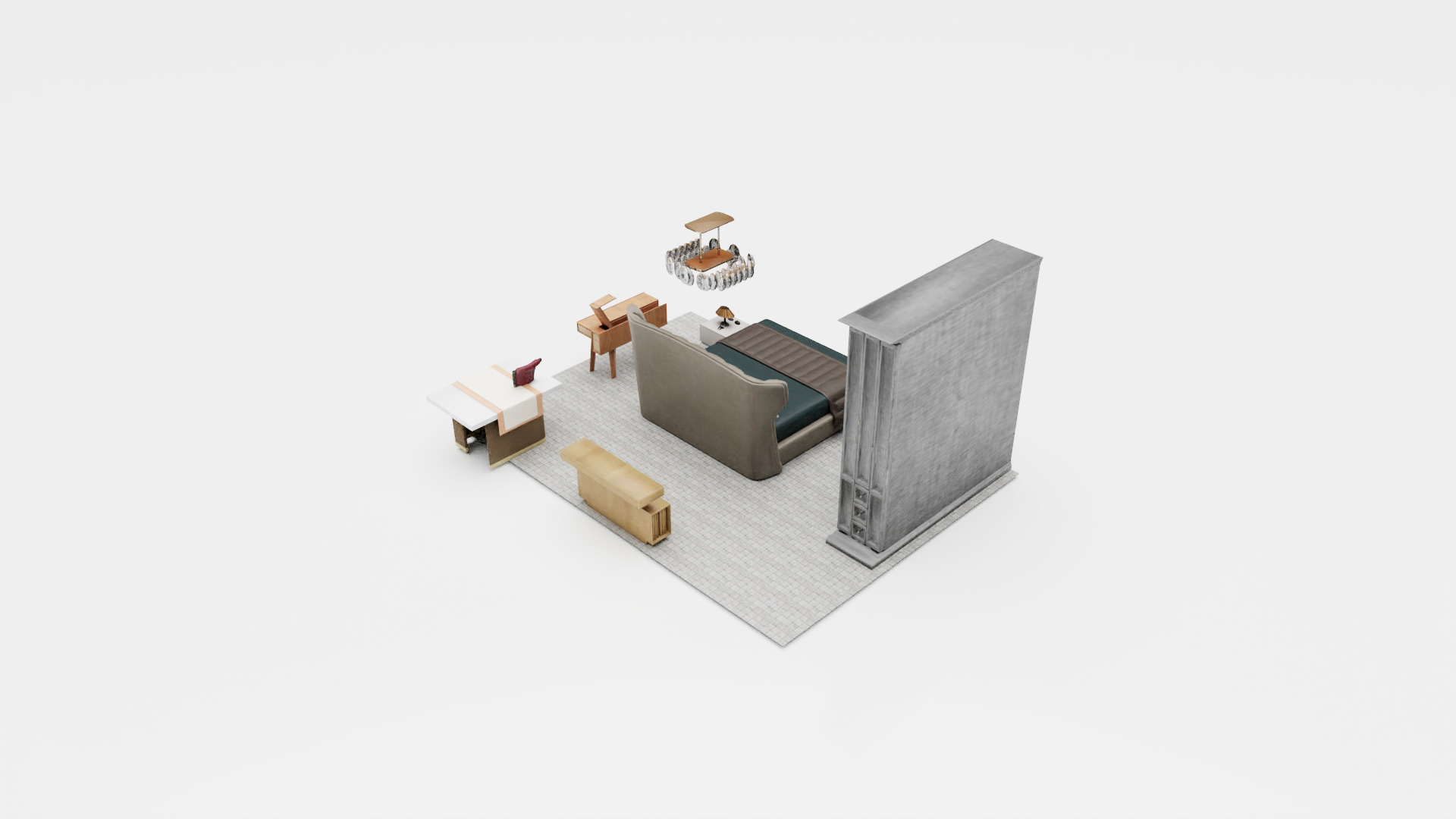}
    \end{subfigure}%
        \begin{subfigure}[b]{0.20\linewidth}
		\centering
		\includegraphics[width=\linewidth, trim=500 200 500 100, clip]{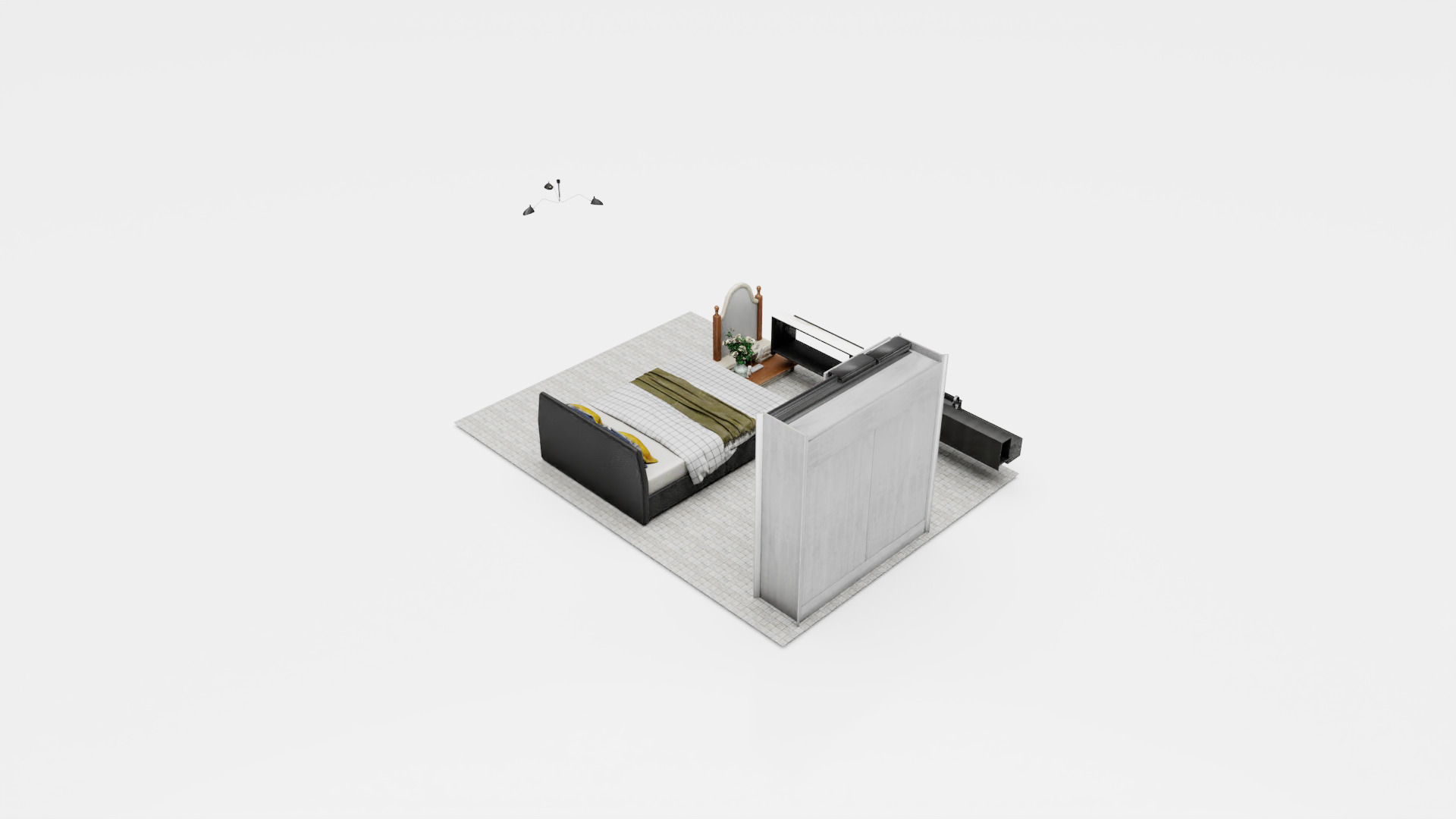}
    \end{subfigure}%
    \begin{subfigure}[b]{0.20\linewidth}
		\centering
		\includegraphics[width=\linewidth, trim=500 200 500 100, clip]{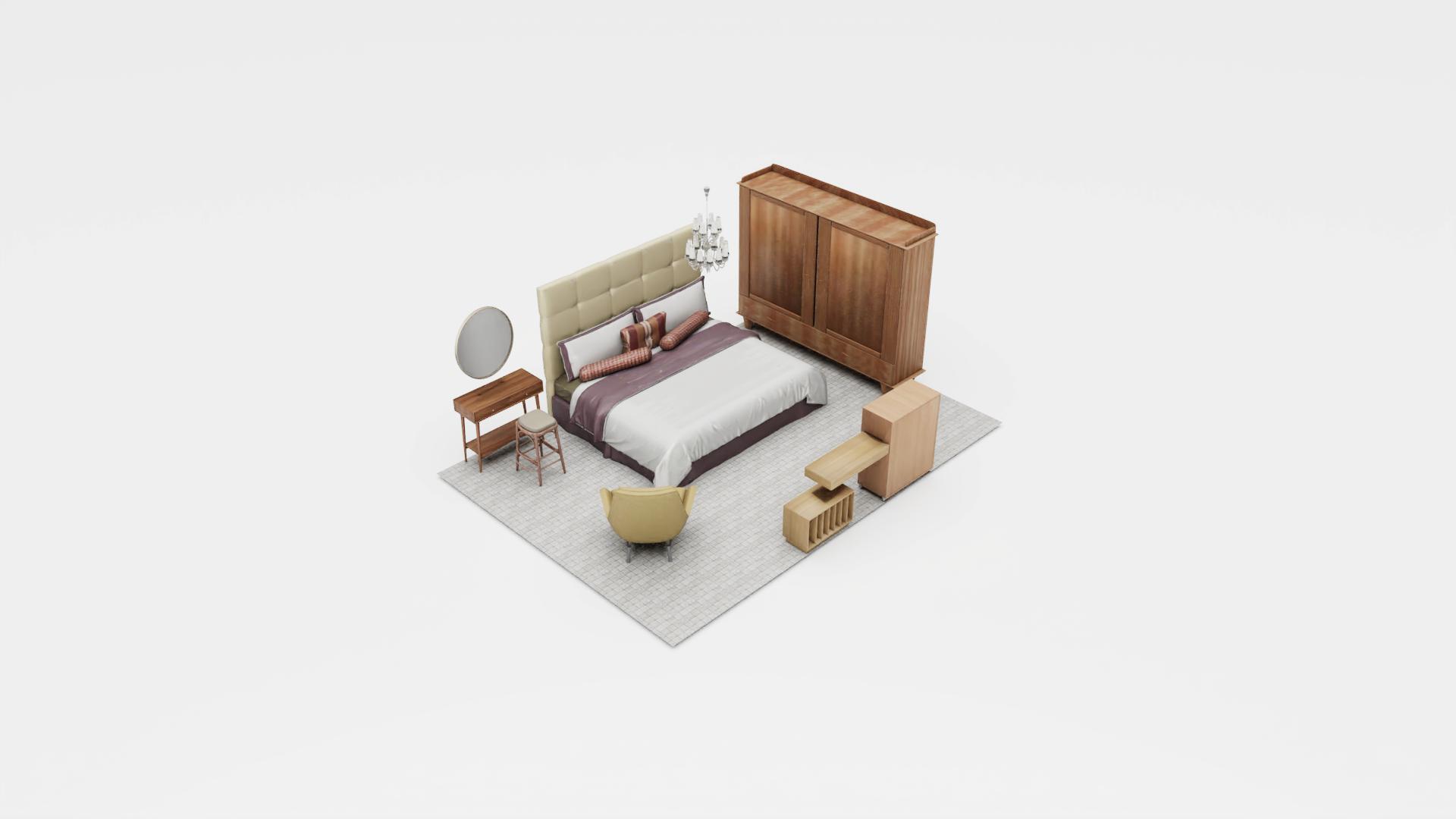}
    \end{subfigure}%
    \vskip\baselineskip%
    \vspace{-2.2em}
    \vskip\baselineskip%
    \begin{subfigure}[b]{0.20\linewidth}
		\centering
		\includegraphics[width=0.8\linewidth]{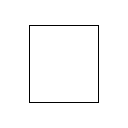}
    \end{subfigure}%
        \begin{subfigure}[b]{0.20\linewidth}
		\centering
		\includegraphics[width=\linewidth, trim=500 200 500 100, clip]{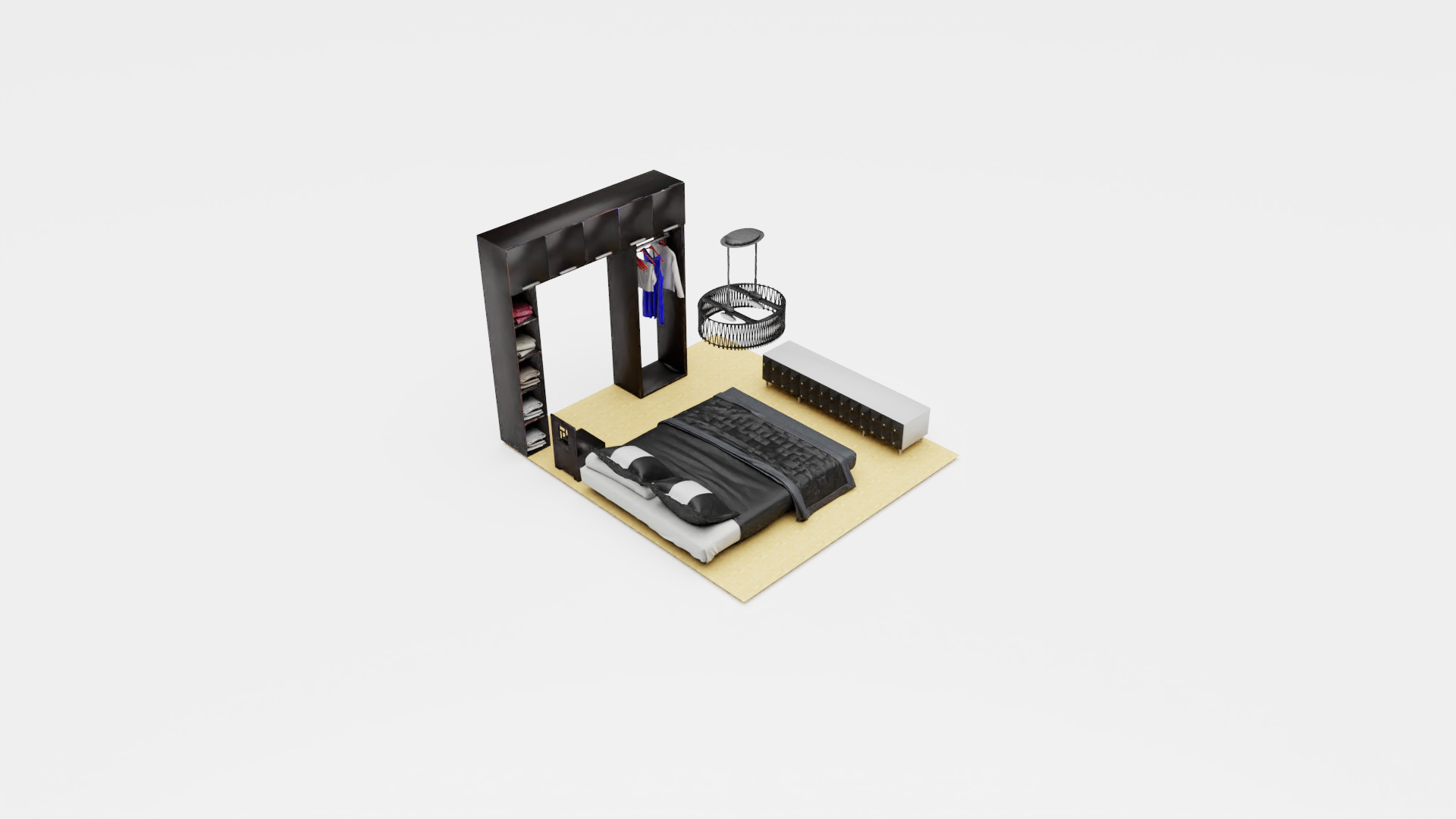}
    \end{subfigure}%
        \begin{subfigure}[b]{0.20\linewidth}
		\centering
		\includegraphics[width=\linewidth, trim=500 200 500 100, clip]{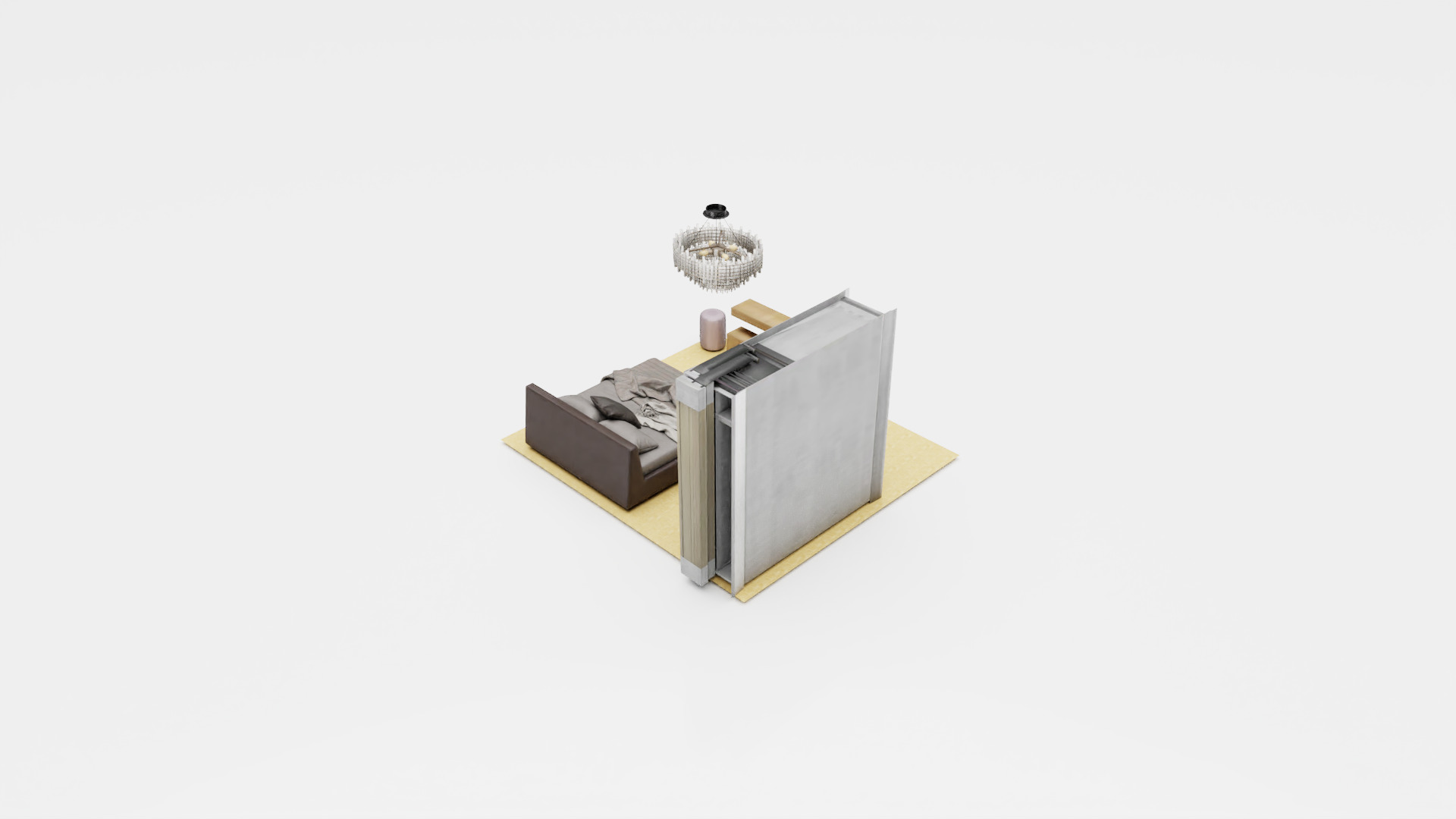}
    \end{subfigure}%
        \begin{subfigure}[b]{0.20\linewidth}
		\centering
		\includegraphics[width=\linewidth, trim=500 200 500 100, clip]{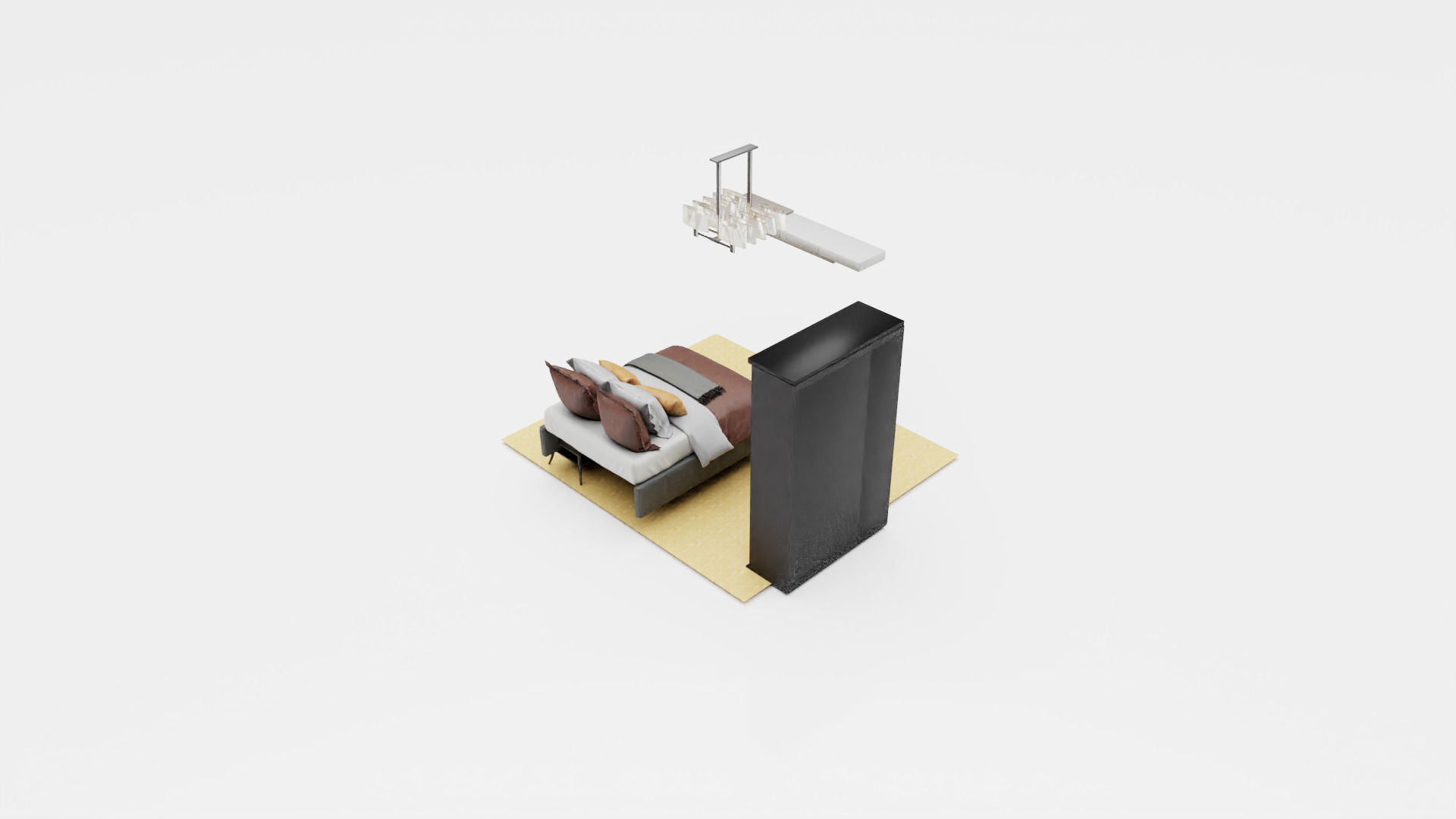}
    \end{subfigure}%
    \begin{subfigure}[b]{0.20\linewidth}
		\centering
		\includegraphics[width=\linewidth, trim=500 200 500 100, clip]{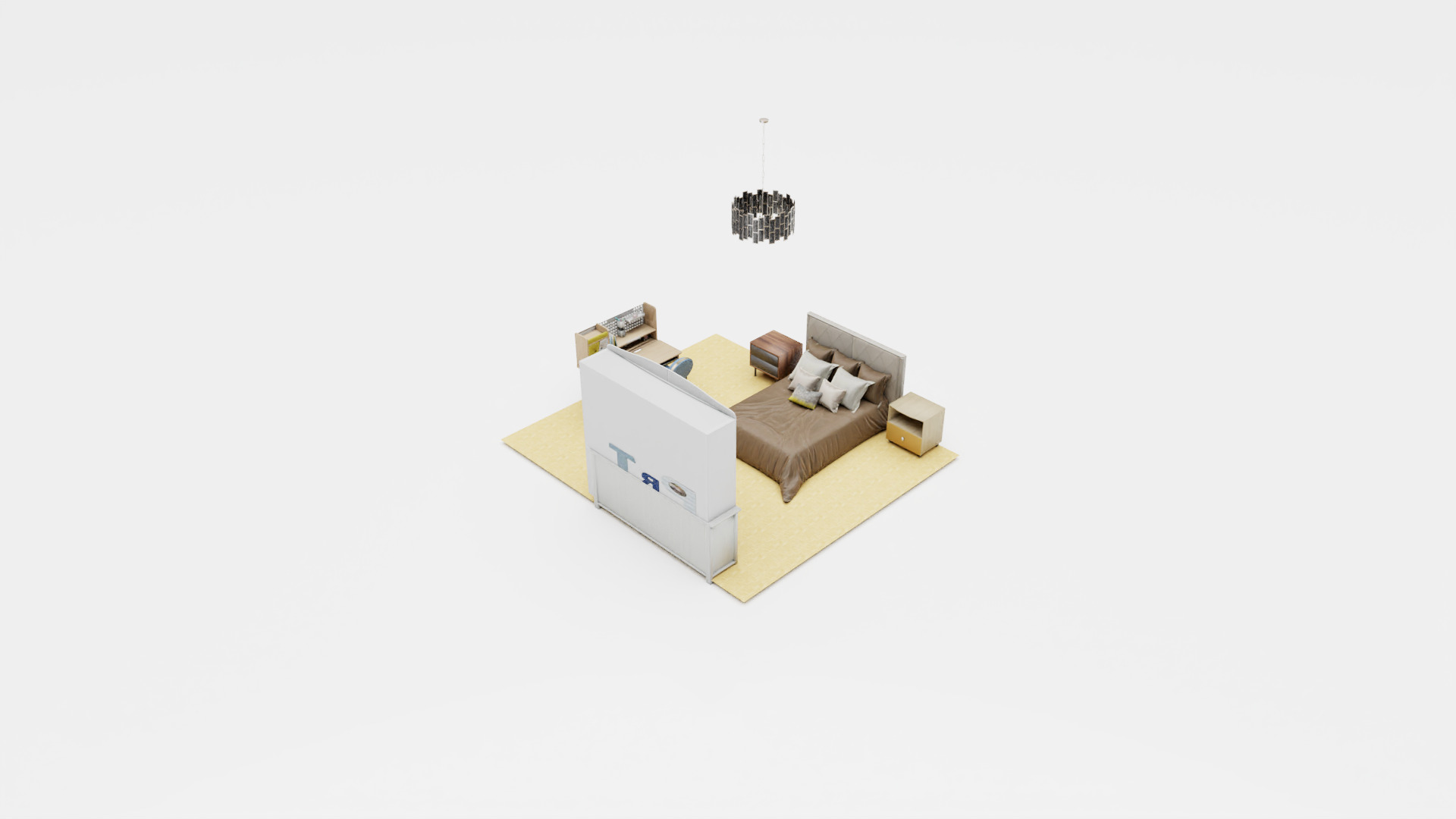}
    \end{subfigure}%
    \vskip\baselineskip%
    \vspace{-2.2em}
    \vskip\baselineskip%
    \begin{subfigure}[b]{0.20\linewidth}
		\centering
		\includegraphics[width=0.8\linewidth]{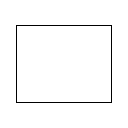}
    \end{subfigure}%
        \begin{subfigure}[b]{0.20\linewidth}
		\centering
		\includegraphics[width=\linewidth, trim=500 200 500 100, clip]{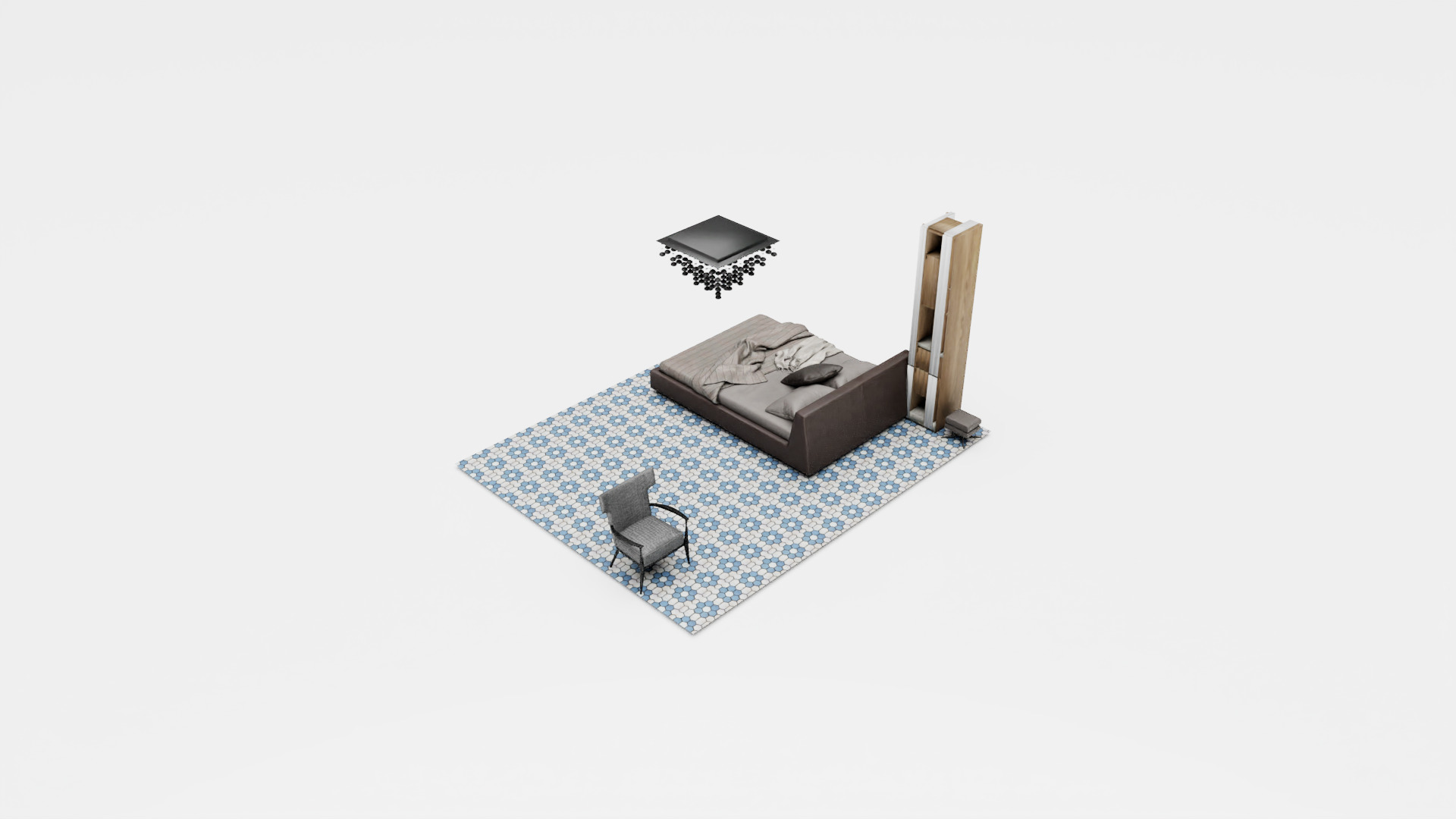}
    \end{subfigure}%
        \begin{subfigure}[b]{0.20\linewidth}
		\centering
		\includegraphics[width=\linewidth, trim=500 200 500 100, clip]{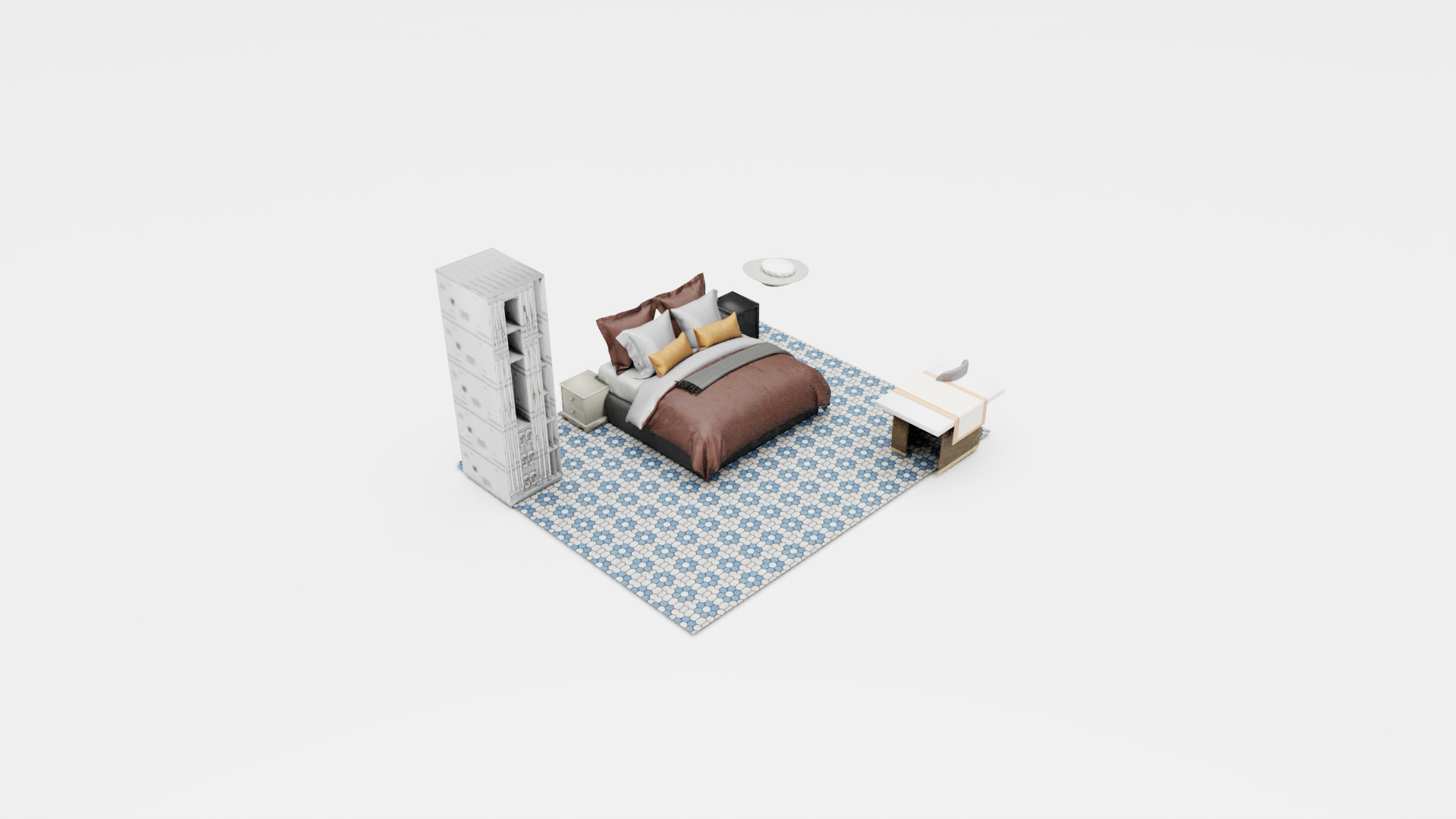}
    \end{subfigure}%
        \begin{subfigure}[b]{0.20\linewidth}
		\centering
		\includegraphics[width=\linewidth, trim=500 200 500 100, clip]{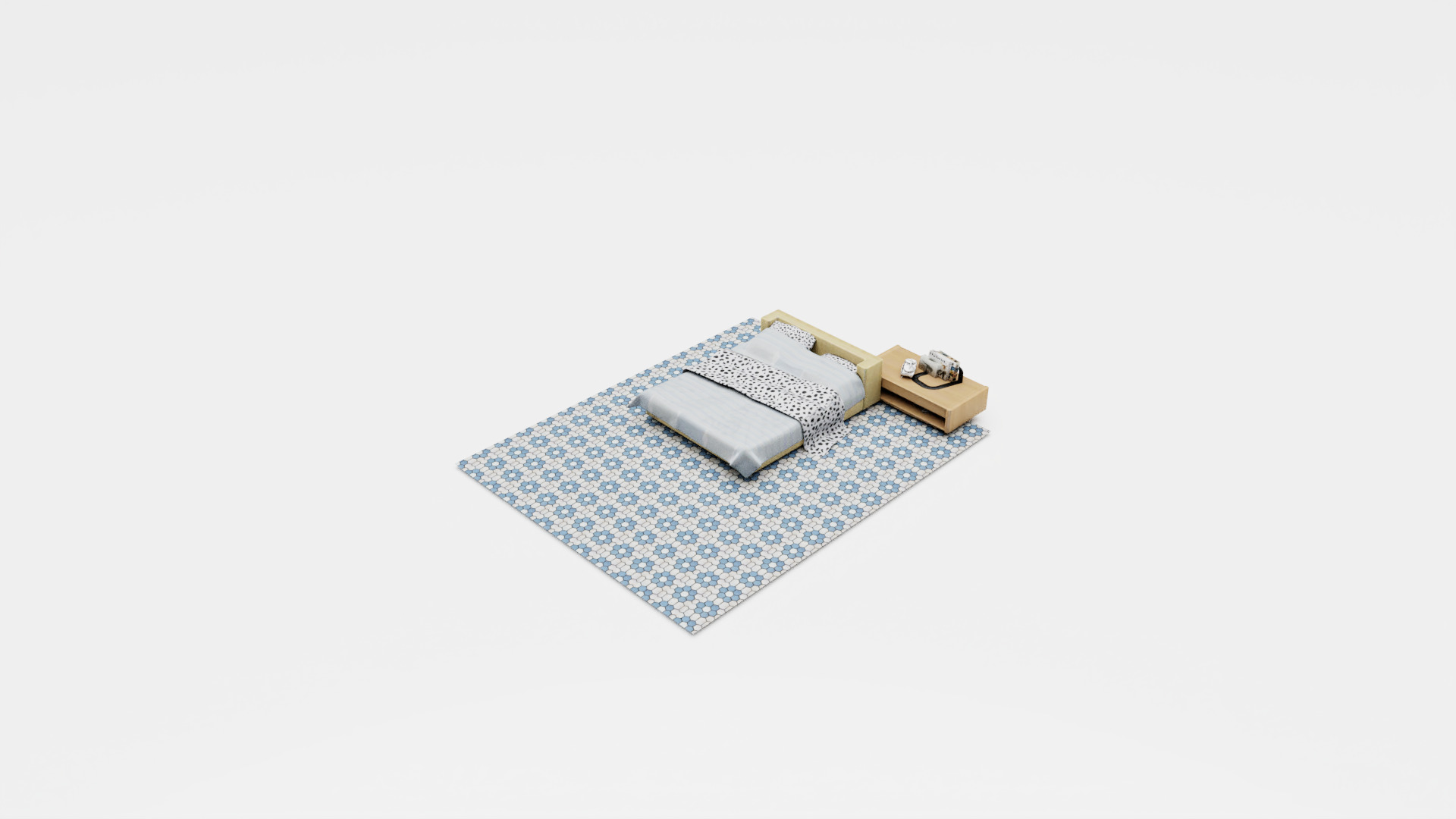}
    \end{subfigure}%
    \begin{subfigure}[b]{0.20\linewidth}
		\centering
		\includegraphics[width=\linewidth, trim=500 200 500 100, clip]{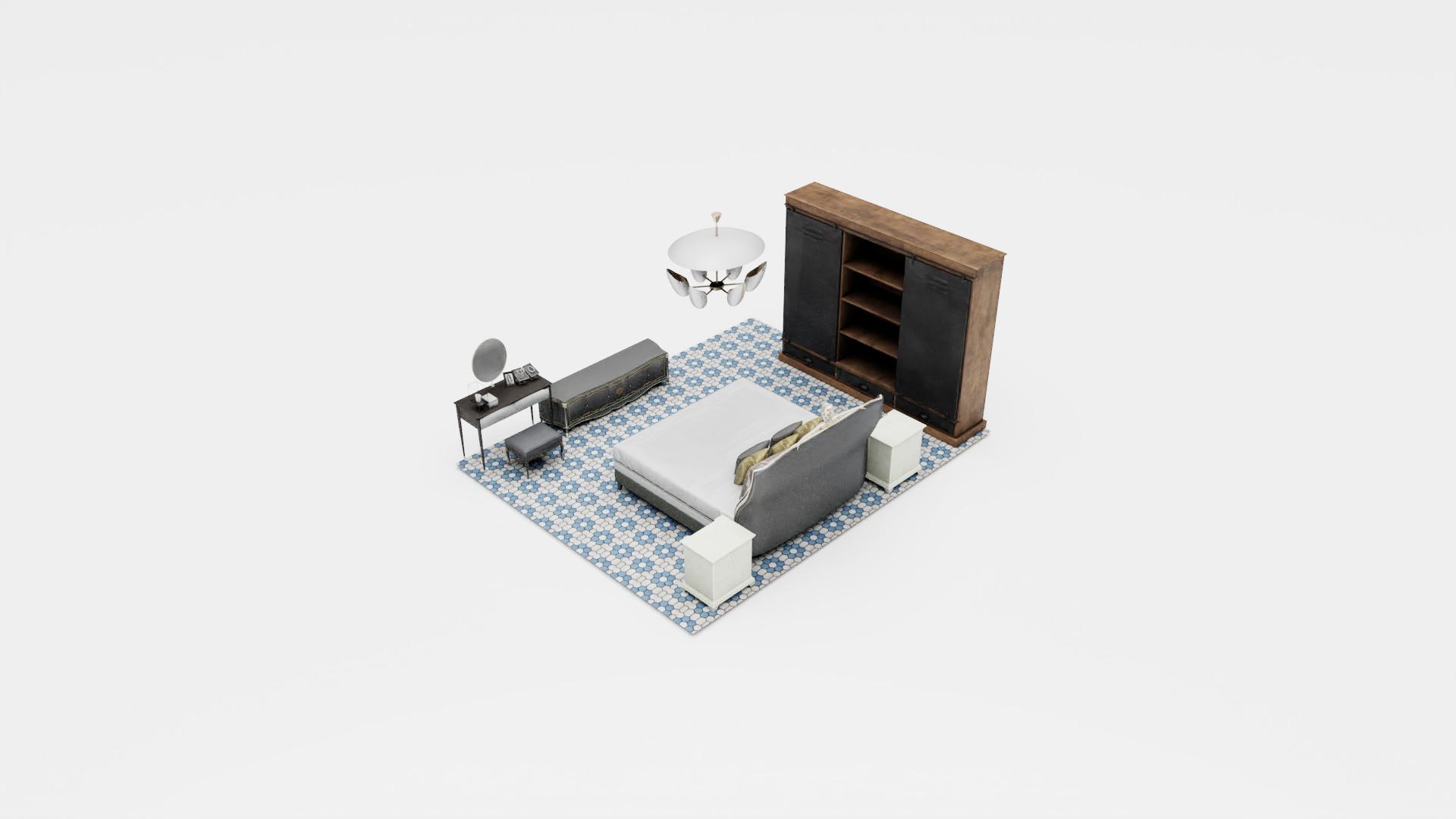}
    \end{subfigure}%
    \vskip\baselineskip%
    \vspace{-2.2em}
    \vskip\baselineskip%
    \begin{subfigure}[b]{0.20\linewidth}
		\centering
		\includegraphics[width=0.8\linewidth]{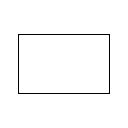}
    \end{subfigure}%
        \begin{subfigure}[b]{0.20\linewidth}
		\centering
		\includegraphics[width=\linewidth, trim=500 200 500 100, clip]{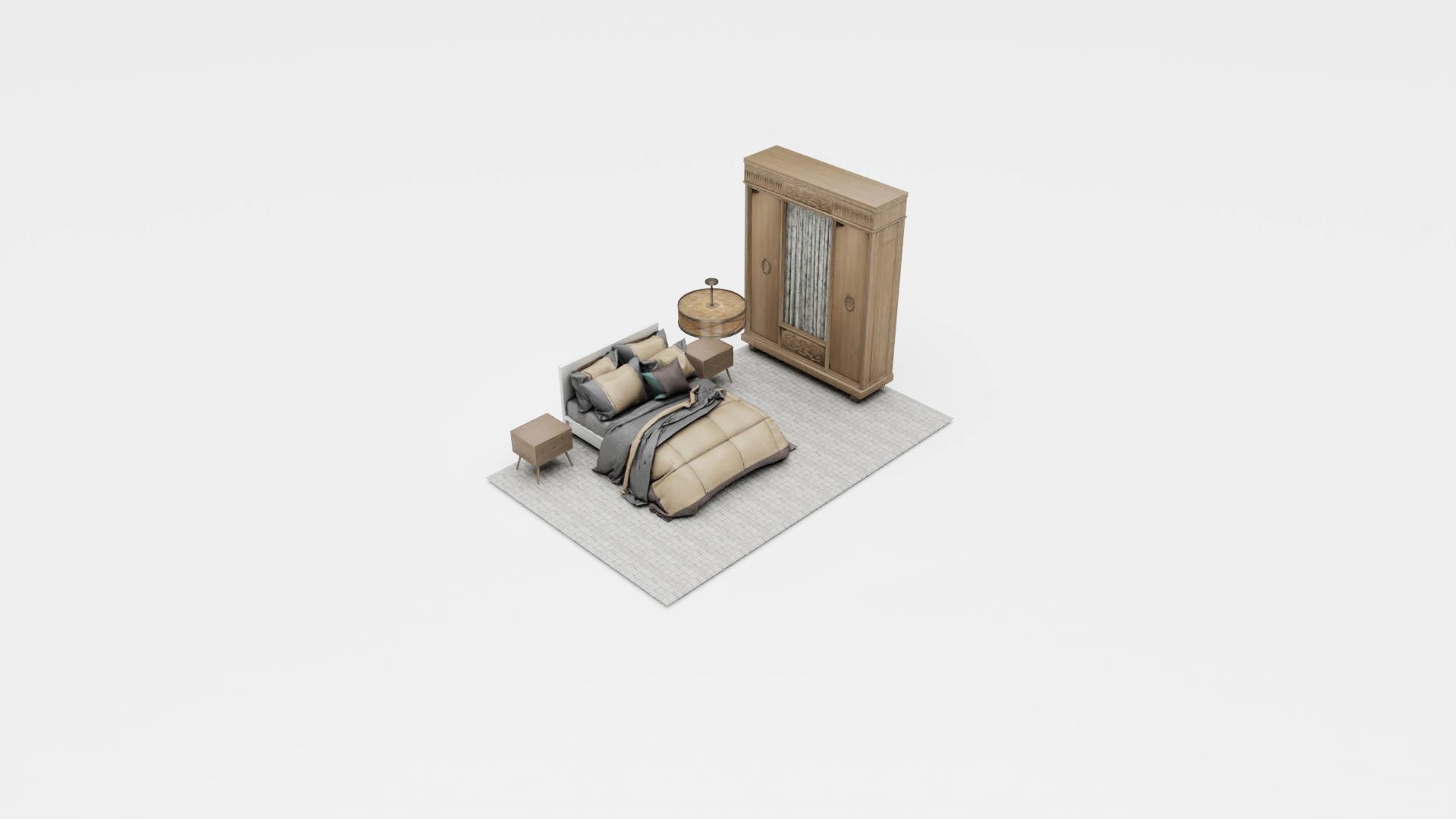}
    \end{subfigure}%
        \begin{subfigure}[b]{0.20\linewidth}
		\centering
		\includegraphics[width=\linewidth, trim=500 200 500 100, clip]{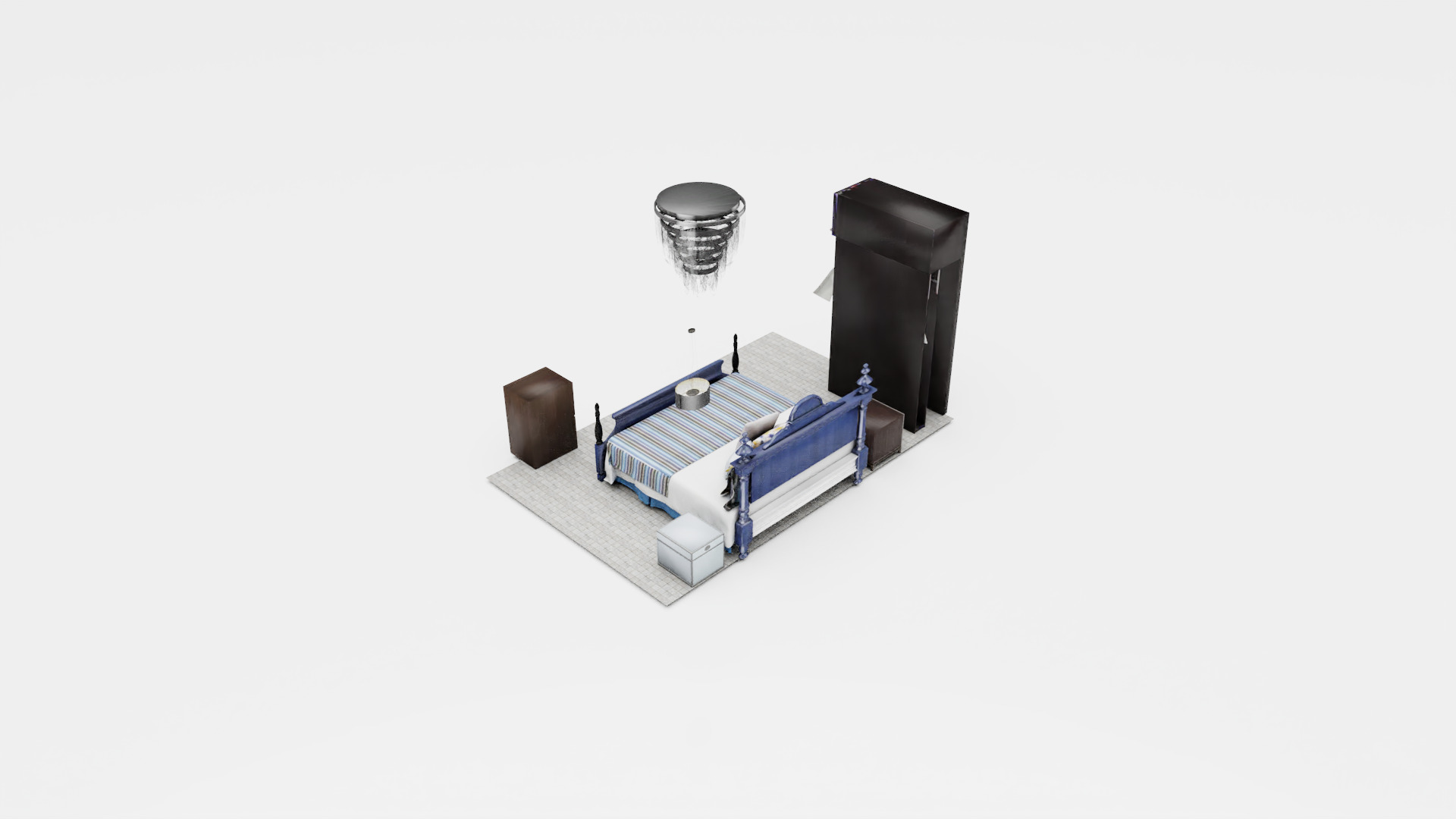}
    \end{subfigure}%
        \begin{subfigure}[b]{0.20\linewidth}
		\centering
		\includegraphics[width=\linewidth, trim=500 200 500 100, clip]{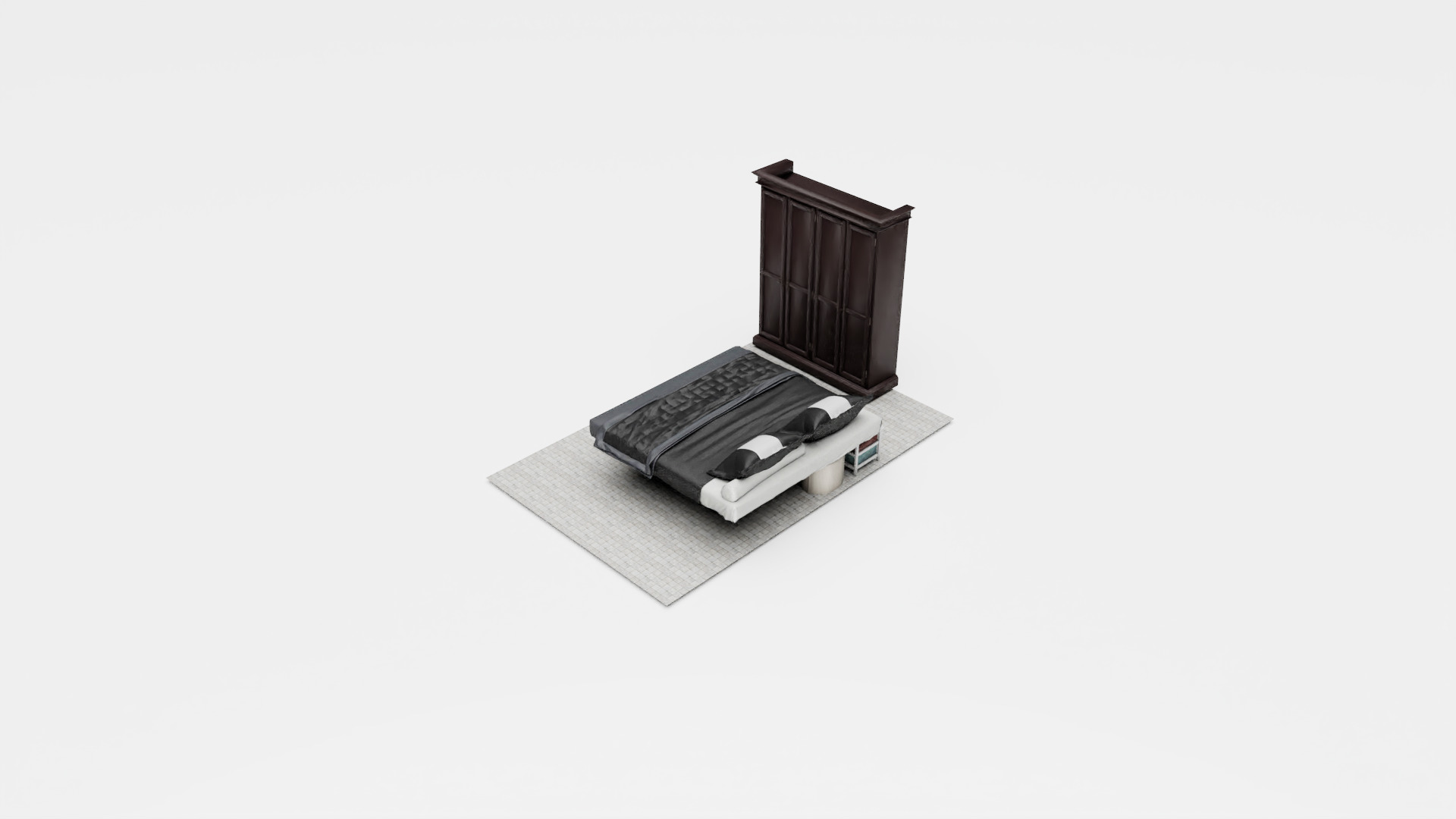}
    \end{subfigure}%
    \begin{subfigure}[b]{0.20\linewidth}
		\centering
		\includegraphics[width=\linewidth, trim=500 200 500 100, clip]{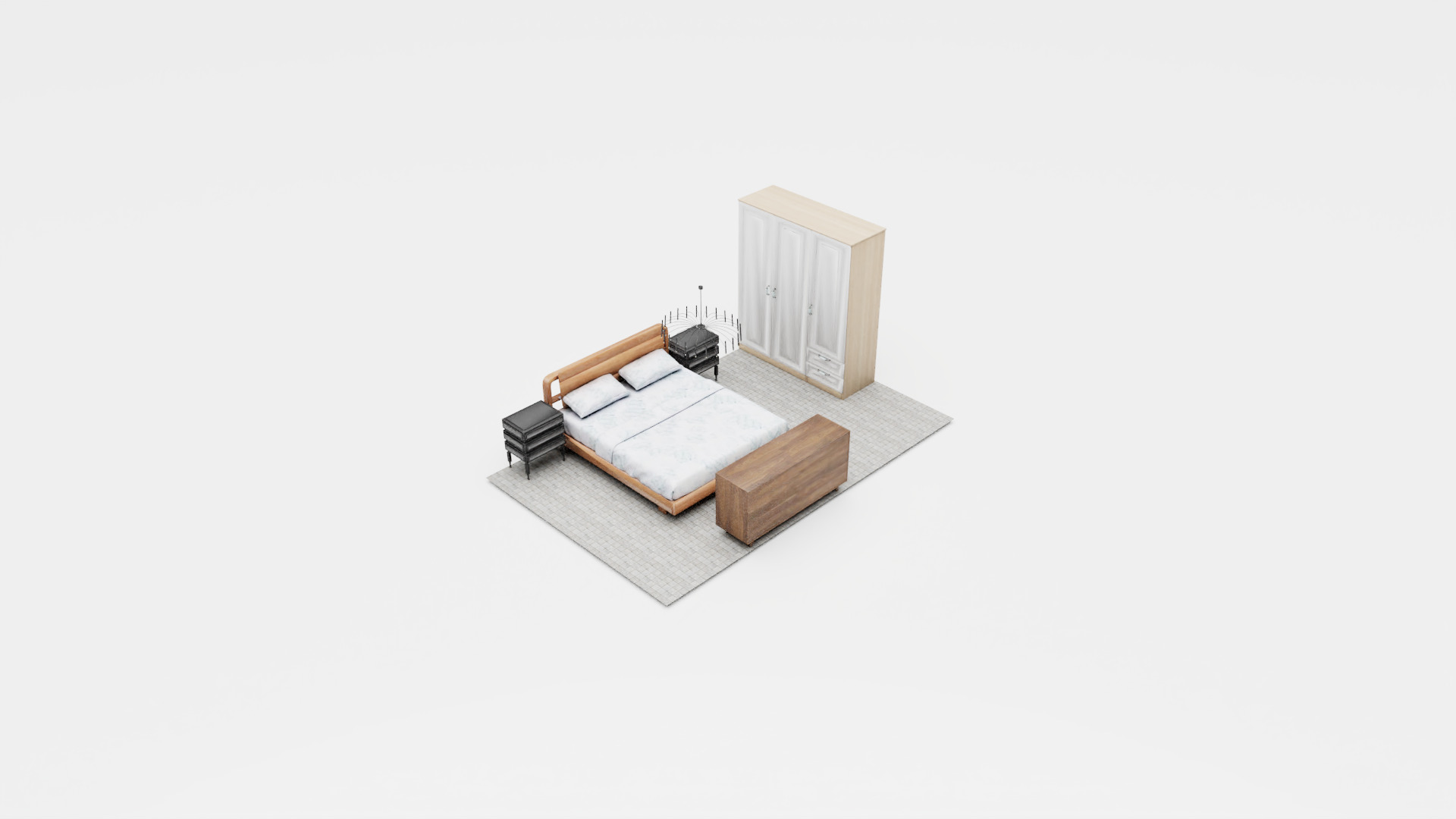}
    \end{subfigure}%
    \vskip\baselineskip%
    \vspace{-2.2em}
    \vskip\baselineskip%
    		            		            		            		        		        \vskip\baselineskip%
    \vspace{-2.2em}
    \vskip\baselineskip%
    \begin{subfigure}[b]{0.20\linewidth}
		\centering
		\includegraphics[width=0.8\linewidth]{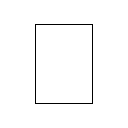}
    \end{subfigure}%
        \begin{subfigure}[b]{0.20\linewidth}
		\centering
		\includegraphics[width=\linewidth, trim=500 200 500 100, clip]{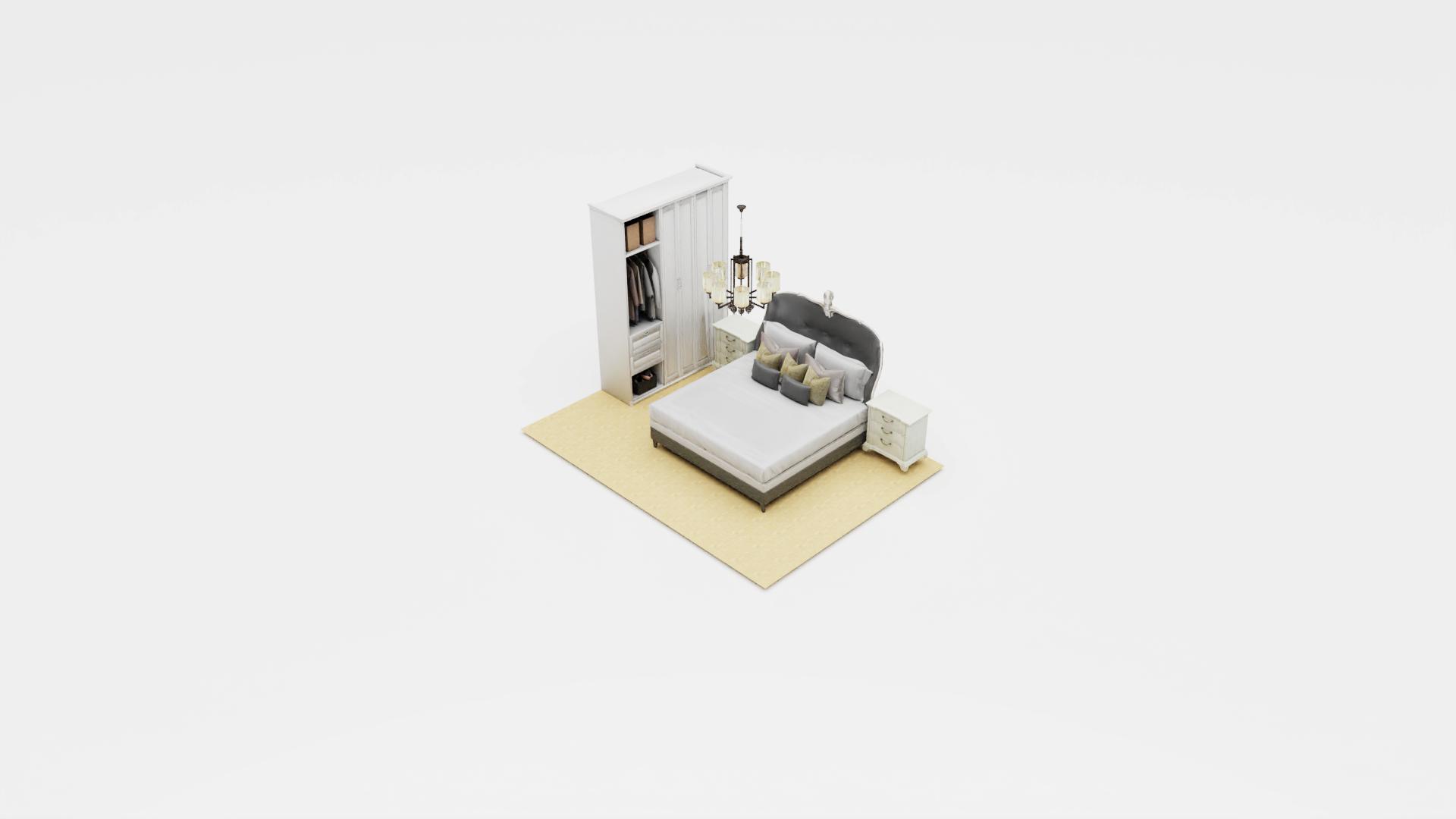}
    \end{subfigure}%
        \begin{subfigure}[b]{0.20\linewidth}
		\centering
		\includegraphics[width=\linewidth, trim=500 200 500 100, clip]{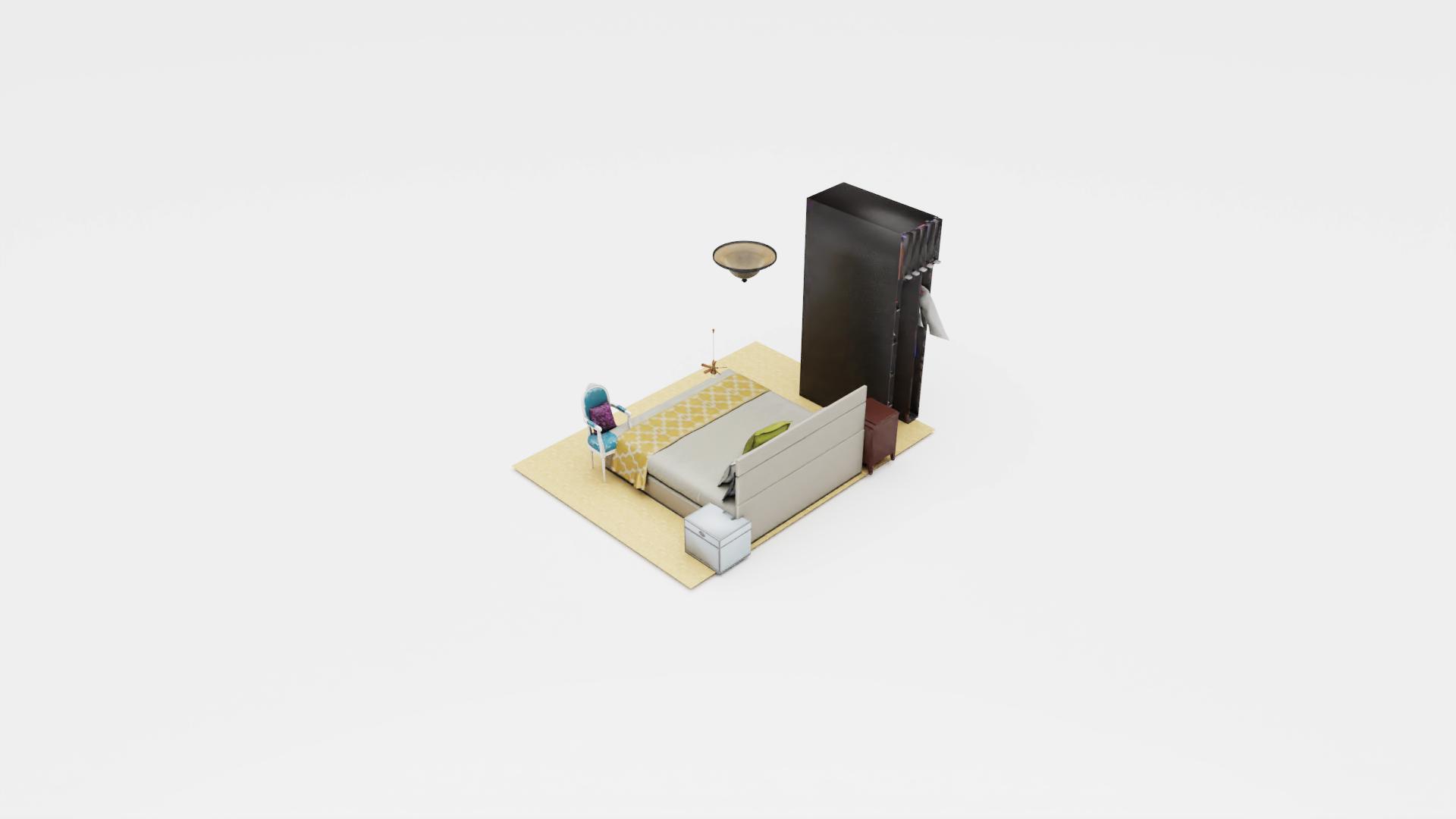}
    \end{subfigure}%
        \begin{subfigure}[b]{0.20\linewidth}
		\centering
		\includegraphics[width=\linewidth, trim=500 200 500 100, clip]{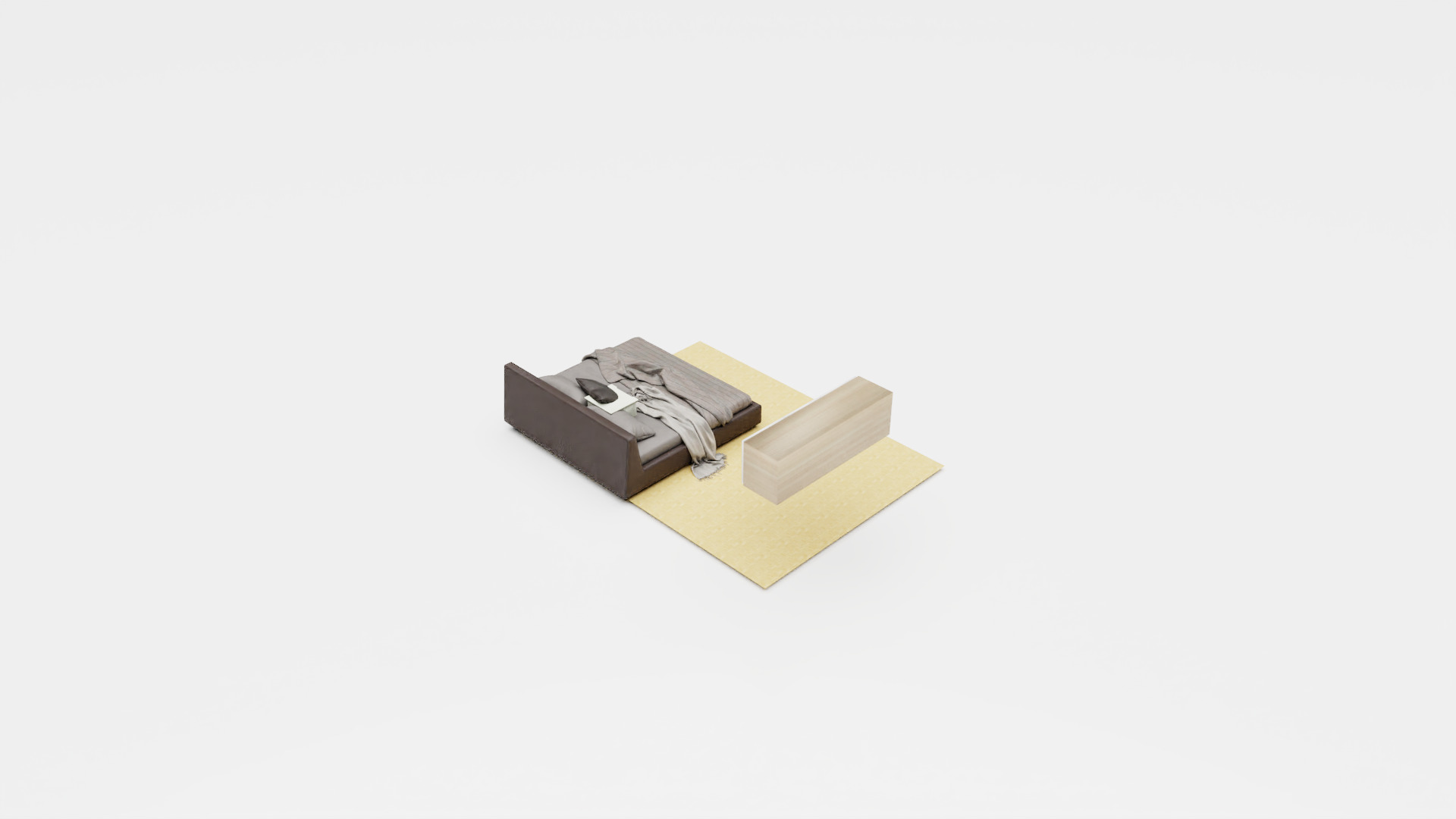}
    \end{subfigure}%
    \begin{subfigure}[b]{0.20\linewidth}
		\centering
		\includegraphics[width=\linewidth, trim=500 200 500 100, clip]{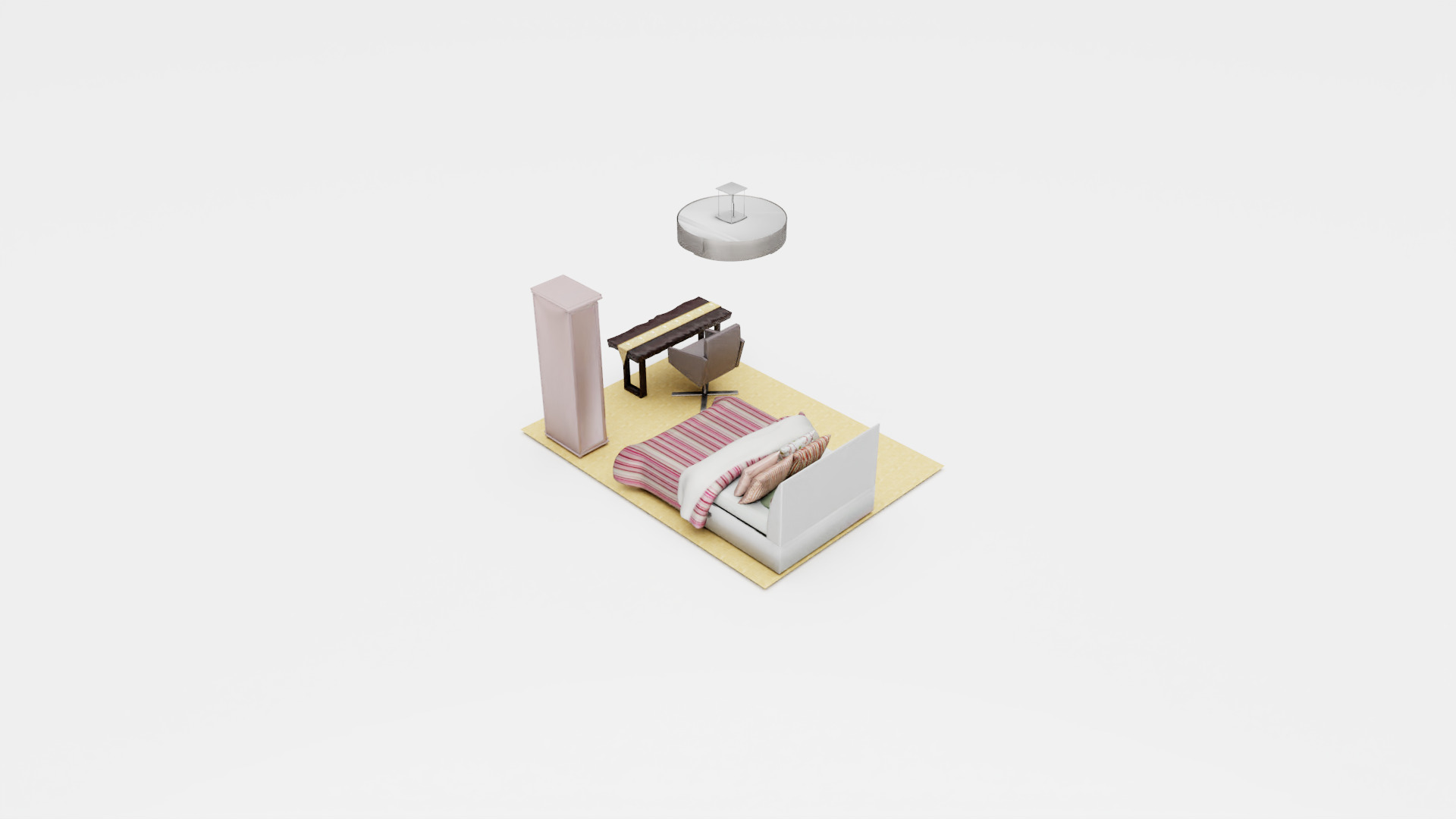}
    \end{subfigure}%
    \vskip\baselineskip%
    \vspace{-2.2em}
    \vskip\baselineskip%
    \begin{subfigure}[b]{0.20\linewidth}
		\centering
		\includegraphics[width=0.8\linewidth]{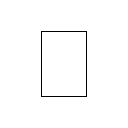}
    \end{subfigure}%
        \begin{subfigure}[b]{0.20\linewidth}
		\centering
		\includegraphics[width=\linewidth, trim=500 200 500 100, clip]{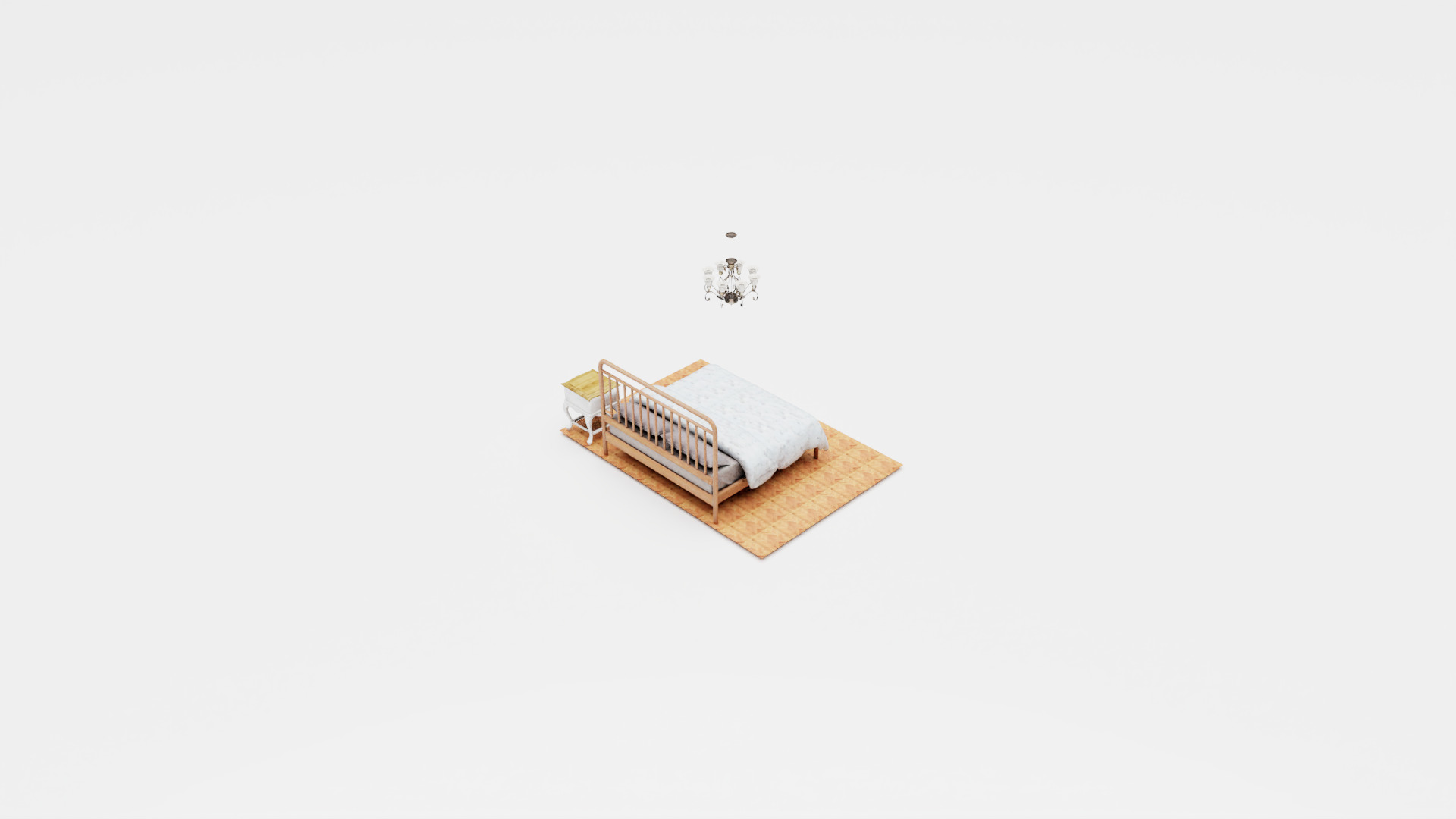}
    \end{subfigure}%
        \begin{subfigure}[b]{0.20\linewidth}
		\centering
		\includegraphics[width=\linewidth, trim=500 200 500 100, clip]{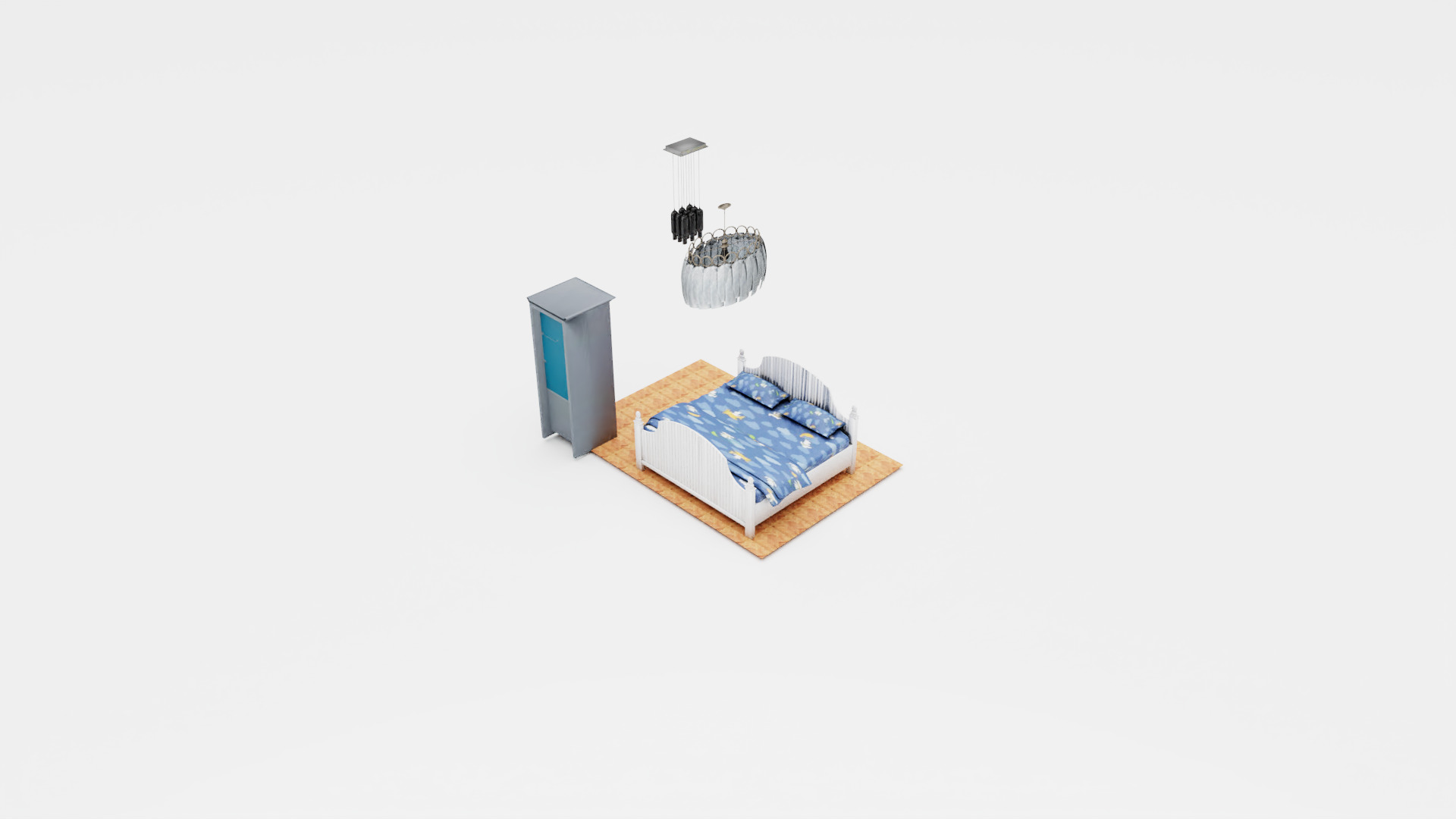}
    \end{subfigure}%
        \begin{subfigure}[b]{0.20\linewidth}
		\centering
		\includegraphics[width=\linewidth, trim=500 200 500 100, clip]{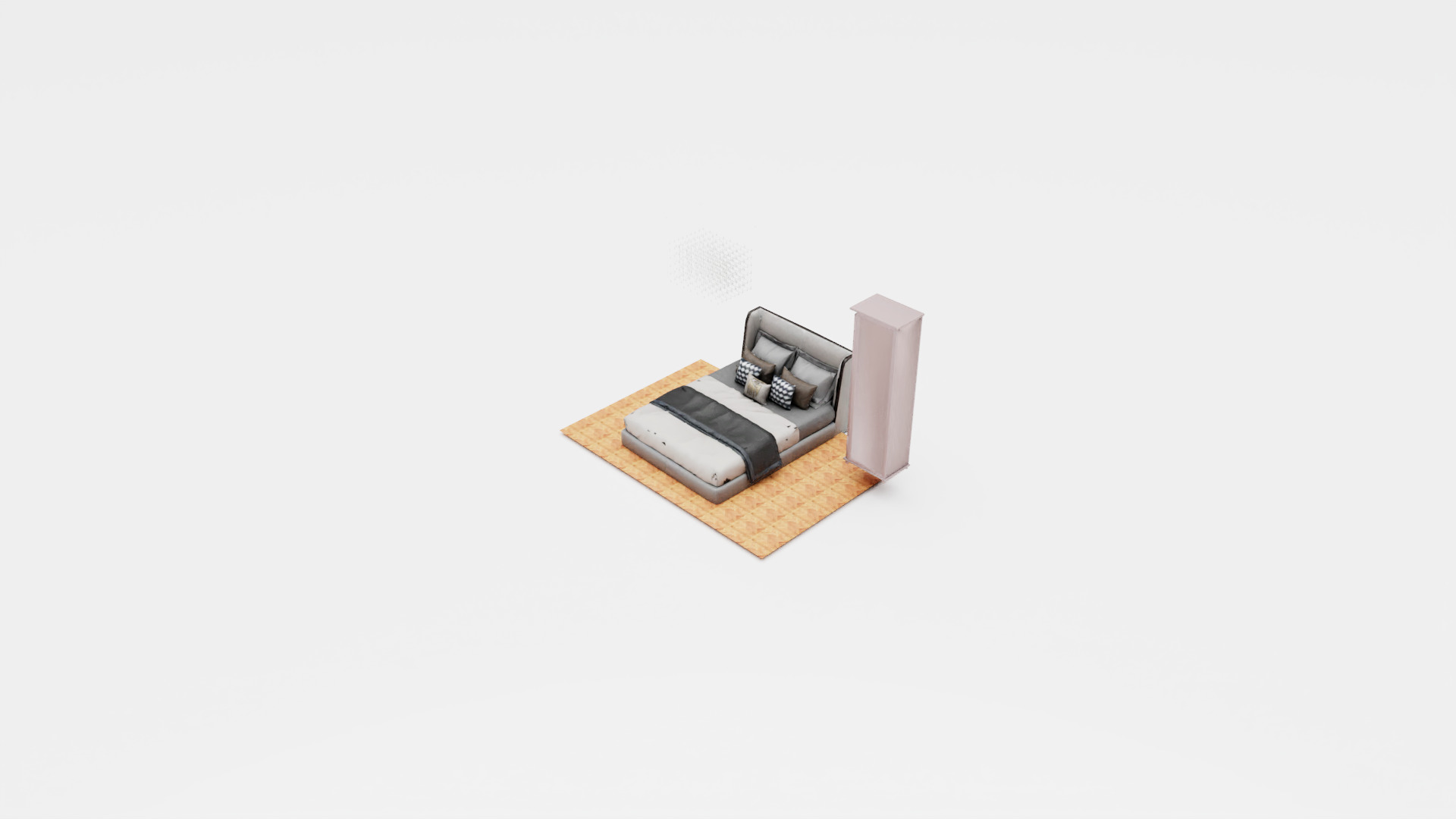}
    \end{subfigure}%
    \begin{subfigure}[b]{0.20\linewidth}
		\centering
		\includegraphics[width=\linewidth, trim=500 200 500 100, clip]{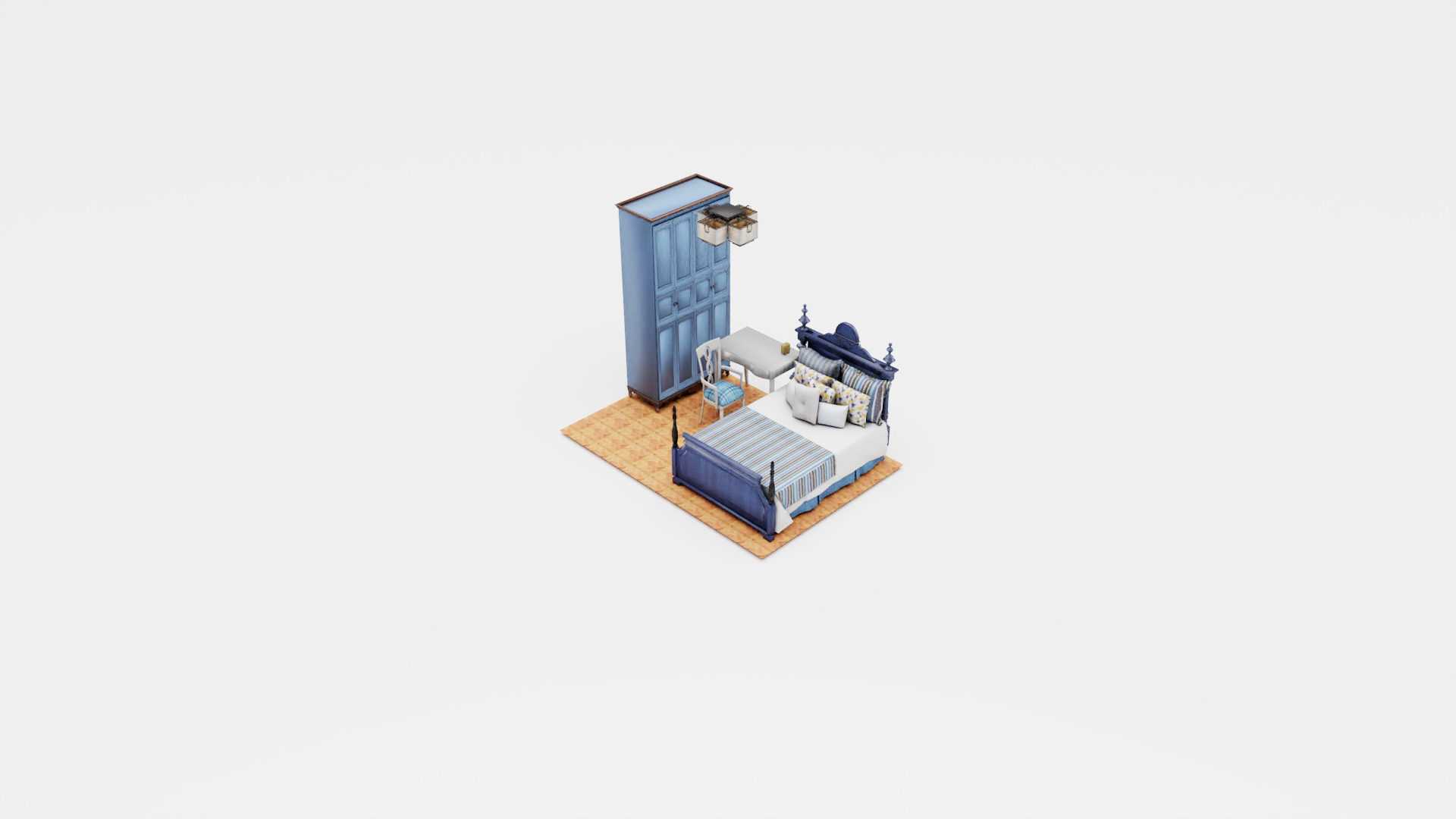}
    \end{subfigure}%
    \vskip\baselineskip%
    \vspace{-2.2em}
    \vskip\baselineskip%
    \begin{subfigure}[b]{0.20\linewidth}
		\centering
		\includegraphics[width=0.8\linewidth]{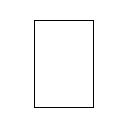}
    \end{subfigure}%
        \begin{subfigure}[b]{0.20\linewidth}
		\centering
		\includegraphics[width=\linewidth, trim=500 200 500 100, clip]{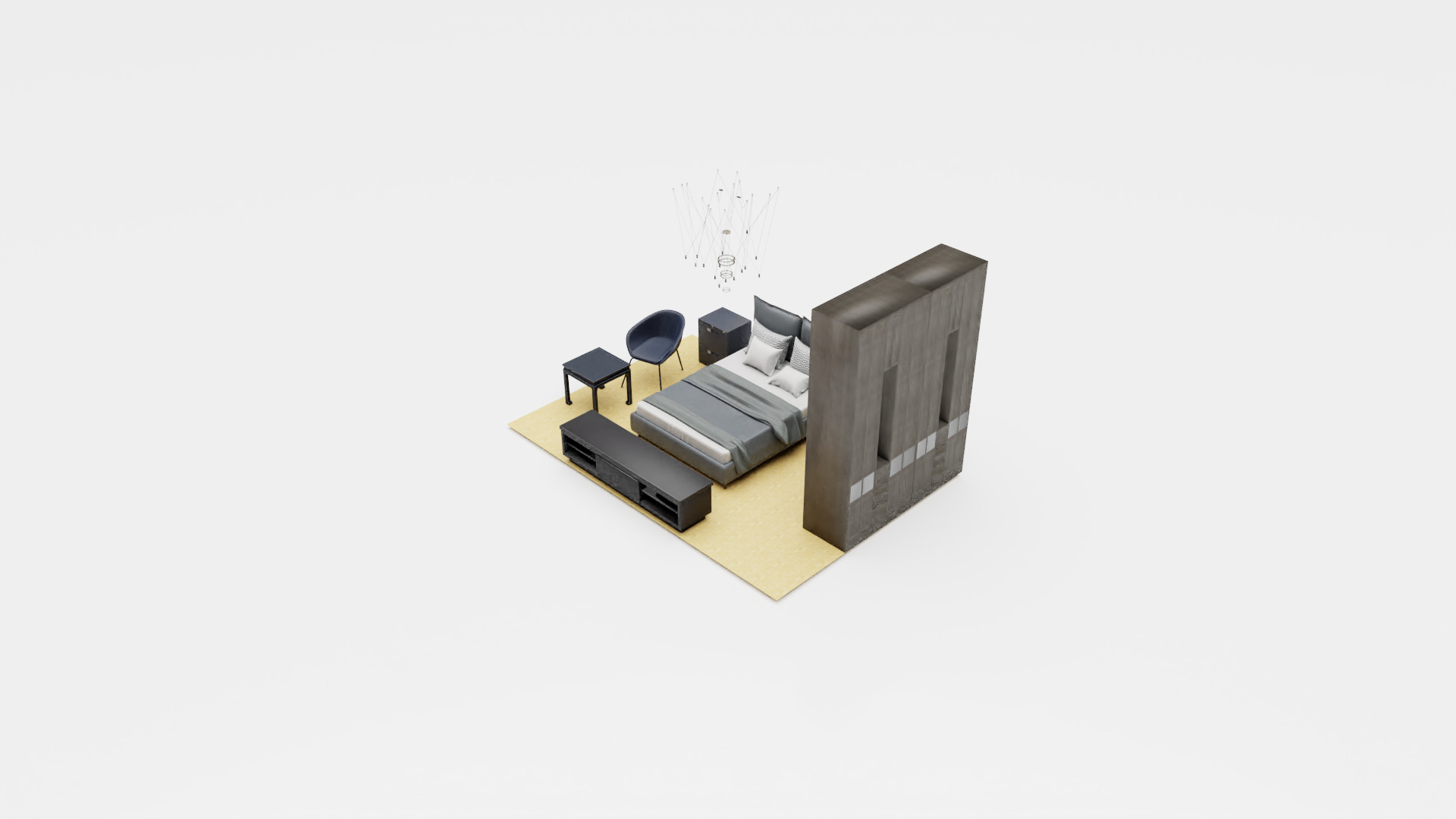}
    \end{subfigure}%
        \begin{subfigure}[b]{0.20\linewidth}
		\centering
		\includegraphics[width=\linewidth, trim=500 200 500 100, clip]{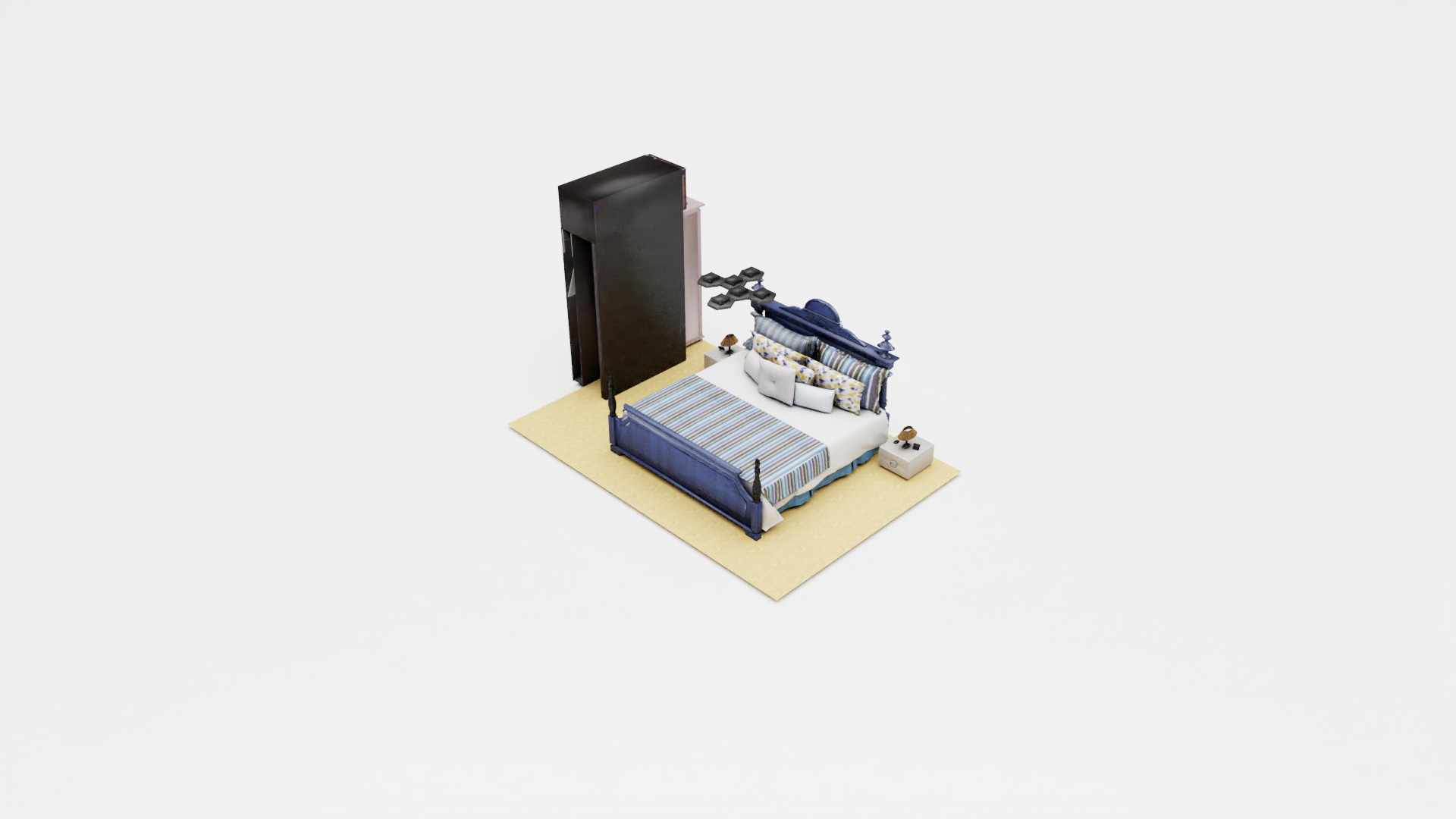}
    \end{subfigure}%
        \begin{subfigure}[b]{0.20\linewidth}
		\centering
		\includegraphics[width=\linewidth, trim=500 200 500 100, clip]{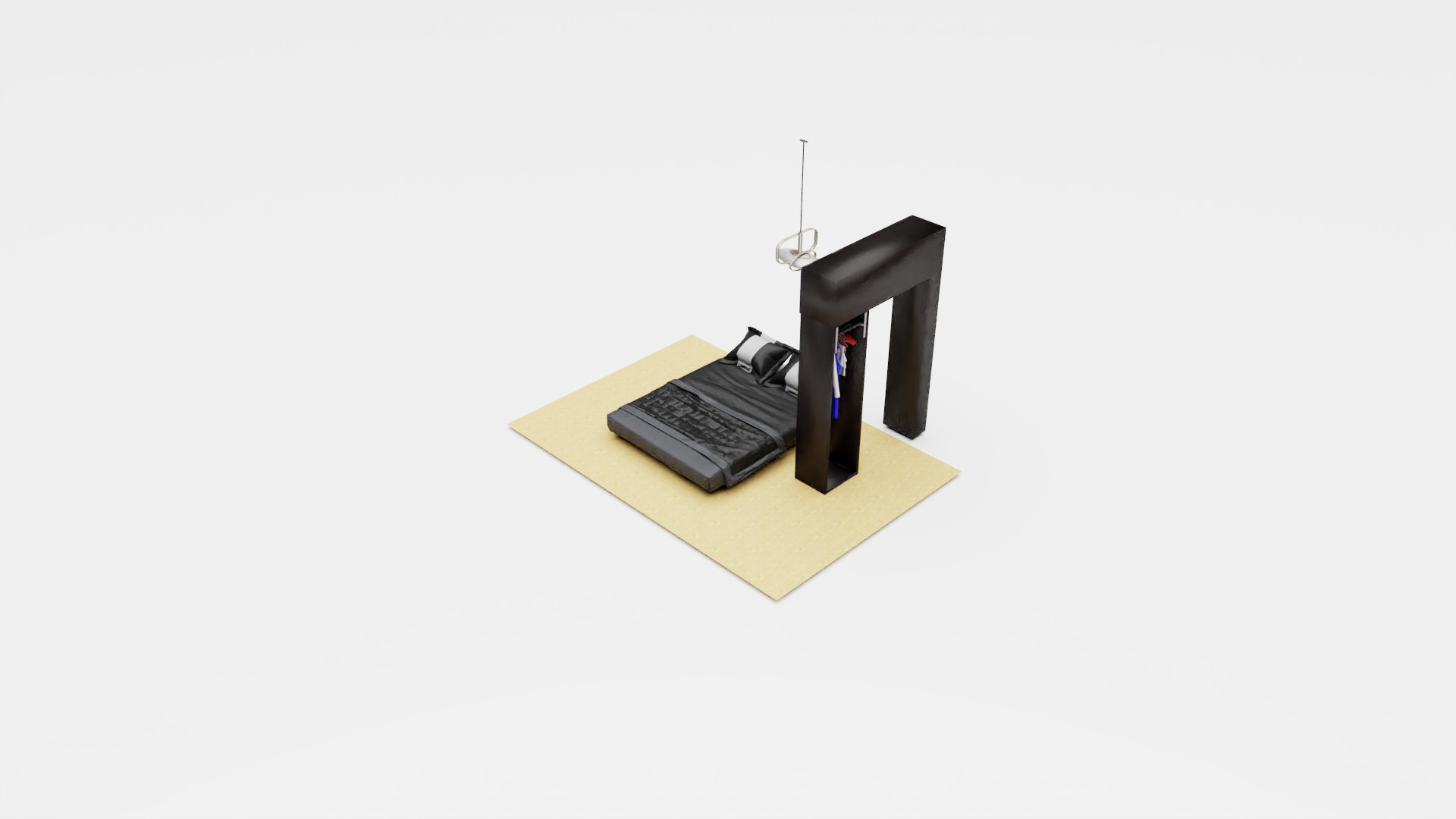}
    \end{subfigure}%
    \begin{subfigure}[b]{0.20\linewidth}
		\centering
		\includegraphics[width=\linewidth, trim=500 200 500 100, clip]{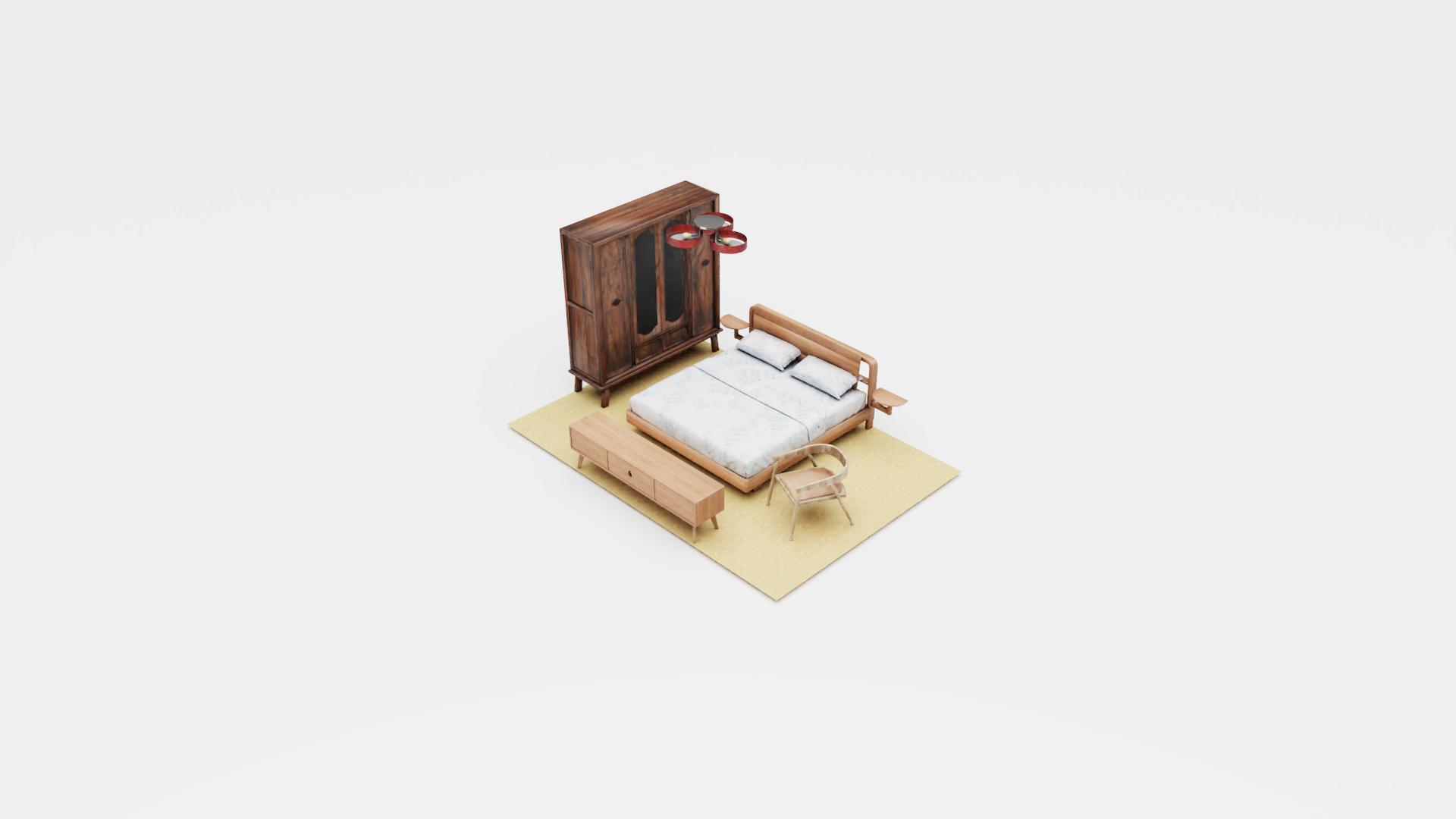}
    \end{subfigure}%
    \vskip\baselineskip%
    \vspace{-2.2em}
    \vskip\baselineskip%
    \begin{subfigure}[b]{0.20\linewidth}
		\centering
		\includegraphics[width=0.8\linewidth]{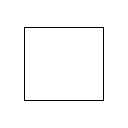}
    \end{subfigure}%
        \begin{subfigure}[b]{0.20\linewidth}
		\centering
		\includegraphics[width=\linewidth, trim=500 200 500 100, clip]{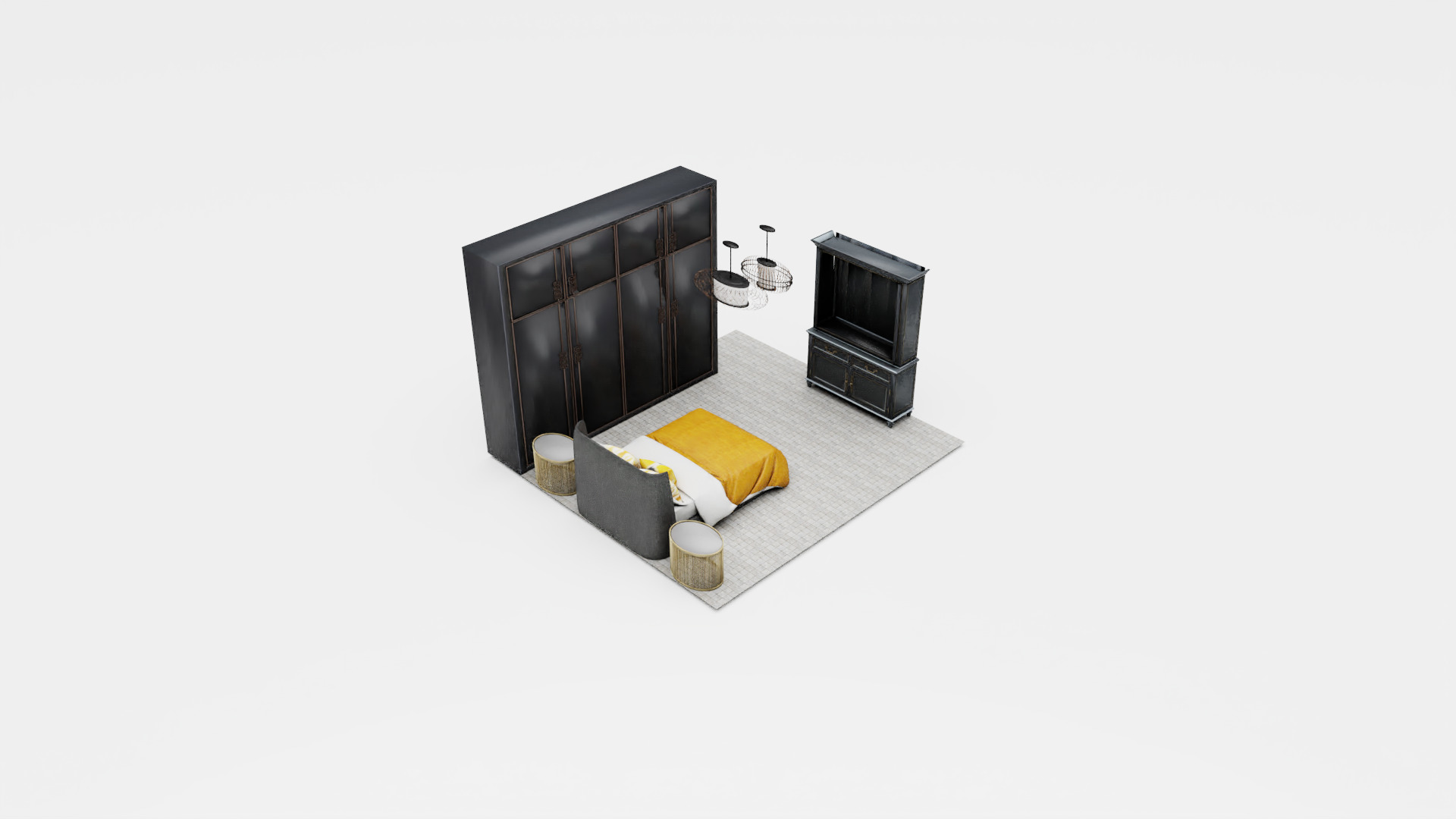}
    \end{subfigure}%
        \begin{subfigure}[b]{0.20\linewidth}
		\centering
		\includegraphics[width=\linewidth, trim=500 200 500 100, clip]{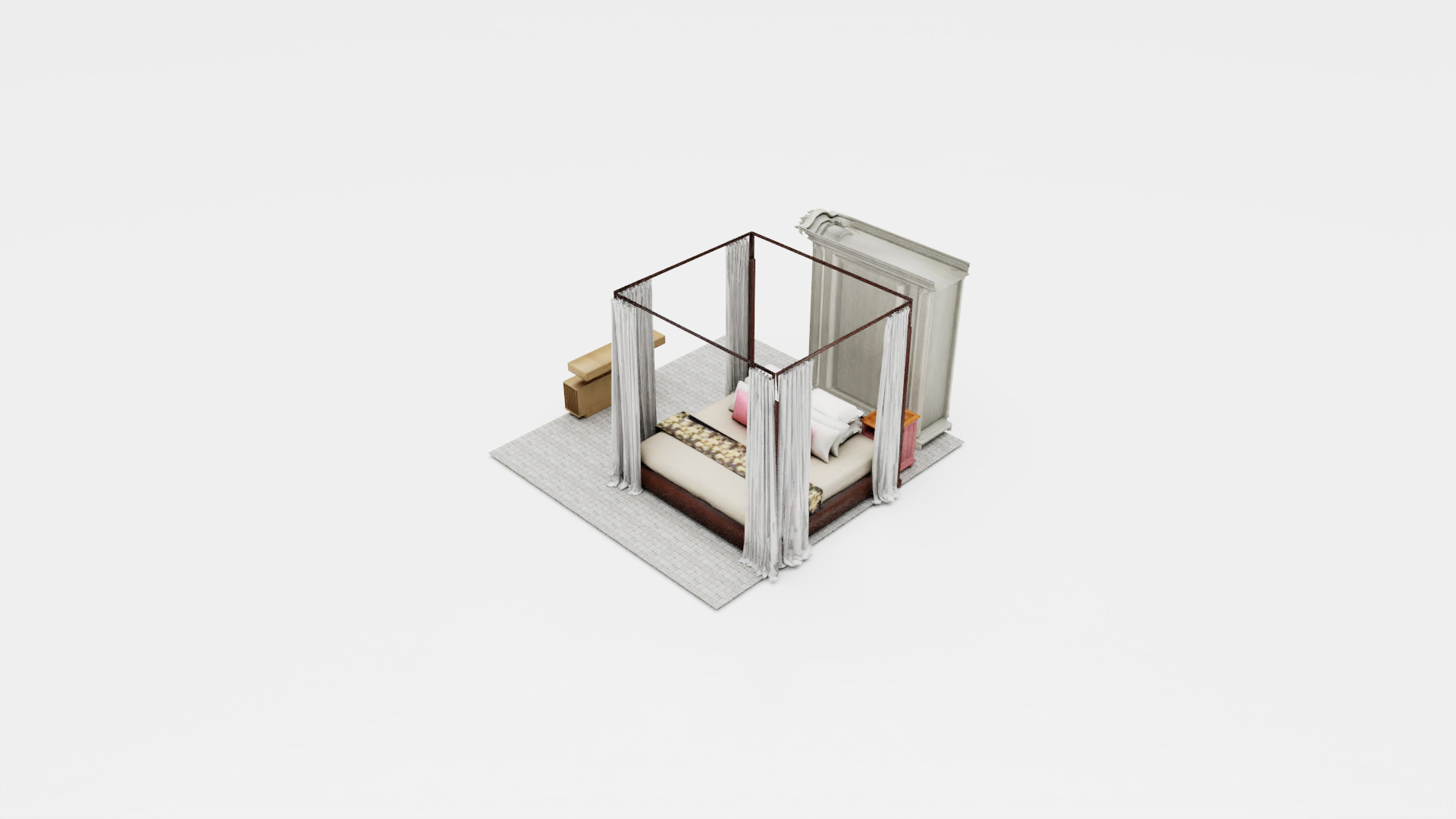}
    \end{subfigure}%
        \begin{subfigure}[b]{0.20\linewidth}
		\centering
		\includegraphics[width=\linewidth, trim=500 200 500 100, clip]{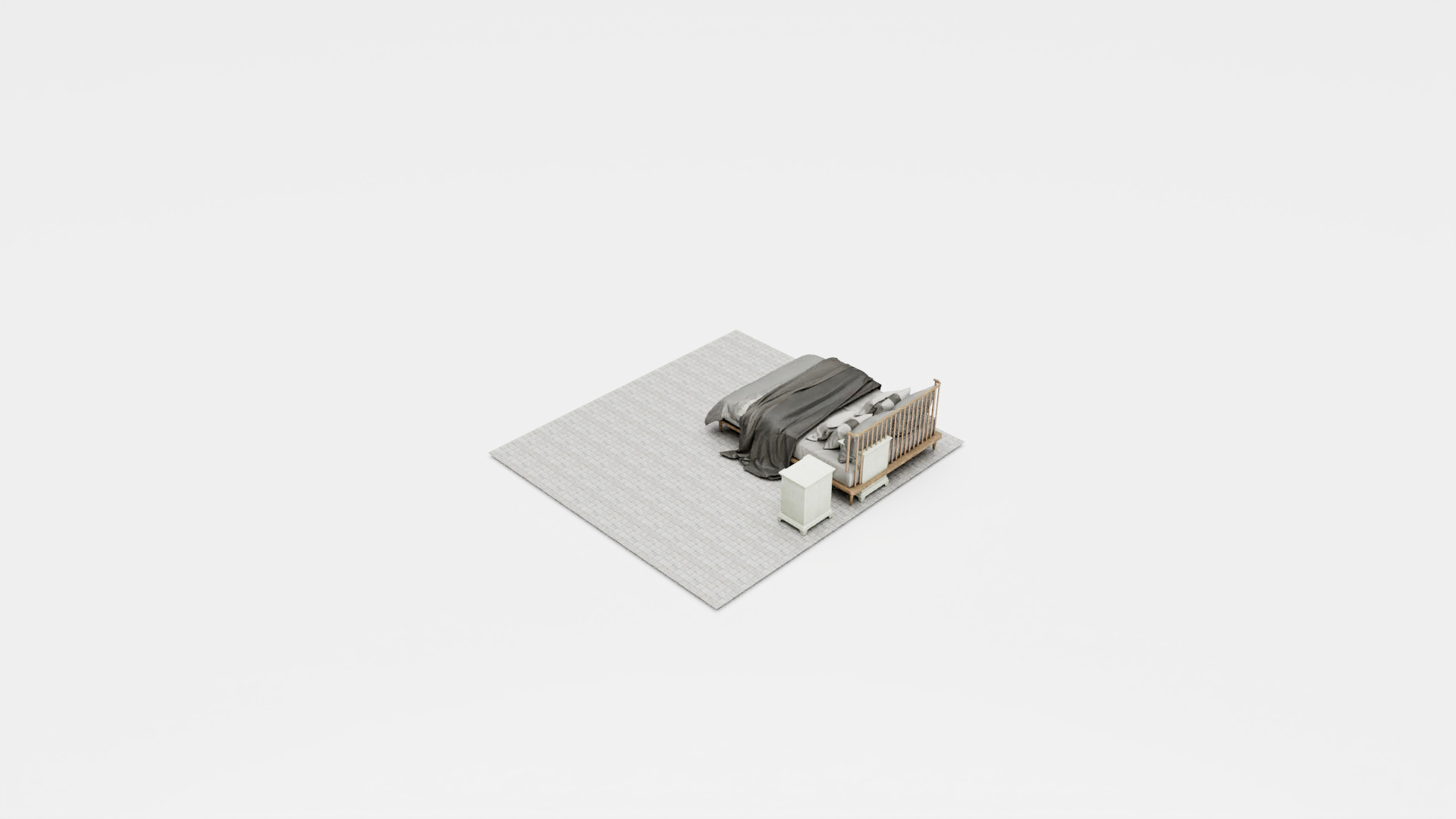}
    \end{subfigure}%
    \begin{subfigure}[b]{0.20\linewidth}
		\centering
		\includegraphics[width=\linewidth, trim=500 200 500 100, clip]{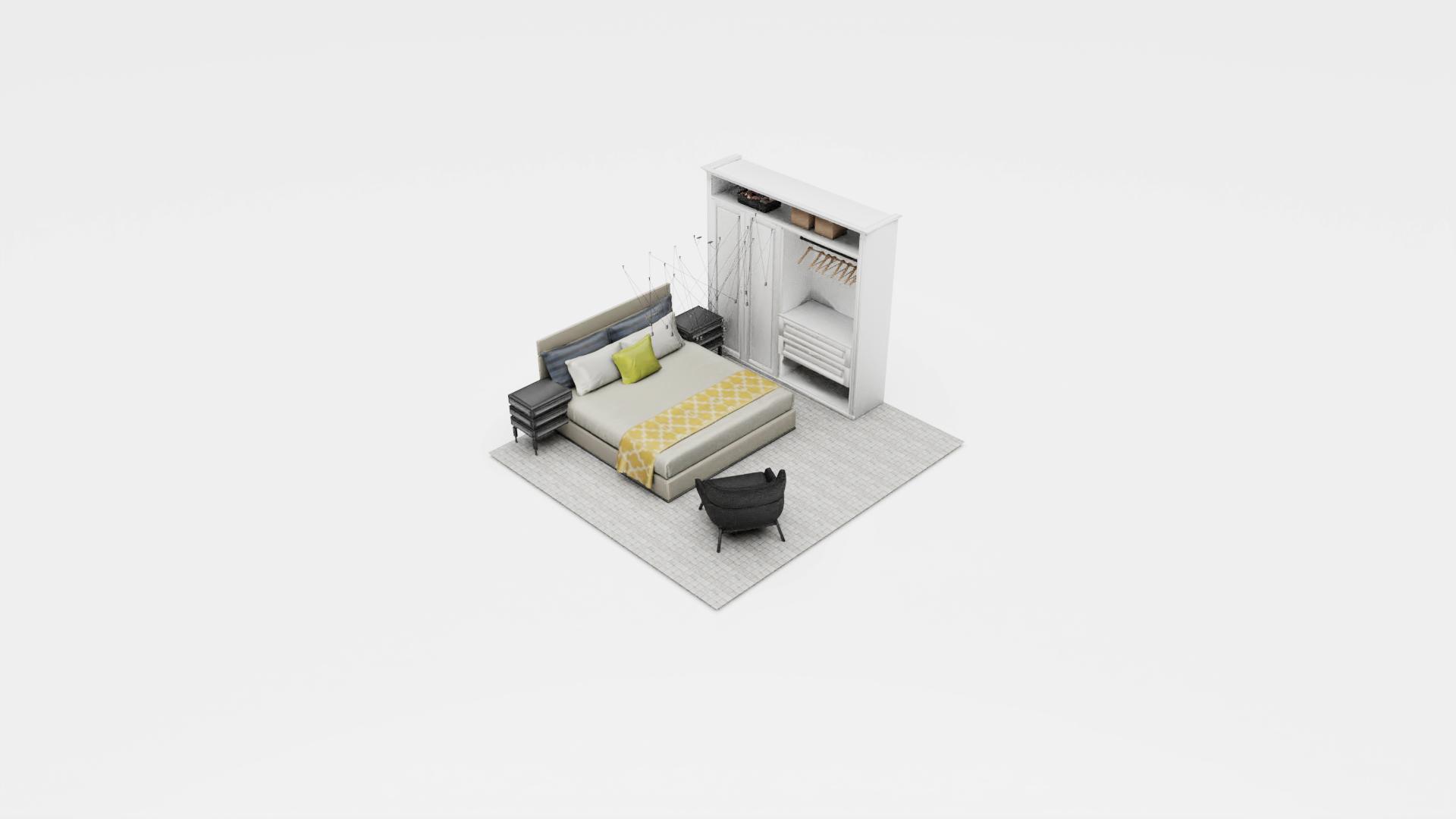}
    \end{subfigure}%
    \vspace{-1.2em}
    \vskip\baselineskip%
    \hfill%
    \caption{{\bf Qualitative Scene Synthesis Results on Bedrooms}.
    Generated scenes for bedrooms using FastSynth, SceneFormer and our method.
    To showcase the generalization abilities of our model we also show the
    closest scene from the training set (2nd column).}
    \label{fig:scene_synthesis_qualitative_bedroom_supp}
    \vspace{-1.2em}
\end{figure}

\begin{figure}[!h]
    \centering
    \begin{subfigure}[b]{0.20\linewidth}
		\centering
        \small Scene Layout
    \end{subfigure}%
    \begin{subfigure}[b]{0.20\linewidth}
		\centering
        \small Training Sample
    \end{subfigure}%
    \begin{subfigure}[b]{0.20\linewidth}
		\centering
        \small FastSynth
    \end{subfigure}%
    \begin{subfigure}[b]{0.20\linewidth}
		\centering
        \small SceneFormer
    \end{subfigure}%
    \begin{subfigure}[b]{0.20\linewidth}
        \centering
        \small Ours
    \end{subfigure}
    \hfill%
    \vskip\baselineskip%
    \vspace{-1.5em}
    \hfill%
    \begin{subfigure}[b]{0.20\linewidth}
		\centering
		\includegraphics[width=0.8\linewidth, trim=0 20 0 20, clip]{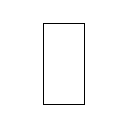}
    \end{subfigure}%
        \begin{subfigure}[b]{0.20\linewidth}
		\centering
		\includegraphics[width=\linewidth, trim=300 50 300 100, clip]{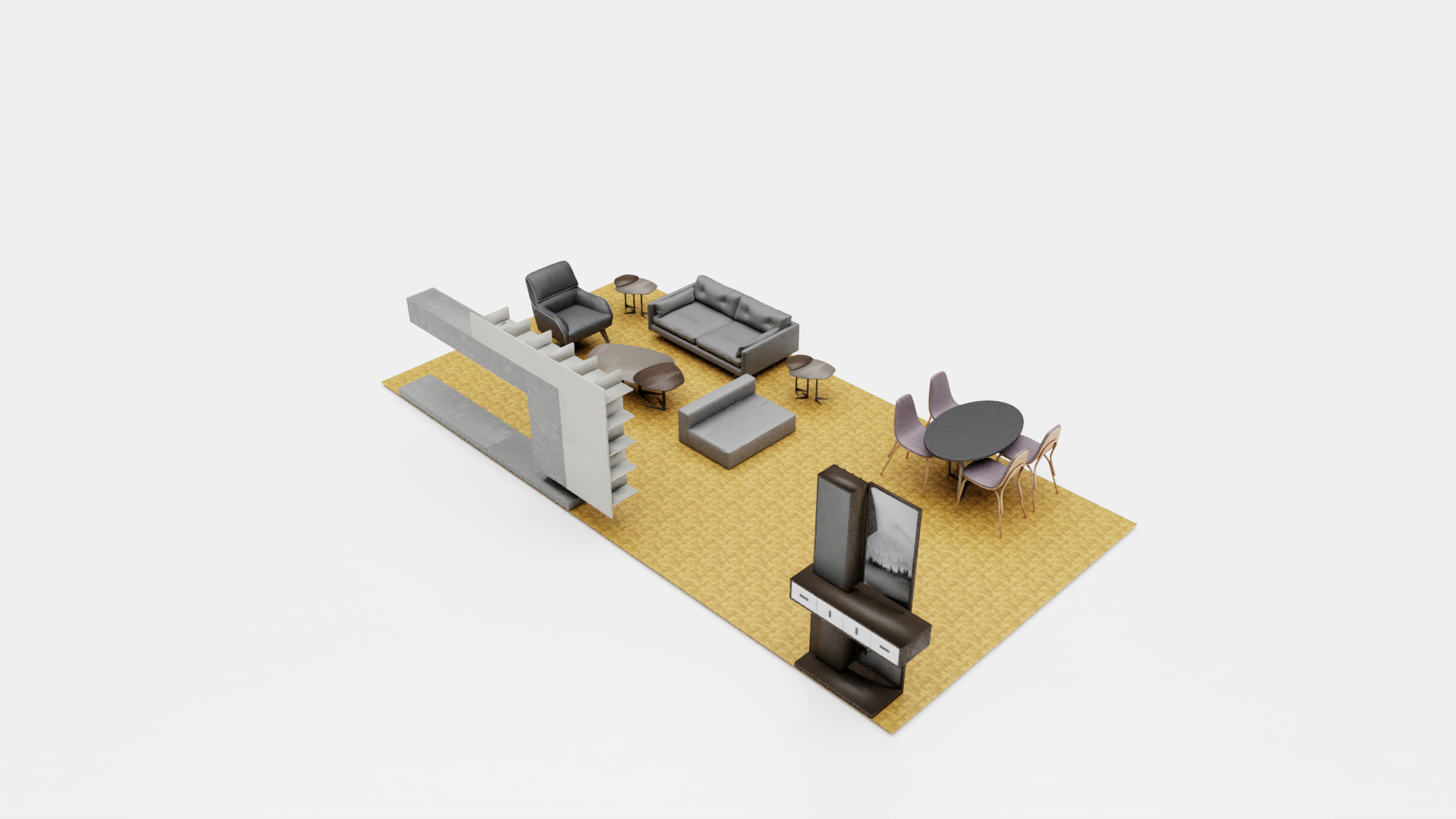}
    \end{subfigure}%
        \begin{subfigure}[b]{0.20\linewidth}
		\centering
		\includegraphics[width=\linewidth, trim=300 50 300 100, clip]{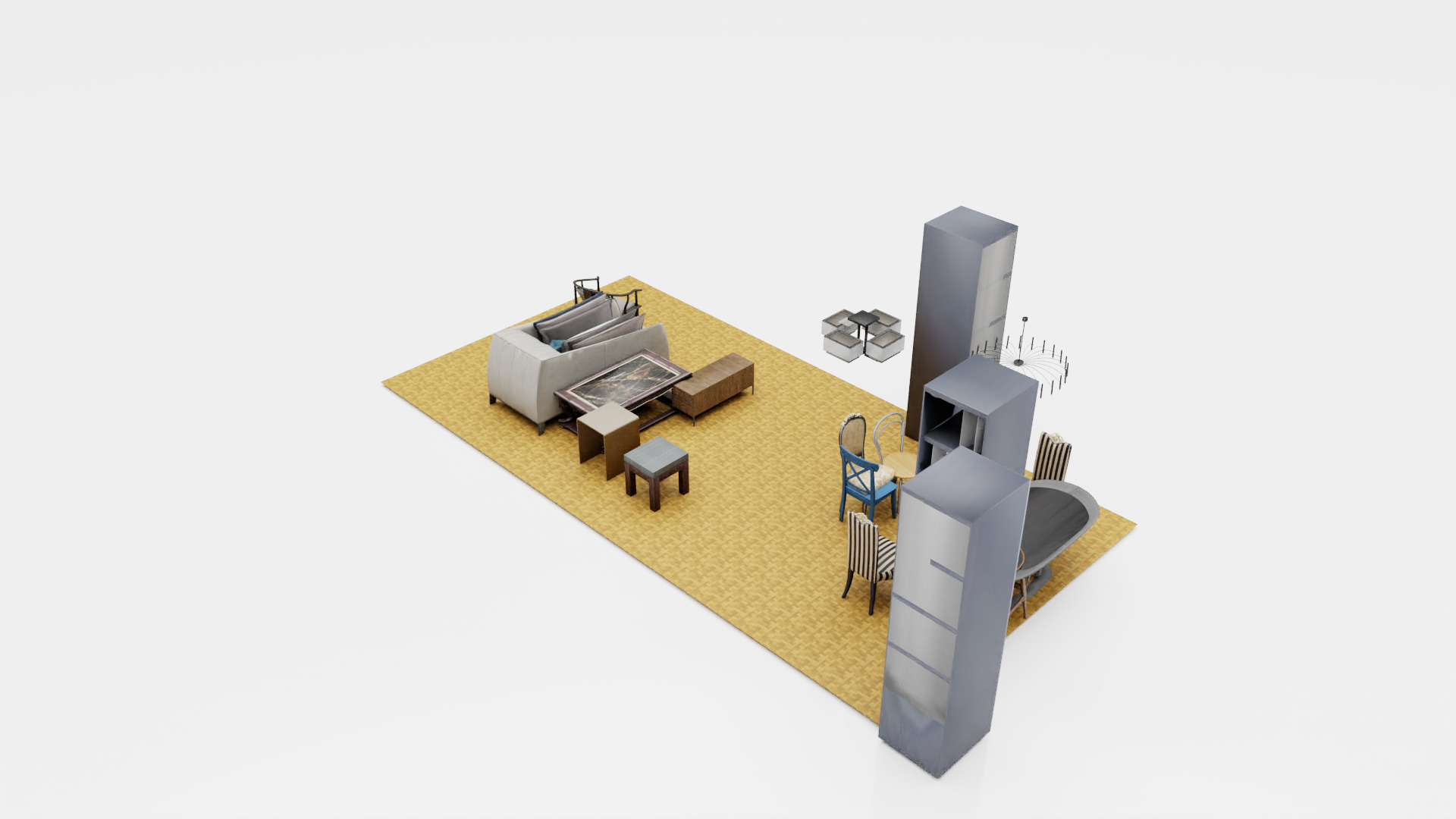}
    \end{subfigure}%
        \begin{subfigure}[b]{0.20\linewidth}
		\centering
		\includegraphics[width=\linewidth, trim=300 50 300 100, clip]{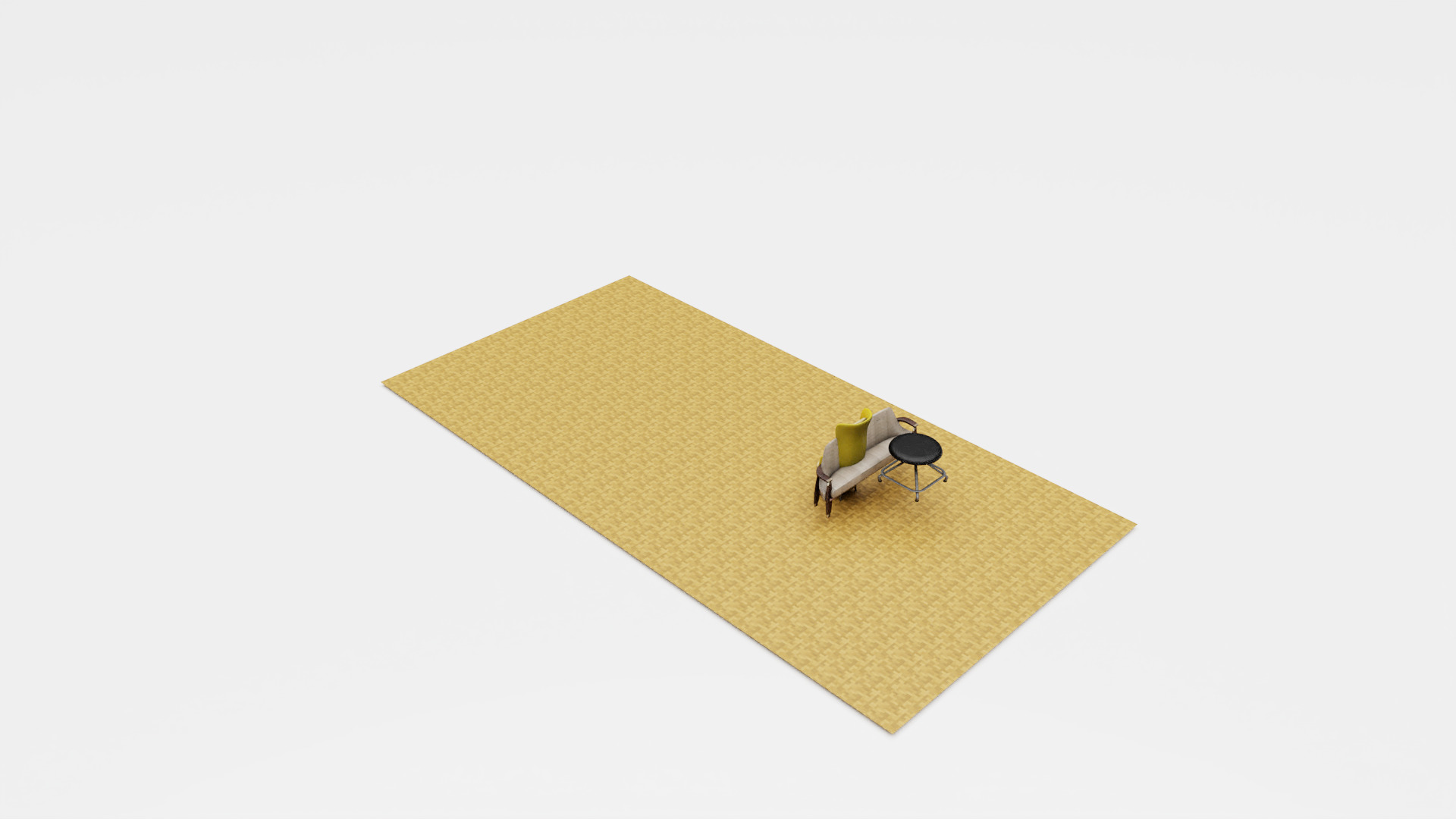}
    \end{subfigure}%
    \begin{subfigure}[b]{0.20\linewidth}
		\centering
		\includegraphics[width=\linewidth, trim=300 50 300 100, clip]{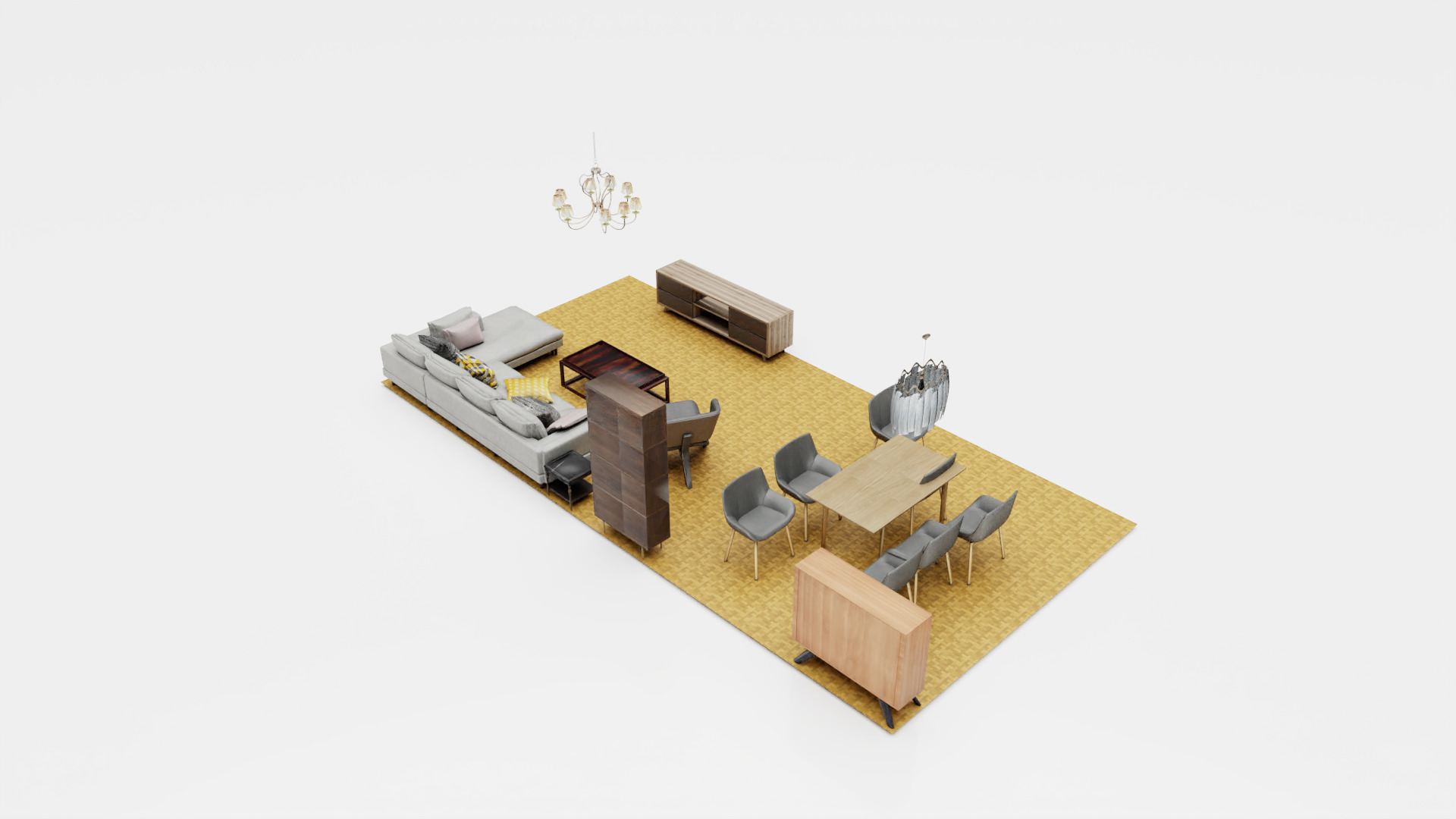}
    \end{subfigure}%
    \vskip\baselineskip%
    \vspace{-2.2em}
    \vskip\baselineskip%
    \begin{subfigure}[b]{0.20\linewidth}
		\centering
		\includegraphics[width=0.8\linewidth, trim=0 20 0 20, clip]{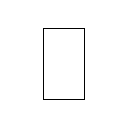}
    \end{subfigure}%
        \begin{subfigure}[b]{0.20\linewidth}
		\centering
		\includegraphics[width=\linewidth, trim=300 50 300 100, clip]{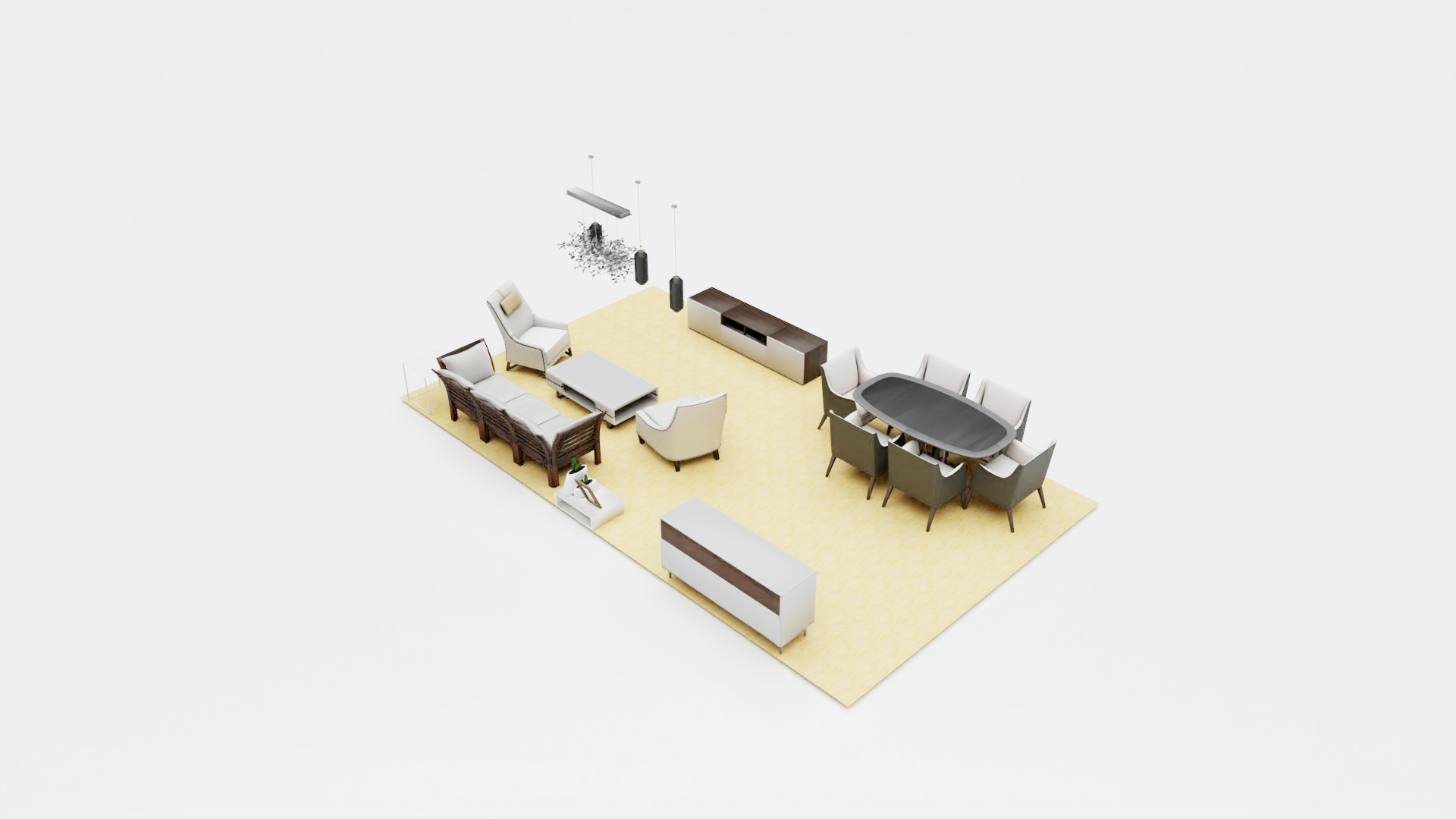}
    \end{subfigure}%
        \begin{subfigure}[b]{0.20\linewidth}
		\centering
		\includegraphics[width=\linewidth, trim=300 50 300 100, clip]{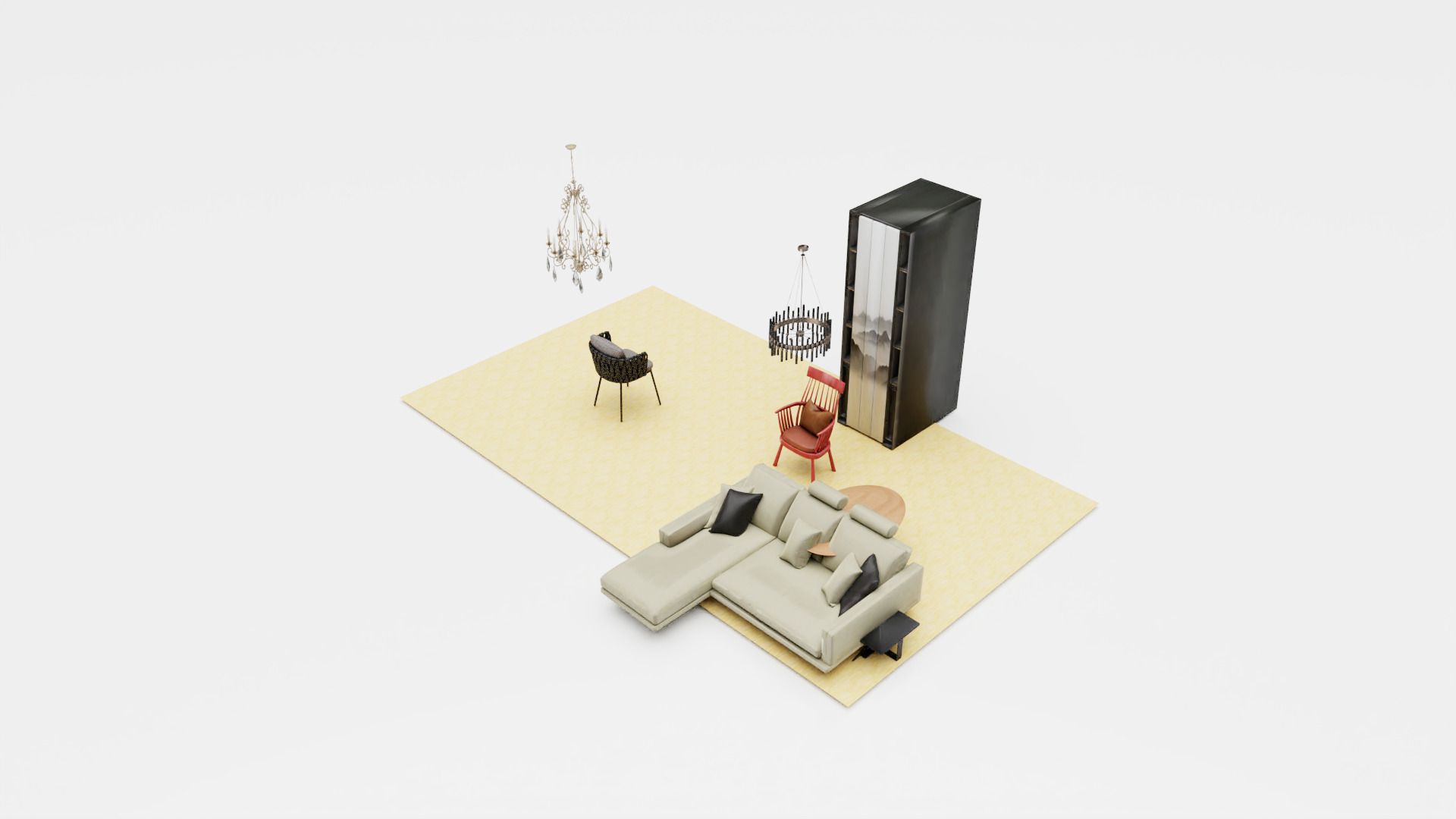}
    \end{subfigure}%
        \begin{subfigure}[b]{0.20\linewidth}
		\centering
		\includegraphics[width=\linewidth, trim=300 50 300 100, clip]{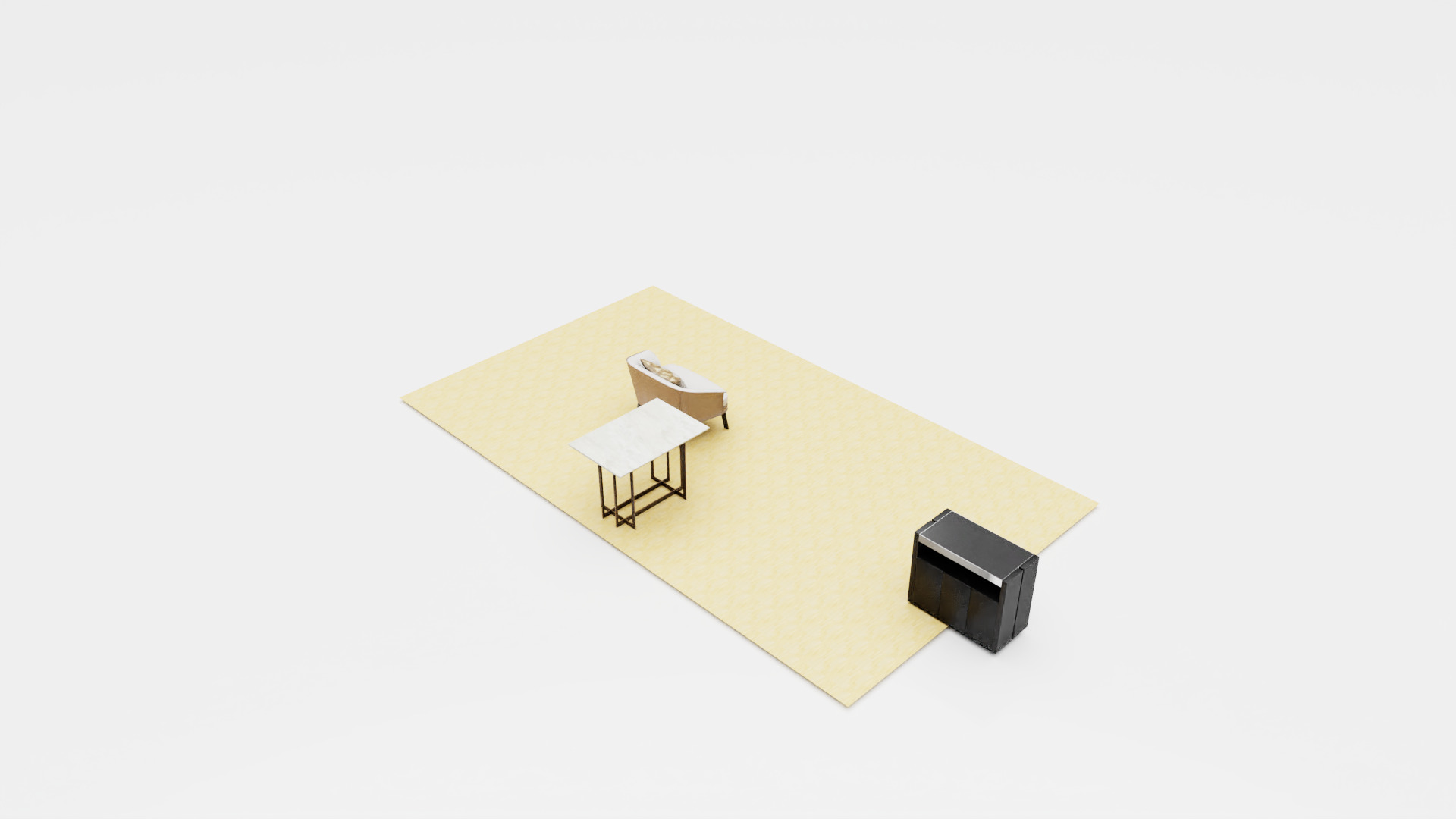}
    \end{subfigure}%
    \begin{subfigure}[b]{0.20\linewidth}
		\centering
		\includegraphics[width=\linewidth, trim=300 50 300 100, clip]{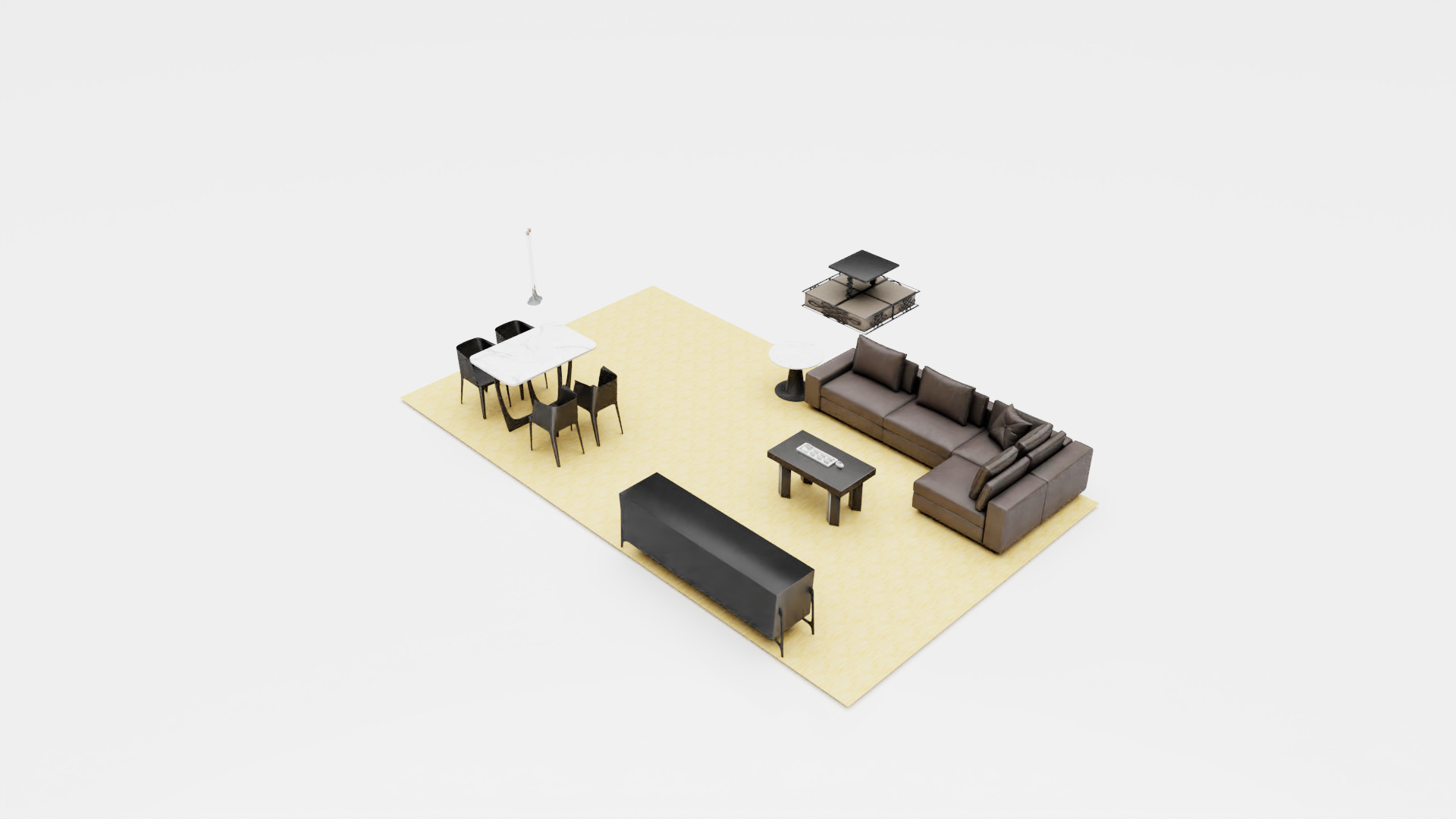}
    \end{subfigure}%
    \vskip\baselineskip%
    \vspace{-2.2em}
    \vskip\baselineskip%
    \begin{subfigure}[b]{0.20\linewidth}
		\centering
		\includegraphics[width=0.8\linewidth, trim=0 20 0 20, clip]{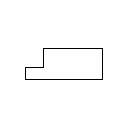}
    \end{subfigure}%
        \begin{subfigure}[b]{0.20\linewidth}
		\centering
		\includegraphics[width=\linewidth, trim=300 50 300 100, clip]{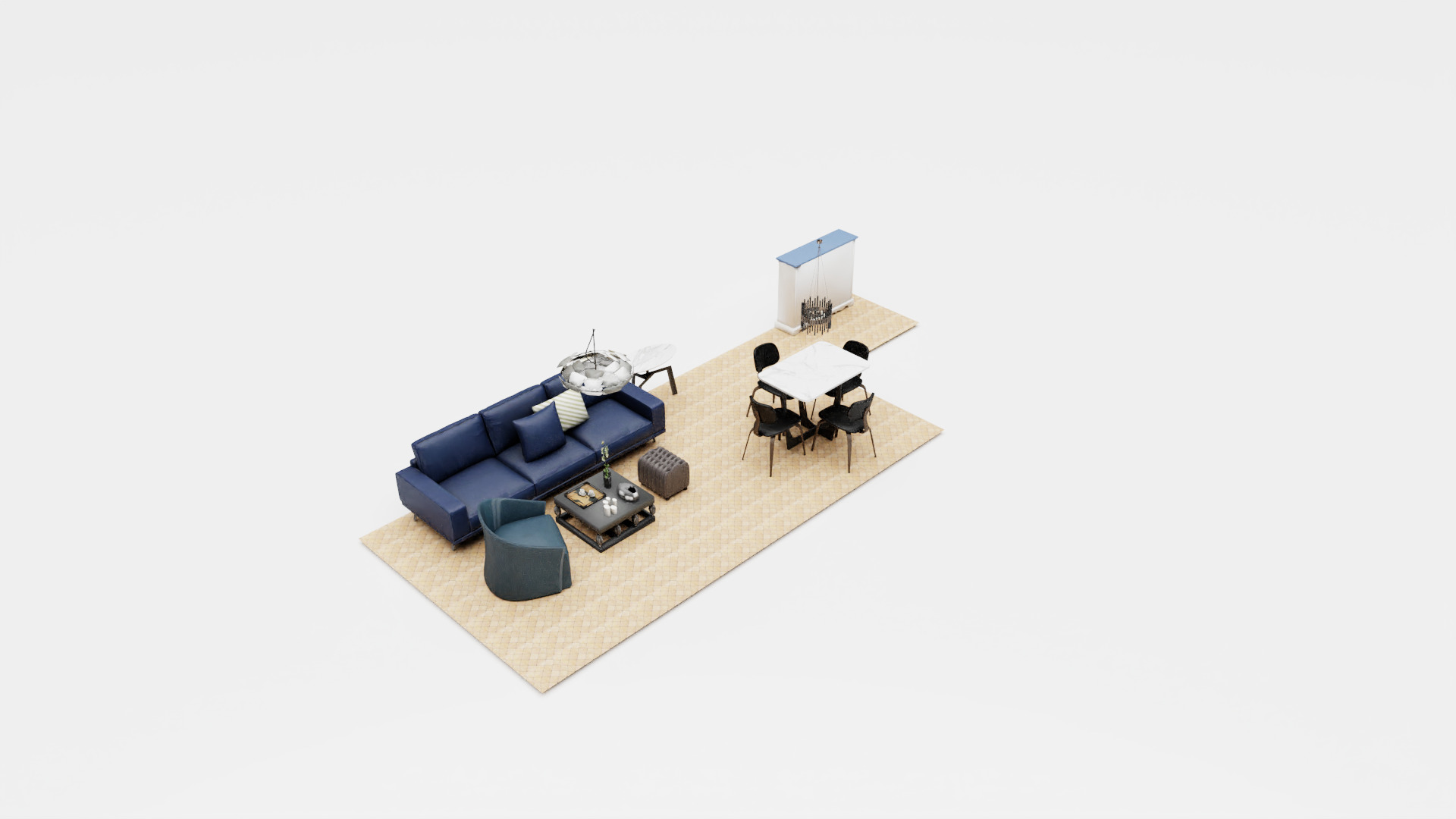}
    \end{subfigure}%
        \begin{subfigure}[b]{0.20\linewidth}
		\centering
		\includegraphics[width=\linewidth, trim=300 50 300 100, clip]{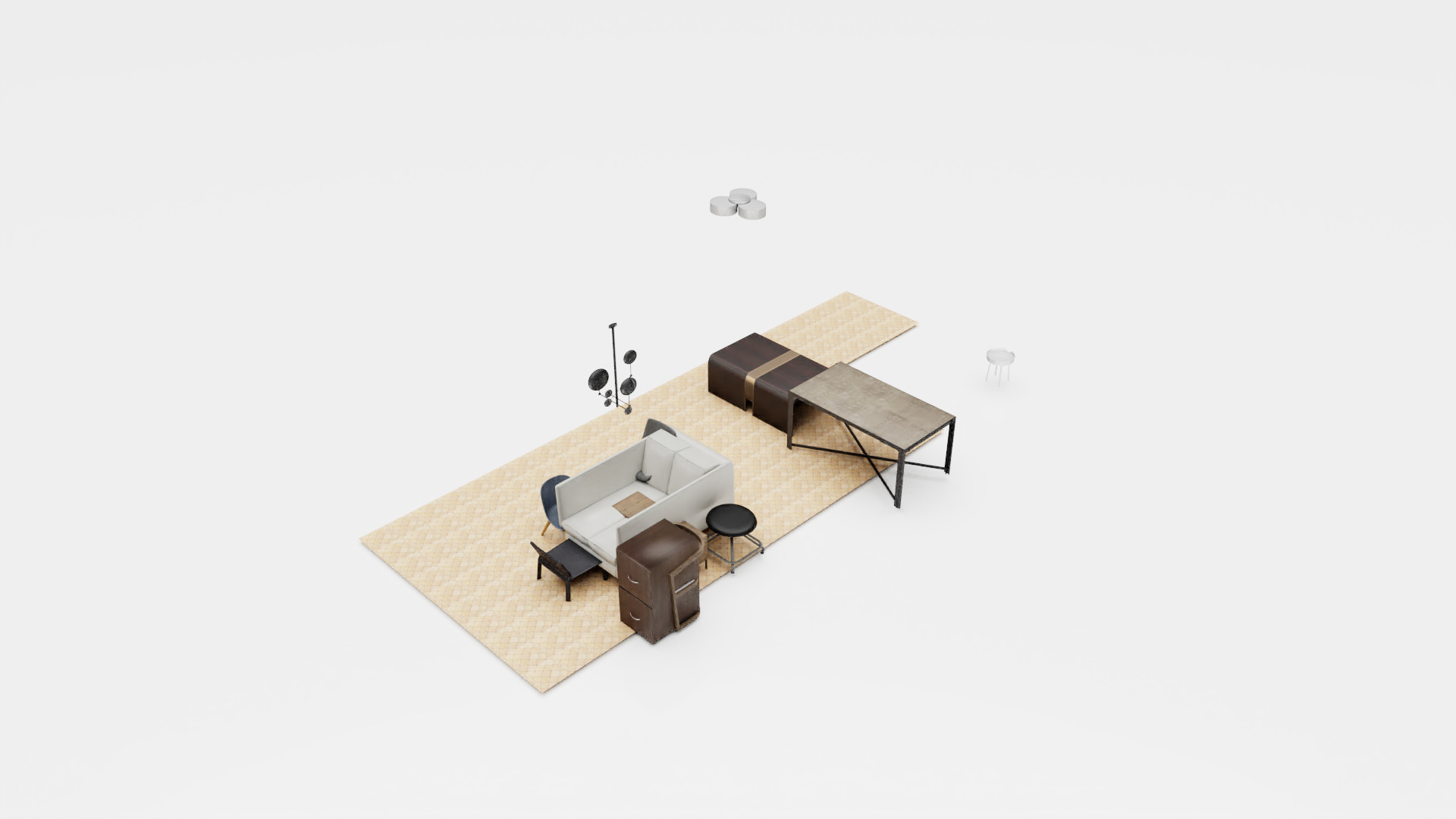}
    \end{subfigure}%
        \begin{subfigure}[b]{0.20\linewidth}
		\centering
		\includegraphics[width=\linewidth, trim=300 50 300 100, clip]{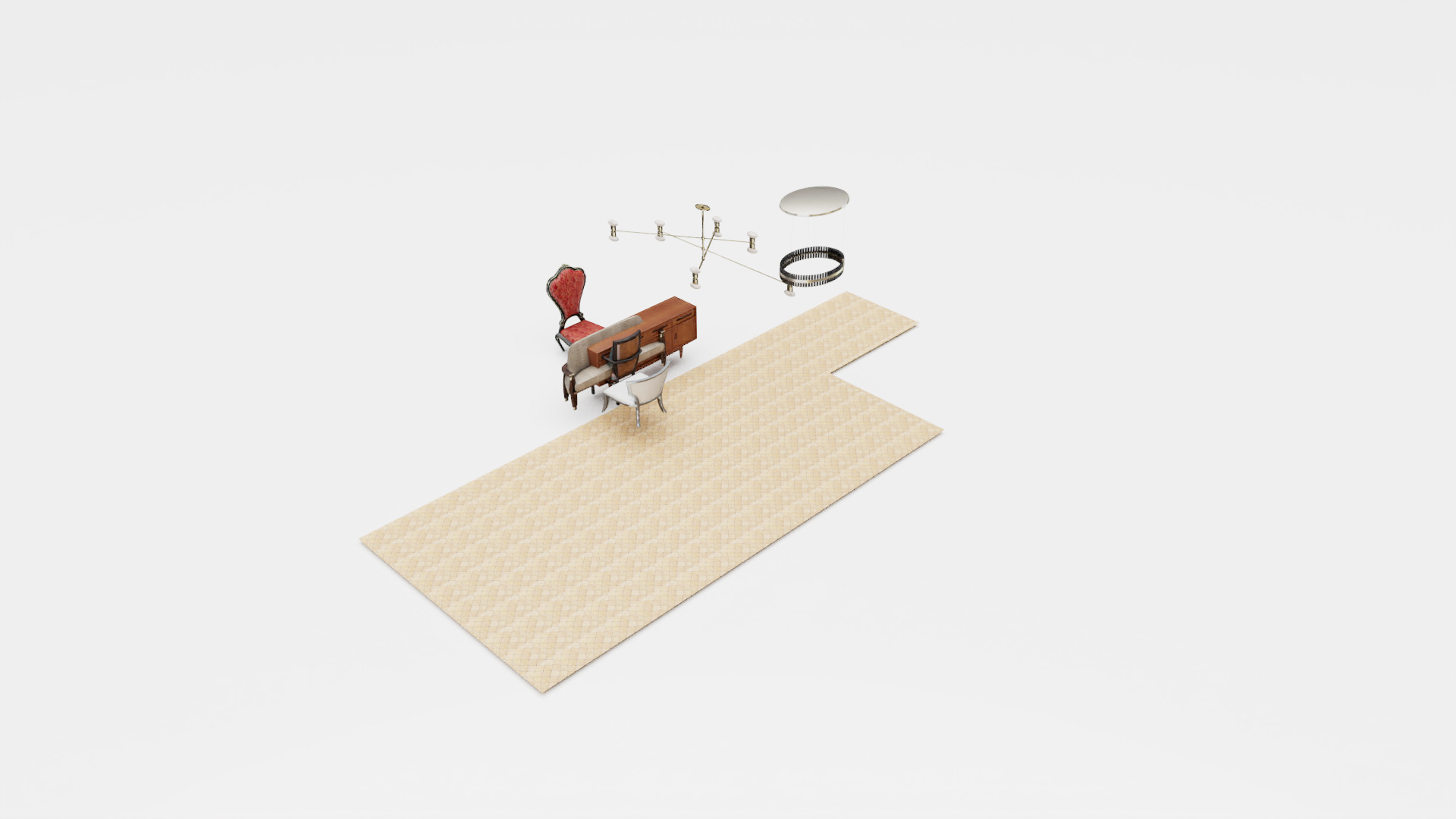}
    \end{subfigure}%
    \begin{subfigure}[b]{0.20\linewidth}
		\centering
		\includegraphics[width=\linewidth, trim=300 50 300 100, clip]{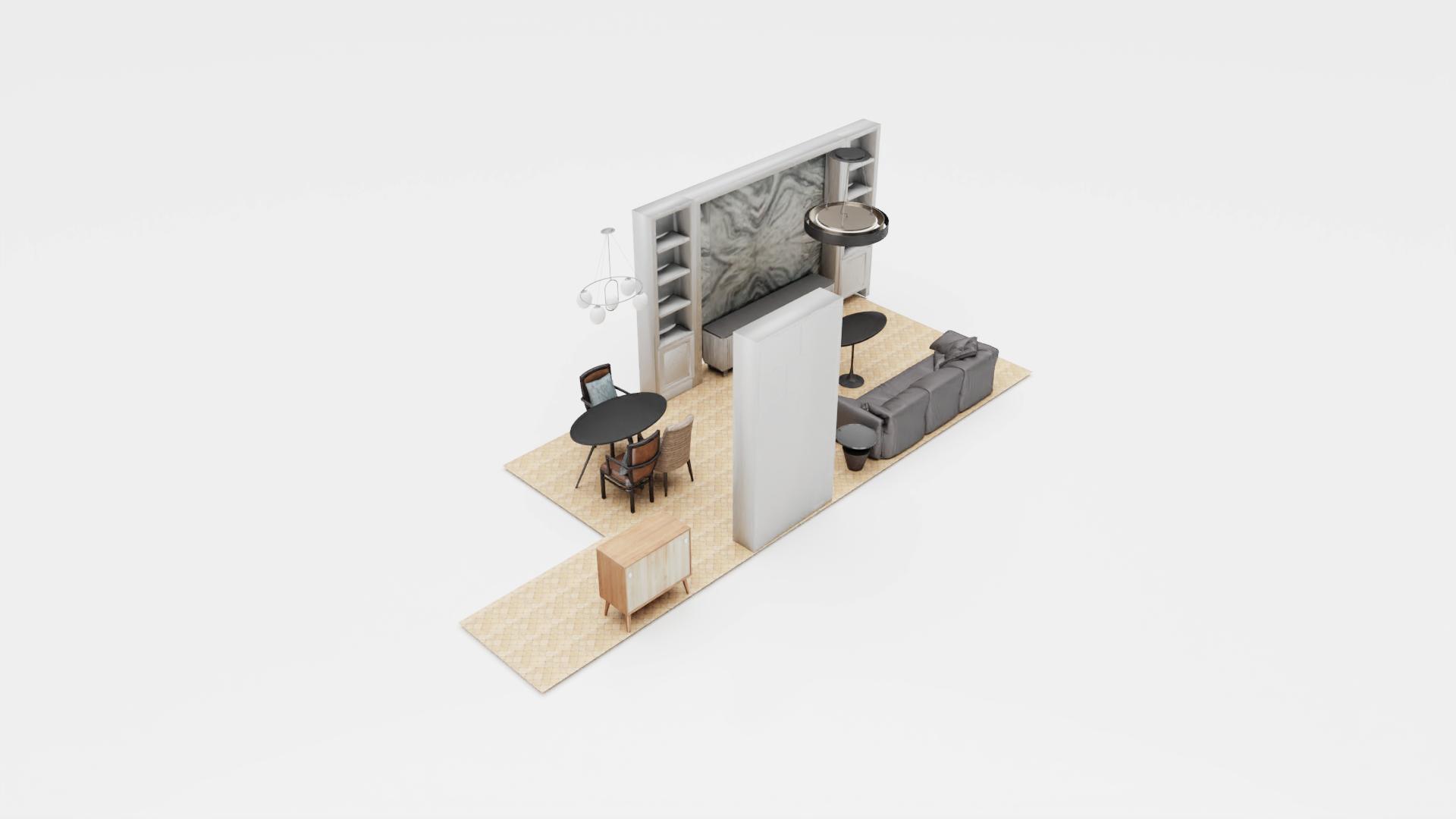}
    \end{subfigure}%
    \vskip\baselineskip%
    \vspace{-2.2em}
    \vskip\baselineskip%
    \begin{subfigure}[b]{0.20\linewidth}
		\centering
		\includegraphics[width=0.8\linewidth, trim=0 20 0 20, clip]{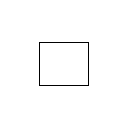}
    \end{subfigure}%
        \begin{subfigure}[b]{0.20\linewidth}
		\centering
		\includegraphics[width=\linewidth, trim=300 50 300 100, clip]{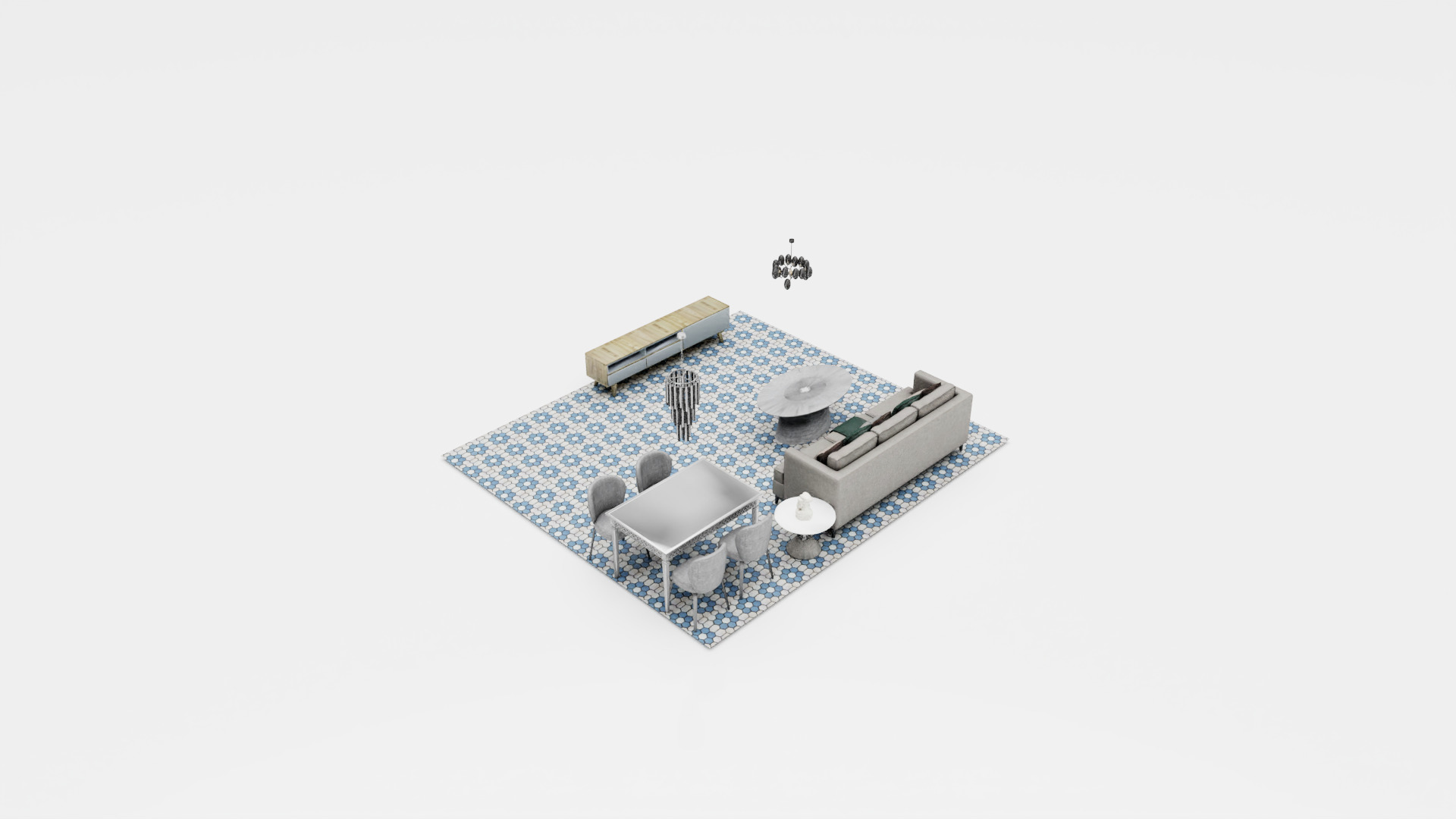}
    \end{subfigure}%
        \begin{subfigure}[b]{0.20\linewidth}
		\centering
		\includegraphics[width=\linewidth, trim=300 50 300 100, clip]{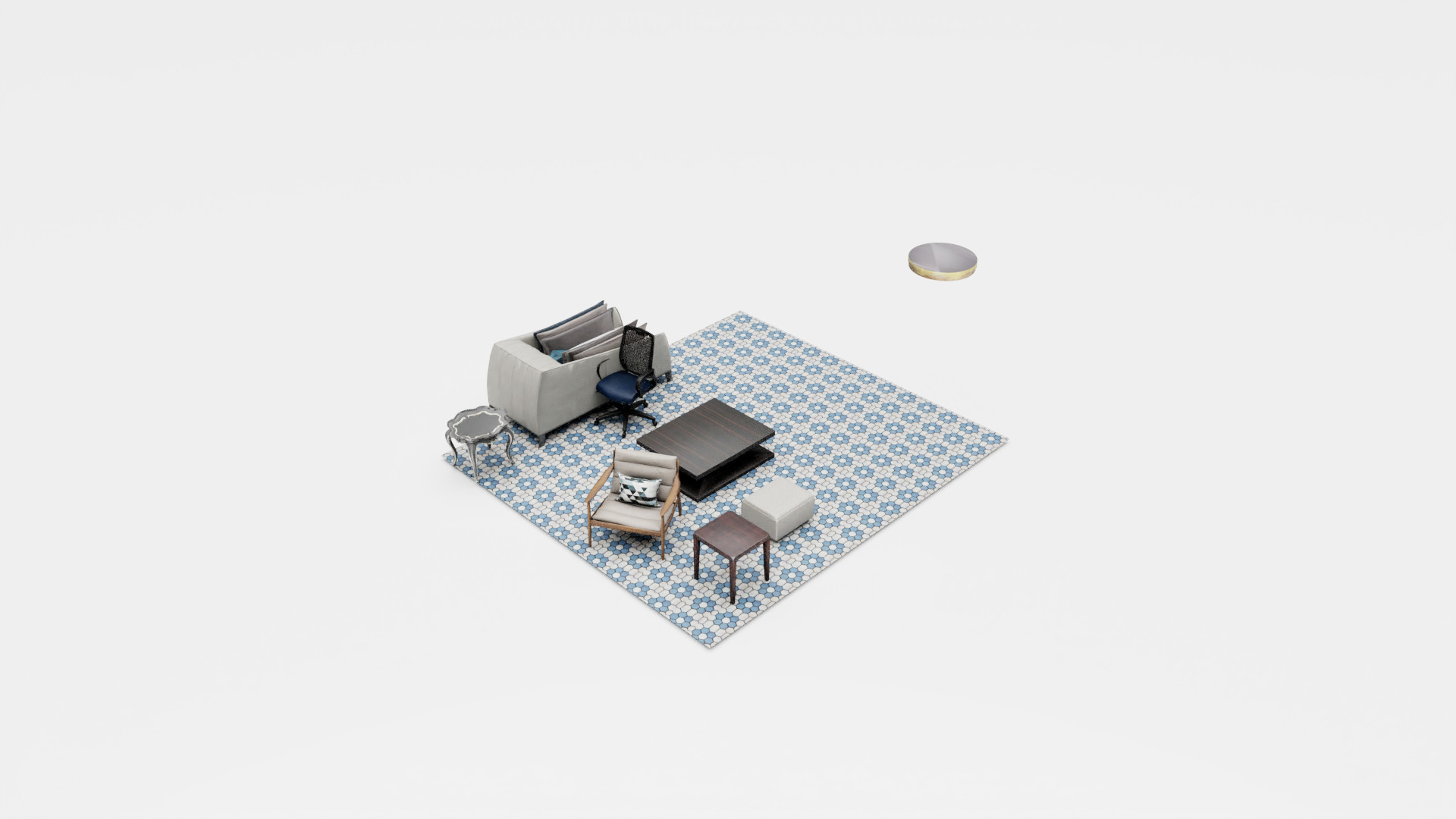}
    \end{subfigure}%
        \begin{subfigure}[b]{0.20\linewidth}
		\centering
		\includegraphics[width=\linewidth, trim=300 50 300 100, clip]{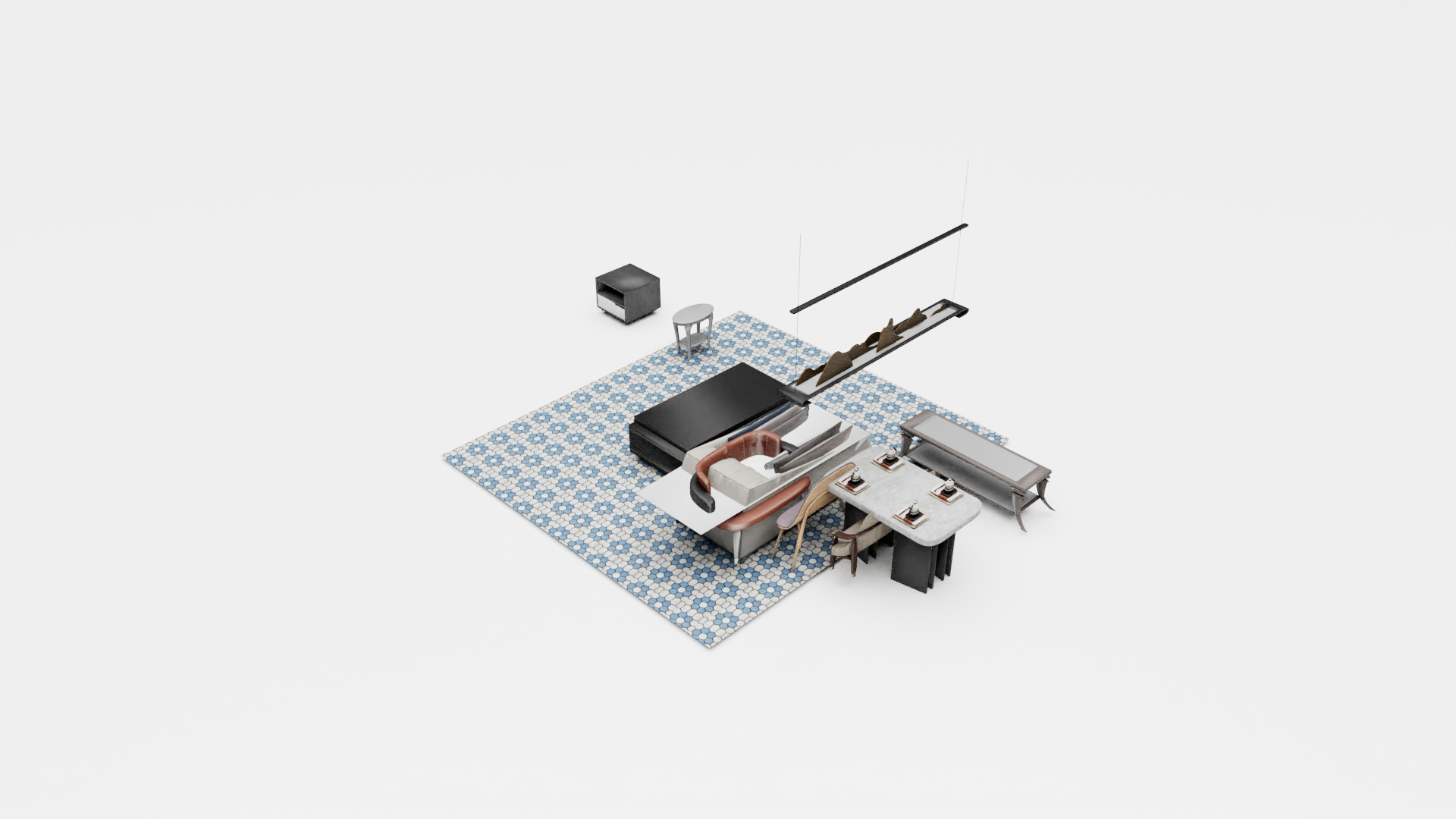}
    \end{subfigure}%
    \begin{subfigure}[b]{0.20\linewidth}
		\centering
		\includegraphics[width=\linewidth, trim=300 50 300 100, clip]{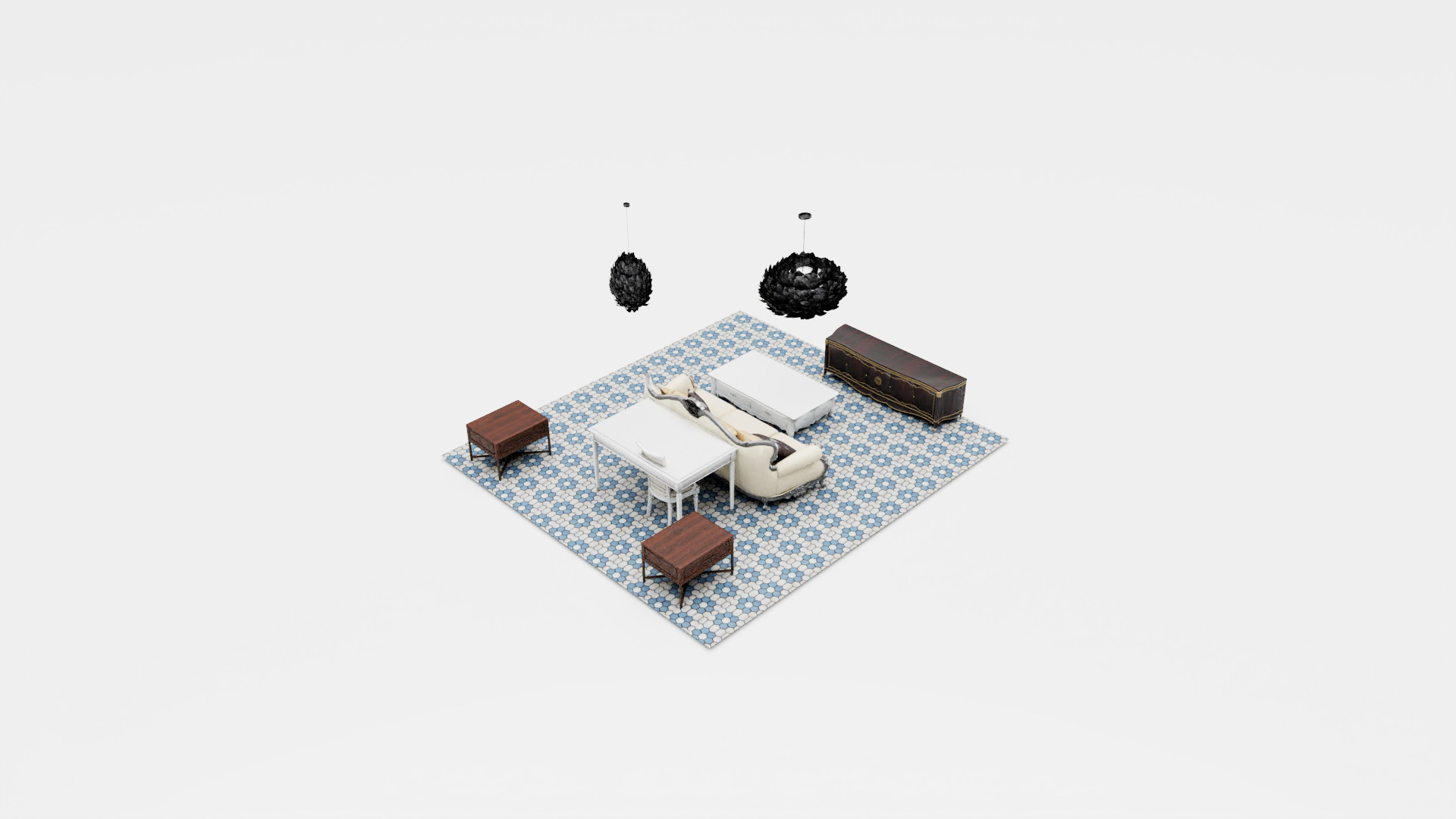}
    \end{subfigure}%
    \vskip\baselineskip%
    \vspace{-2.2em}
    \vskip\baselineskip%
    \begin{subfigure}[b]{0.20\linewidth}
		\centering
		\includegraphics[width=0.8\linewidth, trim=0 20 0 20, clip]{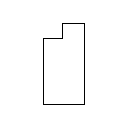}
    \end{subfigure}%
        \begin{subfigure}[b]{0.20\linewidth}
		\centering
		\includegraphics[width=\linewidth, trim=300 50 300 100, clip]{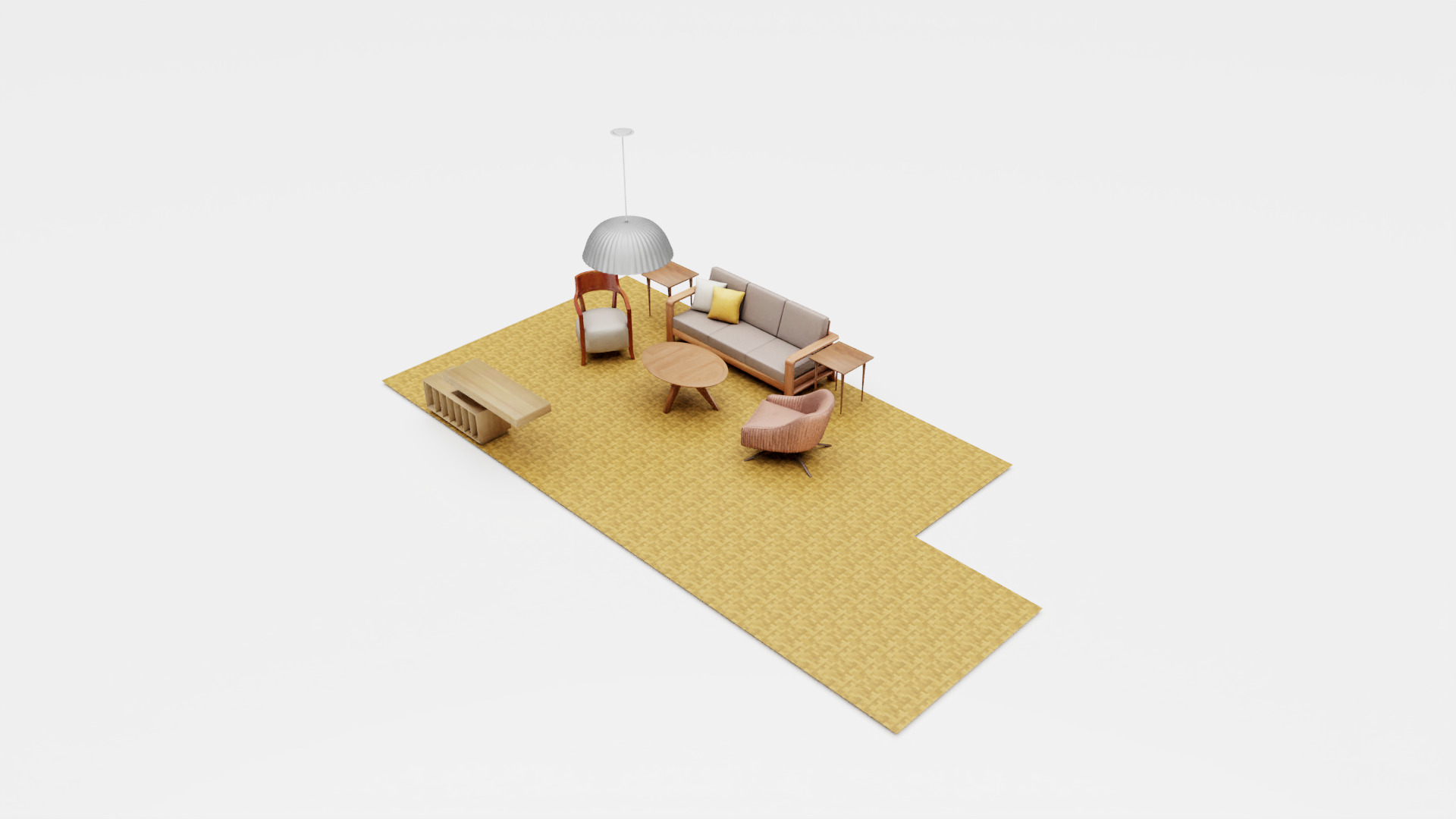}
    \end{subfigure}%
        \begin{subfigure}[b]{0.20\linewidth}
		\centering
		\includegraphics[width=\linewidth, trim=300 50 300 100, clip]{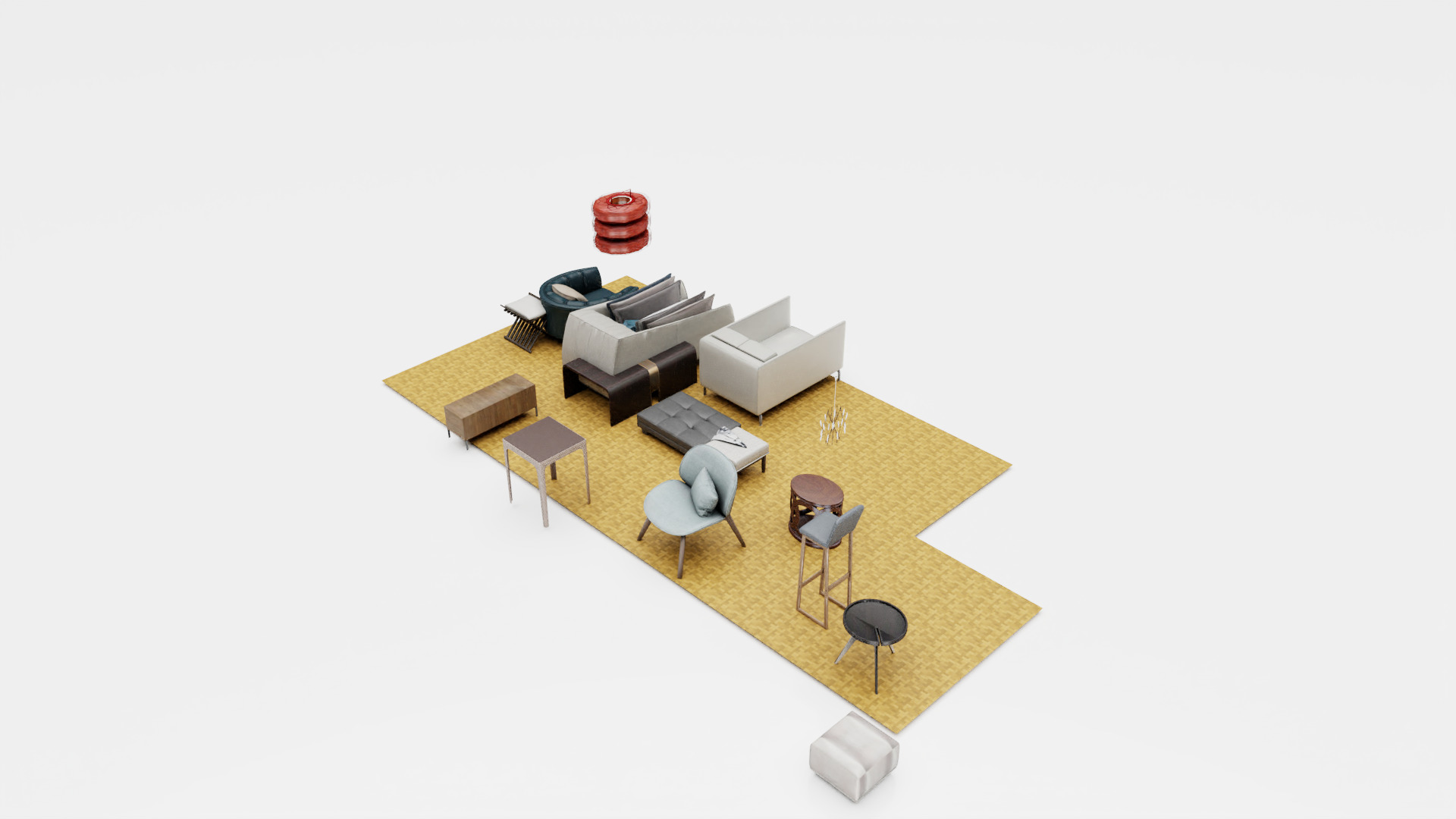}
    \end{subfigure}%
        \begin{subfigure}[b]{0.20\linewidth}
		\centering
		\includegraphics[width=\linewidth, trim=300 50 300 100, clip]{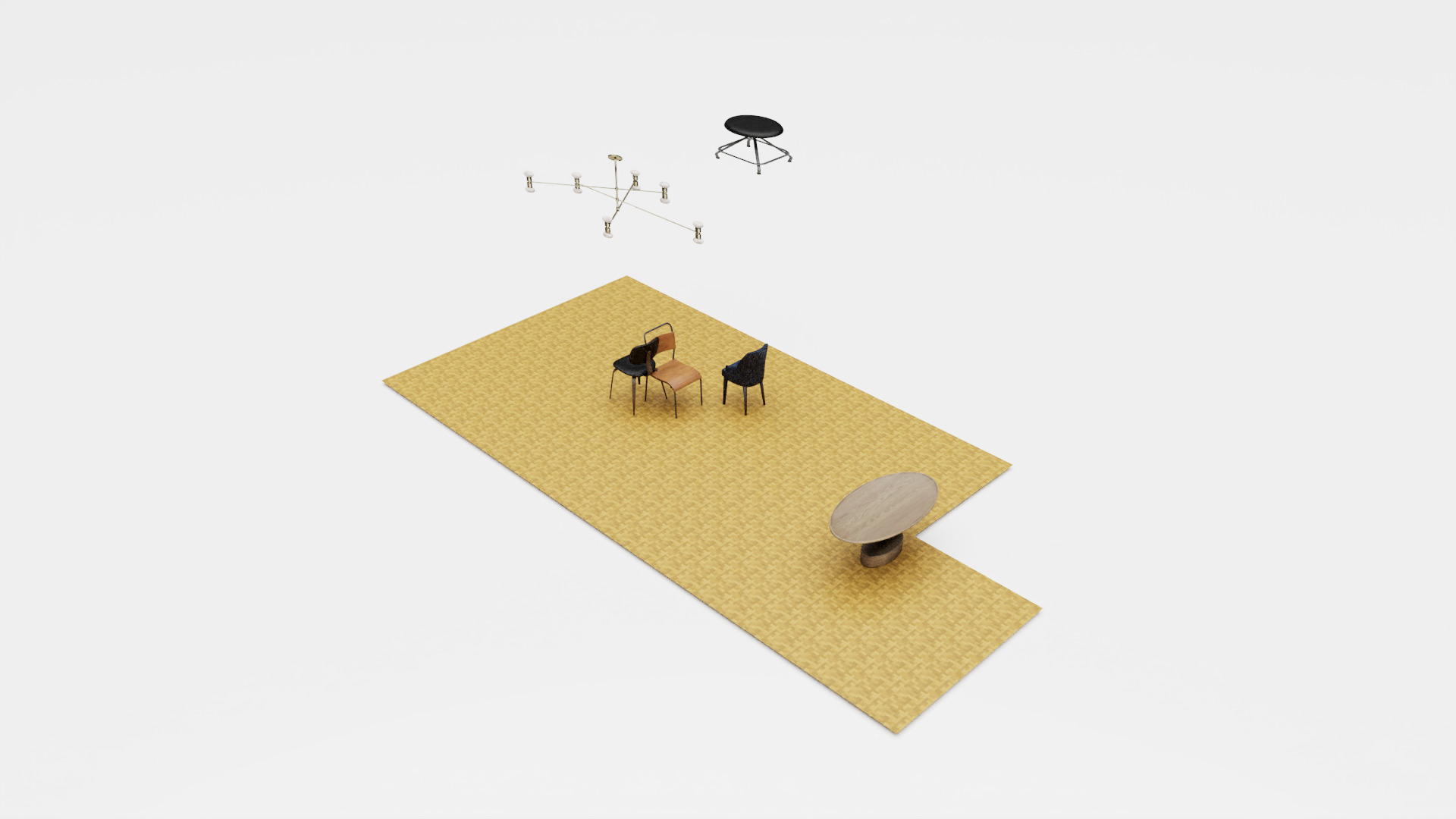}
    \end{subfigure}%
    \begin{subfigure}[b]{0.20\linewidth}
		\centering
		\includegraphics[width=\linewidth, trim=300 50 300 100, clip]{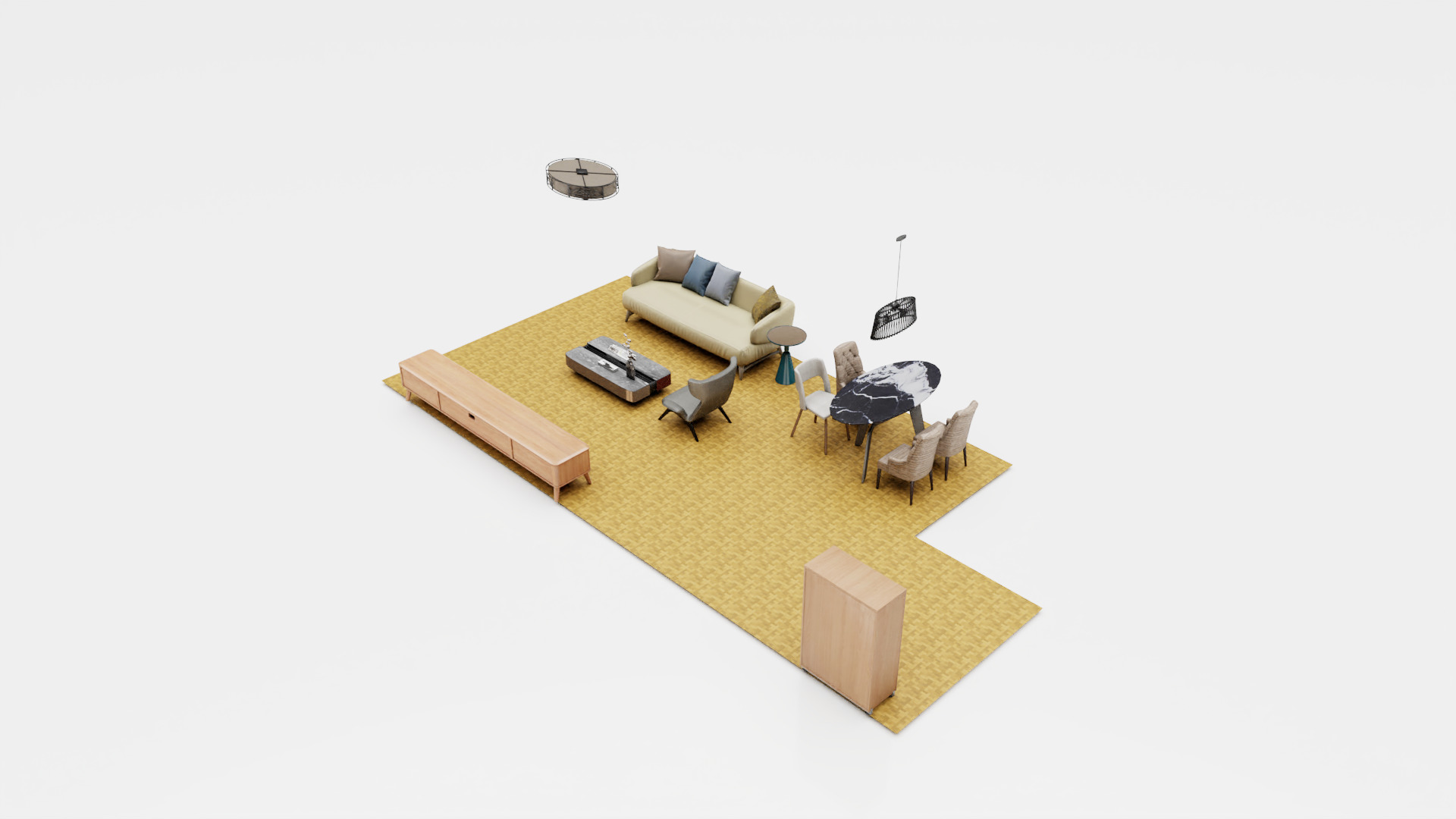}
    \end{subfigure}%
    \vskip\baselineskip%
    \vspace{-2.2em}
    \vskip\baselineskip%
    \begin{subfigure}[b]{0.20\linewidth}
		\centering
		\includegraphics[width=0.8\linewidth, trim=0 20 0 20, clip]{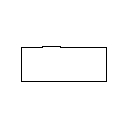}
    \end{subfigure}%
        \begin{subfigure}[b]{0.20\linewidth}
		\centering
		\includegraphics[width=\linewidth, trim=300 50 300 100, clip]{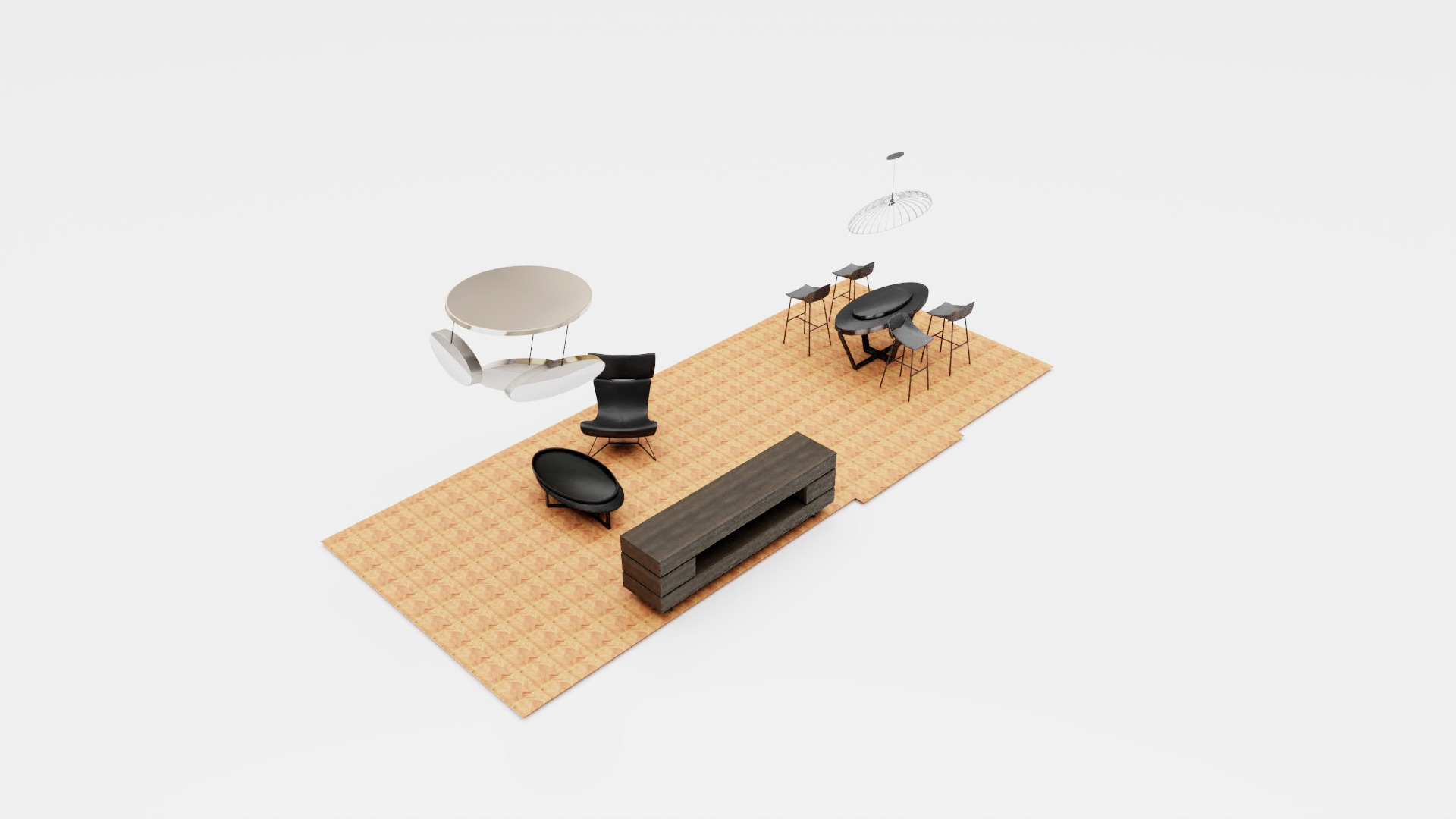}
    \end{subfigure}%
        \begin{subfigure}[b]{0.20\linewidth}
		\centering
		\includegraphics[width=\linewidth, trim=300 50 300 100, clip]{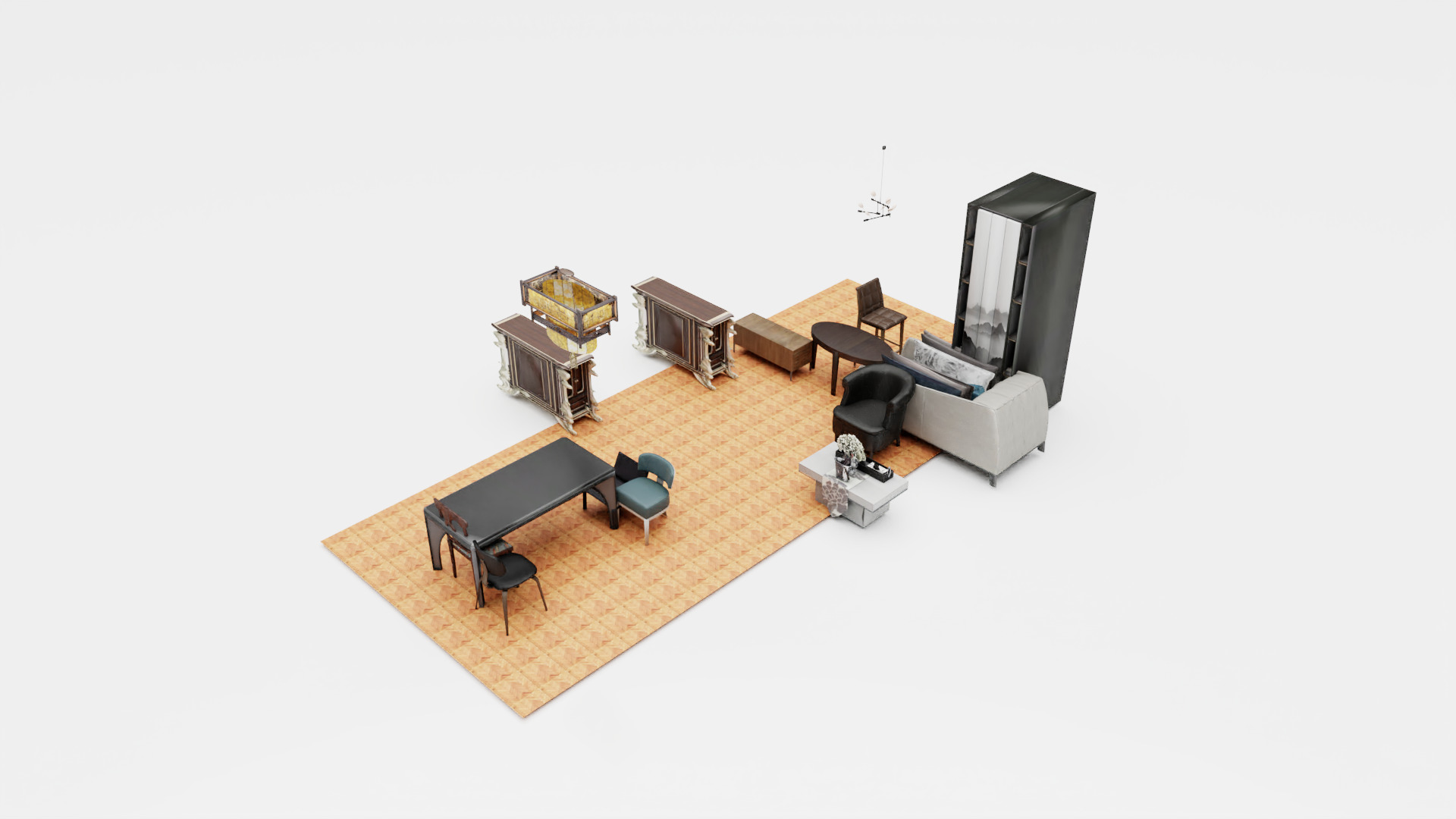}
    \end{subfigure}%
        \begin{subfigure}[b]{0.20\linewidth}
		\centering
		\includegraphics[width=\linewidth, trim=300 50 300 100, clip]{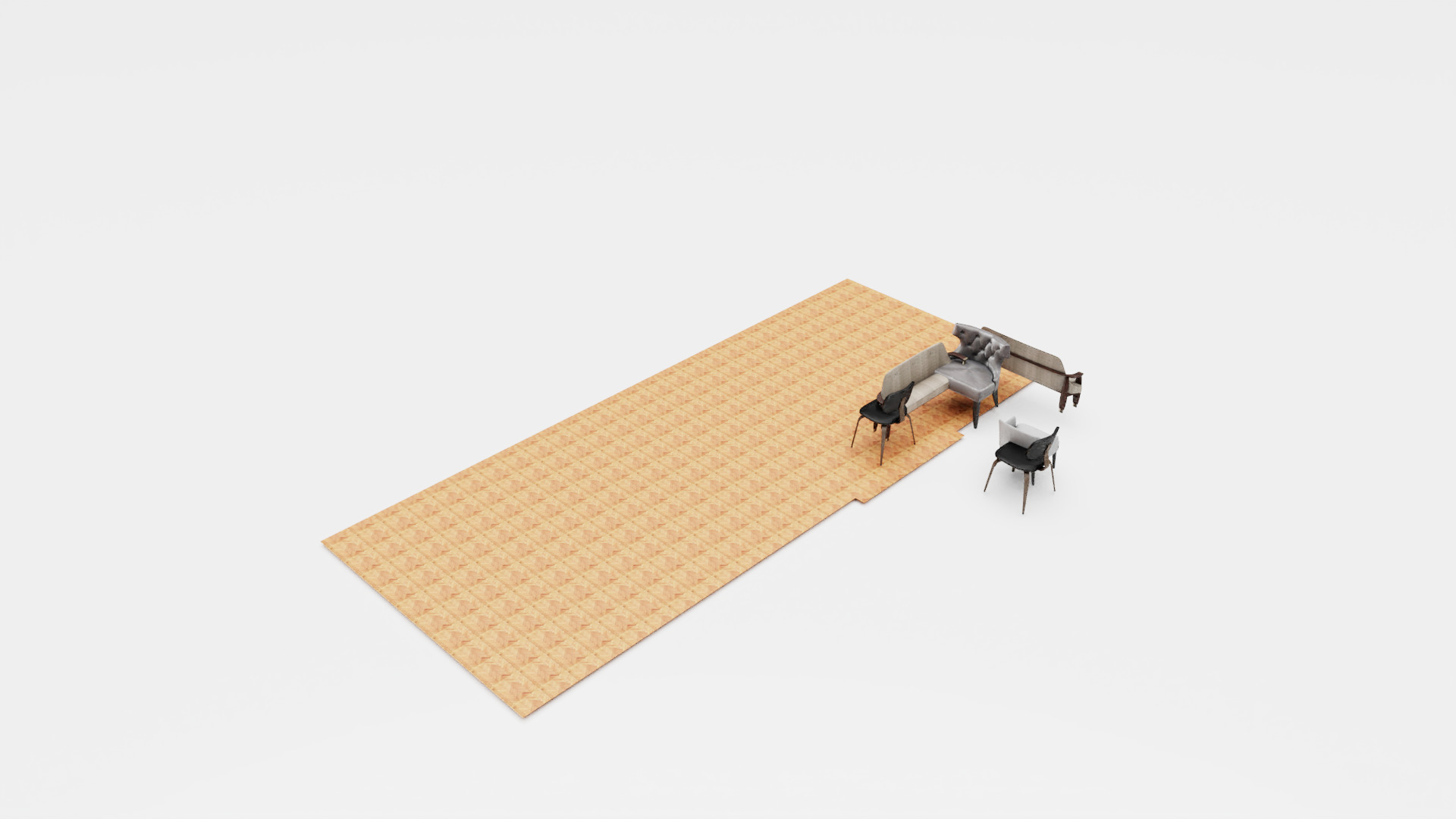}
    \end{subfigure}%
    \begin{subfigure}[b]{0.20\linewidth}
		\centering
		\includegraphics[width=\linewidth, trim=300 50 300 100, clip]{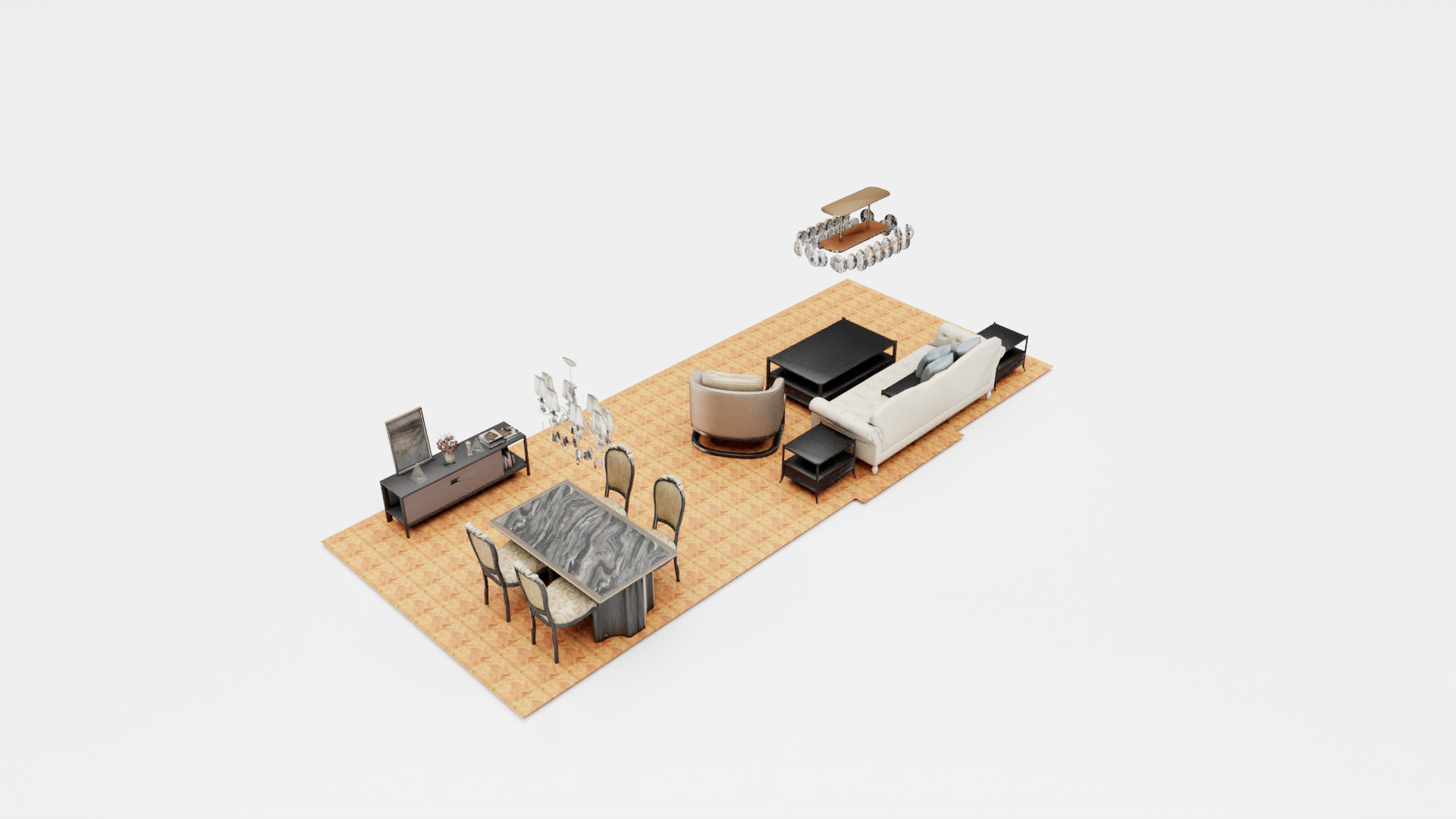}
    \end{subfigure}%
    \vskip\baselineskip%
    \vspace{-2.2em}
    \vskip\baselineskip%
    \begin{subfigure}[b]{0.20\linewidth}
		\centering
		\includegraphics[width=0.8\linewidth, trim=0 20 0 20, clip]{gfx_scene_synthesis_supp_livingroom_walls_LivingDiningRoom-1407}
    \end{subfigure}%
        \begin{subfigure}[b]{0.20\linewidth}
		\centering
		\includegraphics[width=\linewidth, trim=300 50 300 100, clip]{gfx_scene_synthesis_supp_livingroom_omni_train_LivingDiningRoom-1407}
    \end{subfigure}%
        \begin{subfigure}[b]{0.20\linewidth}
		\centering
		\includegraphics[width=\linewidth, trim=300 50 300 100, clip]{gfx_scene_synthesis_supp_livingroom_omni_fast_synth_LivingDiningRoom-1407}
    \end{subfigure}%
        \begin{subfigure}[b]{0.20\linewidth}
		\centering
		\includegraphics[width=\linewidth, trim=300 50 300 100, clip]{gfx_scene_synthesis_supp_livingroom_omni_scene_former_LivingDiningRoom-1407}
    \end{subfigure}%
    \begin{subfigure}[b]{0.20\linewidth}
		\centering
		\includegraphics[width=\linewidth, trim=300 50 300 100, clip]{gfx_scene_synthesis_supp_livingroom_omni_ours_LivingDiningRoom-1407}
    \end{subfigure}%
    \vskip\baselineskip%
    \vspace{-2.2em}
    \vskip\baselineskip%
    \begin{subfigure}[b]{0.20\linewidth}
		\centering
		\includegraphics[width=0.8\linewidth, trim=0 20 0 20, clip]{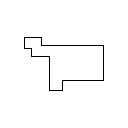}
    \end{subfigure}%
        \begin{subfigure}[b]{0.20\linewidth}
		\centering
		\includegraphics[width=\linewidth, trim=300 50 300 100, clip]{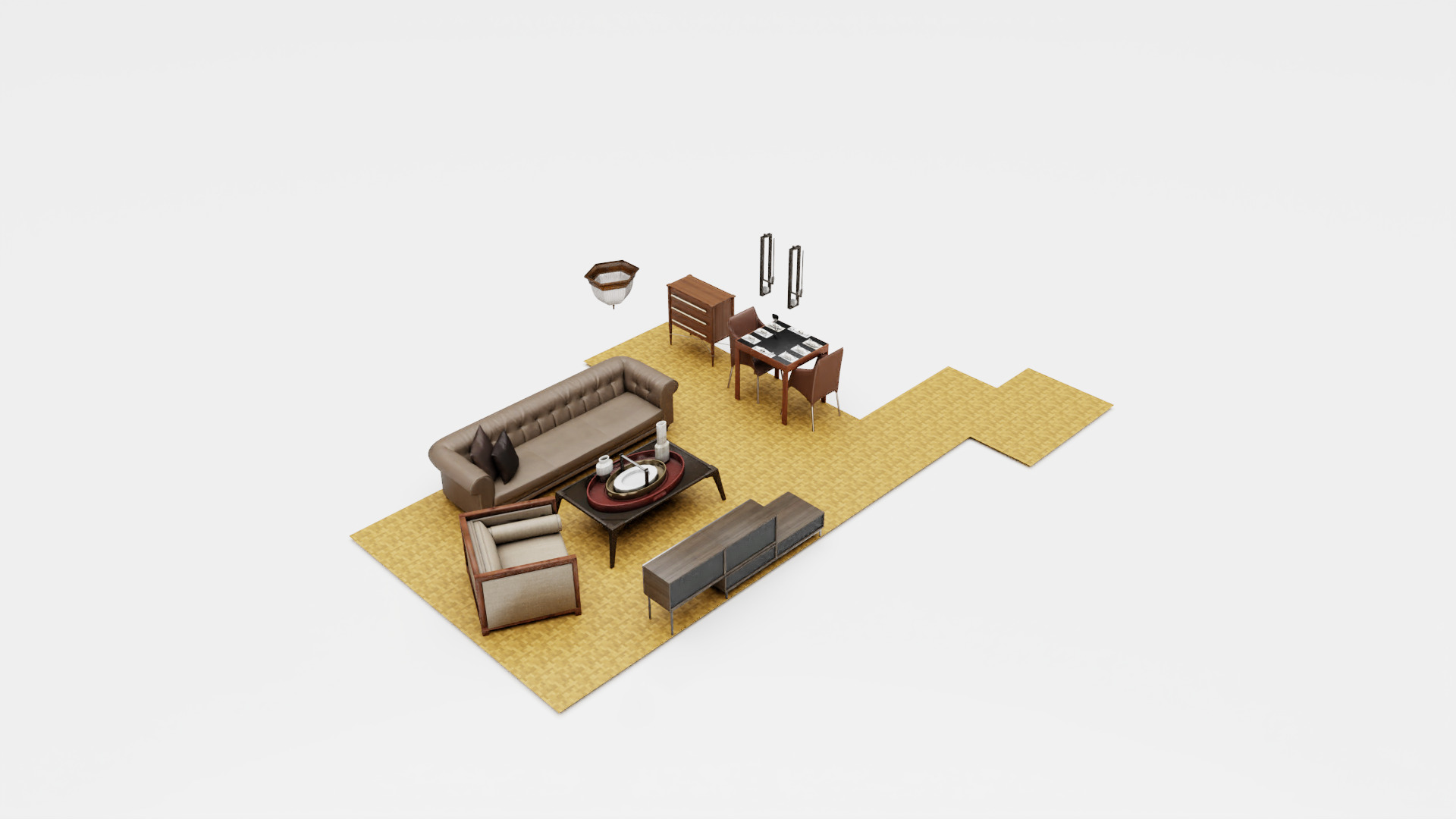}
    \end{subfigure}%
        \begin{subfigure}[b]{0.20\linewidth}
		\centering
		\includegraphics[width=\linewidth, trim=300 50 300 100, clip]{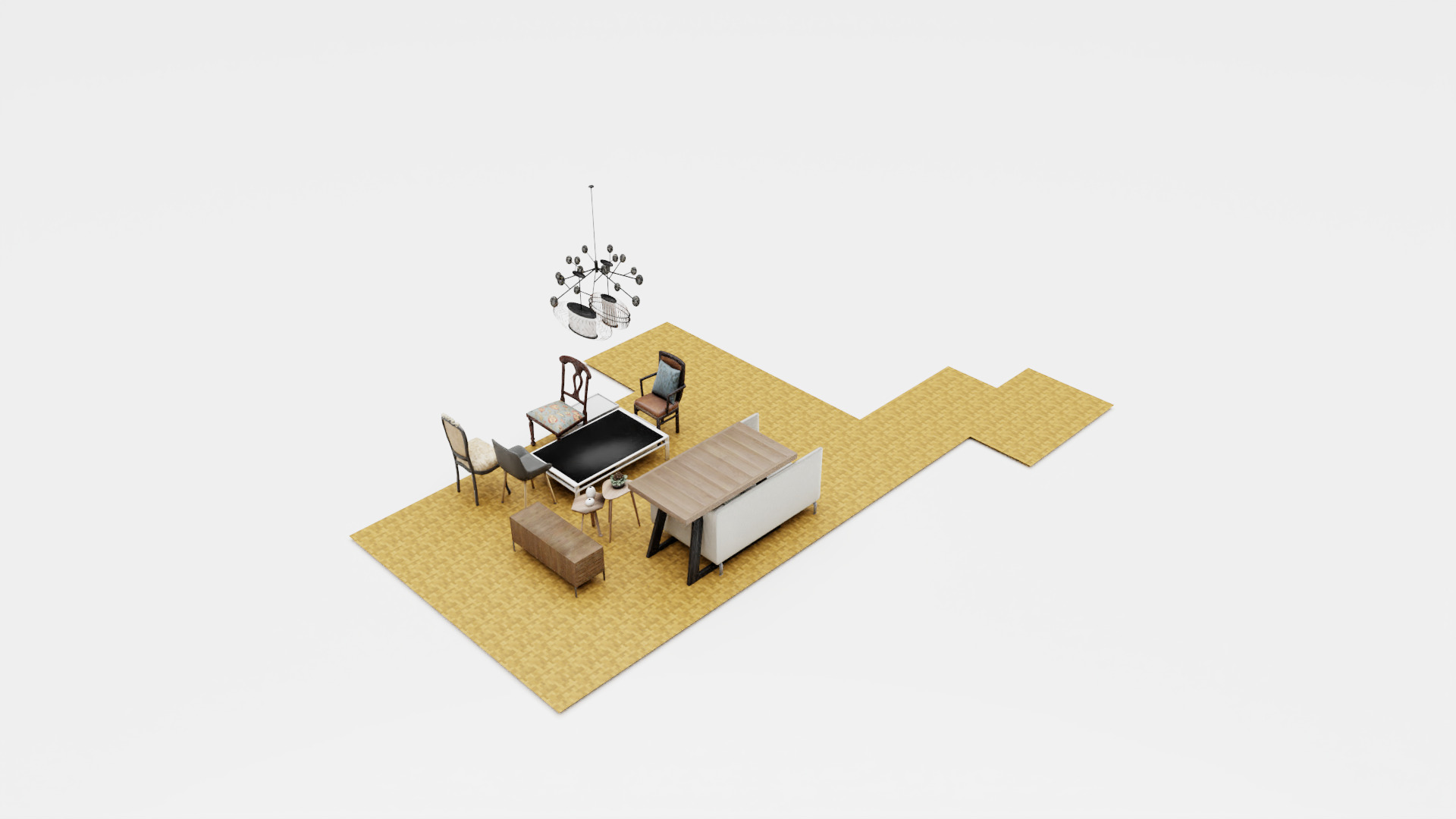}
    \end{subfigure}%
        \begin{subfigure}[b]{0.20\linewidth}
		\centering
		\includegraphics[width=\linewidth, trim=300 50 300 100, clip]{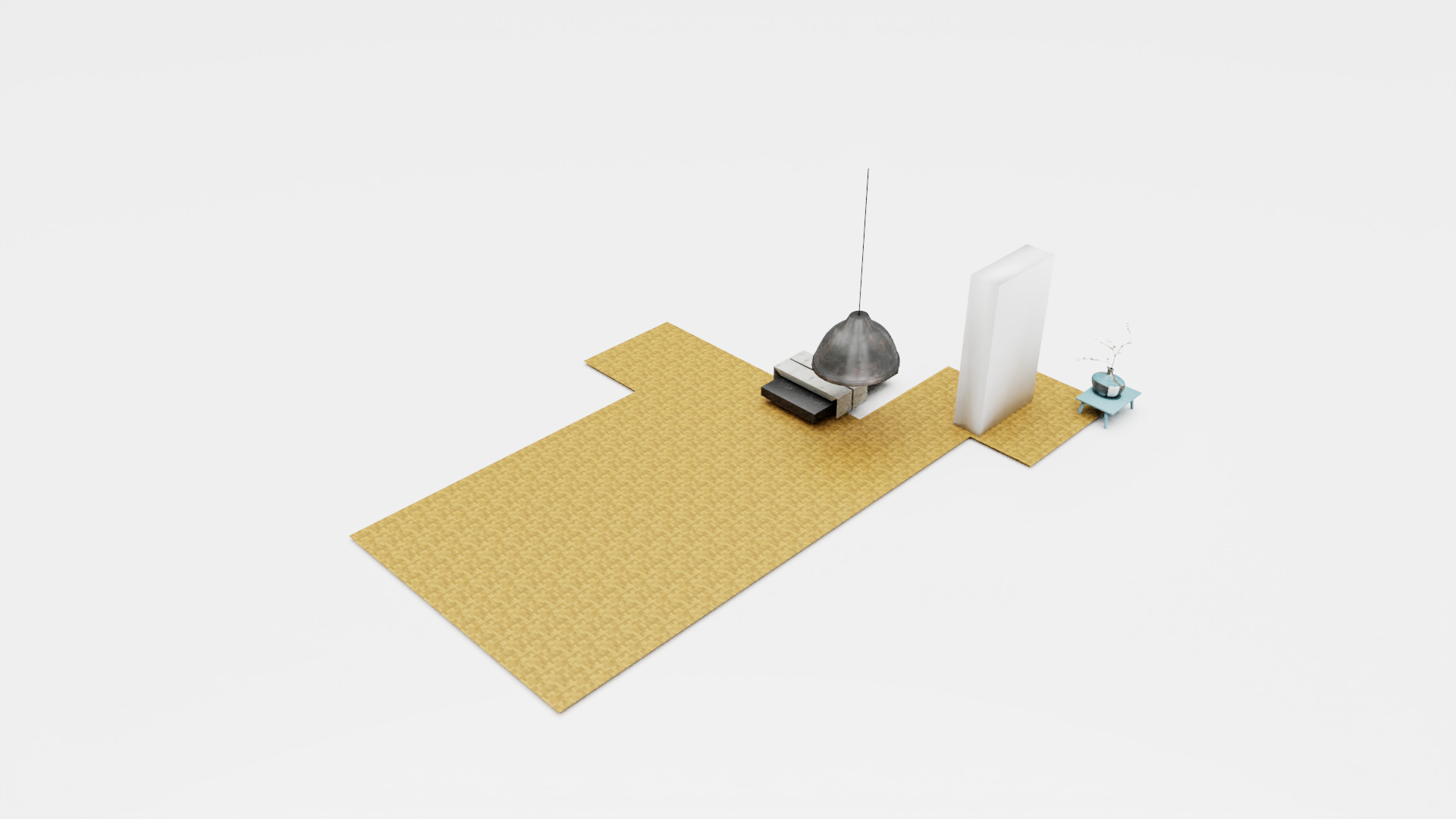}
    \end{subfigure}%
    \begin{subfigure}[b]{0.20\linewidth}
		\centering
		\includegraphics[width=\linewidth, trim=300 50 300 100, clip]{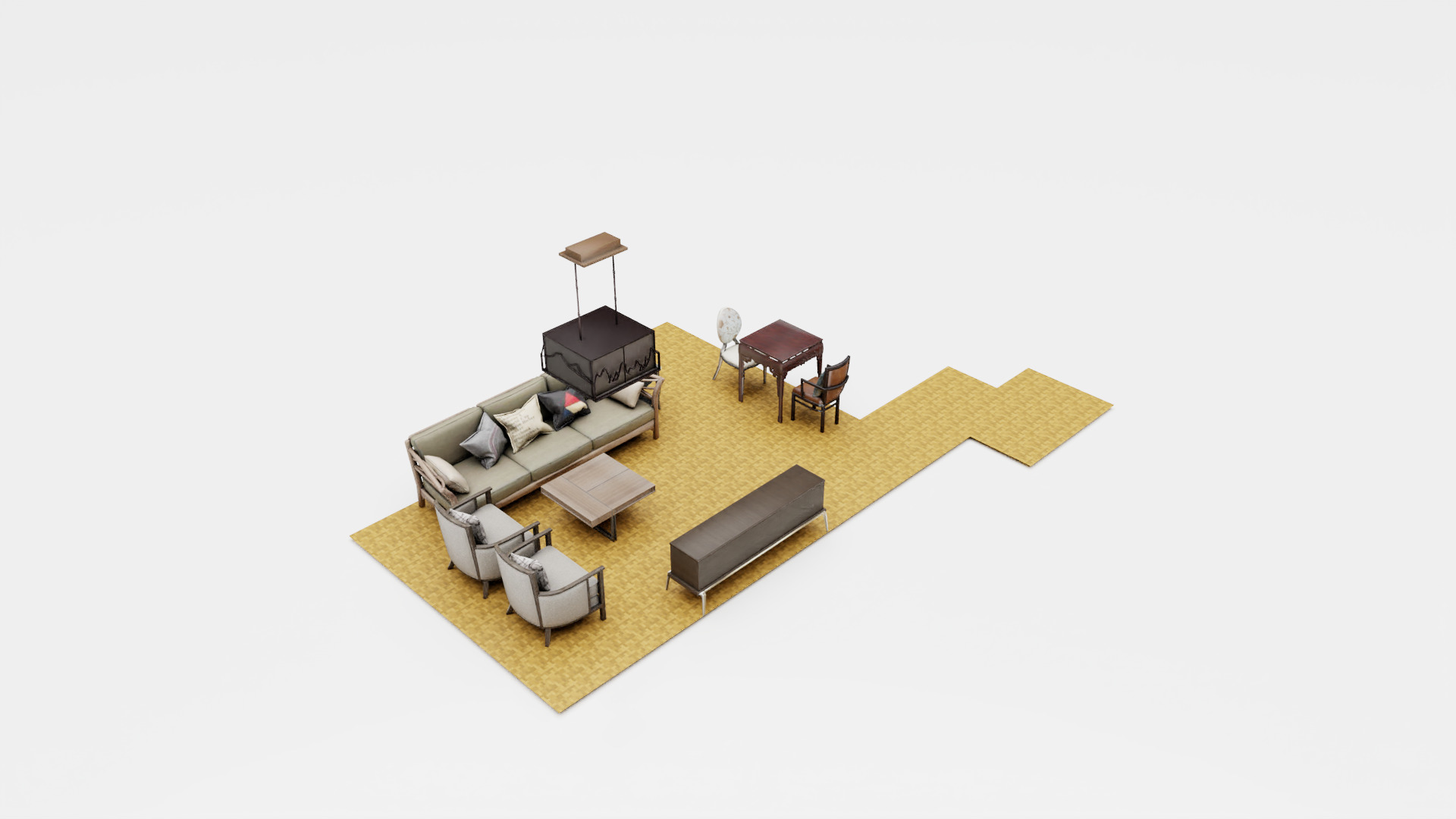}
    \end{subfigure}%
    \vspace{-1.2em}
    \vskip\baselineskip%
    \hfill%
    \caption{{\bf Qualitative Scene Synthesis Results on Living Rooms}.
    Generated scenes for living rooms using FastSynth, SceneFormer and our method.
    To showcase the generalization abilities of our model we also show the
    closest scene from the training set (2nd column).}
    \label{fig:scene_synthesis_qualitative_livingroom_supp}
    \vspace{-1.2em}
\end{figure}

\begin{figure}[!h]
    \centering
    \begin{subfigure}[b]{0.20\linewidth}
		\centering
        \small Scene Layout
    \end{subfigure}%
    \begin{subfigure}[b]{0.20\linewidth}
		\centering
        \small Training Sample
    \end{subfigure}%
    \begin{subfigure}[b]{0.20\linewidth}
		\centering
        \small FastSynth
    \end{subfigure}%
    \begin{subfigure}[b]{0.20\linewidth}
		\centering
        \small SceneFormer
    \end{subfigure}%
    \begin{subfigure}[b]{0.20\linewidth}
        \centering
        \small Ours
    \end{subfigure}
    \hfill%
    \vskip\baselineskip%
    \vspace{-1.5em}
    \hfill%
    \begin{subfigure}[b]{0.20\linewidth}
		\centering
		\includegraphics[width=0.8\linewidth]{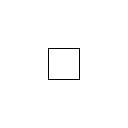}
    \end{subfigure}%
        \begin{subfigure}[b]{0.20\linewidth}
		\centering
		\includegraphics[width=\linewidth, trim=500 200 500 50, clip]{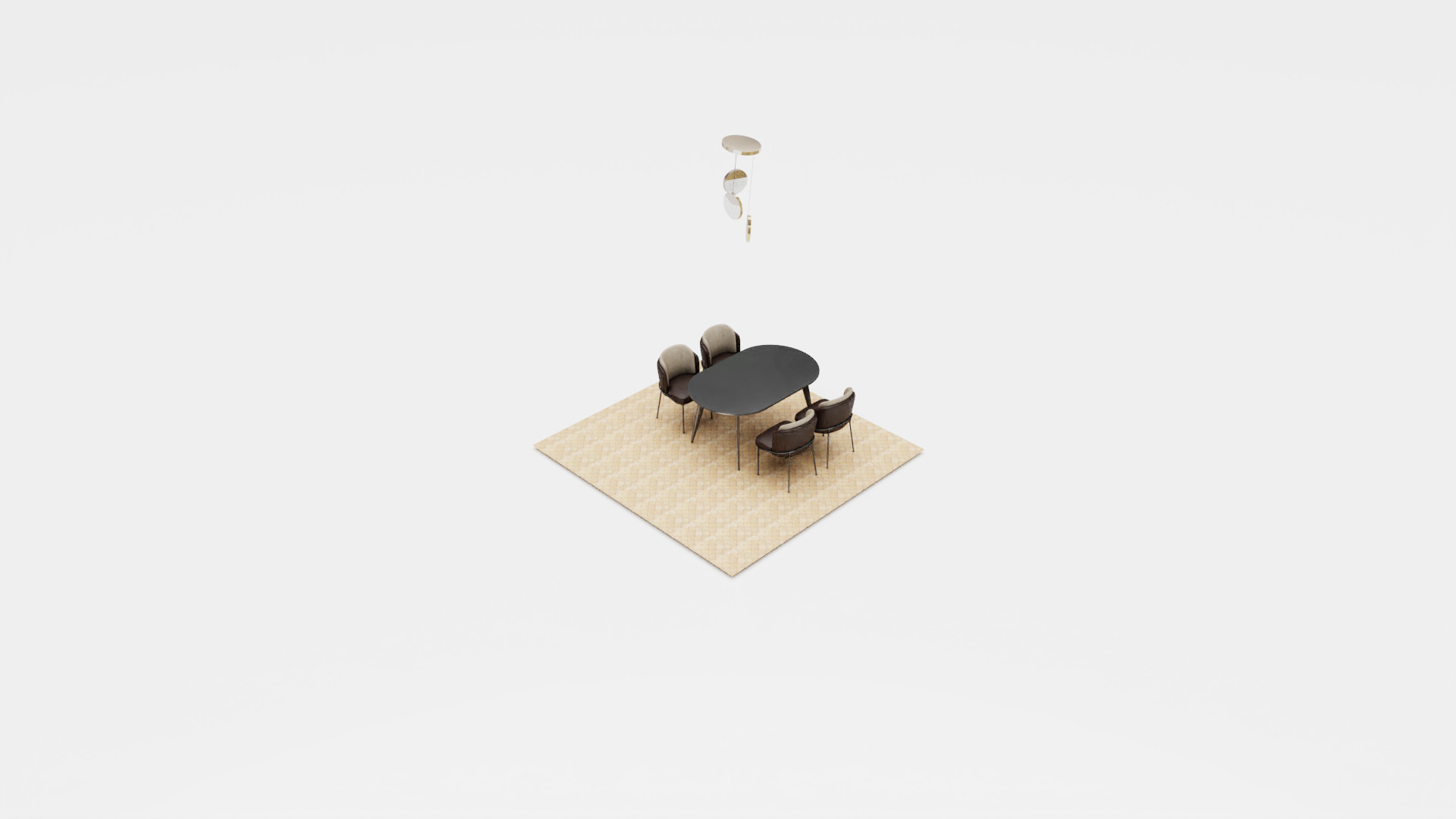}
    \end{subfigure}%
        \begin{subfigure}[b]{0.20\linewidth}
		\centering
		\includegraphics[width=\linewidth, trim=500 200 500 50, clip]{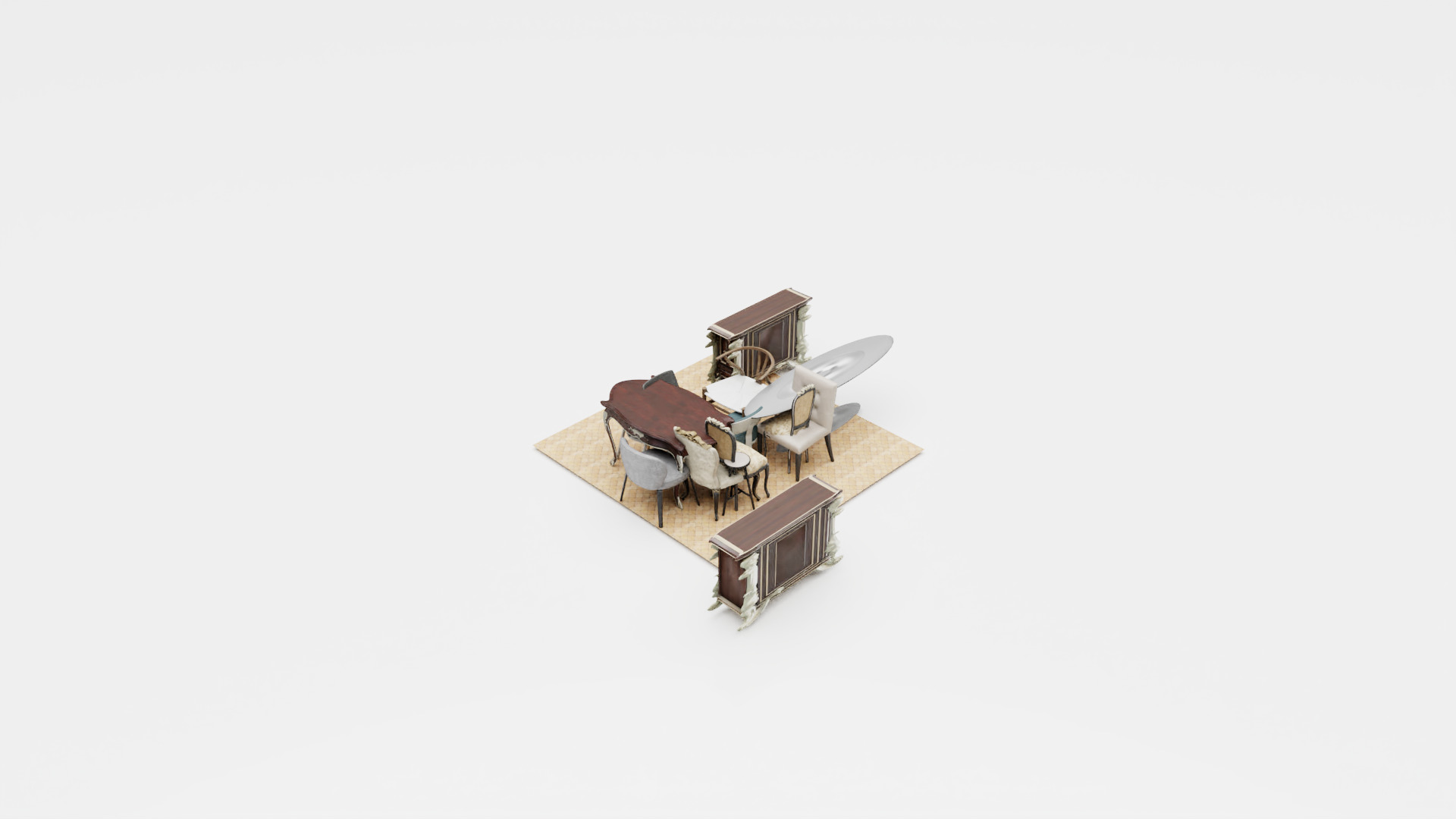}
    \end{subfigure}%
        \begin{subfigure}[b]{0.20\linewidth}
		\centering
		\includegraphics[width=\linewidth, trim=500 200 500 50, clip]{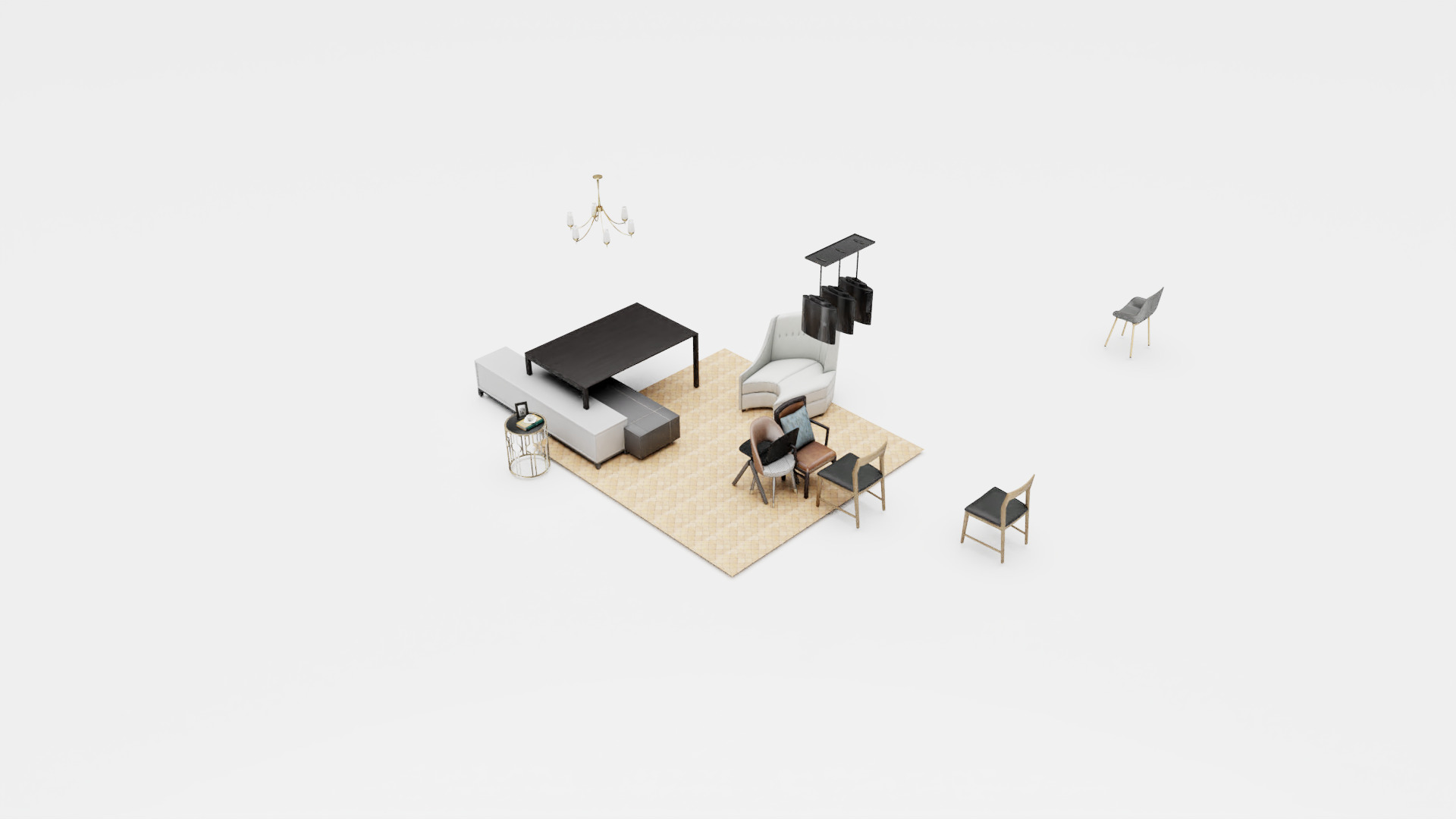}
    \end{subfigure}%
    \begin{subfigure}[b]{0.20\linewidth}
		\centering
		\includegraphics[width=\linewidth, trim=500 200 500 50, clip]{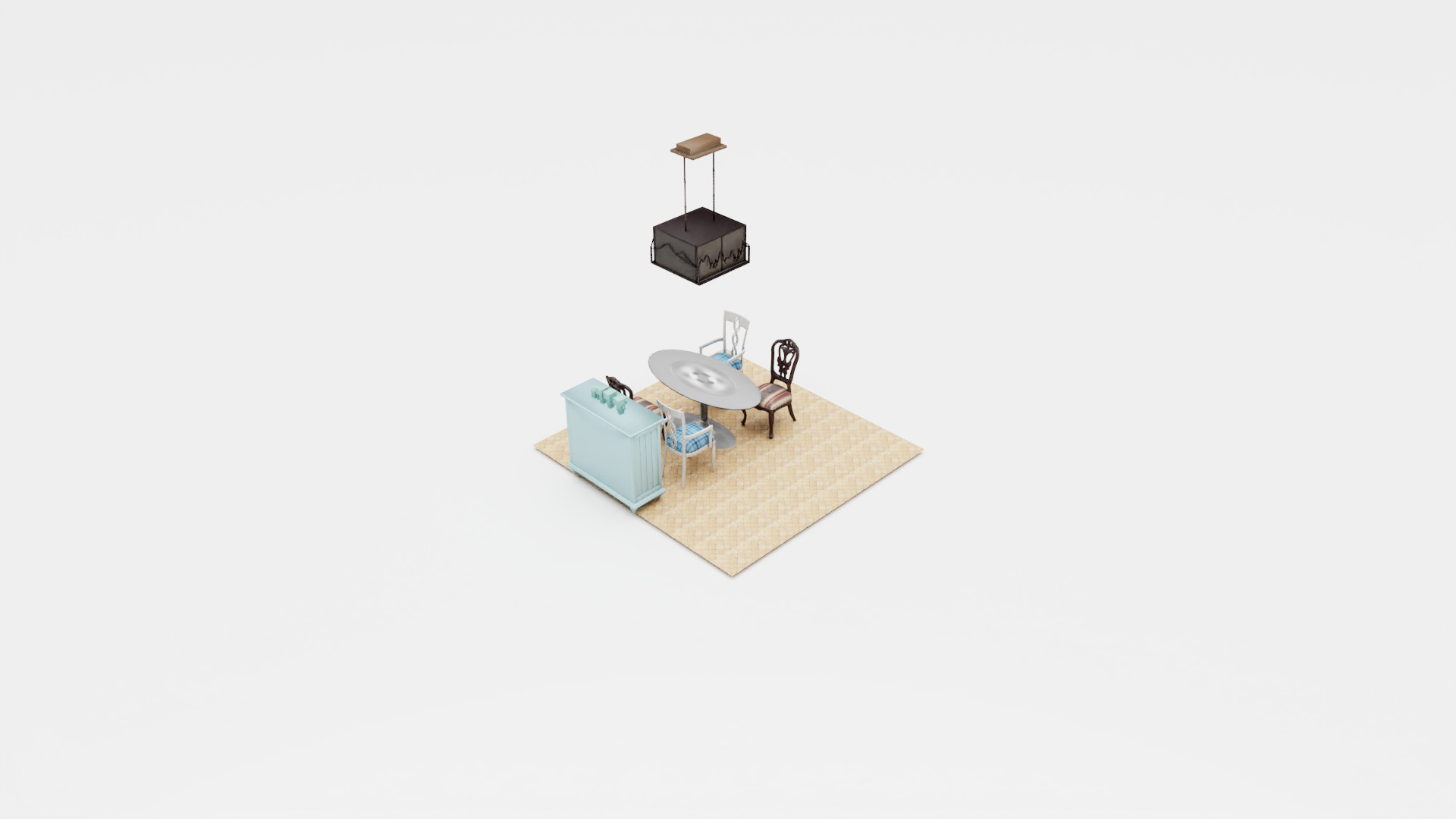}
    \end{subfigure}%
    \vskip\baselineskip%
    \vspace{-2.2em}
    \vskip\baselineskip%
    \begin{subfigure}[b]{0.20\linewidth}
		\centering
		\includegraphics[width=0.8\linewidth]{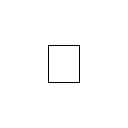}
    \end{subfigure}%
        \begin{subfigure}[b]{0.20\linewidth}
		\centering
		\includegraphics[width=\linewidth, trim=500 200 500 50, clip]{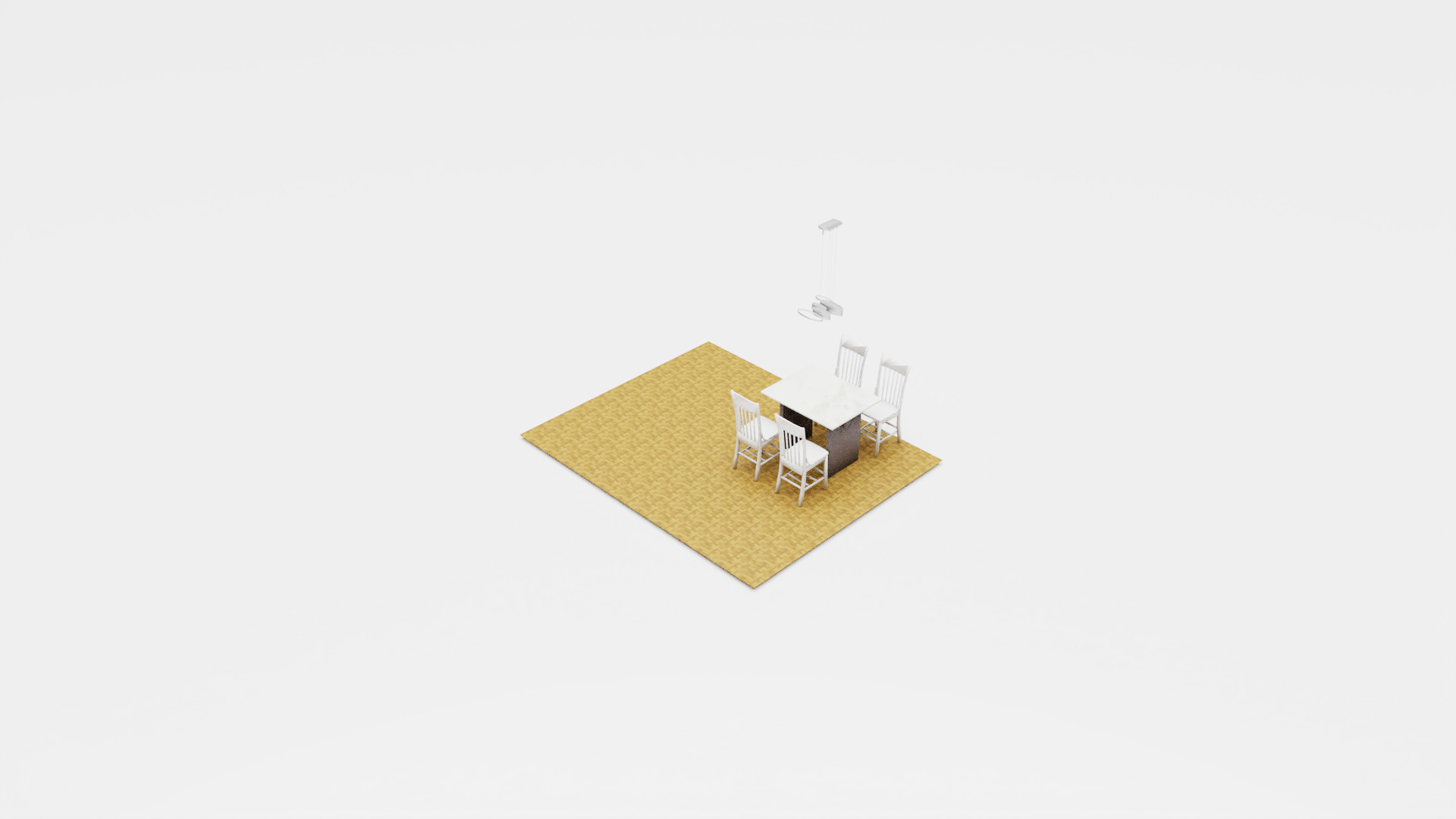}
    \end{subfigure}%
        \begin{subfigure}[b]{0.20\linewidth}
		\centering
		\includegraphics[width=\linewidth, trim=500 200 500 50, clip]{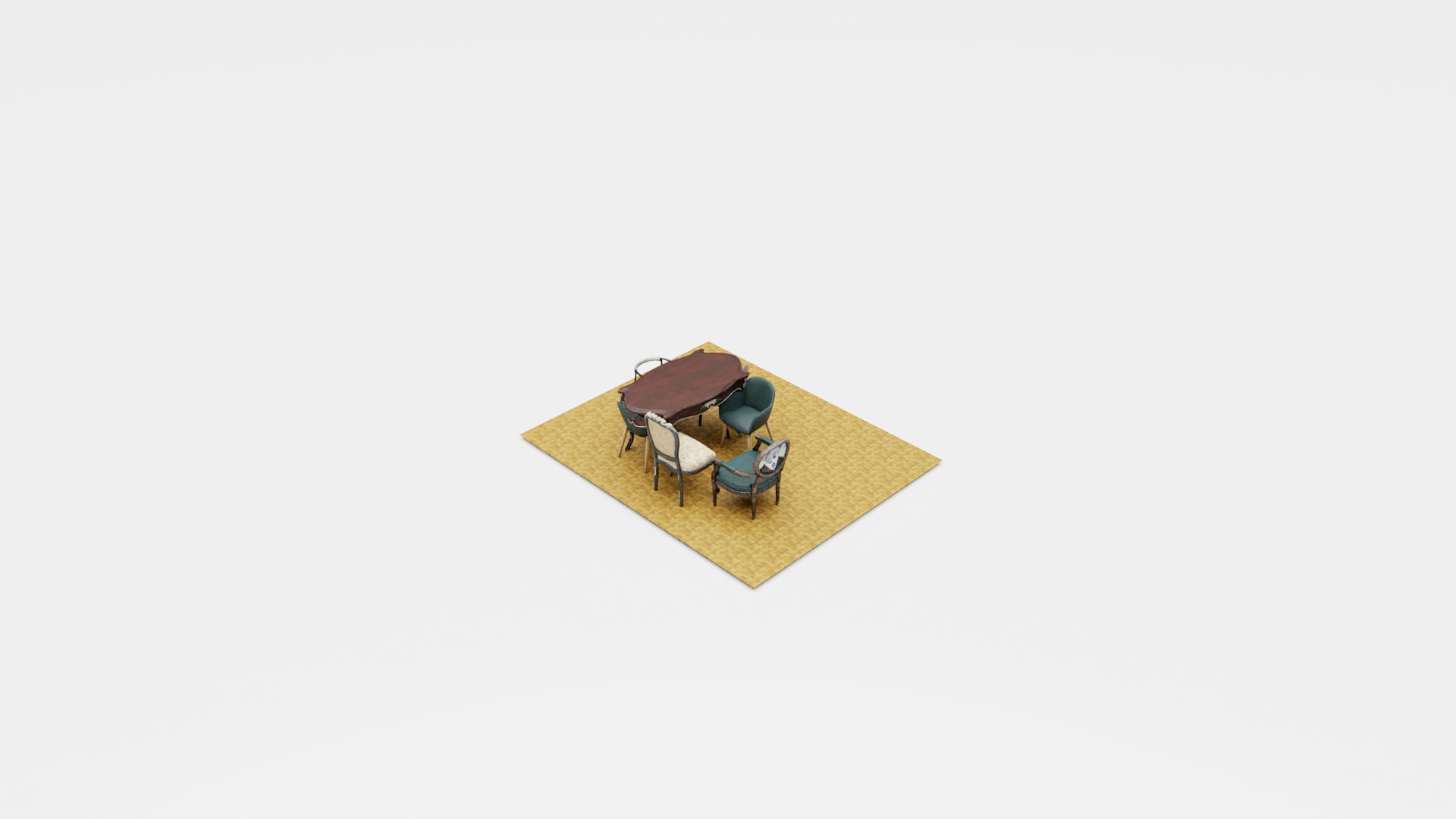}
    \end{subfigure}%
        \begin{subfigure}[b]{0.20\linewidth}
		\centering
		\includegraphics[width=\linewidth, trim=500 200 500 50, clip]{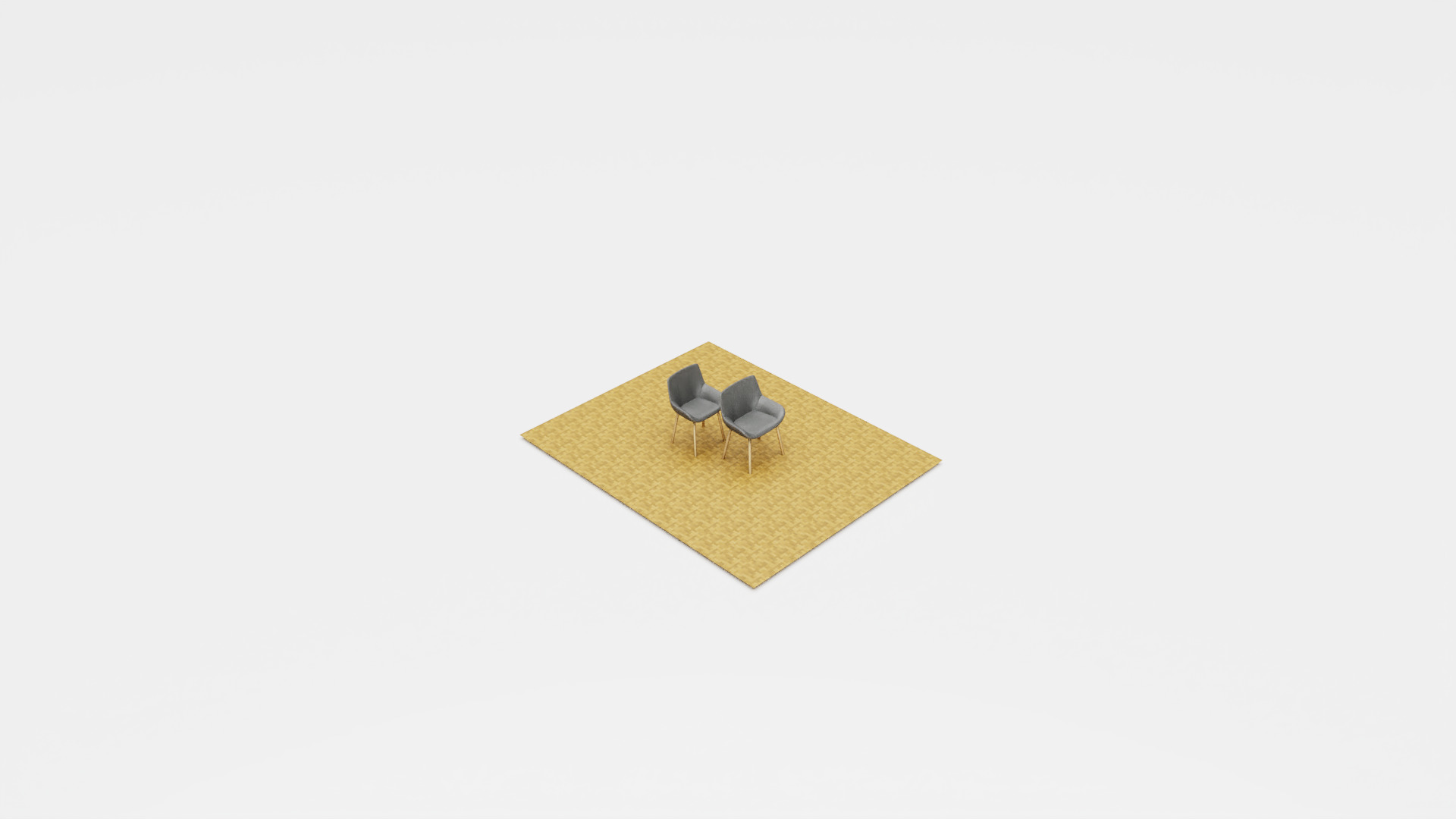}
    \end{subfigure}%
    \begin{subfigure}[b]{0.20\linewidth}
		\centering
		\includegraphics[width=\linewidth, trim=500 200 500 50, clip]{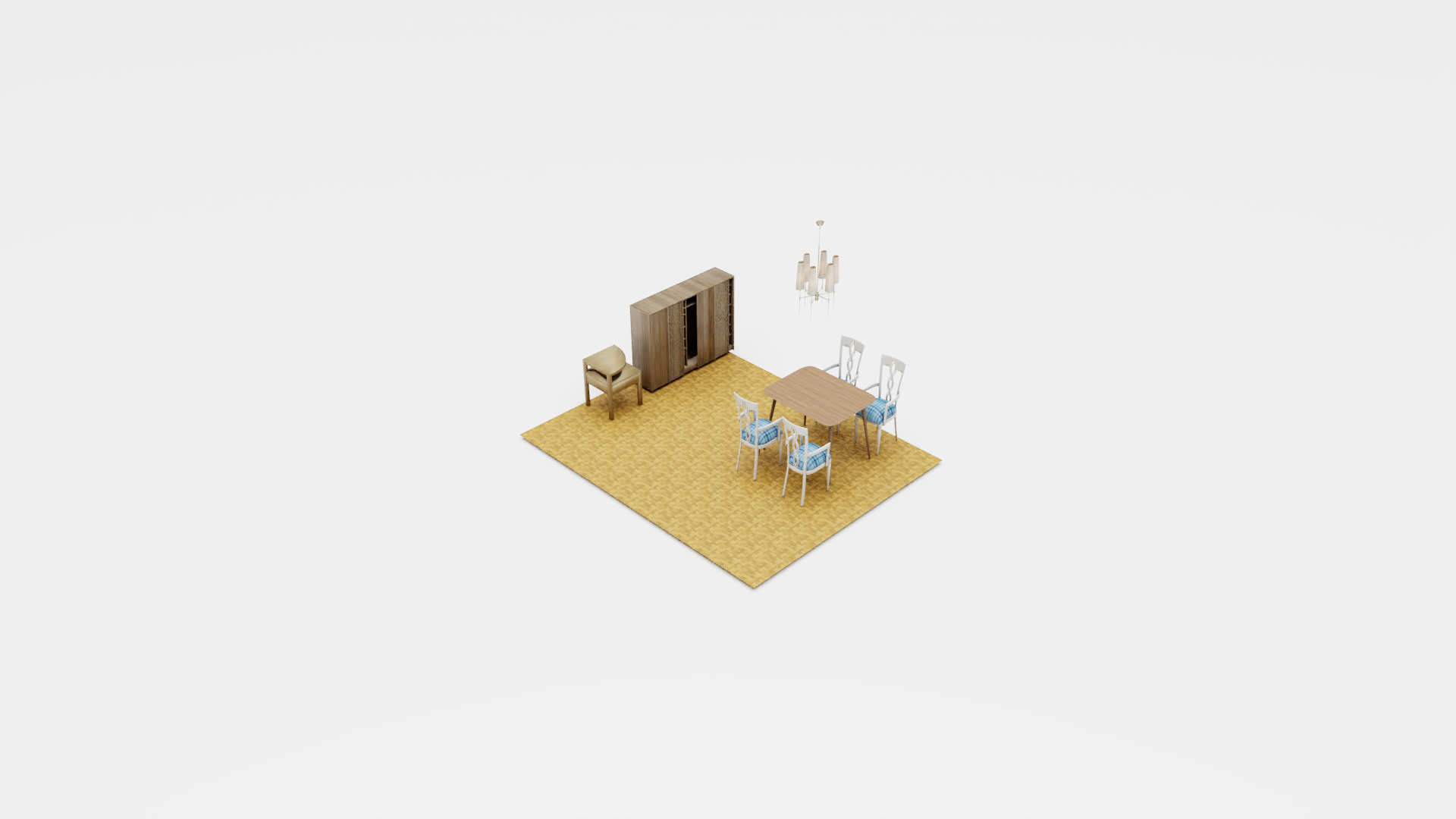}
    \end{subfigure}%
    \vskip\baselineskip%
    \vspace{-2.2em}
    \vskip\baselineskip%
    \begin{subfigure}[b]{0.20\linewidth}
		\centering
		\includegraphics[width=0.8\linewidth]{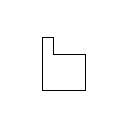}
    \end{subfigure}%
        \begin{subfigure}[b]{0.20\linewidth}
		\centering
		\includegraphics[width=\linewidth, trim=500 200 500 50, clip]{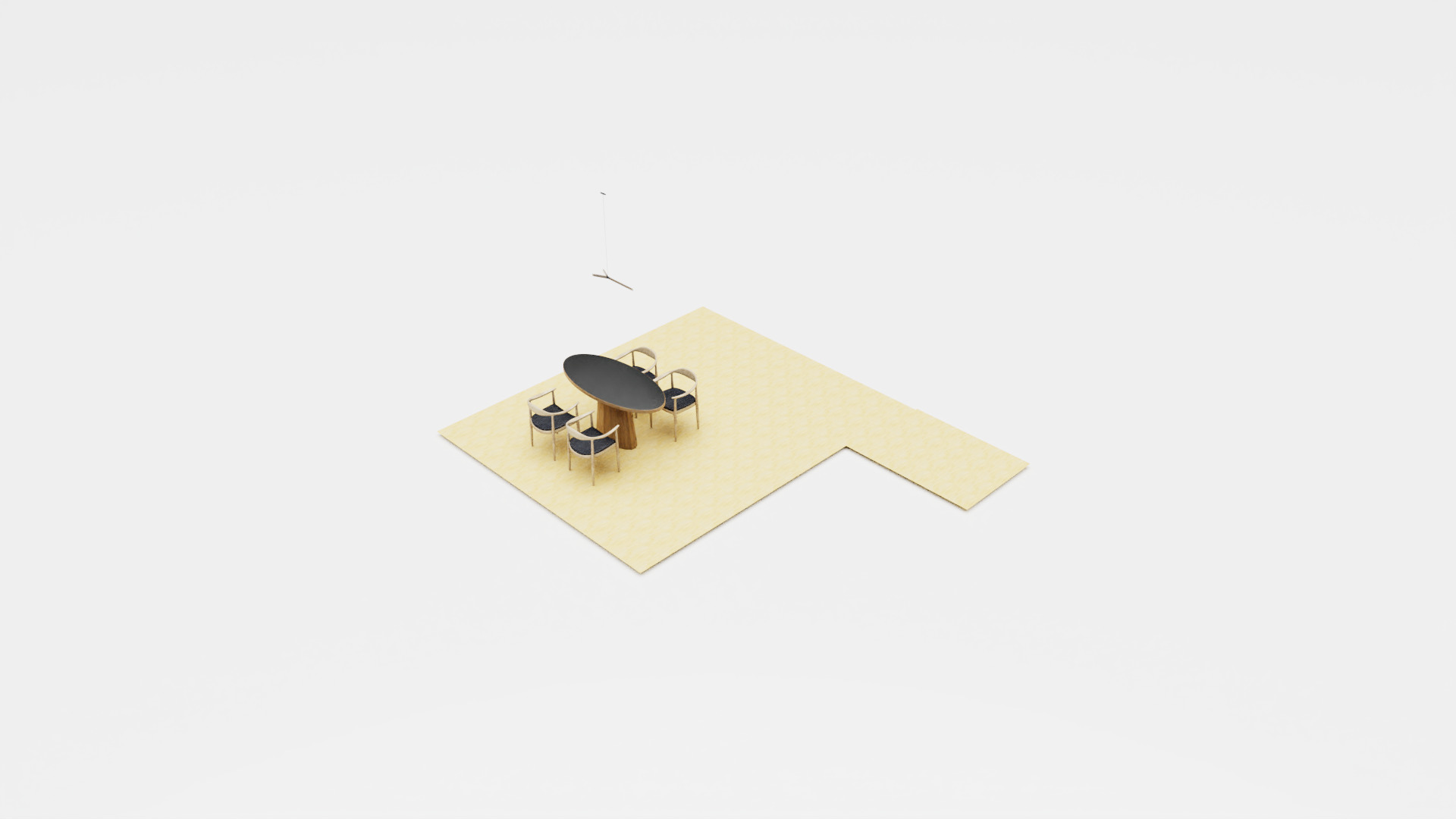}
    \end{subfigure}%
        \begin{subfigure}[b]{0.20\linewidth}
		\centering
		\includegraphics[width=\linewidth, trim=500 200 500 50, clip]{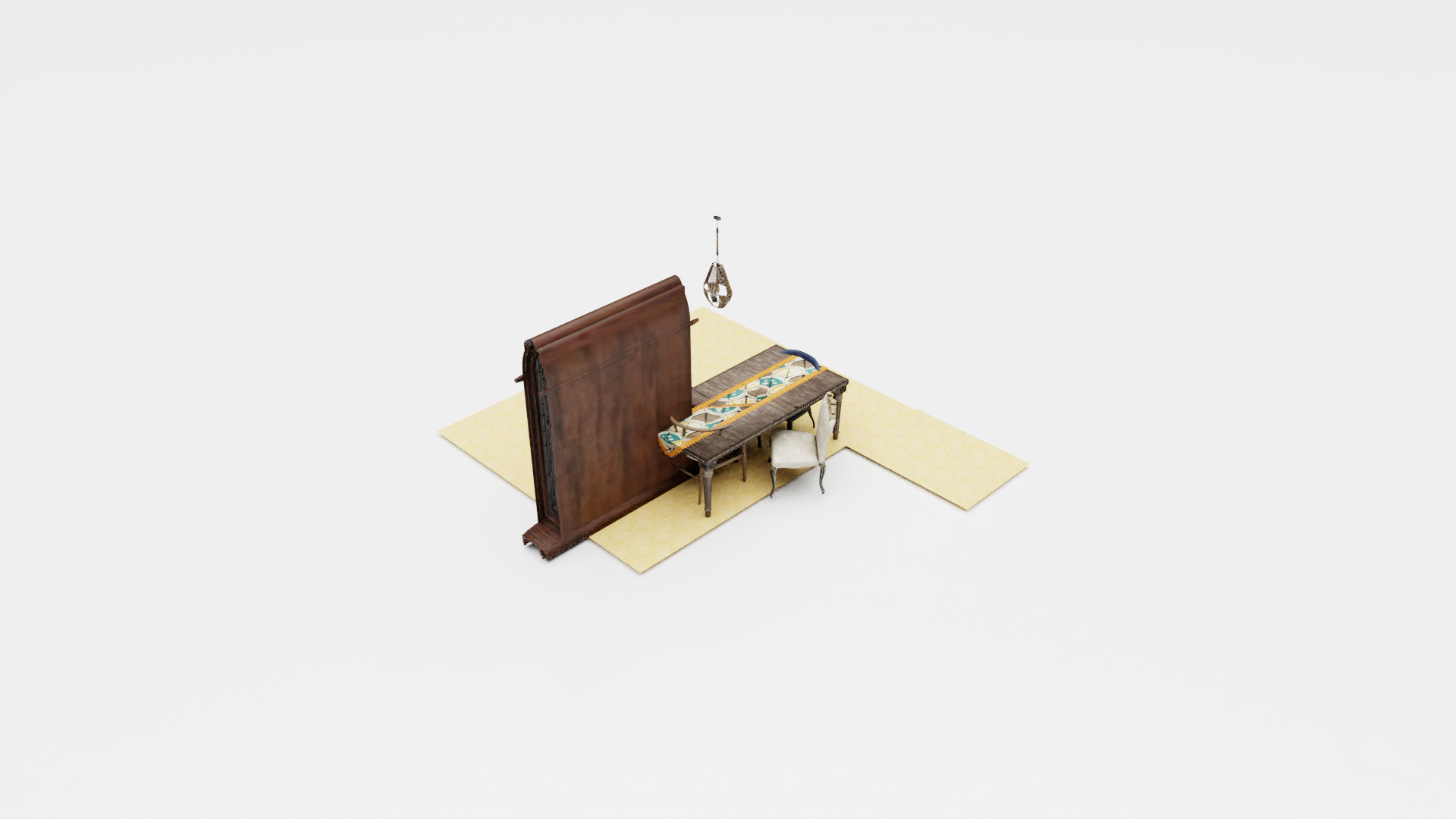}
    \end{subfigure}%
        \begin{subfigure}[b]{0.20\linewidth}
		\centering
		\includegraphics[width=\linewidth, trim=500 200 500 50, clip]{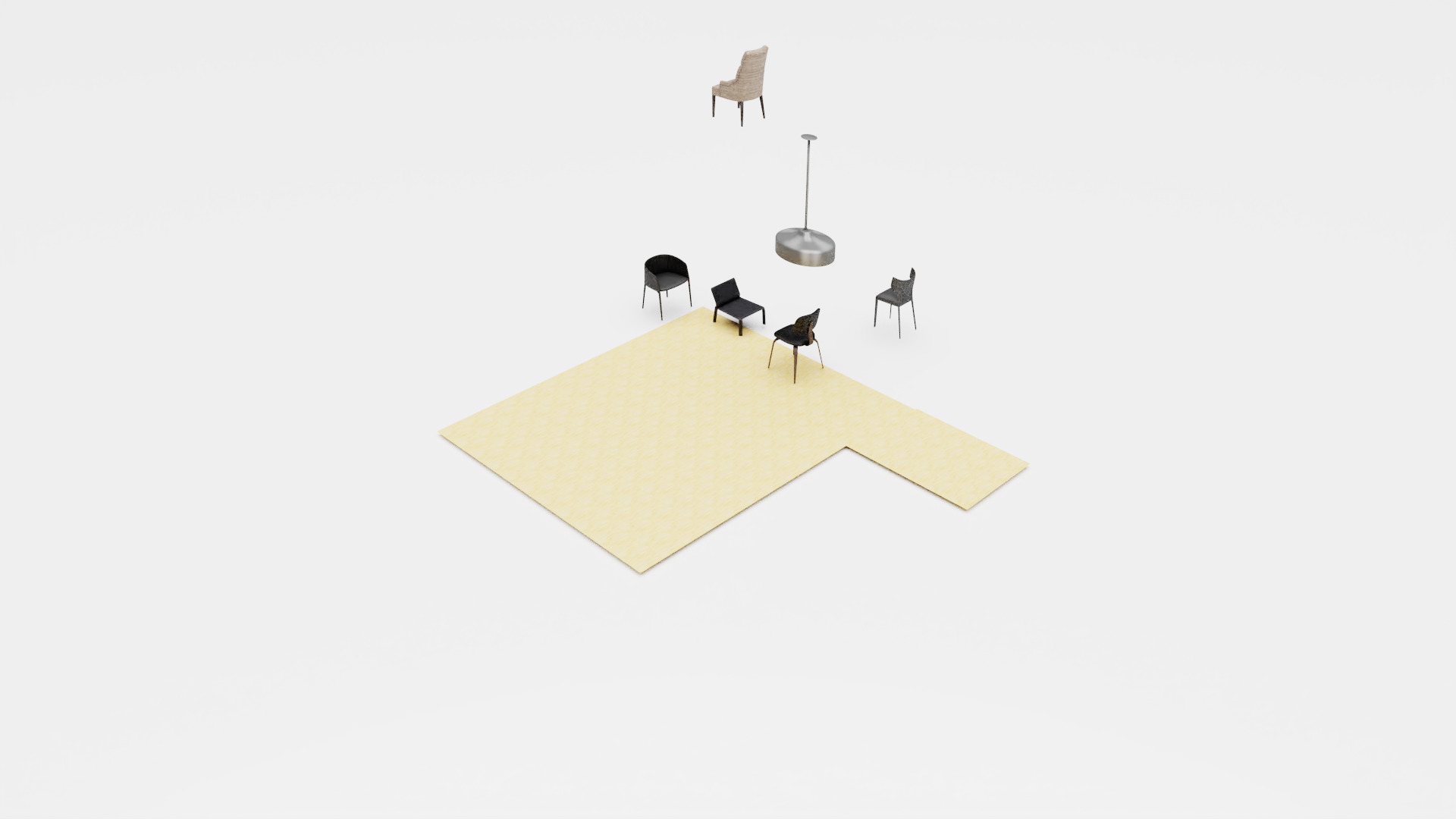}
    \end{subfigure}%
    \begin{subfigure}[b]{0.20\linewidth}
		\centering
		\includegraphics[width=\linewidth, trim=500 200 500 50, clip]{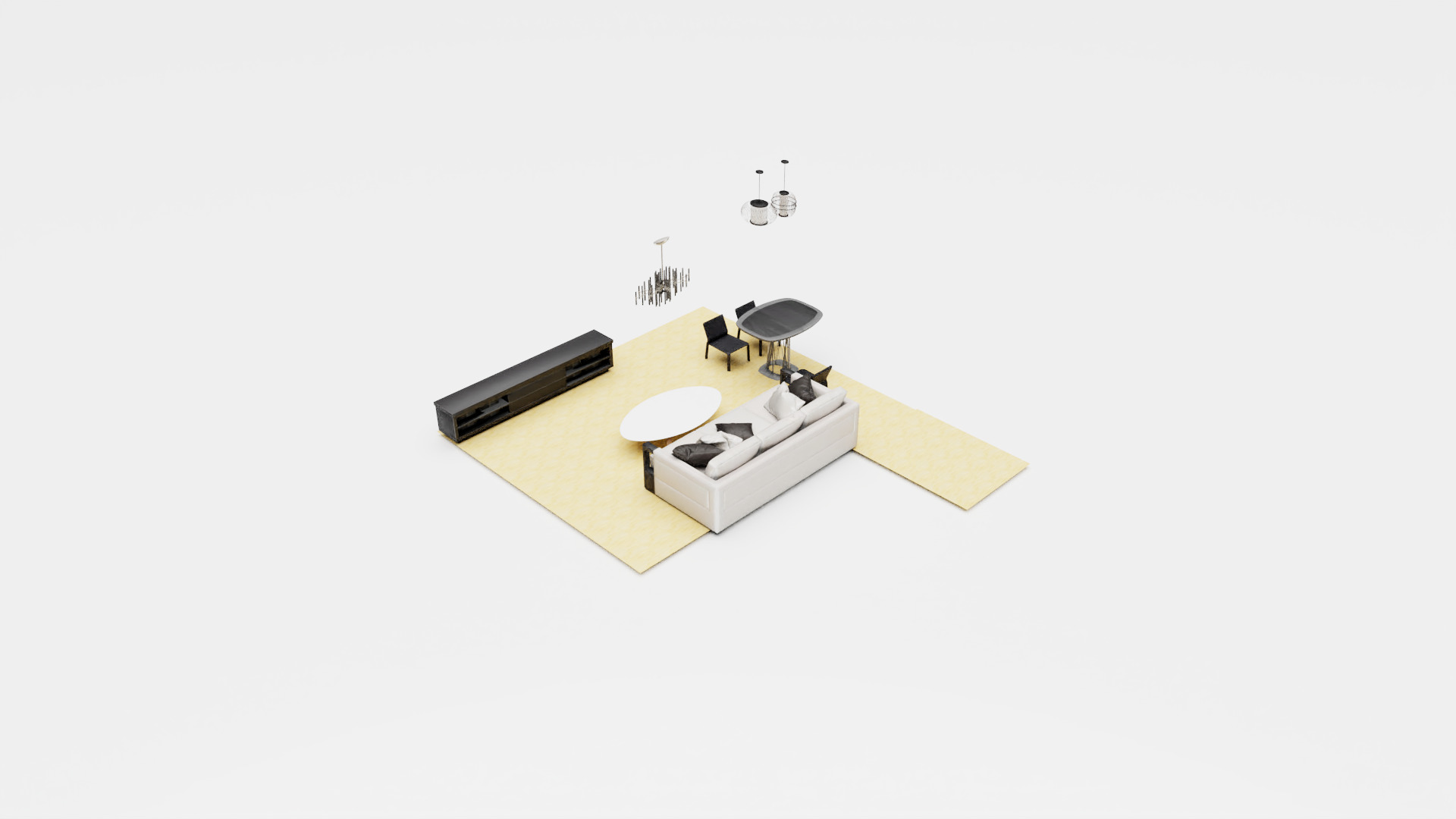}
    \end{subfigure}%
    \vskip\baselineskip%
    \vspace{-2.2em}
    \vskip\baselineskip%
    \begin{subfigure}[b]{0.20\linewidth}
		\centering
		\includegraphics[width=0.8\linewidth]{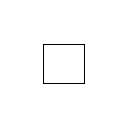}
    \end{subfigure}%
        \begin{subfigure}[b]{0.20\linewidth}
		\centering
		\includegraphics[width=\linewidth, trim=500 200 500 50, clip]{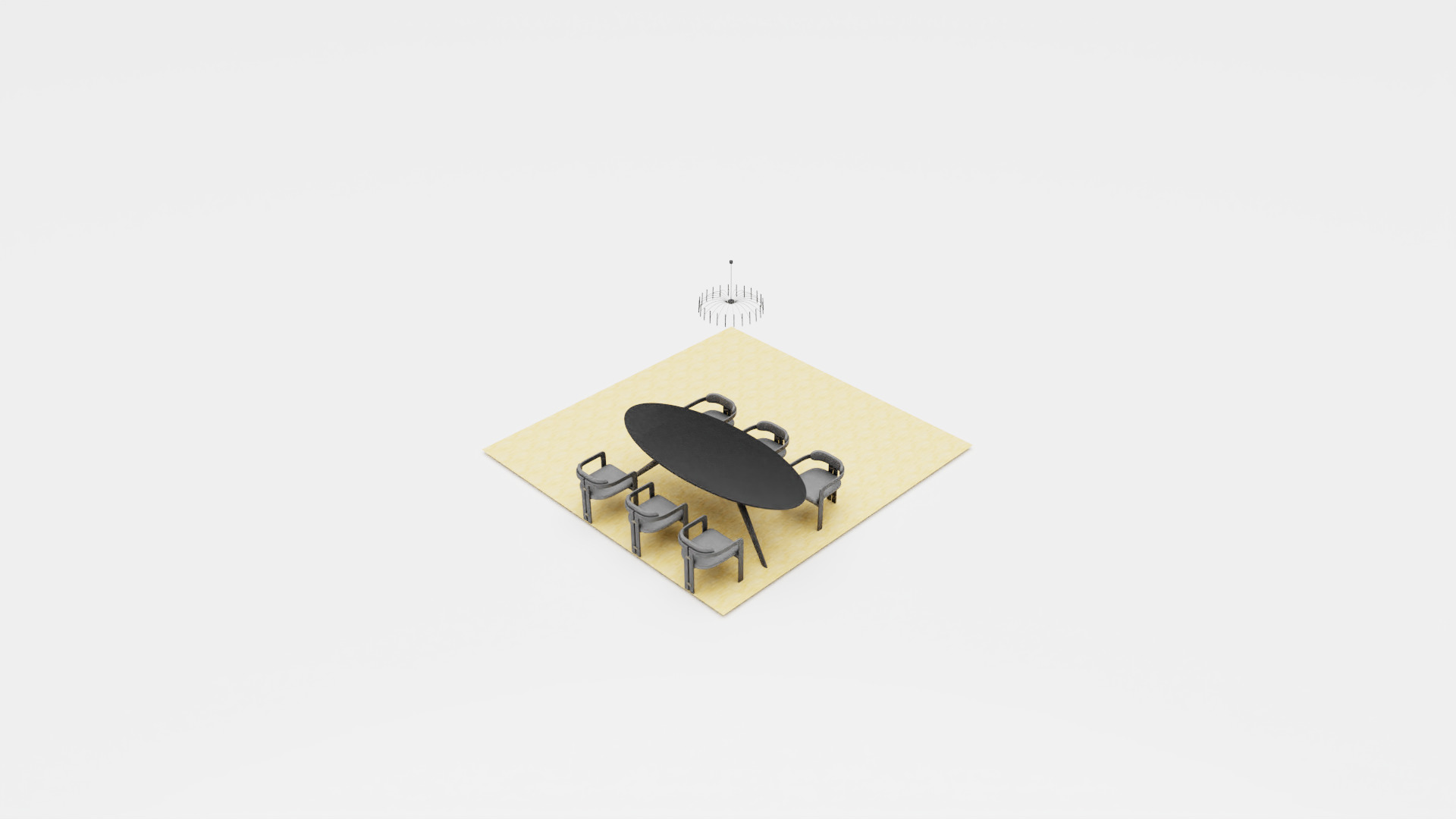}
    \end{subfigure}%
        \begin{subfigure}[b]{0.20\linewidth}
		\centering
		\includegraphics[width=\linewidth, trim=500 200 500 50, clip]{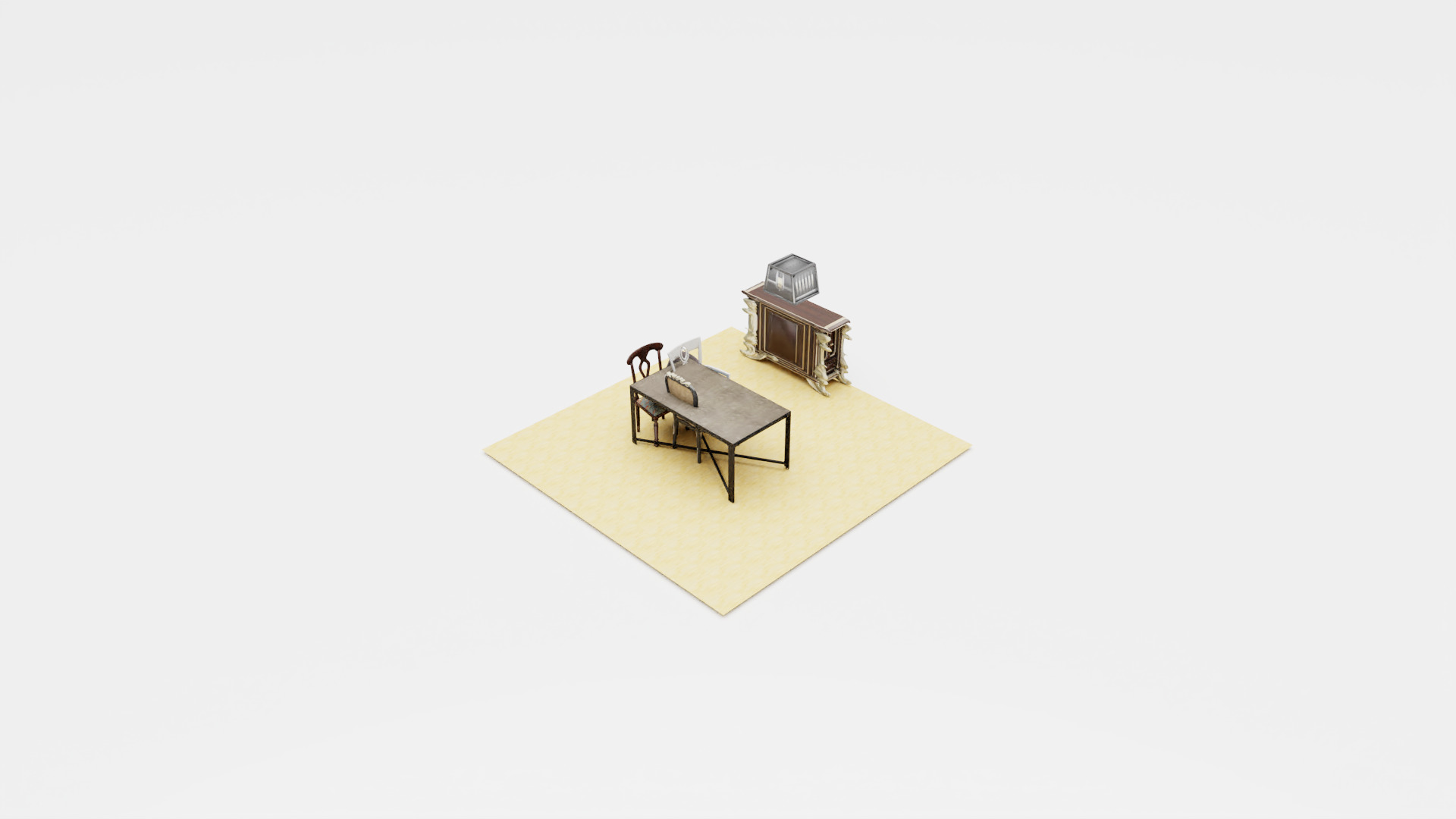}
    \end{subfigure}%
        \begin{subfigure}[b]{0.20\linewidth}
		\centering
		\includegraphics[width=\linewidth, trim=500 200 500 50, clip]{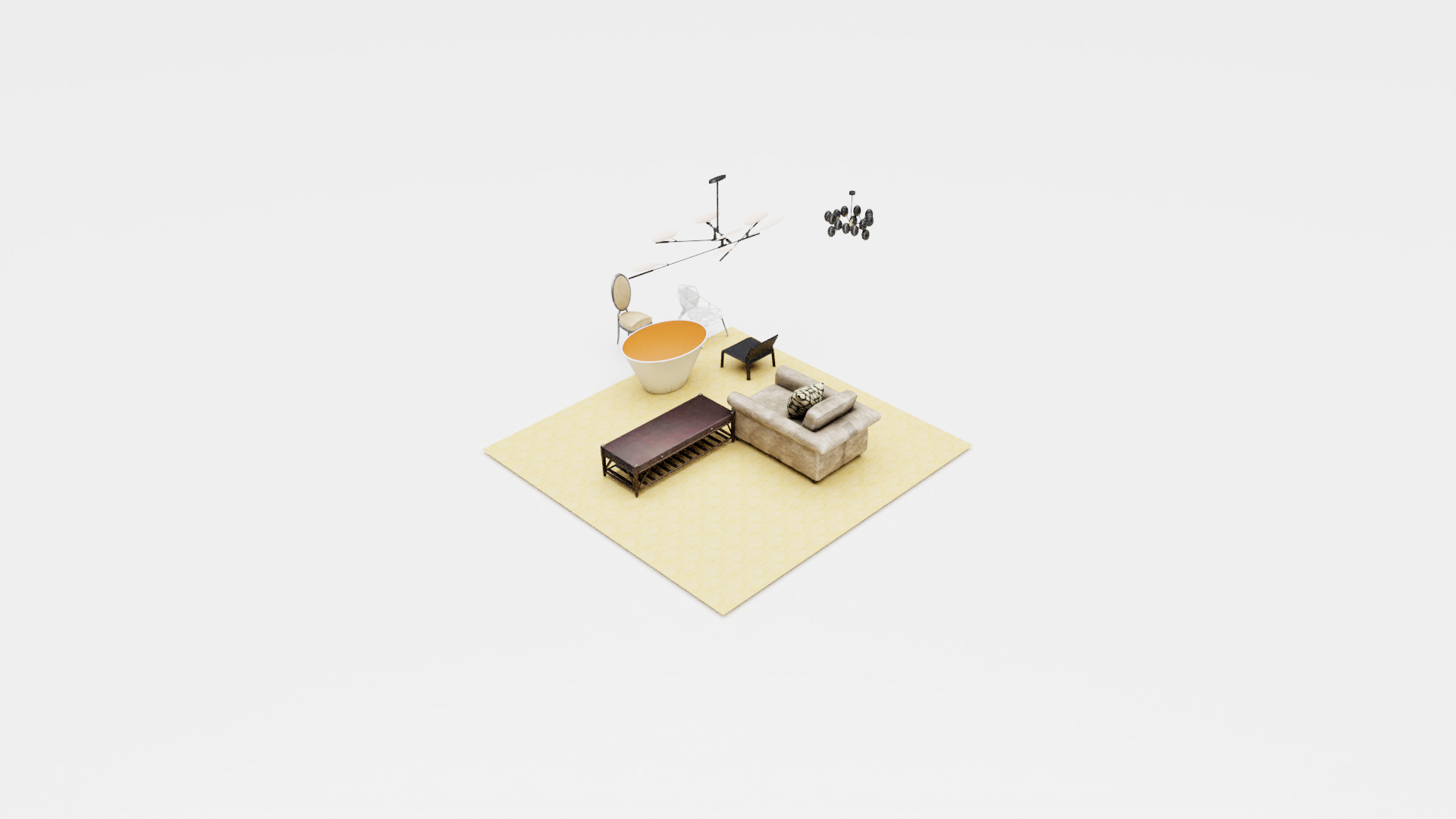}
    \end{subfigure}%
    \begin{subfigure}[b]{0.20\linewidth}
		\centering
		\includegraphics[width=\linewidth, trim=500 200 500 50, clip]{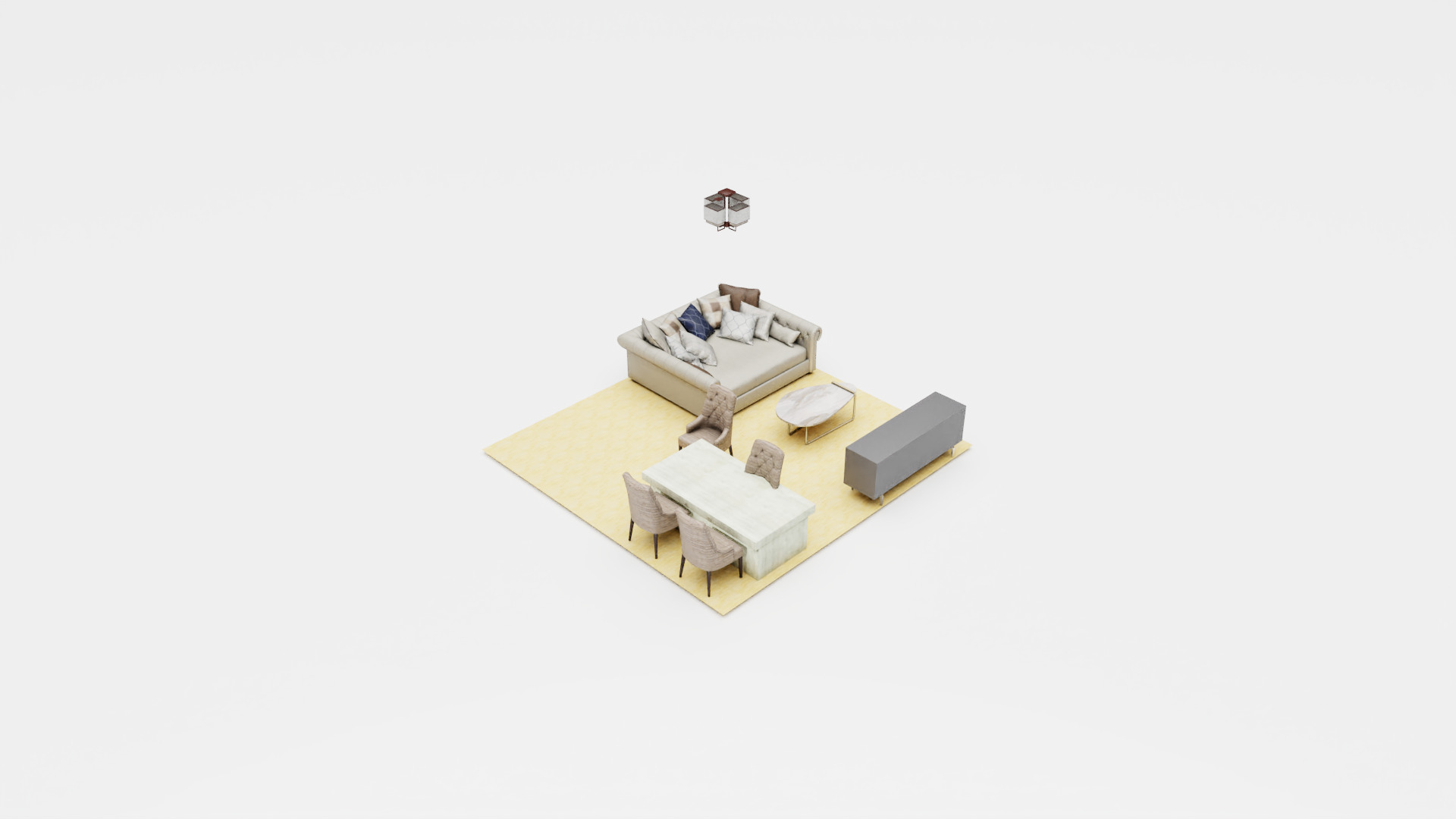}
    \end{subfigure}%
    \vskip\baselineskip%
    \vspace{-2.2em}
    \vskip\baselineskip%
    \begin{subfigure}[b]{0.20\linewidth}
		\centering
		\includegraphics[width=0.8\linewidth]{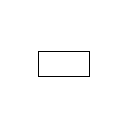}
    \end{subfigure}%
        \begin{subfigure}[b]{0.20\linewidth}
		\centering
		\includegraphics[width=\linewidth, trim=500 200 500 50, clip]{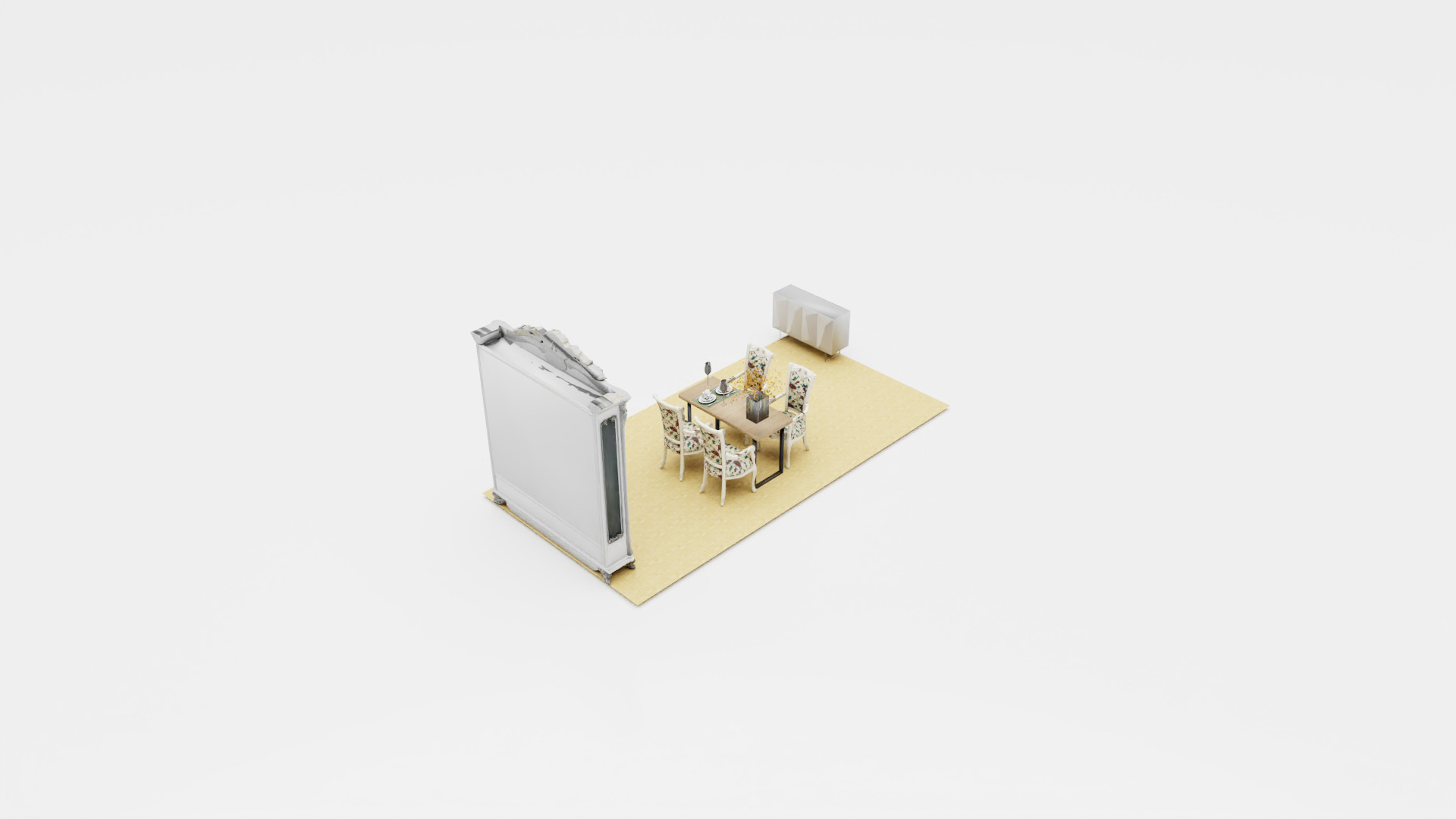}
    \end{subfigure}%
        \begin{subfigure}[b]{0.20\linewidth}
		\centering
		\includegraphics[width=\linewidth, trim=500 200 500 50, clip]{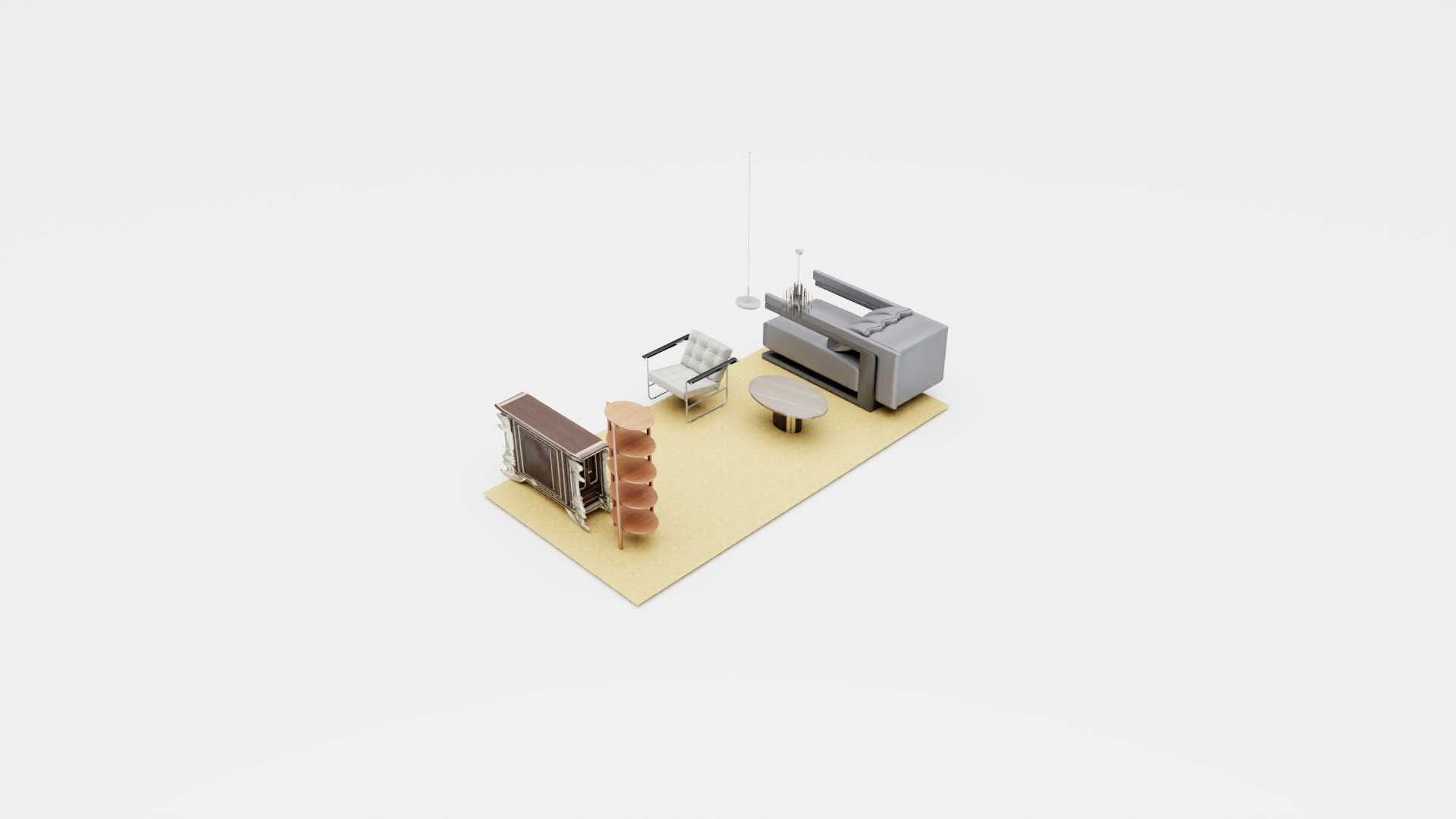}
    \end{subfigure}%
        \begin{subfigure}[b]{0.20\linewidth}
		\centering
		\includegraphics[width=\linewidth, trim=500 200 500 50, clip]{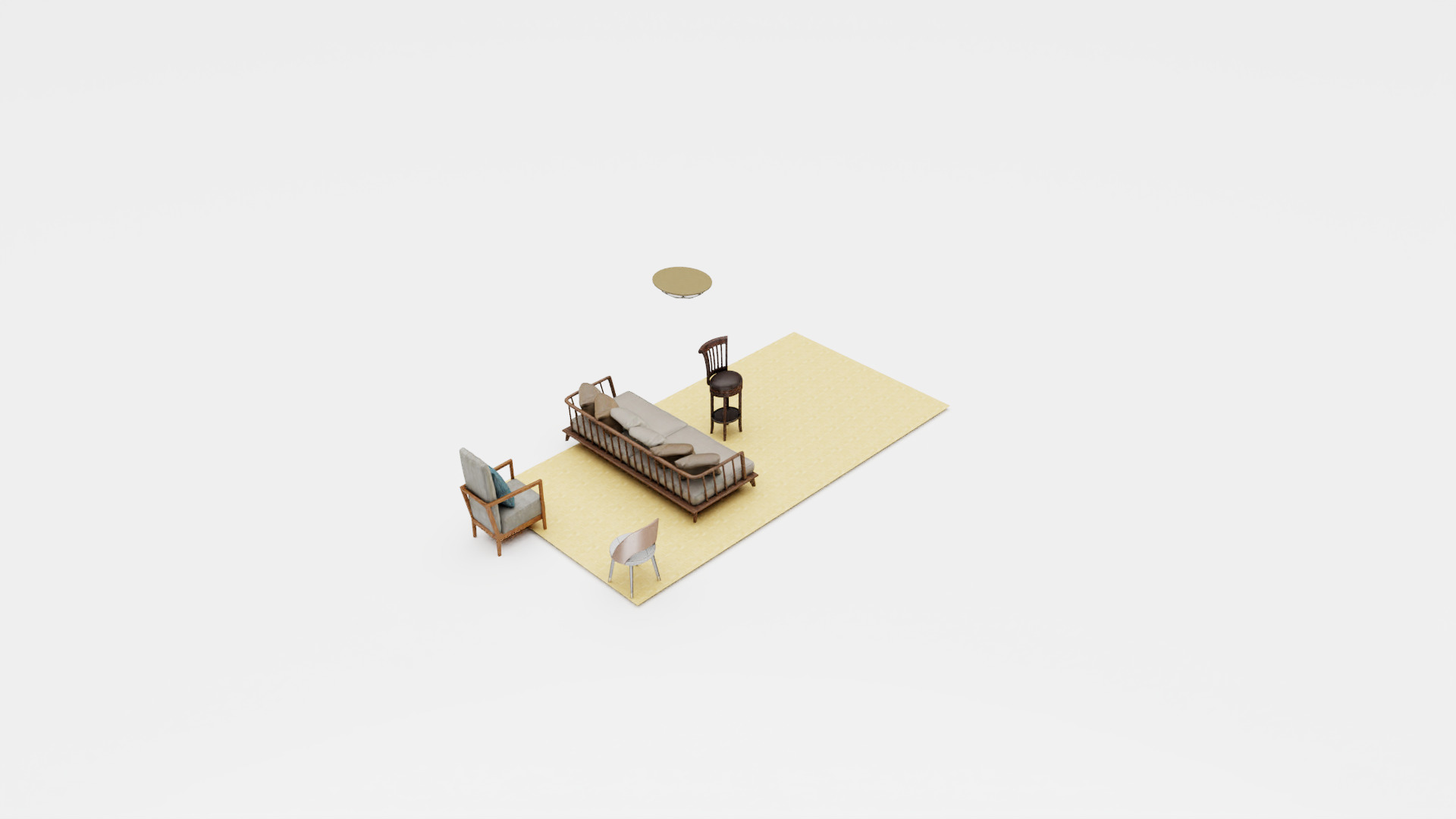}
    \end{subfigure}%
    \begin{subfigure}[b]{0.20\linewidth}
		\centering
		\includegraphics[width=\linewidth, trim=500 200 500 50, clip]{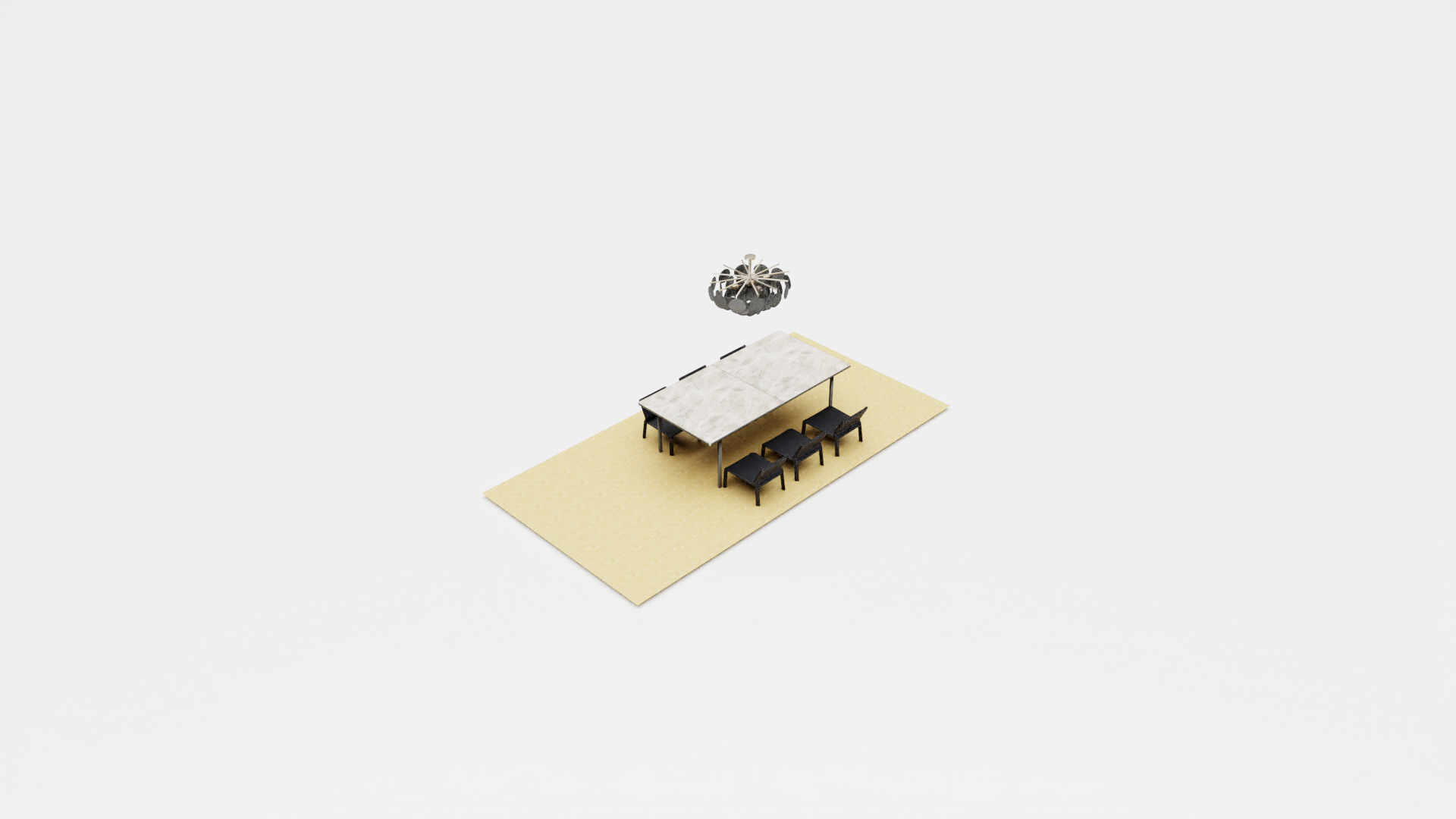}
    \end{subfigure}%
    \vskip\baselineskip%
    \vspace{-2.2em}
    \vskip\baselineskip%
    \begin{subfigure}[b]{0.20\linewidth}
		\centering
		\includegraphics[width=0.8\linewidth, trim=0 20 0 20, clip]{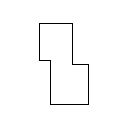}
    \end{subfigure}%
        \begin{subfigure}[b]{0.20\linewidth}
		\centering
		\includegraphics[width=\linewidth, trim=300 50 300 100, clip]{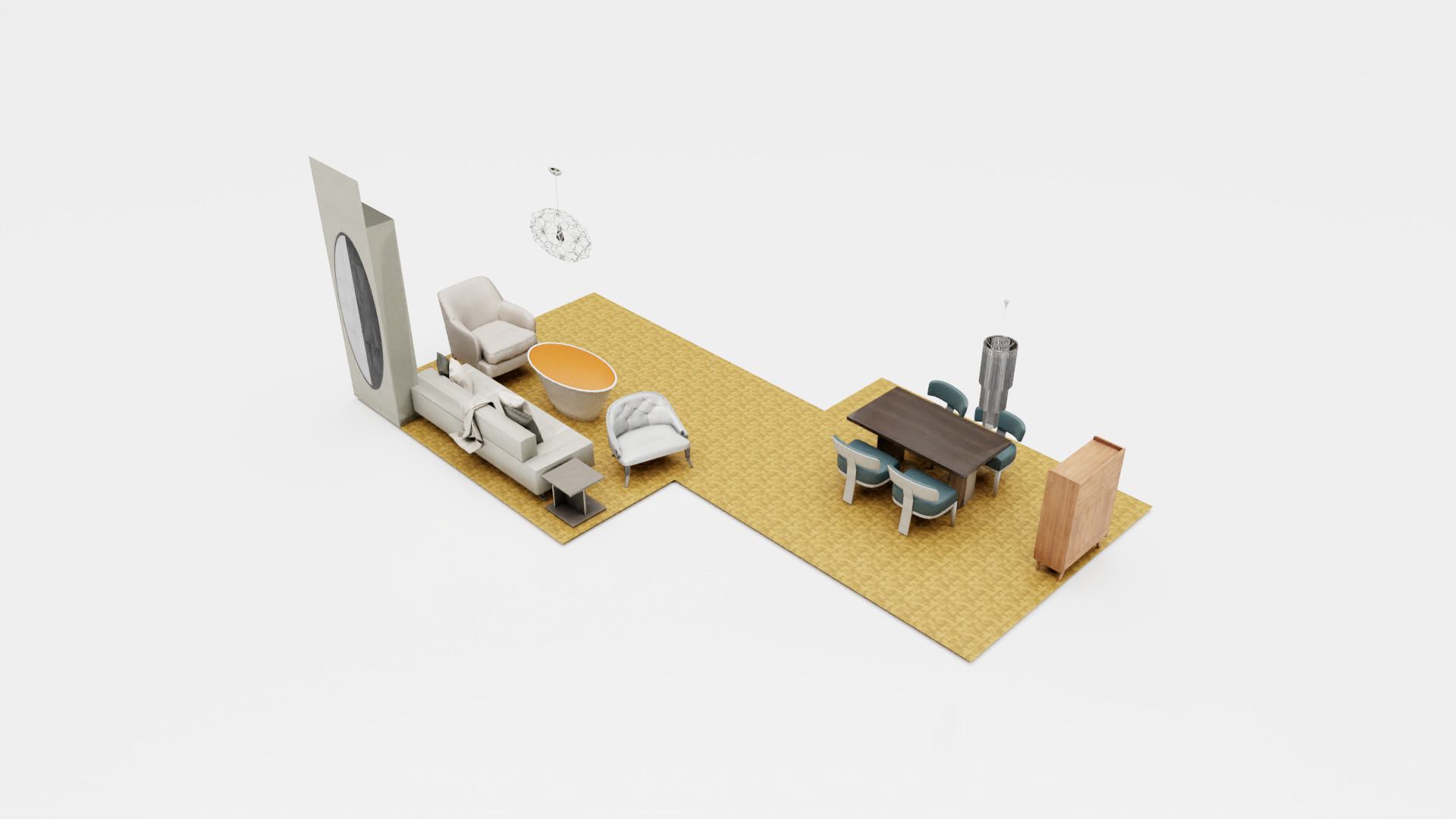}
    \end{subfigure}%
        \begin{subfigure}[b]{0.20\linewidth}
		\centering
		\includegraphics[width=\linewidth, trim=300 50 300 100, clip]{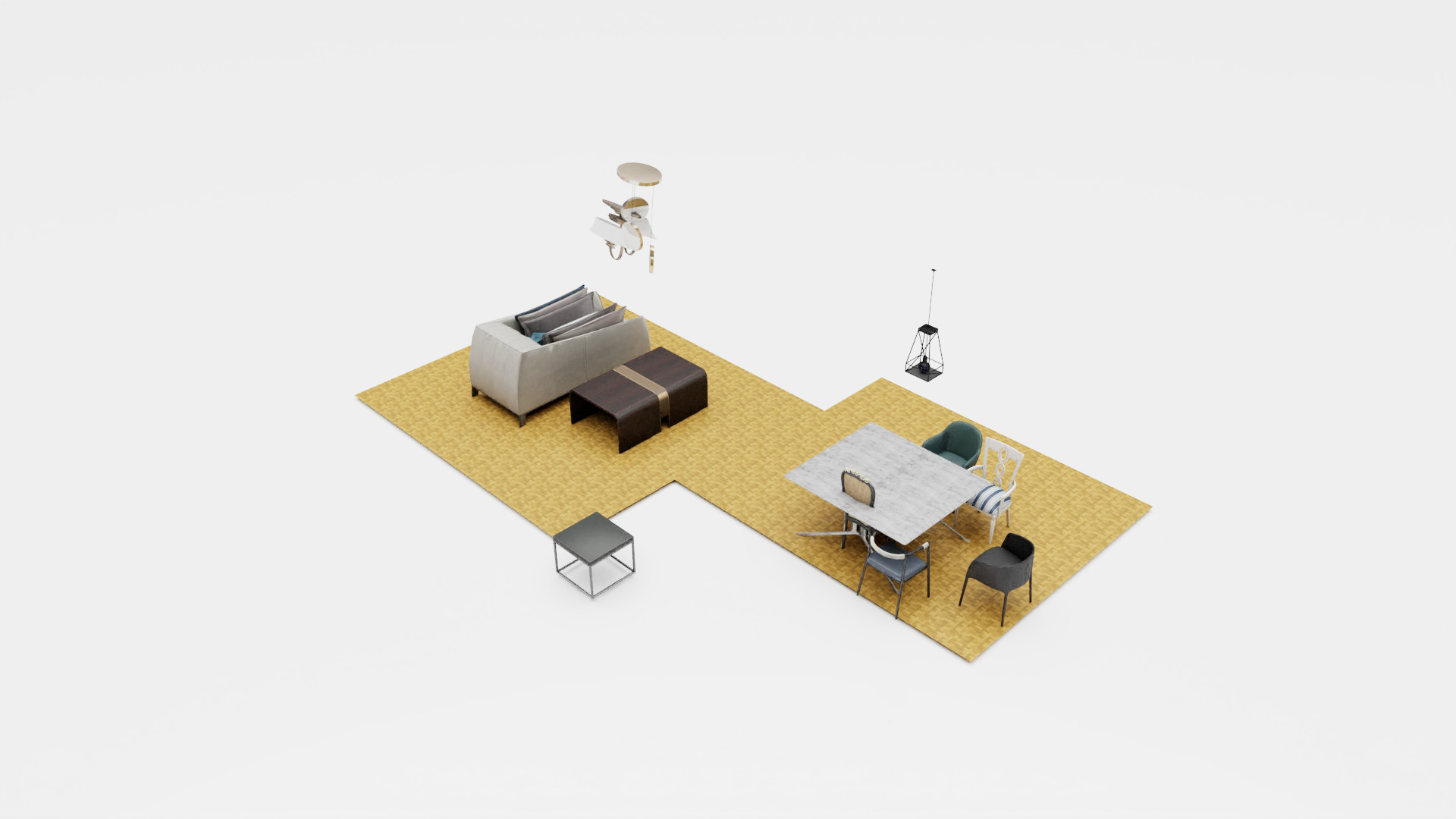}
    \end{subfigure}%
        \begin{subfigure}[b]{0.20\linewidth}
		\centering
		\includegraphics[width=\linewidth, trim=300 50 300 100, clip]{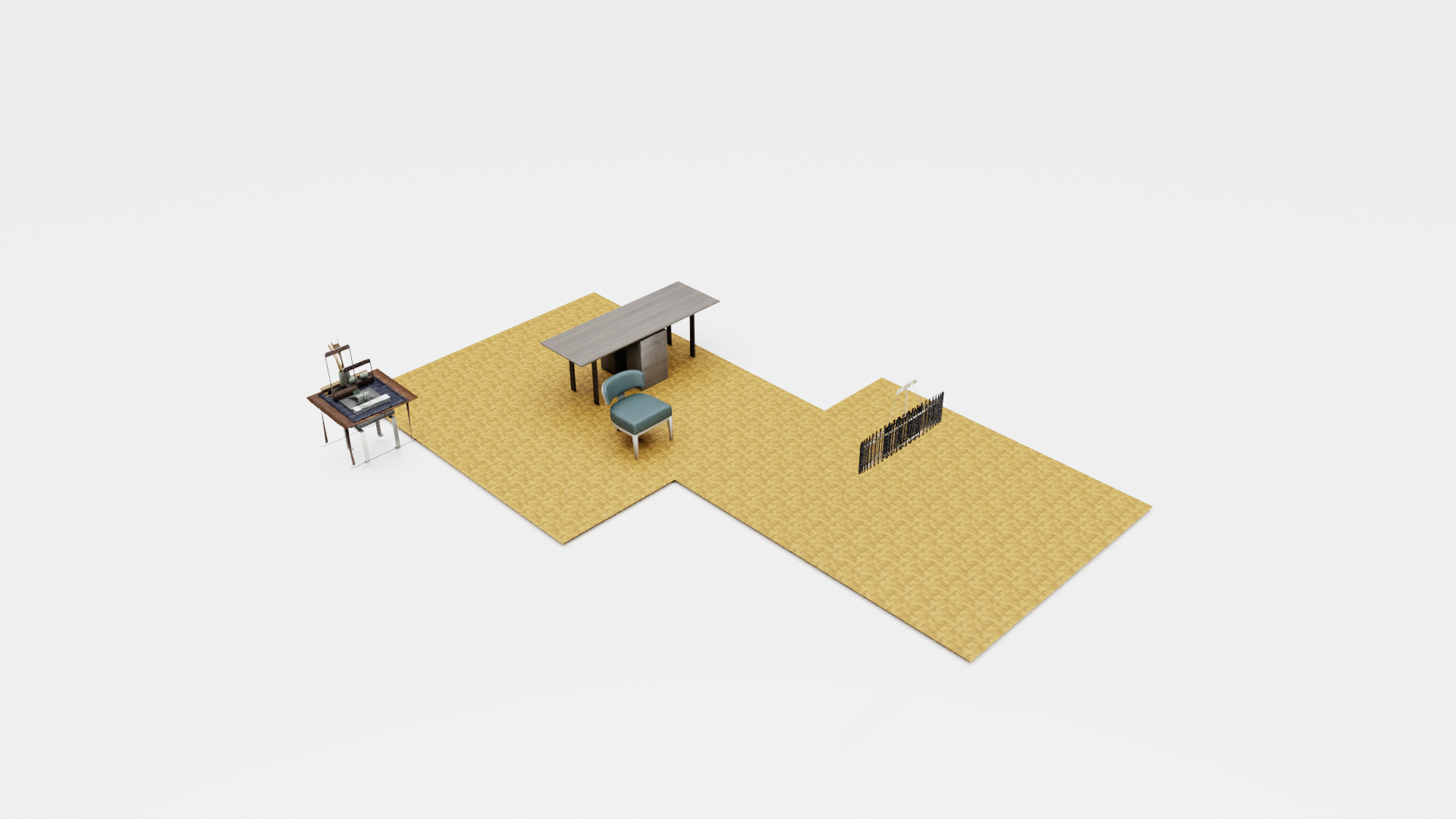}
    \end{subfigure}%
    \begin{subfigure}[b]{0.20\linewidth}
		\centering
		\includegraphics[width=\linewidth, trim=300 50 300 100, clip]{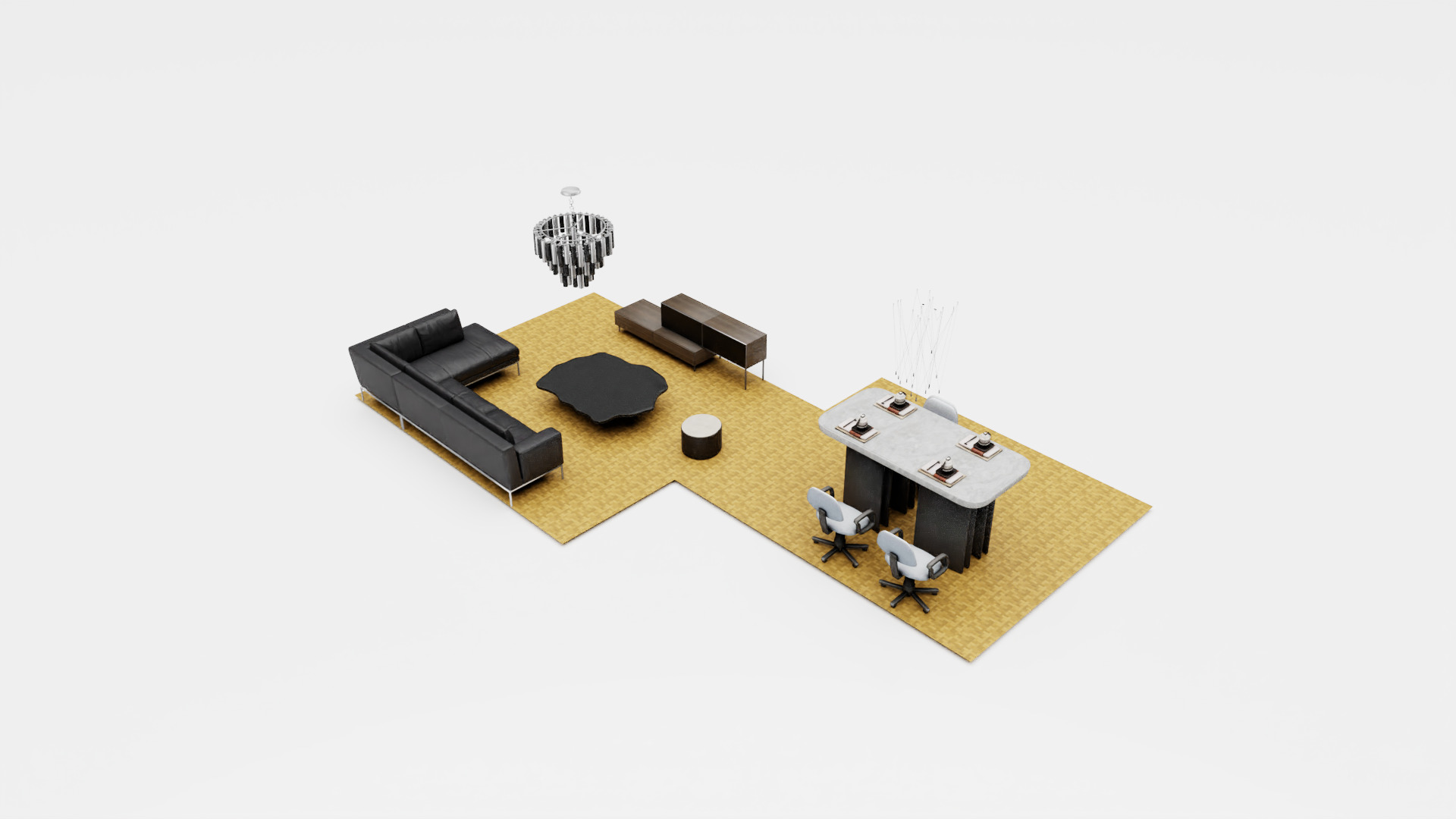}
    \end{subfigure}%
    \vskip\baselineskip%
    \vspace{-2.2em}
    \vskip\baselineskip%
    \begin{subfigure}[b]{0.20\linewidth}
		\centering
		\includegraphics[width=0.8\linewidth, trim=0 20 0 20, clip]{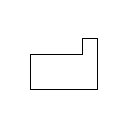}
    \end{subfigure}%
        \begin{subfigure}[b]{0.20\linewidth}
		\centering
		\includegraphics[width=\linewidth, trim=300 50 300 100, clip]{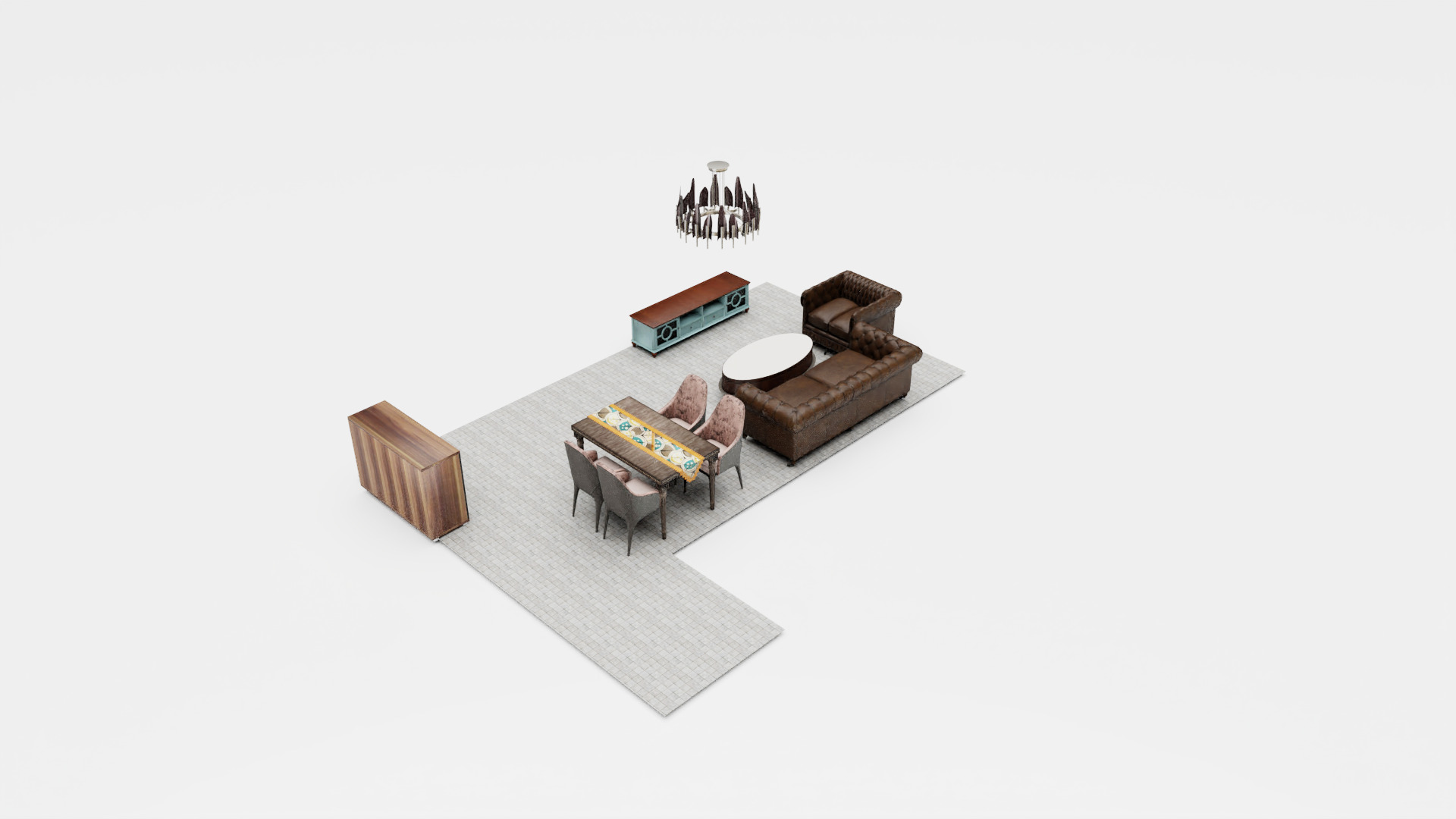}
    \end{subfigure}%
        \begin{subfigure}[b]{0.20\linewidth}
		\centering
		\includegraphics[width=\linewidth, trim=300 50 300 100, clip]{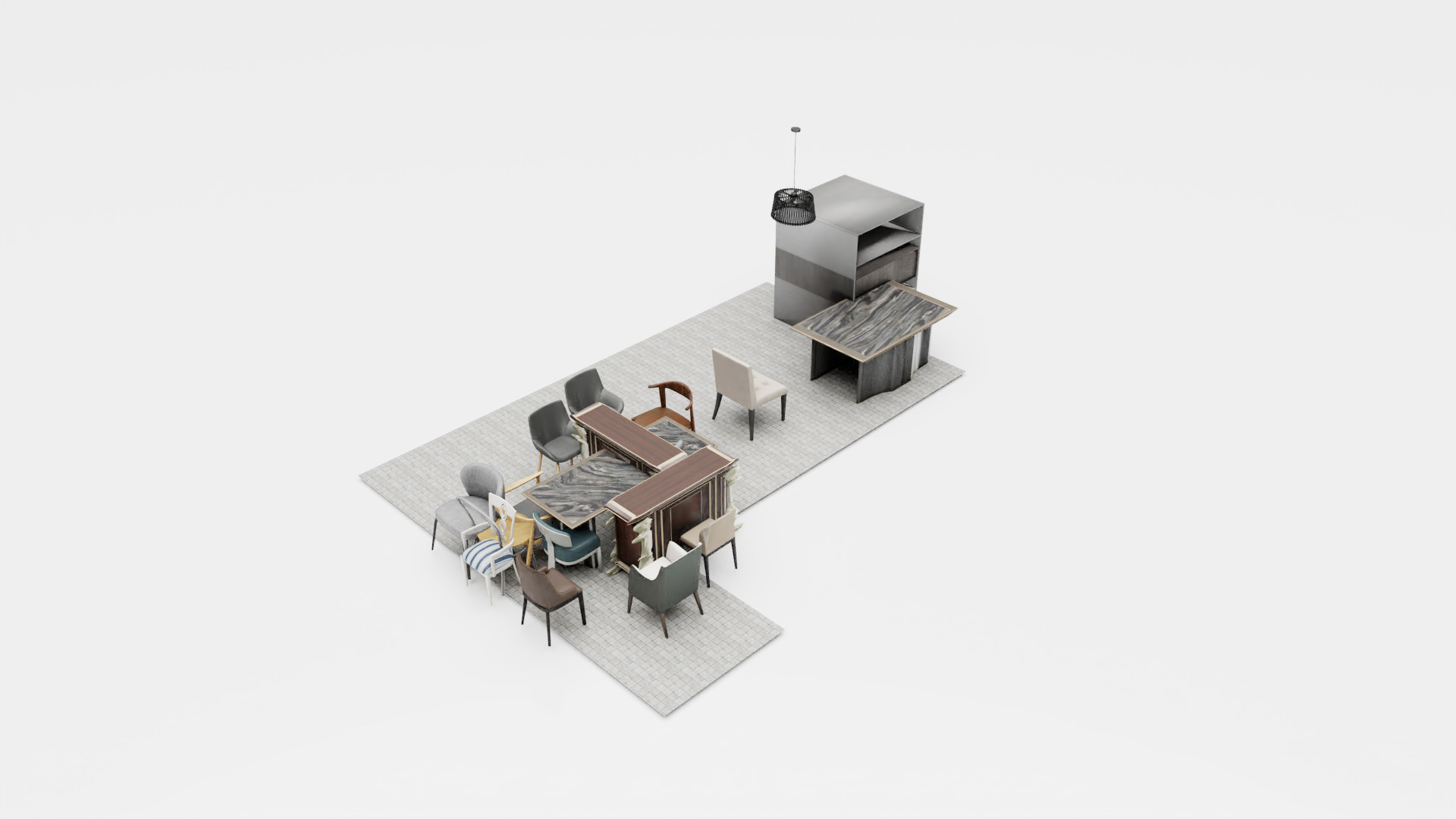}
    \end{subfigure}%
        \begin{subfigure}[b]{0.20\linewidth}
		\centering
		\includegraphics[width=\linewidth, trim=300 50 300 100, clip]{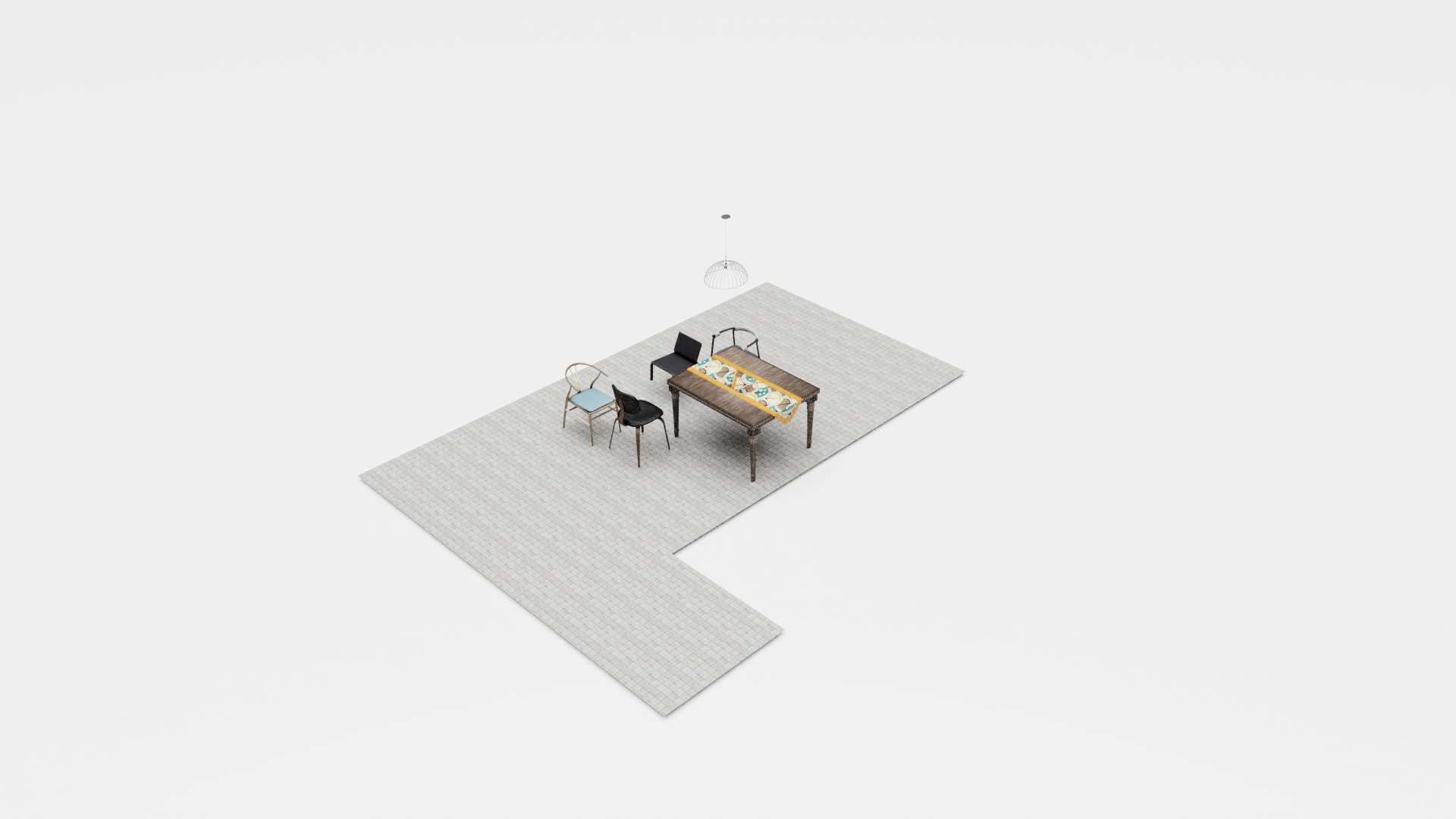}
    \end{subfigure}%
    \begin{subfigure}[b]{0.20\linewidth}
		\centering
		\includegraphics[width=\linewidth, trim=300 50 300 100, clip]{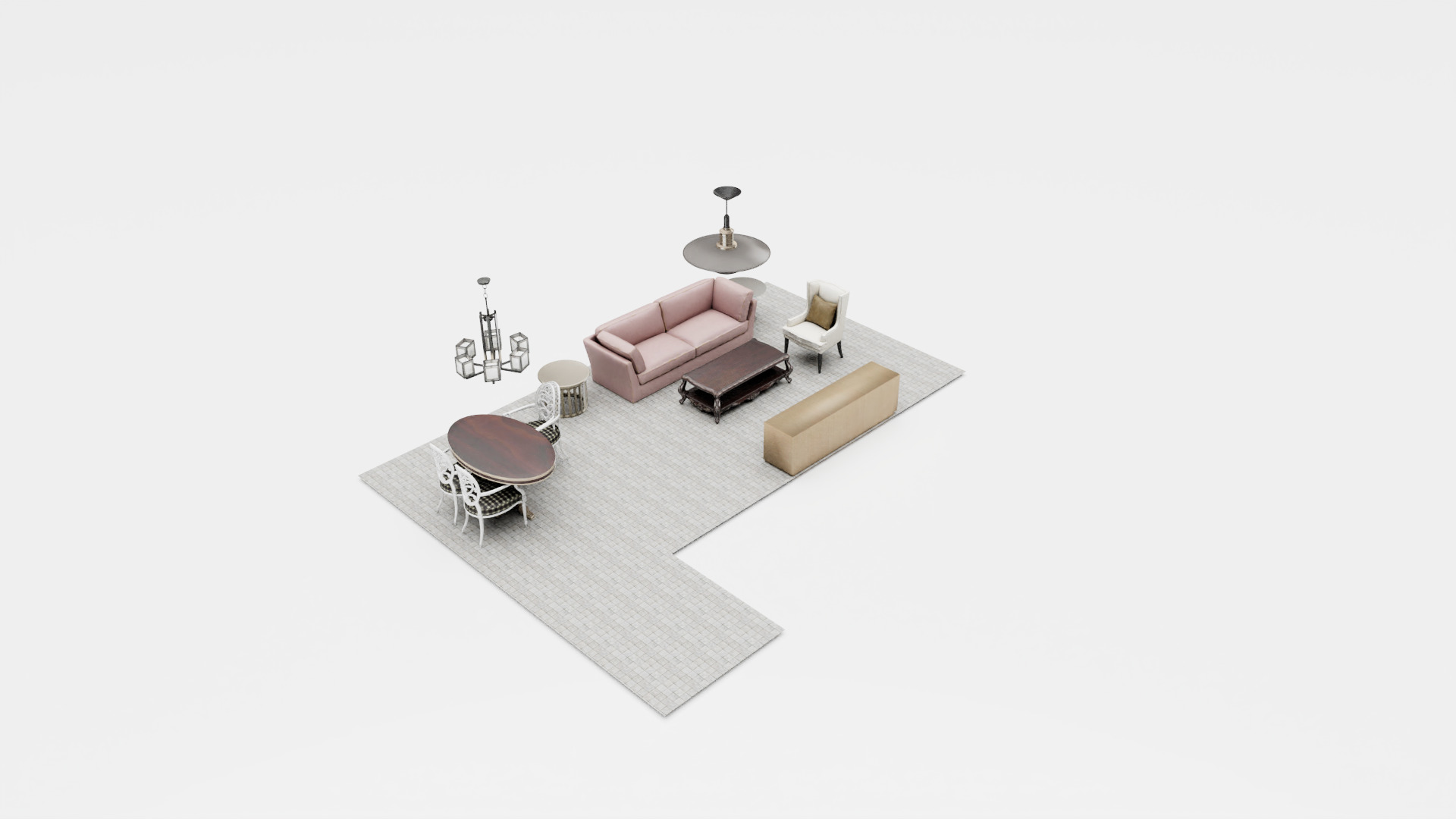}
    \end{subfigure}%
    \vskip\baselineskip%
    \vspace{-2.2em}
    \vskip\baselineskip%
    \begin{subfigure}[b]{0.20\linewidth}
		\centering
		\includegraphics[width=0.8\linewidth, trim=0 20 0 20, clip]{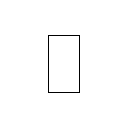}
    \end{subfigure}%
        \begin{subfigure}[b]{0.20\linewidth}
		\centering
		\includegraphics[width=\linewidth, trim=300 50 300 100, clip]{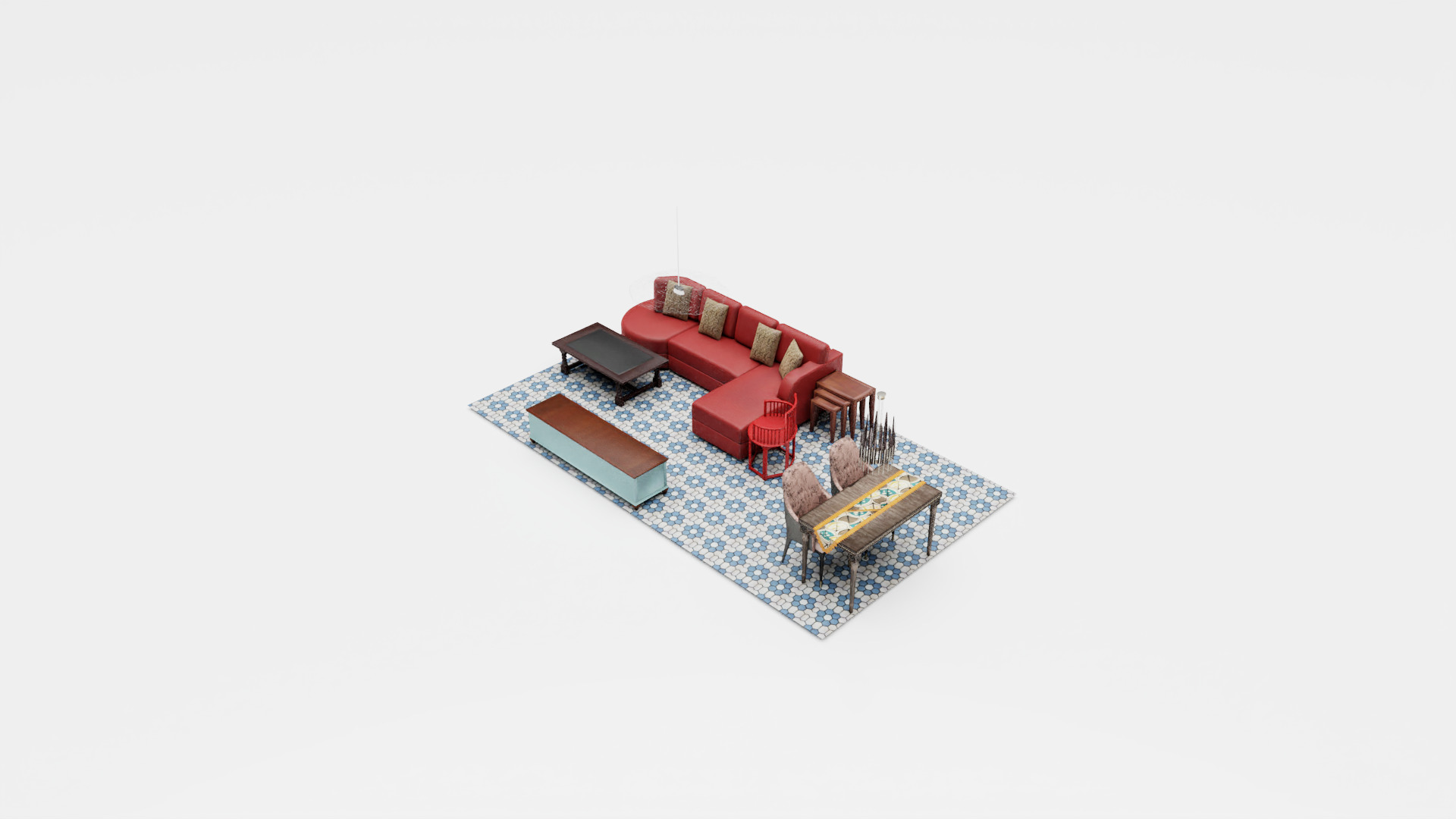}
    \end{subfigure}%
        \begin{subfigure}[b]{0.20\linewidth}
		\centering
		\includegraphics[width=\linewidth, trim=300 50 300 100, clip]{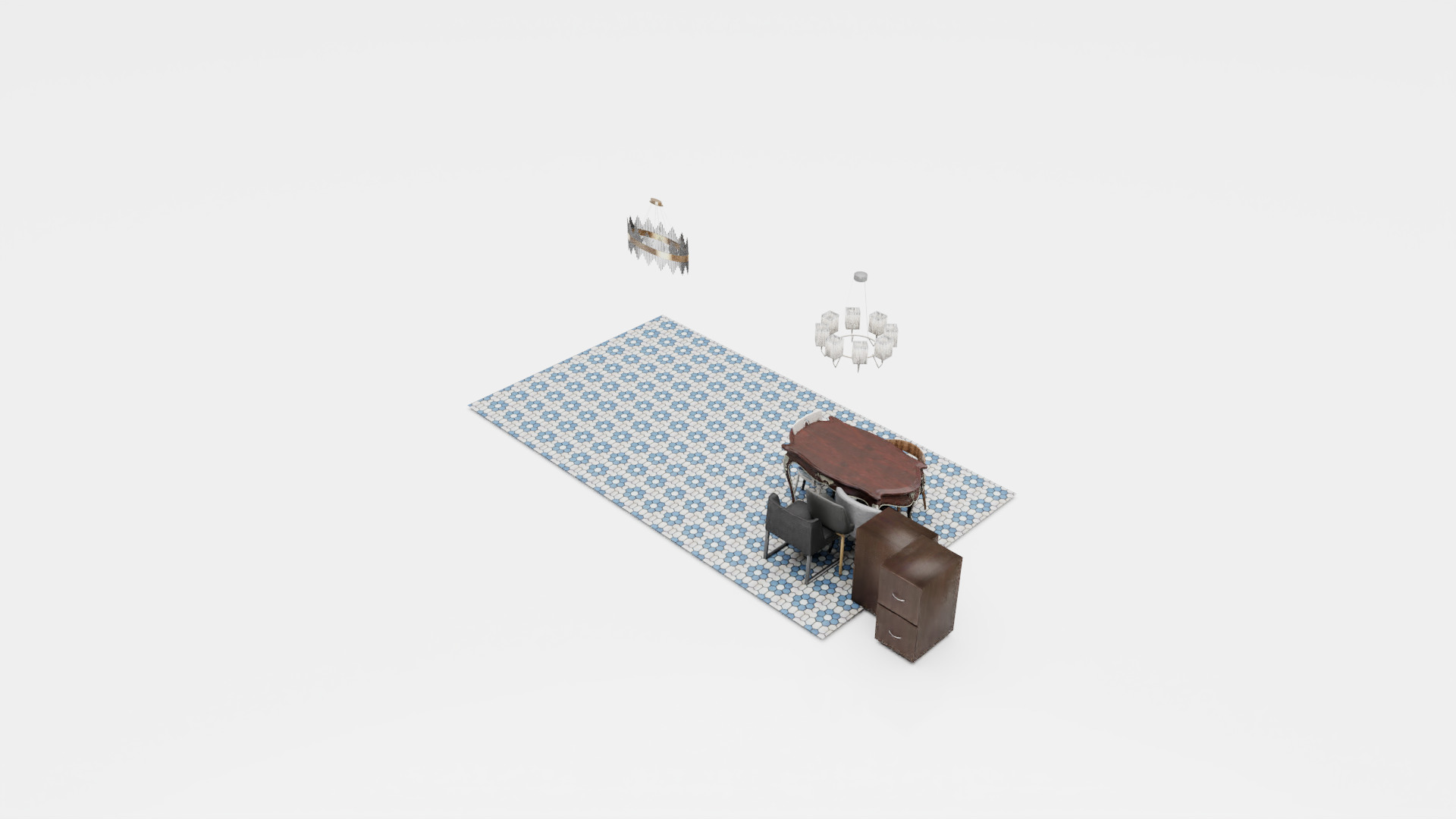}
    \end{subfigure}%
        \begin{subfigure}[b]{0.20\linewidth}
		\centering
		\includegraphics[width=\linewidth, trim=300 50 300 100, clip]{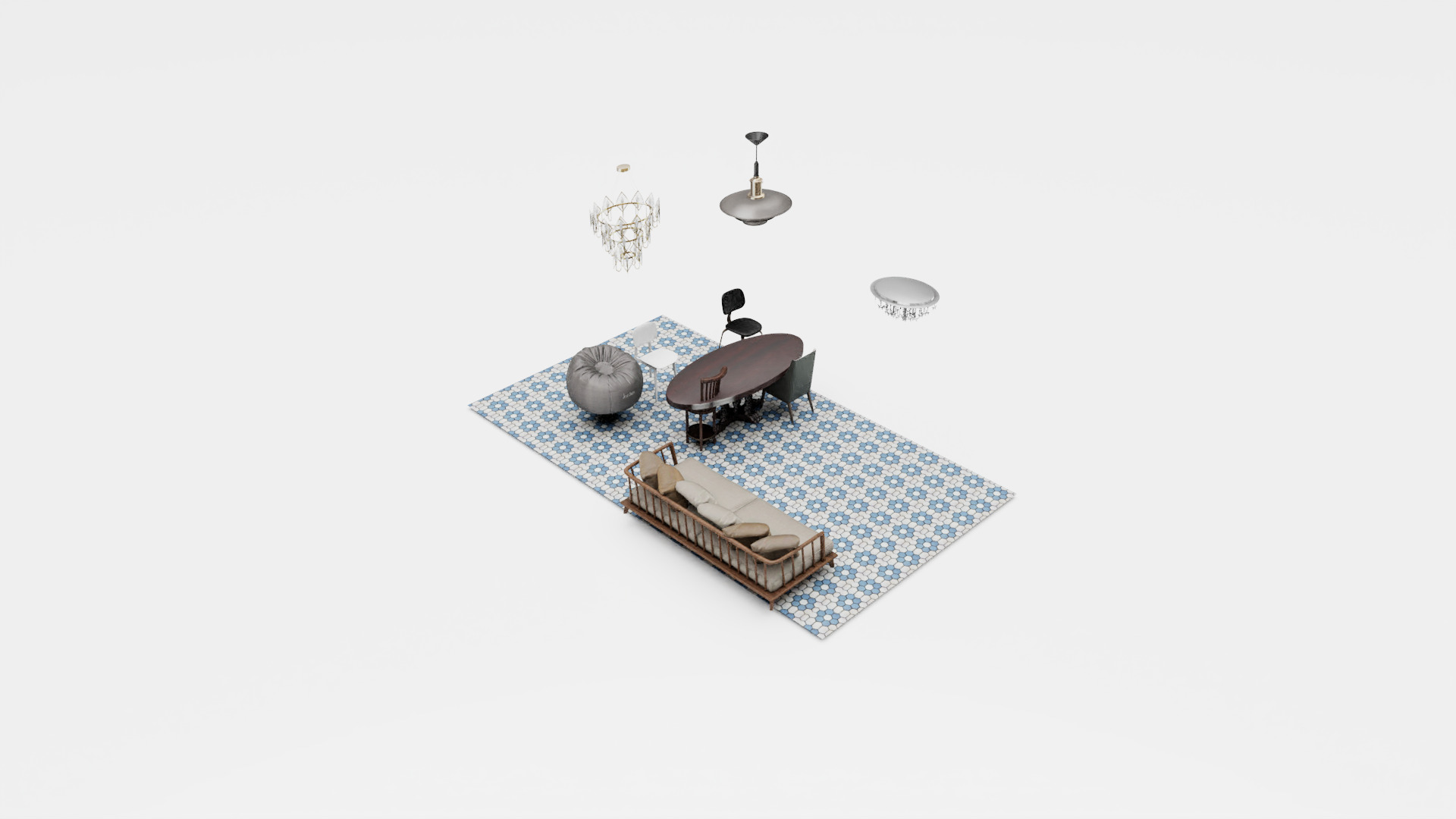}
    \end{subfigure}%
    \begin{subfigure}[b]{0.20\linewidth}
		\centering
		\includegraphics[width=\linewidth, trim=300 50 300 100, clip]{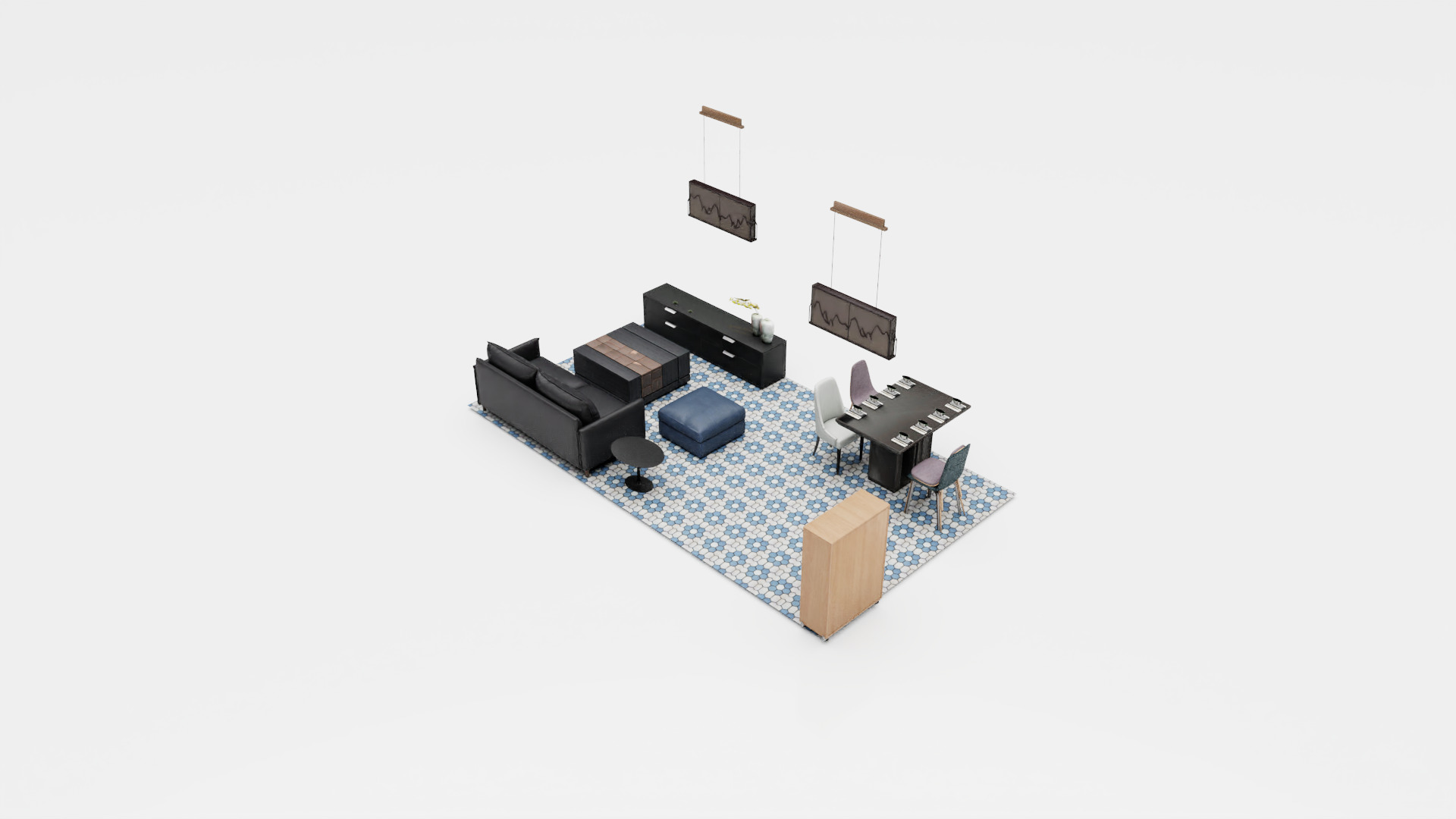}
    \end{subfigure}%
    \vskip\baselineskip%
    \vspace{-1.2em}
    \vskip\baselineskip%
    \hfill%
    \caption{{\bf Qualitative Scene Synthesis Results on Dining Rooms}.
    Generated scenes for dining rooms using FastSynth, SceneFormer and our method.
    To showcase the generalization abilities of our model we also show the
    closest scene from the training set (2nd column).}
    \label{fig:scene_synthesis_qualitative_diningroom_supp}
    \vspace{-1.2em}
\end{figure}

\begin{figure}[!h]
    \centering
    \begin{subfigure}[b]{0.20\linewidth}
		\centering
        \small Scene Layout
    \end{subfigure}%
    \begin{subfigure}[b]{0.20\linewidth}
		\centering
        \small Training Sample
    \end{subfigure}%
    \begin{subfigure}[b]{0.20\linewidth}
		\centering
        \small FastSynth
    \end{subfigure}%
    \begin{subfigure}[b]{0.20\linewidth}
		\centering
        \small SceneFormer
    \end{subfigure}%
    \begin{subfigure}[b]{0.20\linewidth}
        \centering
        \small Ours
    \end{subfigure}
    \hfill%
    \vskip\baselineskip%
    \vspace{-1.5em}
    \hfill%
    \begin{subfigure}[b]{0.20\linewidth}
		\centering
		\includegraphics[width=0.8\linewidth]{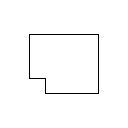}
    \end{subfigure}%
        \begin{subfigure}[b]{0.20\linewidth}
		\centering
		\includegraphics[width=\linewidth, trim=500 200 500 100, clip]{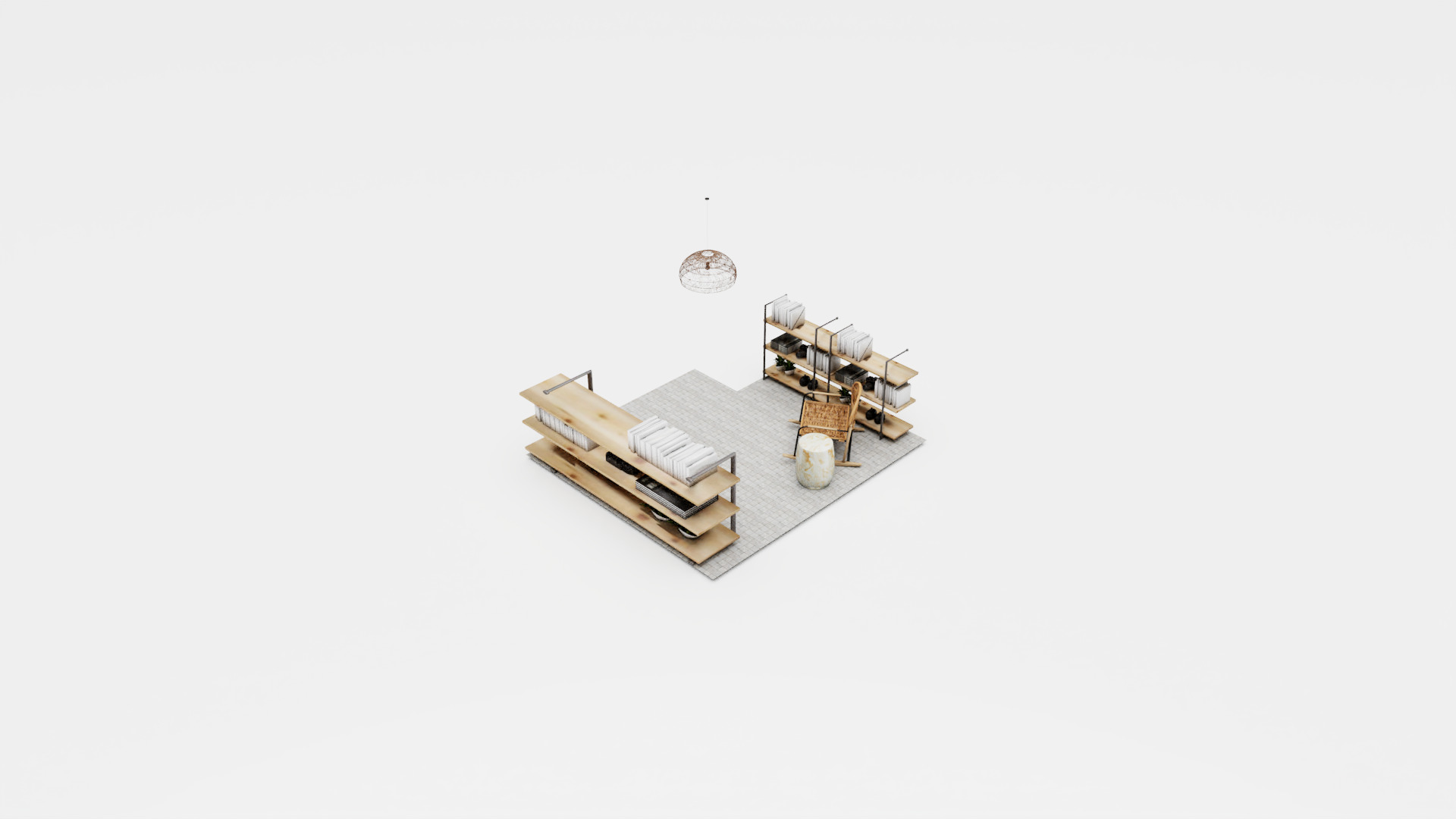}
    \end{subfigure}%
        \begin{subfigure}[b]{0.20\linewidth}
		\centering
		\includegraphics[width=\linewidth, trim=500 200 500 100, clip]{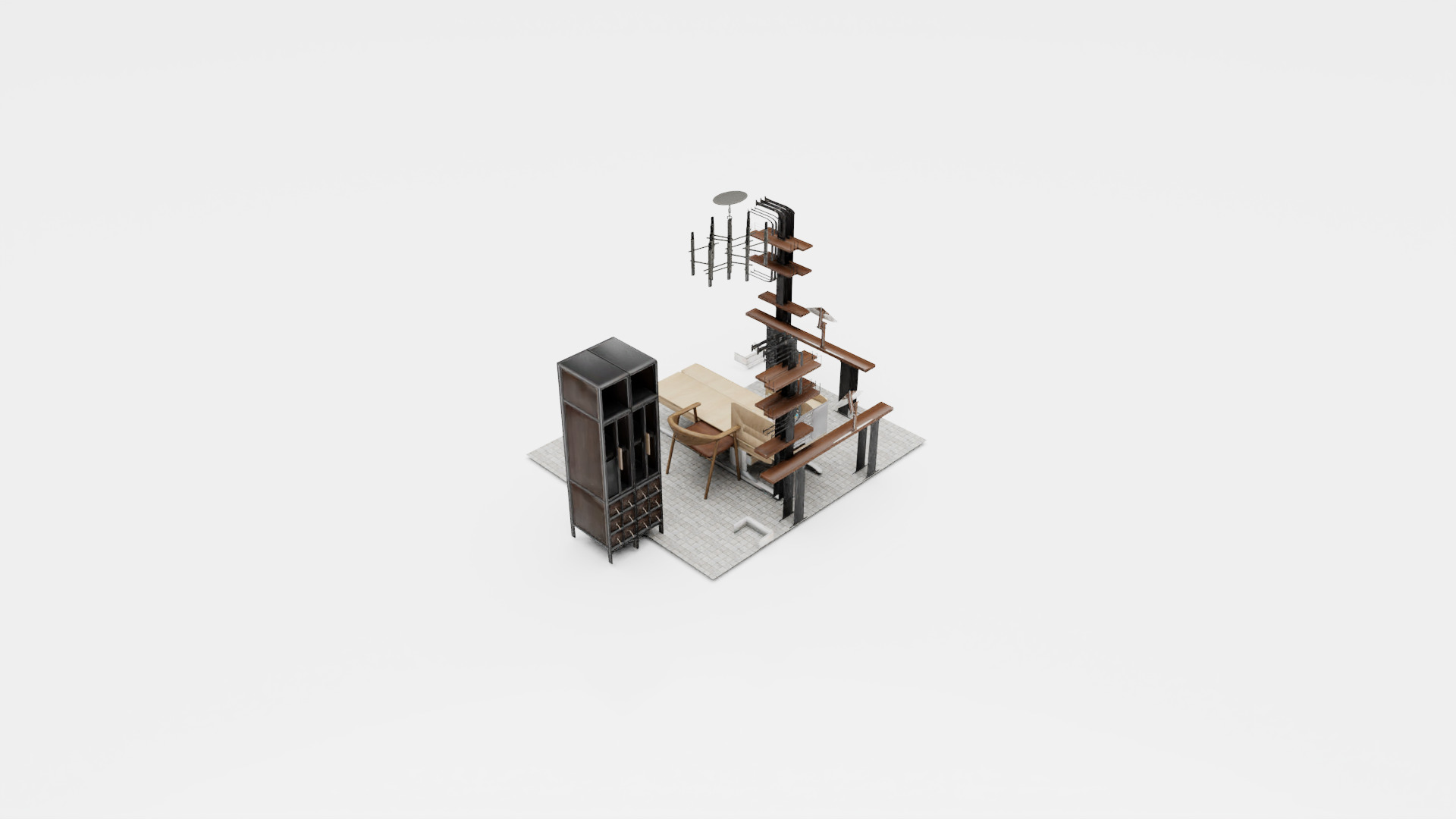}
    \end{subfigure}%
        \begin{subfigure}[b]{0.20\linewidth}
		\centering
		\includegraphics[width=\linewidth, trim=500 200 500 100, clip]{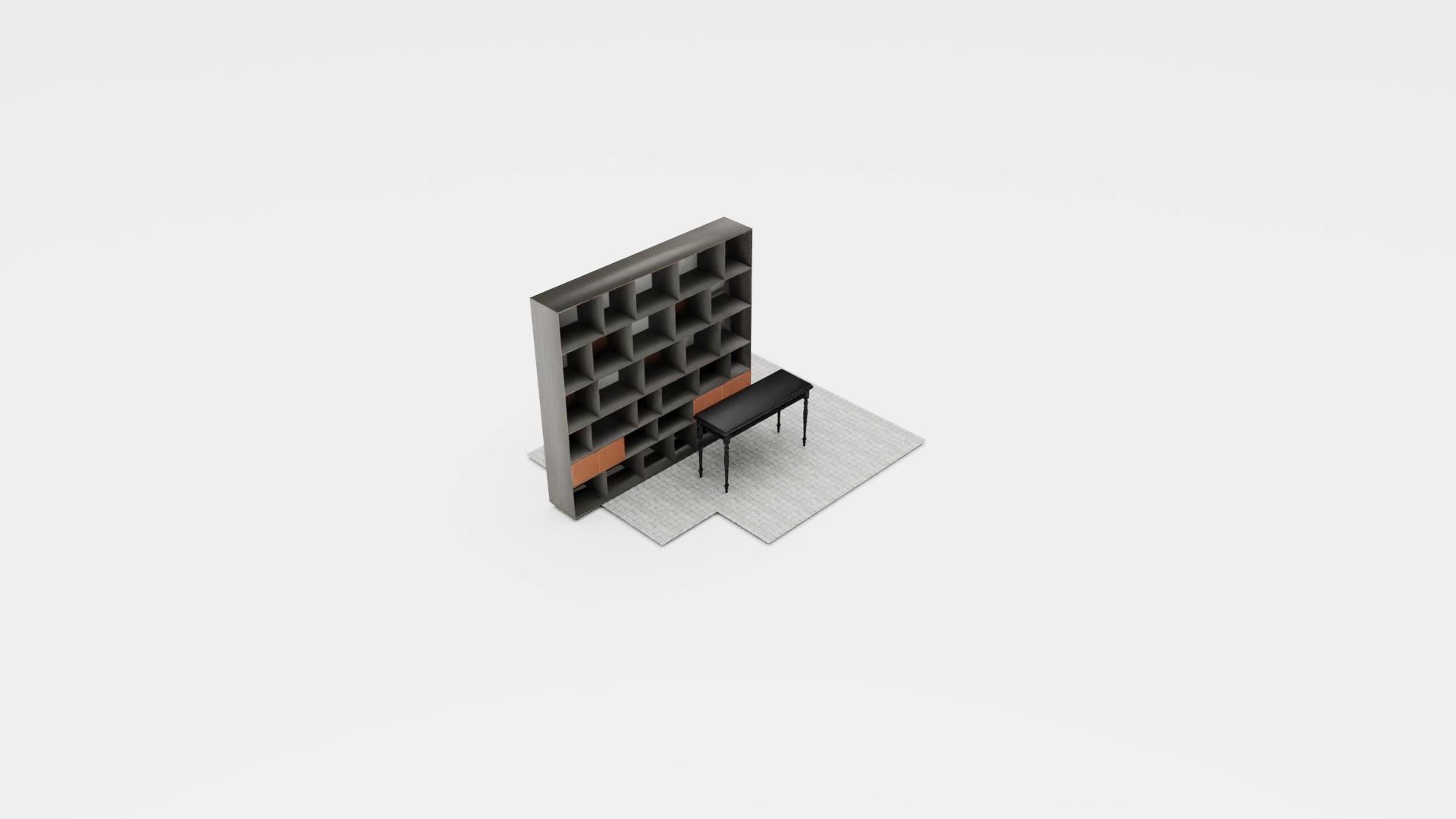}
    \end{subfigure}%
    \begin{subfigure}[b]{0.20\linewidth}
		\centering
		\includegraphics[width=\linewidth, trim=500 200 500 100, clip]{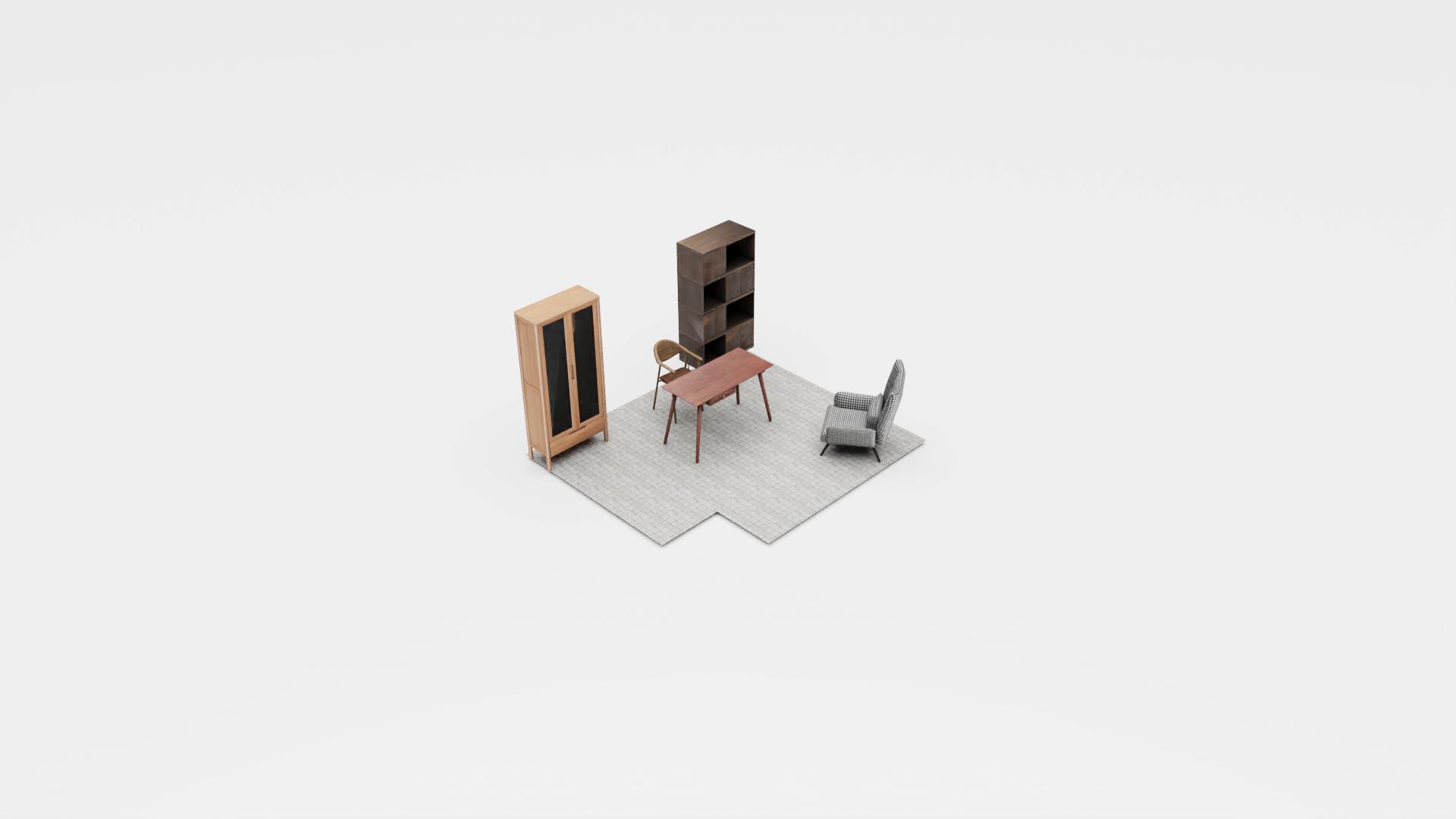}
    \end{subfigure}%
    \vskip\baselineskip%
    \vspace{-2.2em}
    \vskip\baselineskip%
    \begin{subfigure}[b]{0.20\linewidth}
		\centering
		\includegraphics[width=0.8\linewidth]{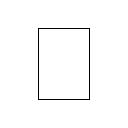}
    \end{subfigure}%
        \begin{subfigure}[b]{0.20\linewidth}
		\centering
		\includegraphics[width=\linewidth, trim=500 200 500 100, clip]{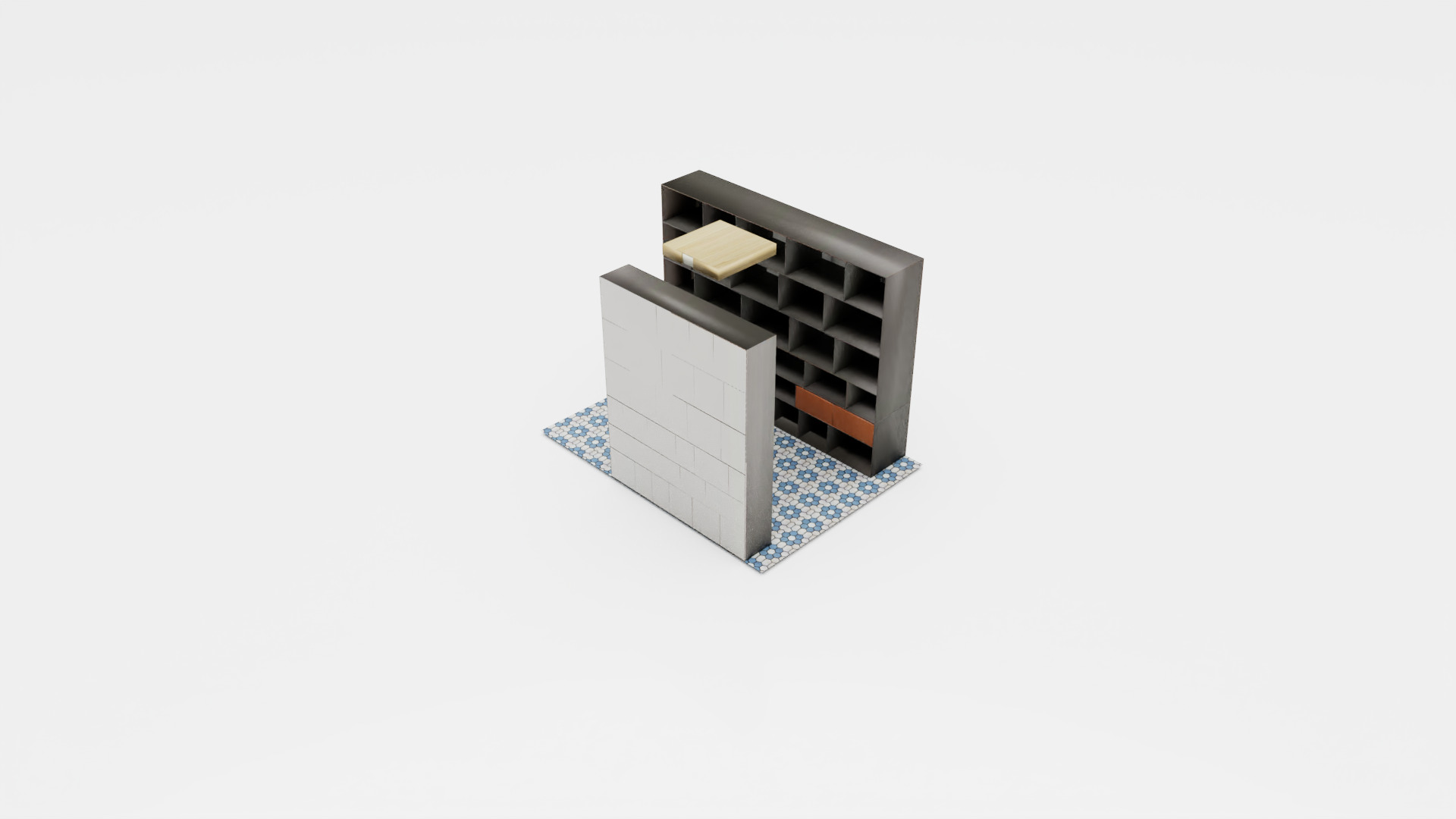}
    \end{subfigure}%
        \begin{subfigure}[b]{0.20\linewidth}
		\centering
		\includegraphics[width=\linewidth, trim=500 200 500 100, clip]{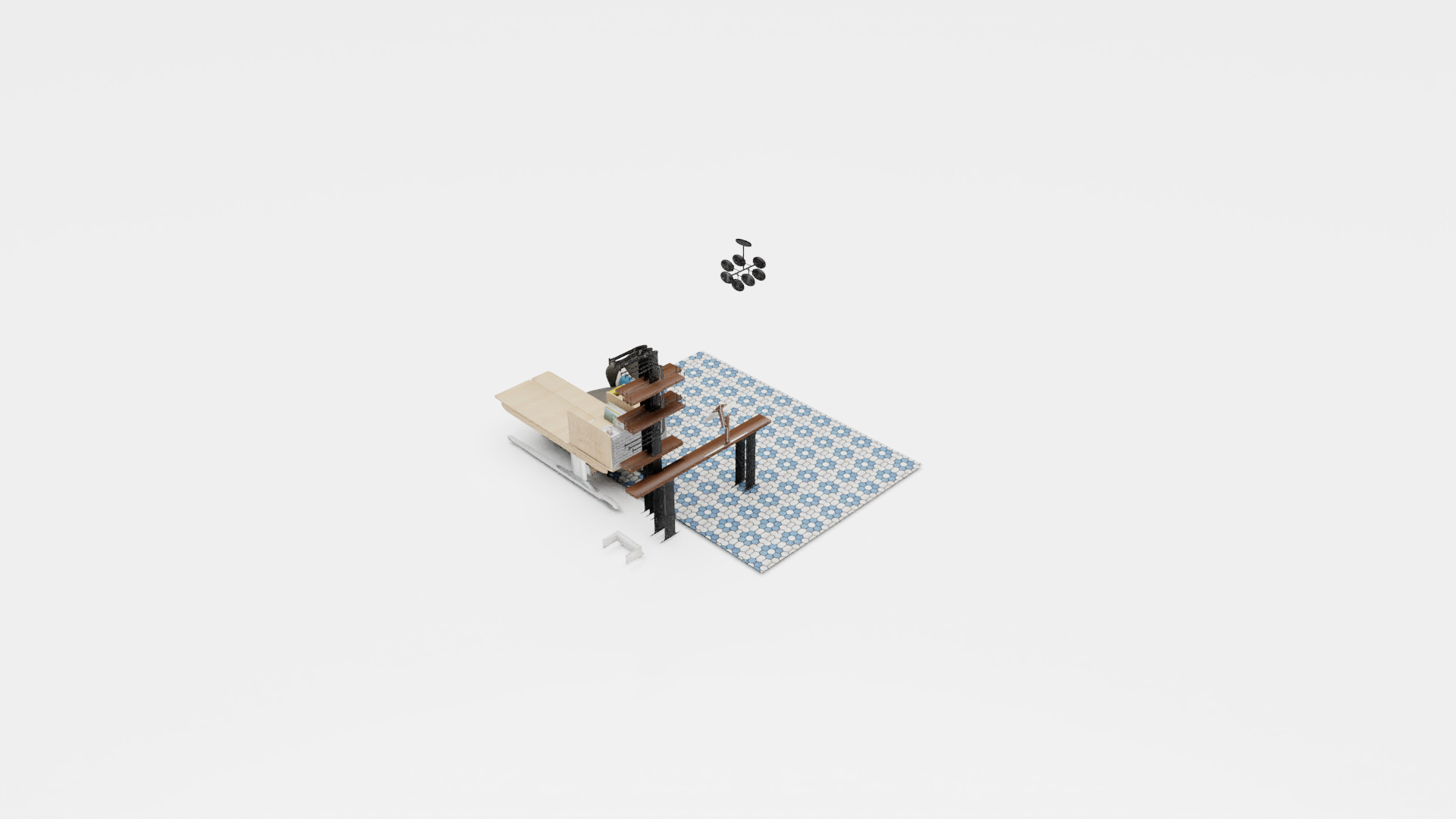}
    \end{subfigure}%
        \begin{subfigure}[b]{0.20\linewidth}
		\centering
		\includegraphics[width=\linewidth, trim=500 200 500 100, clip]{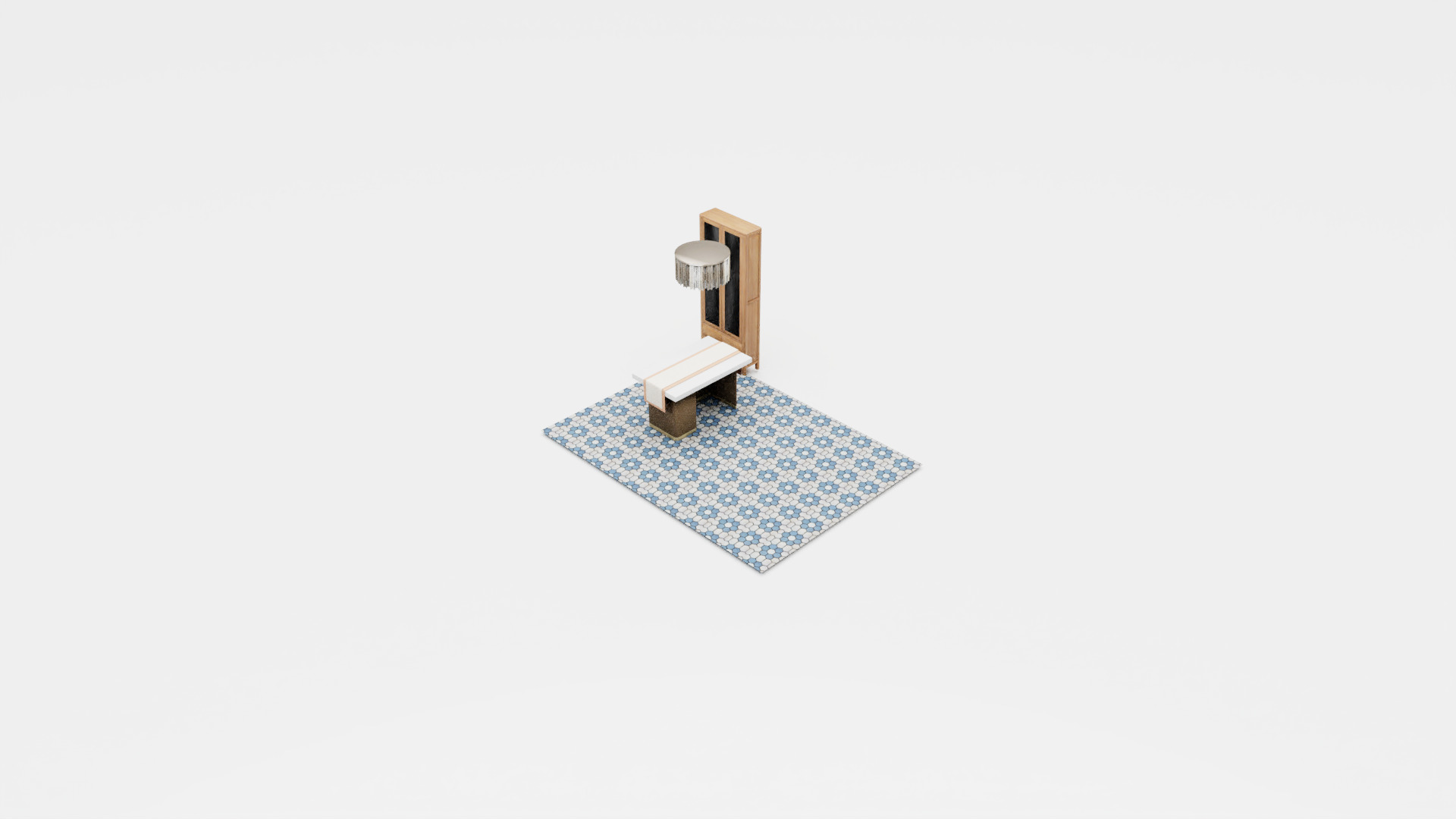}
    \end{subfigure}%
    \begin{subfigure}[b]{0.20\linewidth}
		\centering
		\includegraphics[width=\linewidth, trim=500 200 500 100, clip]{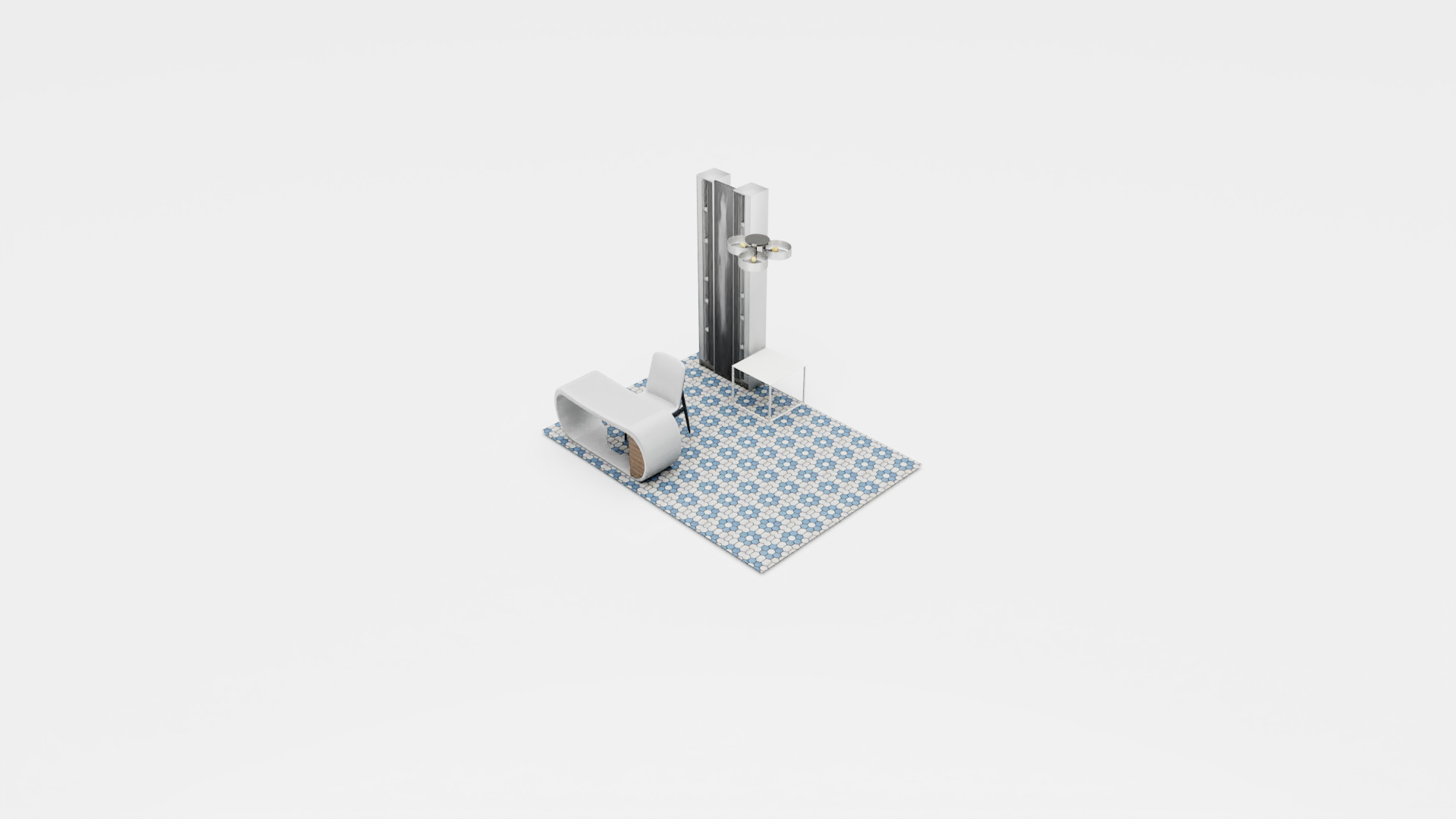}
    \end{subfigure}%
    \vskip\baselineskip%
    \vspace{-2.2em}
    \vskip\baselineskip%
    \begin{subfigure}[b]{0.20\linewidth}
		\centering
		\includegraphics[width=0.8\linewidth]{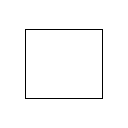}
    \end{subfigure}%
        \begin{subfigure}[b]{0.20\linewidth}
		\centering
		\includegraphics[width=\linewidth, trim=500 200 500 100, clip]{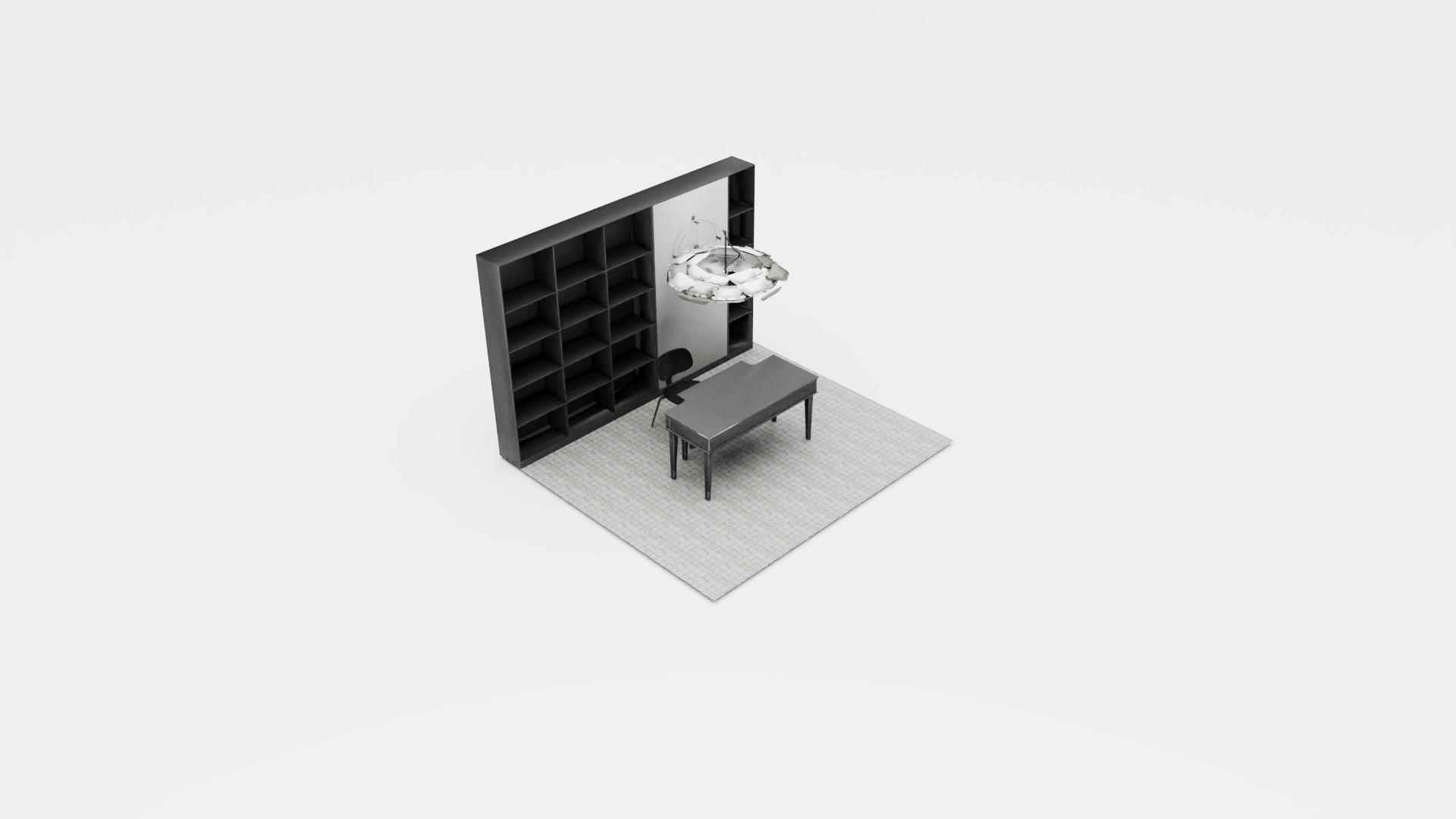}
    \end{subfigure}%
        \begin{subfigure}[b]{0.20\linewidth}
		\centering
		\includegraphics[width=\linewidth, trim=500 200 500 100, clip]{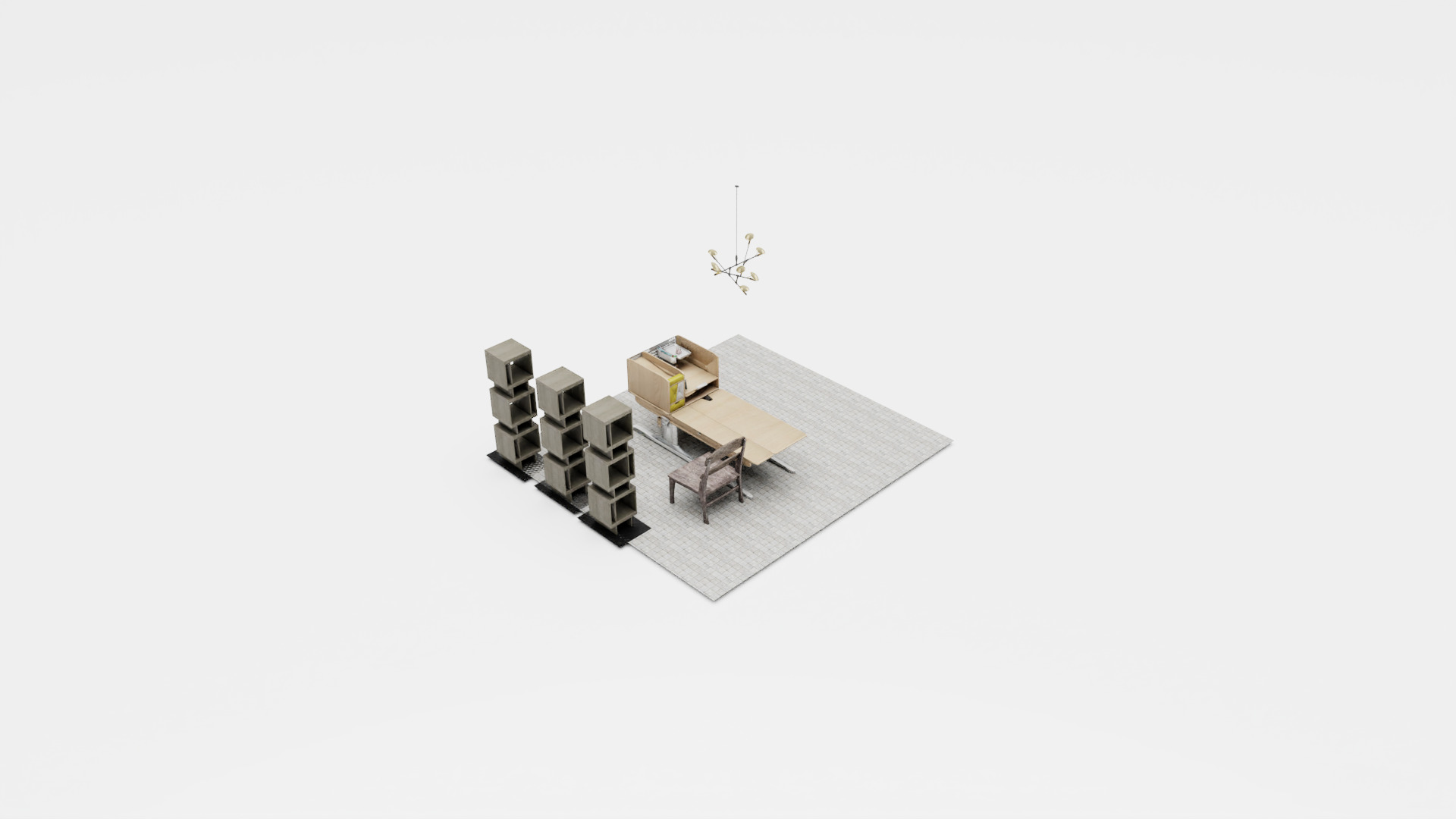}
    \end{subfigure}%
        \begin{subfigure}[b]{0.20\linewidth}
		\centering
		\includegraphics[width=\linewidth, trim=500 200 500 100, clip]{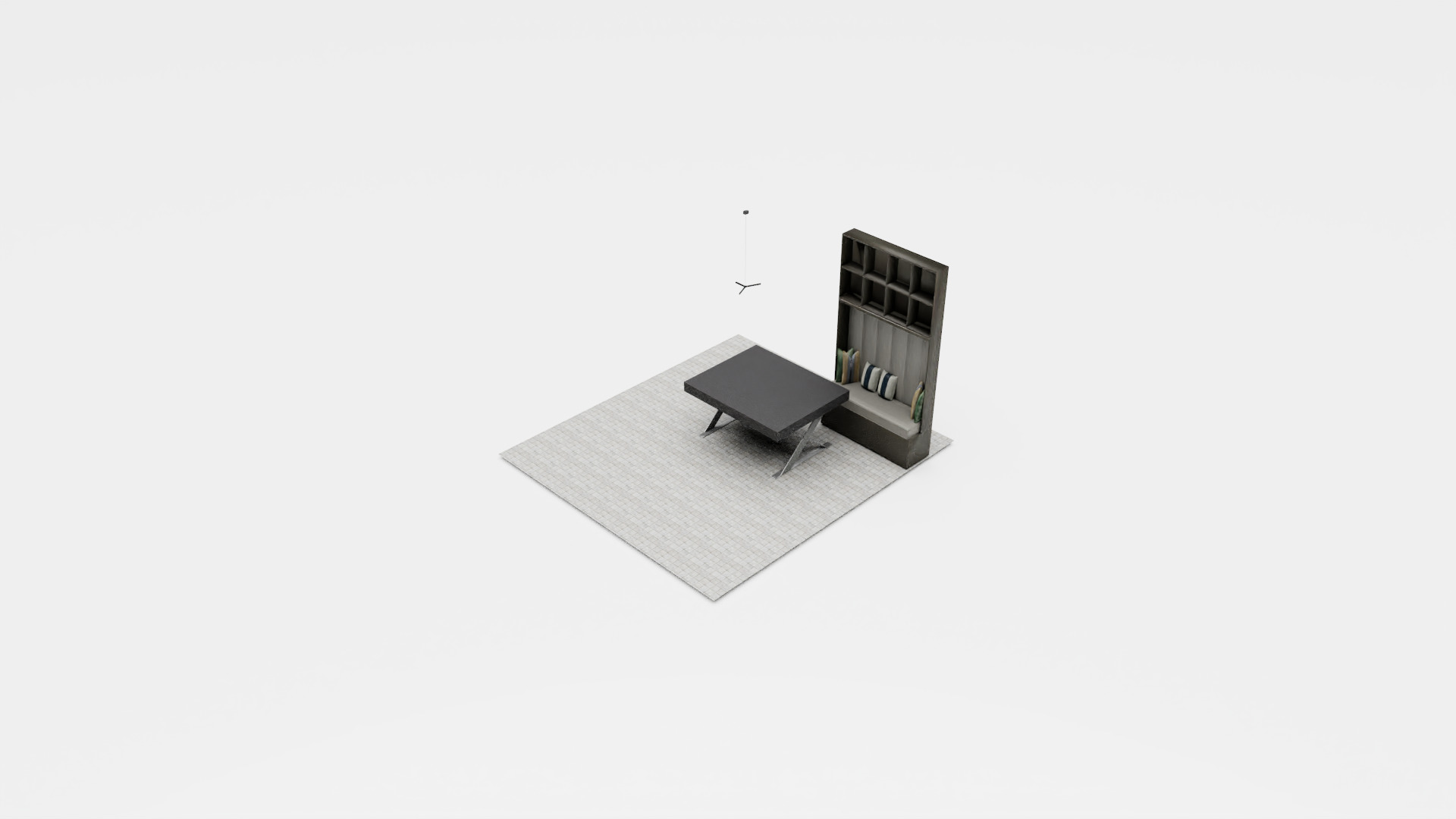}
    \end{subfigure}%
    \begin{subfigure}[b]{0.20\linewidth}
		\centering
		\includegraphics[width=\linewidth, trim=500 200 500 100, clip]{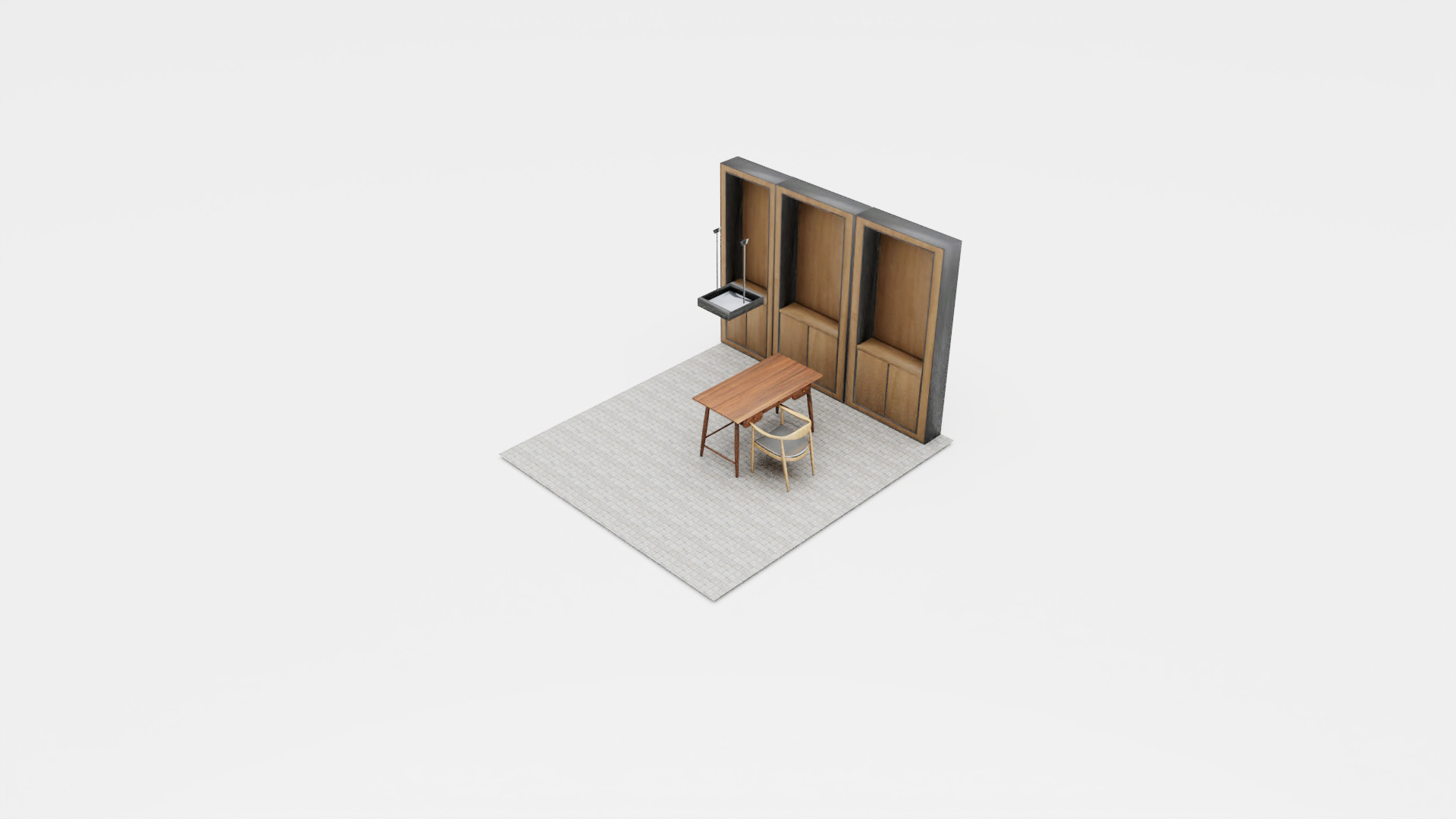}
    \end{subfigure}%
    \vskip\baselineskip%
    \vspace{-2.2em}
    \vskip\baselineskip%
    \begin{subfigure}[b]{0.20\linewidth}
		\centering
		\includegraphics[width=0.8\linewidth]{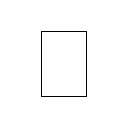}
    \end{subfigure}%
        \begin{subfigure}[b]{0.20\linewidth}
		\centering
		\includegraphics[width=\linewidth, trim=500 200 500 100, clip]{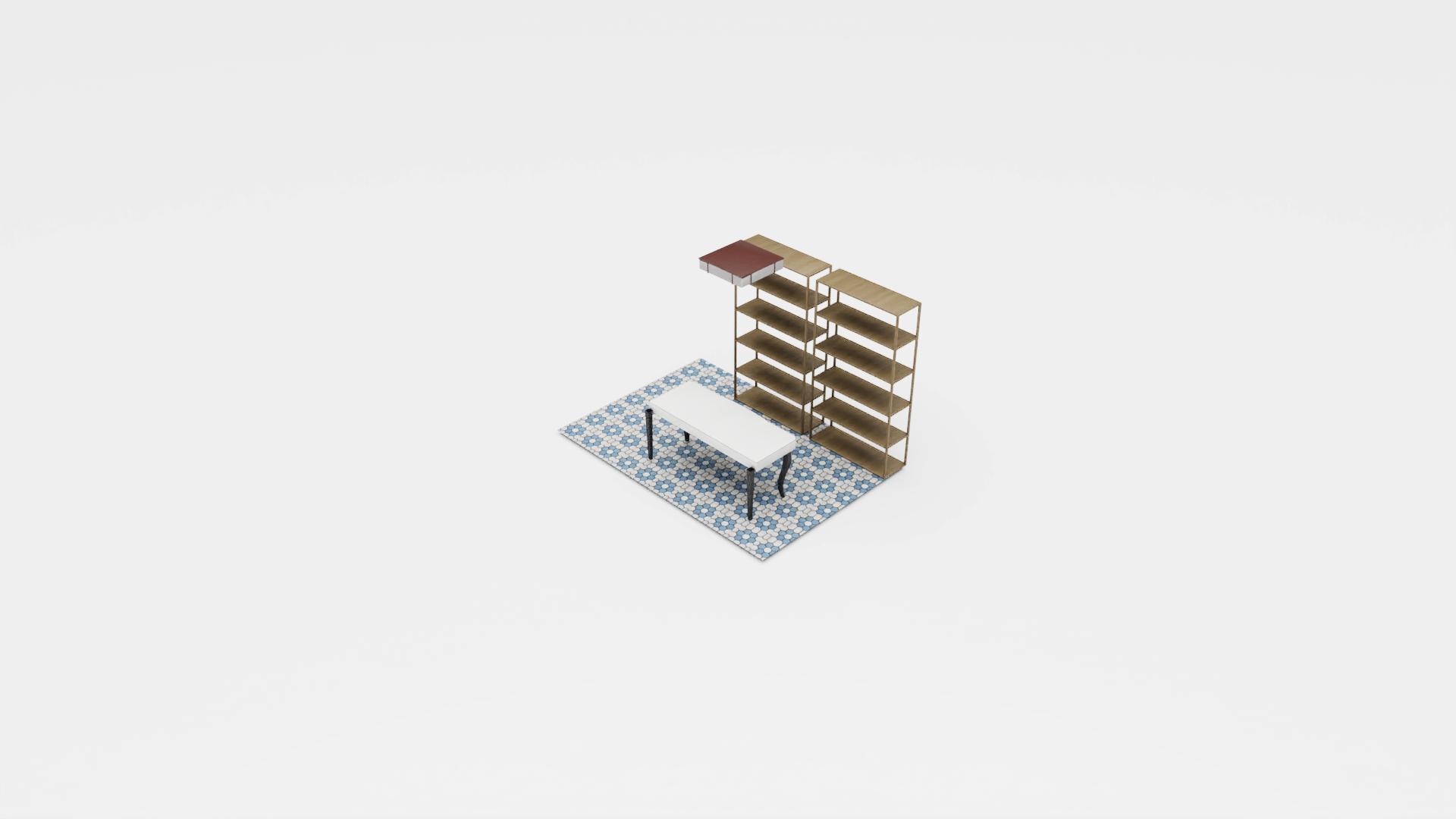}
    \end{subfigure}%
        \begin{subfigure}[b]{0.20\linewidth}
		\centering
		\includegraphics[width=\linewidth, trim=500 200 500 100, clip]{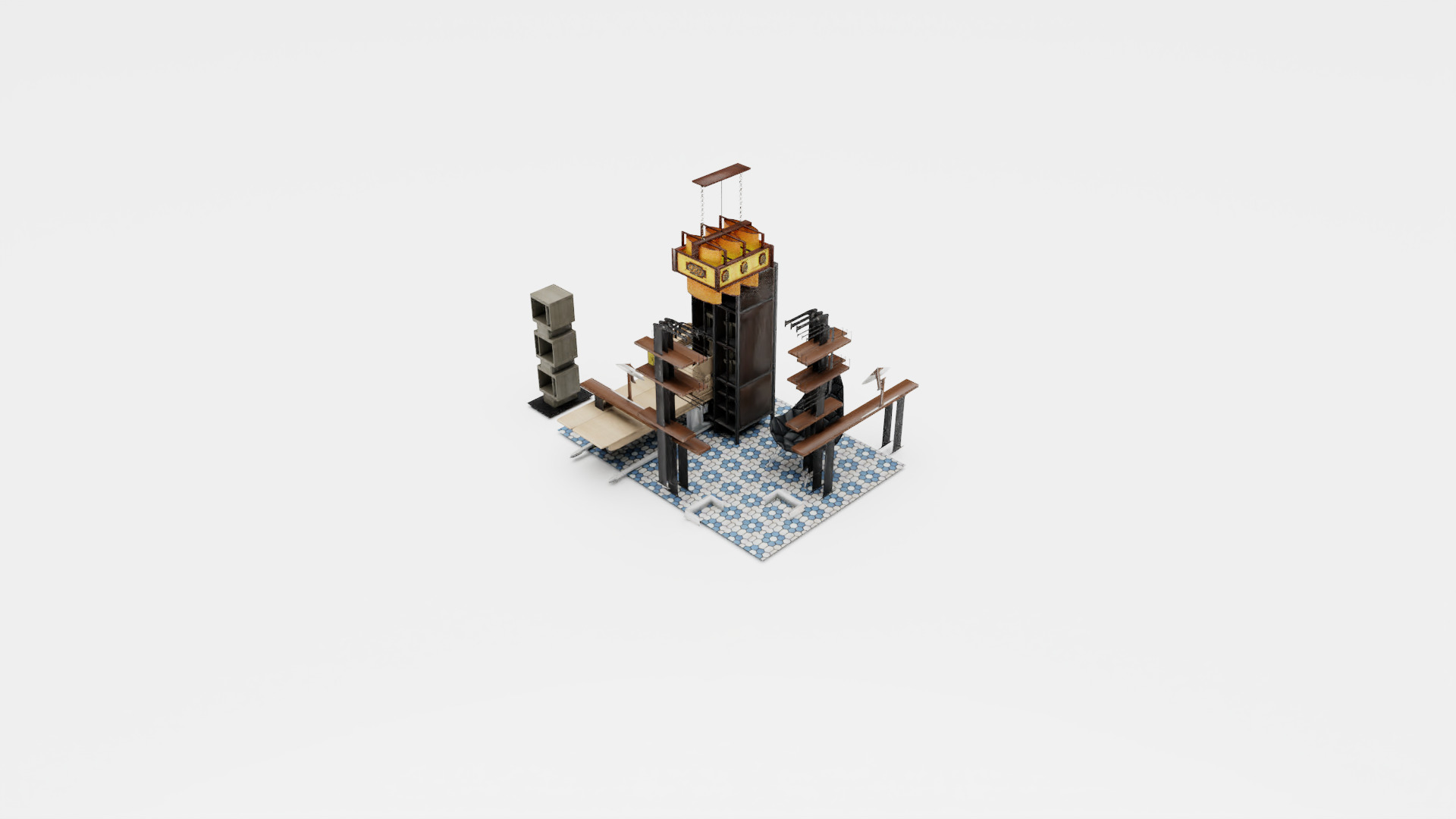}
    \end{subfigure}%
        \begin{subfigure}[b]{0.20\linewidth}
		\centering
		\includegraphics[width=\linewidth, trim=500 200 500 100, clip]{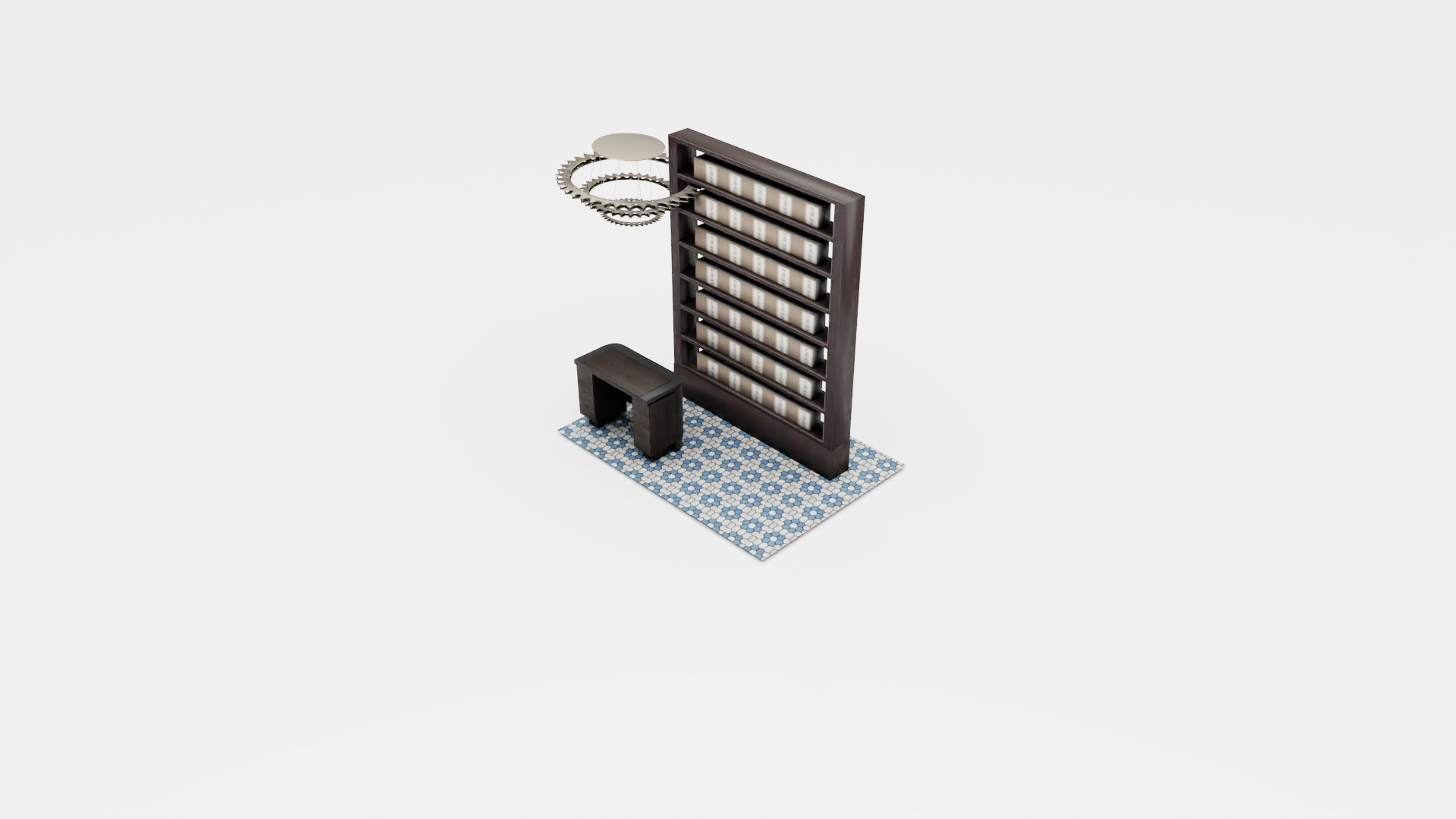}
    \end{subfigure}%
    \begin{subfigure}[b]{0.20\linewidth}
		\centering
		\includegraphics[width=\linewidth, trim=500 200 500 100, clip]{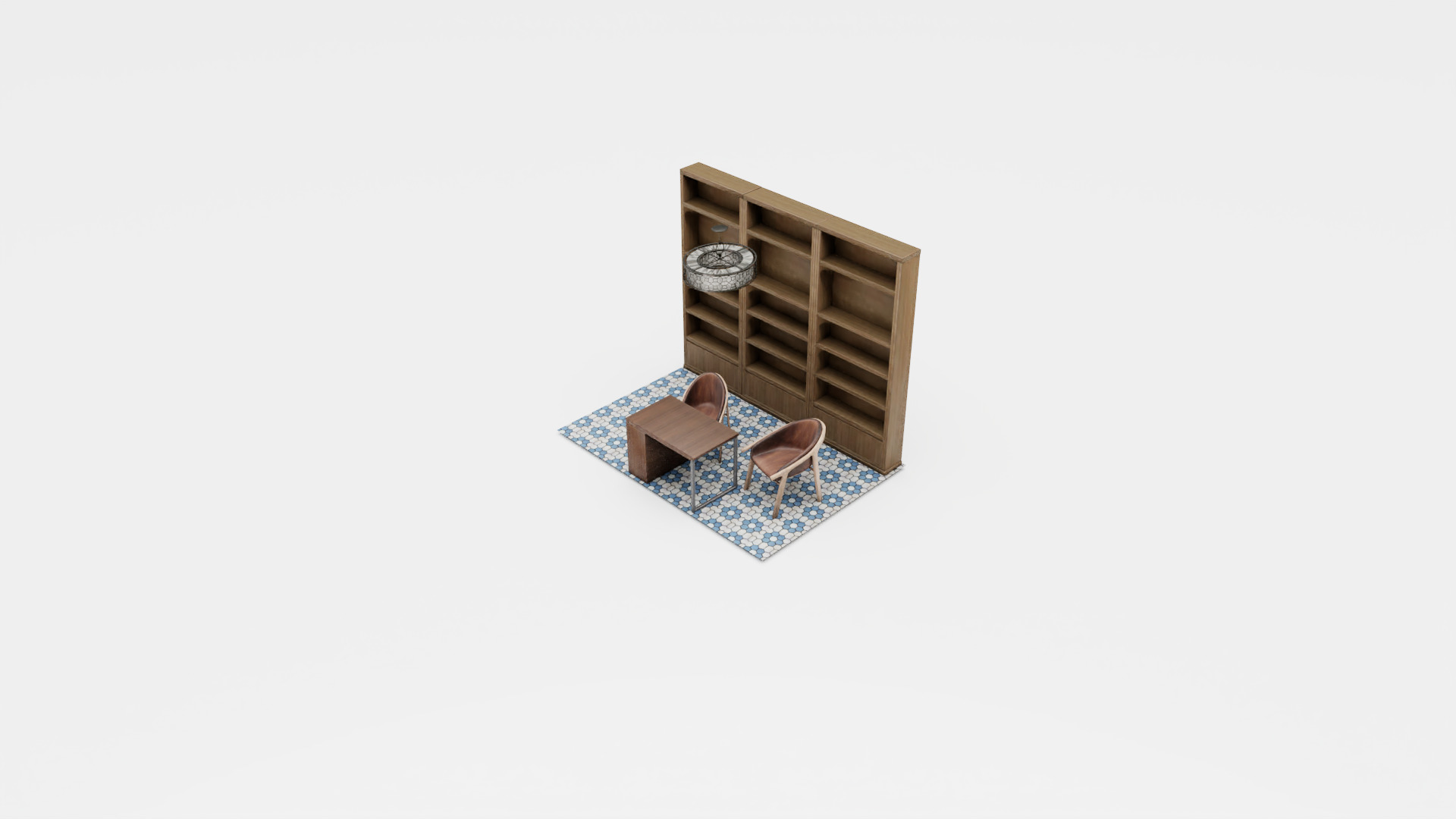}
    \end{subfigure}%
    \vskip\baselineskip%
    \vspace{-2.2em}
    \vskip\baselineskip%
    \begin{subfigure}[b]{0.20\linewidth}
		\centering
		\includegraphics[width=0.8\linewidth]{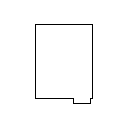}
    \end{subfigure}%
        \begin{subfigure}[b]{0.20\linewidth}
		\centering
		\includegraphics[width=\linewidth, trim=500 200 500 100, clip]{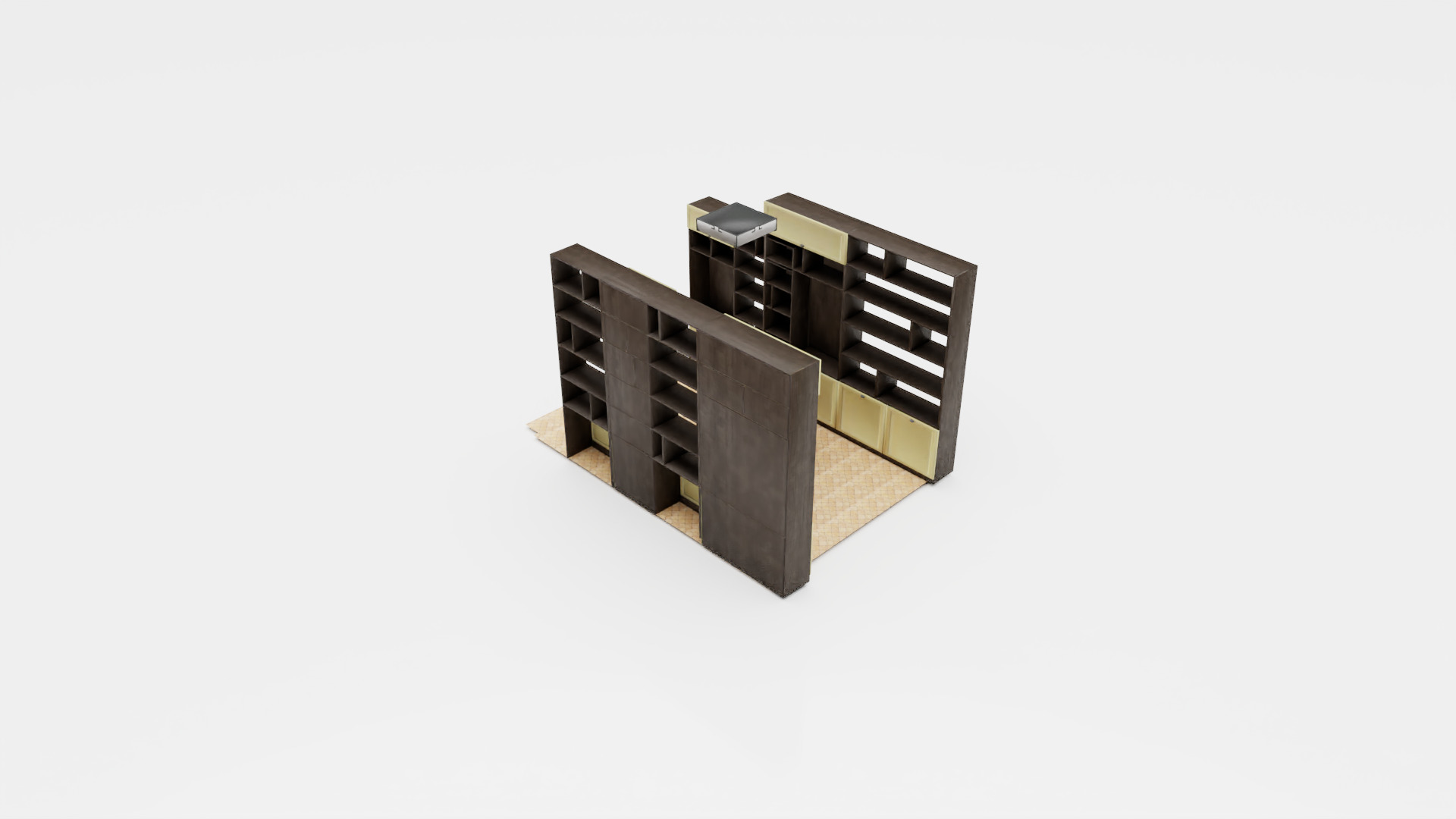}
    \end{subfigure}%
        \begin{subfigure}[b]{0.20\linewidth}
		\centering
		\includegraphics[width=\linewidth, trim=500 200 500 100, clip]{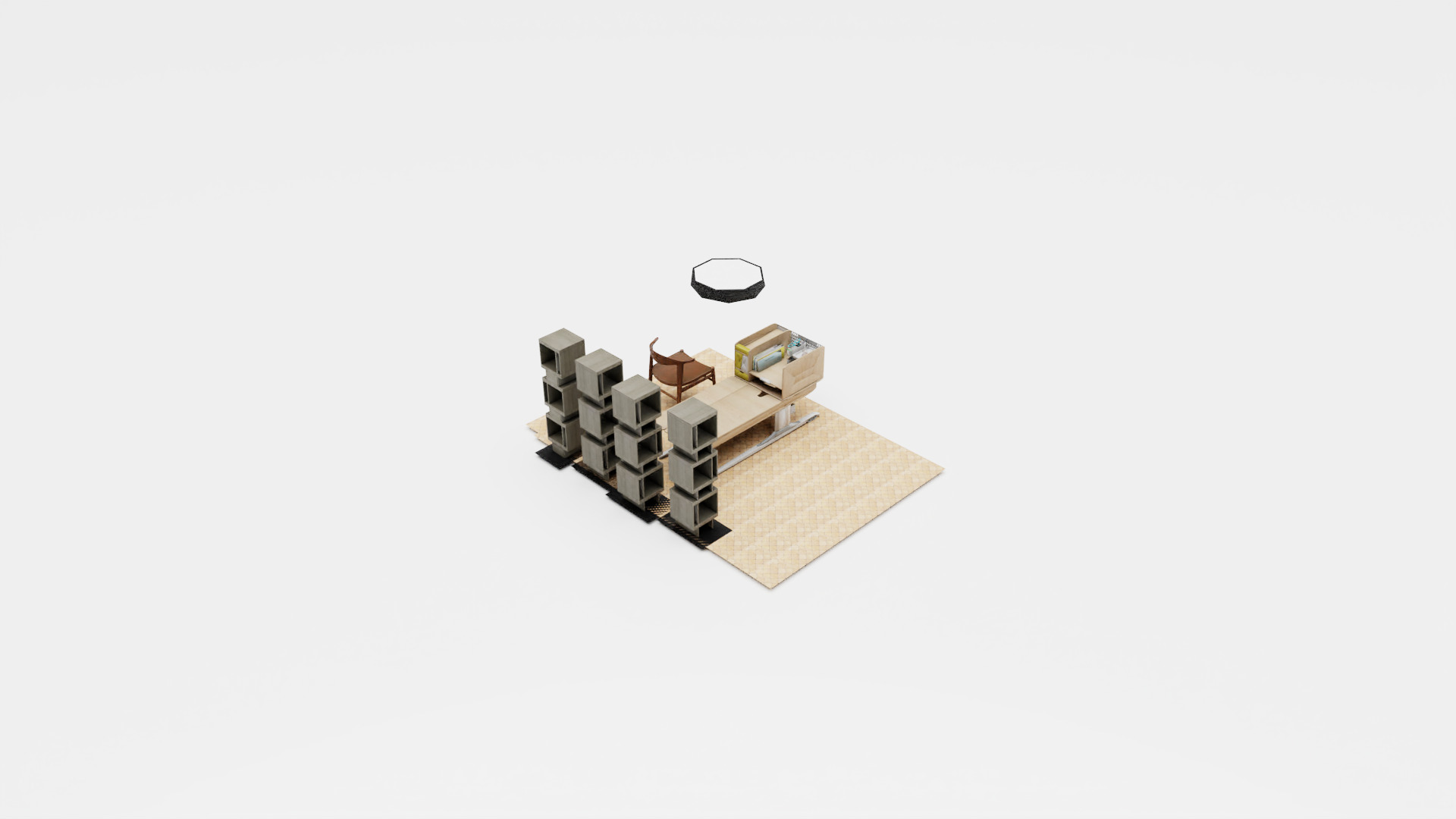}
    \end{subfigure}%
        \begin{subfigure}[b]{0.20\linewidth}
		\centering
		\includegraphics[width=\linewidth, trim=500 200 500 100, clip]{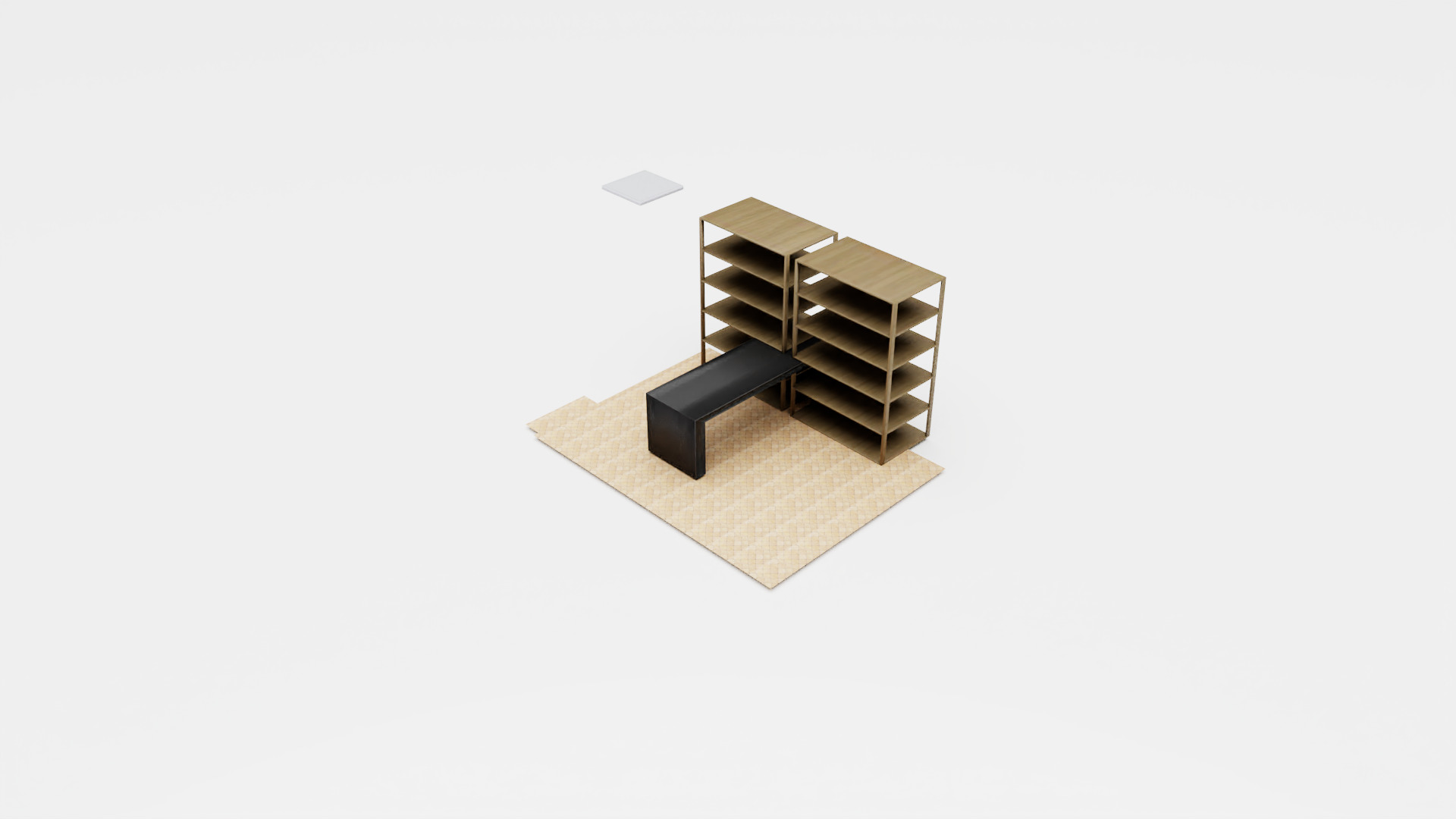}
    \end{subfigure}%
    \begin{subfigure}[b]{0.20\linewidth}
		\centering
		\includegraphics[width=\linewidth, trim=500 200 500 100, clip]{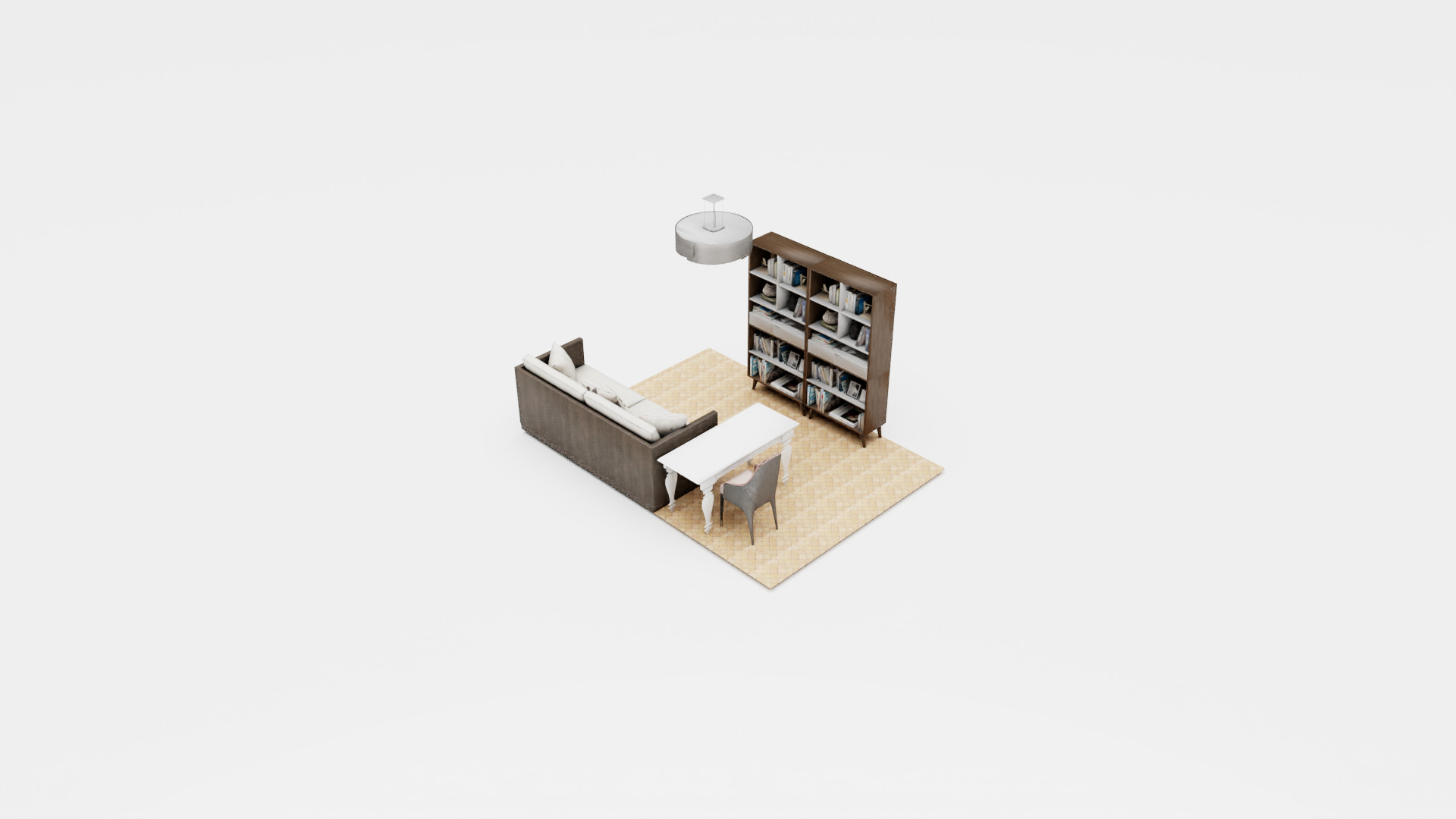}
    \end{subfigure}%
    \vskip\baselineskip%
    \vspace{-2.2em}
        		            		            		            		        		                \vskip\baselineskip%
    \begin{subfigure}[b]{0.20\linewidth}
		\centering
		\includegraphics[width=0.8\linewidth]{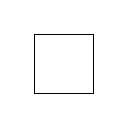}
    \end{subfigure}%
        \begin{subfigure}[b]{0.20\linewidth}
		\centering
		\includegraphics[width=\linewidth, trim=500 200 500 100, clip]{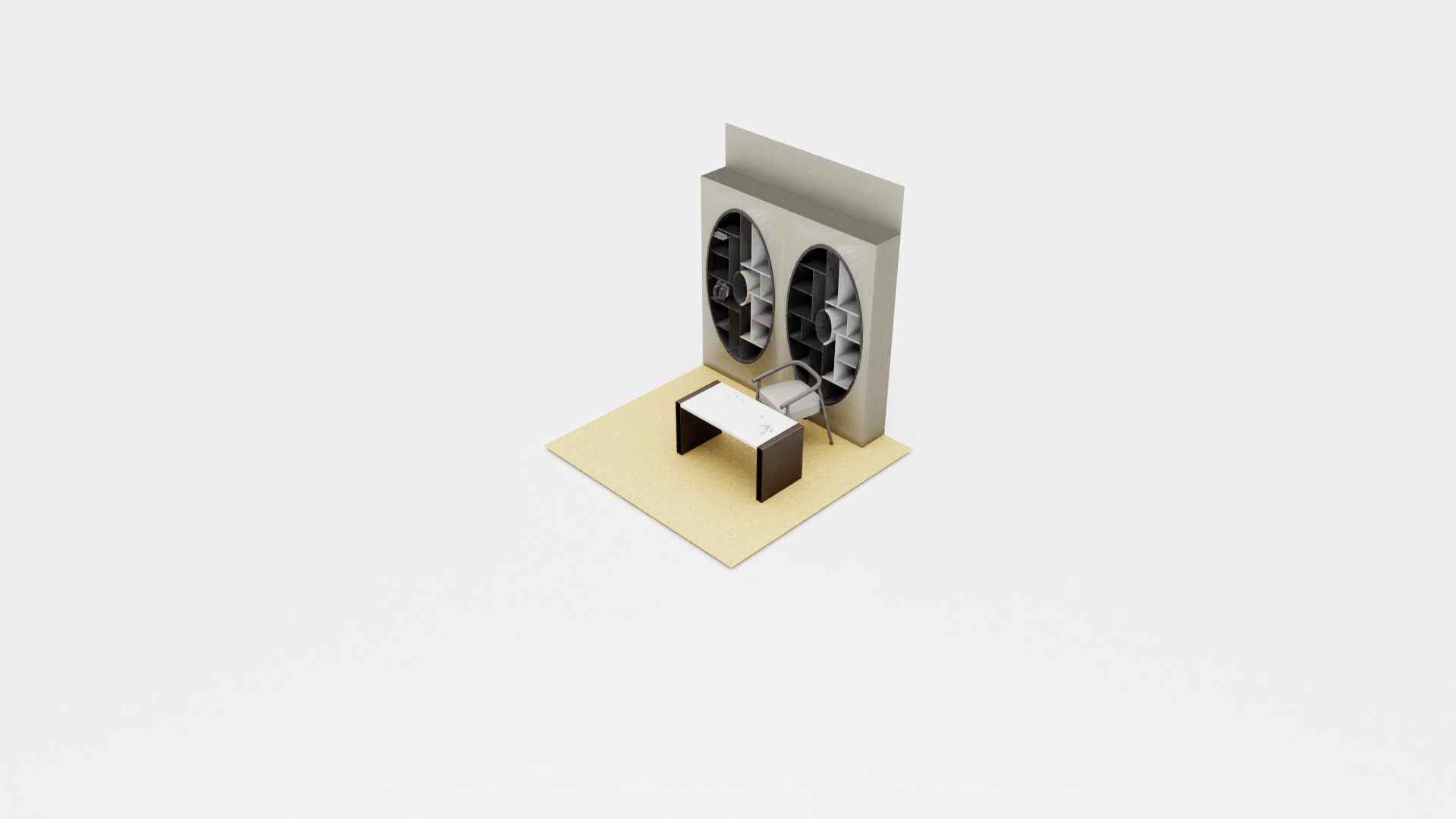}
    \end{subfigure}%
        \begin{subfigure}[b]{0.20\linewidth}
		\centering
		\includegraphics[width=\linewidth, trim=500 200 500 100, clip]{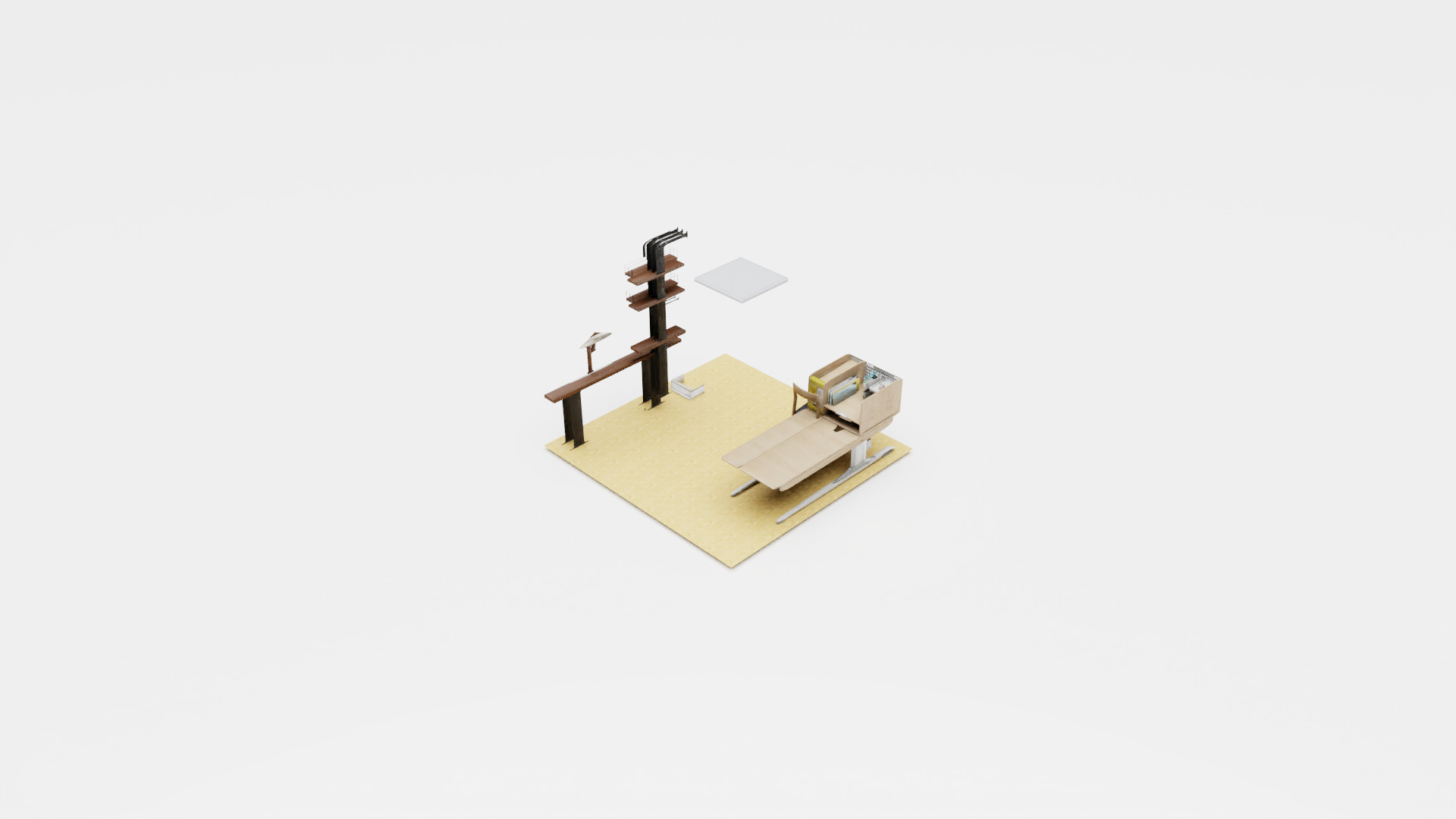}
    \end{subfigure}%
        \begin{subfigure}[b]{0.20\linewidth}
		\centering
		\includegraphics[width=\linewidth, trim=500 200 500 100, clip]{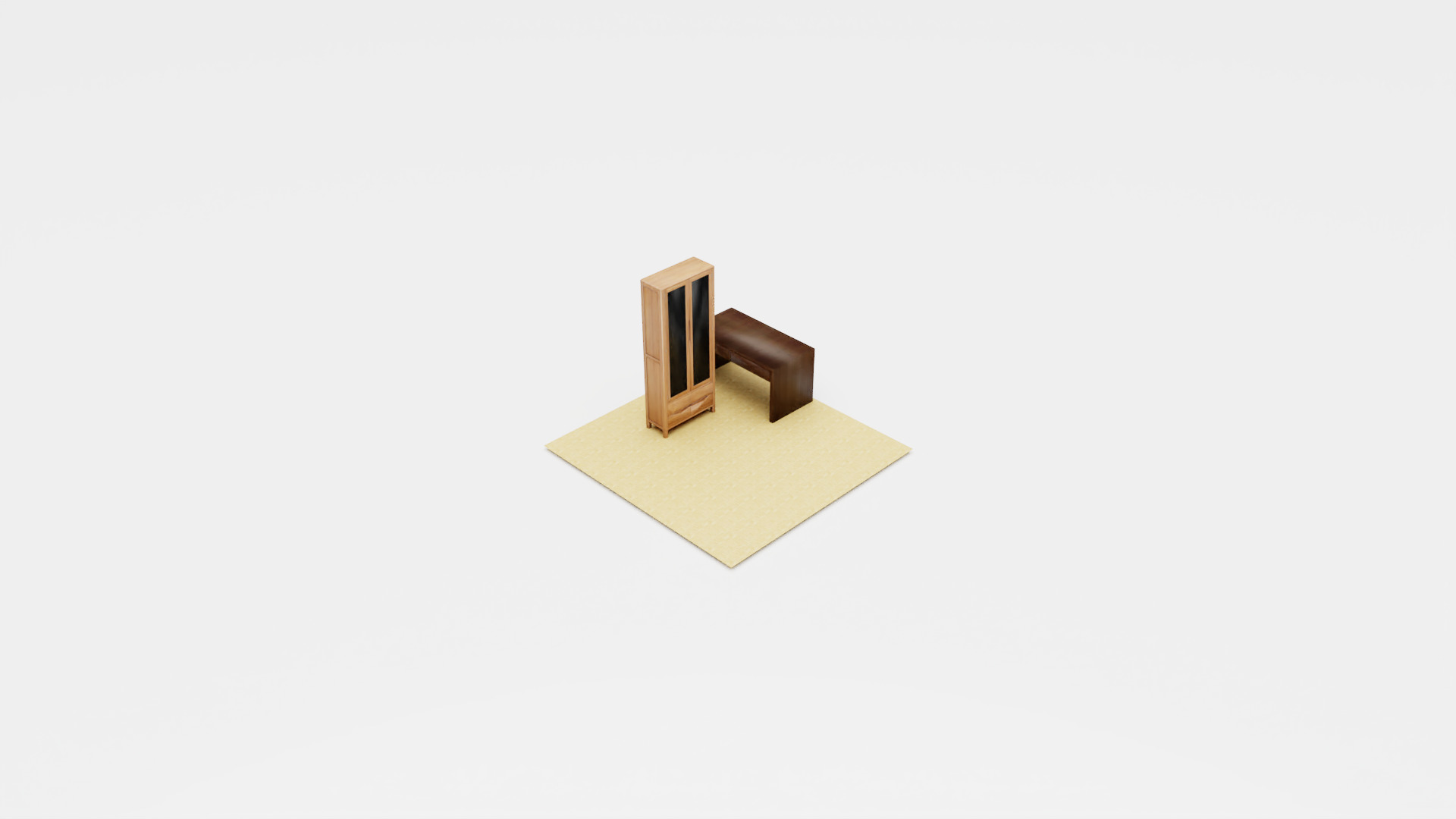}
    \end{subfigure}%
    \begin{subfigure}[b]{0.20\linewidth}
		\centering
		\includegraphics[width=\linewidth, trim=500 200 500 100, clip]{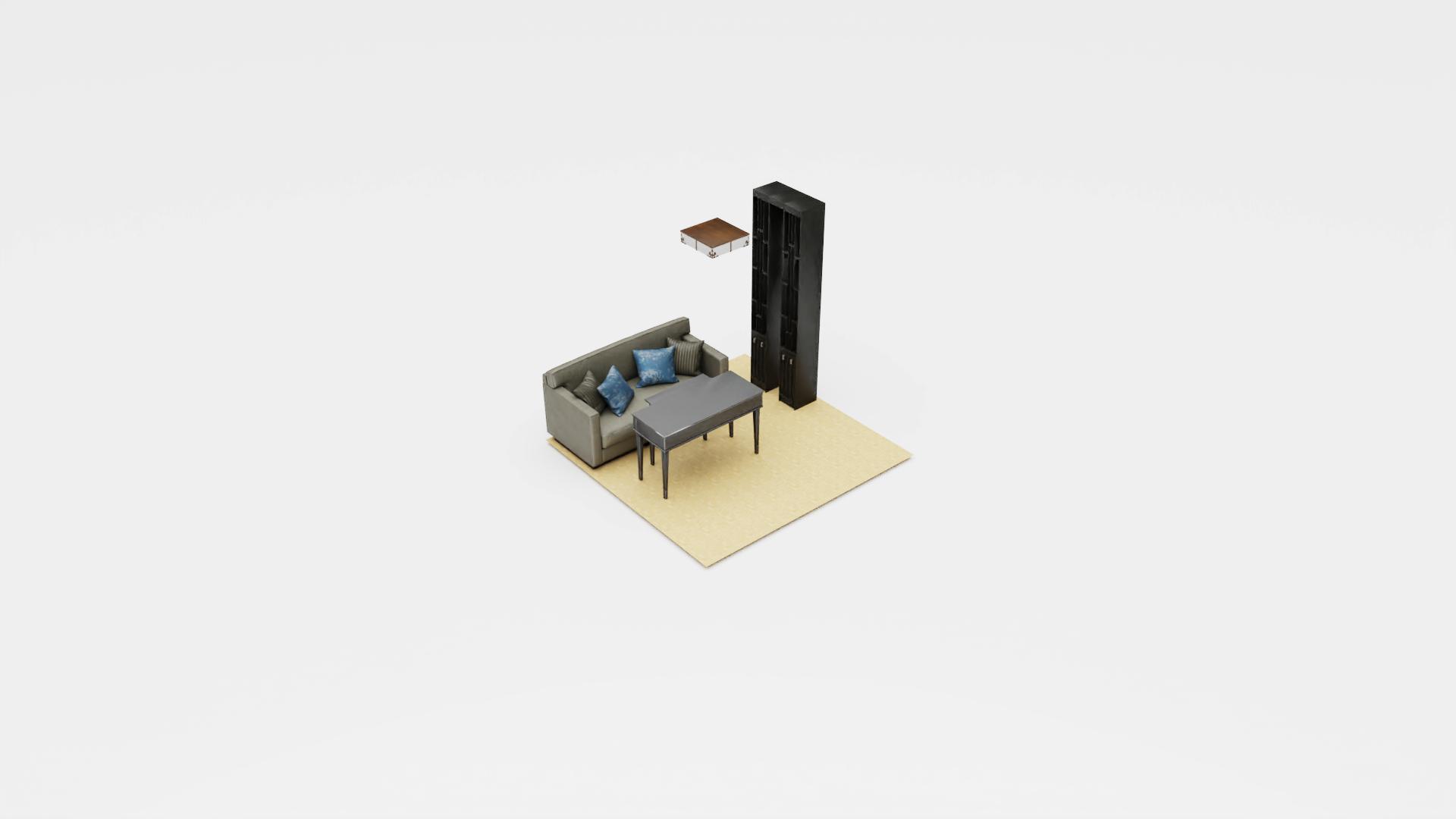}
    \end{subfigure}%
    \vskip\baselineskip%
    \vspace{-2.2em}
    \vskip\baselineskip%
    \begin{subfigure}[b]{0.20\linewidth}
		\centering
		\includegraphics[width=0.8\linewidth]{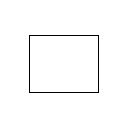}
    \end{subfigure}%
        \begin{subfigure}[b]{0.20\linewidth}
		\centering
		\includegraphics[width=\linewidth, trim=500 200 500 100, clip]{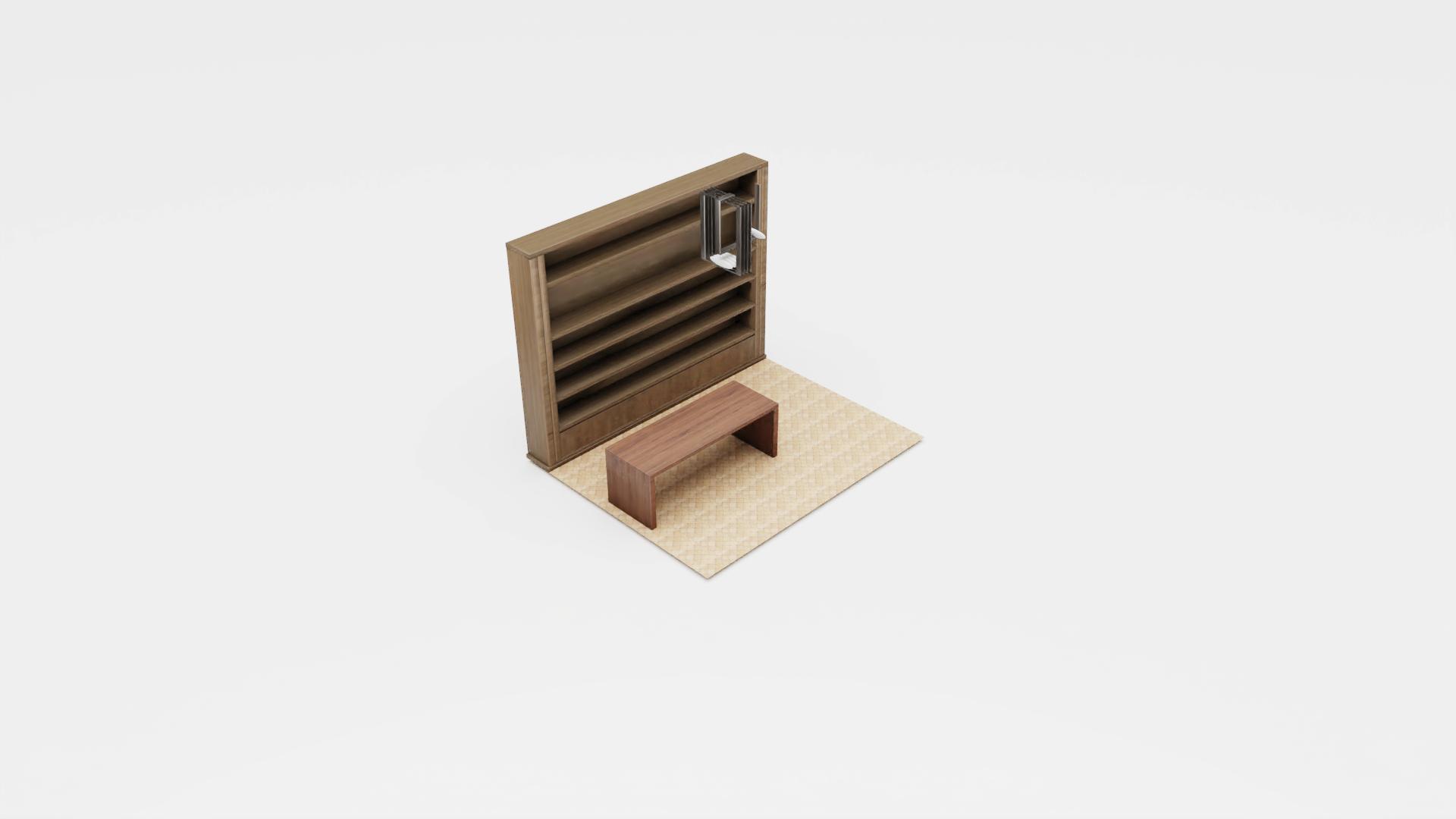}
    \end{subfigure}%
        \begin{subfigure}[b]{0.20\linewidth}
		\centering
		\includegraphics[width=\linewidth, trim=500 200 500 100, clip]{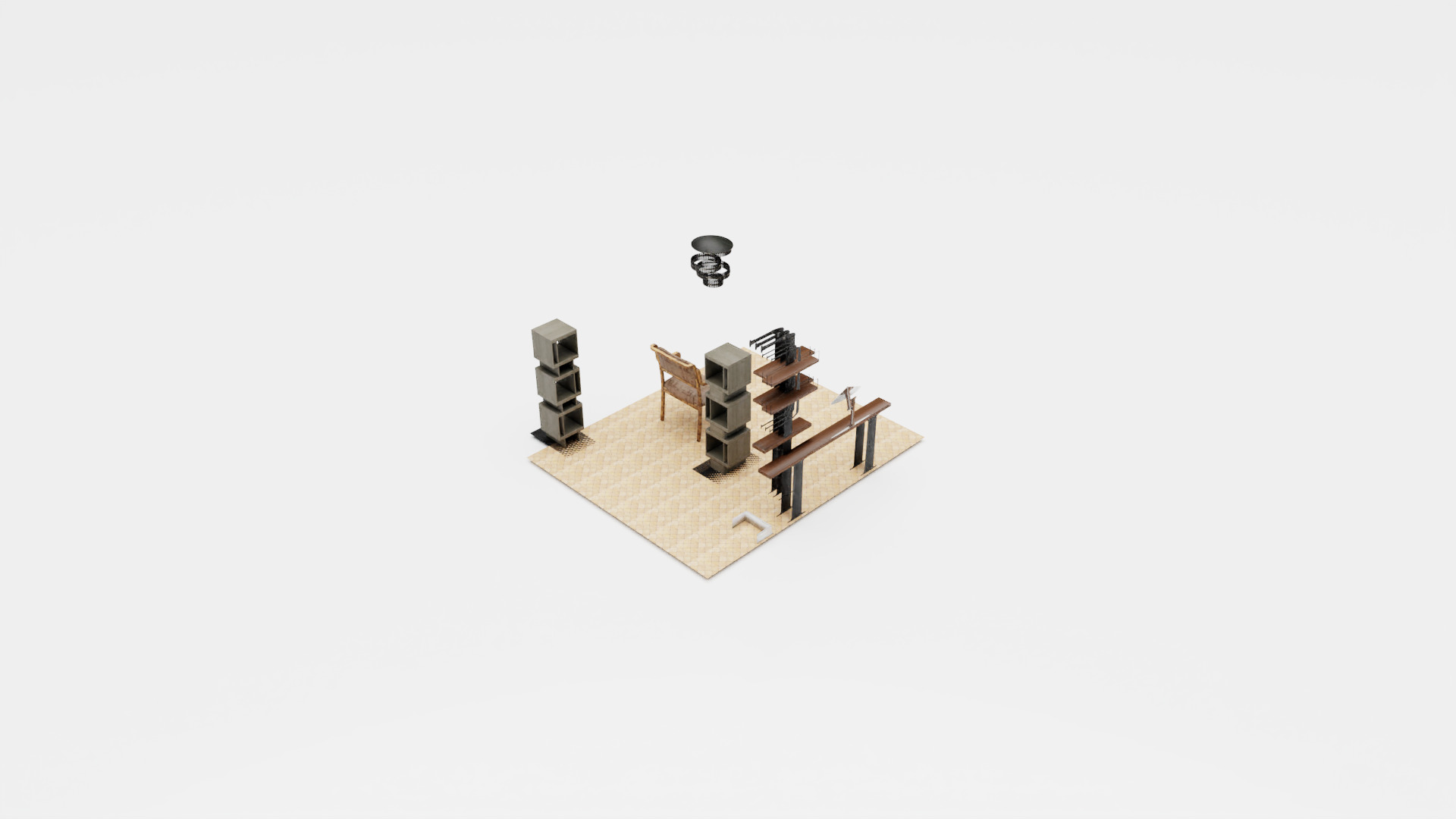}
    \end{subfigure}%
        \begin{subfigure}[b]{0.20\linewidth}
		\centering
		\includegraphics[width=\linewidth, trim=500 200 500 100, clip]{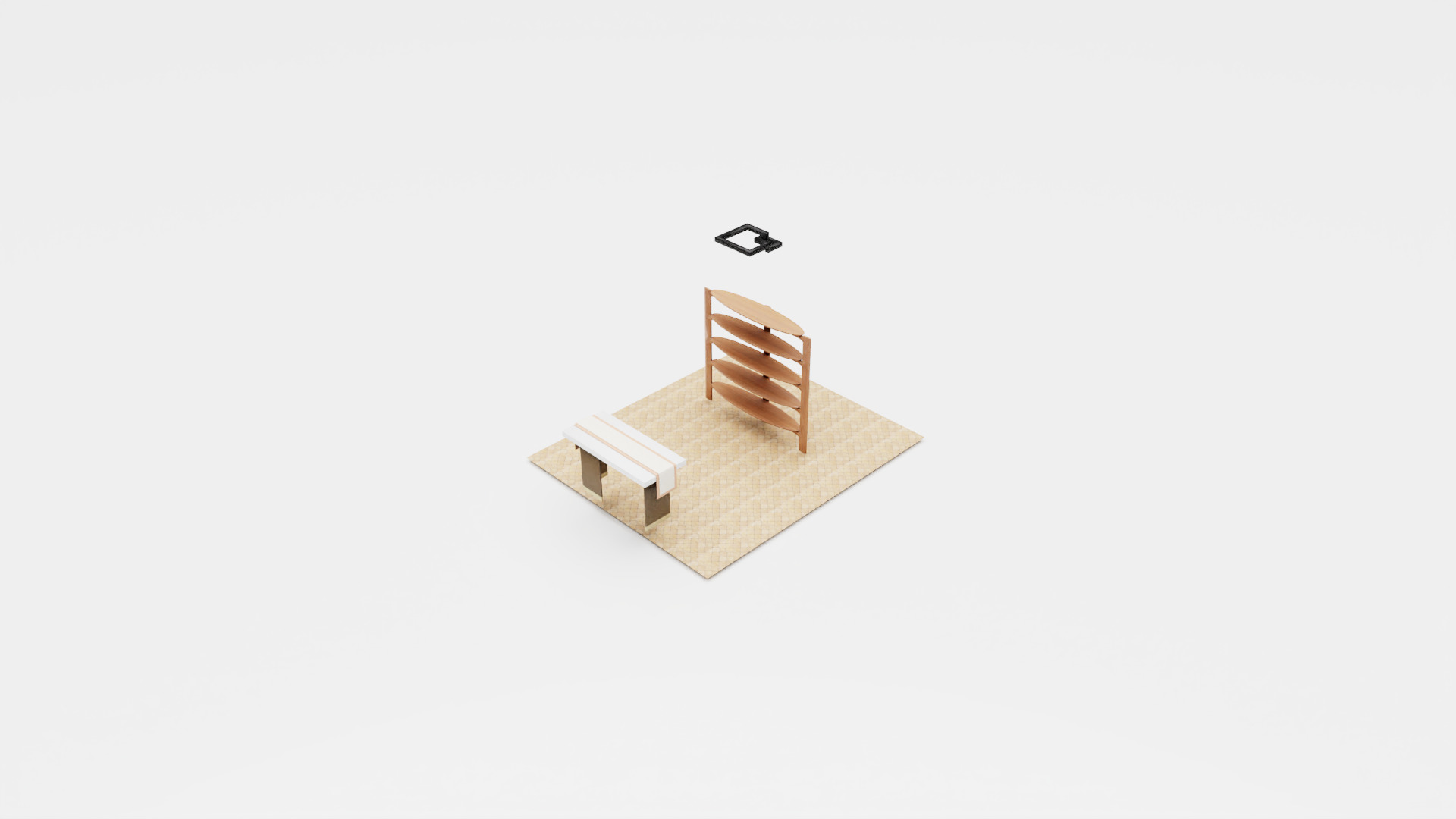}
    \end{subfigure}%
    \begin{subfigure}[b]{0.20\linewidth}
		\centering
		\includegraphics[width=\linewidth, trim=500 200 500 100, clip]{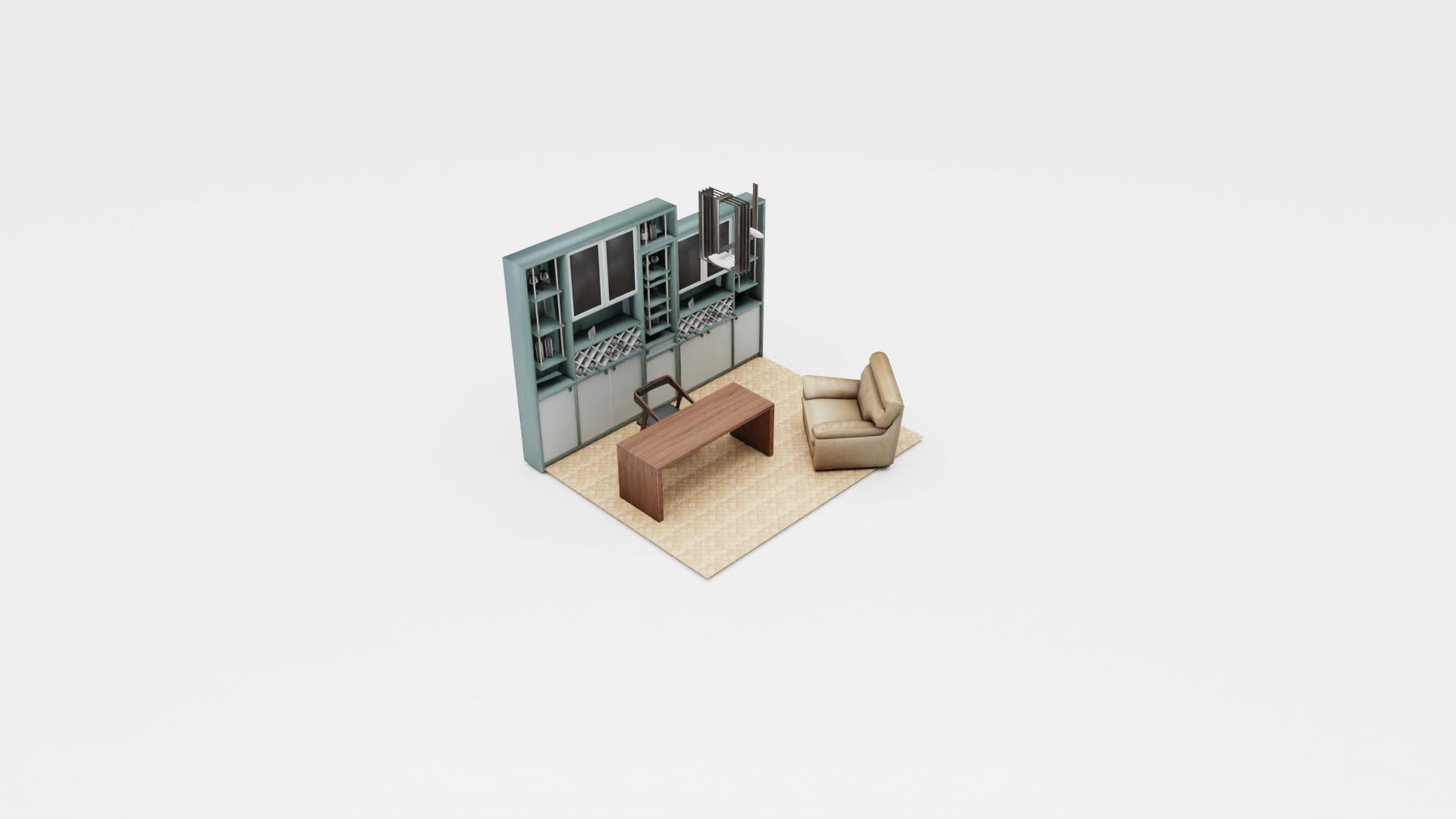}
    \end{subfigure}%
    \vspace{-1.2em}
    \vskip\baselineskip%
    \hfill%
    \caption{{\bf Qualitative Scene Synthesis Results on Libraries}.
    Generated scenes for libraries using FastSynth, SceneFormer and our method.
    To showcase the generalization abilities of our model we also show the
    closest scene from the training set (2nd column).}
    \label{fig:scene_synthesis_qualitative_library_supp}
    \vspace{-1.2em}
\end{figure}

Conditioned on a floor plan, we evaluate the performance of our model on
generating plausible furniture arrangements and compare with 
FastSynth~\cite{Ritchie2019CVPR} and SceneFormer~\cite{Wang2020ARXIV}. 
\figref{fig:scene_synthesis_qualitative_bedroom_supp} provides a qualitative
comparison of generated bedroom scenes conditioned on the same floor layout
using our model and our baselines. We observe that in contrast to
\cite{Ritchie2019CVPR, Wang2020ARXIV}, our model consistently generates layouts
with more diverse objects. In particular, \cite{Wang2020ARXIV} typically
generates bedrooms that consist only of a bed, a wardrobe and less frequently
also a nightstand, whereas both our model and FastSynth synthesize rooms with
more diverse objects. Similarly generated scenes for living rooms and dining
rooms are provided in \figref{fig:scene_synthesis_qualitative_livingroom_supp}
and \figref{fig:scene_synthesis_qualitative_diningroom_supp} respectively. We
observe that for the case of living rooms and dining rooms both baselines
struggle to generate plausible object arrangements, namely generated objects
are positioned outside the room boundaries, have unnatural sizes or populate a
small part of the scene. We hypothesize that this might be related to the
significantly smaller amount of training data compared to bedrooms.
Instead our model, generates realistic living rooms
and dining rooms. For the case of libraries (see
\figref{fig:scene_synthesis_qualitative_library_supp}), again both
\cite{Ritchie2019CVPR, Wang2020ARXIV} struggle to generate functional rooms.

\subsection{Object Co-occurrence}

\begin{figure}[h!]
    \vspace{-0.4em}
    \centering
    \includegraphics[width=1\textwidth]{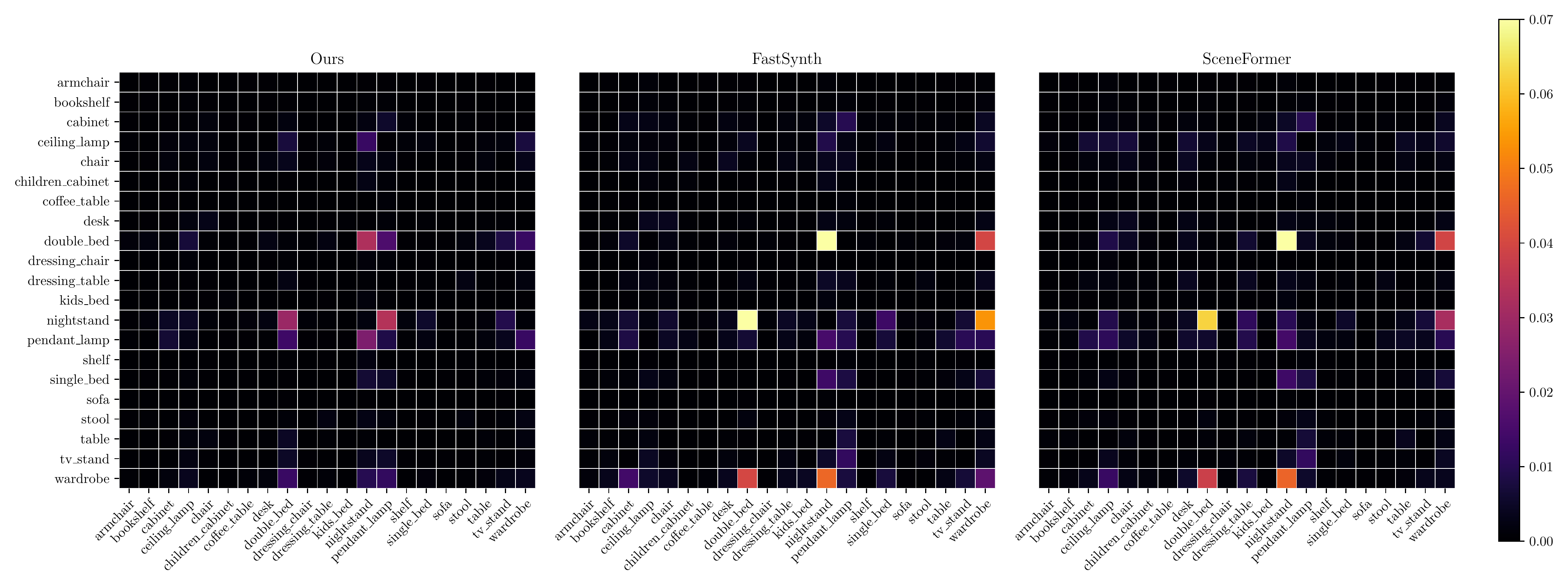}
    \vspace{-1.2em}
    \vskip\baselineskip%
    \caption{\textbf{Absolute Difference between Object Co-occurrence in
    Bedrooms.} We visualize the absolute difference of the probabilities of
    object co-occurrence computed between real and synthesized scenes using
    ATISS (left), FastSynth (middle) and SceneFormer (right). Larger
    differences correspond to warmer colors and are worse.}
    \label{fig:cooccurrences_bedroom}
\end{figure}

To further validate the ability of our model to reproduce the probabilities of
object co-occurrence in the real scenes, we compare the probabilities of object
co-occurrence of synthesized scenes using our model,
FastSynth~\cite{Ritchie2019CVPR} and SceneFormer~\cite{Wang2020ARXIV} for all
room types. In particular, in this experiment, we generate $5000$ scenes using
each method and report the difference between the probabilities of object
co-occurrences between real and synthesized scenes.
\figref{fig:cooccurrences_bedroom} summarizes the absolute differences for the
bedroom scenes. We observe that our model better captures the object
co-occurrence than baselines since the absolute differences for most object
pairs are consistently smaller.

\begin{figure}
    \centering
    \includegraphics[width=1\textwidth]{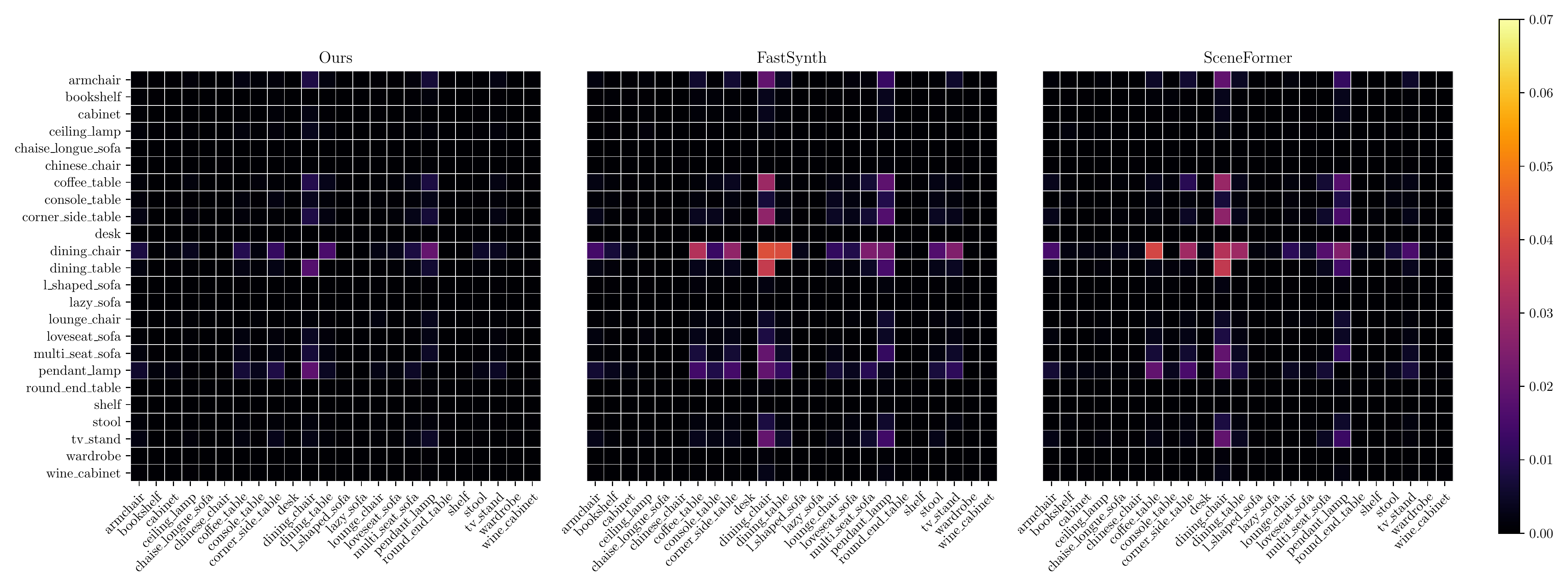}
    \vspace{-1.2em}
    \vskip\baselineskip%
    \caption{\textbf{Absolute Difference between Object Co-occurrence in
    Living Rooms.} We visualize the absolute difference of the probabilities of
    object co-occurrence computed between real and synthesized scenes using
    ATISS (left-most column), FastSynth (middle column), SceneFormer (right-most
    column). Lower is better.}
    \label{fig:cooccurrences_living}
\end{figure}

\begin{figure}
    \centering
    \includegraphics[width=1\textwidth]{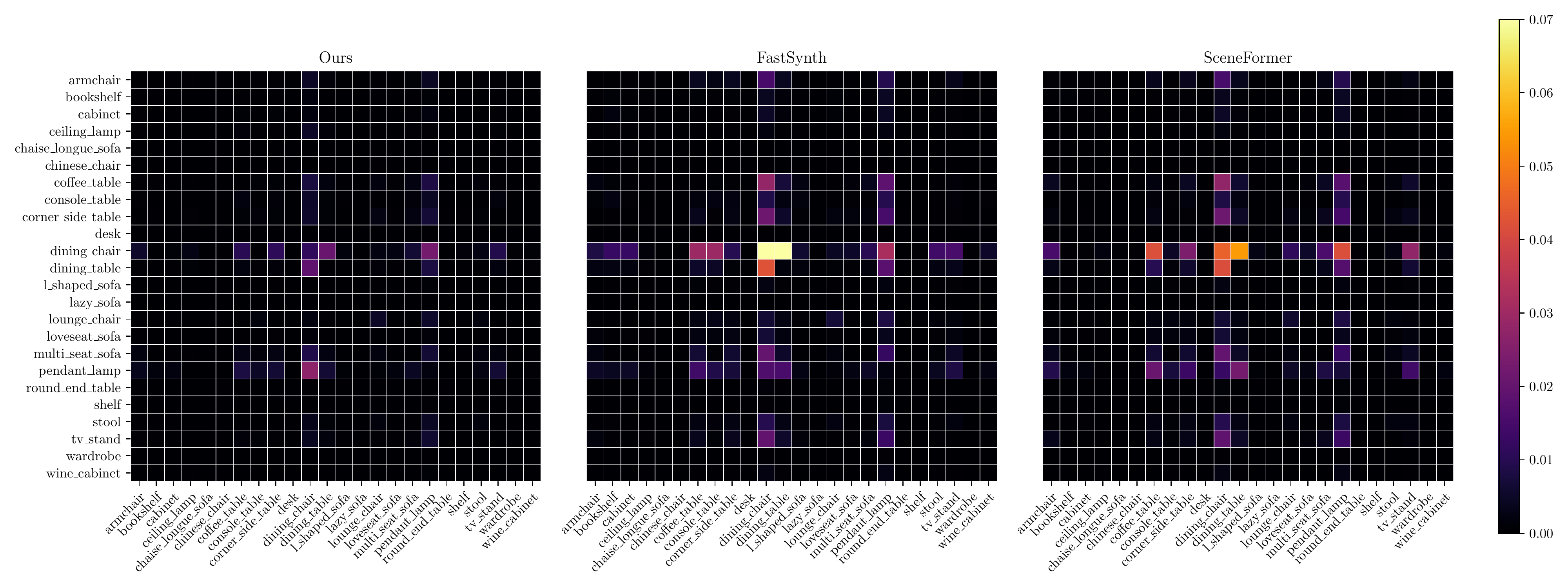}
    \vspace{-1.2em}
    \vskip\baselineskip%
    \caption{\textbf{Absolute Difference between Object Co-occurrence in
    Dining Rooms.} We visualize the absolute difference of the probabilities of
    object co-occurrence computed between real and synthesized scenes using
    ATISS (left-most column), FastSynth (middle column), SceneFormer (right-most
    column). Lower is better.}
    \label{fig:cooccurrences_dining}
\end{figure}

\begin{figure}
    \centering
    \includegraphics[width=1\textwidth]{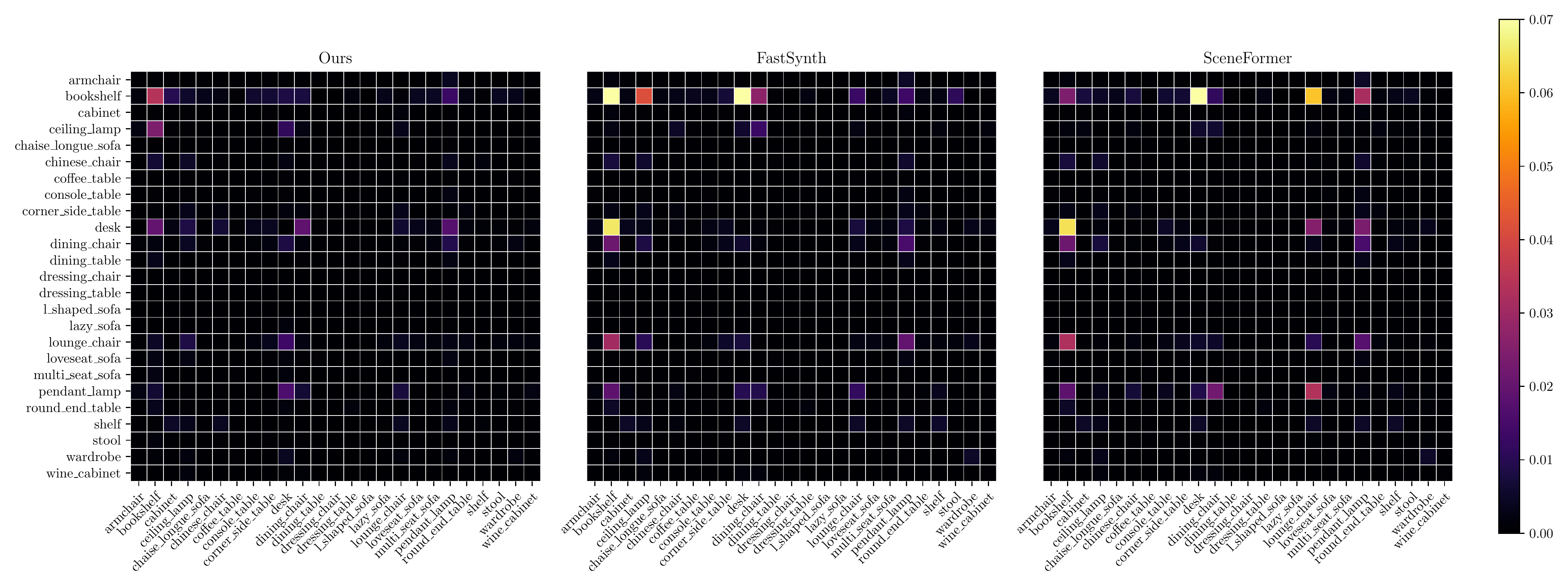}
    \vspace{-1.2em}
    \vskip\baselineskip%
    \caption{\textbf{Absolute Difference between Object Co-occurrence in
    Libraries.} We visualize the absolute difference of the probabilities of
    object co-occurrence computed between real and synthesized scenes using
    ATISS (left-most column), FastSynth (middle column), SceneFormer (right-most
    column). Lower is better.}
    \label{fig:cooccurrences_library}
\end{figure}

This is also validated for the case of living rooms (\figref{fig:cooccurrences_living}), dining rooms
(\figref{fig:cooccurrences_dining}) and libraries
(\figref{fig:cooccurrences_library}), where our model better captures the object
co-occurrences than both FastSynth~\cite{Ritchie2019CVPR} and
SceneFormer~\cite{Wang2020ARXIV}. Note that from our analysis it becomes
evident that while our method better reproduces the probabilities of object
co-occurrence from the real scenes, all methods are able to generate scenes
with plausible object co-occurrences. This is expected, since learning the
categories of objects to be added in a scene is a significantly easier task in
comparison to learning their sizes and positions in 3D space.

\begin{figure}
    \centering
    \begin{subfigure}[b]{\textwidth}
         \centering
         \includegraphics[width=\textwidth]{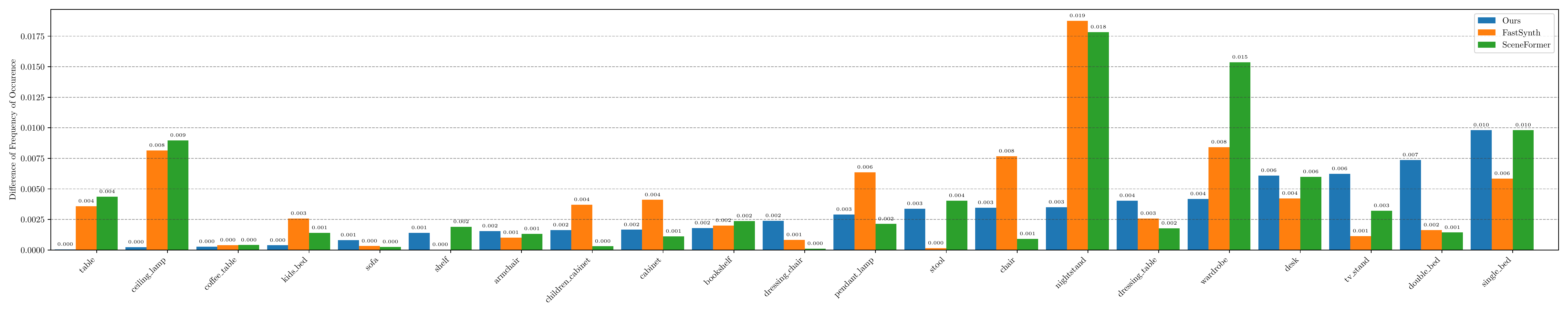}
         \vspace{-2.2em}
         \caption{Bedrooms}
         \label{fig:bedroom_bar_plots}
     \end{subfigure}
     \hfill
     \begin{subfigure}[b]{\textwidth}
         \centering
         \includegraphics[width=\textwidth]{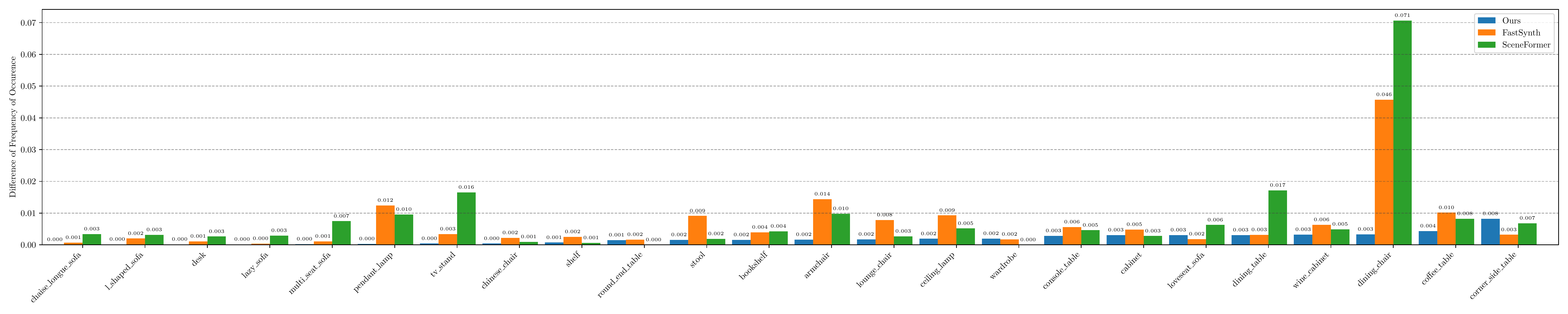}
         \vspace{-2.2em}
         \caption{Living Rooms}
         \label{fig:living_bar_plots}
     \end{subfigure}
     \hfill
     \begin{subfigure}[b]{\textwidth}
         \centering
         \includegraphics[width=\textwidth]{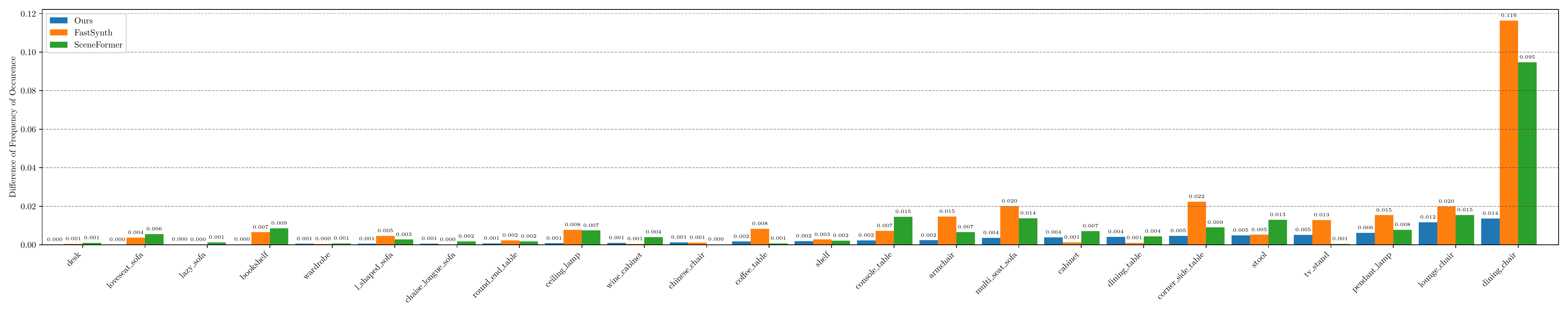}
         \vspace{-2.2em}
         \caption{Dining Rooms}
         \label{fig:dining_bar_plots}
     \end{subfigure}
     \begin{subfigure}[b]{\textwidth}
         \centering
         \includegraphics[width=\textwidth]{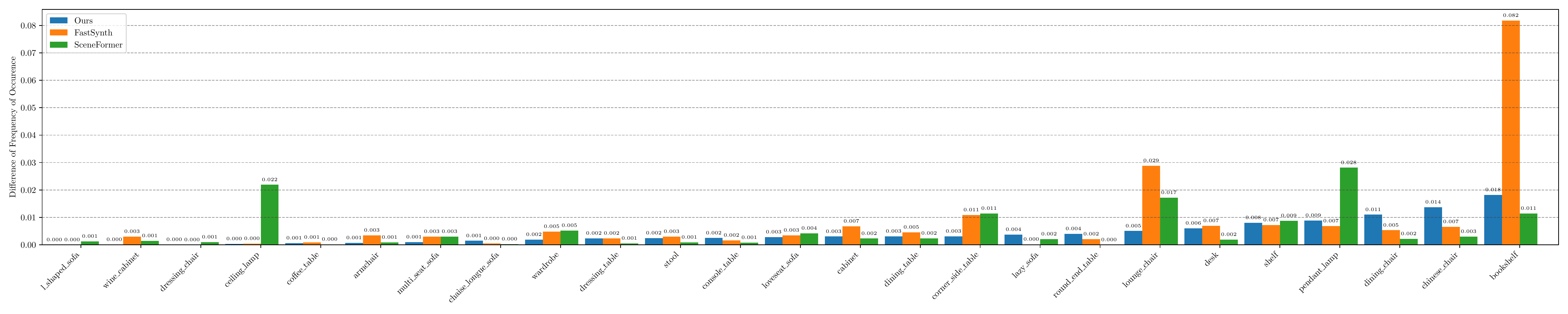}
         \vspace{-2.2em}
         \caption{Libraries}
         \label{fig:library_bar_plots}
     \end{subfigure}
     \caption{\textbf{Difference of Per-Object Frequencies.} We visualize the
     absolute difference between the per-object frequency of generated and real
     scenes using our method, FastSynth~\cite{Ritchie2019CVPR} and
     SceneFormer~\cite{Wang2020ARXIV} for all room types. Lower is better.}
     \label{fig:freq_occurency_rooms}
\end{figure}

Finally, in \figref{fig:freq_occurency_rooms}, we visualize the per-object
difference in frequency of occurrence between synthesized and real scenes from
the test set for all room types. We observe that our model generates object
arrangements with comparable per-object frequencies to real rooms. In
particular, for the case of living rooms (\ref{fig:living_bar_plots}), dining
rooms (\ref{fig:dining_bar_plots}) and libraries
(\ref{fig:library_bar_plots}) that are more challenging rooms types due to
their smaller size, our model has an even smaller discrepancy \wrt the
per-object frequencies.

\subsection{Visualizations of Predicted Distributions}

\begin{figure}
    \centering
     \includegraphics[width=\textwidth]{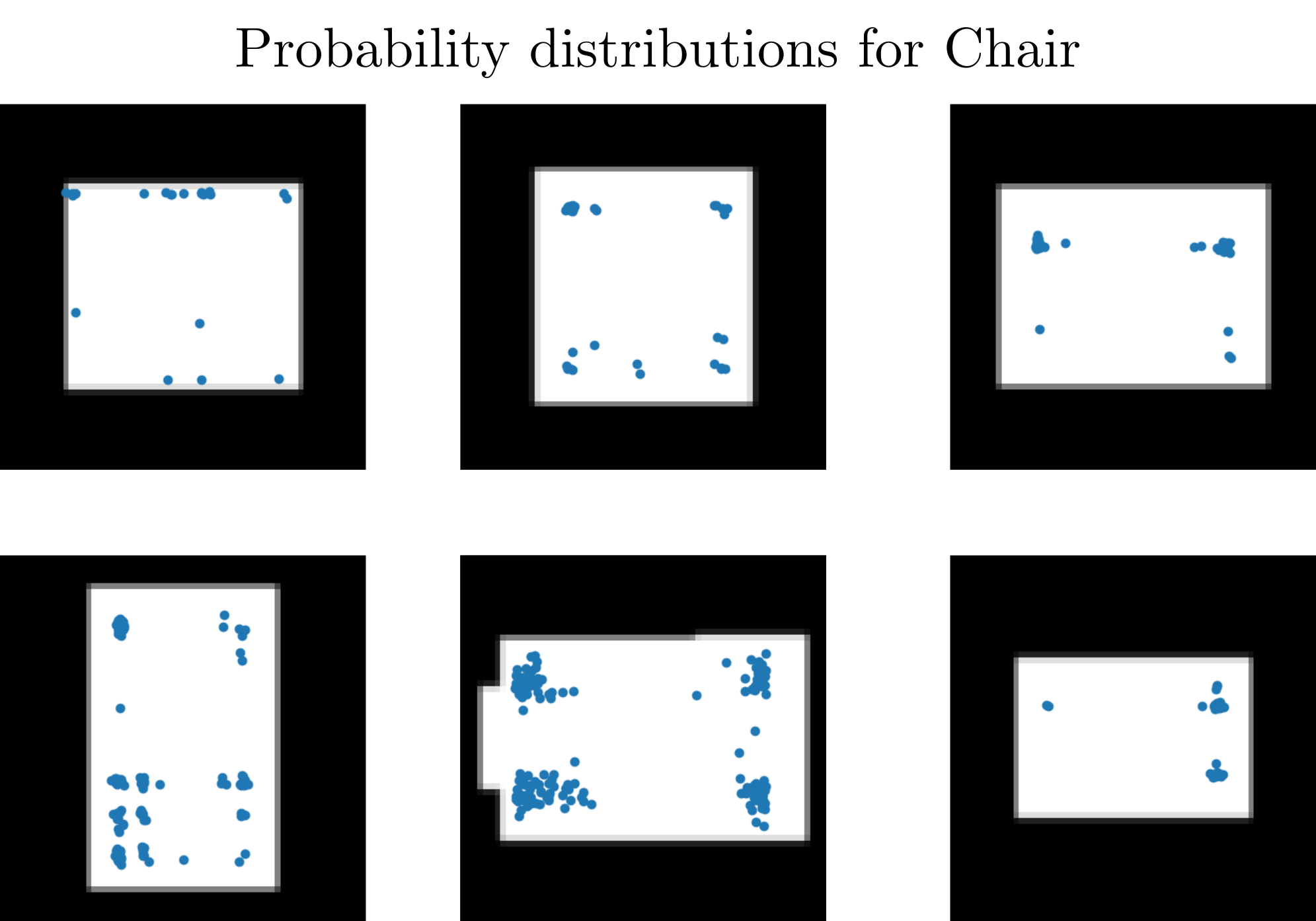}
     \caption{\textbf{Location Distributions for Chair.}}
     \label{fig:chair_location_distributions}
\end{figure}

\begin{figure}
    \centering
     \includegraphics[width=\textwidth]{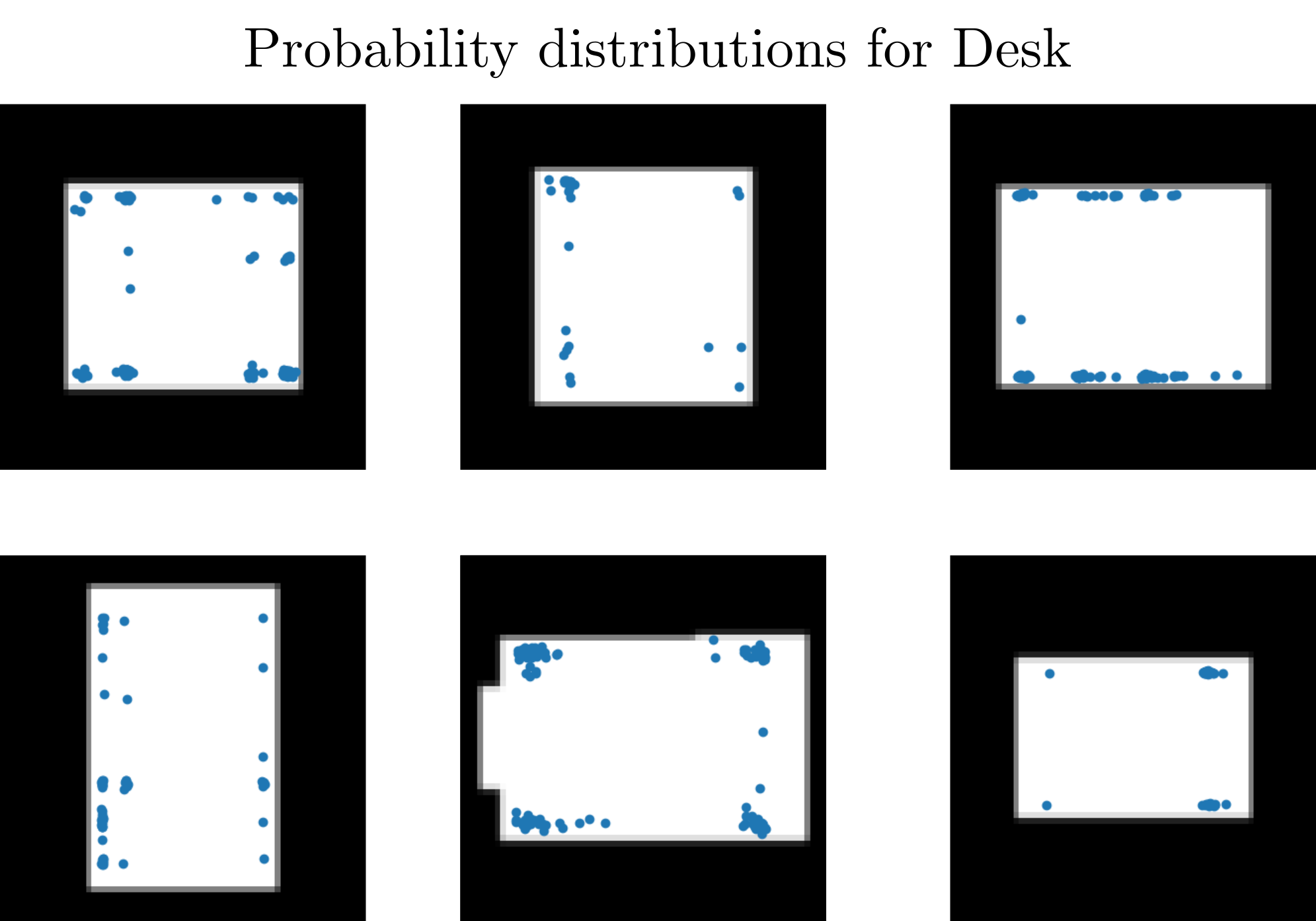}
     \caption{\textbf{Location Distributions for Desk.}}
     \label{fig:desk_location_distributions}
\end{figure}

\begin{figure}
    \centering
     \includegraphics[width=\textwidth]{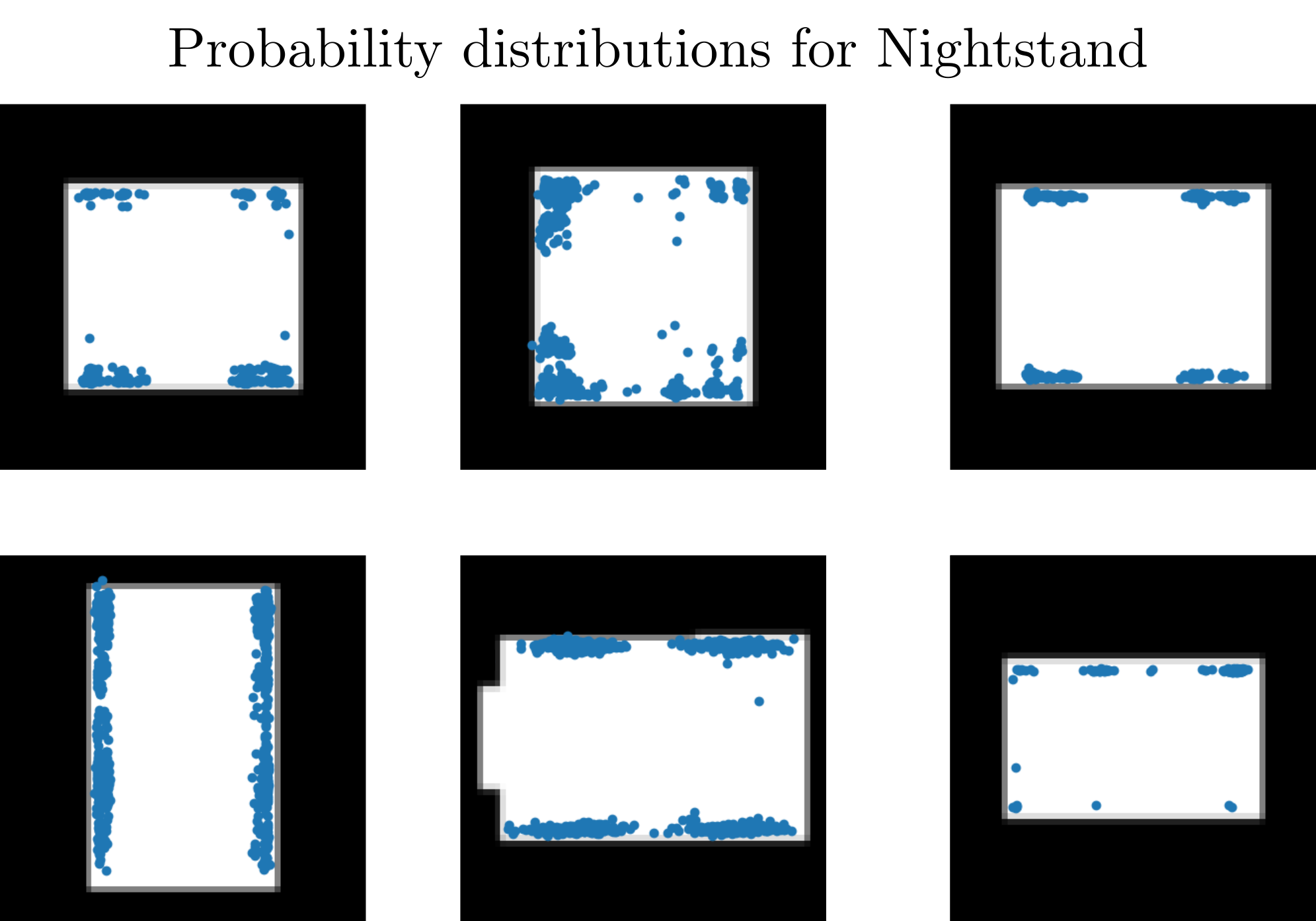}
     \caption{\textbf{Location Distributions for Nightstand.}}
     \label{fig:nightstand_location_distributions}
\end{figure}

\begin{figure}
    \centering
     \includegraphics[width=\textwidth]{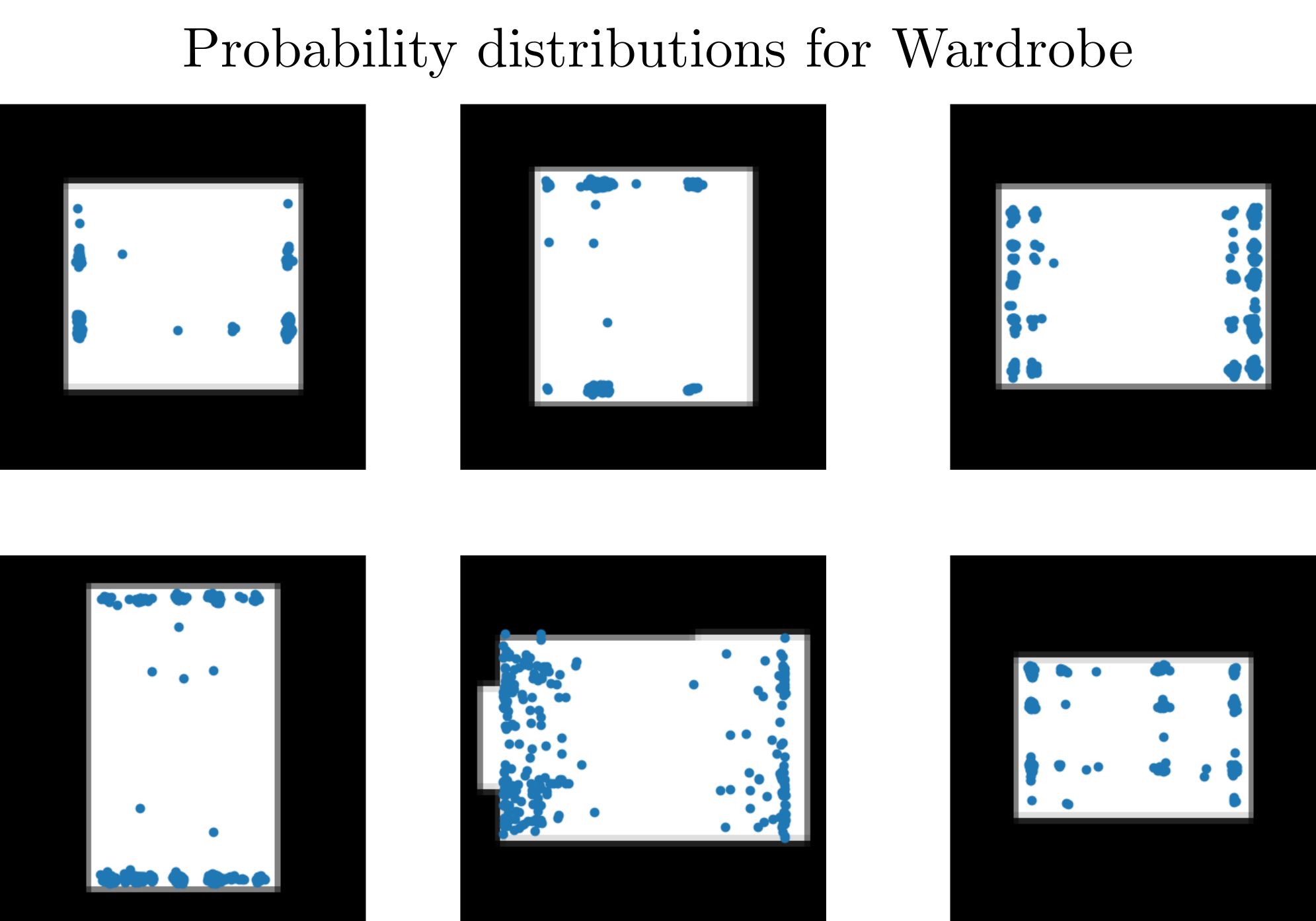}
     \caption{\textbf{Location Distributions for Wardrobe.}}
     \label{fig:wardrobe_location_distributions}
\end{figure}

In this section, we provide examples of the predicted location distributions for
different input scenes. In particular, we randomly select $6$ bedroom floor plans and
conditioned on them we generate $5000$ scenes conditioned on each floor plan. Based on the locations of the generated objects, 
we create scatter plots for the locations of various object categories \ie chair
(\figref{fig:chair_location_distributions}), desk
(\figref{fig:desk_location_distributions}), nightstand
(\figref{fig:nightstand_location_distributions}), wardrobe
(\figref{fig:wardrobe_location_distributions}). We observe that for all object
categories the location distributions of the generated objects are consistently
meaningful.

\subsection{Computational Requirements}

In this section, we provide additional details regarding the computational
requirements of our method, presented in Table 2 and 3 in our main submission.
We observe that ATISS requires significantly less time to generate a scene 
compared to \cite{Wang2020ARXIV, Ritchie2019CVPR}. Note that the computational
cost varies depending on the room type, due to the different average number of
objects for each room type. Living rooms and dining rooms are typically larger
in size, thus more objects need to be generated to cover the empty space. All
reported timings are measured on a machine with an NVIDIA GeForce GTX 1080 Ti
GPU.

Even though the implementations are not directly comparable, since we cannot
guarantee that all have been equally optimized, our findings meet our
expectations. Namely, FastSynth~\cite{Ritchie2019CVPR} requires rendering the
scene each time a new object is added, thus it is expected to be significantly
slower than both SceneFormer and our model. On the other hand,
SceneFormer~\cite{Wang2020ARXIV} utilizes four different transformer models for
generating the attributes of each object, hence it is expected to be at least
four times slower than our model, when generating the same number of objects.

\section{Perceptual Study}

\begin{figure}[h!]
\centering
  \includegraphics[width=1.0\textwidth]{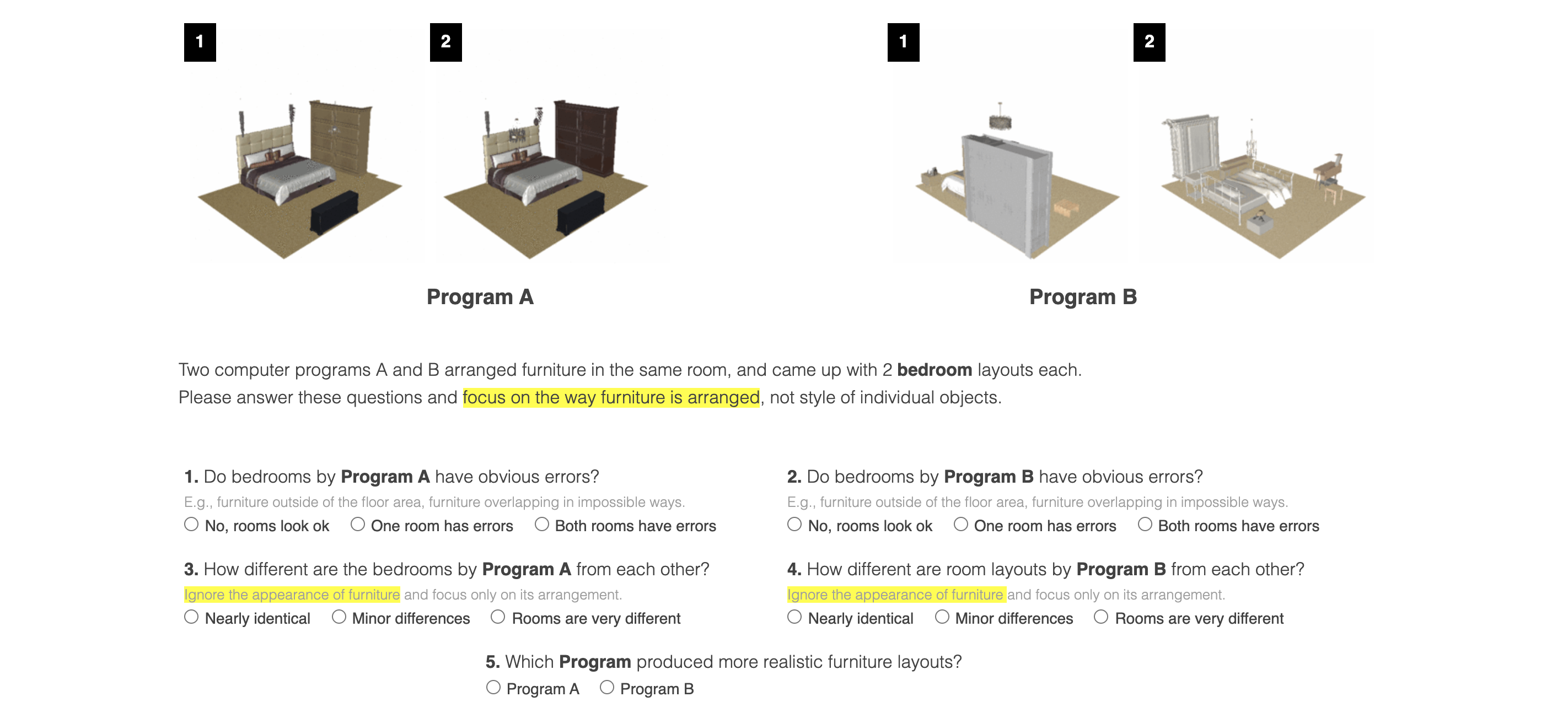}
  \captionof{figure}{\textbf{Perceptual Study UI}. A/B paired questions with rotating
  3D scenes (zoom in).}
  \vspace{-1.2em}
  \label{fig:ustudy_ui}
\end{figure}

We conducted two paired Amazon Mechanical Turk perceptual studies to evaluate
the quality of our generated layouts against FastSynth~\cite{Ritchie2019CVPR} and
SceneFormer~\cite{Wang2020ARXIV}. To this end, we first sampled $211$ floor
plans from the test set and generated $6$ scenes per
floor plan for each method; no filtering or post-processing was used, and samples were randomly
and independently drawn for all methods. Originally, we considered rendering
the rooms with the same furniture objects for each floor plan to allow
participants to only focus on the layout itself, which is the main focus of our
work. However, since the object retrieval is done based on the object
dimensions, rescaling the same furniture piece to fit all predicted dimensions
would result in unrealistically deformed pieces that could skew perceptual
judgements even more heavily.
To avoid having participants focusing on the individual furniture pieces, we
added prominent instructions to focus on the layout and \textbf{not} the
properties of selected objects (see \figref{fig:ustudy_ui}). Each 3D room was
rendered as an animated gif using the same camera rotating around the room. 

In each user study, users were shown paired question sets: two rooms generated
using our method and two generated with the baseline conditioned on the same
floor plan. We randomly selected two out of the 6 pre-rendered scenes for the
given floor plan, and 5 different workers answered the question set about every
floor plan. Namely, the majority of the 6 layouts were shown more than once on
average. A / B order was randomized to avoid bias. The question sets posed the
same two questions about scenes generated with program A and B, in order to let users focus on
the details of the results and to assess errors of the generated layouts.  The
last question forced participants to choose between A or B, based on which
scene looks more realistic.

Specifically,
users were instructed to pay attention to errors like interpenetrating
furniture and furniture outside of the floor area and answer if none, one or
both layouts for each method had errors. We aggregated these statistics to
obtain average error rate per layout, with our method performing nearly twice
better than the best baseline \cite{Ritchie2019CVPR}. The results on realism in Table 4.
in our main submission (first and second row) specify the fraction of the times users chose the
baseline over ours. For example, \cite{Ritchie2019CVPR} was judged more
realistic than ours only 26.9\% of the time. Because there was no intermediate
option, this means that 73.1\% of the time our method was preferred. The last
line in Table 4, in our main submission, aggregated preference for our method
across both studies.

Workers were compensated \$0.05 per question set for a total of USD \$106. The
participation risks involved only the regular risks associated with the use of
a computer.

\section{Additional Related Work on Indoor Synthesis}

In this section, we discuss alternative lines of work on indoor scene
synthesis. Fisher \etal~\cite{Fisher2011SIGGRAPH} propose to represent scenes
using relationship graphs that encode spatial and semantic relationships between
objects in a scene as well as the identity and semantic classification of each
object. Then, they introduce a graph kernel-based scene comparison operator
that allows for retrieving similar scenes, performing context-based model
search \etc. Such representations have been subsequently adopted in models that generate
scenes conditioned on user provided constraints and interior design
guidelines~\cite{Merrell2011SIGGRAPH} or rely on a set of example images 
for generating plausible room layouts \cite{Fisher2012SIGGRAPHASIA}. Another line of research
\cite{Fisher2015SIGGRPAPH, Fu2017SIGGRAPH} leverage activity-associated object
relation graphs for generating semantically meaningful object arrangements.
Finally, another line of research~\cite{Chang2014EMNLP, Ma2018SIGGRAPH} parses
text descriptions into a scene relationship graph that is subsequently used for
arranging objects in a 3D scene.

\section{Discussion and Limitations}

\begin{figure}[!h]
    \centering
    \hfill%
    \begin{subfigure}[b]{0.20\linewidth}
		\centering
		\includegraphics[width=\linewidth]{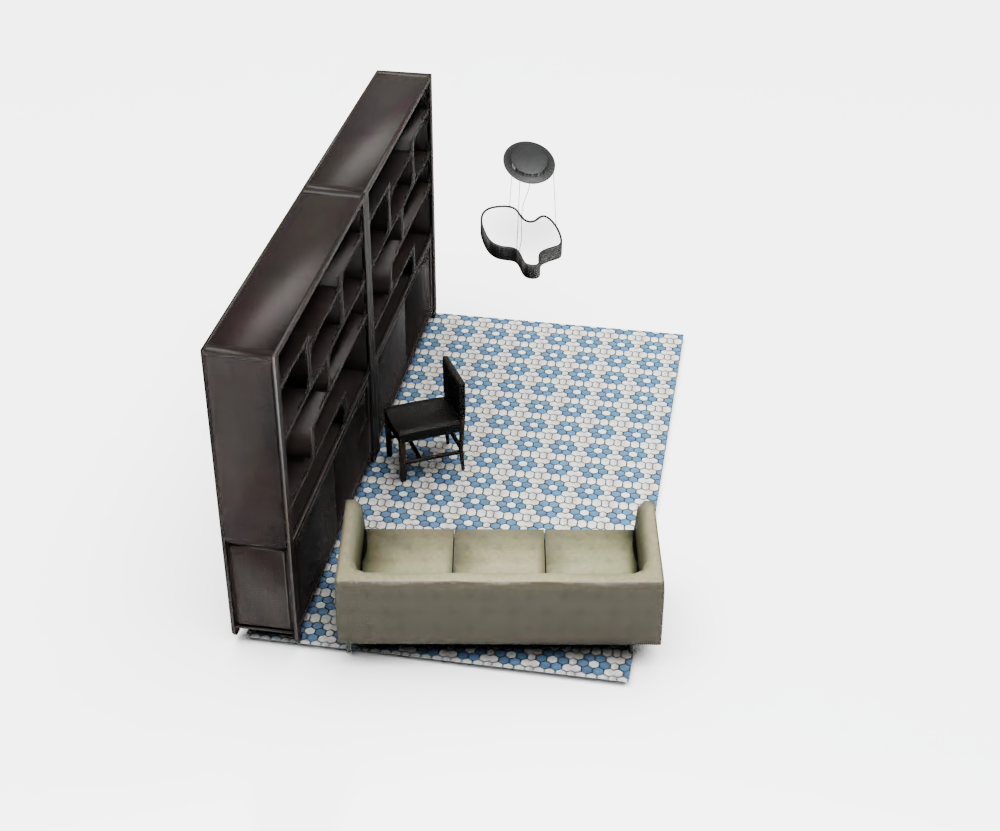}
    \end{subfigure}%
        \begin{subfigure}[b]{0.20\linewidth}
		\centering
		\includegraphics[width=\linewidth]{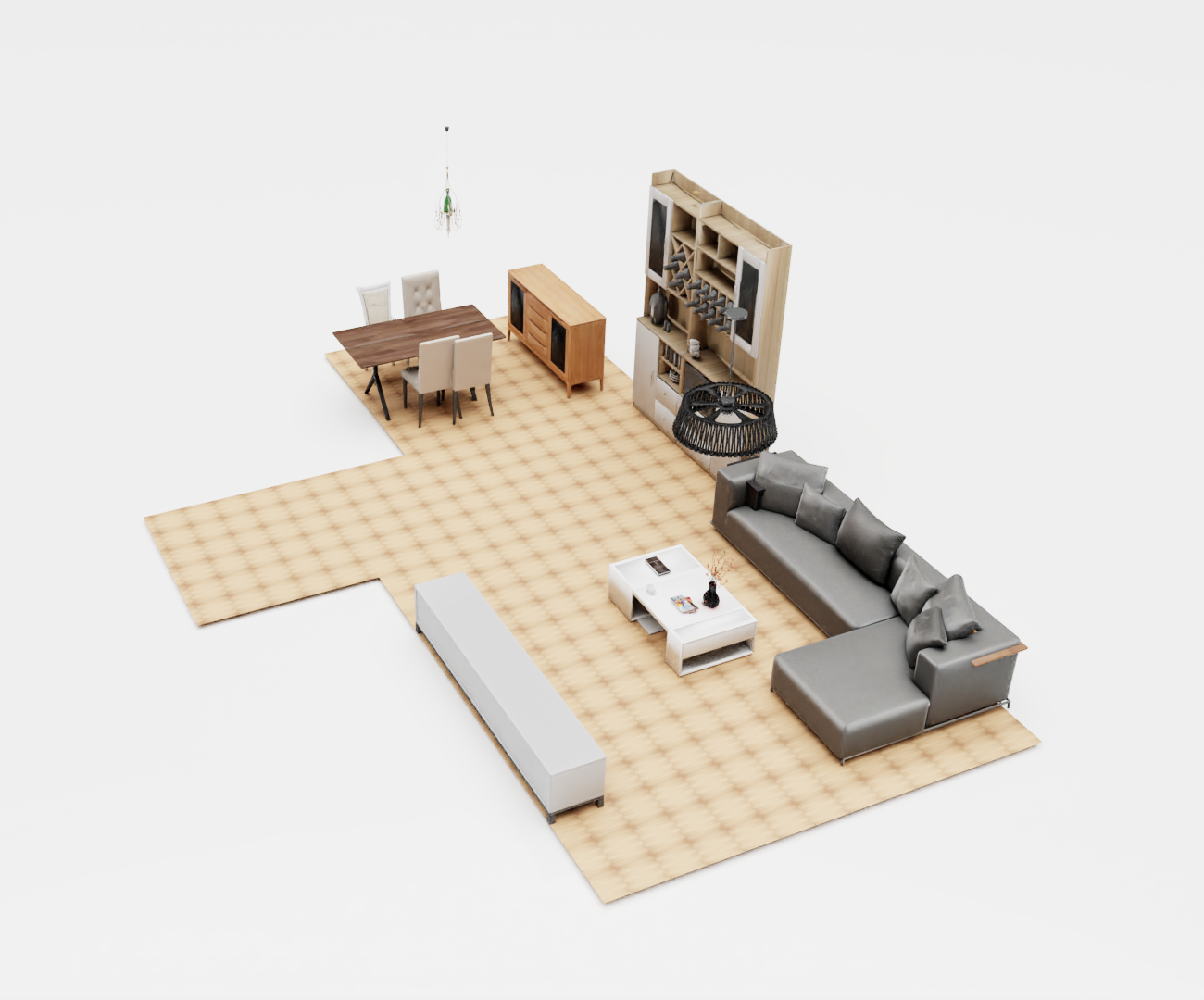}
    \end{subfigure}%
        \begin{subfigure}[b]{0.20\linewidth}
		\centering
		\includegraphics[width=\linewidth]{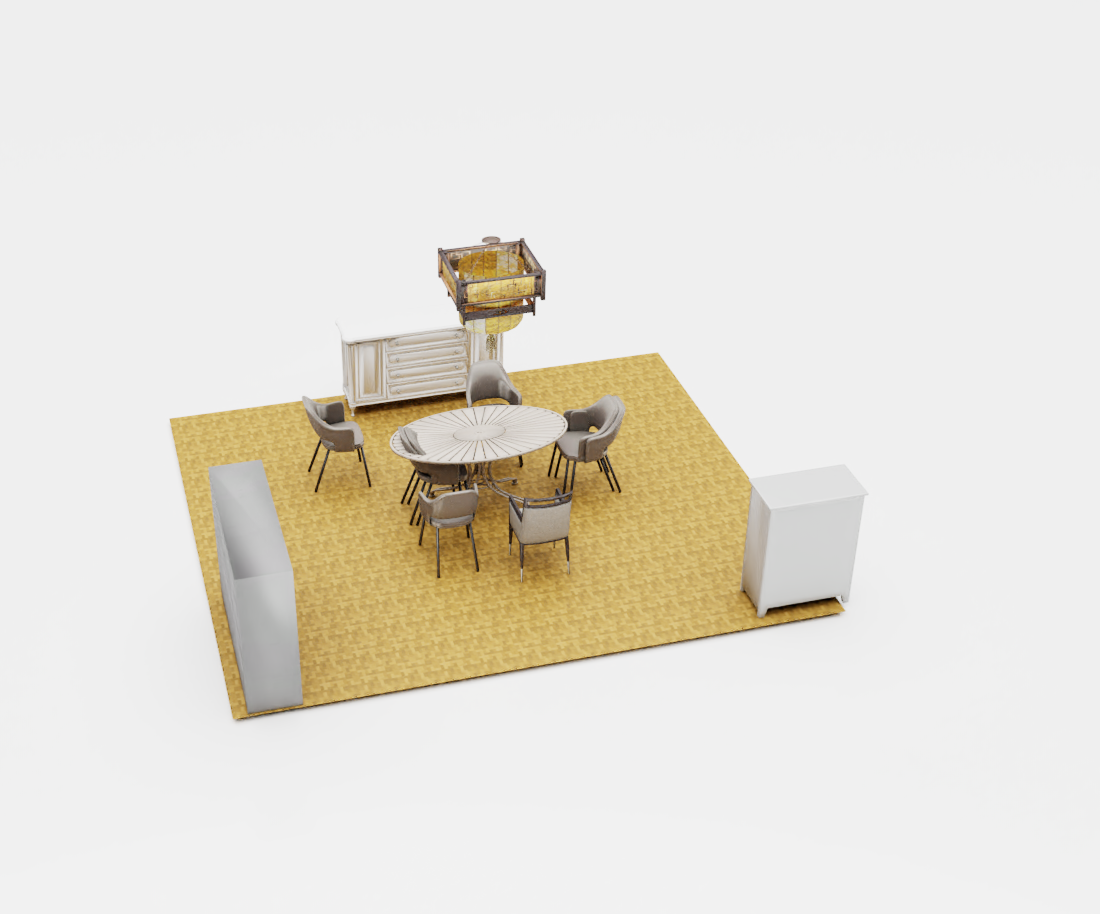}
    \end{subfigure}%
        \begin{subfigure}[b]{0.20\linewidth}
		\centering
		\includegraphics[width=\linewidth]{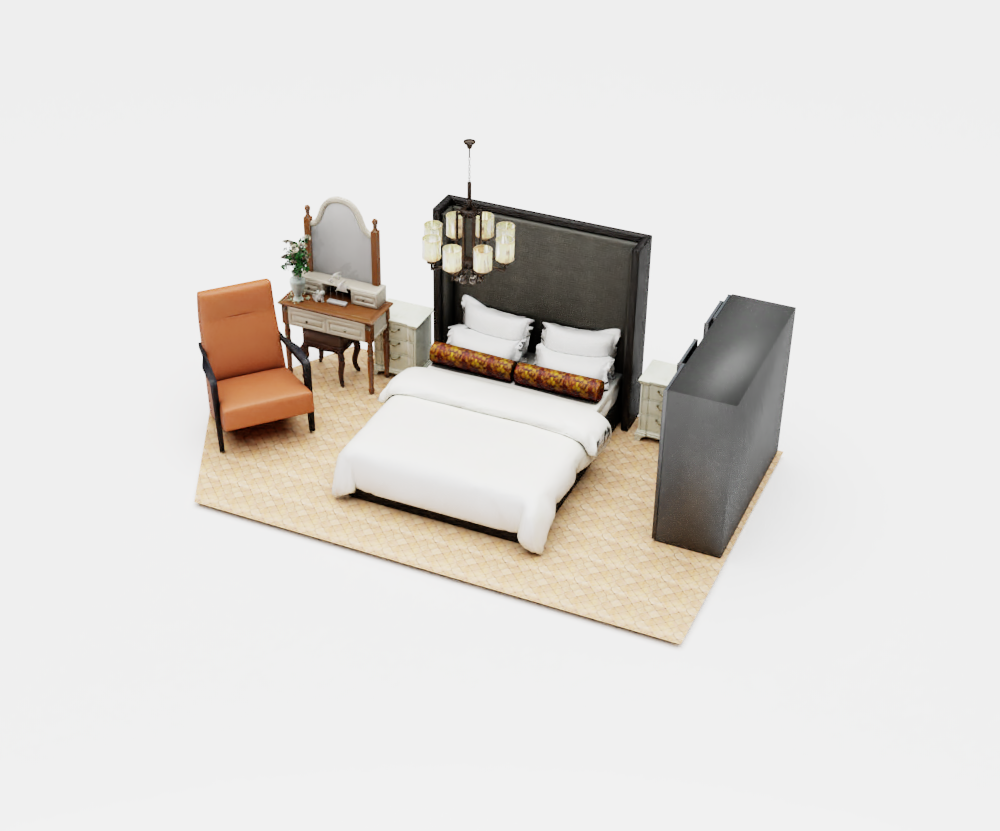}
    \end{subfigure}%
    \begin{subfigure}[b]{0.20\linewidth}
		\centering
		\includegraphics[width=\linewidth]{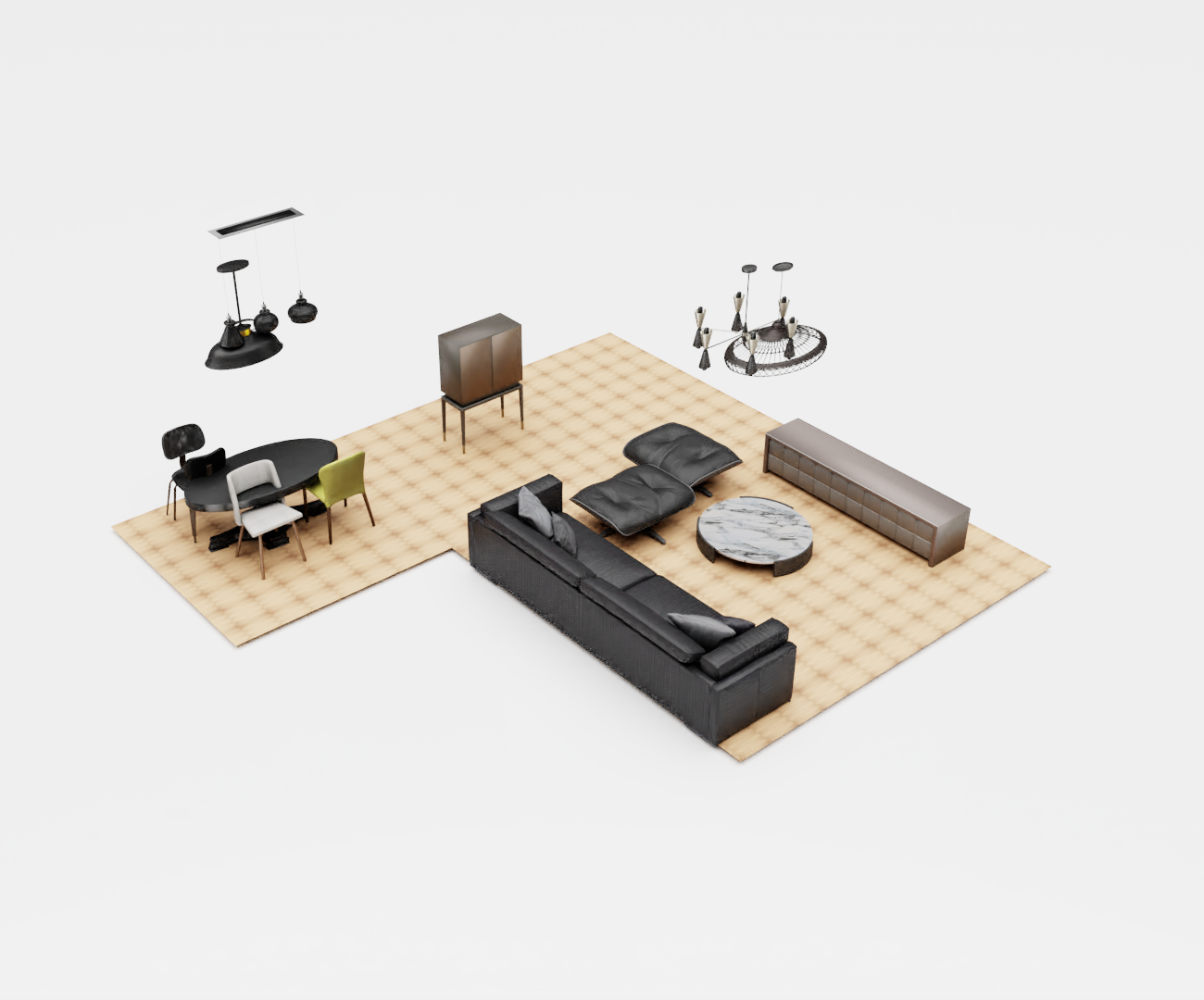}
    \end{subfigure}%
    \vspace{-1.2em}
    \hfill%
    \caption{{\bf Failure Cases}. We visualize various failure cases of
    our model for different toom types.}
    \label{fig:failure_cases}
\end{figure}

Lastly, we discuss the limitations of our model and show some examples of
failure cases in \figref{fig:failure_cases}. One type of failure case that is
illustrated in \figref{fig:failure_cases} is overlapping objects, in
particular chairs for the case of living rooms and dining rooms (see second and
third column in \figref{fig:failure_cases}). As we already discussed in
\secref{sec:3d_front_filtering}, to be able to use the 3D-FRONT dataset, we
performed intense filtering to remove objects that intersect with each other.
However, we found out that not all problematic arrangements were removed from
the dataset, which we hypothesize is the reason for such failure cases. Another type of failure case
that we observed, which is also related to the existence of problematic rooms
in our training data, is the unnatural orientation of objects (\eg chair facing
the bookshelf in first column of \figref{fig:failure_cases} or chair facing
opposite of the table in last column of \figref{fig:failure_cases}.) Note that
these failure cases are quite rare, as also indicated by our quantitative
analysis in Sec. 4.1 in the main submission as well as the perceptual study in
Sec. 4.3, but our method does not guarantee error-free layouts and there is room for improvement.

Our approach is currently limited to generating object properties using a
specific ordering (category first, followed by location, then orientation and
lastly size). To further expand the interactive possibilities of our model, we
believe that also the object attributes should be generated in an order
invariant fashion, similar to the objects in the scene. Furthermore, in our
current formulation, the object retrieval is disconnected from the attribute
generation. As a result we cannot guarantee that the retrieved objects would
match with existing objects in the scene. To address this, in the future, we
plan to also incorporate style as an additional object attribute to allow for
improved object retrieval. Incorporating style information, would also allows
us to generate rooms conditioned on a specific style. Another exciting research
direction that we would like to explore is combining ATISS with existing
compositional representations of objects~\cite{Tulsiani2017CVPRa,
Paschalidou2019CVPR, Mo2019SIGGRAPH, Paschalidou2020CVPR, Paschalidou2021CVPR,
Mo2021ICCV}. This will allow us to generate 3D scenes with control over the
object arrangement, object parts and part relationships. Due to the unique
characteristics of compositional representations representations, our generated
scenes will be fully controllable i.e. it will be possible to manipulate
objects and object parts, edit specific parts of the scene \etc.

\clearpage

\end{document}